\documentclass[12pt]{report} 
\usepackage[]{graphicx} 
\usepackage{bmpsize}
\usepackage{setspace}
\usepackage{bm}
\usepackage{listings}
\usepackage{makeidx}
\usepackage{titlesec}

\makeindex
\usepackage[square,numbers]{natbib}
\usepackage[utf8]{inputenc}
\usepackage{fullpage}
\usepackage{url}
\usepackage[symbols,toc,automake,acronym]{glossaries-extra}

\usepackage{amsmath}
\usepackage{subfig}
\usepackage{xparse}
\usepackage{pgfplotstable}
\usetikzlibrary{matrix,chains,positioning,decorations.pathreplacing,arrows}
\usepackage{listings}
\usepackage{bm,amssymb}
\usepackage{algorithm, float}
\usepackage{algpseudocode}
\floatname{algorithm}{Algorithm}
\usepackage[pdf]{pstricks}
\usepackage{pst-plot}
\usepackage{pst-node}
\usepackage{pst-tree}
\usepackage{pstricks-add}
\usepackage{csvsimple}
\usepackage{xinttools} 
\usepackage{tikz}
\usetikzlibrary{calc}
\usetikzlibrary{shapes.misc}
\usepackage{booktabs}
\usepackage{hyperref}
\usetikzlibrary{decorations.text}

\newcommand\Square[1]{+(-#1,-#1) rectangle +(#1,#1)}
\tikzset{cross/.style={cross out, draw=black, minimum size=2*(#1-\pgflinewidth), inner sep=0pt, outer sep=0pt},
cross/.default={1pt}}

\usetikzlibrary{trees}
\usetikzlibrary{arrows,automata,snakes}
\usetikzlibrary{calc,trees,positioning,arrows,chains,shapes.geometric,%
decorations.pathreplacing,decorations.pathmorphing,shapes,%
matrix,shapes.symbols,automata}

\lstdefinelanguage{Python}{
numbers=left,
numberstyle=\footnotesize,
numbersep=1em,
xleftmargin=1em,
framextopmargin=2em,
framexbottommargin=2em,
showspaces=false,
showtabs=false,
showstringspaces=false,
frame=l,
tabsize=4,
basicstyle=\ttfamily\small\setstretch{1},
backgroundcolor=\color{Background},
commentstyle=\color{Comments}\slshape,
stringstyle=\color{Strings},
morecomment=[s][\color{Strings}]{"""}{"""},
morecomment=[s][\color{Strings}]{'''}{'''},
morekeywords={import,from,class,def,for,while,if,is,in,elif,else,not,and,or,print,break,continue,return,True,False,None,access,as,,del,except,exec,finally,global,import,lambda,pass,print,raise,try,assert},
keywordstyle={\color{Keywords}\bfseries},
morekeywords={[2]@invariant,pylab,numpy,np,scipy},
keywordstyle={[2]\color{Decorators}\slshape},
emph={self},
emphstyle={\color{self}\slshape},
}

\newcommand\pythonstyle{\lstset{
language=Python,
basicstyle=\ttm,
otherkeywords={self},             
keywordstyle=\ttb\color{deepblue},
emph={MyClass,__init__},          
emphstyle=\ttb\color{deepred},    
stringstyle=\color{deepgreen},
frame=tb,                         
showstringspaces=false            %
}}

\newcommand{\boundellipse}[3]
{(#1) ellipse (#2 and #3)
}
\def\biglen{20cm} 
\tikzset{
  half plane/.style={ to path={
       ($(\tikztostart)!.5!(\tikztotarget)!#1!(\tikztotarget)!\biglen!90:(\tikztotarget)$)
    -- ($(\tikztostart)!.5!(\tikztotarget)!#1!(\tikztotarget)!\biglen!-90:(\tikztotarget)$)
    -- ([turn]0,2*\biglen) -- ([turn]0,2*\biglen) -- cycle}},
  half plane/.default={1pt}
}

\def\maxxy{4} 

\lstnewenvironment{python}[1][]
{
\pythonstyle
\lstset{#1}
}
{}

\newcommand\pythoninline[1]{{\pythonstyle\lstinline!#1!}}

\tikzstyle{mybox} = [draw=red,very thick,
    rectangle, rounded corners, inner sep=10pt, inner ysep=20pt]
\tikzstyle{fancytitle} =[fill=red, text=black]

\tikzset{
  treenode/.style = {shape=rectangle, rounded corners,
                     draw, align=center,
                     top color=white, bottom color=blue!20},
  root/.style     = {treenode, font=\Large, bottom color=red!30},
  env/.style      = {treenode, font=\ttfamily\normalsize},
  dummy/.style    = {circle,draw}
}

\tikzset{
	>=stealth',
	punkt/.style={
		rectangle,
		rounded corners,
		draw=black, very thick,
		text width=3cm,
		minimum height=2em,
		text centered},
	pil/.style={
		->,
		thick,
		shorten <=2pt,
		shorten >=2pt,}
}

\usetikzlibrary{arrows,automata,snakes}
\usetikzlibrary{calc,trees,positioning,arrows,chains,shapes.geometric,%
	decorations.pathreplacing,decorations.pathmorphing,shapes,%
	matrix,shapes.symbols,automata}

\tikzset{
	treenode/.style = {shape=rectangle, rounded corners,
		draw, align=center,
		top color=white, bottom color=blue!20},
	root/.style     = {treenode, font=\Large, bottom color=red!30},
	env/.style      = {treenode, font=\ttfamily\normalsize},
	dummy/.style    = {circle,draw}
}

\newcounter{exercise}[chapter]
\newenvironment{exercise}[1][]{\refstepcounter{exercise}\par\medskip
	\textbf{Exercise~\theexercise. #1} \rmfamily}{\medskip}
\renewcommand{\theexercise}{\arabic{chapter}.\arabic{exercise}}

\newcommand\defeq{:=}

\newcommand{\vx}[0]{{\bf x}}
\newcommand{\vv}[0]{{\bf v}}
\newcommand{\vu}[0]{{\bf u}}

\newcommand{\vm}[0]{{\bf m}}
\newcommand{\vq}[0]{{\bf q}}
\newcommand{\mX}[0]{{\bf X}}
\newcommand{\mC}[0]{{\bf C}}
\newcommand{\mA}[0]{{\bf A}}
\newcommand{\mL}[0]{{\bf L}}
\newcommand{\fscore}[0]{F_{1}}
\newcommand{\sparsity}{s}
\newcommand{\mW}[0]{{\bf W}}
\newcommand{\mD}[0]{{\bf D}}
\newcommand{\mZ}[0]{{\bf Z}}
\newcommand{\vw}[0]{{\bf w}}

\newcommand{\mQ}{\mathbf{Q}}

\newcommand{\vy}[0]{{\bf y}}
\newcommand{\va}[0]{{\bf a}}

\newcommand{\vb}[0]{{\bf b}}
\newcommand{\vr}[0]{{\bf r}}
\newcommand{\vz}[0]{{\bf z}}

\newcommand{\vc}[0]{{\bf c}}

\newcommand{\errprob}{p_{\rm err}} 
\newcommand{\prob}[1]{p({#1})} 
 
\def \expect {\mathbb{E} }

\newcommand{\biasterm}{B}
\newcommand{\varianceterm}{V}
\newcommand{\neighbourhood}[1]{\mathcal{N}^{(#1)}}
\newcommand{\nrfolds}{k}
\newcommand{\mseesterr}{E_{\rm est}}
\newcommand{\bootstrapidx}{b}
\newcommand{\modelidx}{l}
\newcommand{\nrbootstraps}{B}
\newcommand{\sampleweight}[1]{q^{(#1)}}
\newcommand{\nrcategories}{K} 
\newcommand{\splitratio}[0]{{\rho}}

\newcommand{\sqeuclnorm}[1]{\big\Vert  {#1} \big\Vert^{2}_{2}}
\newcommand{\bmx}[0]{\begin{bmatrix}}
\newcommand{\emx}[0]{\end{bmatrix}}

\newcommand\eigvecCov{\mathbf{u}} 
\newcommand\eigvecCoventry{u}
\newcommand{\featuredim}{n}
\newcommand{\featurelenraw}{\featuredim'}
\newcommand{\featurelen}{\featuredim}

\newcommand{\samplesize}{m}
\newcommand{\sampleidx}{i} 
\newcommand{\nractions}{A} 
\newcommand{\datapoint}{\vz} 
\newcommand{\actionidx}{a} 
\newcommand{\clusteridx}{c} 
\newcommand{\sizehypospace}{D}
\newcommand{\nrcluster}{k} 
 
\newcommand{\featureidx}{j} 
\newcommand{\clustermean}{{\bm \mu}} 
\newcommand{\clustercov}{{\bm \Sigma}} 

\newcommand{\error}{E}
\newcommand{\augidx}{b}
\newcommand{\task}{\mathcal{T}}
\newcommand{\nrtasks}{T}
\newcommand{\taskidx}{t}
\newcommand\truelabel{y}
\newcommand{\polydegree}{r}
\newcommand\labelvec{\vy}
\newcommand\featurevec{\vx}
\newcommand\feature{x}
\newcommand\predictedlabel{\hat{y}}
\newcommand\dataset{\mathcal{D}}
\newcommand\trainset{\dataset^{(\rm train)}}
\newcommand\valset{\dataset^{(\rm val)}}

\newcommand\effdim[1]{d_{\rm eff} \left( #1 \right)}

\newcommand{\hypospace}{\mathcal{H}}
\newcommand{\emperror}{\widehat{L}}
\newcommand\risk[1]{\expect \big \{ \loss{(\featurevec,\truelabel)}{#1} \big\}}
\newcommand{\featurespace}{\mathcal{X}}
\newcommand{\labelspace}{\mathcal{Y}}
\newcommand{\rawfeaturevec}{\mathbf{z}}
\newcommand{\rawfeature}{z}
\newcommand{\condent}{H}
\newcommand{\explanation}{e}
\newcommand{\explainset}{\mathcal{E}}
\newcommand{\user}{u}
\newcommand{\actfun}{\sigma}
\newcommand{\noisygrad}{g}
\newcommand{\reconstrmap}{r}
\newcommand{\predictor}{h}
\newcommand{\eigval}[1]{\lambda_{#1}}
\newcommand{\regparam}{\lambda}
\newcommand{\lrate}{\alpha}
\newcommand{\edges}{\mathcal{E}}

\DeclareMathOperator{\supp}{supp}
\newcommand{\loss}[2]{L\big({#1},{#2}\big)}

\DeclareMathOperator*{\argmax}{argmax}
\DeclareMathOperator*{\argmin}{argmin}
\newcommand{\itercntr}{r}

\newcommand{\timeidx}{t}

\newcommand{\reward}{r}

\newcommand{\rumba}{Rumba}

\newcommand{\basisfunc}{\phi}
\newcommand{\augparam}{B}
\newcommand{\valerror}{E_{v}}
\newcommand{\trainerror}{E_{t}}
\newcommand{\foldidx}{b}
\newcommand{\testset}{\dataset^{(\rm test)} }
\newcommand{\testerror}{E^{(\rm test)}}
\newcommand{\nrmodels}{M}
\newcommand{\benchmarkerror}{E^{(\rm ref)}}
\newcommand{\lossfun}{L}
 
\newcommand{\cluster}{\mathcal{C}}
\newcommand{\bayeshypothesis}{h^{*}}
\newcommand{\featuremtx}{\mX}
\newcommand{\weight}{w}
\newcommand{\weights}{\vw}
\newcommand{\regularizer}{\mathcal{R}}
\newcommand{\decreg}[1]{\mathcal{R}_{#1}}
\newcommand{\naturalnumbers}{\mathbb{N}}
\newcommand{\featuremapvec}{{\bf \Phi}}
\newcommand{\featuremap}{\phi}
\newcommand{\batchsize}{B}
\newcommand{\batch}{\mathcal{B}}
\newcommand{\foldsize}{B}

\newcommand{\graph}{\mathcal{G}}
\newcommand{\nodes}{\mathcal{V}}

\newcommand{\nodedegree}[1]{d^{(#1)}}
\newcommand{\nodeidx}{i}

\usepackage{amsmath}

\newcommand{\vh}{\mathbf{h}}

\font\myfont=cmr12 at 30pt

\font\myfontb=cmr12 at 17pt
\font\myfontc=cmr12 at 15pt

\usepackage[many]{tcolorbox}
\usepackage{xcolor}

\definecolor{codegreen}{rgb}{0,0.6,0}
\definecolor{codegray}{rgb}{0.5,0.5,0.5}
\definecolor{codepurple}{rgb}{0.58,0,0.82}
\definecolor{Background}{rgb}{0.95,0.95,0.92}
\definecolor{backcolour}{rgb}{0.95,0.95,0.92}
\DeclareTColorBox{mybox}{O{orange}O{0cm}}{
	breakable,
	outer arc=0pt,
	arc=0pt,
	colback=white,
	rightrule=0pt,
	toprule=0pt,
	top=0pt,
	right=0pt,
	bottom=0pt,
	bottomrule=0pt,
	colframe=#1,
	enlarge left by=#2,
	width=\linewidth-#2,
}

\usepackage{amsmath}
\makeglossaries

\newglossaryentry{latex}
{
    name=latex,
    description={LaTeX (short for Lamport TeX) is a document preparation system. The user has to think 
    	about only the content to put in the document and the software will take care of the formatting. }
}

\newglossaryentry{glsy}
{
    name=glossary,
    description={Acronyms and terms which are generally unknown or new to common readers.}
}

\newglossaryentry{minimum}
{
	name=minimum,
	description={Given a set of real numbers, the minimum is the smallest of those numbers.}
}

\newglossaryentry{maximum}
{
	name=maximum,
	description={Given a set of real numbers, the maximum is the largest of those numbers.}
}

\newglossaryentry{mean}
{
	name=mean,
	description={The expectation of a real-valued random variable.}
}

\newglossaryentry{variance}
{
	name={variance},
	description={The variance of a real-valued \gls{rv} $\feature$ is defined as the expectation 
		$\expect\big\{ \big( x - \expect\{x \} \big)^{2} \big\}$ of the squared difference $\feature$ 
		and its expectation $\expect\{x \}$. We extend this definition to vector-valued \gls{rv}s $\featurevec$ 
		as $\expect\big\{ \big\| \featurevec - \expect\{\featurevec \} \big\|_{2}^{2} \big\}$.} ,first={variance},text={variance} 
}

\newglossaryentry{nn}
{
	name={nearest neighbour},
	description={Nearest neighbour methods learn a hypothesis $h: \featurespace \rightarrow \labelspace$ whose 
		function value $h(\featurevec)$ is solely detemined by the nearest neighbours in the 
		feature space $\featurespace$ },
	first={nearest neighbour (NN)},text={NN} 
}

\newglossaryentry{bias}
{
	name={bias},
	description={Consider some unknown quantity $\bar{\weight}$, e.g., the true weight in a linear model $\truelabel = \bar{\weight} \feature + e$ 
		relating feature and label of a \gls{datapoint}. We might use an ML method (e.g., based on \gls{erm}) to 
		compute an estimate $\hat{\weight}$ for the $\bar{\weight}$ based on a set of \gls{datapoint}s that are 
		realizations of \gls{rv}s. The (squared) bias incurred by the estimate $\hat{\weight}$ is typically defined as 
		$\biasterm^{2} \defeq \big( \expect \{ \hat{\weight}  \}- \bar{\weight}\big)^{2}$. We extend this definition to 
		vector-valued quantities using the squared Euclidean norm $\biasterm^{2} \defeq \big\| \expect \{ \widehat{\weights}  \}- \overline{\weights}\big\|_{2}^{2}$.},
first={bias},text={bias} 
}

\newglossaryentry{classification}
{
	name={classification},
	description={Classification is the task of determining a discrete-valued label $\truelabel$ of a \gls{datapoint} 
		based solely on its features $\featurevec$. The label $\truelabel$ belongs to a finite set, such as $\truelabel \in \{ -1,1\}$, 
		or $\truelabel \in \{1,\ldots,19\}$ and represents a category to which the corresponding \gls{datapoint} belongs to.},first={classification},text={classification} 
}

\newglossaryentry{condnr}
{
	name={condition number},
	description={The condition number $\kappa(\mathbf{Q})$ of a \gls{psd} matrix $\mathbf{Q}$ is 
		the ratio of the largest to the smallest eigenvalue of $\mathbf{Q}$.},first={condition number},text={condition number} 
}

\newglossaryentry{classifier}
{
	name={classifier},
	description={A classifier is a hypothesis $h(\featurevec)$ that is used to predict a discrete-valued label. 
		Strictly speaking, a classifier is a hypothesis $h(\featurevec)$ that can take only a finite number of 
		different values. However, we are sometimes sloppy and use the term classifier also for a hypothesis 
		that delivers a real number which is thresholded to obtain the predicted label value. 
		For example, in a binary classification problem with label values $\truelabel \in \{ -1,1\}$, we refer to a 
		linear hypothesis $h(\featurevec) =\weights^{T}\featurevec$ as classifier if it is used to predict the label 
		value according to $\hat{\truelabel} = 1$ when $h(\featurevec) \geq 0$  and $\hat{\truelabel}=-1$ otherwise.},first={classifier},text={classifier} 
}

\newglossaryentry{emprisk}
{name={empirical risk},
  description={The empirical risk of a given hypothesis on a given set of datapoints is the average loss of the hypothesis computed over all datapoints in that set.},first={empirical risk},text={empirical risk} 
}

\newglossaryentry{risk}
{name={risk},
	description={Consider a hypothesis $h$ that is used to predict the label $\truelabel$ of a \gls{datapoint} 
		based on its features $\featurevec$. We measure the quality of a particular prediction using a \gls{lossfunc} $\loss{(\featurevec,\truelabel)}{h}$. If we interpret \gls{datapoint}s as realizations of \gls{iid} \gls{rv}s, 
		also the $\loss{(\featurevec,\truelabel)}{h}$ becomes the realization of a \gls{rv}. Using such an \gls{iidasspt} allows 
		to define the risk of a hypothesis as the expected \gls{loss} $\expect \big\{\loss{(\featurevec,\truelabel)}{h} \big\}$. 
		Note that the risk of $h$ depends on both, the specific choice for the \gls{lossfunc} and the common \gls{probdist} 
		of the \gls{datapoint}s.},
	first={risk},text={risk} 
}

\newglossaryentry{actfun}
{name={activation function},
	description={Each artificial neuron within an \gls{ann} consists of an activation function that maps 
		the inputs of the neuron to a single output value. In general, an activation function is a non-linear 
		map of the weighted sum of neuron inputs (this weighted sum is the activation of the neuron).},
first={activation function},text={activation function} 
}

\newglossaryentry{deepnet}
{name={deep net},
	description={We refer to an \gls{ann} with a (relatively) large number of hidden layers as a deep \gls{ann} or ``deep net''. Deep nets 
	are used to represent the \gls{hypospace}s of deep learning methods \cite{Goodfellow-et-al-2016}.},
	first={deep \gls{ann} (deep net)},text={deep net} 
}

\newglossaryentry{baseline}
{name={baseline},
	description={A reference value or benchmark for the average loss incurred by a hypothesis when applied to the \gls{datapoint}s 
		generated in a specific ML application. Such a reference value might be obtained from human performance (e.g., error rate 
		of dermatologists diagnosing cancer from visual inspection of skin areas) or other ML methods (``competitors'')}, 
	first={baseline},text={baseline} 
}

\newglossaryentry{spectogram}
{name={spectogram},
	description={The spectogram of a time signal, e.g., an audio recording, characterizes the 
	time-frequency distribution of the signal. Loosely speaking, the spectogram quantities the 
	signal strength at a specitic time and frequence.}, 
	first={spectogram},text={spectrogram} 
}

\newglossaryentry{esterr}
{name={estimation error},
	description={Consider \gls{datapoint}s with feature vectors $\featurevec$ and label 
		$\truelabel$. In some applications we can model the relation between features and label of a \gls{datapoint} 
		as $\truelabel = \bar{h}(\featurevec) + \varepsilon$. Here we used some true hypothesis $\bar{h}$ and a noise 
		term $\varepsilon$ which might represent modelling or labelling errors. The estimation error incurred by a ML 
		method that learns a hypothesis $\hat{h}$, e.g., using \gls{erm}, is defined as $\hat{h} - \bar{h}$. 
		For a parametrized hypothesis space, consisting of hypothesis maps that are determined by a parameter vector $\weights$, 
		we define the estimation error in terms of parameter vectors as $\Delta \weights = \widehat{\weights} - \overline{\weights}$.}
	first={estimation error},text={estimation error} 
}

\newglossaryentry{dob}
{name={degree of belonging},
	description={A number that indicats the extend by which a \gls{datapoint} belongs to a \gls{cluster}. 
	The degree of belonging can be interpreted as a soft \gls{cluster} assignment. Soft clustering methods 
   typically represent the degree of belonging by a real number in the interval $[0,1]$. The boundary values 
   $0$ and $1$ correspond to hard cluster assignments.}, first={degree of belonging},text={degree of belonging} 
}

\newglossaryentry{msee}
{name={mean squared estimation error},
	description={Consider a ML method that uses a parametrized hypothesis space. For a given \gls{trainset}, whose \gls{datapoint}s are 
		interpreted as realizations of \gls{rv}s, the ML method learns the parameters incurring the \gls{esterr} $\Delta \weights$.   
		The mean squared estimation error is defined as the expectation $\expect \big\{ \big\| \Delta \weights \big\|^{2} \big\} $ 
		of the squared Euclidean norm of the \gls{esterr}.},
	first={mean squared estimation error (MSEE)},text={MSEE} 
}

\newglossaryentry{expert}
{name={expert},
	description={ML aims at learning a hypothesis $h$ that accurately predicts the label 
		of a \gls{datapoint} based on its features. We measure the prediction error using 
		some \gls{lossfunc}. Ideally we want to find a hypothesis that incurres minimum loss. 
		One approach to make this goal precise is to use the \gls{iidasspt} and use the resulting 
		\gls{bayesrisk} as the benchmark level for the (average) loss of a hypothesis. Alternatively 
		we might know a reference or benchmark hypothesis $h'$ which might be obtained by 
		some existing ML mehtod. We can then compare the loss incurred by $h$ with the loss 
		incurred by $h'$. Such a reference or baseline hypothesis $h'$ is refered to as an \gls{expert}. 
		Note that an expert might deliver very poor predictions. We typically compare against many 
	   different experts and aim at incurring not much more loss than the best among those experts (this is 
	   known as regret minimization) \cite{PredictionLearningGames,HazanOCO}.}
	first={expert},text={expert} 
}

\newglossaryentry{regret}
{name={regret},
	description={The regret of a hypothesis $h$ relative to another hypothesis $h'$, which serves as a reference of baseline, 
		      is the difference between the \gls{loss} incurred by $h$ and the \gls{loss} incurred by $h'$ \cite{PredictionLearningGames}. 
		      The baseline hypothesis $h'$ is also refered to as an \gls{expert}.}
	first={regret},text={regret} 
}

\newglossaryentry{differentiable}
{name={differentiable},
	description={A function $f: \mathbb{R}^{\featuredim} \rightarrow \mathbb{R}$  is differentiable if it 
		has a \gls{gradient} $\nabla f ( \mathbf{x})$ everywhere (for every $\mathbf{x} \in \mathbb{R}^{\featuredim}$).},
	first={differentiable},text={differentiable} 
}

\newglossaryentry{gradient}
{name={gradient},
	description={For a real-valued function $f: \mathbb{R}^{\featuredim} \rightarrow \mathbb{R}: \weights \mapsto f(\weights)$, 
	a vector $\va$ such that $\lim_{\weights \rightarrow \weights'} \frac{f(\weights) - \big(f(\weights')+ \va^{T} (\weights- \weights') \big) }{\| \weights-\weights'\|}=0$ is 
	referred to as the gradient of $f$ at $\weights'$. If such a vector exists it is denoted $\nabla f(\weights')$ or $\nabla f(\weights)\big|_{\weights'}$  .},
	first={gradient},text={gradient} 
}

\newglossaryentry{subgradient}
{name={subgradient},
description={For a real-valued function $f: \mathbb{R}^{\featuredim} \rightarrow \mathbb{R}: \weights \mapsto f(\weights)$, 
		a vector $\va$ such that $f(\weights) \geq  f(\weights') +\big(\weights-\weights' \big)^{T} \va$ is 
		referred to as a subgradient of $f$ at $\weights'$.},
	first={subgradient},text={subgradient} 
}

\newglossaryentry{relu}
{name={ReLU},
	description={The rectified linear unit or ``ReLU'' is a popular choice for the \gls{actfun} of a neuron within an \gls{ann}. 
	It is defined as $g(z) = \max\{0,z\}$ with $z$ being the weighted input of the neuron.}, first = {rectified linear unit (ReLU)}, text={ReLU} 
}

\newglossaryentry{hypothesis}
{name={hypothesis},
	description={A map (or function) $h: \featurespace \rightarrow \labelspace$ from the 
		feature space $\featurespace$ to the label space $\labelspace$. 
		Given a \gls{datapoint} with features $\featurevec$ we use a hypothesis map $h$
		to estimate (or approximate) the label $\truelabel$ using the predicted label 
		$\hat{\truelabel} = h(\featurevec)$. ML is about learning (or finding) a hypothesis map $h$
		such that $\truelabel \approx h(\featurevec)$ for any \gls{datapoint}.},
	first={hypothesis},text={hypothesis}  
}

\newglossaryentry{vcdim}
{name={Vapnik–Chervonenkis (VC) dimension},
	description={The VC dimension of an infinite \gls{hypospace} is a widely-used measure for its size. 
		We refer to \cite{ShalevMLBook} for a precise definition of VC dimension as well as a discussion 
		of its basic properties and use in ML.},
	first={Vapnik–Chervonenkis (VC) dimension},text={VC dimension}  
}

\newglossaryentry{effdim}
{name={effective dimension},
	description={The effective dimension $\effdim{\hypospace}$ of an infinite \gls{hypospace} $\hypospace$ 
		is a measure of its size. Loosely speaking, the effective dimension is equal to the number 
		of ``independent'' tunable parameters of the model. These parameters might be the coefficients  
		used in a linear map or the weights and bias terms of an \gls{ann}.},
	first={effective dimension},text={effective dimension}  
}

\newglossaryentry{labelspace}
{name={label space},
	description={Consider a ML application that involves \gls{datapoint}s characterized by features 
		and labels. The label space of a given ML application or method is constituted by all potential 
		values that the label of a \gls{datapoint} can take on. A popular choice for the label space in 
		regression problems (or methods) is $\labelspace = \mathbb{R}$. Binary classification problems 
		(or methods) use a label space that consists of two different elements, e.g., $\labelspace =\{-1,1\}$, $\labelspace=\{0,1\}$ 
		or $\labelspace = \{ \mbox{``cat image''}, \mbox{''no cat image''} \}$  }, first={label space},text={label space}  
}

\newglossaryentry{histogram}
{name={histogram},
	description={Consider a dataset $\dataset$ consisting of \gls{datapoint}s $\datapoint^{(1)},\ldots,\datapoint^{(\samplesize)}$ 
		that belong to some box in $\mathbb{R}^{\featuredim}$. We partition this hyper-rectangle evenly into small elementary boxes. 
		The histogram of $\dataset$ is the assignment of each elementary box to the corresponding fractions of datapoints in $\dataset$ 
		that belong to this elementary box. 
	},
	first={histogram},text={histogram}  
}

\newglossaryentry{bootstrap}
{name={bootstrap},
	description={Consider a probabilistic model that interprets a given set of \gls{datapoint}s $\dataset = \big\{ \datapoint^{(1)},\ldots,\datapoint^{(\samplesize)}\big\}$ as realizations of \gls{iid} \gls{rv}s with a 
		common \gls{probdist} $p(\datapoint)$. The bootstrap uses the histogram of $\dataset$ as the 
		underlying proability distribution $p(\datapoint)$. 
	},
	first={bootstrap},text={bootstrap}  
}

\newglossaryentry{featurespace}
{name={feature space},
	description={
		The feature space of a given ML application or method is constituted by all potential 
		values that the feature vector of a \gls{datapoint} can take on. Within this book the most 
		frequently used choice for the feature space is the \gls{euclidspace} $\mathbb{R}^{\featuredim}$ 
		with dimension $\featurelen$ being the number of individual features of a \gls{datapoint}.},
	first={feature space},text={feature space}  
}

\newglossaryentry{missingdata}
{name={missing data},
	description={By missing data, we refer to a situation where some feature values of a 
		subset of \gls{datapoint}s are unknown. Data imputation techniaues aim at estimating 
		(predicting) these missing feature values \cite{Abayomi2008DiagnosticsFM}. },
	first={missing data},text={missing data}  
}

\newglossaryentry{psd}
{name={positive semi-definite},
	description={A symmetric matrix $\mQ = \mQ^{T} \in \mathbb{R}^{\featuredim \times \featuredim}$ 
		is referred to as positive semi-definite if $\featurevec^{T} \mQ \featurevec \geq 0$ for every 
		vector $\featurevec \in \mathbb{R}^{\featuredim}$.},
	first={positive semi-definite (psd)},text={psd}  
}

\newglossaryentry{features}
{name={features},
	description={Features are those properties of a \gls{datapoint} that can be measured or computed in an 
		automated fashion. For example, if a \gls{datapoint} is a bitmap image, then we could use 
		the red-green-blue intensities of its pixels as features. Some widely used synonyms for 
		the term feature are ``covariate'',``explanatory variable'', ``independent variable'', ``input (variable)'', ``predictor (variable)'' or ``regressor''  \cite{Gujarati2021,Dodge2003,Everitt2022}. 
		However, this book makes consequent use of the term features for low-level properties 
		of \gls{datapoint}s that can be measured easily.}, first={features},text={features}  
}

\newglossaryentry{label}
{name={label},
	description={A higher level fact or quantity of interest associated with a \gls{datapoint}. If a \gls{datapoint} is an image, 
		its label might be the fact that it shows a cat (or not). Some widely used synonyms for 
		the term label are "response variable", "output variable" or "target"  \cite{Gujarati2021,Dodge2003,Everitt2022}.
 },
	first={label},text={label}  
}

\newglossaryentry{noniid}
{name={non-i.i.d.},
	description={See \gls{noniiddata}.},first={non-i.i.d.},text={non-i.i.d.}  
}

\newglossaryentry{data}
{name={data},
	description={A set of \gls{datapoint}s.},first={data},text={data}  
}

\newglossaryentry{dataset}
{name={dataset},
	description={With a slight abuse of notation we use the terms ``dataset`` or ``set of datapoints'' to 
		refer to an indexed list of \gls{datapoint}s $\datapoint^{(1)},\ldots,$. Thus, there is a first \gls{datapoint} 
		$\datapoint^{(1)}$, a second \gls{datapoint} $\datapoint^{(2)}$ and so on. Strictly speaking a dataset is a list 
		and not a set \cite{HalmosSet}. By using indexed lists of \gls{datapoint}s we avoid some of the challenges 
		arising in concept of an abstract set.},first={dataset},text={dataset}  
}

\newglossaryentry{predictor}
{name={predictor},
	description={A predictor is a \gls{hypothesis} whose function values are numeric, such as real numbers. 
		Given a \gls{datapoint} with features $\featurevec$, the predictor value $h(\featurevec) \in \mathbb{R}$ 
		is used as a prediction (estimate/guess/approximation) for the true numeric label $\truelabel \in \mathbb{R}$ of the \gls{datapoint}. },first={predictor},text={predictor}  
}

\newglossaryentry{labeled datapoint}
{name={labeled datapoint},
 description={A \gls{datapoint} whose label is known or has been determined by some means (might require human experts).},
 first={labeled datapoint},text={labeled datapoint}  
}

\newglossaryentry{rv}
{name={random variable},
 description={A random variable is a mapping for function from a set of elementary events to a set of values. 
 	The set of elementary events is equipped with a probability measure that assigns subsets of elemtary events 
 	a probability. A binary random variable maps elementary events to a set containing two different value, such as 
 	$\{-1,1\}$ or $\{ \mbox{cat}, \mbox{no cat} \}$. A real-valued random variable maps elementary events to 
 	real numbers $\mathbb{R}$. A vector-valued random variable maps elementary events to the \gls{euclidspace} $\mathbb{R}^{\featuredim}$. 
 	Probability theory uses the concept of measurable spaces to rigorously define and study the properties of (large) 
 	collections of random variables \cite{GrayProbBook,BillingsleyProbMeasure}.}, first={random variable (RV)},text={RV}  }

\newglossaryentry{trainset}
{
	name={training set},
	description={A set of data points that is used in \gls{erm} to train a hypothesis $\hat{h}$. The average 
		loss of $\hat{h}$ on the training set is referred to as the training error. The comparison between training and 
		validation error informs adaptations of the ML method (such as using a different \gls{hypospace}).},first={training set},text={training set}  
}

\newglossaryentry{trainerr}
{
	name={training error},
	description={Consider a ML method that aims at learning a hypothesis $h \in \hypospace$ out of a \gls{hypospace}. 
		We refer to the average loss or \gls{emprisk} of a hypothesis $h \in \hypospace$ on a dataset as training error if it 
		is used to choose between different hypotheses. The principle of \gls{erm} is find the hypothesis $h^{*} \in \hypospace$ 
		with smallest training error. Overloading the notation a bit, we might refer by training error also to the minimum \gls{emprisk} 
		achieved by the optimal hypothesis $h^{*} \in \hypospace$.},first={training error},text={training error}  
}

\newglossaryentry{datapoint}
{
	name={data point},
	description={A \gls{datapoint} is any object that conveys information \cite{coverthomas}. Data points might be 
		students, radio signals, trees, forests, images, \gls{rv}s, real numbers or proteins. We characterize data points 
		using two types of properties. One type of property is referred to as a feature. \Gls{features} are properties of a 
		\gls{datapoint} that can be measured or computed in an automated fashion. Another type of property is referred to as a 
		\gls{label}. The label of a \gls{datapoint} represents a higher-level facts or quantities of interest. In contrast to \gls{features}, 
		determining the label of a \gls{datapoint} typically requires human experts (domain experts). Roughly speaking, ML aims 
		at predicting the label of a \gls{datapoint} based solely on its features.},first={data point},text={data point}  
}

\newglossaryentry{valerr}
{name={validation error},
 description={Consider a hypothesis $\hat{h}$ which is obtained by \gls{erm} on a \gls{trainset}. The 
 average loss of $\hat{h}$ on a \gls{valset}, which is different from the \gls{trainset}, is referred to as 
 the validation error.},first={validation error},text={validation error}  
}

\newglossaryentry{valset}
{name={validation set},
  description={A set of \gls{datapoint}s that has not been used as \gls{trainset} in \gls{erm} to train a hypothesis $\hat{h}$. 
  The average loss of $\hat{h}$ on the validation set is referred to as the validation error and used to diagnose 
  the ML method. The comparison between training and validation error informs adaptations of the 
  ML method (such as using a different \gls{hypospace}).},first={validation set},text={validation set}  
}

\newglossaryentry{testset}
{name={test set},
	description={A set of \gls{datapoint}s that have neither been used in a \gls{trainset} to learn parameters of a model 
		nor in a \gls{valset} to choose between different models (by comparing validation errors).},first={test set},text={test set}  
}

\newglossaryentry{linclass}{name={linear classifier}, description={A classifier $h(\featurevec)$ maps the 
		feature vector $\featurevec \in \mathbb{R}^{\featuredim}$ of a datapoint to a predicted label $\hat{\truelabel} \in \labelspace$ out of 
		a finite set of label values $\labelspace$. We can characterize such a classifier equivalently by the decision regions 
		$\decreg{a}$, for every possible label value $a \in \labelspace$. Linear classifiers are such that the boundaries 
		between the regions $\decreg{a}$ are hyperplanes in $\mathbb{R}^{\featuredim}$.  },first={linear classifier},text={linear classifier} }

\newglossaryentry{erm}{name={empirical risk minimization}, description={Empirical risk minimization is the optimization problem 
		of finding the hypothesis with minimum average loss (empirical risk) on a given set of \gls{datapoint}s (the \gls{trainset}).
		 Many ML methods are special cases of empirical risk minimization.},first={empirical risk minimization (ERM)},text={ERM} }

\newglossaryentry{multilabelclass}{name={multi-label classification}, description={Multi-label classification problems and methods involve 
	\gls{datapoint}s that are characterized by several individual labels.},first={multi-label classification},text={multi-label classification} }

\newglossaryentry{ssl}{name={semi-supervised learning}, description={Semi-supervised learning methods use (large amounts of) 
		unlabeled \gls{datapoint}s to support the learning of a hypothesis from (a small number of) labeled \gls{datapoint}s \cite{SemiSupervisedBook}. },first={semi-supervised learning (SSL)},text={SSL} }

\newglossaryentry{regularization}{name={regularization}, description={Regularization techniques modify the \gls{erm} 
		principle such that the learnt hypothesis performs well also outside the \gls{trainset} which is used in \gls{erm}. 
		One specific approach to regularization is by adding a penalty or regularization term to the 
		objective function of \gls{erm} (which is the average loss on the \gls{trainset}). This regulazation term 
	    can be interpreted as an estimate for the increase in the expected loss (risk) compared to the average loss on the \gls{trainset}. },first={regularization},text={regularization} }

\newglossaryentry{rerm}{name={regularized empirical risk minimization}, description={Regularized empirical risk minimization is the problem 
		of finding the hypothesis that optimally balances the average loss (empirical risk) on a \gls{trainset} with 
		a regularization term. The regularization term penalizes a hypothesis that is not robust against (small) perturbations 
		of the \gls{datapoint}s in the \gls{trainset}.},first={regularized empirical risk minimization (RERM)},text={RERM} }
	
\newglossaryentry{gtv}{name={generalized total variation}, description={Generalized total variation measures the changes of 
	vector-valued node attributes of a graph.},first={generalized total varation (GTV)},text={GTV} }
	
\newglossaryentry{srm}{name={structural risk minimization}, description={Structural risk minimization is the problem 
		of finding the hypothesis that optimally balances the average loss (empirical risk) on a \gls{trainset} with 
		a regularization term. The regularization term penalizes a hypothesis that is not robust against (small) perturbations 
	of the \gls{datapoint}s in the \gls{trainset}.},first={structural risk minimization (SRM)},text={SRM} }

\newglossaryentry{netexpfam}{name={networked exponential families}, description={A networked collection of 
		exponential families having a separaate parameter vector for each node of the network. These parameter 
		vectors are coupled via the network structure. },first={networked exponential family (nExpFam)},text={nExpFam} }

\newglossaryentry{scatterplot}{name={scatterplot}, description={A visualization technique that depicts data points by markers in a 
	two-dimensional plane.},first={scatterplot},text={scatterplot} }

\newglossaryentry{stepsize}{name={step size}, description={
		Many ML methods use iterative optimization methods (such as \gls{gdmethods}) 
		to construct a sequence of increasinbly accurate hypothesis 
		maps $h^{(1)},h^{(2)},\ldots$. The $\itercntr$th iteration of such an algorithm starts from the current 
		hypothesis $h^{(\itercntr)}$ and tries to modify it to obtain an improved hypothesis $h^{(\itercntr+1)}$. 
		Iterative algorithms often use a step size (hyper-) parameter. The step size controls the amount by 
		which a single iteration can change or modify the current hypothesis. Since the overall goal of such 
		iteration ML methods is to learn a (approximately) optimal hypothesis we refer to a step size parameter 
		also as a \gls{learnrate}.}, 
	first={step size},text={step size} }

\newglossaryentry{learnrate}{name={learning rate}, description={Consider an iterative method for finding or learning 
				a good choice for a hypothesis. Such an iterative method repeats similar computational (update) steps that adjust 
				or modify the current choice for the hypothesis to obtain an improved hypothesis. A prime example for such an 
				iterative learning method is \gls{gd} and its variants (see \ref{ch_GD}). We refer by learning rate to any parameter 
				of an iterative learning method that controls the extent by which the current hypothesis might be modified or 
				improved in each iteration. A prime example for such a parameter is the step size used in \gls{gd}. Within this book 
		we use the term learning rate mostly as a synonym for the step size of (a variant of) \gls{gd}},
	first={learning rate},text={learning rate} }

\newglossaryentry{featuremap}{name={feature map}, description={A map that transforms some raw features into a 
		new feature vector. The new feature vector might be preferable over the raw features for several reasons. It might be possible 
	to use linear hypothesis with the new feature vectors. Another reason could be that the new feature vector is much shorter and 
therefore avoids overfitting or can be used for a \gls{scatterplot}},
	first={feature map},text={feature map} }
 
  \newglossaryentry{explainability}{name={explainability}, 
 	description={We use the term explainability in a highly informal fashion as a measure for 
 		 the predicatability of a ML method (output). A ML method is perfectly explained to a 
 		 user if she can, upon receiving this explanation, perfectly anticipate the behaviour of the ML method. 
 	},
 	first={explainability},text={explainability} }

  \newglossaryentry{lasso}{name={ least absolute shrinkage and selection operator (Lasso)}, 
	description={The least absolute shrinkage and selection operator (Lasso) is an instance of \gls{srm} for 
		learning the weights $\weights$ of a linear map $h(\featurevec) = \weights^{T} \featurevec$. 
		The Lasso minimizes the sum consisting of an average squared error loss (as in \gls{linreg}) 
		and the scaled $\ell_{1}$ norm of the weight vector $\weights$. 
	},
	first={ least absolute shrinkage and selection operator (Lasso)},text={Lasso} }

 \newglossaryentry{simgraph}{name={similarity graph}, 
 	description={Some applications generate \gls{datapoint}s that are related by a domain-specific 
 		notion of similarity. These similarities can be represented conveniently using a similarity 
 		graph $\graph = \big(\nodes \defeq \{1,\ldots,\samplesize\},\edges\big)$. 
 		The node $\sampleidx \in \nodes$ represents the $\sampleidx$-th \gls{datapoint}. Two 
 		nodes are connected by an undirected edge if the corresponding \gls{datapoint}s are similar. 
 	},
 	first={similarity graph},text={similarity graph} }

\newglossaryentry{LapMat}{name={Laplacian matrix}, 
	description={
		The geometry or structure of a \gls{simgraph} $\graph$ can be analyzed using the properties 
		of special matrices that are associated with $\graph$. One such matrix is the graph Laplacian 
		matrix $\mL$ whose entries are defined in \eqref{equ_def_laplacian_sim_graph}. 
	},
	first={Laplacian matrix},text={Laplacian matrix} }

\newglossaryentry{cm}{name={confusion matrix}, 
	description={Consider \gls{datapoint}s characterized by features $\featurevec$ and label $\truelabel$ 
		having value $\clusteridx \in \{1,\ldots,\nrcluster\}$. The confusion matrix is $\nrcluster \times \nrcluster$ matrix 
		with rows representing different values $\clusteridx$ of the true label of a \gls{datapoint}. 
		The columns of a confusion matrix correspond to different values $\clusteridx'$ delivered by a hypothesis $h(\featurevec)$. 
		The $(\clusteridx,\clusteridx')$-th entry of the confusion matrix is the fraction of \gls{datapoint}s 
		with label $\truelabel\!=\! \clusteridx$ and predicted label $\hat{\truelabel}\!=\!\clusteridx'$.},
	first={confusion matrix},text={confusion matrix} }

\newglossaryentry{dbscan}{name={``density-based spatial clustering of applications with noise'' (DBSCAN)}, 
	description={A clustering algorithm for \gls{datapoint}s that are characterized by numeric feature vectors. 
		Similar to \gls{kmeans} and \gls{softclustering} via \gls{gmm} also DBSCAN uses the Euclidean 
		distances between feature vectors to determine the clusters. However, in contrast to these other 
		clustering methodes, DBSCAN uses a different notion of similarity between \gls{datapoint}s. 
		In particular, DBSCAN considers two \gls{datapoint}s as similar if they are ``connected'' via 
		a sequence (path) of close-by intermediate \gls{datapoint}s. Thus, DBSCAN might consider 
		two \gls{datapoint}s as similar (and therefore belonging to the same cluster) even if 
		their feature vectors have a large Euclidean distance.},
	first={density-based spatial clustering of applications with noise (DBSCAN)},text={DBSCAN} }

\newglossaryentry{fl}{name={federated learning (FL)}, description={Federated learning is an 
		umbrella term for ML methods that train models in a collaborative fashion using decentralized data and computation.},
	first={federated learning (FL)},text={FL} }

\newglossaryentry{iid}{name={i.i.d.}, description={It can be useful to 
		interpret \gls{datapoint}s $\datapoint^{(1)},\ldots,\datapoint^{(\samplesize)}$ 
		as realizatons of independent and identically distributed \gls{rv}s with a common \gls{probdist}. 
		If these \gls{rv}s are continous, their joint \gls{pdf} is $p\big(\datapoint^{(1)},\ldots,\datapoint^{(\samplesize)} \big) = \prod_{\sampleidx=1}^{\samplesize} p \big(\datapoint^{(\sampleidx)}\big)$ with $p(\datapoint)$ being the common 
		marginal \gls{pdf} of the underlying \gls{rv}s.},
	first={independent and identically distributed (iid)},text={iid} }

\newglossaryentry{outlier}{name={outlier}, description={Many ML methods are 
		motivated by the \gls{iidasspt} which interprets \gls{datapoint}s as realizations of 
		\gls{iid} \gls{rv}s with a common \gls{probdist}. The \gls{iidasspt} is useful for applications  
		where the statistical properties of the data generation process are stationary (time-invariant). 
		However, in some applications the data consists of a majority of ``regular'' \gls{datapoint}s 
		that conform with an \gls{iidasspt} and a small number of data points that have fundamentally different 
        statistical properties compared to the regular \gls{datapoint}s. We refer to a \gls{datapoint} that 
        substantially deviates from the statistical properties of the majority of \gls{datapoint}s as an 
        outlier. Different methods for outlier detection use different measures for this deviation.},
	          first={outlier},text={outlier} }

\newglossaryentry{decisionregion}{name={decision region}, description={Consider a hypothesis map $h$ that can 
		only take values from a finite set $\labelspace$. We refer to the set of features $\featurevec \in \featurespace$ 
		that result in the same output $h(\featurevec)=a$ as a decision region of the hypothesis $h$. },first={decision region},text={decision region} }

\newglossaryentry{decisionboundary}{name={decision region}, description={Consider a hypothesis map $h$ that reads in 
		a feature vector $\featurevec \in \mathbb{R}^{\featuredim}$ and delivers a value from a finite set $\labelspace$. 
		The decision boundary induced by $h$ is the set of vectors $\featurevec \in \mathbb{R}^{\featuredim}$ that lie between 
		different\gls{decisionregion}s. More precisely, a vector $\featurevec$ belongs to the decision boundary if and only if 
	each neighborhood $\{ \featurevec': \| \featurevec - \featurevec' \| \leq \varepsilon \}$, for any $\varepsilon >0$, contains 
  at least two vectors with different function values.},first={decision boundary},text={decision boundary} }

\newglossaryentry{euclidspace}{name={Euclidean space}, description={The Euclidean space $\mathbb{R}^{\featuredim}$ of dimension $\featuredim$ refers to the space of all vectors $\featurevec= \big(\feature_{1},\ldots,\feature_{\featurelen}\big)$, with real-valued entries $\feature_{1},\ldots,\feature_{\featuredim} \in \mathbb{R}$, 
whose geometry is defined by the inner product $\featurevec^{T} \featurevec' = \sum_{\featureidx=1}^{\featuredim} \feature_{\featureidx} \feature'_{\featureidx}$ between any two 
vectors $\featurevec,\featurevec' \in \mathbb{R}^{\featuredim}$ \cite{RudinBookPrinciplesMatheAnalysis}.},first={Euclidean space},text={Euclidean space} }

\newglossaryentry{eerm}{name={explainable empirical risk minimization}, description={An instance of structural risk minimization that adds a regularization term 
		to the training error in \gls{erm}. The regularization term is chosen to favour hypotheses that are intrinsically explainable for a user.},first={explainable empirical risk minimization (EERM)},text={EERM} }

\newglossaryentry{kmeans}{name={$k$-means}, description={The $k$-means algorithm is a hard 
		clustering method. It aims at assigning data points to clusters such that they have minimum 
		average distance from the cluster centre.},first={$k$-means},text={$k$-means} }

\newglossaryentry{xml}{name={explainable machine learning}, description={Explainable ML methods aim at 
		complementing predictions with explanations for how the prediction has been obtained.},first={explainable ML},text={explainable ML} }

\newglossaryentry{fmi}{name={Finnish Meteorological Institute}, description={The Finnish Meteorological Institute is a 
		government agency responsible for gathering and reporting weather data in Finland.},first={Finnish Meteorological Institute (FMI)},text={FMI} }

\newglossaryentry{highdimregime}{name={high-dimensional regime}, description={A ML method 
		or problem belongs to the high-dimensional regime if the \gls{effdim} of the \gls{model} is larger 
		than the number of available (labeled) \gls{datapoint}s. For example, linear regression belongs to the 
   	    high-dimensional regime whenever the number $\featuredim$ of features used to characterize datapoints is 
       larger than the number of \gls{datapoint}s in the \gls{trainset}. Another example for the high-dimensional 
       regime are deep learning methods that use a hypothesis space generated by a \gls{ann} with much more tunable 
       weights than the number of \gls{datapoint}s in the \gls{trainset}. The recent field of high-dimensional statistics 
       uses probability theory to analyze ML methods in the high-dimensional regime \cite{Wain2019,BuhlGeerBook}.},
   first={high-dimensional regime},text={high-dimensional regime} }

\newglossaryentry{gmm}{name={Gaussian mixture model}, description={Gaussian mixture models (GMM) are a family of 
		probabilistic models for  \gls{datapoint}s. Within a GMM, the feature vector $\featurevec$ of a \gls{datapoint} is 
		interpreted as being drawn from one out of $\nrcluster$ different multivariate normal (Gaussian) distributions 
		indexed by $\clusteridx=1,\ldots,\nrcluster$. The probability that the feature vector $\featurevec$ is drawn from 
		the $\clusteridx$-th Gaussian distribution is denoted $p_{\clusteridx}$. The GMM is parametrized by the probability 
		$p_{\clusteridx}$ of $\featurevec$ being drawn from the $\clusteridx$-th Gaussian distribution as well as the 
		mean vectors $\clustermean^{(\clusteridx)}$ and covariance matrices $\clustercov^{(\clusteridx)}$ for $\clusteridx=1,\ldots,\nrcluster$. 
	 },first={Gaussian mixture model (GMM)},text={GMM} }
 
\newglossaryentry{ml}{name={maximum likelihood}, description={
		Consider \gls{datapoint}s that are interpreted as \gls{iid} realizations of \gls{rv}s with a common (but unknown) 
		\gls{probdist}. Maximum likelihood methods find a parameter vector $\weights$ for a probabilistic 
		model $\prob{\datapoint; \weights}$ such that the probability (density) of observing the actucal data is maximized. 
		Loosely speaking, we try out all possible parameter vectors $\weights$ and determine the resulting 
		probability of observing the given datapoints if they would be \gls{iid} with common \gls{probdist} $\prob{\datapoint; \weights}$. 
		The maximum likelihood estimator is the parameter vector that results in the highest probability (density). 
	},first={maximum likelihood},text={maximum likelihood}}

\newglossaryentry{em}{name={expectation maximization}, description={Expectation maximization is a generic 
		technique for estimating the parameters of a probabilistic model (a parametrized \gls{probdist}) $\prob{\datapoint; \weights}$ 
		from data \cite{BishopBook,hastie01statisticallearning,GraphModExpFamVarInfWainJor}. Expectation maximization 
		delivers an approximation to the \gls{ml} estimate for the model parameters $\weights$. 
  },first={expectation maximization (EM)},text={EM}}

\newglossaryentry{ppca}{name={probabilistic PCA}, description={Probabilistic PCA extends basic \gls{pca} by using a 
		probabilistic model for \gls{datapoint}s. Within this probabilistic model, the task of dimensionality reduction 
		becomes an estimation problem that can be solved using \gls{em} methods.},first={probabilistic PCA (PPCA)},text={PPCA}}

\newglossaryentry{linreg}{name={linear regression}, description={Linear regression aims at learning a 
		linear hypothesis map to predict a numeric \gls{label} based on numeric \gls{features} of a \gls{datapoint}. The quality 
	of a linear hypothesis map is typically measured using the average squared error loss incurred on a set 
	of labeled \gls{datapoint}s (the \gls{trainset}).},first={linear regression},text={linear regression}}

\newglossaryentry{ridgeregression}{name={ridge regression}, description={Ridge regression aims 
		at learning the weights $\weights$ of linear hypothesis map $h^{(\weights)}(\featurevec)= \weights^{T} \featurevec$. The quality of a particular choice for the weights $\weights$ is measured by the sum of two terms (see \eqref{equ_rerm_ridge_regression}). 
		The first term is the average squared error \gls{loss} incurred by $h^{(\weights)}$ on a set of 
		labeled \gls{datapoint}s (the \gls{trainset}). 
        The second term is the scaled squared Euclidean norm $\regparam \| \weights \|^{2}_{2}$ with a 
         regularization parameter $\regparam > 0$. },first={ridge regression},text={ridge regression}}

\newglossaryentry{logreg}{name={logistic regression}, description={Logistic regression aims at learning a 
		linear hypothesis map to predict a binary \gls{label} based on numeric \gls{features} of a \gls{datapoint}. 
		The quality of a linear hypothesis map (classifier) is measured using its average logistic loss on 
		some labeled datapoints (the \gls{trainset}).},first={logistic regression},text={logistic regression}}
	
\newglossaryentry{logloss}{name={logistic loss}, description={Consider a \gls{datapoint} that is characterized by the features $\featurevec$ 
		and a binary label $\truelabel \in \{-1,1\}$. We use a hypothesis $h$ to predict the label $\truelabel$ solely from the features $\featurevec$. 
		The logistic loss incurred by a specific hypothesis $h$ is defined as \eqref{equ_log_loss}.},first={logistic loss},text={logistic loss}}
	
\newglossaryentry{hingeloss}{name={hinge loss}, description={Consider a \gls{datapoint} that is characterized by a 
		feature vector $\featurevec \in \mathbb{R}^{\featuredim}$ and a binary label $\truelabel \in \{-1,1\}$. The hinge loss 
		incurred by a specific hypothesis $h$ is defined as \eqref{equ_hinge_loss}. A regularized variant of the hinge loss 
		is used by the \gls{svm} to learn a \gls{linclass} with maximum margin between the two classes (see Figure \ref{fig_svm}). 
	},first={hinge loss},text={hinge loss}}

\newglossaryentry{iidasspt}{name={i.i.d.\ assumption}, description={The i.i.d.\ assumption interprets \gls{datapoint}s of a \gls{dataset} 
		as the realizations of \gls{iid} \gls{rv}s.},first={i.i.d.\ assumption},text={i.i.d.\ assumption} }

\newglossaryentry{hypospace}{name={hypothesis space}, description={Every practical ML method uses a specific hypothesis
		 space (or \gls{model}) $\hypospace$. The hypothesis space of a ML method is a subset of all possible maps 
		from the \gls{featurespace} to \gls{labelspace}. The design choice of the hypothesis space should take into 
		account available computational resources and statistical aspects. If the computational infrastructure allows for efficient matrix 
		operations and we expect a linear relation between feature values and label, a resonable first candidate 
		for the hypothesis space is the space of linear maps \eqref{equ_def_hypo_linear_pred}.},first={hypothesis space},text={hypothesis space} }
	
\newglossaryentry{model}{name={model}, description={We use the term model as a synonym for \gls{hypospace}},first={model},text={model} }

\newglossaryentry{ai}{name={artificial intelligence}, description={Artificial intelligence aims at developing systems that behave 
		rational in the sense of maximizing a long-term reward.},first={artificial intelligence (AI)},text={AI} }
	
\newglossaryentry{hardclustering}{name={hard clustering}, description={Hard clustering refers to the task of 
		partitioning a given set of \gls{datapoint}s into (few) non-overlapping clusters. Each \gls{datapoint} is assigned to one specific cluster.},first={hard clustering},text={hard clustering} }
	
\newglossaryentry{softclustering}{name={soft clustering}, description={Soft clustering refers to the task 
		of partitioning a given set of \gls{datapoint}s into (few) overlapping clusters. Each \gls{datapoint} is 
		assigned to several different clusters with varying \gls{dob}. Soft clustering amounts to determining 
		such a \gls{dob} (or soft cluster assignment) for each \gls{datapoint} and each \gls{cluster}.},first={soft clustering},text={soft clustering} }
	
\newglossaryentry{clustering}{name={clustering}, description={Clustering means to decompose a givne 
		set of \gls{datapoint}s into few subsets, which are referred to as clusters, that consist of similar \gls{datapoint}s. 
		Different clustering methods use different measurs for the similarity between \gls{datapoint}s and different 
		representation of cluters. The clustering method \gls{kmeans} uses the average feature vector (``cluster means'') 
		of a cluster as its repsentative (see Section \ref{sec_hard_clustering}). A popular soft clustering method 
		based on \gls{gmm} represents a cluster by a Gaussian (multivariate normal) \gls{probdist} (see Section \ref{sec_soft_clustering}).},first={clustering},text={clustering} }
	
\newglossaryentry{cluster}{name={cluster}, description={A cluster is a subset of \gls{datapoint}s that are more 
	similar to each other than to the \gls{datapoint}s outside the cluster.	The notion and measure of similarity 
	between \gls{datapoint}s is a design choice. If \gls{datapoint}s are characterized by numeric Euclidean feature 
	 vectors it might be reasonable to define similarity between two \gls{datapoint}s using the (inverse of th) Euclidean distance 
    between the corresponding feature vectors },first={cluster},text={cluster} }

\newglossaryentry{huberloss}{name={Huber loss}, description={The Huber loss is a mixture of the squared error loss 
		and the absolute value of the prediction error.},first={Huber loss},text={Huber loss} }

\newglossaryentry{svm}{name={support vector machine}, description={A binary classification method for learning a 
		linear map that maximally seperates \gls{datapoint}s the two classes in the feature space (``maximum margin''). 
		Maximizing this separation is equivalent to minimizing a regularized variant of the \gls{hingeloss} \eqref{equ_hinge_loss}.},first={support vector machine (SVM)},text={SVM} }

\newglossaryentry{eigenvalue}{name={eigenvalue}, description={We refer to a number $\lambda \in \mathbb{R}$ 
		as eigenvalue of a square matrix $\mathbf{A} \in \mathbb{R}^{\featuredim \times \featuredim}$ if there is a 
		non-zero vector $\vx \in \mathbb{R}^{\featuredim} \setminus \{ \mathbf{0} \}$ such that $\mathbf{A} \vx = \lambda \vx$. },first={eigenvalue},text={eigenvalue} }
	
\newglossaryentry{eigenvector}{name={eigenvector}, description={An eigenvector of a matrix $\mathbf{A}$ is a 
		non-zero vector $\vx \in \mathbb{R}^{\featuredim} \setminus \{ \mathbf{0} \}$ such that $\mathbf{A} \vx = \lambda \vx$ 
	with some \gls{eigenvalue} $\lambda$.},first={eigenvector},text={eigenvector} }

\newglossaryentry{evd}{name={eigenvalue decomposition}, description={The task of computing the eigenvalues and corresponding eigenvectors of a matrix.},first={eigenvalue decomposition (EVD)},text={EVD} }

\newglossaryentry{gdmethods}{name={gradient-based method}, description={Gradient-based methods 
		are iterative algorithms for finding the minimum (or maximum) of a differentiable objective function 
		of a parameter vector. These algorithms construct a sequence of approximations to an optimal parameter 
		vector whose function value is minimal (or maximal). As their name indicates, gradient-based methods 
		use the gradients of the objective function evaluated during previous iterations to construct a new (hopefully) 
		improved approximation of an optimal parameter vector.},first={gradient-based methods},text={gradient-based methods} }

\newglossaryentry{sgd}{name={subgradient descent}, description={Subgradient descent is a 
		generalization of \gls{gd} that is obtained by using sub-gradients (instead of gradients) to 
		construct local approximations of an objective function such as the \gls{emprisk} $\emperror\big( h^{(\weights)} \big| \dataset \big)$ 
		as a function of the parameters $\weights$ of a hypothesis $h^{(\weights)}$.},first={subgradient descent},text={subgradient descent} }
	
\newglossaryentry{stochGD}{name={stochastic gradient descent}, description={Stochastic gradient descent is obtained from \gls{gd} by 
	replacing the gradient of the objective function by a noisy (or stochastic) estimate.},first={stochastic gradient descent (SGD)},text={SGD} }

\newglossaryentry{pca}{name={principal component analysis (PCA)}, description={Principal component analysis 
		determines a given number of new features that are obtained by a linear transformation (map) of the 
		raw features. },first={principal component analysis (PCA)},text={PCA} }
	
\newglossaryentry{loss}{name={loss}, description={With a slight abuse of language, we use the term loss either for
		\gls{lossfunc} itself or for its value for a specific pair of \gls{datapoint} and \gls{hypothesis}.},first={loss},text={loss} }

\newglossaryentry{lossfunc}{name={loss function}, description={A loss function is a map 
		$$\lossfun: \featurespace \times \labelspace \times \hypospace \rightarrow \mathbb{R}_{+}: \big( \big(\featurevec,\truelabel\big), h\big) \mapsto  \loss{(\featurevec,\truelabel)}{h}$$ which assigns a pair consisting of a datapoint, with features $\featurevec$ 
		and label $\truelabel$, and a hypothesis $h \in \hypospace$ the non-negative real number $\loss{(\featurevec,\truelabel)}{h}$. 
		The loss value $\loss{(\featurevec,\truelabel)}{h}$ quantifies the discrepancy between the true label $\truelabel$ 
		and the predicted label $h(\featurevec)$. Smaller (closer to zero) values $\loss{(\featurevec,\truelabel)}{h}$ mean 
		a smaller discrepancy between predicted label and true label of a data point. Figure \ref{fig_loss_function} depicts 
		a loss function for a given data point, with features $\featurevec$ and label $\truelabel$, as a function of the 
		hypothesis $h \in \hypospace$.  },first={loss function},text={loss function} }

\newglossaryentry{decisiontree}{name={decision tree}, description={A decision tree is a flow-chart 
		like representation of a hypothesis map $h$. More formally, a decision tree is a directed graph 
		which reads in the feature vector $\featurevec$ of a \gls{datapoint} at its root node. The root node 
		then forwards the \gls{datapoint} to one of its children nodes based on some elementary test on 
		the features $\featurevec$. If the receiving children node is not a leaf node, i.e., it has itself children nodes, 
	  it represents another test. Based on the test result, the \gls{datapoint} is further pushed to one of its neighbours. This testing and 
	  forwarding of the \gls{datapoint} is repeated until the \gls{datapoint} ends up in a leaf node (having no children nodes). 
	  The leaf nodes represent sets (decision regions) constituted by feature vectors $\featurevec$ that are mapped to 
	  the same function value $h(\featurevec)$.},first={decision tree},text={decision tree} }

\newglossaryentry{noniiddata}{name={non i.i.d.\ data}, description={A dataset that cannot be well modelled as 
		realizations of \gls{iid} \gls{rv}s.},first={non-i.i.d.\ data},text={non-i.i.d.\ data} }

\newglossaryentry{API} 
{
	name={Application Programming Interface (API)},
	description={An Application Programming Interface (API) is a particular set of rules and specifications that a software program can follow to access and make use of the services and resources provided by another particular software program that implements that API},
	first={Application Programming Interface (API)},
	text={API}
}

\newglossaryentry{hilbertspace}{name={Hilbert space},description={A Hilbert space is a linear vector space 
		that is equipped with an inner product between pairs of vectors. One important example for a Hilbert 
		space is the Euclidean spaces $\mathbb{R}^{\featuredim}$, for some dimension $\featuredim$, which 
		consists of Euclidean vectors $\vu = \big(u_{1},\ldots,u_{\featurelen}\big)^{T}$ along with the inner 
		product $\vu^{T} \vv$.},first={Hilbert space},text={Hilbert space}}

\newglossaryentry{sample}{name={sample},description={A finite sequence (list) of \gls{datapoint}s $\datapoint^{(1)},\ldots,\datapoint^{(\sampleidx)}$ that 
		is obtained or interpreted as the realizations of $\samplesize$ \gls{iid} \gls{rv}s with the common \gls{probdist} $p(\datapoint)$.
		The length $\samplesize$ of the sequence is referred to as the \gls{samplesize}.},first={sample},text={sample}}
	
\newglossaryentry{samplesize}
{name=sample size,
	description={The number of individual \gls{datapoint}s contained in a dataset that is 
	obtained from realizations of \gls{iid} \gls{rv}s.},first={sample size},text={sample size}
}

\newglossaryentry{ann}
{name=artificial neural network,
	description={An artificial neural network is a graphical (signal-flow) representation of a map from 
		features of a \gls{datapoint} at its input to a predicted label at its output.},first={artificial neural network (ANN)},text={ANN}
}

\newglossaryentry{randomforest}
{name=random forest,
	description={A random forest is a set (ensemble) of different \gls{decisiontree}s. Each of these \gls{decisiontree}s is 
		obtained by fitting a perturbed copy of the original dataset.},first = {random forest}, text={random forest}
}

\newglossaryentry{bagging}{name={bagging},description={bagging (or ``bootstrap aggregation'') is a generic 
		technique to improve or robustify a given ML method. The idea is to use the bootstrap to generate 
		perturbed copy of a given training set and then apply the original ML method to learn a separate 
		hypothesis for each perturbed copy of the training set. The resulting set of hypotheses is then 
		used to predict the label of a \gls{datapoint} by combining or aggregating the individual predictions of each 
		individual hypothesis. For hypotheses that deliver numeric label values (regression methods) this 
		aggregation could be implemented by computing the average of individual predictions.},first={bootstrap aggreation (bagging)},text={bagging}}

\newglossaryentry{gd}{name={gradient descent},description={Gradient descent is an iterative method for finding the minimum 
		of a differentiable function $f(\weights)$. },first={gradient descent (GD)},text={GD}}

\newglossaryentry{ladregression}{name={least absolute deviation regression},description={Least absolute deviation regression uses 
		the average of the absolute precondition errors to find a linear hypothesis.},first={least absolute deviation regression},
	   text={least absolute deviation regression}}

\newglossaryentry{metric}{name={metric},description={A metric refers to a loss function that is used solely 
	    for the final performance evaluation of a learnt hypothesis. The metric is typically a \gls{lossfunc} that 
	    has a ``natural'' interpretation (such as the $0/1$ loss \eqref{equ_def_0_1}) but is not a good choice to guide 
	    the learning process, e.g., via \gls{erm}. For \gls{erm}, we typically prefer \gls{lossfunc}s that depend smoothly 
	    on the (parameters of the) hypothesis. Examples for such smooth loss functions include the squared error 
	    loss \eqref{equ_squared_loss} and the \gls{logloss} \eqref{equ_log_loss}.},first={metric},text={metric}}

\newglossaryentry{bayesrisk}{name={Bayes risk},description={We use the term Bayes risk as a synonym for the \gls{risk} or expected \gls{loss} 
		of a hypothesis. Some authors reserve the term Bayes risk for the \gls{risk} of a hypothesis that achieves minimum \gls{risk}, such a hypothesis 
		being referred to as a \gls{bayesestimator} \cite{LC}.},first={Bayes risk},text={Bayes risk}}
	
\newglossaryentry{bayesestimator}{name={Bayes estimator},description={A hypothesis whose \gls{bayesrisk} is minimal \cite{LC}.},first={Bayes estimator},text={Bayes estimator}}

\newglossaryentry{weights}{name={weights},
	description={We use the term weights synonymously for a finite set of parameters within a \gls{model}. 
		For example, the linear model consists of all linear maps $h(\featurevec)= \weights^{T} \featurevec$ 
		that read in a feature vector $\featurevec=\big(\feature_{1},\ldots,\feature_{\featurelen}\big)^{T}$ of a \gls{datapoint}. 
		Each specific linear map is characterized by specific choices for the parameters for weights $\weights = \big( \weight_{1},\ldots,\weight_{\featuredim}\big)^{T}$.},first={weights},text={weights}}
	
\newglossaryentry{probdist}{name={probability distribution},
	description={The \gls{data} generated in some ML applications can be reasonably well modeled as 
		realizations of a \gls{rv}. The overall statistical properties (or intrinisic structure) of such \gls{data} 
		are then governed by the probability distribution of this \gls{rv}. We use the term probability distribution 
		in a highly informal manner and mean the collection of probabilities assigned to different values or value 
		ranges of a \gls{rv}. The probability distribution of a binary \gls{rv} $\truelabel \in \{0,1\}$ is fully specified 
		by the probabilities $\prob{\truelabel = 0}$ and $\prob{\truelabel=1} \big(\!=\!1\!-\!\prob{\truelabel=0} \big)$. 
	    The probability distribution of a real-valued \gls{rv} $\feature \in \mathbb{R}$ might be specified 
	    by a probability density function $p(\feature)$ such that $\prob{ \feature \in [a,b] } \approx  p(a) |b-a|$. 
	    In the most general case, a probability distribution is defined by a probability measure \cite{GrayProbBook,BillingsleyProbMeasure}.},first={probability distribution},text={probability distribution}}

\newglossaryentry{pdf}{name={probability density function (pdf)},
	description={The probability density function (pdf) $p(\feature)$ of a real-valued 
		\gls{rv} $\feature \in \mathbb{R}$ is a particular representation of its \gls{probdist}. 
		If the pdf exists, it can be used to compute the probability that $\feature$ takes on a 
		value from a (measureable) set $\mathcal{B} \subseteq \mathbb{R}$ via $\prob{\feature \in \mathcal{B}} = \int_{\mathcal{B}} p(\feature') d \feature'$ \cite[Ch. 3]{BertsekasProb}. The pdf of a vector-valued \gls{rv} $\featurevec \in \mathbb{R}^{\featuredim}$ (if it exists) 
        allows to compute the probability that $\featurevec$ falls into a (measurable) region $\mathcal{R}$ via 
        $\prob{\featurevec \in \mathcal{R}} = \int_{\mathcal{R}} p(\featurevec') d \feature_{1}' \ldots d \feature_{\featuredim}' $ \cite[Ch. 3]{BertsekasProb}.},
first={probability density function (pdf)},text={pdf}}

\newglossaryentry{parameters}{name={parameters},
	description={The parameters of a ML \gls{model} are tunable (learnable or adjustable) quantities that 
		allow to choose between different hypothesis maps. For example, the linear model $\hypospace \defeq \{h: h(\feature)= \weight_{1} \feature + \weight_{2}\}$ consists  of all hypothesis maps $h(\feature)= \weight_{1} \feature + \weight_{2}$ with a particular 
		choice for the parameters $\weight_{1},\weight_{2}$. Another example of parameters are the weights 
		assigned to the connections of an \gls{ann}.},first={parameters},text={parameters}}

\newglossaryentry{lln}{name={law of large numbers},
	description={The law of large numbers refers to the convergence of the average of an increasing 
		number of \gls{iid} \gls{rv}s to the mean (or expectation) of their common \gls{probdist}.},first={law of large numbers},text={law of large numbers}}

\newglossaryentry{kCV}{name={$k$-fold cross-validation ($k$-fold CV},
	description={$k$-fold cross-validation divides a dataset evenly into $k$ folds. This algorithm 
	consists of $k$ repetitions, during which one of the folds as the \gls{valset} and the 
	remaining $\nrfolds-1$ folds as a \gls{trainset}.},first={$k$-fold cross-validation ($k$-fold CV)},text={$k$-fold CV}}

\newglossaryentry{nonsmooth}{name={non-smooth},
	description={We refer to a function as non-smooth if it is not \gls{smooth} \cite{nesterov04}.},first={non-smooth},text={non-smooth}}

\newglossaryentry{convex}{name={convex},
	description={A set $\mathcal{C}$ in $\mathbb{R}^{\featuredim}$ is called convex if it contains the line segment 
		between any two points of that set. A function is called convex if its epigraph is a convex set \cite{BoydConvexBook}.},first={convex},text={convex}}

\newglossaryentry{smooth}{name={smooth},
	description={We refer to a real-valued function as smooth if it is differentiable 
		and its \gls{gradient} is continuous \cite{nesterov04,CvxBubeck2015}. In particular, a 
		differentiable function $f(\weights)$ is $\beta$-smooth if the \gls{gradient} 
		$\nabla f(\weights)$ is Lipschitz continuous with Lipschitz constant $\beta$, i.e., $\| \nabla f(\weights) - \nabla f(\weights') \| \leq \beta \| \weights - \weights' \|$. },first={smooth},text={smooth}}

\newglossaryentry{dataug}{name={data augmentation},
	description={Data augmentation methods add synthetic \gls{datapoint}s to an existing set of \gls{datapoint}s. 
		These synthetic \gls{datapoint}s might be obtained by perturbations (adding noise) or 
		transformations (rotations of images) of the original \gls{datapoint}s.},first={data augmentation},text={data augmentation}}

\lstdefinestyle{mystyle}{
	backgroundcolor=\color{backcolour},   
	commentstyle=\color{codegreen},
	keywordstyle=\color{magenta},
	numberstyle=\tiny\color{codegray},
	stringstyle=\color{codepurple},
	basicstyle=\ttfamily\footnotesize,
	breakatwhitespace=false,         
	breaklines=true,                 
	captionpos=b,                    
	keepspaces=true,                 
	numbers=left,                    
	numbersep=5pt,                  
	showspaces=false,                
	showstringspaces=false,
	showtabs=false,                  
	tabsize=2
}

\begin{document}

\pagestyle{plain}

\vspace*{5mm}
\begin{center}
\noindent {\bf {\myfont Machine Learning: The Basics}} \\[4mm]
\noindent {\myfontb Alexander Jung, \today} \\[3mm]
\end{center} 

{\myfontc {\bf please cite as:}}
\vspace*{4mm}

\parbox{\textwidth}{\vrule width.50pt\  {\myfontc A. Jung,``Machine Learning: The Basics," Springer, Singapore, 2022}}


\vspace*{3mm}
\begin{center}
\begin{figure}[htbp]
	\centering
	\hspace*{5mm}
	\begin{tikzpicture}[node distance=1cm, auto,]
	\coordinate (OR) at (0.00, 1.50);
	\node[circle,inner sep=0,minimum size={6cm}](a) at (OR) {};
	\node[circle,inner sep=0,minimum size={8cm}](b) at (OR) {};
	\node[circle,inner sep=0,minimum size={8cm}](c) at (OR) {};
	\draw[red,line width=1,dashed] (a.-150) arc (-150:{-150+120}:3cm);
	\draw[blue,line width=1,dashed] (b.160) arc (160:{160+50}:4cm);
	\draw[black!30!green,line width=2,dashed] (c.20) arc (20:{160}:4cm);
	\draw[black!30!blue,line width=2,dashed] ([shift={(-30:6cm)}]OR) arc (-30:{20}:6cm);
	\draw[black!30!blue,line width=2,dashed] (OR) --([shift={(-30:6cm)}]OR);
	\draw[black!30!blue,line width=2,dashed] (OR) --([shift={(20:6cm)}]OR);
	\draw[black!30!green,line width=2,dashed] (OR) -- (c.20);
	\draw[black!30!green,line width=2,dashed] (OR) -- (c.160);
	\draw[blue,thin,dashed] (OR) -- (b.160);
	\draw[blue,thin,dashed] (OR) -- (b.210);
	\draw[red,thin,dashed] (OR) -- (a.-150);
	\draw[red,thin,dashed] (OR) -- (a.-30);   
	\node[punkt] (data) {observations};
	\node[red,below=0.5cm of data] (data1) {data};
	\node[above=of data] (dummy) {};
	\node[punkt,above=1cm of dummy] (hypothesis) {hypothesis};
	\node[right=1.4cm of dummy] (t) {make prediction} ; 
	\node[left=1.4cm of dummy] (g) {validate/adapt} ;
	\node[blue,below=0.5cm of g,anchor=east] (g1) {loss} ; 
	\node[black!30!blue,below=1cm of t,anchor=west] (g6) {inference} ; 
	\node[black!30!green,above=1cm of hypothesis,anchor=south] (g3) {model} ; 
	\draw [->,line width=0.5mm] (hypothesis.east) to [out=0,in=90] (t.north);
	\draw [->,line width=0.5mm] (t.south) to [out=270,in=0] (data.east);
	\draw [->,line width=0.5mm] (data.west) to [out=180,in=270] (g.south);
	\draw [->,line width=0.5mm] (g.north) to [out=90,in=180] (hypothesis.west);
	\end{tikzpicture}
	\vspace*{-9mm}
	\caption{
		Machine learning combines three main components: model, data and loss. Machine learning 
		methods implement the scientific principle of ``trial and error''. These methods continuously 
		validate and refine a model based on the loss incurred by its predictions about a phenomenon 
		that generates data.}
	\label{fig_AlexMLBP}
\end{figure}
\end{center}



\newpage
\chapter*{Preface}

Machine learning (ML) influences our daily lives in several aspects. We routinely 
ask ML empowered smartphones to suggest lovely restaurants or to guide us 
through a strange place. ML methods have also become standard tools in many 
fields of science and engineering. ML applications transform human lives at 
unprecedented pace and scale. 

This book portrays ML as the combination of three basic components: \index{data}data, \index{model}\gls{model} 
and \index{loss}\gls{loss}. ML methods combine these three components within 
computationally efficient implementations of the basic scientific principle ``trial and error''. 
This principle consists of the continuous adaptation of a hypothesis about a 
phenomenon that generates data. 

ML methods use a hypothesis map to compute predictions of a quantity of interest 
(or higher level fact) that is referred to as \index{label} the label of a \gls{datapoint}. 
A hypothesis map reads in low level properties (referred to as features) of a 
\gls{datapoint} and delivers the prediction for the label of that \gls{datapoint}. ML 
methods choose or learn a hypothesis map from a (typically very) large set of candidate maps. 
We refer to this set as of candidate maps as the \index{hypothesis space} \gls{hypospace} 
or \index{model} \gls{model} underlying an ML method. 

The adaptation or improvement of the hypothesis is based on the discrepancy between 
predictions and observed data. ML methods use a loss function to quantify this discrepancy. 
A plethora of different ML methods is obtained by different design choices for the data 
(representation), \gls{model} and \gls{loss}. ML methods also differ vastly in their practical 
implementations which might obscure their unifying basic principles. Two examples for such 
ML methods are deep learning and \gls{linreg}. 

Deep learning methods use cloud computing frameworks to train large models on large datasets. 
Operating on a much finer granularity for data and computation, \gls{linreg} can typically be implemented 
on small embedded systems. Nevertheless, \index{deep learning}deep learning methods and \gls{linreg} 
use the very same principle of iteratively updating a model based on the discrepancy between model 
predictions and actual observed data. This discrepancy is measured via a \gls{lossfunc}. 

We believe that thinking about ML as combinations of three components given by data, model and \gls{lossfunc} 
helps to navigate the steadily growing offer for ready-to-use ML methods. Our three-component picture 
allows a unified treatment of ML techniques, such as ``early stopping'', ``privacy-preserving'' ML and ``\gls{xml}'', 
that might seem unrelated at first sight. The regularization effect of early stopping in \gls{gdmethods} 
arises from the shrinking of the effective \gls{hypospace}. \index{privacy-preserving ML} 
Privacy-preserving ML methods can be obtained by particular choices for the features used to 
characterize \gls{datapoint}s (see Section \ref{equ_pp_feature_learning}). \index{Explainable ML} \Gls{xml} 
methods can be obtained by particular choices for the \gls{hypospace} and \gls{lossfunc} (see Chapter \ref{chap_explainable_ML}).

To make good use of ML tools it is instrumental to understand its underlying principles at the 
appropriate level of detail. It is typically not necessary to understand the mathematical details  
of advanced optimization methods to successfully apply deep learning methods. On a lower level, 
this tutorial helps ML engineers choose suitable methods for the application at hand. The book also 
offers a higher level view on the implementation of ML methods which is typically required to manage 
a team of ML engineers and data scientists. 

\section*{Acknowledgement}
This book grew from lecture notes prepared - and the student received - for the courses 
CS-E3210 ``Machine Learning: Basic Principles'', CS-E4800 ``Artificial Intelligence'', 
CS-EJ3211 ``Machine Learning with Python'', CS-EJ3311 ``Deep Learning with Python'' 
and CS-C3240 ``Machine Learning'' offered at Aalto University and within the Finnish University network 
\url{fitech.io}. The author is indebted to Shamsiiat Abdurakhmanova, Tomi Janhunen, Yu Tian, Natalia Vesselinova, 
Linli Zhang, Ekaterina Voskoboinik, Buse Atli, Stefan Mojsilovic for carefully reviewing early drafts of this book. 
Some of the figures have been generated with the help of Linli Zhang. The author is also grateful for 
the feedback received from Lasse Leskel{\"a}, Jukka Suomela, Homayun Afrabandpey, Linli Zhang, V{\"a}in{\"o} Mehtola, 
Anselmi Jokinen, Oleg Vlasovetc, Anni Niskanen, Dominik Hodan, Georgios Karakasidis, Yuvrajsinh Chudasama, 
Joni P{\"a}{\"a}kk{\"o}, Harri Wallenius, Nuutti Sten, ST John, Antti Ainamo and Satu Korhonen. 


%

\newpage 
\chapter*{Lists of Symbols}

\section*{Sets} 

\begin{align} 
&a \defeq b & \quad & \mbox{This statement defines $a$ to be shorthand for $b$. } \nonumber \\[4mm]
&|\mathcal{A}| & \quad & \mbox{The cardinality (number of elements) of a finite set $\mathcal{A}$.} \nonumber \\[4mm]
&\mathcal{A} \subseteq \mathcal{B}& \quad & \mbox{$\mathcal{A}$ is a subset of $\mathcal{B}$.} \nonumber \\[4mm]
&\mathcal{A} \subset \mathcal{B}& \quad & \mbox{$\mathcal{A}$ is a strict subset of $\mathcal{B}$.} \nonumber \\[4mm]
&\mathbb{N} & \quad & \mbox{The set of natural numbers $1,2,\ldots$.} \nonumber \\[4mm]
&\mathbb{R}  &\quad &\mbox{The set of real numbers $x$ \cite{RudinBook}.} \nonumber \\[4mm]
&\mathbb{R}_{+}  &\quad &\mbox{The set of non-negative real numbers $x\geq0$.} \nonumber \\[4mm]
&\{0,1\}& \quad & \mbox{The set consisting of two real-number $0$ and $1$.} \nonumber \\[4mm]
&[0,1] &\quad &\mbox{The closed interval of real numbers $x$ with $0 \leq x \leq 1$. } \nonumber \\[4mm]
&f(\cdot),h(\cdot)&\quad &\parbox{.8\textwidth}{A function or map $f(\cdot)$ that accepts any element $a \in \mathcal{A}$ 
	from a set $\mathcal{A}$ as input and delivers a well-defined element $f(a) \in \mathcal{B}$ of a set $\mathcal{B}$. 
	The set $\mathcal{A}$ is the domain of the function $f$ and the set $\mathcal{B}$ is the codomain of $f$. Machine Learning revolves around finding (or learning) a function $h$ (which we call hypothesis) that reads in the features $\featurevec$ of a \gls{datapoint} and delivers a prediction $h(\featurevec)$ for the label $\truelabel$ of the \gls{datapoint}.} \nonumber 
\end{align} 

\newpage
\section*{Matrices and Vectors} 

\begin{align} 
	&\mathbf{I}_{\modelidx \times \featuredim}  & \quad &  \parbox{.8\textwidth}{A generalized identity matrix 
		with $\modelidx$ rows and $\featuredim$ columns. The matrix has entries equal to $1$ along the main diagonal and 
		equal to $0$ otherwise. Some examples: $\mathbf{I}_{1 \times 2} = \big(1, 0\big)$ and $\mathbf{I}_{2 \times 1}= \begin{pmatrix} 1 \\ 0 \end{pmatrix}$.} \nonumber \\[4mm]
	&\mathbf{I} & \quad &  \parbox{.8\textwidth}{A square identity matrix whose shape should be clear from the context.} \nonumber \\[4mm]
	&\mathbb{R}^{\featuredim} & \quad &  \parbox{.8\textwidth}{The set of vectors that consist of $\featuredim$ real-valued entries.} \nonumber \\[4mm] 
	&\featurevec=\big(\feature_{1},\ldots,\feature_{\featuredim})^{T} &\quad & \parbox{.8\textwidth}{A vector of length $\featuredim$. 
		The $\featureidx$th entry of the vector is denoted $\feature_{\featureidx}$.} \nonumber \\[4mm]
	&\| \featurevec\|_{2}  &\quad & \parbox{.8\textwidth}{The Euclidean (or ``$\ell_{2}$'') norm of the vector 
		$\featurevec=\big(\feature_{1},\ldots,\feature_{\featurelen}\big)^{T}$ given as $ \| \featurevec \|_{2} \defeq \sqrt{\sum_{\featureidx=1}^{\featuredim} \feature_{\featureidx}^{2}}$.} \nonumber \\[4mm] 
	&\| \featurevec\|  & \quad &  \parbox{.8\textwidth}{Some norm of the vector $\featurevec$ \cite{golub96}. 
		Unless specified otherwise, we mean the Euclidean norm $\| \featurevec\|_{2}$.} \nonumber \\[4mm]
	&\featurevec^{T} &\quad & \parbox{.8\textwidth}{The transpose of a vector $\featurevec$ that is considered a 
		single column matrix. The transpose is a single-row matrix $\big(\feature_{1},\ldots,\feature_{\featurelen}\big)$.}  \nonumber \\[4mm]
	&\mathbf{X}^{T} &\quad & \parbox{.8\textwidth}{The transpose of a matrix $\mathbf{X} \in \mathbb{R}^{\samplesize \times \featurelen}$. 
		A square real-valued matrix $\mathbf{X} \in \mathbb{R}^{\samplesize \times \samplesize}$ 
		is called symmetric if $\mathbf{X} = \mathbf{X}^{T}$. }  \nonumber\\[4mm]
   &\mathbf{0}= \big(0,\ldots,0\big)^{T}  & \quad &  \parbox{.8\textwidth}{A vector of zero entries whose length should be clear from context.} \nonumber \\[4mm]
      &\big(\vv^{T},\vw^{T} \big)^{T}  & \quad &  \parbox{.8\textwidth}{The vector of length $\featurelen+\featurelen'$ 
      	obtained by concatenating the entries of vector $\vv \in \mathbb{R}^{\featurelen}$ with the entries of $\vw \in \mathbb{R}^{\featurelen'}$.} \nonumber \\[4mm]
    &	{\rm span}\{ \mathbf{B} \}  & \quad &  \parbox{.8\textwidth}{The span of a matrix $\mathbf{B} \in \mathbb{R}^{a \times b}$, 
    	which is the subspace of all linear combinations of columns of $\mathbf{B}$, 
  ${\rm span}\{ \mathbf{B} \} = \big\{  \mathbf{B} \va : \va \in \mathbb{R}^{b} \big\} \subseteq \mathbb{R}^{a}$.} \nonumber \\[4mm]
	&\mathbb{S}^{\featurelen}_{+} &\quad & \parbox{.8\textwidth}{The set of all \gls{psd} 
		matrices of size $\featurelen \times \featurelen$. }  \nonumber
\end{align} 

\newpage
\section*{Probability Theory} 
\begin{align}
\expect_{p} \{ f(\datapoint) \}  \quad\quad & \parbox{.8\textwidth}{The expectation of a function $f(\datapoint)$ of a \gls{rv} 
			$\datapoint$ whose probability distribution is $\prob{\datapoint}$. If the probability distribution is clear from context 
		we just write $\expect \{ f(\datapoint) \}$. }    \nonumber \\[4mm]             
\prob{\featurevec,\truelabel} \quad\quad & \parbox{.8\textwidth}{A (joint) probability distribution of a \gls{rv} 
	whose realizations are \gls{datapoint}s with features $\featurevec$ and label $\truelabel$.} \nonumber            \\[4mm]             
\prob{\featurevec|\truelabel} \quad\quad & \parbox{.8\textwidth}{A conditional probability distribution of a \gls{rv} 
	$\featurevec$ given the value of another \gls{rv} $\truelabel$ \cite[Sec.\ 3.5]{BertsekasProb}. } \nonumber        \\[4mm]             
\prob{\featurevec;\weights} \quad\quad & \parbox{.8\textwidth}{A parametrized probability distribution of a \gls{rv} $\featurevec$. 
   The probability distribution depends on a parameter vector $\weights$. For example, $\prob{\featurevec;\weights}$ could be a 
   multivariate	normal distribution of a Gaussian \gls{rv} $\featurevec$ with the parameter vector $\weights$ being the expectation $\expect \{ \featurevec \}$. } \nonumber              \\[4mm]
\mathcal{N}(\mu, \sigma^{2}) \quad\quad & \parbox{.8\textwidth}{The probability distribution of a scalar normal 
	(``Gaussian'') \gls{rv} $\feature \in \mathbb{R}$ with mean (or expectation) $\mu= \expect \{ \feature \}$ 
	and variance $\sigma^{2} =   \expect \big\{  (  \feature - \mu )^2 \big\}$.} \nonumber       \\[4mm]
\mathcal{N}(\clustermean, \mathbf{C}) \quad\quad & \parbox{.8\textwidth}{The probability distribution of a multivariate (vector-valued) 
	normal (``Gaussian'') \gls{rv} $\featurevec \in \mathbb{R}^{\featuredim}$ with mean (or expectation) $\clustermean= \expect \{ \featurevec \}$ 
	and covariance matrix $\mathbf{C} =  \expect \big\{ \big( \featurevec - \clustermean \big)\big( \featurevec - \clustermean \big)^{T} \big\}$.} \nonumber                                             
\end{align}

\newpage
\section*{Machine Learning}

\begin{align}
\datapoint \quad\quad & \parbox{.8\textwidth}{A \gls{datapoint} which is characterized by several properties that we 
	divide into low-level properties (features) and high-level properties (labels) (see Chapter \ref{ch_Elements_ML}).}    \nonumber   \\[4mm] 
\sampleidx \quad\quad & \parbox{.8\textwidth}{An index $\sampleidx=1,2,\ldots,$ that 
	enumerates the \gls{datapoint}s within a \gls{dataset}.}    \nonumber   \\[4mm] 
\samplesize \quad\quad &\parbox{.8\textwidth}{The number of \gls{datapoint}s in (the size of) a \gls{dataset}.} \nonumber \\[4mm] 
\dataset \quad\quad & \parbox{.8\textwidth}{A dataset $\dataset = \{ \datapoint^{(1)},\ldots, \datapoint^{(\samplesize)} \}$ 
	is a list of individual \gls{datapoint}s $\datapoint^{(\sampleidx)}$, for $\sampleidx=1,\ldots,\samplesize$.}    \nonumber   \\[4mm] 
\featurelen \quad\quad &\parbox{.8\textwidth}{The number of individual \gls{features} used to characterize a \gls{datapoint}.} \nonumber \\[4mm] 
\feature_{\featureidx} \quad\quad &\parbox{.8\textwidth}{The $\featureidx$-th feature of a \gls{datapoint}. The first feature of 
	a given \gls{datapoint} is denoted $\feature_{1}$, the second feature $\feature_{2}$ and so on. } \nonumber \\[4mm] 
\featurevec \quad\quad &\parbox{.8\textwidth}{The feature vector $\featurevec=\big(\feature_{1},\ldots,\feature_{\featuredim}\big)^{T}$ of a \gls{datapoint} whose entries are the individual features of the \gls{datapoint}. With a slight abuse of language, we refer to the feature 
	vector $\featurevec$ also as ``the features'' of a \gls{datapoint}.} \nonumber \\[4mm] 
\rawfeaturevec \quad\quad &\parbox{.8\textwidth}{Beside the symbol $\featurevec$, we sometimes use $\rawfeaturevec$ as another 
	symbol to denote a vector whose entries are features of a \gls{datapoint}. We need two different symbols to distinguish between 
	``raw'' and learnt feature vectors (see Chapter \ref{ch_FeatureLearning}).} \nonumber  \\[4mm] 
\featurevec^{(\sampleidx)} \quad\quad &\parbox{.8\textwidth}{The feature vector of the $\sampleidx$th \gls{datapoint} within a \gls{dataset}. } \nonumber \\[4mm] 
\feature_{\featureidx}^{(\sampleidx)}\quad\quad &\parbox{.8\textwidth}{The $\featureidx$th feature of the $\sampleidx$th 
	\gls{datapoint} within a \gls{dataset}.} \nonumber \\[4mm] 
\batch \quad\quad &\parbox{.8\textwidth}{A mini-batch that consists of a (randomly selected) subset of \gls{datapoint}s from a (typically very large) \gls{dataset} (see Section \ref{sec_sgd}).} \nonumber
\\[4mm] 
\batchsize \quad\quad &\parbox{.8\textwidth}{The size of (the number of \gls{datapoint}s in) a mini-batch using during a \gls{stochGD} step (see Section \ref{sec_sgd}).} \nonumber  
\end{align}

\begin{align}
		\truelabel \quad\quad &\parbox{.8\textwidth}{The label (quantity of interest) of a \gls{datapoint}.} \nonumber   \\[4mm] 
	\truelabel^{(\sampleidx)} \quad\quad &\parbox{.8\textwidth}{The label of the $\sampleidx$th \gls{datapoint}.} \nonumber  \\[4mm] 
	\big(\featurevec^{(\sampleidx)},\truelabel^{(\sampleidx)}\big)  \quad\quad &\parbox{.8\textwidth}{The features and the label of the $\sampleidx$th \gls{datapoint} within a \gls{dataset}.} \nonumber  \\[4mm] 
	\featurespace  \quad\quad & \parbox{.8\textwidth}{The feature space of a ML method consists of all potential 
		feature values that a \gls{datapoint} can have. However, we typically use feature spaces that are much larger than the set of different 
		feature values arising in finite datasets. The majority of the methods discussed in this book uses the feature space 
		$\featurespace=\mathbb{R}^{\featuredim}$ consisting of all Euclidean vectors of length $\featuredim$. }    \nonumber   \\[4mm] 
	\labelspace  \quad\quad & \parbox{.8\textwidth}{The label space $\labelspace$ of a ML method 
		consists of all potential label values that a \gls{datapoint} can have. We often use label spaces that 
		are larger than the set of different label values arising in a give dataset (e.g., a \gls{trainset}). 
		We refer to a ML problems (methods) using a numeric label space, such as $\labelspace=\mathbb{R}$ 
		or $\labelspace=\mathbb{R}^{3}$, as regression problems (methods). ML problems (methods) 
		that use a discrete label space, such as $\labelspace=\{0,1\}$ or $\labelspace=\{\mbox{``cat''},\mbox{``dog''},\mbox{``mouse''}\}$ 
		are referred to as classification problems (methods).}    \nonumber  \\[4mm] 
	\lrate  \quad\quad & \parbox{.8\textwidth}{A \gls{learnrate} or step-size parameter used by \gls{gdmethods}.}    \nonumber       \\[4mm] 
		h(\cdot)  \quad\quad &\parbox{.8\textwidth}{A hypothesis map that reads in features $\featurevec$ of a \gls{datapoint} and delivers a prediction $\hat{\truelabel}=h(\featurevec)$ for its label $\truelabel$.} \nonumber  	  \\[4mm] 
		\hypospace  \quad\quad & \parbox{.8\textwidth}{A hypothesis space or model used by a ML method. 
			The hypothesis space consists of different hypothesis maps $h: \featurespace \rightarrow \labelspace$ between which 
			the ML method has to choose.}   \nonumber
\end{align}

\begin{align}
	\biasterm^2 \quad\quad &\parbox{.8\textwidth}{The squared \gls{bias} of a hypothesis $\hat{h}$ delivered by a ML algorithm that is fed 
		with \gls{datapoint}s which are modelled as realizations of \gls{rv}s. Since the data is modelled as realizations of \gls{rv}s, also the 
		delivered hypothesis $\hat{h}$ is the realization of a \gls{rv}.} \nonumber  \\[4mm] 
	\varianceterm \quad\quad &\parbox{.8\textwidth}{The \gls{variance} of the (parameters of the) hypothesis 
		delivered by a ML algorithm. If the input data for this algorithm is interpreted as realizations of \gls{rv}s, so 
		is the delivered hypothesis a realization of a \gls{rv}.} \nonumber \\[4mm] 
	\loss{(\featurevec,\truelabel)}{h}  \quad\quad & \parbox{.8\textwidth}{The \gls{loss} incurred by predicting the 
		label $\truelabel$ of a \gls{datapoint} using the predicted label $\hat{\truelabel}=h(\featurevec)$. The 
		predicted label $\hat{\truelabel}$ is obtained from evaluating the hypothesis $h \in \hypospace$ for 
		the feature vector $\featurevec$ of the \gls{datapoint}.}    \nonumber   \\[4mm] 
	\valerror \quad\quad &\parbox{.8\textwidth}{The \gls{valerr} of a hypothesis $h$, which is its average loss incurred 
		over a \gls{valset}.} \nonumber \\[4mm] 
	\emperror\big(h| \dataset \big) \quad\quad &\parbox{.8\textwidth}{The \gls{emprisk} or average \gls{loss} incurred by the predictions of 
		hypothesis $h$ for the \gls{datapoint}s in the dataset $\dataset$.} \nonumber 
	\\[4mm] 
	\trainerror \quad\quad &\parbox{.8\textwidth}{The \gls{trainerr} of a hypothesis $h$, which is its average loss incurred 
		over a \gls{trainset}. } \nonumber 	\\[4mm] 
	\timeidx \quad\quad &\parbox{.8\textwidth}{A discrete-time index $\timeidx=0,1,\ldots$ used to enumerate a 
		sequence to sequential events (``time instants''). } \nonumber \\[4mm] 
	\taskidx \quad\quad &\parbox{.8\textwidth}{A generic index used to enumerate a finite set of 
		learning tasks within a multi-task learning problem (see Section \ref{sec_mtl_regularization}).} \nonumber  \\[4mm] 
	\regparam \quad\quad &\parbox{.8\textwidth}{A regularization parameter that scales 
		the regularization term in the objective function of \gls{srm}.} \nonumber \\[4mm] 
	\eigval{\featureidx}\big( \mathbf{Q} \big) \quad\quad &\parbox{.8\textwidth}{The $\featureidx$th 
		eigenvalue (sorted either ascending or descending) of a \gls{psd} matrix $\mathbf{Q}$. We also 
		use the shorthand $\eigval{\featureidx}$ if the corresponding matrix is clear from context. } \nonumber \\[4mm] 
	\actfun(\cdot) \quad\quad &\parbox{.8\textwidth}{The \gls{actfun} used by an artificial neuron within an \gls{ann}.} \nonumber                                             
\end{align}     
                                                                                                            
\begin{align}
		\decreg{\hat{\truelabel}} \quad\quad &\parbox{.8\textwidth}{A hypothesis $h$ partitions the feature space into decision regions.
		A specific decision region 
		$ \decreg{\hat{\truelabel}} $ consists of all feature vectors $\featurevec$ that are mapped to the same predicted label $h(\featurevec)=\hat{\truelabel}$ for all $\featurevec \in \decreg{\hat{\truelabel}} $. } \nonumber   \\[4mm]     
	\weights  \quad\quad & \parbox{.8\textwidth}{A parameter vector $\weights = \big(\weight_{1},\ldots,\weight_{\featuredim}\big)^{T}$ whose entries 
		are parameters of a model. These parameters could be feature weights in linear maps, the weights in \gls{ann}s or the thresholds 
		used for splits in \gls{decisiontree}s.}    \nonumber \nonumber  \\[4mm]  
	h^{(\weights)}(\cdot)  \quad\quad &\parbox{.8\textwidth}{A hypothesis map that involves tunable parameters $\weight_{1},\ldots,\weight_{\featuredim}$ 
		which are stacked into the vector $\weights=\big(\weight_{1},\ldots,\weight_{\featuredim} \big)^{T}$.} \nonumber    \\[4mm]    
	\nabla f(\weights)  \quad\quad & \parbox{.8\textwidth}{The \gls{gradient} of a differentiable real-valued function 
		$f: \mathbb{R}^{\featuredim}\rightarrow \mathbb{R}$ is the vector 
		$\nabla f(\weights) = \big( \frac{\partial f}{\partial \weight_{1}},\ldots,\frac{\partial f}{\partial \weight_{\featuredim}}  \big)^{T} \in \mathbb{R}^{\featuredim}$ \cite[Ch. 9]{RudinBookPrinciplesMatheAnalysis}.}   \nonumber \\[4mm] 
\featuremap(\cdot)  \quad\quad & \parbox{.8\textwidth}{A feature map that reads in features $\featurevec \in \featurespace$ of a \gls{datapoint} and delivers 
	new (transformed) features $\widehat{\featurevec} = \featuremap\big( \featurevec \big) \in \featurespace'$.}    \nonumber                                                                                                                                           
\end{align}

\tableofcontents                        


\chapter{Introduction}
\begin{figure}[htbp]
	\centering
	\includegraphics[width=8cm]{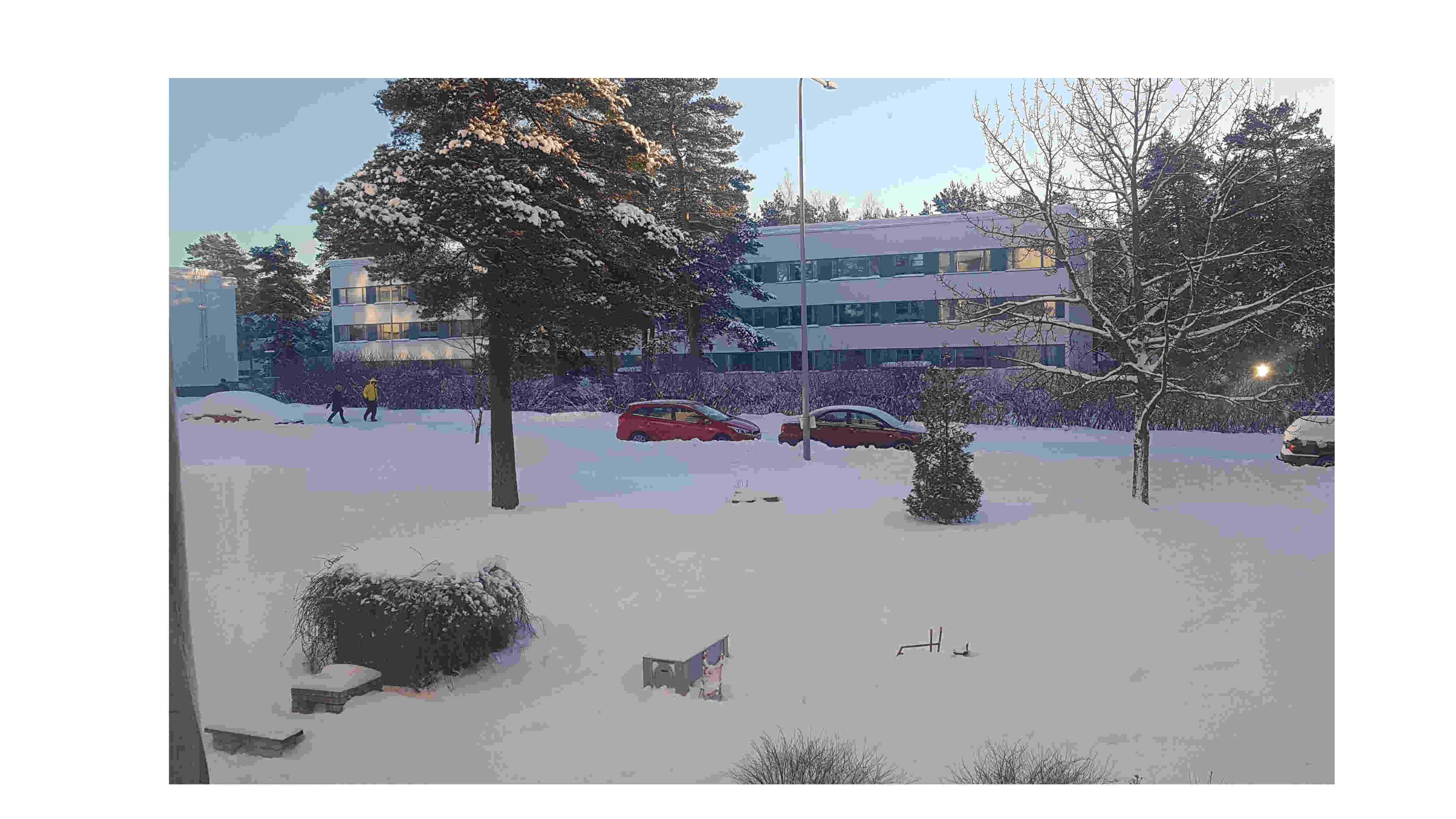}
	\caption{Looking outside the window during the morning of a winter day in Finland.}
	\label{fig:skiday}
\end{figure}

Consider waking up one winter morning in Finland and looking outside the window (see Figure \ref{fig:skiday}). 
It seems to become a nice sunny day which is ideal for a ski trip. To choose the right gear (clothing, wax) it 
is vital to have some idea for the maximum daytime temperature which is typically reached around early 
afternoon. If we expect a maximum daytime temperature of around plus $5$ degrees, we might not put 
on the extra warm jacket but rather take only some extra shirt for change.

We can use ML to learn a predictor for the maximum daytime temperature for the 
specific day depicted in Figure \ref{fig:skiday}. The prediction shall be based solely on the minimum 
temperature observed in the morning of that day. ML methods can learn a predictor in a data-driven 
fashion using historic weather observations provided by the \gls{fmi}. We can download the recordings 
of minimum and maximum daytime temperature for the most recent days and denote the resulting dataset by 
\begin{equation}
	\label{equ_FMI_data}
	\dataset = \big\{ \datapoint^{(1)},\ldots,\datapoint^{(\samplesize)} \big\}.  
\end{equation} 
Each \gls{datapoint} $\datapoint^{(\sampleidx)} = \big(\feature^{(\sampleidx)},\truelabel^{(\sampleidx)}\big)$, 
for $\sampleidx=1,\ldots,\samplesize$, represents some previous day for which the minimum and maximum 
daytime temperature $\feature^{(\sampleidx)}$ and $\truelabel^{(\sampleidx)}$ has been recorded.  We depict 
the data \eqref{equ_FMI_data} in Figure \ref{fig_scatterplot_temp_FMI}. Each dot in Figure \ref{fig_scatterplot_temp_FMI} 
represents a specific day with minimum temperature $\feature$ and maximum temperature $\truelabel$. 

\begin{figure}[htbp]
	\begin{center}
		\begin{tikzpicture}
			\tikzset{x=0.35cm,y=2cm,every path/.style={>=latex},node style/.style={circle,draw}}
			%
			\begin{axis}[axis x line=none,
				axis y line=none,
				ylabel near ticks,
				xlabel near ticks,
				enlarge y limits=true,
				xmin=-40, xmax=40,
				ymin=-40, ymax=40,
				width=10cm, height=10cm, ]
				\addplot[only marks] table [x=mintmp, y=maxtmp, col sep = comma] {FMIData.csv};
				\node at (axis cs:32,1) [anchor=west] {$\feature$};
				\node at (axis cs:0,30) [anchor=west] {$\truelabel$};
				\draw[->] (axis cs:-30,0) -- (axis cs:32,0);
				\draw[->] (axis cs:0,-30) -- (axis cs:0,30);
			\end{axis}
		\end{tikzpicture}
		\vspace*{-14mm}
	\end{center}
	\caption{Each dot represents a specific day that is characterized by its minimum daytime temperature $\feature$ 
		as feature and its maximum daytime temperature $\truelabel$ as label. These temperatures are measured 
		at some \gls{fmi} weather station.}
	\label{fig_scatterplot_temp_FMI}
	\vspace*{-3mm}
\end{figure}

ML methods learn a hypothesis $h(\feature)$, that reads in the minimum temperature $\feature$ and delivers  
a prediction (forecast or approximation) $\hat{\truelabel} = h(\feature)$ for the maximum daytime temperature $\truelabel$. 
Every practical ML method uses a particular \gls{hypospace} out of which the hypothesis $h$ is chosen. This 
\gls{hypospace} of candidates for the hypothesis map is an important design choice and might be based 
on domain knowledge. 

In what follows, we illustrate how to use domain knowledge to motivate a choice for the \gls{hypospace}. 
Let us assume that the minimum and maximum daytime temperature of an arbitrary day are approximately 
related via 
\begin{equation}
	\label{equ_initial_hypo_FMI}
	\truelabel \approx \weight_{1} \feature + \weight_{0} \mbox{ with some \gls{weights} } \weight_{1} \in \mathbb{R}_{+}, \weight_{0} \in \mathbb{R}. 
\end{equation} 
The assumption \eqref{equ_initial_hypo_FMI} reflects the intuition (domain knowledge) that the maximum daytime 
temperature $\truelabel$ should be higher for days with a higher minimum daytime temperature $\feature$. The 
assumption \eqref{equ_initial_hypo_FMI} contains two \index{weights}\gls{weights} $\weight_{1}$ and $\weight_{0}$. 
These \gls{weights} are tuning \index{parameter}\gls{parameters} that allow for some flexibility in our assumption. We 
require the weight $\weight_{1}$ to be non-negative but otherwise leave these \gls{weights} unspecified for the time being. 
The main subject of this book are ML methods that can be used to learn suitable values for the \gls{weights} $\weight_{1}$ 
and $\weight_{0}$ in a data-driven fashion. 

Before we detail how ML can be used to find or learn good values for the weights $\weight_{0}$ in $\weight_{1}$ 
in \eqref{equ_initial_hypo_FMI} let us interpret them. The weight $\weight_{1}$ in \eqref{equ_initial_hypo_FMI} can be 
interpreted as the relative increase in the maximum daytime temperature for an increased minimum daytime 
temperature. Consider an earlier  day with recorded maximum daytime temperature of $10$ degrees and minimum 
daytime temperature of $0$ degrees. The assumption \eqref{equ_initial_hypo_FMI} then means that the maximum 
daytime temperature for another day with minimum temperature of $+1$ degrees would be $10 + \weight_{1}$ 
degrees. The second weight $\weight_{0}$ in our assumption \eqref{equ_initial_hypo_FMI} can be interpreted as the 
maximum daytime temperature that we anticipate for a day with minimum daytime temperature equal to $0$. 

Given the assumption \eqref{equ_initial_hypo_FMI}, it seems reasonable to restrict the ML method to only 
consider linear maps
\begin{equation} 
	\label{equ_def_linear_map_scalar}
	h(\feature) \defeq \weight_{1} \feature + \weight_{0} \mbox{ with some weights } w_{1} \in \mathbb{R}_{+},w_{0} \in \mathbb{R}. 
\end{equation}
Since we require $\weight_{1}\!\geq\!0$, the map \eqref{equ_def_linear_map_scalar} is monotonically increasing with respect 
to the argument $\feature$. Therefore, the prediction $h(\feature)$ for the maximum daytime temperature becomes higher 
with higher minimum daytime temperature $\feature$.    

The expression \eqref{equ_def_linear_map_scalar} defines a whole ensemble of hypothesis maps. 
Each individual map corresponding to a particular choice for $\weight_{1} \geq 0$ and $\weight_{0}$. 
We refer to such an ensemble of potential predictor maps as the \gls{model} or \gls{hypospace} 
that is used by a ML method. 

We say that the map \eqref{equ_def_linear_map_scalar} is parametrized by the vector $\weights= \big(\weight_{1},\weight_{0}\big)^{T}$ 
and indicate this by writing $h^{(\weights)}$. For a given parameter vector $\weights= \big(\weight_{1},\weight_{0}\big)^{T}$, we 
obtain the map 
$h^{(\weights)}(\feature) = \weight_{1} \feature +\weight_{0}$. 
Figure \ref{fig_three_maps_example} depicts three maps $h^{(\weights)}$ 
obtained for three different choices for the parameters $\weights$.  
\begin{figure}[htbp]
	\begin{center}
		\begin{tikzpicture}
			\begin{axis}
				[ylabel=$h(\feature)$,
				xscale=1,
				xlabel=$\feature$, 
				axis x line=center,
				axis y line=center,
				xtick={-3,-2,-1,0,1,2,3},
				ytick={-3,-2,-1,0,1,2,3},
				xlabel={feature $x$},
				ylabel={$h^{(\weights)}(\feature)$},
				xlabel style={right},
				ylabel style={above},
				xmin=-5,
				xmax=5,
				ymin=-5,
				ymax=5
				]
				\addplot[green, ultra thick] (x,4*x);
				\addplot[red, ultra thick] (x,1*x+2);	
				\addplot[blue, ultra thick] (x,2*x+3);	
			\end{axis}
		\end{tikzpicture}
		\vspace*{-4mm}
	\end{center}
	\caption{Three hypothesis maps of the form \eqref{equ_def_linear_map_scalar}.}
	\label{fig_three_maps_example}
\end{figure}

ML would be trivial if there is only one single hypothesis. Having only a single hypothesis means that there is no need to try out 
different hypotheses to find the best one. To enable ML, we need to choose between a whole space of different hypotheses. ML 
methods are computationally efficient methods to choose (learn) a good hypothesis out of (typically very large) \gls{hypospace}s. 
The \gls{hypospace} constituted by the maps \eqref{equ_def_linear_map_scalar} for different weights is uncountably infinite. 

To find, or {\bf learn}, a good hypothesis out of the infinite set \eqref{equ_def_linear_map_scalar}, 
we need to somehow assess the quality of a particular hypothesis map. ML methods use 
a \gls{lossfunc} for this purpose. A \gls{lossfunc} is used to quantify the 
difference between the actual data and the predictions obtained from a hypothesis map 
(see Figure \ref{fig_scatterplot_temp_FMI_linalg}). One widely-used example of a loss 
function is the squared error loss $(\truelabel-h(\feature))^2$. Using this \gls{lossfunc}, ML methods learn 
a hypothesis map out of the model \eqref{equ_def_linear_map_scalar} by tuning $\weight_{1},\weight_{0}$ to minimize 
the average loss $$(1/\samplesize) \sum_{\sampleidx=1}^{\samplesize} \big( \truelabel^{(\sampleidx)} - h\big(\feature^{(\sampleidx)} \big) \big)^{2}.$$

\begin{figure}[htbp]
	\begin{center}
		\begin{tikzpicture}
			\tikzset{x=0.35cm,y=2cm,every path/.style={>=latex},node style/.style={circle,draw}}
			%
			\begin{axis}[axis x line=none,
				axis y line=none,
				ylabel near ticks,
				xlabel near ticks,
				enlarge y limits=true,
				xmin=-40, xmax=40,
				ymin=-40, ymax=40,
				width=10cm, height=10cm, ]
				\addplot[only marks] table [x=mintmp, y=maxtmp, col sep = comma] {FMIData.csv};
				\node at (axis cs:32,1) [anchor=west] {$\feature$};
				\node at (axis cs:0,30) [anchor=west] {$\truelabel$};
				\draw[->] (axis cs:-30,0) -- (axis cs:32,0);
				\draw[->] (axis cs:0,-30) -- (axis cs:0,30);
				\draw[red,line width=3] (axis cs:-30,-10) -- (axis cs:20,25);
			\end{axis}
		\end{tikzpicture}
		\vspace*{-14mm}
	\end{center}
	\caption{Each dot represents a specific days that is characterized by its minimum daytime temperature $\feature$ 
		and its maximum daytime temperature $\truelabel$. We also depict a straight line representing a linear predictor 
		map. A main principle of ML methods is to learn a predictor (or hypothesis) map with minimum discrepancy 
		between predictor map and \gls{datapoint}s. Different ML methods use different types of predictor maps (\gls{hypospace}) and \gls{lossfunc}s to 
		quantify the discrepancy between hypothesis and actual \gls{datapoint}s. 
	}
	\label{fig_scatterplot_temp_FMI_linalg}
	\vspace*{-3mm}
\end{figure}


The above weather prediction is prototypical for many other ML applications. Figure \ref{fig_AlexMLBP} illustrates 
the typical workflow of a ML method. Starting from some initial guess, ML methods repeatedly improve their 
current hypothesis based on (new) observed data. 

Using the current hypothesis, ML methods make predictions or forecasts about future observations. The discrepancy 
between the predictions and the actual observations, as measured using some \gls{lossfunc}, is used to improve the 
hypothesis. Learning happens during improving the current hypothesis based on the discrepancy between its 
predictions and the actual observations. 

ML methods must start with some initial guess or choice for a good hypothesis. This initial guess can be based 
on some prior knowledge or domain expertise \cite{MitchellBias1980}. While the initial guess for a hypothesis might 
not be made explicit in some ML methods, each method must use such an initial guess. In our weather prediction 
application discussed above, we used the linear model \eqref{equ_initial_hypo_FMI} as the initial hypothesis. 

\section{Relation to Other Fields}

ML builds on concepts from several other scientific fields. Conversely, ML provides important 
tools for many other scientific fields. 

\subsection{Linear Algebra} 

Modern ML methods are computationally efficient methods to fit high-dimensional models to 
large amounts of data. The models underlying state-of-the-art ML methods can contain billions of 
tunable or learnable parameters. To make ML methods computationally efficient we need to use 
suitable representations for data and models. 

Maybe the most widely used mathematical structure to represent data is the \gls{euclidspace}
$\mathbb{R}^{\featuredim}$ with some dimension $\featuredim \in \naturalnumbers$ \cite{RudinBookPrinciplesMatheAnalysis}. 
The rich algebraic and geometric structure of $\mathbb{R}^{\featuredim}$ allows us to design 
of ML algorithms that can process vast amounts of data to quickly update a model (parameters). 
Figure \ref{fig_euclideanspace_dim2} depicts the \gls{euclidspace} $\mathbb{R}^{\featuredim}$ 
for $\featuredim=2$, which is used to construct \gls{scatterplot}s.  

\begin{figure}[htbp]
	\begin{center}
		\begin{tikzpicture}
			\begin{axis}
				[ylabel=$h(\feature)$,
				xscale=1,
				xlabel=$\feature$, 
				axis x line=center,
				axis y line=center,
				xtick={-3,-2,-1,0,1,2,3},
				ytick={-3,-2,-1,0,1,2,3},
				xlabel={$z_{1}$},
				ylabel={$z_{2}$},
				xlabel style={right},
				ylabel style={above},
				xmin=-5,
				xmax=5,
				ymin=-5,
				ymax=5
				]
			\end{axis}
			\filldraw[black] (4,4) circle (2pt) node[anchor=west] {$\datapoint= \big( z_{1},z_{2} \big)^{T}$};
			\filldraw[black] (6,2) circle (2pt) node[anchor=west] {$\datapoint'= \big( z'_{1},z'_{2} \big)^{T}$};
		\end{tikzpicture}
		\vspace*{-4mm}
	\end{center}
	\caption{The \gls{euclidspace} $\mathbb{R}^{2}$ is constituted by all vectors (or points) $\datapoint = \big(z_{1},z_{2}\big)^{T}$ (with $z_{1},z_{2} \in \mathbb{R}$) together with the \index{inner product} inner product $\datapoint^{T} \datapoint' = z_{1}z'_{1}+z_{2}z'_{2}$.}
	\label{fig_euclideanspace_dim2}
\end{figure}

The \index{scatterplot}\gls{scatterplot} in Figure \ref{fig_scatterplot_temp_FMI} depicts \gls{datapoint}s (representing individual days) 
as vectors in the \gls{euclidspace} $\mathbb{R}^{2}$. For a given \gls{datapoint}, we obtain its associated 
vector $\datapoint=(\feature,\truelabel)^{T}$ in $\mathbb{R}^{2}$ by stacking the minimum daytime temperature 
$\feature$ and the maximum daytime temperature $\truelabel$ into the vector $\datapoint$ of length two. 

We can use the \gls{euclidspace} $\mathbb{R}^{\featurelen}$ not only to represent \gls{datapoint}s 
but also to represent models for these \gls{datapoint}s. One such class of models is obtained by linear 
maps on $\mathbb{R}^{\featurelen}$. Figure \ref{fig_three_maps_example} depicts some examples for 
such linear maps. We can then use the geometric structure of $\mathbb{R}^{\featurelen}$, defined by the 
Euclidean norm, to search for the best model. As an example, we could search for the linear model, represented by a 
straight line, such that the average (Euclidean) distance to the \gls{datapoint}s in Figure \ref{fig_scatterplot_temp_FMI} 
is as small as possible (see Figure \ref{fig_scatterplot_temp_FMI_linalg}). The properties of linear 
structures are studied within linear algebra \cite{StrangLinAlg2016}. Some important ML methods, 
such as \gls{linclass} (see Section \ref{sec_lin_reg}) or \gls{pca} (see Section \ref{sec_pca}) 
are direct applications of methods from linear algebra. 

\subsection{Optimization} 

A main design principle for ML methods is the formulation of ML problems as \index{optimization}optimization 
problems \cite{OptMLBook}. The weather prediction problem above can be formulated as the problem of optimizing (minimizing) the 
prediction error for the maximum daytime temperature. Many ML methods are obtained by straightforward applications of 
optimization methods to the optimization problem arising from a ML problem (or application). 

The statistical and computational properties of such ML methods can be studied using tools from the theory of 
optimization. What sets the optimization problems in ML apart from ``plain vanilla'' optimization problems (see Figure \ref{fig_optml}-(a)) 
is that we rarely have perfect access to the objective function to be minimized. ML methods learn a hypothesis by 
minimizing a noisy (or even incomplete) version (see Figure \ref{fig_optml}-(b)) of the ultimate objective function. 
The ultimate objective function for ML methods is often defined using an expectation over an unknown \gls{probdist} 
for \gls{datapoint}s. Chapter \ref{ch_Optimization} discusses methods that are based on estimating the objective 
function by empirical averages that are computed over a set of \gls{datapoint}s (which serve as a \gls{trainset}). 
\vspace*{-5mm}
\begin{figure}
	\begin{center}
	
	\tikzset{global scale/.style={
			scale=#1,
			every node/.append style={scale=#1}
		}
	}
	\begin{tikzpicture}[global scale = 1]             
		
		\draw[->, very thick, xshift=-4cm](-3,0)--(3,0) node[above, xshift=0.2cm, yshift=0.2cm, font=\fontsize{12}{0}\selectfont, align=center] {optimization  \\ variable};   
		\draw[->, very thick, xshift=-4cm](0,-0.5)--(0,3) node[above, font=\fontsize{12}{0}\selectfont] {objective}  node[below, yshift=-4cm, font=\fontsize{12}{0}\selectfont] {(a)};   
		\draw[blue, line width=2pt, xshift=-4 cm] (-2.3,2.8) .. controls (1.5,0.3) .. (3,2.5);
		
		\draw[->, very thick, xshift=4cm](-3,0)--(3,0) node[above, xshift=0.2cm, yshift=0.2cm, font=\fontsize{12}{0}\selectfont] {hypothesis};
		\draw[->, very thick, xshift=4cm](0,-0.5)--(0,3) node[above, font=\fontsize{12}{0}\selectfont] {loss}  node[below, yshift=-4cm, font=\fontsize{12}{0}\selectfont] {(b)};
		\draw[dash pattern=on 2pt off 6pt on 10pt off 6pt, line width=2pt, xshift=4 cm] plot [smooth, tension=0.2]
		coordinates {(-2.5,3) (-2,2) (-1.6,2.4) (-1.2,1.7) (-0.6,2) (-0.4,1.4) (-0.2,1.7) (-0.1,1.5) (0.2,1.4) (0.6,1.3) (0.8,1) (1.3,1) (1.4,0.6) (1.7,0.9)};
			\draw[blue, line width=1pt, xshift=4 cm] (-2.3,2.8) .. controls (1.5,0.3) .. (3,2.5);
	\end{tikzpicture}
		\vspace*{-4mm}
	\end{center}
	\caption{(a) A simple optimization problem consists of finding the values of an optimization 
		variable that results in the minimum objective value. (b) ML methods learn (find) a hypothesis 
		by minimizing a (average) \gls{loss}. This average \gls{loss} is a noisy version of the ultimate  
		objective. This ultimate objective function is often defined as an expectation whose underlying 
		probability distribution is unknown (see Section \ref{sec_empirical_risk}).}
	\label{fig_optml}
\end{figure} 

\subsection{Theoretical Computer Science} 

Practical ML methods form a specific subclass of computing systems. Indeed, ML 
methods apply a sequence of computational operations to input data. The result of these computational 
operations are the predictions delivered to the user of the ML method. The interpretation 
of ML as computational systems allows to use tools from theoretical computer science to 
study the feasibility and intrinsic difficulty of ML problems. Even if a ML problem can be solved 
in theoretical sense, every practical ML method must fit the available computational infrastructure 
\cite{PittValiant1988,Valiant1984}. 

The available computational resources, such as processor time, memory and communication bandwidth, 
can vary significantly between different infrastructures. One example for such a computational infrastructure 
is a single desktop computer. Another example for a computational infrastructure is a cloud computing service 
which distributes data and computation over large networks of physical computers \cite{Millard2021}. 

The focus of this book is on ML methods that can be understood as numerical optimization algorithms (see 
Chapter \ref{ch_Optimization} and \ref{ch_GD}). Most of these ML methods amount to (a large number of) 
matrix operations such as matrix multiplication or matrix inversion \cite{golub96}. Numerical linear algebra 
provides a vast algorithmic toolbox for the design of such ML methods \cite{StrangLinAlg2016,Strang2007}. 
The recent success of ML methods in several application domains might be attributed to their efficient use of 
matrices to represent data and models. Using this representation allows us to implement the resulting ML 
methods using highly efficient hard- and software implementations for numerical linear algebra \cite{Goodfellow-et-al-2016}.

\subsection{Information Theory}

\begin{figure}
	\begin{center}
	\tikzset{global scale/.style={
			scale=#1,
			every node/.append style={scale=#1}
		}
	}
	\begin{tikzpicture}[global scale = 0.85,           
		squarednode/.style={rectangle, draw=black!60, line width=2pt, minimum size=15mm, font=\fontsize{12}{12}\selectfont, align=center}]      
		
		\node[squarednode] (information) {information \\ source};         
		\node[squarednode] (transmitter) [right=of information] {transmitter};        
		\node[squarednode] (noisy) [right=of transmitter] {noisy \\ channel};         
		\node[squarednode] (receiver) [right=of noisy] {receiver};
		\node[squarednode, yshift=-2cm] (MLb) [below=of transmitter] {ML};
		\node[squarednode] (channelb) [left=of MLb] {channel};
		\node[squarednode, xshift=-0.5cm] (data) [left=of channelb] {data \\ source};
		\node[below=of channelb, yshift=-0.5cm, font=\fontsize{12}{12}\selectfont] {(b)};
		\node[squarednode, yshift=-2cm] (?) [below=of noisy] {?};
		\node[squarednode] (channelc) [right=of ?] {channel};
		\node[squarednode] (MLc) [right=of channelc] {ML};
		\node[below=of channelc, yshift=-0.5cm, font=\fontsize{12}{12}\selectfont] {(c)};
		
		\draw[->, line width=2pt] (information.east) -- (transmitter.west) node[above, xshift = -0.5cm, yshift=0.5cm, font=\fontsize{12}{12}\selectfont] {$m$};          
		\draw[->, line width=2pt] (transmitter.east) -- (noisy.west) node[below, yshift=-2cm, font=\fontsize{12}{12}\selectfont] {(a)};
		\draw[->, line width=2pt] (noisy.east) -- (receiver.west);
		\draw[->, line width=2pt] (receiver.east) -- +(1,0) node[above, xshift = -0.5cm, yshift=0.5cm, font=\fontsize{12}{12}\selectfont] {$\hat{m}$};
		
		\draw[->, line width=2pt] (data.east) -- (channelb.west) node[above, xshift = -0.8cm, yshift=0.5cm, font=\fontsize{12}{12}\selectfont] {$(\mathbf{X},\mathbf{y})$};
		\draw[->, line width=2pt] (channelb.east) -- (MLb.west) node[above, xshift = -0.5cm, yshift=0.6cm, font=\fontsize{12}{12}\selectfont] {$\mathbf{X}$};
		\draw[->, line width=2pt] (MLb.east) -- +(1,0) node[above, xshift = -0.5cm, yshift=0.5cm, font=\fontsize{12}{12}\selectfont] {$\widehat{\mathbf{y}}$};
		\draw[->, line width=2pt] (?.east) -- (channelc.west) node[above, xshift = -0.5cm, yshift=0.5cm, font=\fontsize{12}{12}\selectfont] {$h^*$};
		\draw[->, line width=2pt] (channelc.east) -- (MLc.west) node[above, xshift = -0.5cm, yshift=0.5cm, font=\fontsize{12}{12}\selectfont] {$\mathcal{D}$};
		\draw[->, line width=2pt] (MLc.east) -- +(1,0) node[above, xshift = -0.5cm, yshift=0.5cm, font=\fontsize{12}{12}\selectfont] {$\hat{h}$};
	\end{tikzpicture}
	\end{center}
	\caption{(a) A basic communication system involves an information source that emits a message $m$. The 
		message is processed by some transmitter and passed through a noisy channel. The receiver tries to recover 
		the original message $m$ by computing the decoded message $\hat{m}$. (b) The inference step of ML (see Figure \ref{fig_AlexMLBP}) 
		corresponds to a communication problem with an information source emitting 
		a \gls{datapoint} with features $\featurevec$ and label $\truelabel$. The ML method receives the features $\featurevec$ 
		and, in an effort to recover the true label $\truelabel$, computes the predicted label $\hat{\truelabel}$. (c) The learning or adaptation 
		step of ML (see Figure \ref{fig_AlexMLBP}) solves a communication problem with some source that selects a true (but unknown) 
		hypothesis $h^{*}$ as the message. The message is passed through an abstract channel that outputs a 
		set $\dataset$ of labeled \gls{datapoint}s which are used as the \gls{trainset} by an ML method. The ML method tries to 
		decode the true hypothesis resulting in the learnt the hypothesis $\hat{h}$.}
	\label{fig_itml}
\end{figure} 

Information theory studies the problem of communication via noisy channels \cite{Tishby2015ITW,Shannon1948,coverthomas,NITBook}. 
Figure \ref{fig_itml} depicts the most simple communication problem that consists of an information 
source that wishes communicate a message $m$ over an imperfect (or noisy) channel to a receiver. 
The receiver tries to reconstruct (or learn) the original message based solely on the noisy channel 
output. Two main goals of information theory are (i) the characterization of conditions that allow reliable, 
i.e., nearly error-free, communication and (ii) the design of efficient transmitter (coding and modulation)
and receiver (demodulation and decoding) methods. 

It turns out that many concepts from information theory are very useful for the analysis and design of ML methods. 
As a point in case, Chapter \ref{chap_explainable_ML} discusses the application of information-theoretic concepts 
to the design of \gls{xml} methods. On a more fundamental level, we can identify two core communication problems 
that arise within ML. These communication problems correspond, respectively, to the inference (making a prediction) 
and the learning (adjusting or improving the current hypothesis) step of a ML method (see Figure \ref{fig_AlexMLBP}). 

We can an interpret the inference step of ML as the problem of decoding the true label of a \gls{datapoint} for 
which we only know its features. This communication problem is depicted in Figure \ref{fig_itml}-(b). Here the message 
to be communicated is the true label of a random \gls{datapoint}. This \gls{datapoint} is ``communicated'' over a 
channel that only passes through its features. The inference step within a ML method then tries to decode the 
original message (true label) from the channel output (features) resulting in the predicted label. A recent line 
of work used this communication problem to study deep learning methods \cite{Tishby2015ITW}. 

A second core communication problem of ML corresponds to the problem of learning (or adjusting) a 
hypothesis (see Figure \ref{fig_itml}-(c)). In this problem, the source selects some ``true'' hypothesis as message. 
This message is then communicated to an abstract channel that models the data generation process. The 
output of this abstract channel are \gls{datapoint}s in a \gls{trainset} $\dataset$ (see Chapter \ref{ch_Optimization}). 
The learning step of a ML method, such as \gls{erm} of Chapter \ref{ch_Optimization}, then amounts to the 
decoding of the message (true hypothesis) based on the channel output (\gls{trainset}). There is significant 
line or research that uses the communication problem in Figure \ref{fig_itml}-(c) to characterize the fundamental 
limits of ML problems and methods such as the minimum required number of training \gls{datapoint}s that makes 
learning feasible \cite{WangWain2010,Wain2009TIT,Santhanam2012,TranAbrJu2020,JungDicLearn2016}. 

The relevance of information theoretic concepts and methods for ML is boosted by the recent trend 
towards distributed or  federated ML \cite{pmlr-v54-mcmahan17a,Smith2017,SattlerClusteredFL2020,Sarchesh2021}. 
We can interpret \gls{fl} applications as a specific type of network communication problems \cite{NITBook}. 
In particular, we can apply network coding techniques to the design and analysis of \gls{fl} methods \cite{NITBook}. 

\subsection{Probability Theory and Statistics}


Consider the \gls{datapoint}s $\rawfeaturevec^{(1)},\ldots,\rawfeaturevec^{(\samplesize)}$ depicted in Figure \ref{fig_scatterplot_temp_FMI}. 
Each \gls{datapoint} represents some previous day that is characterized by its minimum and maximum daytime 
temperature as measured at a specific \gls{fmi} weather observation station. It might be useful to interpret 
these \gls{datapoint}s as realizations of \gls{iid} \gls{rv}s with common (but typically unknown) \gls{probdist} $\prob{\datapoint}$. 
Figure \ref{fig_scatterplot_temp_FMI_stat_model} extends the \gls{scatterplot} in Figure \ref{fig_scatterplot_temp_FMI} 
by adding a contour line of the underlying \gls{probdist} $\prob{\rawfeaturevec}$ \cite{BertsekasProb}. 

\index{probability theory}Probability theory offers principled methods for estimating a probability 
distribution from a set of \gls{datapoint}s (see Section \ref{sec_max_iikelihood}). Let us assume we 
know (an estimate of) the (joint) \gls{probdist} $\prob{\datapoint}$ of features and label of a \gls{datapoint} $\datapoint=\big(\featurevec,\truelabel)$. 
A principled approach to predict the label value of a \gls{datapoint} with features $\featurevec$ is 
based on evaluating the conditional \gls{probdist} $\prob{\truelabel=\hat{\truelabel}|\featurevec}$. 
The conditional \gls{probdist} $\prob{\hat{\truelabel}=\truelabel|\featurevec}$ quantifies how 
likely it is that $\hat{\truelabel}$ is the actual label value of a \gls{datapoint}. We can evaluate the quantity 
$\prob{\hat{\truelabel}=\truelabel|\featurevec}$ for any candidate value $\hat{\truelabel}$ as soon as we know 
the features $\featurevec$ of this \gls{datapoint}. 

Depending on the performance criterion or \gls{lossfunc}, the optimal prediction $\hat{\truelabel}$ is either 
given by the mode of $\prob{\hat{\truelabel}=\truelabel|\featurevec}$ its mean or some other characteristic value. 
It is important to note that this probabilistic approach not only provides a specific prediction (point-estimate) 
but an entire distribution $\prob{\hat{\truelabel}=\truelabel|\featurevec}$ over possible predictions. This distribution 
allows to construct confidence measures, such as the variance, that can be provided along with the prediction. 

Having a probabilistic model, in the form of a \gls{probdist} $\prob{\datapoint}$, for the data arising in an ML application 
not only allows us to compute predictions for labels of \gls{datapoint}s. We can also use $\prob{\datapoint}$ 
to augment the available dataset by randomly drawing (generating) new \gls{datapoint}s from $\prob{\datapoint}$ (see Section \ref{sec_data_augmentation}). 
ML methods that rely on a probabilistic model for data are sometimes referred to as \index{generative method}generative methods. 
A recently popularized class of generative methods, that uses models obtained from \gls{ann}, is known as generative 
adversarial networks \cite{GoodfellowGAN}.

\begin{figure}[htbp]
	\begin{center}
		\begin{tikzpicture}
			\tikzset{x=0.35cm,y=2cm,every path/.style={>=latex},node style/.style={circle,draw}}
			%
			\begin{axis}[axis x line=none,
				axis y line=none,
				ylabel near ticks,
				xlabel near ticks,
				enlarge y limits=true,
				xmin=-40, xmax=40,
				ymin=-40, ymax=40,
				width=10cm, height=10cm, ]
				\addplot[only marks] table [x=mintmp, y=maxtmp, col sep = comma] {FMIData.csv};
				\node at (axis cs:32,1) [anchor=west] {$\feature$};
				\node at (axis cs:0,30) [anchor=west] {$\truelabel$};
				\draw[->] (axis cs:-30,0) -- (axis cs:32,0);
				\draw[->] (axis cs:0,-30) -- (axis cs:0,30);
				\draw [red,rotate=40,thick] \boundellipse{axis cs:20,-36}{300}{50} node[left,xshift=3mm,yshift=10mm]  {$\,\,\prob{\datapoint}$}; 
			\end{axis}
		\end{tikzpicture}
		\vspace*{-14mm}
	\end{center}
	\caption{Each dot represents a \gls{datapoint} $\datapoint=\big(\feature,\truelabel\big)$ that 
		is characterized by a numeric feature $\feature$ and a numeric label $\truelabel$. 
		We also indicate a contour-line of a \gls{probdist} $\prob{\datapoint}$ that could be 
		used to interpret \gls{datapoint}s as realizations of \gls{iid} \gls{rv}s with common \gls{probdist} $\prob{\datapoint}$. 
	}
	\label{fig_scatterplot_temp_FMI_stat_model}
	\vspace*{-3mm}
\end{figure}

\subsection{Artificial Intelligence}

ML theory and methods are instrumental for the analysis and design of \index{artificial intelligence} \gls{ai} \cite{RusselNorvig}. 
An \gls{ai} system, typically referred to as an agent, interacts with its environment by executing (choosing between different) 
actions. These actions influence the environment as well as the state of the \gls{ai} agent. The behaviour of an \gls{ai} system 
is determined by how the perceptions made about the environment are used to form the next action. 

\begin{figure}[htbp]
	\begin{center}
		\includegraphics[width=8cm]{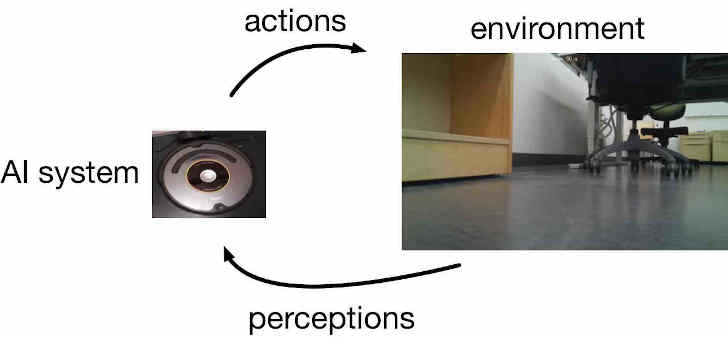}  
		\caption{A cleaning robot chooses actions (moving directions) to maximize 
			a long-term reward measured by the amount of cleaned floor area per day.}
		\label{fig:cleaning_robot}
	\end{center}
\end{figure}

From an engineering point of view, \gls{ai} aims at optimizing behaviour to maximize a long-term 
return. The optimization of behaviour is based solely on the perceptions made by the agent. Let 
us consider some application domains where AI systems can be used:
\begin{itemize}
	\item a forest fire management system: perceptions given by satellite images and local observations using 
	sensors or ``crowd sensing'' via some mobile application which allows humans to notify about relevant events; actions 
	amount to issuing warnings and bans of open fire; return is the reduction of number of forest fires. 
	\item a control unit for combustion engines: 
	perceptions given by various measurements such as temperature, 
	fuel consistency; actions amount to varying fuel feed and timing 
	and the amount of recycled exhaust gas; return is measured in 
	reduction of emissions.   
	\item a severe weather warning service: perceptions given by weather radar; actions 
	are preventive measures taken by farmers or power grid operators; return is measured by 
	savings in damage costs (see \url{https://www.munichre.com/})
	\item an automated benefit application system for the Finnish social insurance institute (``Kela''): 
	perceptions given by information about application and applicant; actions are either to 
	accept or to reject the application along with a justification for the decision; return is 
	measured in reduction of processing time (applicants tend to prefer getting decisions quickly)
	\item  a personal diet assistant: perceived environment is the food preferences of the app 
	user and their health condition; actions amount to personalized suggestions for healthy and tasty food;  
	return is the increase in well-being or the reduction in public spending for health-care.  
	\item the cleaning robot \rumba\, (see\ Figure \ref{fig:cleaning_robot}) perceives its environment 
	using different sensors (distance sensors, on-board camera); actions amount to choosing different 
	moving directions (``north'', ``south'', ``east'', ``west''); return might be the amount of cleaned floor 
	area within a particular time period. 
	\item personal health assistant: perceptions given by current health condition (blood values, weight,\ldots), 
	lifestyle (preferred food, exercise plan); actions amount to personalized suggestions for changing lifestyle habits 
	(less meat, more walking,\ldots); return is measured via the level of well-being (or the reduction in public spending 
	for health-care).  
	\item a government-system for a country: perceived environment is constituted by current economic and 
	demographic indicators such as unemployment rate, budget deficit, age distribution,\ldots; actions involve 
	the design of tax and employment laws, public investment in infrastructure, organization of health-care system; 
	return might be determined by the gross domestic product, the budget deficit or the gross national 
	happiness (cf. \url{https://en.wikipedia.org/wiki/Gross_National_Happiness}). 
\end{itemize}
\vspace*{2mm}

ML methods are used on different levels by an \gls{ai} agent. On a lower level, ML methods help to extract the 
relevant information from raw data. ML methods are used to classify images into different categories which are 
then used an input for higher level functions of the \gls{ai} agent. 

ML methods are also used for higher level tasks of an \gls{ai} agent. To behave optimally, an agent 
is required to learn a good hypothesis for how its behaviour affects its environment. 
We can think of optimal behaviour as a consequent choice of actions that might be predicted 
by ML methods. 

What sets \gls{ai} applications apart from more traditional ML application is that there is an 
strong interaction between ML method and the data generation process. Indeed, \gls{ai} agents 
use the predictions of an ML method to select its next action which, in turn, influences the 
environment which generates new \gls{datapoint}s. The ML subfield of active learning studies 
methods that can influence the data generation \cite{Cohn1996}. 

Another characteristic of \gls{ai} applications is that they typically allow ML methods to 
evaluate the quality of a hypothesis only in hindsight. Within a basic (supervised) ML application  
it is possible for a ML method to try out many different hypotheses on the same \gls{datapoint}. 
These different hypotheses are then scored by their discrepancies with a known correct predictions. 
In contrast to such passive ML applications, AI applications involve \gls{datapoint}s for which it is 
infeasible to determine the correct predictions.

Let us illustrate the above differences between ML and \gls{ai} applications with the help of a self-driving 
toy car. The toy-car is equipped with a small onboard computer, camera, sensors and actors that allow 
to define the steering direction. Our goal is to program the onboard computer such that it implements an 
\gls{ai} agent that optimally steers the toy-car. This \gls{ai} application involves \gls{datapoint}s that represent 
the different (temporal) states of the toy car during its ride. We use a ML method to predict the optimal 
steering direction for the current state. The prediction for the optimal steering angle is obtained by a hypothesis 
map that reads a snapshot form an on-board camera. Since these predictions are used to actually steer the 
car, they influence the future \gls{datapoint}s (states) that will be obtained. 

Note that we typically do not know the actual optimal steering direction for each possible state of the car. 
It is infeasible to let the toy car roam around along any possible path and then manually label each on-board 
camera snapshot with the optimal steering direction (see Figure \ref{fig_rl_donkey}). The usefulness of a 
prediction can be measured only in an indirect fashion by using some form of reward signal. 
Such a \index{reward}reward signal could be obtained from a distance sensor that allows to 
determine if the toy car reduced the distance to a target location.

\section{Flavours of Machine Learning} 

ML methods read in \gls{datapoint}s which are generated within some application domain. 
An individual \gls{datapoint} is characterized by various properties. We find it convenient to divide 
the properties of \gls{datapoint}s into two groups: features and labels (see Chapter \ref{sec_the_data}). 
Features are properties that we measure or compute easily in an automated fashion. Labels are properties 
that cannot be measured easily and often represent some higher level fact (or quantity of interest) 
whose discovery often requires human experts. 

Roughly speaking, ML aims at learning to predict (approximating or guessing) the label of a \gls{datapoint} based solely 
on the features of this \gls{datapoint}. Formally, the prediction is obtained as the function value of a hypothesis map whose 
input argument are the features of a \gls{datapoint}. Since any ML method must be implemented with finite computational 
resources, it can only consider a subset of all possible hypothesis maps. This subset is referred to as the hypothesis 
space or model underlying a ML method. Based on how ML methods assess the quality of different hypothesis maps 
we distinguish three main flavours of ML: supervised, unsupervised and reinforcement learning. 

\subsection{Supervised Learning} 
The main focus of this book is on \index{supervised ML} supervised ML methods. These methods use 
a \gls{trainset} that consists of labeled \gls{datapoint}s (for which we know the correct label values). 
We refer to a \gls{datapoint} as labeled if its label value is known. Labeled \gls{datapoint}s might be obtained 
from human experts that annotate (``label'') \gls{datapoint}s with their label values. There are marketplaces 
for renting human labelling workforce \cite{Sorokin2008}. Supervised ML searches for a hypothesis 
that can imitate the human annotator and allows to predict the label solely from the features of a \gls{datapoint}. 

Figure \ref{fig_curve_fitting} illustrates the basic principle of supervised ML methods. These methods learn 
a hypothesis with minimum discrepancy between its predictions and the true labels of the \gls{datapoint}s 
in the \gls{trainset}. Loosely speaking, supervised ML fits a curve (the graph of the predictor map) to labeled \gls{datapoint}s 
in a \gls{trainset}. For the actual implementing of this curve fitting we need a \gls{lossfunc} that quantifies 
the fitting error. Supervised ML method differ in their choice for a loss function to measure the discrepancy 
between predicted label and true label of a \gls{datapoint}. 

While the principle behind supervised ML sounds trivial, the challenge of modern ML applications is the sheer 
amount of \gls{datapoint}s and their complexity. ML methods must process billions of \gls{datapoint}s with each single 
\gls{datapoint}s characterized by a potentially vast number of features. Consider \gls{datapoint}s representing social network 
users, whose features include all media that has been posted (videos, images, text). Besides the size and complexity 
of datasets, another challenge for modern ML methods is that they must be able to fit highly non-linear predictor maps. 
Deep learning methods address this challenge by using a computationally convenient representation of non-linear maps 
via artificial neural networks \cite{Goodfellow-et-al-2016}.

\begin{figure}[htbp]
	\begin{center}
		\begin{tikzpicture}[auto,scale=1]
			\draw [thick] (1,0.2) rectangle ++(0.1cm,0.1cm) ;
			\draw [thick] (2,3) rectangle ++(0.1cm,0.1cm) ; 
			\draw [thick] (3,2) rectangle ++(0.1cm,0.1cm) node[anchor=west] {\hspace*{0mm}$(x^{(2)},y^{(2)})$};
			\draw [thick] (2,1) rectangle ++(0.1cm,0.1cm) node[anchor=west] {\hspace*{0mm}$(x^{(1)},y^{(1)})$};
			\draw[->] (-0.5,0) -- (3.5,0) node[right] {feature $x$};
			\draw[dashed,line width=1pt] (0,0) -- (3,3) node[anchor=west] {predictor $h(x)$} ; 
			\draw[->] (0,-0.5) -- (0,3.5) node[above] {label $y$};
		\end{tikzpicture}
	\end{center}
	\caption{Supervised ML methods fit a curve to a set of \gls{datapoint}s (which serve as a \gls{trainset}). 
		The curve represents a hypothesis out of some \gls{hypospace} or model. The fitting error (or \gls{trainerr}) is 
		measured using a \gls{lossfunc}. Different ML methods use different combinations of model and \gls{lossfunc}. 
	} 
	\label{fig_curve_fitting}
\end{figure}

\subsection{Unsupervised Learning}
Some ML methods do not require knowing the label value of any \gls{datapoint} and are therefore referred 
to as \index{unsupervised ML} unsupervised ML methods. Unsupervised methods must rely solely on 
the intrinsic structure of \gls{datapoint}s to learn a good hypothesis. Thus, unsupervised methods do not 
need a teacher or domain expert who provides labels for \gls{datapoint}s (used to form a \gls{trainset}). 
Chapters \ref{ch_Clustering} and \ref{ch_FeatureLearning} discuss two large families of unsupervised methods, 
referred to as \index{clustering}clustering and \index{feature learning}feature learning methods. 

\Gls{clustering} methods group \gls{datapoint}s into few subsets or \gls{cluster}. The \gls{datapoint}s within the 
same \gls{cluster} should be more similar with each other than with \gls{datapoint}s outside the \gls{cluster} (see Figure \ref{fig_unsupervised_clustering}). Feature learning methods determine numeric features such that \gls{datapoint}s 
can be processed efficiently using these features. Two important applications of feature learning are dimensionality 
reduction and data visualization.

\begin{figure}[htbp]
	\begin{center}
		\begin{tikzpicture}[auto,scale=0.8]
			\draw [thick] (5,2.5) circle (0.1cm)node[anchor=west] {\hspace*{0mm}$\featurevec^{(3)}$};
			\draw [thick] (4,2) circle (0.1cm)node[anchor=west] {\hspace*{0mm}$\featurevec^{(4)}$};
			\draw [thick] (5,1) circle (0.1cm)node[anchor=west,above] {\hspace*{0mm}$\featurevec^{(2)}$};
			\draw [thick] (1,5)circle (0.1cm) node[anchor=west,above] {\hspace*{0mm}$\featurevec^{(1)}$};
			\draw [thick] (1,3.5)circle (0.1cm)node[anchor=west,above] {\hspace*{0mm}$\featurevec^{(5)}$};
			\draw [thick] (1,2.5) circle (0.1cm)node[anchor=west,above] {\hspace*{0mm}$\featurevec^{(6)}$};
			\draw [thick] (2,4) circle (0.1cm)node[anchor=west,above] {\hspace*{0mm}$\featurevec^{(7)}$};
			\draw[->] (-0.5,0) -- (6.5,0) node[right] {$\feature_{1}$};
			\draw[->] (0,-0.5) -- (0,6.5) node[above] {$\feature_{2}$};
		\end{tikzpicture}
	\end{center}
	\caption{\Gls{clustering} methods learn to predict the cluster (or group) assignments of \gls{datapoint}s 
		based solely on their features. Chapter \ref{ch_Clustering} discusses clustering methods that are 
		unsupervised in the sense of not requiring the knowledge of the true cluster assignment of any \gls{datapoint}. 
	} 
	\label{fig_unsupervised_clustering}
\end{figure}

\subsection{Reinforcement Learning} 
In general, ML methods use a \gls{lossfunc} to evaluate and compare different hypotheses. 
The \gls{lossfunc} assigns a (typically non-negative) loss value to a pair of a \gls{datapoint} and 
a \gls{hypothesis}. ML methods search for a hypothesis, out of (typically large) \gls{hypospace}, 
that incurs minimum loss for any \gls{datapoint}. 
\index{reinforcement learning}Reinforcement learning (RL) studies applications where the predictions 
obtained by a hypothesis influences the generation of future \gls{datapoint}s. RL applications involve \gls{datapoint}s 
that represent the states of a programmable system (an \gls{ai} agent) at different time instants. 
The label of such a \gls{datapoint} has the meaning of an optimal action that the agent should 
take in a given state. Similar to unsupervised ML, RL methods often must learn a hypothesis without 
having access to any labeled \gls{datapoint}. 

In stark contrast to supervised and unsupervised ML methods, RL methods cannot evaluate the \gls{lossfunc} 
for different choices of a hypothesis. Consider a RL method that has to predict the optimal steering angle of a 
car. Naturally, we can only evaluate the usefulness specific combination of predicted label (steering angle) 
and the current state of the car. It is impossible to try out two different hypotheses at the same 
time as the car cannot follow two different steering angles (obtained by the two hypotheses) at the same time. 

Mathematically speaking, RL methods can evaluate the \gls{lossfunc} only point-wise for 
the current hypothesis that has been used to obtain the most recent prediction. These point-wise 
evaluations of the \gls{lossfunc} are typically implemented by using some reward signal \cite{SuttonEd2}. 
Such a reward signal might be obtained from a sensing device and allows to quantify the 
usefulness of the current hypothesis.  

\begin{figure}[htbp]
	\begin{center}
		\includegraphics[width=9cm]{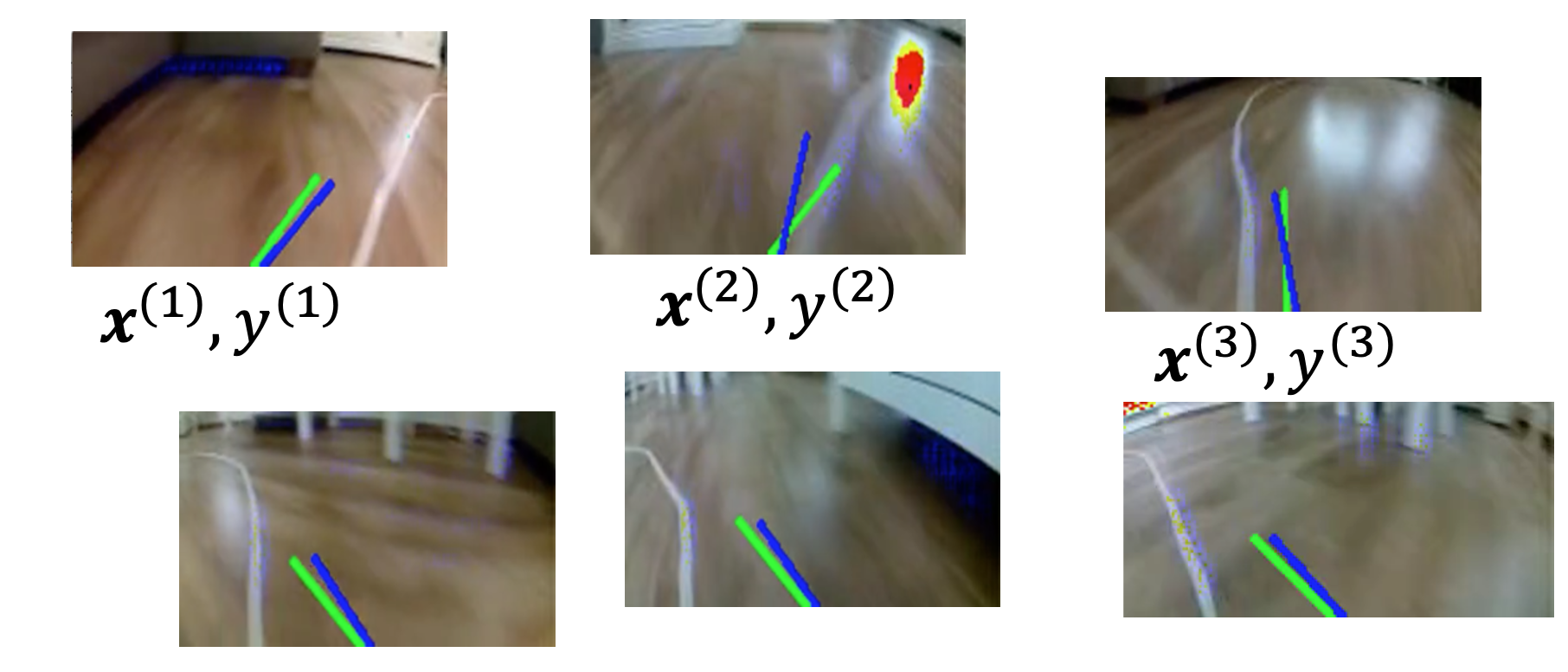}
	\end{center}
	\caption{Autonomous driving requires to predict the optimal steering direction (label) based 
		on an on-board camera snapshot (features) in each time instant. RL methods sequentially adjust 
		a hypothesis for predicting the steering direction from the snapshot. The quality of the current 
		hypothesis is evaluated by the measurement of a distance sensor (to avoid collisions with obstacles).}
	\label{fig_rl_donkey}
\end{figure}

One important application domain for RL methods is autonomous driving (see Figure \ref{fig_rl_donkey}). 
Consider \gls{datapoint}s that represent individual time instants $\timeidx=0,1,\ldots$ during a car ride. The features 
of the $\timeidx$th \gls{datapoint} are the pixel intensities of an on-board camera snapshot taken at time $\timeidx$. 
The label of this \gls{datapoint} is the optimal steering direction at time $\timeidx$ to maximize the distance between 
the car and any obstacle. We could use a ML method to learn hypothesis for predicting the optimal steering direction 
solely from the pixel intensities in the on-board camera snapshot. The \gls{loss} incurred by a particular hypothesis 
is determined from the measurement of a distance sensor after the car moved along the predicted direction. We can 
evaluate the loss only for the hypothesis that has actually been used to predict the optimal steering direction. It is 
impossible to evaluate the loss for other predictions of the optimal steering direction since the car already moved on.

\section{Organization of this Book}

Chapter \ref{ch_Elements_ML} introduces the notions of data, a \gls{model} and a \gls{lossfunc} as 
the three main components of ML. We will also highlight some of the computational and statistical aspects 
that might guide the design choices arising for these three components. A guiding theme of this book is 
the depiction of ML methods as combinations of specific design choices for data representation, the \gls{model} 
and the \gls{lossfunc}. Put differently, we aim at mapping out the vast landscape of ML methods in an abstract 
three-dimensional space spanned by the three dimensions: \gls{data}, \gls{model} and \gls{loss}. 

Chapter \ref{ch_some_examples} details how several well-known ML methods are obtained by 
specific design choices for data (representation), \gls{model} and \gls{lossfunc}. Examples range 
from basic \gls{linreg} (see Section \ref{sec_lin_reg}) via \gls{svm} (see Section \ref{sec_SVM}) to 
deep reinforcement learning (see Section \ref{sec_reinflearning_methods}). 

Chapter \ref{ch_Optimization} discusses a principle approach to combine the three 
components within a practical ML method. In particular, Chapter \ref{ch_Optimization} 
explains how a simple probabilistic model for data lends naturally to the principle of \gls{erm}. 
This principle translates the problem of learning into an optimization problem. ML methods 
based on the \gls{erm} are therefore a special class of optimization methods. The \gls{erm} 
principle can be interpreted as a precise mathematical formulation of the ``learning by trial and error'' paradigm. 

Chapter \ref{ch_GD} discusses a family of iterative methods for solving the \gls{erm} problem 
introduced in Chapter \ref{ch_Optimization}. These methods use the concept of a 
\index{gradient}\gls{gradient} to locally approximate the objective function used in \gls{erm}. 
\Gls{gdmethods} are widely used within deep learning methods to learn useful weights for 
large \gls{ann} (see Section \ref{sec_deep_learning} and \cite{Goodfellow-et-al-2016}). 

The \gls{erm} principle of Chapter \ref{ch_Optimization} requires a hypothesis to accurately predict the labels 
of \gls{datapoint}s in a \gls{trainset}. However, the ultimate goal of ML is to learn a hypothesis that delivers 
accurate predications for any \gls{datapoint} and not only the \gls{trainset}. Chapter \ref{ch_validation_selection} 
discusses some basic validation techniques that allow to probe a hypothesis outside the \gls{trainset} that has 
been used to learn (optimize) this hypothesis via \gls{erm}. Validation techniques are instrumental for model selection, 
i.e., to choose the best model from a given set of candidate models. Chapter \ref{ch_overfitting_regularization} 
presents \gls{regularization} techniques that aim at replacing the \gls{trainerr} of a candidate hypothesis with an 
estimate (or approximation) of its average loss incurred for \gls{datapoint}s outside the \gls{trainset}.

The focus of Chapters \ref{ch_some_examples} - \ref{ch_overfitting_regularization} is on supervised ML 
methods that require a \gls{trainset} of labeled \gls{datapoint}s. Chapters \ref{ch_Clustering} and \ref{ch_FeatureLearning} 
are devoted to unsupervised ML methods which do not require any labeled data. Chapter \ref{ch_Clustering} 
discusses \gls{clustering} methods that partition \gls{datapoint}s into coherent groups which are 
referred to as \index{cluster}\gls{cluster}s. Chapter \ref{ch_FeatureLearning} discusses methods that 
automatically determine the most relevant characteristics (or features) of a \gls{datapoint}. This chapter also 
highlights the importance of using only the most relevant features of a \gls{datapoint}, and to avoid irrelevant 
features, to reduce computational complexity and improve the accuracy of ML methods (such as those discussed in 
Chapter \ref{ch_some_examples}). 

The success of ML methods becomes increasingly dependent on their \gls{explainability} or transparency for 
the user of the ML method \cite{Hagras2018,Mcinerney18}. The \gls{explainability} of a ML method and its
 predictions typically depends on the background knowledge of the user which might vary significantly. 
 Chapter \ref{chap_explainable_ML} discusses two different approaches to obtain personalized \gls{xml}. 
 These techniques use a feedback signal, which is provided for the \gls{datapoint}s in a \gls{trainset}, to 
 either compute personalized explanations for a given ML method or to choose models that are intrinsically 
 explainable to a specific user. 

{\bf Prerequisites.} We assume familiarity with basic notions and concepts of linear algebra, real analysis, and 
probability theory \cite{StrangLinAlg2016,RudinBookPrinciplesMatheAnalysis}. For a brief review of those 
concepts, we recommend \cite[Chapter 2-4]{Goodfellow-et-al-2016} and the references therein. A main 
goal of this book is to develop the basic ideas and principles behind ML methods using a minimum of 
probability theory. However, some rudimentary knowledge about the concept of expectations, probability 
density function of a continuous (real-valued) \gls{rv} and probability mass function of a discrete \gls{rv} 
is helpful \cite{BertsekasProb,GrayProbBook}.

\chapter{Three Components of ML} 
\label{ch_Elements_ML}

\begin{figure}[htbp]
	\begin{center}
		\tikzset{pics/.cd,
			jigsaw/.style={
				code={
					\fill[#1] (-2,-0.35) to[out=90,in=135] (-1.5,-0.45) arc(-135:135:0.6 and
					{0.45*sqrt(2)}) to[out=-135,in=-90] (-2,0.35) |- (-0.35,2)
					to[out=0,in=-45] (-0.45,2.5) arc(225:-45:{0.45*sqrt(2)} and 0.6)
					to[out=-135,in=180] (0.35,2) -| (2,0.35) 
					to[out=-90,in=225] (2.5,0.45) arc(135:-135:0.6 and {0.45*sqrt(2)})
					to[out=135,in=90] (2,-0.35) |- (0.35,-2)
					to[out=180,in=-135] (0.45,-1.5) arc(-45:225:{0.45*sqrt(2)} and 0.6) 
					to[out=-45,in=0] (-0.35,-2) -| cycle;
		}}}
		
		\resizebox{5cm}{!}{
			\begin{tikzpicture}
				\draw (-2,-2) pic{jigsaw=white!80!blue} (2,-2) pic{jigsaw=white!80!green}
				(2,2) pic[rotate=90]{jigsaw=white!80!red};
				\node[anchor=west] at (0.4,3.1) {\bf\Huge \gls{model}};
				\node[anchor=west] at (0.4,-0.9) {\bf\Huge \gls{data}};
				\node[anchor=west] at (-2.4,-0.9) {\bf\Huge \gls{loss}};
		\end{tikzpicture}}
	\end{center} 
	\caption{ML methods fit a \gls{model} to data via minimizing a \gls{lossfunc} (see Figure \ref{fig_AlexMLBP}). 
	 We obtain a variety of ML methods from different design choices for the \gls{model}, \gls{lossfunc} and \gls{data} 
	 representation (see Chapter \ref{ch_some_examples}). A principled approach to combine these three 
	 components is \gls{erm} (see Chapter \ref{ch_Optimization}).}
	\label{fig_ml_problem}
\end{figure}

This book portrays ML as combinations of three components as depicted in Figure \ref{fig_ml_problem}. 
These components are 
\begin{itemize}
	\item \gls{data} as collections of individual \gls{datapoint}s that are characterized by 
	\gls{features} (see Section \ref{sec_feature_space}) and \gls{label}s (see Section \ref{sec_labels}) 
	\item a \gls{model} or \gls{hypospace} that consists of computationally feasible 
	hypothesis maps from feature space to \gls{labelspace} (see Section \ref{sec_hypo_space}) 
	\item a \gls{lossfunc} (see Section \ref{sec_lossfct}) to measure the quality of a hypothesis map. 
\end{itemize}
A ML problem involves specific design choices for \gls{datapoint}s, its features and labels, 
the \gls{hypospace} and the \gls{lossfunc} to measure the quality of a particular hypothesis. 
Similar to ML problems (or applications), we can also characterize ML methods as combinations 
of these three components. This chapter discusses in some depth each of the above 
ML components and their combination within some widely-used ML methods \cite{JMLR:v12:pedregosa11a}. 

We detail in Chapter \ref{ch_some_examples} how some of the most popular ML methods, 
including \gls{linreg} (see Section \ref{sec_lin_reg}) as well as deep learning 
methods (see Section \ref{sec_deep_learning}), are obtained by specific design choices 
for the three components. Chapter \ref{ch_Optimization} will then introduce \gls{erm} as 
a main principle for how to operationally combine the individual ML components. Within the \gls{erm} 
principle, ML problems become optimization problems and ML methods become optimization methods. 

\section{The Data}
\label{sec_the_data}

\begin{figure}[htbp]
	\centering
	\includegraphics[width=7cm]{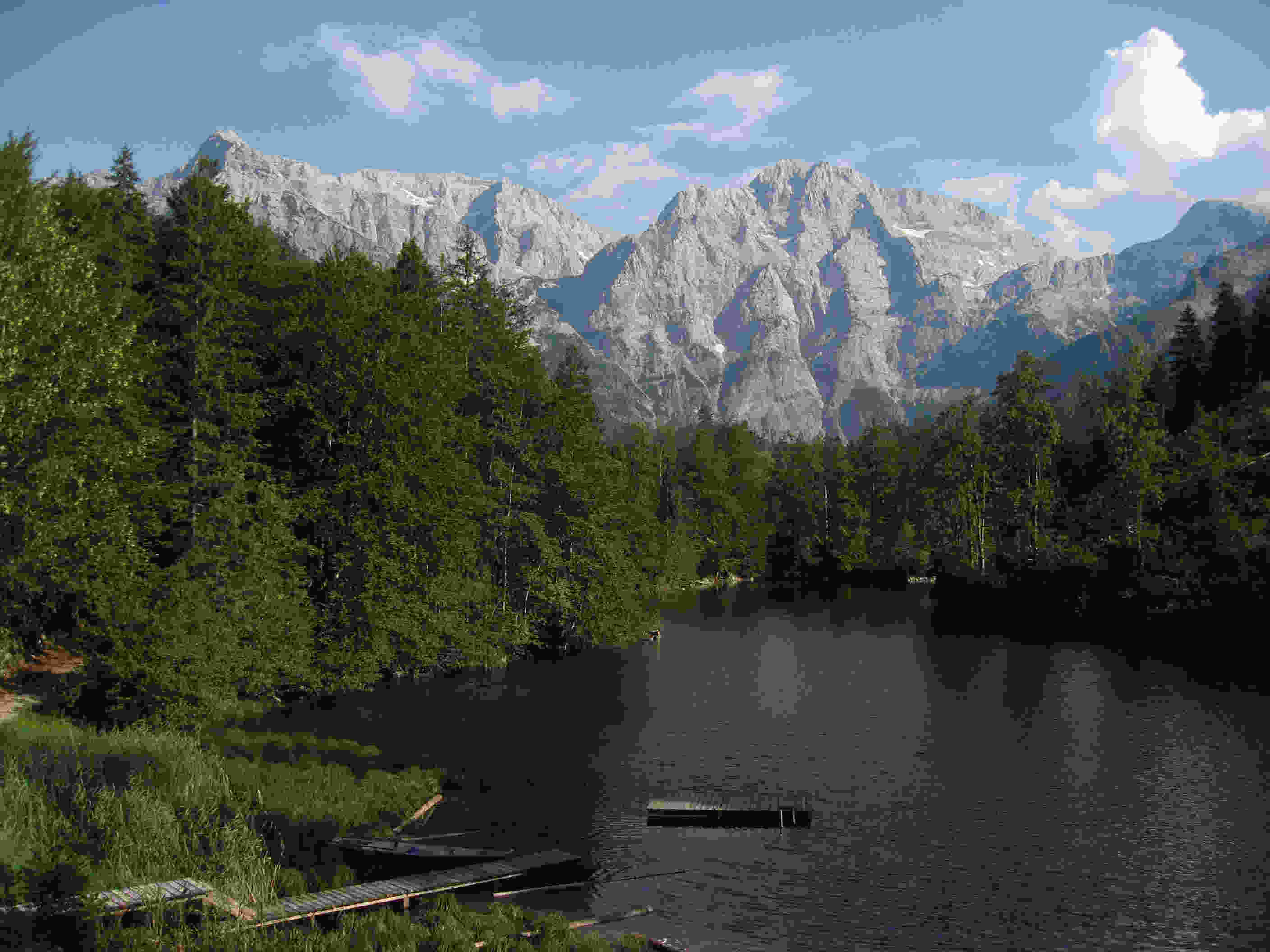}
	\caption{A snapshot taken at the beginning of a mountain hike.}
	\label{fig:image}
\end{figure}

{\bf Data as Collections of \Gls{datapoint}s.} Maybe the most important component of any ML problem 
(and method) is data. We consider data as collections of individual \gls{datapoint}s which are atomic units 
of ``information containers''. \Gls{datapoint}s can represent text documents, signal samples of time series 
generated by sensors, entire time series generated by collections of sensors, frames within a single video, 
random variables, videos within a movie database, cows within a herd, trees within a forest, or forests 
within a collection of forests. Mountain hikers might be interested in \gls{datapoint}s that represent different 
hiking tours (see Figure \ref{fig:image}). 

We use the concept of \gls{datapoint}s in a very abstract and therefore highly flexible manner. \Gls{datapoint}s can represent 
very different types of objects that arise in fundamentally different application domains. For an image processing 
application it might be useful to define \gls{datapoint}s as images. When developing a recommendation system we might 
define \gls{datapoint}s to represent customers. In the development of new drugs we might use \gls{datapoint}s to represent 
different diseases. Ultimately, the choice or definition of \gls{datapoint}s is a design choice. We might refer to the 
task of finding a useful definition of \gls{datapoint}s as \index{data engineering}``\gls{datapoint} engineering''. 

One practical requirement for a useful definition of \gls{datapoint}s is that we should have access 
to many of them. Many ML methods construct estimates for a quantity of interest (such as a prediction 
or forecast) by averaging over a set of reference (or training) \gls{datapoint}s. These estimates become  
more accurate for an increasing number of \gls{datapoint}s used for computing the average. 
A key property of a dataset is the number $\samplesize$ of individual \gls{datapoint}s it contains. 
The number of \gls{datapoint}s within a dataset is also referred to as the \gls{samplesize}. 
From a statistical point of view, the larger the sample size $\samplesize$ the better. 
However, there might be restrictions on computational resources (such as memory size) that 
limit the maximum sample size $\samplesize$ that can be processed. 

For most applications, it is impossible to have full access to every single microscopic property of a \gls{datapoint}. 
Consider a \gls{datapoint} that represents a vaccine. A full characterization of such a \gls{datapoint} would require to 
specify its chemical composition down to level of molecules and atoms. Moreover, there are properties of a 
vaccine that depend on the patient who received the vaccine. 

We find it useful to distinguish between two different groups of properties of a \gls{datapoint}. The first group 
of properties is referred to as \gls{features} and the second group of properties is referred to as \gls{label}s. Roughly 
speaking, features are low-level properties of a \gls{datapoint} that can be measured or computed easily in an 
automated fashion. In contract, labels are high-level properties of a \gls{datapoint}s that represent some quantity 
of interest. Determining the label value of a \gls{datapoint} often requires human labour, e.g., a domain expert who 
has to examine the \gls{datapoint}. Some widely used synonyms for \gls{features} are ``covariate'',``explanatory variable'', 
``independent variable'', ``input (variable)'', ``predictor (variable)'' or ``regressor''  \cite{Gujarati2021,Dodge2003,Everitt2022}. 
Some widely used synonyms for the label of a \gls{datapoint} are "response variable", "output variable" or "target"  \cite{Gujarati2021,Dodge2003,Everitt2022}.

We will discuss the concepts of features and labels in somewhat more detail in Sections \ref{sec_feature_space} and \ref{sec_labels}. 
However, we would like to point out that the distinction between features and labels is blurry. The same 
property of a \gls{datapoint} might be used as a feature in one application, while it might be used as a label 
in another application. Let us illustrate this blurry distinction between features and labels using the 
problem of \index{missing data}\gls{missingdata}. 

Assume we have a list of \gls{datapoint}s each of which is characterized by several properties that could be 
measured easily in principles (by sensors). These properties would be first candidates for being used as features 
of the \gls{datapoint}s. However, some of these properties are unknown (missing) for a small set of \gls{datapoint}s 
(e.g., due to broken sensors). We could then define the properties which are missing for some \gls{datapoint}s as labels 
and try to predict these labels using the remaining properties (which are known for all \gls{datapoint}s) 
as features. The task of determining missing values of properties that could be measured easily in principle 
is referred to as \index{imputation} imputation \cite{Abayomi2008DiagnosticsFM}. 

Missing data might also arise in image processing applications. Consider \gls{datapoint}s being images (snapshots) 
generated by a smartphone camera. Maybe the most intuitive choice for the features of a (bitmap) image are the
 colour intensities for each pixel (see Figure \ref{fig_snapshot_pixels}). Due to hardware failures some of 
the image pixels might be corrupted or (their colour intensities) even completely missing. We could then try to 
use to learn to predict the colour intensities of a pixel based on the colour intensities of the neighbouring pixels. 
To this end, we might define new \gls{datapoint}s as small rectangular regions (patches) of the 
image and use the colour intensity of the centre pixel (``target pixel'') as the label of such a patch. 

Figure \ref{fig:two_main_parameters} illustrates two key properties of a \index{dataset}\gls{dataset}. 
The first property is the \index{sample size}\gls{samplesize} $\samplesize$, i.e., the number of 
individual \gls{datapoint}s that constitute the dataset. The second key property of is the number $\featurelen$ 
of \index{feature}\gls{features} that are used to characterize an individual \gls{datapoint}. The behaviour of 
ML methods often depends crucially on the ratio $\samplesize/\featurelen$. The performance of ML methods 
typically improves with increasing $\samplesize/\featurelen$. As a rule of thumb, we should use datasets for 
which $\samplesize/\featurelen \gg 1$. We will make the informal condition $\samplesize / \featurelen \gg 1$ 
more precise in Chapter \ref{ch_validation_selection}. 

\begin{figure}[htbp]
	\centering
	
	\tikzset{global scale/.style={
			scale=#1,
			every node/.append style={scale=#1}
		}
	}
	\begin{tikzpicture}[node distance=0cm, outer sep = 0pt, global scale = 0.8]
		
		\tikzstyle{wide}=[rectangle, draw, minimum height=0.8cm, minimum width=1.8cm, anchor=north west, text centered]
		\tikzstyle{narrow}=[rectangle, draw, minimum height=0.8cm, minimum width=1.8cm, anchor=north west, text centered]
		
		\node[wide] (year) at (0,0) {Year};
		 \node[] (yearcorner) [above left=0.4cm and 0.1cm of  year] {};
		\node[wide] (2020-1) [below = of year] {2020};
       \node[] (samplesizeup) [below left= 0cm and 0.1cm of year] {};
		\node[wide] (2020-2) [below = of 2020-1] {2020};
		\node[wide] (2020-3) [below = of 2020-2] {2020};
		\node[wide] (2020-4) [below = of 2020-3] {2020};
		\node[wide] (2020-5) [below = of 2020-4] {2020};
		\node[wide] (2020-6) [below = of 2020-5] {2020};
		 \node[] (samplesizedown) [below left= 0cm and 0.1cm of 2020-6] {};
		
		\node[narrow] (m) [right = of year] {m};
		\node[narrow] (1-1) [below = of m] {1};
		\node[narrow] (1-2) [below = of 1-1] {1};
		\node[narrow] (1-3) [below = of 1-2] {1};
		\node[narrow] (1-4) [below = of 1-3] {1};
		\node[narrow] (1-5) [below = of 1-4] {1};
		\node[narrow] (1-6) [below = of 1-5] {1};
		
		\node[narrow] (d) [right = of m] {d};
		\node[narrow] (2) [below = of d] {2};
		\node[narrow] (3) [below = of 2] {3};
		\node[narrow] (4) [below = of 3] {4};
		\node[narrow] (5) [below = of 4] {5};
		\node[narrow] (6) [below = of 5] {6};
		\node[narrow] (7) [below = of 6] {7};
		
		\node[wide] (time) [right = of d] {Time};
		\node[wide] (0-1) [below = of time] {00:00};
		\node[wide] (0-2) [below = of 0-1] {00:00};
		\node[wide] (0-3) [below = of 0-2] {00:00};
		\node[wide] (0-4) [below = of 0-3] {00:00};
		\node[wide] (0-5) [below = of 0-4] {00:00};
		\node[wide] (0-6) [below = of 0-5] {00:00};
		
		\node[wide] (precp) [right = of time] {precp};
		\node[wide] (04) [below = of precp] {0,4};
		\node[wide] (16) [below = of 04] {1,6};
		\node[wide] (01) [below = of 16] {0,1};
		\node[wide] (19) [below = of 01] {1,9};
		\node[wide] (06) [below = of 19] {0,6};
		\node[wide] (41) [below = of 06] {4,1};
		
		\node[wide] (snow) [right = of precp] {snow};
		\node[wide] (55) [below = of snow] {55};
		\node[wide] (53) [below = of 55] {53};
		\node[wide] (51) [below = of 53] {51};
		\node[wide] (52-1) [below = of 51] {52};
		\node[wide] (52-2) [below = of 52-1] {52};
		\node[wide] (52-3) [below = of 52-2] {52};
		
		\node[wide] (airtmp) [right = of snow] {airtmp};
		\node[wide] (25) [below = of airtmp] {2,5};
		\node[wide] (08) [below = of 25] {0,8};
		\node[wide] (58) [below = of 08] {-5,8};
		\node[wide] (135) [below = of 58] {-13,5};
		\node[wide] (24) [below = of 135] {-2,4};
		\node[wide] (04a) [below = of 24] {0,4};
		
		\node[wide] (mintmp) [right = of airtmp] {mintmp};
		\node[wide] (2m) [below = of mintmp] {-2};
		\node[wide] (08m) [below = of 2m] {-0,8};
		\node[wide] (111) [below = of 08m] {-11,1};
		\node[wide] (191) [below = of 111] {-19,1};
		\node[wide] (114) [below = of 191] {-11,4};
		\node[wide] (2m-1) [below = of 114] {-2};
		
		\node[wide] (maxtmp) [right = of mintmp] {maxtmp};
	   \node[] (maxtmpcorner) [above right=0.4cm and 0.1cm of  maxtmp] {};
		\node[wide] (45) [below = of maxtmp] {4,5};
		\node[wide] (46) [below = of 45] {4,6};
		\node[wide] (07) [below = of 46] {-0,7};
		\node[wide] (46-1) [below = of 07] {-4,6};
		\node[wide] (1) [below = of 46-1] {-1};
		\node[wide] (13) [below = of 1] {1,3};
		
		\draw [stealth-stealth, very thick, color=blue] (yearcorner) -- (maxtmpcorner) node[above, xshift=-6.3cm, yshift=0.2cm, black] {\Large $\featuredim$};
		\draw [stealth-stealth, very thick, color=blue](samplesizeup) -- (samplesizedown) node[left, xshift=-0.2cm, yshift=2.1cm, black] {\Large $\samplesize$};
		
	\end{tikzpicture}
	\caption{Two key properties of a dataset are the number (sample size) $\samplesize$ of 
		individual \gls{datapoint}s that constitute the dataset and the number $\featuredim$ of 
		features used to characterize individual \gls{datapoint}s. The behaviour of ML methods typically 
		depends crucially on the ratio $\samplesize/\featuredim$. }
	\label{fig:two_main_parameters}
\end{figure}

\subsection{Features}
\label{sec_feature_space}

Similar to the choice (or definition) of \gls{datapoint}s with an ML application, also the choice of 
which properties to be used as their features is a design choice. In general, features are (low-level) 
properties of a \gls{datapoint} that can be computed or measured easily. This is obviously a highly 
informal characterization since there is no universal criterion for the difficulty of computing of 
measuring a specific property of \gls{datapoint}s. The task of choosing which properties to use as 
features of \gls{datapoint}s might be the most challenging part in the application of ML methods. 
Chapter \ref{ch_FeatureLearning} discusses feature learning methods that automate (to some extend) 
the construction of good features. 

In some application domains there is a rather natural choice for the features of a \gls{datapoint}. For 
\gls{datapoint}s representing audio recording (of a given duration) we might use the signal amplitudes 
at regular sampling instants (e.g., using sampling frequency $44$ kHz) as features.  
For \gls{datapoint}s representing images it seems natural to use the colour (red, green and blue) 
intensity levels of each pixel as a feature (see Figure \ref{fig_snapshot_pixels}).

The feature construction for images depicted in Figure \ref{fig_snapshot_pixels} can be extended to 
other types of \gls{datapoint}s as long as they can be visualized efficiently \cite{Friendly:06:hbook}. 
Audio recordings are typically available a sequence of signal amplitudes $a_{\timeidx}$ collected regularly 
at time instants $\timeidx=1,\ldots,\featuredim$ with sampling frequency $\approx  44$ kHz. From a signal 
processing perspective, it seems natural to directly use the signal amplitudes as features, $\feature_{\featureidx} = a_{\featureidx}$ for $\featureidx=1,\ldots,\featurelen$. However, another choice for the features would be the pixel 
RGB values of some visualization of the audio recording. 

\begin{figure}[htbp]
	\begin{center}
		\includegraphics[width=0.7\textwidth]{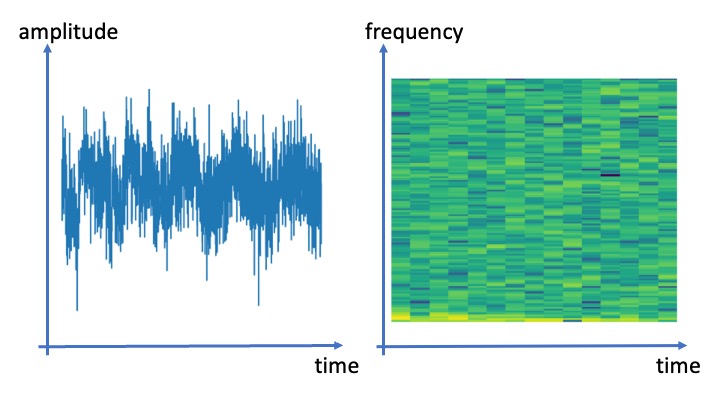}  
	\end{center}
	\caption{Two visualizations of a \gls{datapoint} that represents an \index{audio recording}audio recording. 
		The left figure shows a line plot of the audio signal amplitudes. The right figure shows a \index{spectogram}\gls{spectogram} of the audio recording.}
	\label{fig_visualization_audio}
\end{figure}

Figure \ref{fig_visualization_audio} depicts two possible visualizations of an audio signal. The first visualization is obtained 
from a line plot of the signal amplitudes as a function of time $\timeidx$. Another visualization of an audio recording 
is obtained from an intensity plot of its \gls{spectogram}\cite{TimeFrequencyAnalysisBoashash,MallatBook}. We can 
then use the pixel RGB intensities of these visualizations as the features for an audio recording. Using this 
trick we can transform any ML method for image data to an ML method for audio data. We can use the \gls{scatterplot} 
of a data set to use ML methods for image segmentation to \gls{cluster} the \gls{dataset}(see Chapter \ref{ch_Clustering}). 

\begin{figure}[htbp]
	\begin{center}
		\begin{tikzpicture}
			\draw[step=0.5, help lines] (-2,-2) grid (2,2);
			\filldraw[color=black, fill=red!80] (-2,1.5) -- (-1.5,1.5) -- (-1.5,2) -- (-2,2) -- (-2,1.5);
			\filldraw[color=black, fill=green!80] (-1.5,1.5) -- (-1,1.5) -- (-1,2) -- (-1.5,2) -- (-1.5,1.5);
			\filldraw[color=black, fill=blue!80] (-1,1.5) -- (-0.5,1.5) -- (-0.5,2) -- (-1,2) -- (-1,1.5);
			\node[font=\fontsize{12}{0}\selectfont] at (-4,0) {snapshot $\datapoint^{(\sampleidx)}$};
			\node[font=\fontsize{12}{0}\selectfont] at (0,-3) {$512\times512$ pixels};
			\node[font=\fontsize{12}{0}\selectfont] at (0,-3.5) {(red-green-blue bitmap)};
			\node[font=\fontsize{12}{0}\selectfont] at (4,2.8) {feature vector $\featurevec^{(\sampleidx)}$};
			\node[font=\fontsize{10}{0}\selectfont] at (-3,1.2) {pixel 1};
			\node[font=\fontsize{10}{0}\selectfont] at (-1.8,2.6) {pixel 2};
			\node[font=\fontsize{10}{0}\selectfont] at (3,-2.7) {pixel $512\times512$};
			\node[font=\fontsize{10}{0}\selectfont] at (5.8,0) {$\featurelen$};
			\node[font=\fontsize{10}{0}\selectfont] at (3.5,2) {$255$};
			\node[font=\fontsize{10}{0}\selectfont] at (3.5,1.6) {$0$};
			\node[font=\fontsize{10}{0}\selectfont] at (3.5,1.2) {$0$};
			\node[font=\fontsize{10}{0}\selectfont] at (3.5,0.8) {$0$};
			\node[font=\fontsize{10}{0}\selectfont] at (3.5,0.4) {$255$};
			\node[font=\fontsize{10}{0}\selectfont] at (3.5,0) {$0$};
			\node[font=\fontsize{10}{0}\selectfont] at (3.5,-0.4) {$0$};
			\node[font=\fontsize{10}{0}\selectfont] at (3.5,-0.8) {$0$};
			\node[font=\fontsize{10}{0}\selectfont] at (3.5,-1.2) {$255$};
			\node[font=\fontsize{10}{0}\selectfont] at (3.5,-1.6) {$\vdots$};
			\node[font=\fontsize{10}{0}\selectfont] at (3.5,-2) {$\vdots$};
			\node[font=\fontsize{8}{0}\selectfont] at (4.5,2) {$r[1]$};
			\node[font=\fontsize{8}{0}\selectfont] at (4.5,1.6) {$g[1]$};
			\node[font=\fontsize{8}{0}\selectfont] at (4.5,1.2) {$b[1]$};
			\node[font=\fontsize{8}{0}\selectfont] at (4.5,0.8) {$r[2]$};
			\node[font=\fontsize{8}{0}\selectfont] at (4.5,0.4) {$g[2]$};
			\node[font=\fontsize{8}{0}\selectfont] at (4.5,0) {$b[2]$};
			\draw [-, color=black](-2.8,1.5) -- (-2,1.75);
			\draw [-, color=black](-1.6,2.4) -- (-1.25,2);
			\draw [-, color=black](2,-1.75) -- (2.6,-2.4);
			\draw [stealth-stealth, very thick, color=black](5.5,2.2) -- (5.5,-2.2);
			\draw [-stealth, very thick, color=black](-1.75,2) to [bend left=50] (2.5,1.6);
			\draw [-stealth, very thick, color=black](-1.25,2) to [bend left=45] (2.5,0.4);
			\draw [-stealth, very thick, color=black](-0.5,1.75) to [bend left=30] (2.5,-0.8);
			\draw[decorate,decoration={brace, raise=5pt, amplitude=2mm, mirror}, very thick] (3,2.1) -- (3,1.1);
			\draw[decorate,decoration={brace, raise=5pt, amplitude=2mm, mirror}, very thick] (3,0.9) -- (3,-0.1);
			\draw[decorate,decoration={brace, raise=5pt, amplitude=2mm, mirror}, very thick] (3,-0.3) -- (3,-1.3);
			\draw[color=black, thick] (3.05,1.75) -- (3.05,-1.75);
			\draw[color=black, thick] (3.15,2.2) to [bend right=7] (3.05,1.75);
			\draw[color=black, thick] (3.05,-1.75) to [bend right=7] (3.15,-2.2);
			\draw[color=black, thick] (3.95,1.75) -- (3.95,-1.75);
			\draw[color=black, thick] (3.85,2.2) to [bend left=7] (3.95,1.75);
			\draw[color=black, thick] (3.95,-1.75) to [bend left=7] (3.85,-2.2);
		\end{tikzpicture}
	\end{center}
	\caption{If the snapshot $\datapoint^{(\sampleidx)}$ is stored as a $512 \times 512$ RGB bitmap, we could use as 
		features $\featurevec^{(\sampleidx)} \in \mathbb{R}^{\featuredim}$ the red-, green- and blue component of each pixel 
		in the snapshot. The length of the feature vector would then be $\featuredim= 3 \times 512 \times 512 \approx 786000$. }
	\label{fig_snapshot_pixels}
\end{figure}

Many important ML application domains generate \gls{datapoint}s that are characterized by several numeric 
features $x_{1},\ldots,x_{\featuredim}$. We represent numeric features by real numbers $x_{1},\ldots,x_{\featuredim} \in \mathbb{R}$ 
which might seem impractical. Indeed, digital computers cannot store a real number exactly 
as this would require an infinite number of bits. However, numeric linear algebra soft- and hardware  
allows to approximate real numbers with sufficient accuracy. The majority of ML methods discussed 
in this book assume that \gls{datapoint}s are characterized by real-valued features. Section \ref{sec_discrete_embeddings} 
discusses methods for constructing numeric features of \gls{datapoint}s whose natural representation is non-numeric.  

We assume that \gls{datapoint}s arising in a given ML application are characterized by the same number $\featurelen$ 
of individual features $\feature_{1}.\ldots,\feature_{\featurelen}$. It is convenient to stack the individual features 
of a \gls{datapoint} into a single feature vector 
\begin{equation}
	\nonumber 
	\featurevec=\big(\feature_{1},\ldots,\feature_{\featuredim}\big)^{T}. 
\end{equation}
Each \gls{datapoint} is then characterized by its feature vector $\featurevec$. Note that stacking 
the features of a \gls{datapoint} into a column vector $\featurevec$ is pure convention. We could also 
arrange the features as a row vector or even as a matrix, which might be even more natural 
for features obtained by the pixels of an image (see Figure \ref{fig_snapshot_pixels}). 

We refer to the set of possible feature vectors of \gls{datapoint}s arising in some ML application 
as the \index{feature space}\gls{featurespace} and denote it as $\featurespace$. 
The feature space is a design choice as it depends on what properties of a 
\gls{datapoint} we use as its features. This design choice should take into account 
the statistical properties of the data as well as the available computational infrastructure. 
If the computational infrastructure allows for efficient numerical linear algebra, 
then using $\featurespace = \mathbb{R}^{\featurelen}$ might be a good choice.

The \gls{euclidspace} $\mathbb{R}^{\featuredim}$ is an example of a feature space with 
a rich geometric and algebraic structure \cite{RudinBookPrinciplesMatheAnalysis}. 
The algebraic structure of $\mathbb{R}^{\featuredim}$ is defined by vector addition 
and multiplication of vectors with scalars. The geometric structure of $\mathbb{R}^{\featuredim}$ 
is obtained by the Euclidean norm as a measure for the distance between two elements of $\mathbb{R}^{\featuredim}$. 
The algebraic and geometric structure of $\mathbb{R}^{\featuredim}$ often enables an 
efficient search over $\mathbb{R}^{\featuredim}$ to find elements with desired properties. 
Chapter \ref{sec_ERM_lin_reg} discusses examples of such search problems in the context 
of learning an optimal hypothesis. 

Modern information-technology, including smartphones or wearables, allows us to measure 
a huge number of properties about \gls{datapoint}s in many application domains. Consider a \gls{datapoint} 
representing the book author ``Alex Jung''. Alex uses a smartphone to take roughly five snapshots 
per day (sometimes more, e.g., during a mountain hike) resulting in more than $1000$ snapshots per year. 
Each snapshot contains around $10^{6}$ pixels whose greyscale levels we can use as features of the 
\gls{datapoint}. This results in more than $10^{9}$ features (per year!). 

As indicated above, many important ML applications involve \gls{datapoint}s represented by very long feature 
vectors. To process such high-dimensional data, modern ML methods rely on concepts from high-dimensional 
statistics \cite{BuhlGeerBook,Wain2019}. One such concept from high-dimensional statistics is the notion of 
sparsity. Section \ref{sec_lasso} discusses methods that exploit the tendency of high-dimensional \gls{datapoint}s, 
which are characterized by a large number $\featuredim$ of features, to concentrate near low-dimensional 
subspaces in the \gls{featurespace} \cite{VidalMag}.  

At first sight it might seem that ``the more features the better'' since using more features might 
convey more relevant information to achieve the overall goal. However, as we discuss in Chapter \ref{ch_overfitting_regularization}, 
it can be detrimental for the performance of ML methods to use an excessive amount of (irrelevant) features. 
Computationally, using too many features might result in prohibitive computational resource requirements (such as 
processing time). Statistically, each additional feature typically introduces an additional amount of noise (due 
to measurement or modelling errors) which is harmful for the accuracy of the ML method. 

It is difficult to give a precise and broadly applicable characterization of the maximum number of 
features that should be used to characterize the \gls{datapoint}s. As a rule of thumb, the number $\samplesize$ 
of (labeled) \gls{datapoint}s used to train a ML method should be much larger than the number $\featurelen$ 
of numeric features (see Figure \ref{fig:two_main_parameters}). The informal condition $\samplesize/\featurelen \gg 1$ 
can be ensured by either collecting a sufficiently large number $\samplesize$ of \gls{datapoint}s 
or by using a sufficiently small number $\featurelen$ of features. We next discuss implementations 
for each of these two complementary approaches. 

The acquisition of (labeled) \gls{datapoint}s might be costly, requiring human expert labour. Instead of collecting 
more raw data, it might be more efficient to generate new artificial (synthetic) data via \gls{dataug} techniques. 
Section \ref{sec_data_augmentation} shows how intrinsic symmetries in the data can be used to augment the raw 
data with synthetic data. As an example for an intrinsic symmetry of data, consider \gls{datapoint}s being images. 
We assign each image the label $\truelabel=1$ if it shows a cat and $\truelabel=-1$ otherwise. For each image with known 
label we can generate several augmented (additional) images with the same label. These additional images might be 
obtained by simple image transformation such as rotations or re-scaling (zoom-in or zoom-out) that do not change 
the depicted objects (the meaning of the image). Chapter \ref{ch_overfitting_regularization} shows that some basic 
regularization techniques can be interpreted as an implicit form of \index{data augmentation} \gls{dataug}.  

The informal condition $\samplesize/\featurelen \gg 1$ can also be ensured by reducing the number $\featurelen$ 
of features used to characterize \gls{datapoint}s. In some applications, we might use some domain knowledge to 
choose the most relevant features. For other applications, it might be difficult to tell which quantities are 
the best choice for features. Chapter \ref{ch_FeatureLearning}  discusses methods that learn, based on 
some given dataset, to determine a small number of relevant features of \gls{datapoint}s. 

Beside the available computational infrastructure, also the statistical properties of datasets  
must be taken into account for the choices of the feature space. The linear algebraic structure 
of $\mathbb{R}^{\featuredim}$ allows us to efficiently represent and approximate datasets that are well aligned 
along linear subspaces. Section \ref{sec_pca} discusses a basic method to optimally approximate 
datasets by linear subspaces of a given dimension. The geometric structure of $\mathbb{R}^{\featuredim}$ 
is also used in Chapter \ref{ch_Clustering} to decompose a \gls{dataset} into few \index{groups}groups 
or \index{clusters}clusters that consist of similar \gls{datapoint}s. 

Throughout this book we will mainly use the feature space $\mathbb{R}^{\featuredim}$ 
with dimension $\featuredim$ being the number of features of a \gls{datapoint}. This \gls{featurespace} 
has proven useful in many ML applications due to availability of efficient soft- and hardware for 
numerical linear algebra. Moreover, the algebraic and geometric structure of $\mathbb{R}^{\featuredim}$ 
reflect the intrinsic structure of data arising in many important application domains. This should 
not be too surprising as the \index{Euclidean space}\gls{euclidspace} has evolved as a useful 
mathematical abstraction of physical phenomena \cite{KibbleBerkshireBook}.

In general there is no mathematically correct choice for which properties of a \gls{datapoint} to be used 
as its features. Most application domains allow for some design freedom in the choice of features. Let us illustrate this 
design freedom with a personalized health-care applications. This application involves \gls{datapoint}s 
that represent audio recordings with the fixed duration of three seconds. These recordings are obtained 
via smartphone microphones and used to detect coughing \cite{CougDetection2019}. 

\subsection{Labels}
\label{sec_labels}

Besides its \gls{features}, a \gls{datapoint} might have a different kind of properties. These properties 
represent a higher-level fact or quantity of interest that is associated with the \gls{datapoint}. We 
refer to such properties of a \gls{datapoint} as its \index{label}\gls{label} (or ``output'' or ``target'') 
and typically denote it by $\truelabel$ (if it is a single number) or by $\mathbf{\truelabel}$ (if it is a vector 
of different label values, such as in \gls{multilabelclass}). We refer to the set of all possible label values of 
\gls{datapoint}s arising in a ML application is the \index{label space}\gls{labelspace} $\labelspace$. In general, 
determining the label of a \gls{datapoint} is more difficult (to automate) compared to determining its \gls{features}. 
Many ML methods revolve around finding efficient ways to \index{predict}predict (estimate or approximate) 
the label of a \gls{datapoint} based solely on its \gls{features}. 

The distinction of \gls{datapoint} properties into labels and features is blurry. Roughly speaking, labels are properties 
of \gls{datapoint}s that might only be determined with the help of human experts. For \gls{datapoint}s representing humans we 
could define its label $\truelabel$ as an indicator if the person has flu ($\truelabel=1$) or not ($\truelabel=0$). This 
label value can typically only be determined by a physician. However, in another application we might have 
enough resources to determine the flu status of any person of interest and could use it as a feature that 
characterizes a person. 

Consider a \gls{datapoint} that represents a hike, at the start of which the snapshot in Figure \ref{fig:image} has been 
taken. The features of this \gls{datapoint} could be the red, green and blue (RGB) intensities of each pixel in the 
snapshot in Figure \ref{fig:image}. We stack these RGB values into a vector $\featurevec \in \mathbb{R}^{\featurelen}$ 
whose length $\featurelen$ is three times the number of pixels in the image. The label $\truelabel$ associated with 
a \gls{datapoint} (a hike) could be the expected hiking time to reach the mountain in the snapshot. 
Alternatively, we could define the label $\truelabel$ as the water temperature of the lake that is depicted in the snapshot.

{\bf Numeric Labels - Regression.}
For a given ML application, the \gls{labelspace} $\labelspace$ contains all possible label values of 
\gls{datapoint}s. In general, the \gls{labelspace} is not just a set of different elements but also equipped 
(algebraic or geometric) structure. To obtain efficient ML methods, we should exploit such structure. 
Maybe the most prominent example for such a structured \gls{labelspace} are the real numbers $\labelspace = \mathbb{R}$. 
This \gls{labelspace} is useful for ML applications involving \gls{datapoint}s with numeric labels that can be 
modelled by real numbers. ML methods that aim at predicting a numeric label are referred to as regression methods. 

{\bf Categorical Labels - Classification.} Many important ML applications involve \gls{datapoint}s whose label indicate 
the category or class to which \gls{datapoint}s belongs to. ML methods that aim at predicting such categorical labels 
are referred to as \gls{classification} methods. Examples for classification problems include the diagnosis of 
tumours as benign or maleficent, the classification of persons into age groups or detecting the current floor 
conditions ( ``grass'', ``tiles'' or ``soil'') for a mower robot. 

The most simple type of a classification problems is a \index{binary classification} binary \gls{classification} problem. 
Within binary \gls{classification}, each \gls{datapoint} belongs to exactly one out of two different classes. 
Thus, the label of a \gls{datapoint} takes on values from a set that contains two different elements 
such as $\{0,1\}$ or $\{-1,1\}$ or $\{\mbox{``shows cat''}, \mbox{``shows no cat''}\}$. 

We speak of a \index{multi-class classification} multi-class \gls{classification} problem if \gls{datapoint}s belong to exactly one 
out of more than two categories (e.g., image categories ``no cat shown'' vs. ``one cat shown'' and ``more than one cat shown''). 
If there are $\nrcategories$ different categories we might use the label values $\{1,2,\ldots, \nrcategories\}$. 

There are also applications where \gls{datapoint}s can belong to several categories simultaneously. 
For example, an image can be cat image and a dog image at the same time if it contains a dog and a cat. 
\Gls{multilabelclass} \index{multi-label classification}problems and methods use several labels $\truelabel_{1},\truelabel_{2},\ldots,$ 
for different categories to which a \gls{datapoint} can belong to. The label $\truelabel_{\featureidx}$ 
represents the $\featureidx$th category and its value is $\truelabel_{\featureidx}=1$ if the \gls{datapoint} 
belongs to the $\featureidx$-th category and $\truelabel_{\featureidx}=0$ if not. 

{\bf Ordinal \Gls{label}s.} Ordinal label values are somewhat in between numeric and categorical labels. 
Similar to categorical labels, ordinal labels take on values from a finite set. Moreover, similar to numeric labels, 
ordinal labels take on values from an ordered set. For an example for such an ordered \gls{labelspace}, consider \gls{datapoint}s 
representing rectangular areas of size $1$ km by $1$ km. The features $\featurevec$ of such a \gls{datapoint} 
can be obtained by stacking the RGB pixel values of a satellite image depicting that area (see Figure \ref{fig_snapshot_pixels}). 
Beside the feature vector, each rectangular area is characterized by a label $\truelabel \in \{1,2,3\}$ where 
\begin{itemize} 
	\item $\truelabel=1$ means that the area contains no trees. 
	\item $\truelabel=2$ means that the area is partially covered by trees. 
	\item $\truelabel=3$ means that the are is entirely covered by trees. 
\end{itemize} 
Thus we might say that label value $\truelabel=2$ is ``larger'' than label value $\truelabel=1$ and label value $\truelabel=3$ is 
``larger'' than label value $\truelabel=2$. 

The distinction between regression and classification problems and methods is somewhat blurry. 
Consider a binary classification problem based on \gls{datapoint}s whose label $\truelabel$ takes on 
values $-1$ or $1$. We could turn this into a regression problem by using a new label $\truelabel'$ 
which is defined as the confidence in the label $\truelabel$ being equal to $1$. On the other hand, 
given a prediction $\hat{\truelabel'}$ for the numeric label $\truelabel' \in \mathbb{R}$ we can obtain a 
prediction $\hat{\truelabel}$ for the binary label $\truelabel \in \{-1,1\}$ by setting $\hat{\truelabel} \defeq 1$ 
if $\hat{\truelabel'} \geq0$ and $\hat{\truelabel}\defeq-1$ otherwise. A prominent example for this link 
between regression and classification is \gls{logreg} which is discussed in Section \ref{sec_LogReg}. 
\Gls{logreg} is a binary classification method that uses the same \gls{model} as \gls{linreg} but 
a different \gls{lossfunc}. 

We refer to a \gls{datapoint} as being \emph{labeled} if, besides its features $\featurevec$, the value of 
its label $\truelabel$ is known. The acquisition of labeled \gls{datapoint}s typically involves human labour, 
such as verifying if an image shows a cat. In other applications, acquiring labels might require sending 
out a team of marine biologists to the Baltic sea \cite{MLMarineBiology}, to run a particle physics 
experiment at the European organization for nuclear research (CERN) \cite{MLCERN}, or to conduct 
animal trials in pharmacology \cite{MLPharma}. 

Let us also point out online market places for human labelling workforce \cite{Mort2018}. These market places, 
allow to upload \gls{datapoint}s, such as collections of images or audio recordings, and then offer an hourly 
rate to humans that label the \gls{datapoint}s. This labeling work might amount to marking images that show a cat. 

Many applications involve \gls{datapoint}s whose features can be determined easily, but whose labels are 
known for few \gls{datapoint}s only. Labeled data is a scarce resource. Some of the most successful 
ML methods have been devised in application domains where label information can be acquired 
easily \cite{UnreasonableData}. ML methods for  speech recognition and machine translation can 
make use of massive labeled datasets that are freely available \cite{Koehn2005}. 

In the extreme case, we do not know the label of any single \gls{datapoint}. Even in the absence of any 
labeled data, ML methods can be useful for extracting relevant information from features only. We 
refer to ML methods which do not require any labeled \gls{datapoint}s as ``unsupervised'' ML methods. 
We discuss some of the most important unsupervised ML methods in Chapter \ref{ch_Clustering} 
and Chapter \ref{ch_FeatureLearning}).  

ML methods learn (or search for) a ``good'' predictor $h: \featurespace \rightarrow \labelspace$ 
which takes the features $\featurevec \in \featurespace$ of a \gls{datapoint} as its input and outputs a 
predicted label (or output, or target) $\hat{\truelabel} = h(\featurevec) \in \labelspace$. A good predictor 
should be such that $\hat{\truelabel} \approx \truelabel$, i.e., the predicted label $\hat{\truelabel}$ 
is close (with small error $\hat{\truelabel} - \truelabel$) to the true underlying label $\truelabel$. 

\subsection{Scatterplot} 
\label{equ_subsection_scatterplot}

Consider \gls{datapoint}s characterized by a single numeric feature $\feature$ and single numeric label $\truelabel$. 
To gain more insight into the relation between the features and label of a \gls{datapoint}, it can be instructive to 
generate a \gls{scatterplot} as shown in Figure \ref{fig_scatterplot_temp_FMI}. A \gls{scatterplot} depicts 
the \gls{datapoint}s $\datapoint^{(\sampleidx)}=(\feature^{(\sampleidx)},\truelabel^{(\sampleidx)})$ 
in a two-dimensional plane with the axes representing the values of feature $\feature$ and label $\truelabel$. 

The visual inspection of a \gls{scatterplot} might suggest potential relationships between 
feature $\feature$ (minimum daytime temperature) and label $\truelabel$ (maximum daytime temperature). 
From Figure \ref{fig_scatterplot_temp_FMI}, it seems that there might be a relation between 
feature $\feature$ and label $\truelabel$ since \gls{datapoint}s with larger $\feature$ tend to have 
larger $\truelabel$. This makes sense since having a larger minimum daytime temperature typically implies 
also a larger maximum daytime temperature. 

To construct a \gls{scatterplot} for \gls{datapoint}s with more than two 
features we can use feature learning methods (see Chapter \ref{ch_FeatureLearning}). 
These methods transform high-dimensional \gls{datapoint}s, having billions of raw features, 
to three or two new features. These new features can then be used as the coordinates of 
the \gls{datapoint}s in a \gls{scatterplot}.

\subsection{Probabilistic Models for Data} 
\label{equ_prob_models_data}
A powerful idea in ML is to interpret each \gls{datapoint}s as the realization of a \gls{rv}. 
For ease of exposition let us consider \gls{datapoint}s that are characterized by a single feature $\feature$. 
The following concepts can be extended easily to \gls{datapoint}s characterized by 
a feature vector $\featurevec$ and a label $\truelabel$. 

One of the most basic examples of a probabilistic model for \gls{datapoint}s in ML is the \gls{iid} assumption. 
This assumption interprets \gls{datapoint}s $\feature^{(1)},\ldots,\feature^{(\samplesize)}$ as realizations of statistically 
independent \gls{rv}s with the same \gls{probdist} $p(\feature)$. It might not be immediately clear why it is 
a good idea to interpret \gls{datapoint}s as realizations of \gls{rv}s with the common \gls{probdist} $p(\feature)$. 
However, this interpretation allows us to use the properties of the \gls{probdist} to characterize overall 
properties of entire datasets, i.e., large collections of \gls{datapoint}s. 

The \gls{probdist} $p(\feature)$ underlying the \gls{datapoint}s within the \gls{iidasspt} is either known 
(based on some domain expertise) or estimated from data. It is often enough to estimate only some parameters 
of the distribution $p(\feature)$. Section \ref{sec_max_iikelihood} discusses a principled approach to estimate 
the parameters of a \gls{probdist} from a given set of \gls{datapoint}s. This approach is sometimes referred to as maximum 
likelihood and aims at finding (parameter of) a \gls{probdist} $p(\feature)$ such that the probability (density) 
of observing the given \gls{datapoint}s is maximized \cite{LC,kay,BertsekasProb}.

Two of the most basic and widely used parameters of a \gls{probdist} $p(\feature)$ are the 
expected value or \index{mean}\gls{mean} \cite{BillingsleyProbMeasure}
$$\mu_{\feature} = \expect\{\feature\} \defeq \int_{\feature'} \feature' p(\feature') dx'$$ 
and the \index{variance}\gls{variance} 
$$\sigma_{\feature}^{2} \defeq \expect\big\{\big(\feature-\expect\{\feature\}\big)^{2}\big\}.$$ 
These parameters can be estimated using the sample mean (average) and sample variance, 
\begin{align} 
	\label{equ_sample_mean_var}
	\hat{\mu}_{\feature} & \defeq (1/\samplesize) \sum_{\sampleidx=1}^{\samplesize} \feature^{(\sampleidx)}\mbox{ , and } \nonumber \\
	\widehat{\sigma^{2}_{\feature}}&  \defeq (1/\samplesize) \sum_{\sampleidx=1}^{\samplesize} \big( \feature^{(\sampleidx)} - \hat{\mu}_{\feature} \big)^{2}.  
\end{align} 
The sample mean and sample variance \eqref{equ_sample_mean_var} are the \index{maximum likelihood}\gls{ml} 
estimators for the mean and variance of a normal (Gaussian) distribution $p(\feature)$ (see \cite[Chapter 2.3.4]{BishopBook}). 

Most of the ML methods discussed in this book are motivated by an \gls{iidasspt}. It is important to note that 
this \gls{iidasspt} is only a modelling assumption (or hypothesis). There is no means to verify if an arbitrary 
set of \gls{datapoint}s are ``exactly'' realizations of \gls{iid} \gls{rv}s. There are principled statistical 
methods (hypothesis tests) that allow to verify if a given set of \gls{datapoint} can be well approximated 
as realizations of \gls{iid} \gls{rv}s \cite{Luetkepol2005}. Alternatively, we can enforce the \gls{iidasspt} if 
we generate synthetic data using a random number generator. Such synthetic \gls{iid} \gls{datapoint}s 
could be obtained by sampling algorithms that incrementally build a synthetic dataset by adding 
randomly chosen raw \gls{datapoint}s \cite{Efron97}.

\section{The Model}
\label{sec_hypo_space}

Consider some ML application that generates \gls{datapoint}s, each characterized by features $\featurevec \in \featurespace$ 
and label $\truelabel \in \labelspace$. The goal of a ML method is to learn a 
hypothesis map $h: \featurespace \rightarrow \labelspace$ such that 
\begin{equation} 
	\label{equ_approx_label_pred}
	\truelabel \approx \underbrace{h(\featurevec)}_{\hat{\truelabel}} \mbox{ for any \gls{datapoint}}. 
\end{equation}  
The informal goal \eqref{equ_approx_label_pred} will be made precise in several aspects throughout the 
rest of our book. First, we need to quantify the approximation error \eqref{equ_approx_label_pred} incurred by a given 
hypothesis map $h$. Second, we need to make precise what we actually mean by requiring \eqref{equ_approx_label_pred} to hold 
for ``any'' \gls{datapoint}. We solve the first issue by the concept of a loss function in Section \ref{sec_lossfct}. 
The second issue is then solved in Chapter \ref{ch_Optimization} by using a simple probabilistic model 
for data. 

\begin{figure}[htbp]
	\begin{minipage}{\textwidth}
		\begin{center}
			\includegraphics[width=0.7\textwidth]{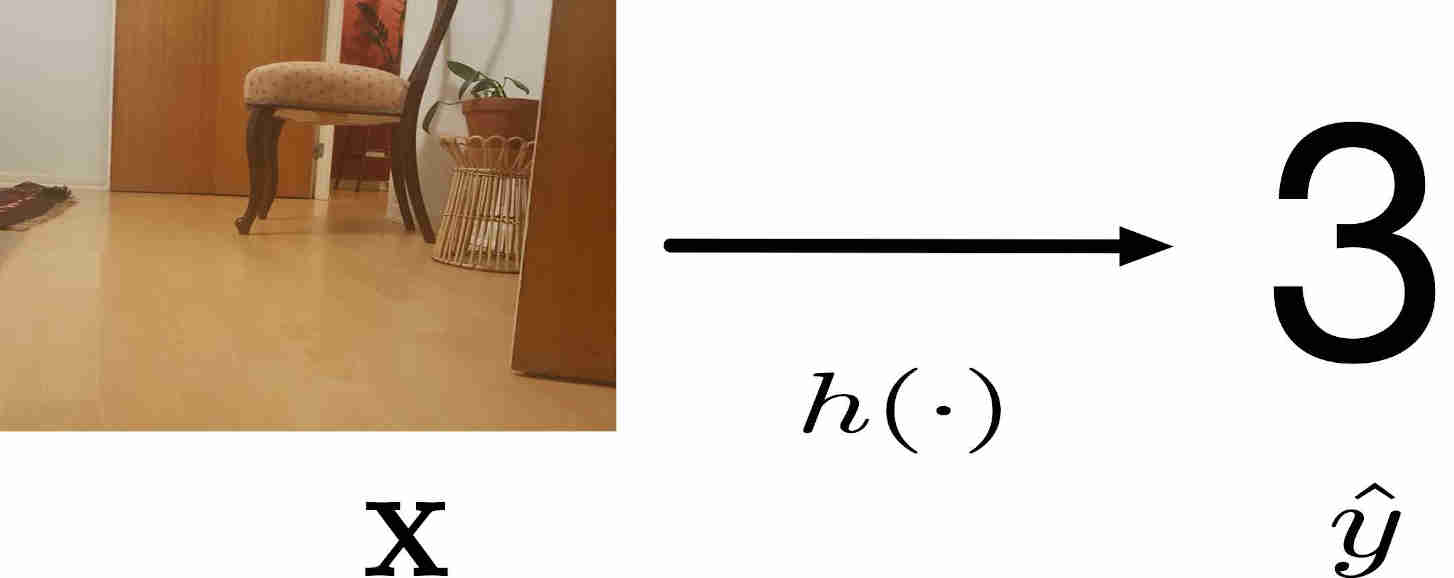}  
		\end{center}
	\end{minipage}
	\caption{A hypothesis (predictor) $h$ maps features $\featurevec\!\in\!\featurespace$, of an on-board 
		camera snapshot, to the prediction $\hat{\truelabel}\!=\! h(\featurevec)\!\in\!\labelspace$ for the coordinate of 
		the current location of a cleaning robot. ML methods use data to learn predictors $h$ 
		such that $\hat{\truelabel}\!\approx\!\truelabel$ (with true label $\truelabel$).}
	\label{fig_feature_map_eval}
\end{figure}

\begin{figure}[htbp]
	\centering
	\includegraphics[width=0.7\textwidth]{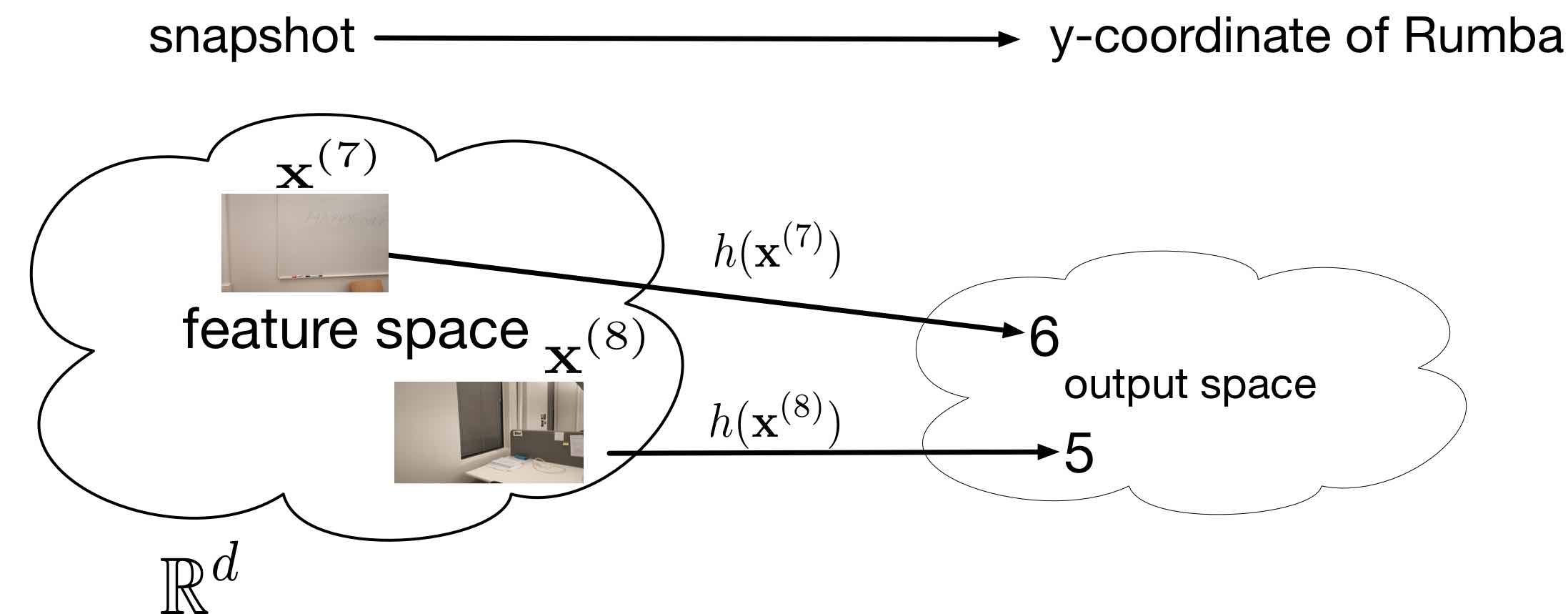}  
	\caption{
		Consider a hypothesis $h: \featurespace \rightarrow \labelspace$ that is used for locating a cleaning robot. 
		The hypothesis $h$ reads in the feature vector $\featurevec^{(\timeidx)} \in \featurespace$, that might be 
		RGB pixel intensities of an on-board camera snapshot, at time $\timeidx$. It then outputs a prediction  $\hat{\truelabel}^{(\timeidx)}=h(\featurevec^{(\timeidx)} )$ for the $\truelabel$-coordinate $\truelabel^{(\timeidx)}$ 
		of the cleaning robot at time $\timeidx$. A key problem studied within ML is how to automatically learn 
		a good (accurate) predictor $h$ such that $\truelabel^{(\timeidx)} \approx h(\featurevec^{(\timeidx)} )$. }
	\label{fig:Hypothesis Map}
\end{figure}

Let us assume for the time being that we have found a reasonable hypothesis $h$ in 
the sense of \eqref{equ_approx_label_pred}. We can then use this hypothesis to predict the 
label of any \gls{datapoint} for which we know its features. The prediction $\hat{\truelabel}=h(\featurevec)$ 
is obtained by evaluating the hypothesis for the features $\featurevec$ of a \gls{datapoint} (see Figure \ref{fig_feature_map_eval} and \ref{fig:Hypothesis Map}). 
We refer to a hypothesis map also as a predictor map since it is used 
or compute the prediction $\hat{\truelabel}$ of a (true) label $\truelabel$.

For ML problems using a finite \gls{labelspace} $\labelspace$ (e..g, $\labelspace=\{-1,1\}$, we refer to a hypothesis also 
as a \gls{classifier}. For a finite \label{labelspace} $\labelspace$, we can characterize a particular classifier map $h$ 
using its different \gls{decisionregion}s
\begin{equation} 
	\label{equ_decision_region}
	\decreg{a} \defeq \big\{ \featurevec \in \mathbb{R}^{\featuredim}: h(\featurevec) = a \big\} \subseteq \featurespace. 
\end{equation}
Each label value $a \in \labelspace$ is associated with a specific \gls{decisionregion} $\decreg{a}$.
For a given label value $a \in \labelspace$, the decision region $\decreg{a}$ is constituted by all 
feature vectors $\featurevec \in \featurespace$ which are mapped to this label value, $h(\featurevec)=a$. 

In principle, ML methods could use any possible map $h: \featurespace \rightarrow \labelspace$ 
to predict the label $\truelabel \in \labelspace$ via computing $\hat{\truelabel} = h(\featurevec)$. 
The set of all maps from the feature space $\featurespace$ to the \gls{labelspace} is typically  
denoted as $\labelspace^{\featurespace}$.\footnote{The notation $\labelspace^{\featurespace}$ is to be understood 
	as a symbolic shorthand and should not be understood literately as a power such as $4^5$.} 
In general, the set $\labelspace^{\featurespace}$ is way too large to be search over by a practical ML methods. 
As a point in case, consider \gls{datapoint}s characterized by a single numeric feature $\feature \in \mathbb{R}$ and 
label $\truelabel \in \mathbb{R}$. The set of all real-valued maps $h(\feature)$ of a real-valued argument already 
contains uncountably infinite many different hypothesis maps \cite{HalmosSet}. 

Practical ML methods can search and evaluate only a (tiny) subset of all possible hypothesis maps. This subset of 
computationally feasible (``affordable'') hypothesis maps is referred to as the \gls{hypospace} or \gls{model} 
underlying a ML method. As depicted in Figure \ref{fig_hypo_space}, ML methods typically use a \gls{hypospace} $\hypospace$ 
that is a tiny subset of $\labelspace^{\featurespace}$. Similar to the features and labels used to characterize 
\gls{datapoint}s, also the \gls{hypospace} underlying a ML method is a design choice. As we will see, 
the choice for the \gls{hypospace} involves a trade-off between computational complexity and 
statistical properties of the resulting ML methods. 

The preference for a particular \gls{hypospace} often depends on the computational infrastructure that is 
available to a ML method. Different computational infrastructures favour different \gls{hypospace}s. ML 
methods implemented in a small embedded system, might prefer a linear \gls{hypospace} which results in 
algorithms that require a small number of arithmetic operations. Deep learning methods implemented in a 
cloud computing environment typically use much larger \gls{hypospace}s obtained from large \gls{ann} (see Section \ref{sec_deep_learning}). 

ML methods can also be implemented using a spreadsheet software. Here, we might use a \gls{hypospace} consisting 
of maps $h: \featurespace \rightarrow \labelspace$ that are represented by look up tables (see Table \ref{table_lookup}). 
If we instead use the programming language Python to implement a ML method, we can obtain a hypothesis class by 
collecting all possible Python subroutines with one input (scalar feature $\feature$), one output 
argument (predicted label $\hat{\truelabel}$) and consisting of less than $100$ 
lines of code. 

\vspace*{2mm}
\begin{table}
	\begin{center}
		\begin{tabular}{ |c|c| }
			\multicolumn{1}{c}{feature $x$}
			&  \multicolumn{1}{c}{prediction $h(x)$} \\
			\cline{1-2}
			0 & 0 \\
			\cline{1-2}
			1/10 & 10 \\
			\cline{1-2}
			2/10 & 3 \\
			\cline{1-2}
			\vdots & \vdots \\
			\cline{1-2}
			1 & 22.3 \\
			\cline{1-2}
		\end{tabular}
	\end{center}
	\caption{	\label{table_lookup} A look-up table defines a hypothesis map $h$. The value $h(\feature)$ is given
		by the entry in the second column of the row whose first column entry is $\feature$. 
		We can construct a \gls{hypospace} $\hypospace$ by using a collection of different look-up tables.}
\end{table}
\vspace*{2mm}

Broadly speaking, the design choice for the \gls{hypospace} $\hypospace$ of a ML method 
has to balance between two conflicting requirements.  
\begin{itemize} 
	\item It has to be sufficiently large such that it contains at least 
	one accurate predictor map $\hat{h} \in \hypospace$. A hypothesis 
	space $\hypospace$ that is too small might fail to include a predictor 
	map required to reproduce the (potentially highly non-linear) relation 
	between features and label. 
	
	Consider the task of grouping or classifying images into ``cat'' images and ``no cat image''. The classification of 
	each image is based solely on the feature vector obtained from the pixel colour intensities. The relation between 
	features and label ($\truelabel \in \{ \mbox{cat}, \mbox{no cat} \}$) is highly non-linear. Any ML method that uses a 
	\gls{hypospace} consisting only of linear maps will most likely fail to learn a good predictor (classifier). 
	We say that a ML method is \index{underfitting} underfitting if it uses a \gls{hypospace} that does not contain 
	any hypotheses maps that can accurately predict the label of any \gls{datapoint}s. 
	\item It has to be sufficiently small such that its processing fits the available computational resources (memory, bandwidth, processing time). 
	We must be able to efficiently search over the \gls{hypospace} to find good predictors (see Section \ref{sec_lossfct} and Chapter \ref{ch_Optimization}). This requirement implies also that the maps $h(\featurevec)$ contained in $\hypospace$ 
	can be evaluated (computed) efficiently \cite{Austin2018}. Another important reason for using a \gls{hypospace} $\hypospace$ 
	that is not too large is to avoid overfitting (see Chapter \ref{ch_overfitting_regularization}). If the \gls{hypospace} $\hypospace$ 
	is too large, then we can easily find a hypothesis which (almost) perfectly predicts the labels of \gls{datapoint}s in a \gls{trainset} 
	which is used to learn a hypothesis. However, such a hypothesis might deliver poor predictions for labels of \gls{datapoint}s outside 
	the \gls{trainset}. We say that the hypothesis does not \index{generalization}generalize well. 
\end{itemize}

\subsection{Parametrized \Gls{hypospace}s}
A wide range of current scientific computing environments allow for efficient numerical linear algebra. 
This hard- and software allows to efficiently process data that is provided in the form of numeric 
arrays such as vectors, matrices or tensors \cite{JMLR:v12:pedregosa11a}. To take advantage of 
such computational infrastructure, many ML methods use the \gls{hypospace}  
\begin{align}
	\label{equ_def_hypo_linear_pred}
	\hspace*{-3mm}\hypospace^{(\featuredim)} \!\defeq\!\{ h^{(\weights)}\!:\!\mathbb{R}^{\featuredim}\!\rightarrow\!\mathbb{R}: h^{(\weights)}(\featurevec)\!\defeq\!\featurevec^{T} \weights \mbox{ with some parameter vector } \weights\!\in\!\mathbb{R}^{\featuredim} \}.  
\end{align}
The \gls{hypospace} \eqref{equ_def_hypo_linear_pred} is constituted by linear maps (functions) 
\begin{equation} 
	\label{equ_def_linear_map}
	h^{(\weights)}\big(\featurevec\big): \mathbb{R}^{\featuredim} \rightarrow \mathbb{R}: \featurevec \mapsto \weights^{T} \featurevec.
\end{equation}  
The function $h^{(\weights)}$ \eqref{equ_def_linear_map} maps, in a linear fashion, the feature vector 
$\featurevec \in \mathbb{R}^{\featuredim}$ to the predicted \gls{label} $h^{(\weights)}(\featurevec) = \featurevec^{T} \weights \in \mathbb{R}$. 
For $\featuredim\!=\!1$ the feature vector reduces a single feature $\feature$ and the hypothesis 
space \eqref{equ_def_hypo_linear_pred} consists of all maps $h^{(\weight)}(\feature) = \weight \feature$ 
with weight $\weight \in \mathbb{R}$ (see Figure \ref{scalar_lin_space}).

\begin{figure}[htbp]
	\begin{center}
		\begin{tikzpicture}
			\begin{axis}
				[ylabel=$h(x)$,scale=0.9,
				xlabel=$x$,
				axis x line=center,
				axis y line=center,
				xtick={1},
				ytick={1},
				xlabel={feature $\feature$},
				ylabel={$h^{(\weight)}(\feature)$},
				xlabel style={right},
				ylabel style={above},
				xmin=-1.4,
				xmax=3.5,
				ymin=-1.1,
				ymax=1.1
				]
				\addplot [smooth, color=red, ultra thick] table [x=a, y=b, col sep=comma] {linear.csv} node [right,color=red] {$h^{(1)}(\feature)\!=\!\feature$};
				\addplot [color=blue, ultra thick] table [x=a, y=c, col sep=comma] {linear.csv} node [right,color=blue] {$h^{(0.2)}(\feature)\!=\!0.2 \feature$}; 
				\addplot [color=black, ultra thick] table [x=a, y=d, col sep=comma] {linear.csv}node [right,color=black] {$h^{(0.7)}(\feature)\!=\!0.7 \feature$}; 
			\end{axis}
		\end{tikzpicture}
		\vspace*{-4mm}
	\end{center}
	\caption{Three particular members of the \gls{hypospace} 
		$\hypospace =\{ h^{(\weight)}: \mathbb{R} \rightarrow \mathbb{R}, h^{(\weight)}(\feature) = \weight  \feature \}$ 
		which consists of all linear functions of the scalar feature $x$. We can parametrize 
		this \gls{hypospace} conveniently using the weight $ \weight \in \mathbb{R}$ as $h^{(\weight)}(\feature) = \weight  \feature$.}
	\label{scalar_lin_space}
\end{figure}

The elements of the \gls{hypospace} $\hypospace$ in \eqref{equ_def_hypo_linear_pred} are  
parametrized by the parameter vector $\weights \in \mathbb{R}^{\featuredim}$. Each map $h^{(\weights)} \in \hypospace$ 
is fully specified by the parameter vector $\weights \in \mathbb{R}^{\featuredim}$. This parametrization 
of the \gls{hypospace} $\hypospace$ allows to process and manipulate hypothesis maps 
by vector operations. In particular, instead of searching over the function space $\hypospace$ 
(its elements are functions!) to find a good hypothesis, we can equivalently search over all 
possible parameter vectors $\weights \in \mathbb{R}^{\featuredim}$. 

The search space $\mathbb{R}^{\featuredim}$ is still (uncountably) infinite but it has a rich 
geometric and algebraic structure that allows to efficiently search over this space. Chapter \ref{ch_GD} 
discusses methods that use the concept ot a \gls{gradient} to implement an efficient 
search for useful parameter vectors $\weights \in \mathbb{R}^{\featurelen}$. 

The \gls{hypospace} \eqref{equ_def_hypo_linear_pred} is also appealing because 
of the broad availability of computing hardware such as graphic processing units. 
Another factor boosting the widespread use of \eqref{equ_def_hypo_linear_pred} 
might be the offer for optimized software libraries for numerical linear algebra. 

The \gls{hypospace} \eqref{equ_def_hypo_linear_pred} can also be used for classification problems, e.g., 
with \gls{labelspace} $\labelspace = \{-1,1\}$. Indeed, given a linear predictor map $h^{(\weights)}$ we can classify data 
points according to $\hat{\truelabel}\!=\!1$ for $h^{(\weights)}(\featurevec) \geq 0$ and $\hat{\truelabel}\!=\!-1$ otherwise. 
We refer to a classifier that computes the predicted label by first applying a linear map to 
the features as a \index{linear classifier}\gls{linclass}. 

Figure \ref{fig_lin_dec_boundary} illustrates the decision regions \eqref{equ_decision_region} of a \gls{linclass} for binary labels.  
The decision regions are half-spaces and, in turn, the \gls{decisionboundary} is a hyperplane $\{ \featurevec: \weights^{T} \featurevec = b \}$. Note that 
each \gls{linclass} corresponds to a particular linear hypothesis map from the \gls{hypospace} \eqref{equ_def_hypo_linear_pred}. 
However, we can use different loss functions to measure the quality of a \gls{linclass}. Three widely-used examples for ML methods 
that learn a \gls{linclass} are \gls{logreg} (see Section \ref{sec_LogReg}), the \gls{svm} (see Section \ref{sec_SVM}) 
and the naive Bayes classifier (see Section \ref{sec_NaiveBayes}). 
\begin{figure} 
	\begin{center}
		\begin{tikzpicture}[allow upside down, scale=1]
			\draw[dashed,line width=3pt] (0,0) -- (10,5)  [smooth, tension=0.5]
			node[sloped,inner sep=0cm,above,pos=.5,
			anchor=south west,
			minimum height=2cm,minimum width=1cm](N){adf};
			\path [fill=yellow] (0,0) -- (0,5) -- (10,5) -- (0,0); 
			\node [below] at (5,1) {$h(\featurevec)<0, \hat{\truelabel} = -1$};
			\node [below] at (9,3.7) {\gls{decisionboundary}};
			\node [below] at (2,4) {$h(\featurevec)\geq0, \hat{\truelabel} = 1$};
			\path (N.south west)
			edge[->,blue,thick] node[left] {$\weights$} (N.north west);
		\end{tikzpicture}
	\end{center}
	\caption{A hypothesis $h: \featurespace\!\rightarrow\!\labelspace$ for a binary classification problem, 
		with \gls{labelspace} $\labelspace=\{ -1,1\}$ and feature space $\featurespace=\mathbb{R}^{2}$, can be 
		represented conveniently via the \gls{decisionboundary} (dashed line) which separates all feature vectors $\featurevec$ 
		with $h(\featurevec) \geq 0 $ from the region of feature vectors with $h(\featurevec)<0$. If the \gls{decisionboundary} 
		is a hyperplane $\{ \featurevec : \weights^{T} \featurevec = b \}$ (with normal vector $\weights \in \mathbb{R}^{\featuredim}$), we refer 
		to the map $h$ as a \gls{linclass}.}
	\label{fig_lin_dec_boundary}
\end{figure}

In some application domains, the relation between features $\featurevec$ and label $\truelabel$ 
of a \gls{datapoint} is highly non-linear. As a case in point, consider \gls{datapoint}s that are images 
of animals. The map that relates the pixel intensities of an image to a label value indicating if the image shows 
a cat is highly non-linear. For such applications, the \gls{hypospace} \eqref{equ_def_hypo_linear_pred}  
is not suitable as it only contains linear maps. The second main example for a parametrized \gls{hypospace} 
studied in this book contains also non-linear maps. This parametrized \gls{hypospace} is obtained from a 
parametrized signal flow diagram which is referred to as an \gls{ann}. Section \ref{sec_deep_learning} 
discusses the construction of non-linear parametrized hypothesis spaces using an \gls{ann}. 

{\bf Upgrading a Hypothesis Space via Feature Maps.}
Let us discuss a simple but powerful technique for enlarging (``upgrading'') a given \gls{hypospace} $\hypospace$ 
to a larger \gls{hypospace} $\hypospace' \supseteq \hypospace$ that offers a wider selection of hypothesis maps. 
The idea is to replace the original features $\featurevec$ of a \gls{datapoint} with new (transformed) 
features $\rawfeaturevec = \featuremapvec (\featurevec)$. The transformed features are obtained by 
applying a \index{feature map}feature map $\featuremap(\cdot)$ to the original features $\featurevec$. 
This upgraded \gls{hypospace} $\hypospace'$ consists of all concatenations of the feature map $\featuremap$ and 
some hypothesis $h \in \hypospace$, 
\begin{equation} 
	\label{equ_def_enlarged_hypospace}
	\hypospace' \defeq \big\{ h'(\featurevec) \defeq h\big( \featuremapvec (\featurevec) \big) : h \in \hypospace\big\}. 
\end{equation}
The construction \eqref{equ_def_enlarged_hypospace} used for arbitrary combinations of a feature map $\featuremap(\cdot)$ 
and a ``base'' \gls{hypospace} $\hypospace$. The only requirement is that the output of the feature map 
can be used as input for a hypothesis $h \in \hypospace$. More formally, the range of the feature map 
must belong to the domain of the maps in $\hypospace$. Examples for ML methods that use a \gls{hypospace} 
of the form \eqref{equ_def_enlarged_hypospace} include polynomial regression (see Section \ref{sec_polynomial_regression}), 
Gaussian basis regression (see Section \ref{sec_linbasreg}) and the important family of kernel methods (see Section \ref{sec_kernel_methods}). 
The feature map in \eqref{equ_def_enlarged_hypospace} might also be obtained from \index{clustering}\gls{clustering} or feature 
learning methods (see Section \ref{sec_clust_preproc} and Section \ref{sec_comb_PcA_Linreg}).

For the special case of the linear \gls{hypospace} \eqref{equ_def_hypo_linear_pred}, the resulting enlarged 
\gls{hypospace} \eqref{equ_def_enlarged_hypospace} is given by all linear maps $\weights^{T} \rawfeaturevec$ 
of the transformed features $\featuremapvec(\featurevec)$. Combining the \gls{hypospace} \eqref{equ_def_hypo_linear_pred} with a non-linear 
feature map results in a \gls{hypospace} that contains non-linear maps from the original feature vector $\featurevec$ 
to the predicted label $\hat{\truelabel}$, 
\begin{equation}
	\label{equ_conc_featuremap_linear}
	\hat{\truelabel} = \weights^{T} \rawfeaturevec = \weights^{T} \featuremapvec (\featurevec). 
\end{equation}

{\bf Non-Numeric Features.}
The \gls{hypospace} \eqref{equ_def_hypo_linear_pred} can only be used for \gls{datapoint}s whose features are numeric 
vectors $\featurevec = (\feature_{1},\ldots,\feature_{\featuredim})^{T} \in \mathbb{R}^{\featuredim}$. In some application domains, such as 
natural language processing, there is no obvious natural choice for numeric features. However, since ML methods 
based on the \gls{hypospace} \eqref{equ_def_hypo_linear_pred} are well developed (using numerical linear algebra), 
it might be useful to construct numerical features even for non-numeric data (such as text). For text data, there has 
been significant progress recently on methods that map a human-generated text into sequences of vectors 
(see \cite[Chap. 12]{Goodfellow-et-al-2016} for more details). Moreover, Section \ref{sec_discrete_embeddings} will 
discuss an approach to generate numeric features for \gls{datapoint}s that have an intrinsic notion of similarity. 

\subsection{The Size of a Hypothesis Space}
The notion of a \gls{hypospace} being too small or being too large can be made precise in different ways. The size of 
a finite \gls{hypospace} $\hypospace$ can be defined as its cardinality $|\hypospace|$ which is simply the number of 
its elements. For example, consider \gls{datapoint}s represented by $100 \times 10\!=\!1000$ black-and-white pixels and 
characterized by a binary label $\truelabel \in \{0,1\}$. We can model such \gls{datapoint}s using the \gls{featurespace} $\featurespace = \{0,1\}^{1000}$ 
and \gls{labelspace} $\labelspace = \{0,1\}$. The largest possible \gls{hypospace} $\hypospace = \labelspace^\featurespace$ 
consists of all maps from $\featurespace$ to $\labelspace$. The size or cardinality of this space is $|\hypospace| = 2^{2^{1000}}$.

Many ML methods use a \gls{hypospace} which contains infinitely many different predictor maps (see, e.g., \eqref{equ_def_hypo_linear_pred}). For an infinite \gls{hypospace}, we cannot use the number of its elements as a measure for its size. Indeed, for an 
infinite \gls{hypospace}, the number of elements is not well-defined. Therefore, we measure the size of a 
\gls{hypospace} $\hypospace$ using its \index{effective dimension}\gls{effdim} $\effdim{\hypospace}$. 

Consider a \gls{hypospace} $\hypospace$ consisting of maps $h: \featurespace\rightarrow \labelspace$ that read in 
the features $\featurevec \in \featurespace$ and output an predicted label $\hat{\truelabel} = h(\featurevec) \in \labelspace$. 
We define the \gls{effdim} $\effdim{\hypospace}$ of $\hypospace$ as the maximum number $\sizehypospace \in \mathbb{N}$ such that for any set $\dataset = \big\{ \big(\featurevec^{(1)},\truelabel^{(1)}\big), \ldots, \big(\featurevec^{(\sizehypospace)},\truelabel^{(\sizehypospace)}\big) \}$ of $\sizehypospace$ \gls{datapoint}s with 
different features, we can always find a hypothesis $h \in \hypospace$ that perfectly fits the labels, $\truelabel^{(\sampleidx)} = h\big( \featurevec^{(\sampleidx)} \big)$ for $\sampleidx=1,\ldots,\sizehypospace$. 

The \gls{effdim} of a \gls{hypospace} is closely related to the \gls{vcdim} \cite{VapnikBook}. 
The \gls{vcdim} is maybe the most widely used concept for measuring the size of infinite \gls{hypospace}s  \cite{VapnikBook,ShalevMLBook,BishopBook,hastie01statisticallearning}. 
However, the precise definition of the \gls{vcdim} are beyond the scope of this book. Moreover, the 
\gls{effdim} captures most of the relevant properties of the \gls{vcdim} for our 
purposes. For a precise definition of the \gls{vcdim} and discussion of its applications 
in ML we refer to \cite{ShalevMLBook}. Let us illustrate the concept of \gls{effdim} as a measure 
for the size of a \gls{hypospace} with two examples: \gls{linreg} and polynomial regression. 

\Gls{linreg} uses the \gls{hypospace} 
$$\hypospace^{(\featuredim)} = \{ h: \mathbb{R}^{\featuredim} \rightarrow \mathbb{R}: h(\featurevec) = \weights^{T} \featurevec \mbox{ with some vector } \weights \in \mathbb{R}^{\featurelen}\}.$$ 
Consider a dataset $\dataset=\big\{ \big(\featurevec^{(1)} ,\truelabel^{(1)}\big),\ldots, \big(\featurevec^{(\samplesize)} ,\truelabel^{(\samplesize)}\big) \big\} $ consisting of $\samplesize$ \gls{datapoint}s. We refer to this 
number also as the \gls{samplesize} of the dataset.  Each \gls{datapoint} is characterized by a feature 
vector $\featurevec^{(\sampleidx)} \in \mathbb{R}^{\featuredim}$ and a numeric label 
$\truelabel^{(\sampleidx)} \in \mathbb{R}$. Let us further assume that \gls{datapoint}s are 
realizations of \gls{iid} \gls{rv}s with a common \gls{probdist}. 

Under the \gls{iidasspt}, the matrix $$\featuremtx = \big(\featurevec^{(1)},\ldots,\featurevec^{(\samplesize)}\big) \in \mathbb{R}^{\featuredim\times \samplesize},$$ 
which is obtained by stacking (column-wise) the feature vectors $\featurevec^{(\sampleidx)}$ 
(for $\sampleidx=1,\ldots,\samplesize$), is full (column-) rank with probability one. Basic results of 
linear algebra allow to show that the \gls{datapoint}s in $\dataset$ can be perfectly fit by a linear 
map $h \in \hypospace^{(\featuredim)}$ as long as $\samplesize \leq \featuredim$. 

As long as the number $\samplesize$ of \gls{datapoint}s does not exceed the number of features 
characterizing each \gls{datapoint}, i.e.,  as long as $\samplesize \leq \featuredim$, we can find (with probability one) 
a parameter vector $\widehat{\weights}$ such that $\truelabel^{(\sampleidx)} = \widehat{\weights}^{T} \featurevec^{(\sampleidx)}$ 
for all $\sampleidx=1,\ldots,\samplesize$. The \gls{effdim} of the linear \gls{hypospace} $\hypospace^{(\featuredim)}$ 
is therefore $\sizehypospace = \featuredim$. 

As a second example, consider the \gls{hypospace} $\hypospace_{\rm poly}^{(\featuredim)}$ which is constituted 
by the set of polynomials with maximum degree $\featuredim$. The fundamental theorem of algebra tells us that any set 
of $\samplesize$ \gls{datapoint}s with different features can be perfectly fit by a polynomial of degree $\featuredim$ 
as long as $\featuredim \geq \samplesize$. Therefore, the \gls{effdim} of the \gls{hypospace} $\hypospace_{\rm poly}^{(\featuredim)}$ 
is $\sizehypospace=\featuredim$. Section \ref{sec_polynomial_regression} discusses polynomial regression in more detail. 

\begin{figure}[htbp]
	\begin{center}
		\begin{tikzpicture}[allow upside down, scale=0.4]
			\node [below] at (5,-3) {$\labelspace^{\featurespace}$};
			\draw [ultra thick] (5,0) circle (5cm);
			\draw [ultra thick,fill=black!20] (5,0) circle (1cm);
			\node [] at (5,0) {$\hypospace$};
		\end{tikzpicture}
	\end{center}
	\caption{The \index{hypothesis space}\gls{hypospace} $\hypospace$ is a (typically very small) subset 
		of the (typically very large) set $\labelspace^{\featurespace}$ of all possible maps from feature 
		space $\featurespace$ into the label space $\labelspace$.}
	\label{fig_hypo_space}
\end{figure}
		
\section{The Loss}
\label{sec_lossfct}
		
		Every ML method uses a (more of less explicit) \gls{hypospace} $\hypospace$ which consists of all 
		computationally feasible predictor maps $h$. Which predictor map $h$ out of all the maps 
		in the \gls{hypospace} $\hypospace$ is the best for the ML problem at hand? To answer this questions, 
		ML methods use the concept of a \gls{lossfunc}. Formally, a \gls{lossfunc} is a map 
		$$\lossfun: \featurespace \times \labelspace \times \hypospace \rightarrow \mathbb{R}_{+}: \big( \big(\featurevec,\truelabel\big), h\big) \mapsto  \loss{(\featurevec,\truelabel)}{h}$$ which assigns a pair consisting of a \gls{datapoint}, itself characterized by 
		features $\featurevec$ and label $\truelabel$, and a hypothesis $h \in \hypospace$ the non-negative real number $\loss{(\featurevec,\truelabel)}{h}$. 
		
		\hspace*{10mm}
		\begin{figure}[htbp]
			\begin{center}
				\begin{tikzpicture}
					\begin{axis}
						[
						axis x line=center,
						axis y line=center,
						xlabel={hypothesis $h$},
						xlabel style={above right},
						ylabel style={above right},
						xtick=\empty,
						ytick=\empty,
						xmin=-4,
						xscale = 1.4, 
						xmax=4,
						ymin=-0.5,
						ymax=2.5
						]
						\addplot [smooth, ultra thick] table [x=a, y=b, col sep=comma] {logloss.csv};    
					\end{axis}
					\node [above] at (1,5) {$\loss{(\featurevec,\truelabel)}{h}$};
				\end{tikzpicture}
			\end{center}
			\vspace*{-7mm}
			\caption{Some \gls{lossfunc} $\loss{(\featurevec,\truelabel)}{h}$ for a fixed \gls{datapoint}, with features $\featurevec$ and label $\truelabel$, 
				and varying hypothesis $h$. ML methods try to find (learn) a hypothesis that incurs minimum loss.}
			\label{fig_loss_function}
		\end{figure}

		
		The loss value $\loss{(\featurevec,\truelabel)}{h}$ quantifies the discrepancy between the true label $\truelabel$ 
		and the predicted label $h(\featurevec)$. A small (close to zero) value $\loss{(\featurevec,\truelabel)}{h}$ indicates  
		a low discrepancy between predicted \gls{label} and true label of a \gls{datapoint}. Figure \ref{fig_loss_function} depicts 
		a loss function for a given \gls{datapoint}, with features $\featurevec$ and label $\truelabel$, as a function of the 
		hypothesis $h \in \hypospace$. The basic principle of ML methods can then be formulated as: Learn (find) a 
		hypothesis out of a given \gls{hypospace} $\hypospace$ that incurs a minimum loss $\loss{(\featurevec,\truelabel)}{h}$ 
		for any \gls{datapoint} (see Chapter \ref{ch_Optimization}). 
		
		Much like the choice for the \gls{hypospace} $\hypospace$ used in a ML method, also the loss function is a 
		design choice. We will discuss some widely used examples for \gls{lossfunc} in Section \ref{sec_loss_numeric_label} 
		and Section \ref{sec_loss_categorical}. The choice for the \gls{lossfunc} should take into account the computational 
		complexity of searching the \gls{hypospace} for a hypothesis with minimum loss. Consider a ML method that uses 
		a parametrized \gls{hypospace} and a \gls{lossfunc} that is a convex and \index{differentiable} 
		differentiable (smooth) function of the parameters of a hypothesis. In this case, searching for a hypothesis with small \gls{loss} 
		can be done efficiently using the \gls{gdmethods} discussed in Chapter \ref{ch_GD}. The minimization of a \gls{lossfunc} 
		that is either non-convex or non-differentiable is typically computationally much more difficult. Section \ref{sec_comp_stat_ERM} 
		discusses the computational complexities of different types of \gls{lossfunc}s in more detail.
		
		Beside computational aspects, the choice for the \gls{lossfunc} should also take into account statistical aspects. 
		For example, some \gls{lossfunc}s result in ML methods that are more robust against \gls{outlier}s (see Section \ref{sec_lad} and 
		Section \ref{sec_SVM}). The choice of \gls{lossfunc} might also be guided by probabilistic models for the 
		data generated in an ML application. Section \ref{sec_max_iikelihood} details how the \gls{ml} principle of 
		statistical inference provides an explicit construction of \gls{lossfunc}s in terms of an (assumed) 
		\gls{probdist} for \gls{datapoint}s. 
		
		The choice for the \gls{lossfunc} used to evaluate the quality of a hypothesis might also be influenced by its 
		interpretability. Section \ref{sec_loss_categorical} discusses loss functions for hypotheses that are used to 
		classify \gls{datapoint}s into two categories. It seems natural to measure the quality of such a hypothesis 
		by the average number of wrongly classified \gls{datapoint}s, which is precisely the average $0/1$ \gls{loss} \eqref{equ_def_0_1} 
		(see Section \ref{sec_loss_categorical}). 
		
		In contrast to its appealing interpretation as error-rate, the computational aspects of the average $0/1$ \gls{loss} are 
		less pleasant. Minimizing the average $0/1$ loss to learn an accurate hypothesis amounts to a non-convex 
		and \gls{nonsmooth} optimization problem which is computationally challenging. Section \ref{sec_loss_categorical} 
		introduces the \gls{logloss} as a computationally attractive alternative choice 
		for the \gls{lossfunc} in binary \gls{classification} problems. Learning a hypothesis that minimizes a (average) \gls{logloss} 
		amounts to a smooth convex optimization problem. Chapter \ref{ch_GD} discusses computationally cheap \gls{gdmethods} 
	    for solving smooth convex optimization problems. 
		
		The above aspects (computation, statistic, interpretability) result typically in conflicting goals for the choice 
		of a loss function. A \gls{lossfunc} that has favourable statistical properties might incur a high computational 
		complexity of the resulting ML method. \Gls{lossfunc}s that result in computationally efficient ML methods might 
		not allow for an easy interpretation (what does it mean intuitively if the \gls{logloss} of a hypothesis in a binary classification 
		problem is $10^{-1}$?). It might therefore be useful to use different \gls{lossfunc}s for the search of a good 
		hypothesis (see Chapter \ref{ch_Optimization}) and for its final evaluation. Figure \ref{fig_loss_function_and_metric} 
		depicts an example for two such loss functions, one of them used for learning a hypothesis by minimizing the loss 
		and the other one used for the final performance evaluation. 
		
		For example, in a binary classification problem, we might use the \gls{logloss} to search for (learn) an accurate 
		hypothesis using the optimization methods in Chapter \ref{ch_Optimization}. The \gls{logloss} is appealing for 
		this purpose as it can be minimized via efficient \gls{gdmethods} (see Chapter \ref{ch_GD}). After having found 
		(learnt) an accurate hypothesis, we use the average $0/1$ loss for the final performance evaluation. The $0/1$ loss 
		is appealing for this purpose as it can be interpreted as an error or misclassification rate. The \gls{lossfunc} used for 
		the final performance evaluation of a learnt hypothesis is sometimes referred to as \gls{metric}.
		
		\hspace*{10mm}
		\begin{figure}[htbp]
			\begin{center}
				\begin{tikzpicture}
					\begin{axis}
						[
						axis x line=center,
						axis y line=center,
						xlabel={hypothesis $h$},
						xlabel style={above right},
						ylabel style={above right},
						xtick=\empty,
						ytick=\empty,
						xmin=-4,
						xscale = 1.4, 
						xmax=4,
						ymin=-0.5,
						ymax=2.5
						]
						\addplot [smooth, ultra thick] table [x=a, y=b, col sep=comma] {logloss.csv};    
						\addplot [color=black, dashed, ultra thick] table [x=a, y=b, col sep=comma] {zerooneloss.csv};     
					\end{axis}
					\node [above] at (1,5) {loss for learning a good $h$};
					\node [above] at (1,2.3) {loss (metric) for final evaluation};
				\end{tikzpicture}
			\end{center}
			\vspace*{-7mm}
			\caption{Two different loss functions for a given \gls{datapoint} and varying hypothesis $h$. 
				One of these \gls{loss} functions (solid curve) is used to learn a good hypothesis by minimizing the loss. 
				The other \gls{loss} function (dashed curve) is used to evaluate the performance of the learnt 
				hypothesis. The \gls{lossfunc} used for this final performance evaluation is sometimes referred to 
				as a \gls{metric}.}
			\label{fig_loss_function_and_metric}
		\end{figure}
		
		\subsection{Loss Functions for Numeric Labels} 
		\label{sec_loss_numeric_label}
		For ML problems involving \gls{datapoint}s with a numeric label $\truelabel \in \mathbb{R}$, i.e., for regression problems 
		(see Section \ref{sec_labels}), a widely used (first) choice for the loss function can be the squared error \gls{loss}
		\begin{equation} 
			\label{equ_squared_loss}
			\loss{(\featurevec,\truelabel)}{h} \defeq \big(\truelabel - \underbrace{h(\featurevec)}_{=\predictedlabel} \big)^{2}. 
		\end{equation} 
		The squared error loss \eqref{equ_squared_loss} depends on the features $\featurevec$ only via the 
		predicted label value $\predictedlabel= h(\featurevec)$. We can evaluate the squared error loss solely 
		using the prediction $h(\featurevec)$ and the true label value $\truelabel$. Besides the prediction $h(\featurevec)$, 
		no other properties of the features $\featurevec$ are required to determine the squared error loss. 
		We will slightly abuse notation and use the shorthand $\loss{\truelabel}{\predictedlabel}$ for any loss 
		function that depends on the features $\featurevec$ only via the predicted label $\predictedlabel=h(\featurevec)$. 
		Figure \ref{fig_squarederror_loss} depicts the squared error loss as a function of the prediction error $\truelabel - \predictedlabel$. 
		
		\begin{figure}[htbp]
			\begin{center}
				\begin{tikzpicture}
					\begin{axis}
						[ylabel=$h(x)$,
						xlabel=$x$,grid, 
						axis x line=center,
						axis y line=center,
						xtick={-2,-1,...,2},
						ytick={0,1,...,2},
						xlabel={prediction error $y-h(x)$},
						ylabel={squared error loss $\lossfun$},
						xlabel style={below right},
						ylabel style={above},
						xmin=-2.5,
						xmax=2.5,
						ymin=-0.5,
						ymax=2.5
						]
						\addplot [color=black, ultra thick] table [x=a, y=b, col sep=comma] {squarederrorloss.csv};      
					\end{axis}
				\end{tikzpicture}
				\vspace*{-4mm}
			\end{center}
			\caption{A widely used choice for the \gls{lossfunc} in regression problems (with \gls{datapoint}s having numeric labels) is 
				the squared error loss \eqref{equ_squared_loss}. Note that, for a given hypothesis $h$, we can evaluate the squared error 
				loss only if we know the \gls{features} $\featurevec$ and the \gls{label} $\truelabel$ of the \gls{datapoint}.}
			\label{fig_squarederror_loss}
		\end{figure}
		
		The squared error loss \eqref{equ_squared_loss} has appealing computational and statistical properties. 
		For linear predictor maps $h(\featurevec) = \weights^{T} \featurevec$, the squared error loss is a convex and differentiable 
		function of the parameter vector $\weights$. This allows, in turn, to efficiently search for the optimal linear 
		predictor using efficient iterative optimization methods (see Chapter \ref{ch_GD}). The squared error loss 
		also has a useful interpretation in terms of a probabilistic model for the features and labels. 
		Minimizing the squared error loss is equivalent to maximum likelihood estimation within a linear Gaussian model \cite[Sec. 2.6.3]{hastie01statisticallearning}. 
		
		Another loss function used in regression problems is the absolute error loss $|\hat{y} - y|$. 
		Using this loss function to guide the learning of a predictor results in methods that are robust 
		against a few \gls{outlier}s in the \gls{trainset} (see Section \ref{sec_lad}). However, this improved robustness 
		comes at the expense of increased computational complexity of minimizing the (non-differentiable) 
		absolute error loss compared to the (differentiable) squared error loss \eqref{equ_squared_loss}. 
		
		\subsection{Loss Functions for Categorical Labels} 
		\label{sec_loss_categorical}
		
		Classification problems involve \gls{datapoint}s whose labels take on values from a discrete \gls{labelspace} $\labelspace$. 
		In what follows, unless stated otherwise, we focus on binary classification problems. Moreover, without loss of 
		generality, we use the \gls{labelspace} $\labelspace = \{-1,1\}$. Classification methods aim at learning a hypothesis 
		or classifier that maps the features $\featurevec$ of a \gls{datapoint} to a predicted label $\hat{\truelabel} \in \labelspace$. 
		
		It is often convenient to implement a classifier by thresholding the value $h(\featurevec) \in \mathbb{R}$ of a hypothesis map 
		that can deliver arbitrary real numbers. We then classify a \gls{datapoint} as $\predictedlabel=1$ if $h(\featurevec)>0$ 
		and $\predictedlabel=-1$ otherwise. Thus, the predicted label is obtained from the sign of  the value $h(\featurevec)$. 
		While the sign of $h(\featurevec)$ determines the predicted label $\predictedlabel$, we can interpret the absolute value
		$|h(\featurevec)|$ as the confidence in this classification. Is is customary to abuse notation and refer to both, the 
		final classification rule (obtained by a thresholding step) $\featurevec \mapsto \predictedlabel$ and the hypothesis $h(\featurevec)$ (whose 
		outout is thresholded) as a binary classifier. 
		
		In principle, we can measure the quality of a hypothesis when used to classify \gls{datapoint}s 
		using the squared error loss \eqref{equ_squared_loss}. However, the squared error is typically a poor 
		measure for the quality of a hypothesis $h(\featurevec)$ that is used to classify a \gls{datapoint} 
		with binary label $\truelabel \in \{-1,1\}$. Figure \ref{fig_squarederrornotgoodclass} illustrates 
		how the squared error loss of a hypothesis can be (highly) misleading for binary classification.

		\begin{figure}[htbp]
			\begin{center}
				\begin{tikzpicture}[auto,scale=1]
					\draw  [line width=0.4mm]  (1,0) circle (0.1cm)   node[anchor=south]  {$(1,-1)$};
					\draw  [line width=0.4mm]  (2,0) circle (0.1cm)  node[anchor=west]  {$(2,-1)$};
					\draw [thick] (5,2) \Square{3pt}  node[label={[xshift=0.2cm, yshift=-0.3cm]$(5,1)$}] {};
					\draw [thick] (7,2) \Square{3pt} node[label={[xshift=0.6cm, yshift=-0.3cm]$(\feature\!=\!7,\truelabel\!=\!1)$}] {};
					\draw[->] (-0.5,1) -- (10,1) node[right] {feature $\feature$};
					\draw[dashed,line width=1pt] (2,-1) -- (5,5) node[anchor=west] {predictor $h^{(2)}(\feature)=2(\feature\!-\!3)$} ;
					\draw[line width=1pt] (2,2) -- (10,2) node[anchor=west] {predictor $h^{(1)}(\feature)=1$} ;
					\draw[->] (0,-0.5) -- (0,3.5) node[above] {label $\truelabel$};
				\end{tikzpicture}
			\end{center}
			\caption{A \gls{trainset} consisting of four \gls{datapoint}s with binary labels $\predictedlabel^{(\sampleidx)} \in \{-1,1\}$. 
				Minimizing the squared error loss \eqref{equ_squared_loss} would prefer the (poor) classifier $h^{(1)}$ 
				over the (reasonable) classifier $h^{(2)}$.}
			\label{fig_squarederrornotgoodclass}
		\end{figure}
		
		Figure \ref{fig_squarederrornotgoodclass} depicts a dataset $\dataset$ consisting of $\samplesize=4$ \gls{datapoint}s with 
		binary labels $\truelabel^{(\sampleidx)} \in \{-1,1\}$, for $\sampleidx=1,\ldots,\samplesize$. The figure also depicts 
		two candidate hypotheses $h^{(1)}(\feature)$ and $h^{(2)}(\feature)$ that can be used for classifying \gls{datapoint}s. 
		The classifications $\predictedlabel$ obtained with the hypothesis $h^{(2)}(\feature)$ would perfectly match the 
		labels of the four training \gls{datapoint}s since $h^{(2)}\big(\feature^{(\sampleidx)}\big)\geq 0$ if and if only if $\truelabel^{(\sampleidx)}=1$. 
		In contrast, the classifications $\predictedlabel^{(\sampleidx)}$ obtained by thresholding $h^{(1)}(\feature)$
		 are wrong for \gls{datapoint}s with $\truelabel=-1$. 
		
		Looking at $\dataset$, we might prefer using $h^{(2)}(\feature)$ over $h^{(1)}$ to classify \gls{datapoint}s. 
		However, the squared error loss incurred by the (reasonable) classifier $h^{(2)}$ is much larger 
		than the squared error loss incurred by the (poor) classifier $h^{(1)}$. The squared error loss is 
		typically a bad choice for assessing the quality of a hypothesis map that is used for classifying 
		\gls{datapoint}s into different categories. 
		
		Generally speaking, we want the \gls{lossfunc} to punish (deliver large values for) a hypothesis that 
		is very confident ($|h(\featurevec)|$ is large) in a wrong classification ($\predictedlabel \neq \truelabel$). 
		Moreover, a useful \gls{lossfunc} loss function should not punish (deliver small values for) a hypothesis 
		is very confident ($|h(\featurevec)|$ is large) in a correct classification ($\predictedlabel = \truelabel$). 
		However, by its very definition, the squared loss \eqref{equ_squared_loss} yields large values if 
		the confidence $|h(\featurevec)|$ is large, no matter if the resulting (after thresholding) classification 
		is correct or wrong. 
		
We now discuss some \gls{lossfunc}s which have proven useful for assessing the quality of a 
hypothesis that is used to classify \gls{datapoint}s. Unless noted otherwise, the formulas for these \gls{lossfunc}s 
are valid only if the label values are the real numbers $-1$ and $1$ (the \gls{labelspace} is $\labelspace = \{-1,1\}$). 
These formulas need to modified accordingly if different label values are used. For example, instead of 
the \gls{labelspace}  $\labelspace = \{-1,1\}$, we could equally well use the \gls{labelspace}  $\labelspace =\{0,1\}$, 
or $\labelspace = \{ \square, \triangle \}$ or $\labelspace = \{ \mbox{ ``Class 1''}, \mbox{ ``Class 2''} \}$. 
		
A natural choice for the \gls{lossfunc} can be based on the requirement that a reasonable hypothesis should 
deliver a correct classifications, $\predictedlabel = \truelabel$ for any \gls{datapoint}. 
This suggests to learn a hypothesis $h(\featurevec)$ by minimizing the $0/1$ loss
\begin{equation}
\label{equ_def_0_1}
\loss{(\featurevec,\truelabel)}{h} \defeq \begin{cases} 1 &\mbox{ if } \truelabel \neq \predictedlabel  \\ 0 & \mbox{ else,} \end{cases} \mbox{ with } 
\predictedlabel = 1 \mbox{ for }h(\featurevec) \geq 0\mbox{, and }
\predictedlabel = -1 \mbox{ for }h(\featurevec) < 0.
\end{equation} 
Figure \ref{fig_class_loss} illustrates the $0/1$ loss \eqref{equ_def_0_1} for a \gls{datapoint} with features $\featurevec$ 
and label $\truelabel\!=\!1$ as a function of the hypothesis value $h(\featurevec)$. The $0/1$ loss is equal to zero if 
the hypothesis yields a correct classification $\predictedlabel=\truelabel$. For a wrong classification $\predictedlabel\neq\truelabel$, 
the $0/1$ loss yields the value one.  
		
The $0/1$ loss \eqref{equ_def_0_1} is conceptually appealing when \gls{datapoint}s are interpreted as 
realizations of \gls{iid} \gls{rv}s with the same \gls{probdist} $p(\featurevec,\truelabel)$. 
Given $\samplesize$ realizations $(\featurevec^{(\sampleidx)},\truelabel^{(\sampleidx)}) \big\}_{\sampleidx=1}^{\samplesize}$ of such \gls{iid} \gls{rv}s,
\begin{equation} 
\label{equ_0_1_approx_prob}
(1/\samplesize) \sum_{\sampleidx=1}^{\samplesize} \loss{(\featurevec^{(\sampleidx)},\truelabel^{(\sampleidx)})}{h} \approx \prob{ \truelabel \neq \predictedlabel }
\end{equation} 
with high probability for sufficiently large sample size $\samplesize$. A precise formulation of the 
approximation \eqref{equ_0_1_approx_prob} can be obtained from the \gls{lln} \cite[Section 1]{BillingsleyProbMeasure}. 
We can apply the \gls{lln} since the loss values $\loss{(\featurevec^{(\sampleidx)},\truelabel^{(\sampleidx)})}{h}$ are realizations 
of \gls{iid} \gls{rv}s. It is customary to indicate the average $0/1$ loss of a hypothesis as the \index{accuracy}accuracy 
$1 - (1/\samplesize) \sum_{\sampleidx=1}^{\samplesize} \loss{(\featurevec^{(\sampleidx)},\truelabel^{(\sampleidx)})}{h}$. 
		
In view of \eqref{equ_0_1_approx_prob}, the $0/1$ loss \eqref{equ_def_0_1} seems a very natural choice for 
assessing the quality of a classifier if our goal is to enforce correct classifications $\hat{\truelabel}=\truelabel$. 
This appealing statistical property of the $0/1$ loss comes at the cost of a high computational complexity. 
Indeed, for a given \gls{datapoint} $(\featurevec,\truelabel)$, the $0/1$ loss \eqref{equ_def_0_1} is 
non-convex and non-differentiable when viewed as a function of the hypothesis $h$. Thus, ML 
methods that use the $0/1$ loss to learn a hypothesis map typically involve advanced 
optimization methods to solve the resulting learning problem (see Section \ref{sec_NaiveBayes}). 
		
To avoid the non-convexity of the $0/1$ loss \eqref{equ_def_0_1} we might approximate it by a 
convex \gls{lossfunc}. One popular convex approximation of the $0/1$ loss is the \index{hinge loss}\gls{hingeloss}
\begin{equation} 
\label{equ_hinge_loss}
\loss{(\featurevec,\truelabel)}{h} \defeq \max \{ 0 , 1 - \truelabel h(\featurevec) \}. 
\end{equation}
Figure \ref{fig_class_loss} depicts the \gls{hingeloss} \eqref{equ_hinge_loss} as a function of the 
hypothesis $h(\featurevec)$. The \gls{hingeloss} \eqref{equ_hinge_loss} becomes minimal (equal to zero) 
for a correct classification ($\hat{\truelabel} = \truelabel$) with sufficient confidence $h(\featurevec) \geq 1$. 
For a wrong classification  ($\hat{\truelabel} \neq \truelabel$), the \gls{hingeloss} increases monotonically 
with the confidence $|h(\feature)|$ in the wrong classification. While the \gls{hingeloss} avoids the non-convexity 
of the $0/1$ \gls{loss}, it still is a non-differentiable function of $h(\featurevec)$. A non-differentiable 
\gls{lossfunc} cannot be minimized by simple \gls{gdmethods} (see Chapter \ref{ch_GD}) but require 
more advanced optimization methods. 
		
Beside the $0/1$ loss and the \gls{hingeloss}, another popular \gls{lossfunc} for binary \gls{classification} 
problems is the \gls{logloss}
\begin{equation} 
\label{equ_log_loss}
\loss{(\featurevec,\truelabel)}{h} \defeq  \log ( 1 + \exp(- \truelabel h(\featurevec))).
\end{equation}
The \gls{logloss} \eqref{equ_log_loss} is used in \gls{logreg} (see Section \ref{sec_LogReg}) to measure the usefulness 
of a linear hypothesis $h(\featurevec) = \weights^{T} \featurevec$. 
Figure \ref{fig_class_loss} depicts the \gls{logloss} \eqref{equ_log_loss} as a function of the hypothesis $h(\featurevec)$. 
The \gls{logloss} \eqref{equ_log_loss} is a convex and differentiable function of $h(\featurevec)$. For a correct 
classification ($\hat{\truelabel}\!=\!\truelabel$), the \gls{logloss} \eqref{equ_log_loss} decreases monotonically with  
increasing confidence $h(\featurevec)$. For a wrong classification  ($\hat{\truelabel}\!\neq\!\truelabel$), the \gls{logloss} 
increases monotonically with increasing confidence $|h(\featurevec)|$ in the wrong classification.

Both the \gls{hingeloss} \eqref{equ_hinge_loss} and the \gls{logloss} \eqref{equ_log_loss} are \gls{convex} functions  
of the weights $\weights \in \mathbb{R}^{\featuredim}$ in a linear hypothesis map $h(\featurevec) = \weights^{T} \featurevec$.
However, in contrast to the \gls{hingeloss}, the \gls{logloss} \eqref{equ_log_loss} is also a differentiable function of the $\weights$. 
The convex and differentiable \gls{logloss} function can be minimized using simple \gls{gdmethods} such 
as \gls{gd} (see Chapter \ref{sec_GD_logistic_regression}). In contrast, we cannot use basic \gls{gdmethods} to 
minimize the \gls{hingeloss} since it is not differentiable (it does not have a \gls{gradient} everywhere). However, 
we can apply a generalization of \gls{gd} which is known as \index{subgradient descent}\gls{sgd} \cite{CvxBubeck2015}. 
\Gls{sgd} is obtained from \gls{gd} by generalizing the concept of a \gls{gradient} to that of a \index{subgradient}\gls{subgradient}.

			\begin{figure}[htbp]
					\begin{center}
				\begin{tikzpicture}
				
					\begin{axis}
						[ylabel=$h(\featurevec)$,
						xlabel=$\featurevec$,grid, 
						axis x line=center,
						axis y line=center,
						xtick={-2,-1,...,2},
						ytick={0,1,...,2},
						xlabel={hypothesis $h(\featurevec)$},
						ylabel={\hspace*{3mm} loss $\lossfun$},
						xlabel style={below right},
						ylabel style={above right},
						xmin=-4,
						xscale = 1.4, 
						xmax=4,
						ymin=-0.5,
						ymax=2.5
						]
						\addplot [smooth, color=red, ultra thick] table [x=a, y=b, col sep=comma] {logloss.csv};    
						\addplot [color=blue, ultra thick] table [x=a, y=b, col sep=comma] {hingeloss.csv};     
						\addplot [color=black, ultra thick] table [x=a, y=b, col sep=comma] {zerooneloss.csv};     
						\addplot[domain=-1:4, thin,dashed] {(x-1)^2};
					\end{axis}
					
					\node [right] at (9,6) {very confident in $\predictedlabel\!=\!1 \Rightarrow$}; 
					\node [right,color=red] at (7,1.5) {\gls{logloss} (for $\truelabel\!=\!1$)};
					\node [right] at (7,3) {squared error (for $\truelabel\!=\!1$)};
					\node [right,color=blue] at (4,4) {hinge loss (for $\truelabel\!=\!1)$};
					\node [left] at (2.5,3) {$0/1$ loss (for $\truelabel\!=\!1)$};
					\node [left] at (3,6) {$\Leftarrow$ very confident in $\predictedlabel\!=\!-1$};
				\end{tikzpicture}
				\end{center}
				\vspace*{-4mm}
				\caption{The solid curves depict three widely-used \gls{lossfunc}s for binary classification. 
					A \gls{datapoint} with features $\featurevec$ and label $\truelabel=1$ is classified as $\predictedlabel=1$ if $h(\featurevec)\geq 0$ 
					and classified as $\predictedlabel=-1$ if $h(\featurevec) <0$. We can interpret the absolute value $|h(\featurevec)|$ 
					as the confidence in the \gls{classification} $\hat{\truelabel}$. The more confident we are in a 
					correct \gls{classification} ($\predictedlabel\!=\!\truelabel\!=\!1$), i.e, the more positive $h(\featurevec)$, 
					the smaller the \gls{loss}. Note that each of the three \gls{lossfunc}s for binary \gls{classification} 
					tends monotonically towards $0$ for increasing $h(\featurevec)$. The dashed curve depicts the squared 
					error loss \eqref{equ_squared_loss}, which increases for increasing $h(\featurevec)$.   }
				\label{fig_class_loss}
			\end{figure}

\subsection{Loss Functions for Ordinal Label Values}
Some \gls{lossfunc}s are particularly suited for predicting ordinal label values (see Section \ref{sec_the_data}). 
Consider \gls{datapoint}s representing areal images of rectangular areas of size $1 \mbox{km}$ by $1 \mbox{km}$. 
We characterize each \gls{datapoint} (rectangular area) by the feature vector $\featurevec$ obtained by stacking 
the RGB values of each image pixel (see Figure \ref{fig_snapshot_pixels}). Beside the feature vector, 
each rectangular area is characterized by a label $\truelabel \in \{1,2,3\}$ where 
\begin{itemize} 
\item $\truelabel=1$ means that the area contains no trees. 
\item $\truelabel=2$ means that the area is partially covered by trees. 
\item $\truelabel=3$ means that the are is entirely covered by trees. 
\end{itemize} 
We might consider the label value $\truelabel=2$ to be ``larger'' than label value $\truelabel=1$ 
and label value $\truelabel=3$ to be ``larger'' than label value $\truelabel=2$. Let us construct a
\gls{lossfunc} that takes such an ordering of label values into account when evaluating the quality 
of the predictions $h(\featurevec)$. 
		
Consider a \gls{datapoint} with feature vector $\featurevec$ and label $\truelabel=1$ as well as two different 
hypotheses $h^{(a)}, h^{(b)} \in \hypospace$. The hypothesis $h^{(a)}$ delivers the predicted label $\hat{\truelabel}^{(a)}= h^{(a)}(\featurevec) =2$, 
while the other hypothesis $h^{(b)}$ delivers the predicted label $\hat{\truelabel}^{(a)}= h^{(a)}(\featurevec) =3$. Both 
predictions are wrong, since they are different form the true label value $\truelabel=1$. It seems reasonable to consider 
the prediction $\hat{\truelabel}^{(a)}$ to be less wrong than the prediction $\hat{\truelabel}^{(b)}$ and therefore we would 
prefer the hypothesis $h^{(a)}$ over $h^{(b)}$. However, the $0/1$ loss is the same for $h^{(a)}$ and $h^{(b)}$ and therefore 
does not reflect our preference for $h^{(a)}$. We need to modify (or tailor) the $0/1$ loss to take into account the 
application-specific ordering of label values. For the above application, we might define a loss function via 
\begin{equation}
\loss{(\featurevec,\truelabel)}{h} \defeq \begin{cases}  0 &\mbox{ , when } \truelabel = h(\featurevec) \\ 10 &\mbox{ , when } |\truelabel-h(\featurevec)|=1  \\ 100 & \mbox{ otherwise.} \end{cases} 
\end{equation} 
		
		\subsection{Empirical Risk} 
		\label{sec_empirical_risk}
		The basic idea of ML methods (including those discussed in Chapter \ref{ch_some_examples}) is to find (or learn) 
		a hypothesis (out of a given \gls{hypospace} $\hypospace$) that incurs minimum loss when applied to 
		arbitrary \gls{datapoint}s. To make this informal goal precise we need to specify what we mean by ``arbitrary \gls{datapoint}''. 
		One of the most successful approaches to define the notion of ``arbitrary \gls{datapoint}'' is by probabilistic models 
		for the observed \gls{datapoint}s. 
		
		The most basic and widely-used probabilistic model interprets \gls{datapoint}s 
		$\big( \featurevec^{(\sampleidx)},\truelabel^{(\sampleidx)} \big)$ as realizations of \gls{iid} \gls{rv}s 
		with a common \gls{probdist} $p(\featurevec,\truelabel)$. Given such a probabilistic 
		model, it seems natural to measure the quality of a hypothesis $h$ by the expected loss or \gls{bayesrisk} \cite{LC} 
		\begin{equation} 
			\label{equ_def_bayes_risk_int}
			\expect \big \{ \loss{(\featurevec,\truelabel)}{h} \} \defeq \int_{\featurevec,\truelabel}  \loss{(\featurevec,\truelabel)}{h} d p(\featurevec,\truelabel).
		\end{equation} 
		The \gls{bayesrisk} of $h$ is the expected value of the loss $\loss{(\featurevec,\truelabel)}{h}$ incurred when applying the 
		hypothesis $h$ to (the realization of) a random \gls{datapoint} with features $\featurevec$ and label $\truelabel$. Note 
		that the computation of the \gls{bayesrisk} \eqref{equ_def_bayes_risk_int} requires the joint \gls{probdist} 
		$p(\featurevec,\truelabel)$ of the (random) features and label of \gls{datapoint}s. 
		
		The \gls{bayesrisk} \eqref{equ_def_bayes_risk_int} seems to be reasonable performance measure for a hypothesis $h$. Indeed, the \gls{bayesrisk} of a 
		hypothesis is small only if the hypothesis incurs a small loss on average for \gls{datapoint}s drawn from the \gls{probdist} 
		$p(\featurevec,\truelabel)$. However, it might be challenging to verify if the \gls{datapoint}s generated in a
		particular application domain can be accurately modelled as realizations (draws) from a \gls{probdist} $p(\featurevec,\truelabel)$. 
		Moreover, it is also often the case that we do not know the correct \gls{probdist} $p(\featurevec,\truelabel)$. 
		
		Let us assume for the moment, that \gls{datapoint}s are generated as \gls{iid} realizations of a common \gls{probdist} 
		$p(\featurevec,\truelabel)$ which is known. It seems reasonable to learn a hypothesis $\bayeshypothesis$ 
		that incurs minimum \gls{bayesrisk}, 
		\begin{equation} 
			\label{equ_def_bayes_risk_int}
			\expect \big \{ \loss{(\featurevec,\truelabel)}{\bayeshypothesis} \}  \defeq \min_{h \in \hypospace} \expect \big \{ \loss{(\featurevec,\truelabel)}{h} \}. 
		\end{equation} 
		A hypothesis that solves \eqref{equ_def_bayes_risk_int}, i.e., that achieves the minimum possible \gls{bayesrisk}, 
		is referred to as a \gls{bayesestimator} \cite[Chapter 4]{LC}. The main computational challenge for learning the 
		optimal hypothesis is the efficient (numerical) solution of the optimization problem \eqref{equ_def_bayes_risk_int}. 
		Efficient methods to solve the optimization problem \eqref{equ_def_bayes_risk_int} are studied within 
		estimation theory \cite{LC,GraphModExpFamVarInfWainJor}.  
		
		The focus of this book is on ML methods which do not require knowledge of the underlying \gls{probdist} 
		$p(\featurevec,\truelabel)$. One of the most widely used principle for these ML methods is to approximate the 
		\gls{bayesrisk} by an empirical (sample) average over a finite set of labeled \gls{datapoint}s $\dataset=\big(\featurevec^{(1)},\truelabel^{(1)}\big),\ldots,\big(\featurevec^{(\samplesize)},\truelabel^{(\samplesize)}\big)$. 
		In particular, we define the \index{empirical risk}\gls{emprisk} of a hypothesis $h \in \hypospace$ for a dataset $\dataset$ as 
		\begin{equation} 
			\label{eq_def_emp_error_101}
			\emperror(h|\dataset) = (1/\samplesize)\sum_{\sampleidx=1}^{\samplesize} \loss{(\featurevec^{(\sampleidx)},\truelabel^{(\sampleidx)})}{h}.  
		\end{equation} 
		The \gls{emprisk} of the hypothesis $h \in \hypospace$ is the average loss on the \gls{datapoint}s in $\dataset$. 
		To ease notational burden, we use $\emperror(h)$ as a shorthand for $\emperror(h|\dataset)$ if the underlying 
		dataset $\dataset$ is clear from the context. Note that in general the \gls{emprisk} depends on both, 
		the hypothesis $h$ and the (features and labels of the) \gls{datapoint}s in the dataset $\dataset$. 
		
		If the \gls{datapoint}s used to compute the \gls{emprisk} \eqref{eq_def_emp_error_101} are (can be modelled as) 
		realizations of \gls{iid} \gls{rv}s whose common distribution is $p(\featurevec,\truelabel)$, basic results of 
		probability theory tell us that  
		\begin{equation} 
			\label{equ_emp_risk_approximates_Bayes_risk}
			\expect \big \{ \loss{(\featurevec,\truelabel)}{h} \} \approx  (1/\samplesize)\sum_{\sampleidx=1}^{\samplesize} \loss{(\featurevec^{(\sampleidx)},\truelabel^{(\sampleidx)})}{h} \mbox{ for sufficiently large sample size } \samplesize.
		\end{equation}
		The approximation error in \eqref{equ_emp_risk_approximates_Bayes_risk} can be quantified precisely by 
		some of the most basic results of probability theory. These results are often summarized under the umbrella term 
		\gls{lln} \cite{BillingsleyProbMeasure,GrayProbBook,BertsekasProb}. 
		
		Many (if not most) ML methods are motivated by \eqref{equ_emp_risk_approximates_Bayes_risk} which suggests that 
		a hypothesis with small \gls{emprisk} \eqref{eq_def_emp_error_101} will also result in a small expected loss. The minimum 
		possible expected loss is achieved by the \gls{bayesestimator} of the label $\truelabel$, given the features $\featurevec$. 
		However, to actually compute the optimal estimator we would need to know the (joint) \gls{probdist} $p(\featurevec,\truelabel)$ 
		of features $\featurevec$ and label $\truelabel$. 
		
		\subsubsection{Confusion Matrix}
		\label{sec_confustion_matrix} 
		
		Consider a dataset $\dataset$ with \gls{datapoint}s characterized by feature vectors $\featurevec^{(\sampleidx)}$ 
		and labels $\truelabel^{(\sampleidx)} \in \{1,\ldots,\nrcluster\}$. We might interpret the label value of a \gls{datapoint} 
		as the index of a category or class to which the \gls{datapoint} belongs to. Multi-class classification problems 
		aim at learning a hypothesis $h$ such that $h(\featurevec) \approx \truelabel$ for any \gls{datapoint}. 
		
		In principle, we could measure the quality of a given hypothesis $h$ by the average $0/1$ loss incurred on the 
		labeled \gls{datapoint}s in (the \gls{trainset}) $\dataset$. However, if the dataset $\dataset$ contains mostly \gls{datapoint}s
		with one specific label value, the average $0/1$ loss might obscure the performance of $h$ for \gls{datapoint}s 
		having one of the rare label values. Indeed, even if the average $0/1$ loss is very small, the hypothesis might 
		perform poorly for \gls{datapoint}s of a minority category. 
		
		The \index{confusion matrix}\gls{cm} generalizes the concept of the $0/1$ loss to application domains 
		where the relative frequency (fraction) of \gls{datapoint}s with a specific label value varies significantly (\index{imbalanced data}imbalanced data). 
		Instead of considering only the average $0/1$ loss incurred by a hypothesis on a dataset $\dataset$, we use a 
		whole family of \gls{lossfunc}s. In particular, for each pair of label values $\clusteridx,\clusteridx' \in \{1,\ldots,\nrcluster\}$, we define the loss 
		\begin{equation} 
			\label{equ_def_loss_funs_conf_matrix}
			\lossfun^{(\clusteridx\!\rightarrow\!\clusteridx')}\big( \big(\featurevec, \truelabel \big),h \big) \defeq \begin{cases} 1 & \mbox{ if } \truelabel= \clusteridx \mbox{ and } h(\featurevec) = \clusteridx' \\ 0 &\mbox{ otherwise} \end{cases}.
		\end{equation} 
		We then compute the average loss \eqref{equ_def_loss_funs_conf_matrix} incurred on the 
		dataset $\dataset$, 
		\begin{equation} 
			\label{eq_def_emp_error_conf_matrix}
			\emperror^{(\clusteridx\!\rightarrow\!\clusteridx')}(h|\dataset) \defeq (1/\samplesize)\sum_{\sampleidx=1}^{\samplesize} \lossfun^{(\clusteridx\!\rightarrow\!\clusteridx')} \big( (\featurevec^{(\sampleidx)},\truelabel^{(\sampleidx)}),h \big) \mbox{ for } \clusteridx,\clusteridx' \in \{1,\ldots,\nrcluster\}. 
		\end{equation} 
		It is convenient to arrange the values \eqref{eq_def_emp_error_conf_matrix} as a matrix which is referred 
		to as a \gls{cm}. The rows of a \gls{cm} correspond to different values $\clusteridx$ of the true label of a \gls{datapoint}. 
		The columns of a \gls{cm} correspond to different values $\clusteridx'$ delivered by the hypothesis $h(\featurevec)$. 
		The $(\clusteridx,\clusteridx')$-th entry of the \gls{cm} is $\emperror^{(\clusteridx\!\rightarrow\!\clusteridx')}(h|\dataset)$.
		
		
		\subsubsection{Precision, Recall and F-Measure}
		\label{sec_prec_recall_f1}
		
		Consider an object detection application where \gls{datapoint}s are images. The label of \gls{datapoint}s might indicate 
		the presence ($\truelabel\!=\!1$) or absence ($\truelabel\!=\!-1$) of an object, it is then customary to define the \cite{InfoRetrievaBook}
		\begin{equation} 
			\mbox{recall } \defeq \emperror^{(1 \rightarrow 1)}(h|\dataset) \mbox{, and the precision } \defeq \frac{\emperror^{(1 \rightarrow 1)}(h|\dataset)} {\emperror^{(1 \rightarrow 1)}(h|\dataset)  + \emperror^{(-1 \rightarrow 1)}(h|\dataset)  }. 
		\end{equation} 
		Clearly, we would like to find a hypothesis with both, large recall and large precision. However, these two goals 
		are typically conflicting, a hypothesis with a high recall will have small precision. Depending on the 
		application, we might prefer having a high recall and tolerate a lower precision. 
		
		It might be convenient to combine the recall and precision of a hypothesis into a single quantity, 
		\begin{equation} 
			\label{equ_f1_score}
			\fscore \defeq 2  \frac{\mbox{precision} \times \mbox{recall}}{\mbox{precision} +\mbox{recall}}
		\end{equation} 
		The $F$ measure \eqref{equ_f1_score} is the harmonic mean \cite{AbramowitzStegun} of the precision 
		and recall of a hypothesis $h$. It is a special case of the $F_{\beta}$-score
		\begin{equation} 
			\label{equ_fbeta_score}
			F_{\beta} \defeq \big( 1 + \beta^2 \big)  \frac{\mbox{precision} \times \mbox{recall}}{\beta^{2} \mbox{precision} +\mbox{recall}}. 
		\end{equation} 
		The $F$ measure \eqref{equ_f1_score} is obtained from \eqref{equ_fbeta_score} for the choice $\beta=1$. It is 
		therefore customary to refer to \eqref{equ_f1_score} as the $F_{1}$-score of a hypothesis $h$. 		
		
		\subsection{Regret} 
		
		In some ML applications, we might have access to the predictions obtained from some reference 
		methods which are referred to as \index{expert} \gls{expert}s \cite{PredictionLearningGames,HazanOCO}. 
		The quality of a hypothesis $h$ is measured via the difference between the loss incurred by its predictions $h(\featurevec)$ and the loss incurred by the 
		predictions of the experts. This difference, which is referred to as the \index{regret}regret, 
		measures by how much we \gls{regret} to have used the prediction $h(\featurevec)$ instead of using(or following) 
		the prediction of the expert. The goal of \index{regret minimization} \gls{regret} minimization is to learn a 
		hypothesis with a small \gls{regret} compared to given set of \gls{expert}s. 
		
		The concept of \gls{regret} minimization is useful when we do not make any probabilistic assumptions 
		(see Section \ref{equ_prob_models_data}) about the data. Without a probabilistic model we cannot use 
		the \gls{bayesrisk} of the (optimal) \gls{bayesestimator} as a baseline (or benchmark). 
		The concept of \gls{regret} minimization avoids the need for a probabilistic model of the data to obtain a 
		baseline \cite{PredictionLearningGames}. This approach replaces the \gls{bayesrisk} with the \gls{regret} 
		relative to given reference methods (the \gls{expert}s). 
		
		\subsection{Rewards as Partial Feedback}  
		Some applications involve \gls{datapoint}s whose labels are so difficult or costly to determine that we 
		cannot assume to have any labeled data available. Without any labeled data, we cannot evaluate the 
		loss function for different choices for the hypothesis. Indeed, the evaluation of the loss function 
		typically amounts to measuring the distance between predicted label and true label of a \gls{datapoint}. 
		Instead of evaluating a loss function, we must rely on some indirect feedback or ``reward'' that 
		indicates the usefulness of a particular prediction \cite{PredictionLearningGames,SuttonEd2}. 
		
		Consider the ML problem of predicting the optimal steering directions for an autonomous car. The 
		prediction has to be recalculated for each new state of the car. ML methods can sense the state via a 
		feature vector $\featurevec$ whose entries are pixel intensities of a snapshot. The goal is to learn a hypothesis 
		map from the feature vector $\featurevec$ to a guess $\hat{\truelabel} = h(\featurevec)$ for the optimal steering direction $\truelabel$ 
		(true label). Unless the car circles around in small area with fixed obstacles, we have no access to 
		labeled \gls{datapoint}s or reference driving scenes for which we already know the optimum steering direction. 
		Instead, the car (control unit) needs to learn the hypothesis $h(\featurevec)$ based solely on the feedback signals 
		obtained from various sensing devices (cameras, distance sensors).

		\section{Putting Together the Pieces} 
		\label{sec_putting_togehter_the_pieces}
		
		A guiding theme of this book is that ML methods are obtained by different combinations of data, 
		model and loss. We will discuss some key principles behind these methods in depth in the following 
		chapters. Let us develop some intuition for how ML methods operate by considering a very simple 
		ML problem. This problem involves \gls{datapoint}s that are characterized by a single numeric 
		feature $\feature \in \mathbb{R}$ and a numeric label $\truelabel  \in \mathbb{R}$. We assume 
		to have access to $\samplesize$ labeled \gls{datapoint}s 
		\begin{equation} 
			\big(\feature^{(1)},\truelabel^{(1)}\big),\ldots, \big(\feature^{(\samplesize)},\truelabel^{(\samplesize)}\big)  \label{equ_labeled_data_putting_together}
		\end{equation} 
		for which we know the true label values $\truelabel^{(\sampleidx)}$. 
		
		The assumption of knowing the exact true label values $\truelabel^{(\sampleidx)}$ for any \gls{datapoint} is an 
		idealization. We might often face labelling or measurement errors such that the observed labels are noisy 
		versions of the true label. Later on, we will discuss techniques that allow ML 
		methods to cope with noisy labels in Chapter \ref{ch_overfitting_regularization}. 
		
		Our goal is to learn a (hypothesis) map $h: \mathbb{R} \rightarrow \mathbb{R}$ such that 
		$h(\feature) \approx \truelabel$ for any \gls{datapoint}. Given a \gls{datapoint} with feature $\feature$, 
		the function value $h(\feature)$ should be an accurate approximation of its label value $\truelabel$. 
		We require the map to belong to the \gls{hypospace} $\hypospace$ of linear maps,  
		\begin{equation} 
			\label{equ_def_lin_pred_intercept}
			h^{(\weight_{0},\weight_{1})}(\feature) = \weight_{1} \feature + \weight_{0}. 
		\end{equation}
		
		The predictor \eqref{equ_def_lin_pred_intercept} is parameterized by the slope 
		$\weight_{1}$ and the intercept (bias or offset) $\weight_{0}$. We indicate this by the 
		notation $h^{(\weight_{0},\weight_{1})}$. A particular choice for the weights $\weight_{1},\weight_{0}$ 
		defines a linear hypothesis $h^{(\weight_{0},\weight_{1})}(\feature) = \weight_{1}\feature +\weight_{0}$. 
		
		Let us use the linear hypothesis map $h^{(\weight_{0},\weight_{1})}(\feature)$ to predict the labels of 
		training \gls{datapoint}s. In general, the predictions $\hat{\truelabel}^{(\sampleidx)} = h^{(\weight_{0},\weight_{1})}\big(\feature^{(\sampleidx)}\big)$ 
		will not be perfect and incur a non-zero \index{prediction error}prediction error $\hat{\truelabel}^{(\sampleidx)} - \truelabel^{(\sampleidx)}$ (see Figure \ref{fig_emp_error}).  
		
		We measure the goodness of the predictor map $h^{(\weight_{0},\weight_{1})}$ 
		using the average squared error loss (see \eqref{equ_squared_loss})
		\begin{align}
			\label{equ_def_cost_fun_putting_togheter}
			f(\weight_{0},\weight_{1}) & \defeq (1/\samplesize) \sum_{\sampleidx=1}^{\samplesize} \big( \truelabel^{(\sampleidx)} - h^{(\weight_{0},\weight_{1})}(\feature^{(\sampleidx)})  \big)^{2} \nonumber \\
			& \stackrel{\eqref{equ_def_lin_pred_intercept}}{=}  (1/\samplesize) \sum_{\sampleidx=1}^{\samplesize} \big( \truelabel^{(\sampleidx)} - ( \weight_{1} \feature^{(\sampleidx)} + \weight_{0}) \big)^{2}. 
		\end{align}
		The \gls{trainerr} $f(\weight_{0},\weight_{1})$ is the average of the squared 
		prediction errors incurred by the predictor $h^{(\weight_{0},\weight_{1})}(x)$ 
		to the labeled \gls{datapoint}s \eqref{equ_labeled_data_putting_together}. 
		
		It seems natural to learn a good predictor \eqref{equ_def_lin_pred_intercept} 
		by choosing the parameters $\weight_{0},\weight_{1}$ to minimize the \gls{trainerr} 
		\begin{equation}
			\min_{\weight_{0},\weight_{1} \in \mathbb{R}} f(\weight_{0},\weight_{1}) \stackrel{\eqref{equ_def_cost_fun_putting_togheter}}{=} \min_{\weight_{1},\weight_{0} \in \mathbb{R}}   (1/\samplesize) \sum_{\sampleidx=1}^{\samplesize} \big( \truelabel^{(\sampleidx)} - ( \weight_{1} \feature^{(\sampleidx)} + \weight_{0}) \big)^{2}.
		\end{equation} 
		
		The optimal parameters $\hat{\weight}_{0},\hat{\weight}_{1}$ are characterized by the 
		zero-gradient condition,\footnote{A necessary and sufficient condition for $\widehat{\weights}$ to minimize a 
			convex differentiable function $f(\weights)$ is $\nabla f(\widehat{\weights}) = \mathbf{0}$ \cite[Sec.\ 4.2.3]{BoydConvexBook}.} 
		\begin{equation}
			\label{equ_zero_grad_condition_putting_together}
			\frac{\partial f(\weight_{0},\weight_{1})}{\partial \weight_{0}}\bigg|_{\weight_{0} = \hat{\weight}_{0}, \weight_{1} = \hat{\weight}_{1}} = 0 \mbox{, and }\frac{\partial f(\weight_{0},\weight_{1})}{\partial \weight_{1}}\bigg|_{\weight_{0} = \hat{\weight}_{0}, \weight_{1} = \hat{\weight}_{1}} = 0. 
		\end{equation} 
		Inserting \eqref{equ_def_cost_fun_putting_togheter} into \eqref{equ_zero_grad_condition_putting_together} and 
		by using basic rules for calculating derivatives (see, e.g., \cite{RudinBookPrinciplesMatheAnalysis}), we obtain the 
		following optimality conditions 
		\begin{align}
			\label{equ_zero_grad_condition_putting_together_special_form}
		&	(1/\samplesize) \sum_{\sampleidx=1}^{\samplesize} \big( \truelabel^{(\sampleidx)} - ( \hat{\weight}_{1} \feature^{(\sampleidx)} + \hat{\weight}_{0}) \big)= 0 \mbox{, and } \\ 
		& (1/\samplesize) \sum_{\sampleidx=1}^{\samplesize} \feature^{(\sampleidx)} \big( \truelabel^{(\sampleidx)} - ( \hat{\weight}_{1} \feature^{(\sampleidx)} + \hat{\weight}_{0}) \big) = 0.  \nonumber
		\end{align} 
		
		Any parameter values $\hat{\weight}_{0},\hat{\weight}_{1} \in \mathbb{R}$ that satisfy \eqref{equ_zero_grad_condition_putting_together_special_form} 
		define a hypothesis map $h^{(\hat{\weight}_{0},\hat{\weight}_{1})}(\feature) = \hat{\weight}_{1}\feature + \hat{\weight}_{0}$ that 
		is optimal in the sense of incurring a minimum \gls{trainerr}, 
		$$f(\hat{\weight}_{0},\hat{\weight}_{1}) = \min_{\weight_{0},\weight_{1} \in \mathbb{R}} f(\weight_{0},\weight_{1}).$$
		
		Let us rewrite the optimality condition \eqref{equ_zero_grad_condition_putting_together_special_form} 
		using matrices and vectors. To this end, we first rewrite the hypothesis \eqref{equ_def_lin_pred_intercept} as 
		$$ h(\featurevec) = \weights^{T} \featurevec \mbox{ with } \weights = \big(\weight_{0},\weight_{1}\big)^{T}, \featurevec = \big(1,\feature\big)^{T}.$$
		Let us stack the feature vectors $\featurevec^{(\sampleidx)} = \big(1,\feature^{(\sampleidx)} \big)^{T}$ 
		and labels $\truelabel^{(\sampleidx)}$ of the \gls{datapoint}s \eqref{equ_labeled_data_putting_together} 
		into a \index{feature matrix}feature matrix and a \index{label vector}label vector, 
		\begin{equation}
			\featuremtx  = \big(\featurevec^{(1)},\ldots,\featurevec^{(\samplesize)}\big)^{T} \in \mathbb{R}^{\samplesize \times 2}, \labelvec = \big(\truelabel^{(1)},\ldots,\truelabel^{(\samplesize)}\big)^{T} \in \mathbb{R}^{\samplesize}. 
		\end{equation} 
		We can then reformulate \eqref{equ_zero_grad_condition_putting_together_special_form} as 
		\begin{equation}
			\label{equ_zero_grad_condition_putting_together_special_form_mtx}
			\featuremtx^{T} \big( \labelvec - \featuremtx \widehat{\weights} \big) = \mathbf{0}. 
		\end{equation} 
The optimality conditions \eqref{equ_zero_grad_condition_putting_together_special_form_mtx} and 
\eqref{equ_zero_grad_condition_putting_together_special_form} are equivalent in the following sense. 
The entries of any parameter vector $\widehat{\weights} = \big(\hat{\weight}_{0},\hat{\weight}_{1}\big)$ that satisfies  \eqref{equ_zero_grad_condition_putting_together_special_form_mtx} are solutions to 
\eqref{equ_zero_grad_condition_putting_together_special_form} and vice-versa. 
		
		\begin{figure}[htbp]
			\centering
			
			\tikzset{global scale/.style={
					scale=#1,
					every node/.append style={scale=#1}
				}
			}
			\begin{tikzpicture}[global scale = 1]             
				\draw[->, very thick](-0.5,0)--(10,0) node[above, xshift=1cm, yshift=0.2cm, font=\fontsize{12}{0}\selectfont] {feature $\feature$};
				\draw[->, very thick](0,-0.5)--(0,6) node[above, font=\fontsize{12}{0}\selectfont] {label $\truelabel$};
				
				\node[font=\fontsize{12}{0}\selectfont] (x1) at (3, -0.7) {$\feature^{(1)}$};      
				\node[font=\fontsize{12}{0}\selectfont] (x2) at (6, -0.7) {$\feature^{(2)}$};
				\node[font=\fontsize{12}{0}\selectfont] (x3) at (8, -0.7) {$\feature^{(3)}$};
				
				\node[yshift=2.1cm] (hx1) at (x1) {};      
				\node[yshift=4.3cm] (hx2) at (x2) {};
				\node[yshift=5cm] (hx3) at (x3) {};
				\node (hx) at (9,4.6) {};
				
				\node[yshift=5.5cm] (y1) at (x1) {};        
				\node[yshift=3.5cm] (y2) at (x2) {};
				\node[yshift=6.5cm] (y3) at (x3) {};
				
				\draw[line width=2pt] plot [smooth, tension=1]
				coordinates {(-0.5,0.5) (1.3,0.7) (hx1) (4.2,2.7) (hx2) (hx3) (hx)};
				
				\draw[fill=white] (hx1) circle (0.1) node[below, xshift=1cm, font=\fontsize{12}{0}\selectfont] {$h(\feature^{(1)})$};     
				\draw[fill=white] (hx2) circle (0.1) node[below, xshift=1cm, font=\fontsize{12}{0}\selectfont] {$h(\feature^{(2)})$}; 
				\draw[fill=white] (hx3) circle (0.1) node[below, xshift=1cm, font=\fontsize{12}{0}\selectfont] {$h(\feature^{(3)})$}; 
				\node[xshift=0.5cm, yshift=0.3cm, font=\fontsize{12}{0}\selectfont] at (hx) {$h(\feature)$};
				
				\draw[fill=white] (y1) circle (0.1) node[above, yshift=0.2cm, font=\fontsize{12}{0}\selectfont] {$\truelabel^{(1)}$};     
				\draw[fill=white] (y2) circle (0.1) node[below, xshift=0.5cm, font=\fontsize{12}{0}\selectfont] {$\truelabel^{(2)}$}; 
				\draw[fill=white] (y3) circle (0.1) node[above, yshift=0.2cm, font=\fontsize{12}{0}\selectfont] {$\truelabel^{(3)}$}; 
				
				\draw[dashed] (x1) -- (hx1);     
				\draw[dashed] (x2) -- (y2);
				\draw[dashed] (x3) -- (hx3);
				
				\draw[<->, line width=1pt] (hx1) -- (y1) node[above, yshift=-1.5cm, rotate=90, font=\fontsize{12}{0}\selectfont] {$\truelabel^{(1)} - h(\feature^{(1)})$};
				\draw[<->, line width=1pt] (hx2) -- (y2);
				\draw[<->, line width=1pt] (hx3) -- (y3);
				
			\end{tikzpicture}
			\caption{We can evaluate the quality of a particular predictor $h \in \hypospace$ by measuring the 
				prediction error $\truelabel- h(\feature)$ obtained for a labeled \gls{datapoint} $(\feature,\truelabel)$. }
		\label{fig_emp_error}
	\end{figure}

\newpage
\section{Exercises} 
\vspace*{-10mm}
\
\begin{exercise}[Perfect Prediction.]
Consider \gls{datapoint}s that are characterized by a single numeric feature $\feature\!\in\!\mathbb{R}$ and a numeric label 
$\truelabel\!\in\mathbb{R}$. We use a ML method to learn a hypothesis map $h: \mathbb{R} \rightarrow \mathbb{R}$ 
based on a \gls{trainset} consisting of three \gls{datapoint}s $$(\feature^{(1)}=1,\truelabel^{(1)} = 3), (\feature^{(2)}=4,\truelabel^{(2)}=-1), (\feature^{(3)}=1,\truelabel^{(3)}=5).$$ 
Is there any chance for the ML method to learn a hypothesis map that perfectly fits the  
\gls{datapoint}s such that $h\big( \feature^{(\sampleidx)} \big) = \truelabel^{(\sampleidx)}$ for $\sampleidx=1,\ldots,3$. 
Hint: Try to visualize the \gls{datapoint}s in a \gls{scatterplot} and various hypothesis maps (see Figure \ref{fig_three_maps_example}). 
\end{exercise} 

\begin{exercise}[Temperature Data.]
Consider a \gls{dataset} of daily air temperatures $\feature^{(1)},\ldots,\feature^{(\samplesize)}$ measured at the 
\gls{fmi} observation station ``Utsjoki Nuorgam'' during 01.12.2019 and 29.02.2020. Thus, $\feature^{(1)}$ is the daily 
temperature measured on 01.12.2019, $\feature^{(2)}$ is the daily temperature measure don 02.12.2019, and 
$\feature^{(\samplesize)}$ is the daily temperature measured on 29.02.2020. You can download this \gls{dataset} from the link \url{https://en.ilmatieteenlaitos.fi/download-observations}. ML methods often determine few parameters to 
characterize large collections of \gls{datapoint}s. Compute, for the above temperature measurement dataset, 
the following quantities: 
\begin{itemize}
	\item the \gls{minimum} $A \defeq \min_{\sampleidx=1,\ldots,\samplesize} \feature^{(\sampleidx)}$
	\item the \gls{maximum} $B \defeq \max_{\sampleidx=1,\ldots,\samplesize} \feature^{(\sampleidx)}$
	\item the average $C \defeq (1/\samplesize) \sum_{\sampleidx=1,\ldots,\samplesize} \feature^{(\sampleidx)}$
	\item the standard deviation $D \defeq \sqrt{(1/\samplesize)\sum_{\sampleidx=1,\ldots,\samplesize} \big( \feature^{(\sampleidx)}-C \big)^2}$
\end{itemize} 
\end{exercise}

\begin{exercise}[Deep Learning on Raspberry PI.]
Consider the tiny desktop computer ``RaspberryPI'' equipped with a total of $8$ Gigabytes memory \cite{DLRaspberryPI}. 
We want implement a ML algorithm that learns a hypothesis map that is represented 
by a deep \gls{ann} involving $\featurelen=10^6$ numeric parameters. Each parameter 
is quantized using $8$ bits ($=1$ Byte). How many different hypotheses can we store at most on a 
RaspberryPI computer? (You can assume that $1 {\rm Gigabyte} = 10^{9} {\rm Bytes}$.)
\end{exercise} 

\begin{exercise}[Ensembles.]
For some applications it can be a good idea to not learn a single hypothesis but to learn 
a whole ensemble of hypothesis maps $h^{(1)},\ldots,h^{(\augparam)}$. These hypotheses 
might even belong to different \gls{hypospace}s, $h^{(1)} \in \hypospace^{(1)},\ldots,h^{(\augparam)} \in \hypospace^{(\augparam)}$. 
These \gls{hypospace}s can be arbitrary except that they are defined for the same feature space and \gls{labelspace}. 
Given such an ensemble we can construct a new (``meta'') hypothesis $\tilde{h}$ by combining (or aggregating) the individual predictions obtained 
from each hypothesis, 
\begin{equation} 
	\label{equ_def_ensemble}
	\tilde{h}(\featurevec) \defeq a\big( h^{(1)}(\featurevec), \ldots,h^{(\augparam)}(\featurevec) \big). 
\end{equation}
Here, $a(\cdot)$ denotes some given (fixed) combination or aggregation function. One example for such 
an aggregation function is the average $a\big( h^{(1)}(\featurevec), \ldots,h^{(\augparam)}(\featurevec) \big) \defeq (1/\augparam) \sum_{\augidx=1}^{\augparam} h^{(\augidx)}(\featurevec)$. We obtain a new ``meta'' \gls{hypospace} $\widetilde{\hypospace}$, 
that consists of all hypotheses of the form \eqref{equ_def_ensemble} with $h^{(1)} \in \hypospace^{(1)},\ldots,h^{(\augparam)} \in \hypospace^{(\augparam)}$.
Which conditions on the aggregation function $a(\cdot)$ and the individual \gls{hypospace}s $\hypospace^{(1)},\ldots,\hypospace^{(\augparam)}$ 
ensure that $\widetilde{\hypospace}$ contains each individual \gls{hypospace}, i.e.,  $\hypospace^{(1)},\ldots,\hypospace^{(\augparam)} \subseteq  \widetilde{\hypospace}$. 
\end{exercise} 

\begin{exercise}[How Many Features?]
\label{ex_compML_shazaam}
Consider the ML problem underlying a music information retrieval smartphone app \cite{ShazamPaper}. 
Such an app aims at identifying a song title based on a short audio recording of a song interpretation. 
Here, the feature vector $\featurevec$ represents the sampled audio signal and the label $\truelabel$
is a particular song title out of a huge music database. What is the length $\featuredim$ of the feature 
vector $\featurevec \in \mathbb{R}^{\featuredim}$ if its entries are the signal amplitudes of a $20$-second 
long recording which is sampled at a rate of $44$ kHz?
\end{exercise}

\begin{exercise}[Multilabel Prediction.]
\label{ex_ch2_multilabel}
Consider \gls{datapoint}s that are characterized by a feature vector $\featurevec \in \mathbb{R}^{10}$ and 
a vector-valued label $\labelvec \in \mathbb{R}^{30}$. Such vector-valued labels arise in \gls{multilabelclass} 
problems. We want to predict the label vector using a linear predictor map 
\begin{equation}
	\label{equ_lin_predictor_multilabel}
	\vh(\featurevec) = \mathbf{W} \featurevec \mbox{ with some matrix } \mathbf{W} \in \mathbb{R}^{30 \times 10}. 
\end{equation} 
How many different linear predictors \eqref{equ_lin_predictor_multilabel} are there ? $10$, $30$, $40$, or infinite?
\end{exercise}

\begin{exercise}[Average Squared Error Loss as Quadratic Form]
\label{ex_2_0}
Consider the \gls{hypospace} constituted by all linear maps $h(\featurevec) = \weights^{T} \featurevec$ with some 
weight vector $\weights \in \mathbb{R}^{\featuredim}$. We try to find the best linear map by minimizing the average 
squared error loss (the \gls{emprisk}) incurred on labeled \gls{datapoint}s (\gls{trainset}) 
$(\featurevec^{(1)},\truelabel^{(1)}),(\featurevec^{(2)},\truelabel^{(2)}),\ldots,(\featurevec^{(\samplesize)},\truelabel^{(\samplesize)})$.  
Is it possible to represent the resulting \gls{emprisk} as a convex quadratic function$
f(\weights) = \weights^{T} \mathbf{C} \weights + \vb \weights + c$? If this is possible, how are the matrix 
$\mathbf{C}$, vector $\vb$ and constant $c$ related to the features and labels of \gls{datapoint}s in the \gls{trainset}? 
\end{exercise}

\begin{exercise}[Find Labeled Data for Given Empirical Risk.]
\label{ex_2_1}
Consider linear \gls{hypospace} consisting of linear maps $h^{(\weights)}(\featurevec) = \weights^{T} \featurevec$ that are 
parametrized by a weight vector $\weights$. We learn an optimal weight vector by minimizing the average 
squared error loss $f(\weights) = \emperror \big( h^{(\weights)} | \dataset\big)$ incurred by $h^{(\weights)}(\featurevec)$ 
on the \gls{trainset} $\dataset = \big(\featurevec^{(1)},\truelabel^{(1)}\big),\ldots,\big(\featurevec^{(\samplesize)},\truelabel^{(\samplesize)}\big)$. 
Is it possible to reconstruct the \gls{dataset} $\dataset$ just from knowing the function $f(\weights)$?. 
Is the resulting labeled training data unique or are there different \gls{trainset}s that could have 
resulted in the same \gls{emprisk} function? 
Hint: Write down the \gls{trainerr} $f(\weights)$ in the form $f(\weights) = \weights^{T} \mathbf{Q} \weights + c + \vb^{T} \weights$ 
with some matrix $\mathbf{Q}$, vector $\vb$ and scalar $c$ that might depend on 
the features and labels of the training \gls{datapoint}s. 
\end{exercise}

\begin{exercise}[Dummy Feature Instead of Intercept.]
\label{sec_dummy_feature}
Show that any hypothesis map of the form $h(\feature) = \weight_{1} \feature +\weight_{0}$ can be obtained 
from the concatenation of a feature map $\featuremap: \feature \mapsto \rawfeaturevec$ with the linear 
map $\tilde{h}(\rawfeaturevec) \defeq \widetilde{\weights}^{T} \rawfeaturevec$ using parameter vector $\widetilde{\weights} = \big( \weight_{1}, \weight_{0} \big)^{T}  \in \mathbb{R}^{2}$.  
\end{exercise}

\begin{exercise}[Approximate Non-Linear Maps Using Indicator Functions for Feature Maps.]
\label{ex_2_2}
Consider an ML application generating \gls{datapoint}s characterized by a scalar feature $x \in \mathbb{R}$ 
and numeric label $\truelabel \in \mathbb{R}$. We construct a non-linear map by first transforming the 
feature $\feature$ to a new feature vector $\rawfeaturevec=(\featuremap_{1}(\feature),\featuremap_{2}(\feature),\featuremap_{3}(\feature),\featuremap_{4}(\feature))^{T} \in \mathbb{R}^{4}$. 
The components $\featuremap_{1}(\feature),\ldots,\featuremap_{4}(\feature)$ are indicator functions of intervals 
$[-10,-5), [-5,0),[0,5),[5,10]$. In particular, $\phi_{1}(\feature) = 1$ for $\feature \in [-10,-5)$ and $\phi_{1}(\feature)=0$ otherwise. 
We obtain a \gls{hypospace} $\hypospace^{(1)}$ by collecting all maps from feature $\feature$ to predicted label $\hat{\truelabel}$ 
that can written as a a weighted linear combination $\weights^{T}\rawfeaturevec$ (with some parameter vector $\weights$) 
of the transformed features. Which of the following hypothesis maps belong to $\hypospace^{(1)}$?


\tikzset{global scale/.style={
		scale=#1,
		every node/.append style={scale=#1}
	}
}
\begin{tikzpicture}[global scale = 1]             
	
	\foreach \x in {-4,4}
	{
		
		\draw[->, very thick, xshift=\x cm](-3,0)--(3,0) node[above, xshift=0.5cm, yshift=-0.2cm, font=\fontsize{12}{0}\selectfont] {$x$};    
		\draw[->, very thick, xshift=\x cm](0,-0.5)--(0,2.5) node[above, yshift=0.2cm, font=\fontsize{12}{0}\selectfont] {$h(x)$};
		\node[font=\fontsize{12}{0}\selectfont, xshift=\x cm] (-10) at (-2,-0.8) {-10};
		\node[font=\fontsize{12}{0}\selectfont, xshift=\x cm] (-5) at (-1,-0.8) {-5};
		\node[font=\fontsize{12}{0}\selectfont, xshift=\x cm] (0) at (0,-0.8) {0};
		\node[font=\fontsize{12}{0}\selectfont, xshift=\x cm] (5) at (1,-0.8) {5};
		\node[font=\fontsize{12}{0}\selectfont, xshift=\x cm] (10) at (2,-0.8) {10};
		\draw[blue!30, dashed, thick, xshift=\x cm](-10) -- +(0,3);     
		\draw[blue!30, dashed, thick, xshift=\x cm](-5) -- +(0,3);
		\draw[blue!30, dashed, thick, xshift=\x cm](5) -- +(0,3);
		\draw[blue!30, dashed, thick, xshift=\x cm](10) -- +(0,3);
		
	};
	
	\draw[blue, line width=2pt, , xshift=-4 cm](-3,0)--(-2,0)--(-2,1)--(-1,1)--(-1,0)--(2.5,0);
	\node[font=\fontsize{12}{0}\selectfont, xshift=-4 cm] at (0,-2) {(a)};
	\draw[blue, line width=2pt, , xshift=4 cm](-3,0)--(-2,0)--(-1,1);
	\draw[blue, line width=2pt, , xshift=4 cm](-1,0)--(2.5,0);
	\node[font=\fontsize{12}{0}\selectfont, xshift=4 cm] at (0,-2) {(b)};
	
\end{tikzpicture}

%
%
\end{exercise}

\begin{exercise}[Python \Gls{hypospace}.]
\label{ex_2_3}
Consider the source codes below for five different Python functions that read in the numeric 
feature $\feature$, perform some computations that result in a prediction $\hat{\truelabel}$. 
How large is the \gls{hypospace} that is constituted by all maps that can be represented by 
one of those Python functions. 

\begin{minipage}{0.3\textwidth}
	\begin{lstlisting}
		def func1(x):
		hat_y = 5*x+3
		return hat_y
	\end{lstlisting}
\end{minipage}
\hspace*{4mm}
\begin{minipage}{0.3\textwidth}
	\begin{lstlisting}
		def func2(x):
		tmp = 3*x+3 
		hat_y = tmp+2*x
		return hat_y
	\end{lstlisting}
\end{minipage}
\hspace*{4mm}
\begin{minipage}{0.3\textwidth}
	\begin{lstlisting}
		def func3(x):
		tmp = 3*x+3 
		hat_y = tmp-2*x
		return hat_y
	\end{lstlisting}
\end{minipage}

\begin{minipage}{0.3\textwidth}
	\begin{lstlisting}
		def func4(x):
		tmp = 3*x+3 
		hat_y = tmp-2*x+4
		return hat_y
	\end{lstlisting}
\end{minipage}
\hspace*{4mm}
\begin{minipage}{0.3\textwidth}
	\begin{lstlisting}
		def func5(x):
		tmp = 3*x+3 
		hat_y = 4*tmp-2*x
		return hat_y
	\end{lstlisting}
\end{minipage}
\end{exercise}

\begin{exercise}[A Lot of Features.]
\label{ex_2_toomanyfeatures}
One important application domain for ML methods is healthcare. Here, \gls{datapoint}s represent 
human patients that are characterized by health-care records. These records might contain physiological 
parameters, CT scans along with various diagnoses provided by healthcare professionals. Is it a 
good idea to use every data field of a healthcare record as features of the \gls{datapoint} ? 
\end{exercise}

\begin{exercise}[Over-Parametrization.]
\label{ex_2_overparm}
Consider \gls{datapoint}s characterized by feature vectors $\featurevec \in \mathbb{R}^{2}$ and 
a numeric label $\truelabel \in \mathbb{R}$. We want to learn the best 
predictor out of the \gls{hypospace} 
$$ \hypospace = \big\{ h(\featurevec) = \featurevec^{T} \mathbf{A} \weights: \weights \in \mathcal{S} \}. $$
Here, we used the matrix 
$\mathbf{A} = \begin{pmatrix} 1 & -1 \\ -1 & 1 \end{pmatrix}$ and 
the set $$\mathcal{S} = \big\{ (1,1)^{T}, (2,2)^{T}, (-1,3)^{T}, (0,4)^{T} \big\} \subseteq \mathbb{R}^{2}.$$ 
What is the cardinality of the \gls{hypospace} $\hypospace$, i.e., how many 
different predictor maps does $\hypospace$ contain?
\end{exercise}

\begin{exercise}[Squared Error Loss.]
\label{ex_2_4}
Consider a \gls{hypospace} $\hypospace$ constituted by three predictors $h^{(1)}(\cdot), h^{(2)}(\cdot),h^{(3)}(\cdot)$. 
Each predictor $h^{(\featureidx)}(\feature)$ is a real-valued function of a real-valued argument $\feature$. 
Moreover, for each $\featureidx \in \{1,2,3\}$, 
\begin{equation} 
h^{(\featureidx)}(\feature) = \begin{cases} 0 & \mbox{ if } \feature^2 \leq j \\   j & \mbox{ otherwise.} \end{cases} 
\end{equation}
Can you tell which of these hypothesis is optimal in the sense of having smallest average 
squared error \gls{loss} on the three \gls{datapoint}s $(\feature=1/10,\truelabel=3)$, $(0,0)$ and $(1,-1)$. 
\end{exercise}

\begin{exercise}[Classification Loss.]
\label{ex_2_classif_loss}
The Figure \ref{fig_class_loss} depicts different \gls{lossfunc}s for a fixed \gls{datapoint} with label $\truelabel=1$ 
and varying hypothesis $h \in \hypospace$. How would Figure \ref{fig_class_loss} change if we 
evaluate the same l\gls{lossfunc}s for another \gls{datapoint} $\datapoint=(\feature,\truelabel)$ 
with label $\truelabel=-1$?
\end{exercise}

\begin{exercise}[Intercept Term.]
\label{ex_2_5}
\Gls{linreg} methods model the relation between the label $\truelabel$ and feature $\feature$ 
of a \gls{datapoint} as $\truelabel = h(\feature) + e$ with some small additive term $e$. The predictor 
map $h(\feature)$ is assumed to be linear $h(\feature) =\weight_1 \feature + \weight_0$. The parameter  
$\weight_{0}$ is sometimes referred to as the intercept (or bias) term. Assume we know for a given 
linear predictor map its values $h(\feature)$ for $\feature=1$ and $\feature=3$. Can you determine 
the weights $\weight_{1}$ and $\weight_{0}$ based on $h(1)$ and $h(3)$?
\end{exercise}

\begin{exercise}[Picture Classification.]
\label{ex_2_6}
Consider a huge collection of outdoor pictures you have taken during 
your last adventure trip. You want to organize these pictures as three 
categories (or classes) \emph{dog}, \emph{bird} and \emph{fish}. How 
could you formalize this task as a ML problem? 
\end{exercise}

\begin{exercise}[Maximal \Gls{hypospace}.]
\label{ex_2_7}
Consider \gls{datapoint}s characterized by a single real-valued feature $\feature$ and a single 
real-valued label $\truelabel$. How large is the largest possible \gls{hypospace} of predictor maps 
$h(\feature)$ that read in the feature value of a \gls{datapoint} and deliver a real-valued prediction $\hat{\truelabel}=h(\feature)$ ? 
\end{exercise}

\begin{exercise}[A Large but Finite \Gls{hypospace}.]
\label{ex_2_8}
Consider \gls{datapoint}s whose features are $10 \times 10$ pixel black-and-white (bw) images. 
Besides the pixels, each \gls{datapoint} is also characterized by a binary label $\truelabel \in \{0,1\}$. 
Consider the \gls{hypospace} which is constituted by all maps that take a $10 \times 10$ pixel bw image 
as input and deliver a prediction for the label. How large is this \gls{hypospace}? 
\end{exercise}

\begin{exercise}[Size of Linear \Gls{hypospace}.]
\label{ex_size_lin_hypospace} 
Consider a \gls{trainset} of $\samplesize$ \gls{datapoint}s with feature vectors $\featurevec^{(\sampleidx)} \in \mathbb{R}^{\featuredim}$ and 
numeric labels $\truelabel^{(1)},\ldots,\truelabel^{(\samplesize)}$. The feature vectors and labels of the \gls{datapoint}s in 
the \gls{trainset} are arbitrary except that we assume the feature matrix $\featuremtx = \big(\featurevec^{(1)},\ldots,\featurevec^{(\samplesize)} \big)$ 
is full rank. What condition on $\samplesize$ and $\featurelen$ guarantees that we can find a linear 
predictor $h(\featurevec) = \weights^{T} \featurevec$ that perfectly fits the 
\gls{trainset}, i.e., $\truelabel^{(1)} = h\big(\featurevec^{(1)} \big),\ldots, \truelabel^{(\samplesize)} = h\big(\featurevec^{(\samplesize)} \big)$. 
\end{exercise}

\begin{exercise}[Classification with Imbalanced Data.]
\label{ex_imbalanced_data_binary_class} 
Consider a \gls{dataset} $\dataset$ of $\samplesize$ \gls{datapoint}s with feature vectors $\featurevec^{(\sampleidx)} \in \mathbb{R}^{\featuredim}$ 
and discrete-valued labels $\truelabel^{(\sampleidx)} \in \{1,2,\ldots,10\}$. The data is highly imbalanced, 
more than $90$ percent of \gls{datapoint}s have a label $\truelabel =1$. We learn a hypothesis out 
of the \gls{hypospace} $\hypospace'$ that is constituted by the ten maps $h^{(\clusteridx)}(\featurevec)= \clusteridx$ for $\clusteridx=1,2,\ldots,10$. Is there a hypothesis $h\in \hypospace'$ whose average $0/1$ \gls{loss} 
on $\dataset$ does not exceed $0.3$ ?
\end{exercise}

\begin{exercise}[Accuracy and Average \Gls{logloss}.]
\label{ex_different_loss} 
Consider a dataset $\dataset$ that consists of $\samplesize$ \gls{datapoint}s , indexed by $\sampleidx=1,\ldots,\samplesize$. 
Each \gls{datapoint} is characterized by a feature vector $\featurevec^{(\sampleidx)} \in \mathbb{R}^{2}$ and by a binary label $\truelabel^{(\sampleidx)} \in \{-1,1\}$. We use a linear hypothesis map $h^{(\weights)}(\featurevec) = \weights^{T} \featurevec$ 
to classify \gls{datapoint}s according to $\hat{\truelabel} = 1$ if $h^{(\weights)}(\featurevec) \geq 0$ and $\hat{\truelabel}=-1$ 
otherwise. Two popular quality measures of a hypothesis are the accuracy and the average \gls{logloss} $(1/\samplesize) \sum_{\sampleidx} \loss{\big(\featurevec^{(\sampleidx)},\truelabel^{(\sampleidx)} \big)}{h^{(\weights)}}$ with 
the \gls{logloss} \eqref{equ_log_loss}. The accuracy of a hypothesis $h^{(\weights)}(\featurevec)$ is defined as $1- (1/\samplesize) \sum_{\sampleidx} \loss{\big(\featurevec^{(\sampleidx)},\truelabel^{(\sampleidx)} \big)}{h^{(\weights)}}$ with the $0/1$ loss \eqref{equ_def_0_1}. 
Loosely speaking, the accuracy of a hypothesis is ``one minus the average $0/1$ loss ''. Can you construct a 
specific dataset with arbitrary but finite size $\samplesize$ such that there are two different linear 
hypotheses $h^{(\weights)}$ and $h^{(\weights')}$ with accuracy and average \gls{logloss} of $h^{(\weights)}$ 
being strictly larger than accuracy and average \gls{logloss} of $h^{(\weights')}$. 
\end{exercise}

\chapter{The Landscape of ML} 
\label{ch_some_examples}
\begin{figure}[htbp]
	\resizebox{1\textwidth}{!}{
		\begin{tikzpicture}[auto]
			\coordinate (OR) at (0.00, 0.0);  
			\coordinate (RIGHT) at (16,0); 
			
			\node[](a) at (OR) {};
			\node[](b) at (RIGHT) {};
			
			\node[]  at  ($ (b) + (1,0) $)  {\gls{model}};
			\node[above=0.5cm of OR] (absloss) {} ;
			\node[above=1.5cm of absloss] (sqerror) {} ;
			\node[above=1.5cm of sqerror] (regsqloss) {} ;
			\node[above=1.5cm of regsqloss] (logloss) {} ;
			\node[above=1.5cm of logloss] (hingeloss) {} ;
			\node[above=1.5cm of hingeloss] (regret) {} ;
			\node[above=1.5cm of regret] (zeroone) {} ;
			\coordinate (TOP) at ($ (zeroone) + (0,1.5) $) ; 
			\node[](c) at (TOP) {};
			\node[align=left]  at  ($ (absloss) - (1,0) $)  {absolute\\loss};
			\node[align=left]  at  ($ (logloss) - (1,0) $)  {logistic\\loss};
			\node[align=left]  at  ($ (sqerror) - (1,0) $)  {squared\\error};
			\node[align=left,text width=2cm]  at  ($ (regsqloss) - (1,0) $)  {regularized squared error};
			\node[align=left]  at  ($ (hingeloss) - (1,0) $)  {hinge\\loss};
			\node[align=left]  at  ($ (regret) - (1,0) $)  {regret};
			\node[align=left]  at  ($ (zeroone) - (1,0) $)  {$0/1$ loss};
			\node[anchor=east]  at  ($ (c) + (0,0.1) $)  {\gls{lossfunc}};
			\node[right=2cm of OR] (linpred) {} ;
			\node[right=5cm of OR] (upgrlinpred) {} ;
			\node[right=15cm of OR] (neuralnets) {} ;
			\node[right=10cm of OR] (piecewise) {} ;
			\node[text width=3cm,above=0.01cm]  at  ($ (upgrlinpred) -  (0,1.2) $)  {upgraded linear maps \eqref{equ_conc_featuremap_linear}}; 
			\node[]  at  ($ (linpred) - (0,0.5) $)  {linear maps};
			\node[]  at  ($ (neuralnets) - (0,0.5) $)  {\gls{ann}};
			\node[align=left]  at  ($ (piecewise) - (0,0.5) $)  {piecewise \\ constant};
			\node[draw,text width=2cm]  at  ($ (linpred) + (absloss) $)  {Sec.\ \ref{sec_lad}};
			\node[draw,text width=2cm]  at  ($ (linpred) + (sqerror) $)  {Sec.\ \ref{sec_lin_reg}};
			\node[draw,text width=2cm]  at  ($ (upgrlinpred) + (sqerror) $)  {Sec.\ \ref{sec_polynomial_regression} \\ Sec.\ \ref{sec_linbasreg}\\ Sec.\ \ref{sec_kernel_methods}};
			\node[draw,text width=2cm]  at  ($ (upgrlinpred) + (logloss) $)  {Sec.\ \ref{sec_kernel_methods}};
			\node[draw,text width=2cm]  at  ($ (upgrlinpred) + (hingeloss) $)  {Sec.\ \ref{sec_kernel_methods}};
			\node[draw,text width=2cm]  at  ($ (linpred) + (logloss) $)  {Sec.\ \ref{sec_LogReg}};
			\node[draw,text width=2cm]  at  ($ (linpred) + (hingeloss) $)  {Sec.\ \ref{sec_SVM}};
			\node[draw,text width=2cm]  at  ($ (linpred) + (regsqloss) $)  {Sec.\ \ref{sec_lasso}};
			\node[draw,text width=2cm]  at  ($ (upgrlinpred) + (regsqloss) $)  {Sec.\ \ref{sec_kernel_methods}};
			\node[draw,text width=2cm]  at  ($ (linpred) + (regret) $)  {Sec.\ \ref{sec_lin_ucb}};
			\node[draw,text width=2cm]  at  ($ (linpred) + (zeroone) $)  {Sec.\ \ref{sec_NaiveBayes}};
			\node[draw,text width=2cm,,align=left]  at  ($ (piecewise) + (zeroone) $)  {Sec.\ \ref{sec_decision_trees}\\ Sec.\ \ref{sec_nearest_neighbour_methods}};
			\node[draw,text width=2cm,,align=left]  at  ($ (piecewise) + (sqerror) $)  {Sec.\ \ref{sec_decision_trees}\\ Sec.\ \ref{sec_nearest_neighbour_methods}};
			\node[draw,text width=2cm]  at  ($ (neuralnets) + (regret) $)  {Sec.\ \ref{sec_deep_learning}};
			\node[draw,text width=2cm,align=left]  at  ($ (neuralnets) + (logloss) $)  {Sec.\ \ref{sec_deep_learning}};
			\draw[black,line width=1] ($ (linpred) - (0,0.1) $) to ($ (linpred) + (0,0.1) $) ;
			\draw[black,line width=1] ($ (neuralnets) - (0,0.1) $) to ($ (neuralnets) + (0,0.1) $) ;
			\draw[black,line width=1] ($ (upgrlinpred) - (0,0.1) $) to ($ (upgrlinpred) + (0,0.1) $) ;
			\draw[black,line width=1] ($ (piecewise) - (0,0.1) $) to ($ (piecewise) + (0,0.1) $) ;
			\draw[black,line width=1] ($ (absloss) - (0.1,0) $) to ($ (absloss) + (0.1,0) $) ;
			\draw[black,line width=1] ($ (sqerror) - (0.1,0) $) to ($ (sqerror) + (0.1,0) $) ;
			\draw[black,line width=1] ($ (regsqloss) - (0.1,0) $) to ($ (regsqloss) + (0.1,0) $) ;
			\draw[black,line width=1] ($ (logloss) - (0.1,0) $) to ($ (logloss) + (0.1,0) $) ;
			\draw[black,line width=1] ($ (hingeloss) - (0.1,0) $) to ($ (hingeloss) + (0.1,0) $) ;
			\draw[black,line width=1] ($ (regret) - (0.1,0) $) to ($ (regret) + (0.1,0) $) ;
			\draw[black,line width=1] ($ (zeroone) - (0.1,0) $) to ($ (zeroone) + (0.1,0) $) ;
			\draw[->,black,line width=1] ($ (a) - (1,0) $) to (b);
			\draw[->,black,line width=1] ($ (a) - (0,0.4) $) to (c);
		\end{tikzpicture}
	}
	\caption{ML methods fit a model to data by minimizing a \gls{lossfunc}. Different ML 
		methods use different design choices for data, \gls{model} and loss.} \label{fig_ML_methods_loss_hypo_2d}
\end{figure}
As discussed in Chapter \ref{ch_Elements_ML}, ML methods combine 
three main components: 
\begin{itemize} 
	\item  a set of \gls{datapoint}s that are characterized by \gls{features} and \gls{label}s  
	\item a \gls{model} or \gls{hypospace} $\hypospace$ that consists of different hypotheses $h \in \hypospace$. 
	\item a \gls{lossfunc} to measure the quality of a particular hypothesis $h$. 
\end{itemize} 
Each of these three components involves design choices for the representation of data, their 
features and labels, the \gls{model} and \gls{lossfunc}. This chapter details the high-level design choices 
used by some of the most popular ML methods. Figure \ref{fig_ML_methods_loss_hypo_2d} depicts 
these ML methods in a two-dimensional plane whose horizontal axes represents different \gls{hypospace}s 
and the vertical axis represents different \gls{lossfunc}s. 

To obtain a practical ML method we also need to combine the above components. 
The basic principle of any ML method is to search the model for a hypothesis that incurs 
minimum \gls{loss} on any \gls{datapoint}. Chapter \ref{ch_Optimization} will then discuss 
a principled way to turn this informal statement into actual ML algorithms that could be 
implemented on a computer. 

\section{Linear Regression} 
\label{sec_lin_reg}

Consider \gls{datapoint}s characterized by feature vectors $\featurevec \in \mathbb{R}^{\featuredim}$ and numeric 
label $\truelabel \in \mathbb{R}$. \index{linear regression} \Gls{linreg} methods learn a hypothesis out of 
the linear \gls{hypospace} 
\begin{align}
	\label{equ_lin_hypospace}
	\hypospace^{(\featuredim)} & \defeq \{ h^{(\weights)}: \mathbb{R}^{\featuredim}\!\rightarrow\!\mathbb{R}: h^{(\weights)}(\featurevec)\!=\!\weights^{T} \featurevec \mbox{ with some parameter vector } \weights \in \mathbb{R}^{\featuredim} \}.
\end{align}  
Figure \ref{fig_three_maps_example} depicts the graphs of some maps from $\hypospace^{(2)}$ for \gls{datapoint}s 
with feature vectors of the form $\featurevec = (1,\feature)^{T}$. The quality of a particular predictor $h^{(\weights)}$ 
is measured by the squared error loss \eqref{equ_squared_loss}. Using labeled data 
$\dataset =\{ (\featurevec^{(\sampleidx)},\truelabel^{(\sampleidx)}) \}_{\sampleidx=1}^{\samplesize}$, 
\gls{linreg} learns a linear predictor $\hat{h}$ which minimizes the average squared error loss  \eqref{equ_squared_loss}, 
or ``mean squared error'',
\begin{align} 
	\label{equ_opt_pred_linreg}
	\hat{h}  & = \argmin_{h \in \hypospace^{(\featuredim)}  }\emperror(h|\dataset)  \stackrel{\eqref{eq_def_emp_error_101}}{=} \argmin_{h \in \hypospace^{(\featuredim)}  }  (1/\samplesize) \sum_{\sampleidx=1}^{\samplesize} (\truelabel^{(\sampleidx)} - h(\featurevec^{(\sampleidx)}))^{2}.
\end{align} 

Since the \gls{hypospace} $\hypospace^{(\featuredim)} $ is parametrized by 
the parameter vector $\weights$ (see \eqref{equ_lin_hypospace}), we can rewrite \eqref{equ_opt_pred_linreg} 
as an optimization of the parameter vector $\weights$, 
\begin{align} 
	\label{equ_opt_weight_vector_linreg_weight}
	\widehat{\weights} &  = \argmin_{\weights \in \mathbb{R}^{\featuredim}} (1/\samplesize) \sum_{\sampleidx=1}^{\samplesize} (\truelabel^{(\sampleidx)} - h^{(\weights)}(\featurevec^{(\sampleidx)}))^{2} \nonumber \\
	& \stackrel{h^{(\weights)}(\featurevec) = \weights^{T} \featurevec}{=} \argmin_{\weights \in \mathbb{R}^{\featuredim}} (1/\samplesize) \sum_{\sampleidx=1}^{\samplesize} (\truelabel^{(\sampleidx)} - \weights^{T} \featurevec^{(\sampleidx)})^{2}.
\end{align} 
The optimization problems \eqref{equ_opt_pred_linreg} and \eqref{equ_opt_weight_vector_linreg_weight} 
are equivalent in the following sense: Any optimal parameter vector $\widehat{\weights}$ which solves 
\eqref{equ_opt_weight_vector_linreg_weight}, can be used to construct an optimal predictor $\hat{h}$, 
which solves \eqref{equ_opt_pred_linreg}, via $\hat{h}(\featurevec) = h^{(\widehat{\weights})}(\featurevec) = \big(\widehat{\weights}\big)^{T} \featurevec$.

\section{Polynomial Regression} 
\label{sec_polynomial_regression}


Consider an ML problem involving \gls{datapoint}s which are characterized by a 
single numeric feature $\feature \in  \mathbb{R}$ (the feature space is $\featurespace= \mathbb{R}$) 
and a numeric label $\truelabel \in \mathbb{R}$ (the \gls{labelspace} is $\labelspace= \mathbb{R}$). 
We observe a bunch of labeled \gls{datapoint}s which are depicted in Figure \ref{fig_scatterplot_poly}. 

\begin{figure}[htbp]
	\begin{center}
		\begin{tikzpicture}[scale=0.8]
			\begin{axis}[
				axis x line=middle,
				axis y line=middle,
				ylabel near ticks,
				xlabel near ticks,
				enlarge y limits=true,
				width=10cm, height=8cm,     
				grid = major,
				grid style={dashed, gray!30},
				ylabel=label $\truelabel$,
				xlabel=feature $\feature$,
				]        
				\addplot[only marks] table [x=BTCNorm, y=ETHNorm, col sep = comma] {CryptoScatter.csv};
			\end{axis}
		\end{tikzpicture}
	\end{center}
	\caption{A \gls{scatterplot} that depicts a set of \gls{datapoint}s $(\feature^{(1)},\truelabel^{(1)}),\ldots,(\feature^{(\samplesize)},\truelabel^{(\samplesize)})$. 
		The $\sampleidx$th \gls{datapoint} is depicted by a dot whose coordinates 
		are the feature $\feature^{(\sampleidx)}$ and label $\truelabel^{(\sampleidx)}$ of that \gls{datapoint}.} 
	\label{fig_scatterplot_poly}
\end{figure}

Figure \ref{fig_scatterplot_poly} suggests that the relation $\feature \mapsto \truelabel$ between feature 
$\feature$ and label $\truelabel$ is highly non-linear. For such non-linear relations between features and 
labels it is useful to consider a \gls{hypospace} which is constituted by polynomial maps
\begin{align}
	\label{equ_def_poly_hyposapce}
	\hypospace^{(\featuredim)}_{\rm poly}& = \{ h^{(\weights)}: \mathbb{R} \rightarrow \mathbb{R}: h^{(\weights)}(\feature) = \sum_{\featureidx=1}^{\featuredim} \weight_{\featureidx} \feature^{\featureidx-1}, \nonumber \\ 
 &	\mbox{with some } \weights\!=\!(\weight_{1},\ldots,\weight_{\featuredim})^{T}\!\in\!\mathbb{R}^{\featuredim} \}. 
\end{align}
We can approximate any non-linear relation $\truelabel\!=\!h(\feature)$ with any desired level of 
accuracy using a polynomial $\sum_{\featureidx=1}^{\featuredim} \weight_{\featureidx} \feature^{\featureidx-1}$ of sufficiently large 
degree $\featuredim$.\footnote{The precise formulation of this statement is known as the 
	``Stone-Weierstrass Theorem'' \cite[Thm. 7.26]{RudinBookPrinciplesMatheAnalysis}.}

For \gls{linreg} (see Section \ref{sec_lin_reg}), we measure the quality of a predictor by the squared 
error loss \eqref{equ_squared_loss}. Based on labeled \gls{datapoint}s $\dataset =\{ (\feature^{(\sampleidx)},\truelabel^{(\sampleidx)}) \}_{\sampleidx=1}^{\samplesize}$, each having a scalar feature $\feature^{(\sampleidx)}$ and label $\truelabel^{(\sampleidx)}$, polynomial 
regression minimizes the average squared error loss (see \eqref{equ_squared_loss}):
\begin{equation} 
	\label{opt_hypo_poly}
	\min_{h \in \hypospace_{\rm poly}^{(\featuredim)} } (1/\samplesize) \sum_{\sampleidx=1}^{\samplesize} (\truelabel^{(\sampleidx)} - h^{(\weights)}(\feature^{(\sampleidx)}))^{2}.
\end{equation} 
It is customary to refer to the average squared error loss also as the \index{mean squared error} mean squared error.  

We can interpret polynomial regression as a combination of a feature map (transformation) 
(see Section \ref{sec_feature_space}) and \gls{linreg} (see Section \ref{sec_lin_reg}). 
Indeed, any polynomial predictor $h^{(\weights)} \in \hypospace_{\rm poly}^{(\featuredim)}$ 
is obtained as a concatenation of the \gls{featuremap}  
\begin{equation}
	\label{equ_poly_feature_map} 
	\featuremap(\feature) \mapsto (1,\feature,\ldots,\feature^{\featuredim})^{T} \in \mathbb{R}^{\featuredim+1}
\end{equation}
with some linear map $\tilde{h}^{(\weights)}: \mathbb{R}^{\featuredim+1} \rightarrow \mathbb{R}: \featurevec \mapsto \weights^{T} \featurevec$, i.e., 
\begin{equation}
	\label{equ_concact_phi_g_poly}
	h^{(\weights)}(\feature) = \tilde{h}^{(\weights)}(\featuremap(\feature)). 
\end{equation}

Thus, we can implement polynomial regression by first applying the \gls{featuremap} $\featuremap$ 
(see \eqref{equ_poly_feature_map}) to the scalar features $\feature^{(\sampleidx)}$, resulting in the transformed 
feature vectors 
\begin{equation} 
	\label{equ_def_poly_feature_vectors}
	\featurevec^{(\sampleidx)} = \featuremap \big(\feature^{(\sampleidx)}\big) = \big(1,\feature^{(\sampleidx)},\ldots,\big( \feature^{(\sampleidx)} \big)^{\featuredim-1} \big)^{T} \in \mathbb{R}^{\featuredim}, 
\end{equation} 
and then applying \gls{linreg} (see Section \ref{sec_lin_reg}) to these new feature vectors. 

By inserting \eqref{equ_concact_phi_g_poly} into \eqref{opt_hypo_poly}, we obtain a \gls{linreg}
problem \eqref{equ_opt_weight_vector_linreg_weight} with feature vectors \eqref{equ_def_poly_feature_vectors}. 
Thus, while a predictor $h^{(\vw)} \in \hypospace_{\rm poly}^{(\featuredim)}$ 
is a non-linear function $h^{(\vw)}(\feature)$ of the original feature $\feature$, it is a linear 
function $\tilde{h}^{(\weights)}(\featurevec) = \weights^{T}\featurevec$ (see \eqref{equ_concact_phi_g_poly}), 
of the transformed features $\featurevec$ \eqref{equ_def_poly_feature_vectors}. 

\section{Least Absolute Deviation Regression}
\label{sec_lad}

Learning a linear predictor by minimizing the average squared error loss incurred on training data is not 
robust against the presence of \gls{outlier}s. This sensitivity to outliers is rooted in the properties of the squared 
error loss $(\truelabel - h(\featurevec))^{2}$. Minimizing the average squared error forces the resulting 
predictor $\hat{\truelabel}$ to not be too far away from any \gls{datapoint}. However, it might be useful to 
tolerate a large prediction error $\truelabel - h ( \featurevec)$ for an unusual or exceptional \gls{datapoint} that 
can be considered an \gls{outlier}.  

Replacing the squared loss with a different \gls{lossfunc} can make the learning robust against \gls{outlier}s. 
One important example for such a ``robustifying'' \gls{lossfunc} is the \gls{huberloss} \cite{HuberRobustBook}
\begin{equation}
	\label{equ_def_Huber_loss}
	\loss{\big(\featurevec,\truelabel \big)}{h} = \begin{cases} (1/2) (\truelabel-h(\featurevec))^{2} & \mbox{ for } |\truelabel-h(\featurevec)| \leq   \varepsilon \\ 
		\varepsilon(|\truelabel-h(\featurevec)| - \varepsilon/2) & \mbox{ else. }\end{cases}
\end{equation}
Figure \ref{fig_Huber_loss} depicts the \gls{huberloss} as a function of the prediction error $\truelabel - h(\featurevec)$. 

\tikzset{global scale/.style={
		scale=#1,
		every node/.append style={scale=#1}
	}
}

\begin{figure}
\begin{center}
\begin{tikzpicture}[global scale = 1]  
	\draw[<->, very thick](4.5,0)--(0,0)--(0,3.5);
	\draw[very thick](0,-0.3)--(0,0)--(-4.5,0);
	\draw[red, very thick, domain=2:3] plot(\x, \x-1);
	\draw[red, very thick, domain=-3:-2] plot(\x, -\x-1);
	\draw[blue, very thick, domain=-2:2] plot(\x, 0.25*\x*\x);
	
	\draw[blue, very thick, <->](-2,-0.3)--(2,-0.3);
	\draw[blue](-2,0.7*1.4*1.4)--(-2,-0.5);
	\draw[blue](2,0.7*1.4*1.4)--(2,-0.5);
	\node[font=\fontsize{10}{0}\selectfont] at (0,-1) {squared error loss};
	\node[font=\fontsize{10}{0}\selectfont] at (4,0.5) {prediction error $\hat{\truelabel}-\truelabel$};
	\node[font=\fontsize{10}{0}\selectfont, red] at (3.5,2.5) {absolute difference loss};
	
\end{tikzpicture}
\end{center}
	\caption{The \gls{huberloss} \eqref{equ_def_Huber_loss} resembles the squared error loss \eqref{equ_squared_loss} 
	for small prediction error and the absolute difference loss for larger prediction errors.}
\label{fig_Huber_loss}
\end{figure}

The \index{Huber loss}\gls{huberloss} definition \eqref{equ_def_Huber_loss} contains a tuning parameter $\epsilon$. 
The value of this tuning parameter defines when a \gls{datapoint} is considered as an \gls{outlier}. 
Figure \ref{fig_huber_loss_scatter} illustrates the role of this parameter as the width of a band around a hypothesis map. 
The prediction error of this hypothesis map for \gls{datapoint}s within this band are measured 
used squared error loss \eqref{equ_squared_loss}. For \gls{datapoint}s outside this band (outliers) we use instead 
the absolute value of the prediction error as the resulting loss. 

\begin{figure}
	\begin{center}
		\begin{tikzpicture}[global scale = 1]  
			
			\draw[blue, fill=blue!5](-0.35, -0.35*0.4+0.6 - 0.03)--(4.95, 4.95*0.4+0.6 - 0.03)--(5.05, 5.05*0.4+0.6 - 0.7)--(-0.25, -0.25*0.4+0.6 - 0.7)--(-0.35, -0.35*0.4+0.6 - 0.03);
			\draw[blue, fill=blue!5](-0.4, -0.4*0.4+0.6 + 0.03)--(4.8, 4.8*0.4+0.6 + 0.03)--(4.7, 4.7*0.4+0.6 + 0.7)--(-0.5, -0.5*0.4+0.6 + 0.7)--(-0.4, -0.4*0.4+0.6 + 0.03);
			\draw[<->, very thick](6,0)--(0,0)--(0,3);
			\draw[very thick](0,-0.3)--(0,0)--(-0.3,0);
			\draw[red, very thick, domain=-0.3:5] plot(\x, 0.4*\x+0.6);     
			\draw[<->, very thick](4.2, 4.2*0.4+0.6 - 0.05)--(4.2, 4.2*0.4+0.6 - 0.7);
			\draw[color=blue!60, fill=blue!60](0.5, 0.5*0.4+0.6 - 0.2) circle (0.08);
			\draw[color=blue!60, fill=blue!60](1, 1*0.4+0.6 - 0.3) circle (0.08);
			\draw[color=blue!60, fill=blue!60](1.1, 1.1*0.4+0.6 + 0.15) circle (0.08);
			\draw[color=blue!60, fill=blue!60](1.5, 1.5*0.4+0.6 - 0.3) circle (0.08);
			\draw[color=blue!60, fill=blue!60](1.8, 1.8*0.4+0.6 + 0.4) circle (0.08);
			\draw[color=blue!60, fill=blue!60](2.2, 2.2*0.4+0.6 - 0.3) circle (0.08);
			\draw[color=blue!60, fill=blue!60](3, 3*0.4+0.6 + 0.2) circle (0.08);
			\draw[color=blue!60, fill=blue!60](3.2, 3.2*0.4+0.6 - 0.25) circle (0.08);
			\draw[color=blue!60, fill=blue!60](3.8, 3.8*0.4+0.6 - 0.35) circle (0.08);
			\draw[color=blue!60, fill=blue!60](3.9, 3.9*0.4+0.6 + 0.35) circle (0.08);
			\draw[color=blue!60, fill=blue!60](3.1,0.55) circle (0.08);
			
			\node[font=\fontsize{10}{0}\selectfont] at (6.3,0.3) {feature $\feature$};
			\node[font=\fontsize{10}{0}\selectfont] at (0,3.3) {label $\truelabel$};
			\node[font=\fontsize{10}{0}\selectfont] at (5.5,2.7) {$h(\feature)$};
			\node[font=\fontsize{10}{0}\selectfont] at (4.5,2) {$\varepsilon$};
			\node[font=\fontsize{10}{0}\selectfont] at (4,0.6) {``\gls{outlier}''};
			
		\end{tikzpicture}
	\end{center}
	\caption{The \gls{huberloss} measures prediction errors via squared error loss for 
		regular \gls{datapoint}s inside the band of width $\varepsilon$ around the 
		hypothesis map $h(\featurevec)$ and via the absolute difference loss for 
		an \gls{outlier} outside the band. }
\label{fig_huber_loss_scatter}
\end{figure}

The \gls{huberloss} is robust to outliers since the corresponding (large) prediction errors 
$\truelabel - \hat{\truelabel}$ are not squared. \Gls{outlier}s have a smaller effect on the average 
\gls{huberloss} (over the entire dataset) compared to the average squared error loss. The improved 
robustness against outliers of the \gls{huberloss} comes at the expense of increased computational 
complexity. The squared error loss can be minimized using efficient gradient based methods 
(see Chapter \ref{ch_GD}). In contrast, for $\varepsilon=0$, the \gls{huberloss} is non-differentiable and 
requires more advanced optimization methods. 

The \gls{huberloss} \eqref{equ_def_Huber_loss} contains two important special cases. The first 
special case occurs when $\varepsilon$ is chosen to be very large, such that the condition $|\truelabel-\hat{\truelabel}| \leq \varepsilon$ 
is satisfied for most \gls{datapoint}s. In this case, the \gls{huberloss} resembles the squared error loss 
\eqref{equ_squared_loss} (up to a scaling factor $1/2$). The second special case is obtained for $\varepsilon=0$. 
Here, the \gls{huberloss} reduces to the scaled absolute difference loss $|\truelabel - \hat{\truelabel}|$.

\section{The Lasso}
\label{sec_lasso}
We will see in Chapter \ref{ch_validation_selection} that \gls{linreg} (see Section \ref{sec_lin_reg}) 
typically requires a \gls{trainset} larger than the number of features used to characterized a \gls{datapoint}. 
However, many important application domains generate \gls{datapoint}s with a number $\featuredim$ 
of features much higher than the number $\samplesize$ of available labeled \gls{datapoint}s in the \gls{trainset}. 

In the \gls{highdimregime}, where $\samplesize \ll \featurelen$, basic \gls{linreg} methods 
will not be able to learn useful weights $\weights$ for a linear hypothesis. 
Section \ref{sec_gen_linreg} shows that for $\samplesize \ll \featurelen$, \gls{linreg} will 
typically learn a hypothesis that perfectly predicts labels of \gls{datapoint}s in the \gls{trainset} 
but delivers poor predictions for \gls{datapoint}s outside the \gls{trainset}. This phenomenon is referred 
to as overfitting and poses a main challenge for ML applications in the \gls{highdimregime}. 

Chapter \ref{ch_overfitting_regularization} discusses basic \gls{regularization} techniques that allow to 
prevent ML methods from overfitting. We can regularize \gls{linreg} by augmenting the squared 
error loss \eqref{equ_squared_loss} of a hypothesis $h^{(\weights)}(\featurevec) = \weights^{T} \featurevec$ 
with an additional penalty term. This penalty term depends solely on the weights $\weights$ and 
serves as an estimate for the increase of the average loss on \gls{datapoint}s outside the \gls{trainset}. 
Different ML methods are obtained from different choices for this penalty term. 
The \gls{lasso} is obtained from \gls{linreg} by replacing the squared error loss with the regularized loss 
\begin{equation}
	\label{equ_def_reg_loss_lin_reg}
	\loss{(\featurevec,\truelabel)}{h^{(\vw)}} = (\truelabel-\weights^{T}\featurevec)^{2} + \regparam \| \weights \|_{1}. 
\end{equation}
Here, the penalty term is given by the scaled norm $\regparam \| \vw \|_{1}$.
The value of $\regparam$ can be chosen based on some probabilistic model that interprets a \gls{datapoint} as 
the realization of a \gls{rv}. The label of this \gls{datapoint} (which is a realization of a \gls{rv}) is 
related to its features via $$\truelabel = \overline{\weights}^{T} \featurevec + \varepsilon.$$ 
Here, $\overline{\weights}$ denotes some true underlying parameter vector and $\varepsilon$ is a 
realization of an a \gls{rv} that is independent of the features $\featurevec$. We need the 
``noise'' term $\varepsilon$ since the labels of \gls{datapoint}s collected in some ML application are typically 
not exactly obtained by a linear combination $\overline{\weights}^{T} \featurevec$ of its features.   

The tuning of $\regparam$ in \eqref{equ_def_reg_loss_lin_reg} can be guided by the statistical properties 
(such as the variance) of the noise $\varepsilon$, the number of non-zero entries in $\overline{\weights}$ 
and a lower bound on the non-zero values \cite{Wain2019,BuhlGeerBook}. Another option for choosing the 
value $\regparam$ is to try out different candidate values and pick the one resulting in smallest validation 
error (see Section \ref{sec_validate_predictor}).

\section{Gaussian Basis Regression}
\label{sec_linbasreg}

Section \ref{sec_polynomial_regression} showed how to extend linear regression by first 
transforming the feature $\feature$ using a vector-valued \gls{featuremap} $\featuremap: \mathbb{R} \rightarrow \mathbb{R}^{\featuredim}$. 
The output of this \gls{featuremap} are the transformed features $\featuremap(\feature)$ which 
are fed, in turn, to a linear map $h\big(\featuremap(\feature)\big) = \weights^{T} \featuremap(\feature)$. 
Polynomial regression in Section \ref{sec_polynomial_regression} has been obtained for the specific 
\gls{featuremap} \eqref{equ_poly_feature_map} whose entries are the powers $\feature^{l}$ of the 
scalar original feature $\feature$. However, it is possible to use other functions, different from 
polynomials, to construct the feature map $\featuremap$. We can extend linear regression using an 
arbitrary feature map 
\begin{equation} 
	\featuremap(\feature) = (\basisfunc_{1}(\feature),\ldots,\basisfunc_{\featuredim}(\feature))^{T}  
\end{equation} 
with the scalar maps $\basisfunc_{\featureidx}: \mathbb{R} \rightarrow \mathbb{R}$ which 
are referred to as ``basis functions''. The choice of basis functions 
depends heavily on the particular application and the underlying relation 
between features and labels of the observed \gls{datapoint}s. The basis 
functions underlying polynomial regression are $\basisfunc_{\featureidx}(\feature)= \feature^{\featureidx}$. 

Another popular choice for the basis functions are ``Gaussians'' 
\begin{equation} 
	\label{equ_basis_Gaussian}
	\phi_{\sigma,\mu}(\feature) = \exp(-(1/(2\sigma^{2})) (\feature\!-\!\mu)^{2}). 
\end{equation}
The family \eqref{equ_basis_Gaussian} of maps is parameterized by the variance $\sigma^{2}$ and the mean (shift) $\mu$. 
Gaussian basis \gls{linreg} combines the feature map 
\begin{equation} 
	\featuremap(\feature) = \big (\phi_{\sigma_{1},\mu_{1}}(\feature) ,\ldots,\phi_{\sigma_{\featuredim},\mu_{\featuredim}}(\feature) \big)^{T} 
\end{equation}
with \gls{linreg} (see Figure \ref{fig_lin_bas_expansion}). The resulting \gls{hypospace} is 
\begin{align}
	\label{equ_def_Gauss_hypospace}
	\hypospace^{(\featuredim)}_{\rm Gauss} & = \{ h^{(\weights)}: \mathbb{R} \rightarrow \mathbb{R}: h^{(\weights)}(\feature)\!=\!\sum_{\featureidx=1}^{\featuredim}  \weight_{\featureidx}\phi_{\sigma_{\featureidx},\mu_{\featureidx}}(\feature) \nonumber \\
	& \mbox{ with weights } \weights=(\weight_{1},\ldots,\weight_{\featuredim})^{T} \in \mathbb{R}^{\featuredim}\}.
\end{align}

Different choices for the variance $\sigma_{\featureidx}^{2}$ and shifts $\mu_{\featureidx}$ of the Gaussian 
function in \eqref{equ_basis_Gaussian} results in different \gls{hypospace}s $\hypospace_{\rm Gauss}$. 
Chapter \ref{sec_modsel} will discuss model selection techniques that allow to find useful values 
for these parameters.

The hypotheses of \eqref{equ_def_Gauss_hypospace} are parametrized by a parameter vector $\weights \in \mathbb{R}^{\featuredim}$. 
Each hypothesis in $\hypospace_{\rm Gauss}$ corresponds to a particular choice for the parameter vector $\weights$. 
Thus, instead of searching over $\hypospace_{\rm Gauss}$ to find a good hypothesis, we can search over the \gls{euclidspace} 
$\mathbb{R}^{\featuredim}$. Highly developed methods for searching over the space $\mathbb{R}^{\featuredim}$, for a wide range 
of values for $\featuredim$, are provided by numerical linear algebra \cite{golub96}. 


\begin{figure}[htbp]
	\begin{center}
		\begin{tikzpicture}
				\draw[blue,  thick, domain=-3:3] plot (\x, {2/(1+exp(-\x))});
				\draw[red,  thick, domain=-3:3] plot (\x,  {exp(-(\x+1)^2)+exp(-(\x-1)^2)}) ;
				\node [red,right] at (2,0.4) {$\hat{\truelabel}=h^{(\weights)}(\feature)$ with $h^{(\weights)} \!\in\!\hypospace^{(2)}_{\rm Gauss}$} ; 
				\node [blue,right] at (2,1.6) {$\truelabel=h(\feature)$} ; 
				\draw[->] (-3.5,0) -- (3.5,0) node[right] {$\feature$};
				\draw[->] (0,-0.25) -- (0,1.5) node[above] {$\truelabel$};
				\foreach \y/\ytext in {0/0, 1/1}
				\draw[shift={(0,\y)}] (2pt,0pt) -- (-2pt,0pt) node[left] {$\ytext$};  
				\foreach \x/\xtext in {-3/-3,-2/-2,-1/-1,0/0,1/1, 2/2,3/3}
				\draw[shift={(\x,0)}] (0pt,2pt) -- (0pt,-2pt) node[below] {$\xtext$};  
		\end{tikzpicture}
		\vspace*{-5mm}
	\end{center}\caption{The true relation $\feature \mapsto \truelabel$ (blue) between feature $\feature$ 
	 and label $\truelabel$ of \gls{datapoint}s is highly non-linear. Therefore it seems reasonable to predict 
	 the label using a non-linear hypothesis map $h^{(\weights)}(\feature) \!\in\!\hypospace^{(2)}_{\rm Gauss}$ with some parameter vector $\weights \in \mathbb{R}^{2}$.} 
	\label{fig_lin_bas_expansion}
\end{figure}

\section{Logistic Regression} 
\label{sec_LogReg}

\Gls{logreg} is a ML method that allows to classify \gls{datapoint}s according to two categories. 
Thus, \gls{logreg} is a binary classification method that can be applied to \gls{datapoint}s 
characterized by feature vectors $\featurevec \in \mathbb{R}^{\featuredim}$ 
(feature space $\featurespace=\mathbb{R}^{\featuredim}$) and binary labels $\truelabel$. 
These binary labels take on values from a label space that contains two different label values. 
Each of these two label values represents one of the two categories to which the \gls{datapoint}s can belong. 

It is convenient to use the \gls{labelspace} $\labelspace = \mathbb{R}$ and encode the two label values as $\truelabel=1$ 
and $\truelabel=-1$. However, it is important to note that \gls{logreg} can be used with an arbitrary \gls{labelspace} 
which contains two different elements. Another popular choice for the \gls{labelspace} is $\labelspace=\{0,1\}$.  
\Gls{logreg} learns a hypothesis out of the \gls{hypospace} $\hypospace^{(\featuredim)}$ (see \eqref{equ_lin_hypospace}). 
Note that \gls{logreg} uses the same \gls{hypospace} as \gls{linreg} (see Section \ref{sec_lin_reg}). 

At first sight, it seems wasteful to use a linear hypothesis $h(\featurevec) = \weights^T \featurevec$, 
with some parameter vector $\weights \in \mathbb{R}^{\featuredim}$, to predict a 
binary label $\truelabel$. Indeed, while the prediction $h(\featurevec)$ can take any real number, 
the label $\truelabel \in \{-1,1\}$ takes on only one of the two real numbers $1$ and $-1$. 

It turns out that even for binary labels it is quite useful to use 
a hypothesis map $h$ which can take on arbitrary real numbers. 
We can always obtain a predicted label $\hat{\truelabel} \in \{-1,1\}$ by comparing 
hypothesis value $h(\featurevec)$ with a threshold. A \gls{datapoint} with \gls{features} $\featurevec$, 
is classified as $\hat{\truelabel}=1$ if $h(\featurevec)\geq 0$ and $\hat{\truelabel}=-1$ for $h(\featurevec)< 0$. 
Thus, we use the sign of the \gls{predictor} $h$ to determine the 
final prediction for the label. The absolute value $|h(\featurevec)|$ is then used 
to quantify the reliability of (or confidence in) the classification $\hat{\truelabel}$. 

Consider two \gls{datapoint}s with feature vectors $\featurevec^{(1)}, \featurevec^{(2)}$ and 
a linear classifier map $h$ yielding the function values $h(\featurevec^{(1)}) = 1/10$ 
and $h(\featurevec^{(2)}) = 100$. Whereas the predictions for both \gls{datapoint}s result 
in the same label predictions, i.e., $\hat{\truelabel}^{(1)}\!=\!\hat{\truelabel}^{(2)}\!=\!1$, the 
classification of the \gls{datapoint} with feature vector $\featurevec^{(2)}$ seems to be 
much more reliable. 

\Gls{logreg} uses the \gls{logloss} \eqref{equ_log_loss} to assess the quality of a particular hypothesis 
$h^{(\weights)} \in \hypospace^{(\featuredim)}$. In particular, given some labeled \gls{trainset} 
$\dataset =\{ \featurevec^{(\sampleidx)},\truelabel^{(\sampleidx)} \}_{\sampleidx=1}^{\samplesize}$, 
\gls{logreg} tries to minimize the \gls{emprisk} (average \gls{logloss}) 
\begin{align} 
	\label{equ_def_emp_risk_logreg}
	\emperror ( \weights | \dataset ) & = (1/\samplesize) \sum_{\sampleidx=1}^{\samplesize}  \log( 1+ \exp ( - \truelabel^{(\sampleidx)} h^{(\weights)}(\featurevec^{(\sampleidx)}))) \nonumber \\
	& \stackrel{h^{(\weights)}(\featurevec)=\weights^{T} \featurevec}{=}  (1/\samplesize) \sum_{\sampleidx=1}^{\samplesize} \log( 1+ \exp ( - \truelabel^{(\sampleidx)} \weights^{T} \featurevec^{(\sampleidx)})).
\end{align} 

Once we have found the optimal parameter vector $\widehat{\weights}$, which minimizes \eqref{equ_def_emp_risk_logreg}, 
we can classify any \gls{datapoint} solely based on its features $\featurevec$. Indeed, we just need to evaluate 
the hypothesis $h^{(\widehat{\weights})}$ for the features $\featurevec$ to obtain the predicted label 
\begin{equation}
	\label{equ_class_logreg}
	\hat{\truelabel} =  \begin{cases} 1 & \mbox{ if } h^{(\widehat{\weights})}(\featurevec) \geq 0 \\ -1 & \mbox{ otherwise.} \end{cases}
\end{equation} 
Since $h^{(\widehat{\weights})}(\featurevec) = \big(\widehat{\weights}\big)^{T} \featurevec$ (see \eqref{equ_lin_hypospace}), 
the classifier \eqref{equ_class_logreg} amounts to testing whether $\big(\widehat{\weights}\big)^{T} \featurevec \geq 0$ 
or not. 

The classifier \eqref{equ_class_logreg} partitions the feature space $\featurespace\!=\!\mathbb{R}^{\featuredim}$ 
into two half-spaces $\decreg{1}\!=\!\big\{ \featurevec: \big(\widehat{\weights}\big)^{T} \featurevec\!\geq\!0 \big\}$ and 
$\decreg{-1}\!=\!\big\{ \featurevec: \big(\widehat{\weights}\big)^{T} \featurevec\!<\!0 \big\}$ which are separated by the 
hyperplane $\big(\widehat{\weights}\big)^{T} \featurevec =0$ (see Figure \ref{fig_lin_dec_boundary}). Any \gls{datapoint} 
with features $\featurevec \in \decreg{1}$ ($\featurevec \in \decreg{-1}$) is classified as $\hat{\truelabel}\!=\!1$ ($\hat{\truelabel}\!=\!-1$). 

\Gls{logreg} can be interpreted as a statistical estimation method for a particular probabilistic model 
for the \gls{datapoint}s. This probabilistic model interprets the label $\truelabel \in \{-1,1\}$ of a \gls{datapoint} as a 
\gls{rv} with the probability distribution 
\begin{align}
	\label{equ_prob_model_logistic_model}
	\prob{\truelabel=1;\weights}& = 1/(1 + \exp(- \weights^{T} \featurevec) )  \nonumber \\
	& \stackrel{h^{(\weights)}(\featurevec)\!=\!\weights^{T} \featurevec}{=} 1/(1 + \exp(- h^{(\weights)}(\featurevec))) )  . 
\end{align} 
As the notation indicates, the probability \eqref{equ_prob_model_logistic_model} 
is parametrized by the parameter vector $\weights$ of the linear hypothesis $h^{(\weights)}(\featurevec)\!=\!\weights^{T} \featurevec$. 
Given the probabilistic model \eqref{equ_prob_model_logistic_model}, we 
can interpret the classification \eqref{equ_class_logreg} as choosing $\hat{\truelabel}$ 
to maximize the probability $\prob{\truelabel=\hat{\truelabel};\weights}$. 


Since $\prob{\truelabel=1} + \prob{\truelabel=-1}=1$, 
\begin{align}
	\label{equ_prob_model_logistic_model_minus_1}
	\prob{\truelabel=-1} & =  1 - \prob{\truelabel=1} \nonumber \\
	& \stackrel{\eqref{equ_prob_model_logistic_model}}{=} 1 - 1/(1+ \exp(-\weights^{T} \featurevec)) \nonumber \\
	& = 1/(1+ \exp(\weights^{T} \featurevec)).
\end{align}

In practice we do not know the parameter vector in \eqref{equ_prob_model_logistic_model}. 
Rather, we have to estimate the parameter vector $\weights$ in \eqref{equ_prob_model_logistic_model} 
from observed \gls{datapoint}s. A principled approach to estimate the parameter 
vector is to maximize the probability (or likelihood) of actually obtaining the 
dataset $\dataset=\{ (\featurevec^{(\sampleidx)},\truelabel^{(\sampleidx)}) \}_{\sampleidx=1}^{\samplesize}$ 
as realizations of \gls{iid} \gls{datapoint}s whose labels are distributed 
according to \eqref{equ_prob_model_logistic_model}. This yields the \gls{ml} estimator 
\begin{align}
\label{equ_deriv_prog_logreg}
\widehat{\weights} & = \argmax_{\weights \in \mathbb{R}^{\featuredim}} \prob{ \{\truelabel^{(\sampleidx)}\}_{\sampleidx=1}^{\samplesize}} \nonumber \\
& \stackrel{\truelabel^{(\sampleidx)}\mbox{\gls{iid}}}{=} \argmax_{\vw \in \mathbb{R}^{\featuredim}} \prod_{\sampleidx=1}^{\samplesize} \prob { \truelabel^{(\sampleidx)}}\nonumber \\
& \stackrel{\eqref{equ_prob_model_logistic_model},\eqref{equ_prob_model_logistic_model_minus_1}}{=} \argmax_{\vw \in \mathbb{R}^{\featuredim}} \prod_{\sampleidx=1}^{\samplesize} 1/(1 + \exp(- \truelabel^{(\sampleidx)} \weights^{T} \featurevec^{(\sampleidx)}) ). 
\end{align}
Note that the last expression \eqref{equ_deriv_prog_logreg} is only valid 
if we encode the binary labels using the values $1$ and $-1$. Using different 
label values results in a different expression. 

Maximizing a positive function $f(\weights)>0$ is equivalent to maximizing $\log f(\weight)$, $$\argmax\limits_{\weights \in \mathbb{R}^{\featuredim}} f(\weights)\!=\!\argmax\limits_{\weights \in \mathbb{R}^{\featuredim}}\log f(\weights).$$ 
Therefore, \eqref{equ_deriv_prog_logreg} can be further developed as 
\begin{align}
\label{equ_deriv_prog_logreg_2}
\widehat{\weights} & \stackrel{\eqref{equ_deriv_prog_logreg}}{=} \argmax_{\weights \in \mathbb{R}^{\featuredim}} \sum_{\sampleidx=1}^{\samplesize} - \log\big(1\!+\!\exp(- \truelabel^{(\sampleidx)} \weights^{T} \featurevec^{(\sampleidx)}) \big)  \nonumber \\
& = \argmin_{\weights \in \mathbb{R}^{\featuredim}} (1/\samplesize)\sum_{\sampleidx=1}^{\samplesize} \log\big(1\!+\!\exp(- \truelabel^{(\sampleidx)}\vw^{T} \featurevec^{(\sampleidx)}) \big). 
\end{align}
Comparing \eqref{equ_deriv_prog_logreg_2} with \eqref{equ_def_emp_risk_logreg} 
reveals that \gls{logreg} is nothing but \gls{ml} estimation of the parameter vector $\weights$ in 
the probabilistic model \eqref{equ_prob_model_logistic_model}.

\section{Support Vector Machines} 
\label{sec_SVM} 

\Gls{svm}s are a family of ML methods for learning a hypothesis to predict a binary label $\truelabel$ of a \gls{datapoint} 
based on its features $\featurevec$. Without loss of generality we consider binary labels taking values in the 
\gls{labelspace} $\labelspace = \{-1,1\}$. A \gls{svm} uses the linear \gls{hypospace} \eqref{equ_lin_hypospace} which consists 
of linear maps $h(\featurevec) = \weights^{T} \featurevec$ with some parameter vector $\weights \in \mathbb{R}^{\featuredim}$. 
Thus, the \gls{svm} uses the same \gls{hypospace} as \gls{linreg} and \gls{logreg} which we have 
discussed in Section \ref{sec_lin_reg} and Section \ref{sec_LogReg}, respectively. What sets the 
\gls{svm} apart from these other methods is the choice of \gls{lossfunc}. 

Different instances of a \gls{svm} are obtained by using different constructions for the features of a \gls{datapoint}. 
\index{kernel SVM}Kernel \gls{svm}s use the concept of a kernel map to construct (typically high-dimensional) 
features (see Section \ref{sec_kernel_methods} and \cite{LampertNowKernel}). In what follows, 
we assume the feature construction has been solved and we have access to a feature vector  $\featurevec \in \mathbb{R}^{\featuredim}$ 
for each \gls{datapoint}. 

Figure \ref{fig_svm} depicts a dataset $\dataset$ of labeled \gls{datapoint}s, each characterized by a feature vector 
$\featurevec^{(\sampleidx)} \in \mathbb{R}^{2}$ (used as coordinates of a marker) and 
a binary label $\truelabel^{(\sampleidx)} \in \{-1,1\}$ (indicated by different marker shapes). 
We can partition dataset $\dataset$ into two classes 
\begin{equation} 
\label{equ_two_classes_svm}
\mathcal{C}^{(\truelabel=1)}\!=\!\{ \featurevec^{(\sampleidx)} : \truelabel^{(\sampleidx)}\!=\!1\}\mbox{,  and }\mathcal{C}^{(\truelabel=-1)}\!=\!\{ \featurevec^{(\sampleidx)} : \truelabel^{(\sampleidx)}\!=\!-1 \}.
\end{equation} 
The \gls{svm} tries to learn a linear map $h^{(\weights)}(\featurevec) = \weights^{T} \featurevec$ 
that perfectly separates the two classes in the sense of 
\begin{align} 
\label{equ_two_classes_svm_perf_sep}
\underbrace{h \big(  \featurevec^{(\sampleidx)} \big)}_{\weights^{T} \featurevec^{(\sampleidx)} }  > 0 \mbox{ for } \featurevec^{(\sampleidx)} \in  \mathcal{C}^{(\truelabel=1)} \mbox{ and } 
\underbrace{h \big(  \featurevec^{(\sampleidx)} \big)}_{\weights^{T} \featurevec^{(\sampleidx)} }  < 0 \mbox{ for } \featurevec^{(\sampleidx)} \in  \mathcal{C}^{(\truelabel=-1)}.
\end{align} 
We refer to a dataset, whose \gls{datapoint}s have binary labels. as \index{linear separable}linear separable 
if we can find at least one linear map that separates in the sense of \eqref{equ_two_classes_svm_perf_sep}. 
The dataset in Figure \ref{fig_svm} is linearly separable. 

As can be verified easily, any linear map $h^{(\weights)}(\featurevec) = \weights^{T} \featurevec$ achieving zero  
average \gls{hingeloss} \eqref{equ_hinge_loss} on the dataset $\dataset$ perfectly satisfies this dataset \eqref{equ_two_classes_svm_perf_sep}. 
It seems reasonable to learn a linear map by minimizing the average \gls{hingeloss} \eqref{equ_hinge_loss}. 
However, one drawback of this approach is that there might be (infinitely) many different linear maps 
that achieve zero average \gls{hingeloss} and, in turn, perfectly separate the \gls{datapoint}s in Figure \ref{fig_svm}. 
Indeed, consider a linear map $h^{(\weights)}$ that achieves zero average \gls{hingeloss} for the $\dataset$ 
in Figure \ref{fig_svm} (and therefore perfectly separates it). Then, any other linear map $h^{(\weights')}$ 
with weights $\weights' = \regparam \weights$, using an arbitrary number $\regparam > 1$ also achieves zero 
average \gls{hingeloss} (and perfectly separates the dataset). 

Neither the separability requirement \eqref{equ_two_classes_svm_perf_sep} nor the \gls{hingeloss} \eqref{equ_hinge_loss} 
are sufficient as a sole training criterion. Indeed, there are many (if not most) datasets that are not linearly separable. 
Even for a linearly separable dataset (such as the one Figure \ref{fig_svm}), there are infinitely many linear maps 
with zero average \gls{hingeloss}. Which one of these infinitely many different maps should we use? To settle these issues, 
the \gls{svm} uses a ``regularized'' \gls{hingeloss},   
\begin{align}
\loss{(\featurevec,\truelabel)}{h^{(\weights)}} & \defeq  \max \{ 0 , 1 - \truelabel\cdot h^{(\weights)}(\featurevec) \}  
+ \regparam \sqeuclnorm{ \weights }\nonumber \\
& \hspace*{-3mm}\stackrel{h^{(\weights)}(\featurevec) = \weights^{T} \featurevec}{=}  
\max \{ 0 , 1 - \truelabel\cdot \weights^{T} \featurevec \} + \regparam \sqeuclnorm{ \weights }.  \label{equ_loss_svm}
\end{align}
The loss \eqref{equ_loss_svm} augments the \gls{hingeloss} \eqref{equ_hinge_loss} by the term $\regparam \sqeuclnorm{\weights }$. 
This term is the scaled (by $\regparam >0$) squared Euclidean norm of the weights $\weights$ of the 
linear hypothesis $h$ used to classify \gls{datapoint}s. it can be shown that adding the term $\regparam \sqeuclnorm{ \weights }$ 
to the \gls{hingeloss} \eqref{equ_hinge_loss} has an \index{regularization}\gls{regularization} effect. 

The loss \label{equ_loss_svm} favours linear maps $h^{(\weights)}$ that are robust against (small) perturbations 
of the \gls{datapoint}s. The tuning parameter $\regparam$ in \eqref{equ_loss_svm} controls the strength of this 
regularization effect and might therefore also be referred to as a \index{regularization paramter}\gls{regularization} 
parameter. We will discuss \gls{regularization} on a more general level in Chapter \ref{ch_overfitting_regularization}. 

Let us now develop a useful geometric interpretation of the linear hypothesis obtained by minimizing the \gls{lossfunc} \eqref{equ_loss_svm}. 
According to \cite[Chapter 2]{LampertNowKernel}, a classifier $h^{(\weights_{\rm SVM})}$ that minimizes the average loss \eqref{equ_loss_svm}, 
maximizes the distance (margin) $\xi$ between its \gls{decisionboundary} and each of the two classes $\cluster^{(\truelabel=1)}$ and $\cluster^{(\truelabel=-1)}$ (see \eqref{equ_two_classes_svm}). The \index{decision boundary}\gls{decisionboundary} 
is given by the set of feature vectors $\featurevec$ satisfying $\weights_{\rm SVM}^{T} \featurevec=0$, 

Making the margin as large as possible is reasonable as it ensures that the resulting classifications are robust 
against small perturbations of the features (see Section \ref{sec_robustness}). As depicted in Figure \ref{fig_svm}, 
the margin between the \gls{decisionboundary} and the classes $\mathcal{C}_{1}$ and $\mathcal{C}_{2}$ is typically 
determined by few \gls{datapoint}s (such as $\featurevec^{(6)}$ in Figure \ref{fig_svm}) which are closest 
to the \gls{decisionboundary}. These \gls{datapoint}s have minimum distance to the \gls{decisionboundary} and are 
referred to as \index{support vectors} support vectors. 

\begin{figure}[htbp]
\begin{center}
	\begin{tikzpicture}[auto,scale=0.8]
		\draw [thick] (1,2) circle (0.1cm)node[anchor=west] {\hspace*{0mm}$\featurevec^{(5)}$};
		\draw [thick] (0,1.6) circle (0.1cm)node[anchor=west] {\hspace*{0mm}$\featurevec^{(4)}$};
		\draw [thick] (0,3) circle (0.1cm)node[anchor=west] {\hspace*{0mm}$\featurevec^{(3)}$};
		\draw [thick] (2,1) circle (0.1cm)node[anchor=east,above] {\hspace*{0mm}$\featurevec^{(6)}$};
		\node[] (B) at (-2,0) {``support vector''};
		\draw[->,dashed] (B) to (1.9,1) ; 
		\draw [|<->|,thick] (2.05,0.95)  -- (2.75,0.25)node[pos=0.5] {$\xi$} ; 
		\draw [thick] (1,-1.5) -- (4,1.5) node [right] {$h^{(\weights)}$} ; 
		\draw [thick] (3,-1.9) rectangle ++(0.1cm,0.1cm) node[anchor=west,above]  {\hspace*{0mm}$\featurevec^{(2)}$};
		\draw [thick] (4,.-1) rectangle ++(0.1cm,0.1cm) node[anchor=west,above] {\hspace*{0mm}$\featurevec^{(1)}$};
	\end{tikzpicture}
	\caption{The \gls{svm} learns a hypothesis (or classifier) $h^{(\weights)}$ with minimum average soft-margin \gls{hingeloss} \eqref{equ_loss_svm}. 
		Minimizing this loss is equivalent to maximizing the margin $\xi$ between the \gls{decisionboundary} of $h^{(\weights)}$ 
		and each class of the \gls{trainset}.}
	\label{fig_svm}
\end{center}
\end{figure}

We highlight that both, the \gls{svm} and \gls{logreg} use the same \gls{hypospace} of linear maps. 
Both methods learn  a \gls{linclass} $h^{(\weights)} \in \hypospace^{(\featuredim)}$ (see \eqref{equ_lin_hypospace}) 
whose \gls{decisionboundary} is a hyperplane in the feature space $\featurespace = \mathbb{R}^{\featuredim}$ 
(see Figure \ref{fig_lin_dec_boundary}). The difference between \gls{svm} and \gls{logreg} is in their choice 
for the \gls{lossfunc} used to evaluate the quality of a hypothesis $h^{(\weights)} \in \hypospace^{(\featuredim)}$. 

The \gls{hingeloss} \eqref{equ_hinge_loss} is a (in some sense optimal) convex approximation to the $0/1$ loss \eqref{equ_def_0_1}. 
Thus, we expect the classifier obtained by the \gls{svm} to yield a smaller classification error probability $\prob{ \predictedlabel \neq \truelabel }$ (with $\predictedlabel =1$ if $h(\featurevec)\geq0$ and $\predictedlabel=-1$ otherwise) compared to \gls{logreg} 
which uses the \gls{logloss} \eqref{equ_log_loss}. The \gls{svm} is also statistically appealing as it 
learns a robust hypothesis. Indeed, the hypothesis map with maximum margin is maximally robust 
against perturbations of the feature vectors of \gls{datapoint}s. Section \ref{sec_robustness} 
discusses the importance of robustness in ML methods in more detail. 

The statistical superiority (in terms of robustness) of the \gls{svm} comes at the cost of increased 
computational complexity. The \gls{hingeloss} \eqref{equ_hinge_loss} is non-differentiable 
which prevents the use of simple \gls{gdmethods} (see Chapter \ref{ch_GD}) and requires more 
advanced optimization methods. In contrast, the \gls{logloss} \eqref{equ_log_loss} is \gls{convex} 
and \gls{differentiable}. \Gls{logreg} allows for the use of \gls{gdmethods} to minimize the average \gls{logloss} 
incurred on a \gls{trainset} (see Chapter \ref{ch_GD}).

\section{Bayes Classifier}
\label{sec_NaiveBayes}

Consider \gls{datapoint}s characterized by \gls{features} $\featurevec \in \featurespace$ 
and some binary \gls{label} $\truelabel \in \labelspace$. We can use any two different label values 
but let us assume that the two possible label values are $\truelabel=-1$ or $\truelabel=1$. 
We would like to find (or learn) a classifier $h: \featurespace \rightarrow \labelspace$ 
such that the predicted (or estimated) label $\hat{\truelabel} = h(\featurevec)$ agrees with the true label $\truelabel \in \labelspace$ 
as much as possible. Thus, it is reasonable to assess the quality of a classifier $h$ using the $0/1$ 
loss \eqref{equ_def_0_1}. We could then learn a \gls{classifier} using the \gls{erm} with the \gls{lossfunc}
\eqref{equ_def_0_1}. However, the resulting optimization problem is typically intractable since the 
\gls{loss} \eqref{equ_def_0_1} is non-convex and non-differentiable. 

Instead of solving the (intractable) \gls{erm} for $0/1$ \gls{loss} \eqref{equ_def_0_1}, we can take a different route 
to construct a classifier. This construction is based on a simple probabilistic model for \gls{data}. Using this 
model, we can interpret the average $0/1$ \gls{loss} incurred by a hypothesis on a \gls{trainset} as an 
approximation to the probability $\errprob = \prob{ \truelabel \neq h(\featurevec) }$. Any classifier $\hat{h}$ that 
minimizes the error probability $\errprob$, which is the expected $0/1$ \gls{loss}, is referred to as a \gls{bayesestimator}. 
Section \ref{sec_ERM_Bayes} will discuss ML methods using \gls{bayesestimator} in more detail.    

Let us derive the \gls{bayesestimator} for a the special case of a binary classification problem. 
Here, \gls{datapoint}s are characterized by features $\featurevec$ and label $\truelabel \in \{-1,1\}$. 
Elementary probability theory allows to derive the \gls{bayesestimator}, which is the hypothesis 
minimizing the expected $0/1$ \gls{loss}, as 
\begin{equation} 
	\label{equ_def_Bayes_est_binary_class}
\hat{h}(\featurevec) = \begin{cases} 1 & \mbox{ if } \prob{\truelabel = 1| \featurevec } > \prob{\truelabel = -1| \featurevec } \\ - 1 & \mbox{ otherwise.} \end{cases}. 
\end{equation} 

Note that the \gls{bayesestimator} \eqref{equ_def_Bayes_est_binary_class} depends on the \gls{probdist} $\prob{\featurevec,\truelabel}$ 
underlying the \gls{datapoint}s.\footnote{Remember that we interpret \gls{datapoint}s as realizations of \gls{iid} \gls{rv}s with common 
	\gls{probdist} $\prob{\featurevec,\truelabel}$.} We obtain different \gls{bayesestimator}s for different probabilistic models. 
One widely used probabilistic model results in a \gls{bayesestimator} that belongs to the linear \gls{hypospace} \eqref{equ_lin_hypospace}. 
Note that this \gls{hypospace} underlies also \gls{logreg} (see Section \ref{sec_LogReg}) and the \gls{svm} (see Section \ref{sec_SVM}). 
Thus, \gls{logreg}, \gls{svm} and \gls{bayesestimator} are all examples of a \gls{linclass} (see Figure \ref{fig_lin_dec_boundary}). 

A \gls{linclass} partitions the feature space $\featurespace$ into two half-spaces. One half-space consists 
of all feature vectors $\featurevec$ which result in the predicted label $\predictedlabel=1$ and the other half-space 
constituted by all feature vectors $\featurevec$ which result in the predicted label $\predictedlabel=-1$. The family of 
ML methods that learn a \gls{linclass} differ in their choices for the \gls{lossfunc}s used to assess the quality of  
these half-spaces.

\section{Kernel Methods} 
\label{sec_kernel_methods}

Consider a ML (classification or regression) problem with an underlying feature space $\featurespace$. In order to predict 
the label $y \in \labelspace$ of a \gls{datapoint} based on its \gls{features} $\featurevec \in \featurespace$, we apply a \gls{predictor} $h$ 
selected out of some \gls{hypospace} $\hypospace$. Let us assume that the available computational infrastructure only 
allows us to use a linear \gls{hypospace} $\hypospace^{(\featuredim)}$ (see \eqref{equ_lin_hypospace}).

For some applications, using a linear hypothesis $h(\featurevec)=\weights^{T}\featurevec$ is not suitable since the relation between 
\gls{features} $\featurevec$ and label $\truelabel$ might be highly non-linear. One approach to extend the capabilities of linear hypotheses 
is to transform the raw \gls{features} of a \gls{datapoint} before applying a linear hypothesis $h$. 

The family of kernel methods is based on transforming the \gls{features} $\featurevec$ to new features $\hat{\featurevec} \in \featurespace'$ 
which belong to a (typically very) high-dimensional space $\featurespace'$ \cite{LampertNowKernel}. It is not uncommon that, 
while the original feature space is a low-dimensional \gls{euclidspace} (e.g., $\featurespace = \mathbb{R}^{2}$), 
the transformed feature space $\featurespace'$ is an infinite-dimensional function space. 

The rationale behind transforming the original features into a new (higher-dimensional) feature space $\featurespace'$ is to 
reshape the intrinsic geometry of the feature vectors $\featurevec^{(\sampleidx)} \in \featurespace$ such that the transformed 
feature vectors $\hat{\featurevec}^{(\sampleidx)}$ have a ``simpler'' geometry (see Figure \ref{fig_kernelmethods}). 

Kernel methods are obtained by formulating ML problems (such as \gls{linreg} or \gls{logreg}) using the 
transformed features $\hat{\featurevec}= \phi(\featurevec)$. A key challenge within kernel methods is the choice of the feature 
map $\phi: \featurespace \rightarrow \featurespace'$ which maps the original feature vector $\featurevec$ to a new 
feature vector $\hat{\featurevec}= \phi(\featurevec)$.

\begin{figure}[htbp]
\begin{center}
	\begin{minipage}{0.45\textwidth}
		\begin{tikzpicture}[auto,scale=0.6]
			\draw [thick] (-3,-3) rectangle (4,4) node [anchor=east,above] {$\featurespace$} ;
			\draw [thick] (2,2) circle (0.1cm)node[anchor=west] {\hspace*{0mm}$\featurevec^{(5)}$};
			\draw [thick] (-2,1.6) circle (0.1cm)node[anchor=west] {\hspace*{0mm}$\featurevec^{(4)}$};
			\draw [thick] (-2,-1.7) circle (0.1cm)node[anchor=west] {\hspace*{0mm}$\featurevec^{(3)}$};
			\draw [thick] (2,-1.9) circle (0.1cm) node[anchor=west] {\hspace*{0mm}$\featurevec^{(2)}$};
			\draw [thick] (.5,.5) rectangle ++(0.1cm,0.1cm) node[anchor=west,above] {\hspace*{0mm}$\featurevec^{(1)}$};
		\end{tikzpicture}
	\end{minipage}
	\begin{minipage}{0.45\textwidth}
		\begin{tikzpicture}[auto,scale=0.6]
			\draw [thick] (0,-4) rectangle (7,3) node [anchor=east,above] {$\featurespace'$} ;
			\draw [thick] (3,0) circle (0.1cm)node[anchor=north] {\hspace*{0mm}$\hat{\featurevec}^{(5)}$};
			\draw [thick] (4,0) circle (0.1cm)node[anchor=north] {\hspace*{0mm}$\hat{\featurevec}^{(4)}$};
			\draw [thick] (5,0) circle (0.1cm)node[anchor=north] {\hspace*{0mm}$\hat{\featurevec}^{(3)}$};
			\draw [thick] (6,0) circle (0.1cm) node[anchor=north] {\hspace*{0mm}$\hat{\featurevec}^{(2)}$};
			\draw [thick] (1,0) rectangle ++(0.1cm,0.1cm) node[anchor=west,above] {\hspace*{0mm}$\hat{\featurevec}^{(1)}$};
		\end{tikzpicture}
	\end{minipage}
	\caption{The data set $\dataset = \{ (\featurevec^{(\sampleidx)},\truelabel^{(\sampleidx)}) \}_{\sampleidx=1}^{5}$ consists of  
		$5$ \gls{datapoint}s with features $\featurevec^{(\sampleidx)}$ and binary labels $\truelabel^{(\sampleidx)}$. Left: In the original feature 
		space $\featurespace$, the \gls{datapoint}s cannot be separated perfectly by any \gls{linclass} Right: The feature 
		map $\phi: \featurespace \rightarrow \featurespace'$ transforms the features $\featurevec^{(\sampleidx)}$ to the new features 
		$\hat{\featurevec}^{(\sampleidx)}=\phi\big(\featurevec^{(\sampleidx)}\big)$ in the new feature space $\featurespace'$. In the new 
		feature space $\featurespace'$ the \gls{datapoint}s can be separated perfectly by a \gls{linclass}. }
	\label{fig_kernelmethods}
\end{center}
\end{figure}

\newpage
\section{Decision Trees} 
\label{sec_decision_trees}

A \gls{decisiontree} is a flowchart-like description of a map $h: \featurespace \rightarrow \labelspace$ which maps 
the features $\featurevec \in \featurespace$ of a \gls{datapoint} to a predicted label $h(\featurevec) \in \labelspace$ \cite{hastie01statisticallearning}. 
While we can use \gls{decisiontree}s for an arbitrary feature space $\featurespace$ and \gls{labelspace} $\labelspace$, 
we will discuss them for the particular feature space $\featurespace = \mathbb{R}^{2}$ and \gls{labelspace} $\labelspace=\mathbb{R}$. 

Figure \ref{fig_decision_tree} depicts an example for a \gls{decisiontree}. A \gls{decisiontree}, consists of nodes 
which are connected by directed edges. We can think of a \gls{decisiontree},as a step-by-step instruction, 
or a ``recipe'', for how to compute the function value $h(\featurevec)$ given the features $\featurevec \in \featurespace$ 
of a \gls{datapoint}. This computation starts at the ``root'' node and ends at one of the ``leaf'' nodes 
of the \gls{decisiontree}. 

A leaf node $\hat{\truelabel}$, which does not have any outgoing edges, represents a \gls{decisionregion} $\decreg{\hat{\truelabel}} \subseteq \featurespace$ in the feature space. The hypothesis $h$ associated with a \gls{decisiontree},
is constant over the regions $\decreg{\hat{\truelabel}}$, such that $h(\featurevec) = \hat{\truelabel}$ 
for all $\featurevec \in \decreg{\hat{\truelabel}}$ and some label value $\hat{\truelabel}\in \mathbb{R}$. 

The nodes in a \gls{decisiontree} are of two different types, 
\begin{itemize} 
\item decision (or test) nodes, which represent particular ``tests'' about the feature vector $\featurevec$, 
e.g., ``is the norm of $\featurevec$ larger than $10$?'').
\item leaf nodes, which correspond to subsets of the feature space. 
\end{itemize} 
The particular \gls{decisiontree},depicted in Figure \ref{fig_decision_tree} consists 
of two decision nodes (including the root node) and three leaf nodes. 

Given limited computational resources, we can only use \gls{decisiontree}s with a limited depth. 
The depth of a \gls{decisiontree}, is the maximum number of hops it takes to reach a leaf 
node starting from the root and following the arrows. The \gls{decisiontree},depicted 
in Figure \ref{fig_decision_tree} has depth $2$. We obtain an entire \gls{hypospace} by 
collecting all hypothesis maps that are obtained from the \gls{decisiontree} in Figure \ref{fig_decision_tree} 
with some vectors $\vu$ and $\vv$, some positive radius $\weight >0$. The resulting \gls{hypospace} 
is parametrized by the vectors $\vu, \vv$ and the number $\weight$. 

To assess the quality of a particular \gls{decisiontree},we can use various  
loss functions. Examples of loss functions used to measure the quality 
of a \gls{decisiontree}, are the squared error loss for numeric labels (regression) 
or the impurity of individual \gls{decisionregion} for categorical labels (classification). 

\Gls{decisiontree} methods use as a \gls{hypospace} the set of all hypotheses 
which represented by a family of \gls{decisiontree}s.  Figure \ref{fig_hypospace_DT_depth_2} 
depicts a collection of \gls{decisiontree}s which are characterized by having depth at most two. 
More generally, we can construct a collection of \gls{decisiontree}s using a fixed set of 
``elementary tests'' on the input feature vector such as $\| \featurevec \| > 3$, $\feature_{3} < 1$. 
These tests might also involve a continuous (real-valued) parameter such as $\{ \feature_{2} > \weight \}_ {\weight \in [0,10]}$. 
We then build a \gls{hypospace} by considering all \gls{decisiontree}s not exceeding a maximum 
depth and whose decision nodes carry out one of the elementary tests. 

\begin{figure}[htbp]
\begin{minipage}{.45\textwidth} %
	\scalebox{0.6}{
		\begin{tikzpicture}
			[->,>=stealth',level/.style={sibling distance = 5cm/#1,
				level distance = 1.5cm}] 
			\node [env] {$\| \featurevec-\vu \| \leq \weight$?}
			child { node [env] {$h(\featurevec) = h_{1} $}
				edge from parent node [left,align=center] {no} }
			child { node [env] {$\| \featurevec\!-\!\vv \|\!\leq\!\weight$?}
				child { node [env] {$h(\featurevec)\!=\!h_{2}$}
					edge from parent node [left, align=center] {no} }
				child { node [env] {$h(\featurevec)\!=\!h_{3}$}
					edge from parent node [right, align=center]
					{yes} }
				edge from parent node [right,align=center] {yes} };
		\end{tikzpicture}
	}
\end{minipage}
\begin{minipage}{.45\textwidth} %
	\hspace*{15mm}
	\begin{tikzpicture}
		\draw (-2,2) rectangle (2,-2);
		\begin{scope}
			\clip (-0.5,0) circle (1cm);
			\clip (0.5,0) circle (1cm);
			\fill[color=gray] (-2,1.5) rectangle (2,-1.5);
		\end{scope}
		\draw (-0.5,0) circle (1cm);
		\draw (0.5,0) circle (1cm);
		\draw[fill] (-0.5,0) circle [radius=0.025];
		\node [below right, red] at (-0.5,0) {$\decreg{\hat{\truelabel}_{3}}$};
		\node [below left, blue] at (-0.7,0) {$\decreg{\hat{\truelabel}_{2}}$};
		\node [above left] at (-0.7,1) {$\decreg{\hat{\truelabel}_{1}}$};
		\node [left] at (-0.4,0) {$\mathbf{u}$};
		\draw[fill] (0.5,0) circle [radius=0.025];
		\node [right] at (0.6,0) {$\mathbf{v}$};
	\end{tikzpicture}
\end{minipage}
\caption{A \gls{decisiontree} represents a hypothesis $h$ which is constant on the \gls{decisionregion} $\decreg{\hat{\truelabel}}$, i.e., 
	$h(\featurevec)\!=\!\hat{\truelabel}$ for all $\featurevec\!\in\!\decreg{\hat{\truelabel}}$. Each \gls{decisionregion}  
	$\decreg{\hat{\truelabel}} \!\subseteq\!\featurespace$ corresponds to a specific leaf node in the \gls{decisiontree}.}
\label{fig_decision_tree}
\end{figure} 

\begin{figure}[htbp]
\begin{center}
	\begin{minipage}{.4\textwidth}
		\scalebox{0.8}{
			\begin{tikzpicture} [->,>=stealth',level/.style={sibling distance = 3cm/#1,
					level distance = 1.5cm}] 
				\node [env] {$\| \featurevec-\vu \| \leq r$?}
				child {node [env] {$h(\featurevec)=1$}
					edge from parent node [left,align=center] {no} }
				child {node [env] {$h(\featurevec)=2$}
					edge from parent node [right,align=center] {yes} };
		\end{tikzpicture}}
	\end{minipage}
	\hspace*{3mm}
	\begin{minipage}{.4\textwidth}
		\scalebox{0.8}{
			\begin{tikzpicture} [->,>=stealth',level/.style={sibling distance = 3.5cm,
					level distance = 1.5cm}] 
				\node [env] {$\| \featurevec-\vu \| \leq \weight$?}
				child {node [env] {$h(\featurevec)=1$}
					edge from parent node [left,align=center] {no} }
				child {node  [env] {$\| \featurevec-\vv \| \leq \weight$?}    child {node  [env] {$h(\featurevec) =10$} edge from parent node [left,align=center] {no} }
					child {node  [env] {$h(\featurevec) =20$} edge from parent node [right,align=center] {yes} }
					edge from parent node [right,align=center] {yes} 
				};
		\end{tikzpicture}}
	\end{minipage}
\end{center}
\caption{A \gls{hypospace} $\hypospace$ which consists of two \gls{decisiontree}s 
	with depth $2$ and using the tests $\| \featurevec\!-\!\vu \|\!\leq\! \weight$ and $\|\featurevec\!-\!\vv \|\!\leq\! \weight$ 
	with a fixed radius $\weight$ and vectors $\vu,\vv \in \mathbb{R}^{\featurelen}$. }
\label{fig_hypospace_DT_depth_2}
\end{figure}

A \gls{decisiontree} represents a map $h: \featurespace \rightarrow \labelspace$, which 
is piecewise-constant over regions of the feature space $\featurespace$. These 
non-overlapping \gls{decisionregion}s  partition the feature space into subsets of features 
that are all mapped to the same predicted label. Each leaf node of a \gls{decisiontree} corresponds 
to one particular \gls{decisionregion}. Using large \gls{decisiontree}s that contain many different 
test nodes, we can learn a hypothesis with highly complicated \gls{decisionregion}s. 
These \gls{decisionregion}s can be chosen such they perfectly align with almost any given 
labeled dataset (see Figure \ref{fig_decisiontree_overfits}).

Using a sufficiently large (deep) \gls{decisiontree}, we can obtain a hypothesis 
map that closely approximates any given non-linear map (under mild technical conditions such 
as Lipschitz continuity). This is quite different from ML methods using the linear \gls{hypospace} \eqref{equ_lin_hypospace}, 
such as \gls{linreg}, \gls{logreg} or the \gls{svm}. These methods learn linear hypothesis maps with 
a rather simple geometry. Indeed, a linear map is constant along hyperplanes. Moreover, the 
decision regions obtained from linear classifiers are always entire half-spaces (see Figure \ref{fig_lin_dec_boundary}). 


\begin{figure}[htbp]
\begin{minipage}{0.4\textwidth}
	\begin{tikzpicture}[auto,scale=0.5]
		\draw [thick] (5,5) circle (0.1cm)node[anchor=west] {\hspace*{0mm}$\featurevec^{(3)}$};
		\draw [dashed] (3,3) rectangle (6,6) ;
		\draw [dashed] (3,0) rectangle (6,3) ;
		\draw [dashed] (0,3) rectangle (3,6) ;
		\draw [dashed] (0,0) rectangle (3,3) ;
		\draw [thick] (1,1) circle (0.1cm)node[anchor=west] {\hspace*{0mm}$\featurevec^{(4)}$};
		\draw [thick] (5,1) rectangle ++(0.2cm,0.2cm) node[anchor=west,above] {\hspace*{0mm}$\featurevec^{(2)}$};
		\draw [thick] (1,5) rectangle ++(0.2cm,0.2cm) node[anchor=west,above] {\hspace*{0mm}$\featurevec^{(1)}$};
		\draw[->] (-0.5,0) -- (6.5,0) node[right] {$\feature_{1}$};
		\draw[->] (0,-0.5) -- (0,6.5) node[above] {$\feature_{2}$};
		\foreach \y/\ytext in {0/0, 1/1,2/2,3/3,4/4,5/5,6/6} \draw[shift={(0,\y)}] (2pt,0pt) -- (-2pt,0pt) node[left] {$\ytext$};  
		\foreach \x/\xtext in{0/0, 1/1,2/2,3/3,4/4,5/5,6/6}\draw[shift={(\x,0)}] (0pt,2pt) -- (0pt,-2pt) node[below] {$\xtext$};  
	\end{tikzpicture}
\end{minipage}
\begin{minipage}{0.44\textwidth}
	\scalebox{0.6}{
		\begin{tikzpicture} [->,>=stealth',level/.style={sibling distance = 3cm/#1,
				level distance = 1.5cm},scale=0.9] 
			\tikzstyle{level 1}=[sibling distance=60mm]
			\tikzstyle{level 2}=[sibling distance=30mm]
			\node [env] {$\feature_{1}\!\leq\!3$?}
			child  {node [env] {$\feature_{2}\!\leq\!3?$} 
				child {node [env] {$h(\featurevec)\!=\!\truelabel^{(3)}$}  edge from parent node [left,align=center] {no}}
				child {node [env] {$h(\featurevec)\!=\!\truelabel^{(2)}$}   edge from parent node [left,align=center] {yes}}
				edge from parent node [left,align=center] {no} }
			child { node [env]  {$x_{2}\!\leq\!3?$}
				child {node [env] {$h(\featurevec)\!=\!\truelabel^{(1)}$}   edge from parent node [left,align=center] {no} }
				child {node [env] {$h(\featurevec)\!=\!\truelabel^{(4)}$}   edge from parent node [left,align=center] {yes} }
				edge from parent node [right,align=center] {yes}};
	\end{tikzpicture}}
\end{minipage}
\caption{Using a sufficiently large (deep) \gls{decisiontree}, we can construct a map $h$ that perfectly fits any given 
	labeled dataset $\{(\featurevec^{(\sampleidx)},\truelabel^{(\sampleidx)} ) \}_{\sampleidx=1}^{\samplesize}$ such that 
	$h(\featurevec^{(\sampleidx)})\!=\!\truelabel^{(\sampleidx)}$ for $\sampleidx=1,\ldots,\samplesize$.}
\label{fig_decisiontree_overfits}
\end{figure}

\newpage
\section{Deep Learning} 
\label{sec_deep_learning}

Another example of a \gls{hypospace} uses a signal-flow representation of a hypothesis map $h: \mathbb{R}^{\featuredim} \rightarrow \mathbb{R}$. 
This signal-flow representation is referred to as a \gls{ann}. As the name indicates, an \gls{ann} is a network of 
interconnected elementary computational units. These computational units might be referred to as \index{artificial neuron}artificial 
neurons or just neurons. 

Figure \ref{fig_activate_neuron} depicts the simplest possible \gls{ann} that consists of a single neuron. 
The neuron computes a weighted sum of the inputs and then applies an \gls{actfun} $\actfun(z)$ to produce 
the output $\actfun(z)$. The $\featureidx$-th input of a neuron is assigned a parameter or weight $\weight_{\featureidx}$. 
For a given choice of weights the \gls{ann} in Figure \ref{fig_activate_neuron} represents a hypothesis map 
$h^{(\weights)}(\featurevec) = \actfun(z) = \actfun \big( \sum_{\featureidx} \weight_{\featureidx} \feature_{\featureidx} \big)$. 

The \gls{ann} in Figure \ref{fig_activate_neuron} defines a hypothesis space that is constituted by all 
maps $h^{(\weights)}$ obtained for different choices for the weights $\weights$ in Figure \ref{fig_activate_neuron}. 
Note that the single-neuron \gls{ann} in Figure \ref{fig_activate_neuron} reduces to a linear map 
when we use the \gls{actfun} $\actfun(z) =  z$. However, even when we use a non-linear \gls{actfun} in Figure \ref{fig_activate_neuron}, the resulting 
hypothesis space is essentially the same as the space of linear maps \eqref{equ_lin_hypospace}. In particular, if we threshold 
the output of the \gls{ann} in Figure \ref{fig_activate_neuron} to obtain a binary label, we will always 
obtain a \gls{linclass} like \gls{logreg} and \gls{svm} (see Section \ref{sec_LogReg} and Section \ref{sec_SVM}). 

\begin{figure}[htbp]
	\centering
	\begin{tikzpicture}[
		plain/.style={
			draw=none,
			fill=none,
		},
		net/.style={
			matrix of nodes,
			nodes={
				draw,
				circle,
				inner sep=10pt
			},
			nodes in empty cells,
			column sep=2cm,
			row sep=-9pt
		},
		>=latex
		]
		\matrix[net] (mat)
		{
			|[plain]| &|[plain]| \\
			|[plain]|& |[plain]| \\
			|[plain]| & |[plain]| \\
			|[plain]|& & |[plain]|\\
			|[plain]|& |[plain]| \\
			|[plain]| & |[plain]| \\
			|[plain]| &|[plain]| \\  };
		\draw[->] (mat-2-1) node[]{$\feature_{1}$} --  node[above]{$\weight_{1}$}  (mat-4-2);
		\draw[->] (mat-4-1) node[]{$\feature_{2}$}  --  node[above]{$\weight_{2}$}  (mat-4-2) ;
		\draw[->] (mat-6-1) node[]{$\feature_{3}$}--  node[above]{$\weight_{3}$}  (mat-4-2);
		\draw[->] (mat-4-2) -- node[above] {$\actfun(z)$} +(2cm,0);
	\end{tikzpicture}
	\caption{\Gls{ann} consisting of a single neuron that implements 
		a weighted summation $z=\sum_{\featureidx} \weight_{\featureidx} \feature_{\featureidx}$ of 
		its inputs $\feature_{\featureidx}$ followed by applying a non-linear \gls{actfun} $\actfun(z)$.}
	\label{fig_activate_neuron}
\end{figure}

Deep learning methods use \gls{ann} consisting of many (thousands to millions) interconnected neurons \cite{Goodfellow-et-al-2016}. 
In principle the interconnections between neurons can be arbitrary. One widely used approach however is to organize neurons as \index{layers} 
layers and place connections mainly between neurons in consecutive layers \cite{Goodfellow-et-al-2016}. 
Figure \ref{fig_ANN} depicts an example for a \gls{ann} consisting of one hidden layer to 
represent a (parametrized) hypothesis  $h^{(\weights)}: \mathbb{R}^{\featuredim} \rightarrow \mathbb{R}$. 
\begin{figure}[htbp]
\centering
\begin{tikzpicture}[
	plain/.style={
		draw=none,
		fill=none,
	},
	net/.style={
		matrix of nodes,
		nodes={
			draw,
			circle,
			inner sep=10pt
		},
		nodes in empty cells,
		column sep=1cm,
		row sep=-11pt
	},
	>=latex, 
	scale=0.6
	]
	\matrix[net] (mat)
	{
		|[plain]| \parbox{1cm}{\centering input\\layer} & |[plain]| \parbox{1cm}{\centering hidden\\layer} & |[plain]| \parbox{1cm}{\centering output\\layer} \\
		|[plain]|&  |[plain]|\\
		|[plain]| & \\
		& |[plain]| \\
		|[plain]| & |[plain]| \\
		& & \\
		|[plain]| & |[plain]| \\
		|[plain]| & |[plain]| \\
		|[plain]| & \\
		|[plain]| & |[plain]| \\    };
	\draw[<-] (mat-4-1) -- node[above] {$\feature_{1}$} +(-2cm,0);
	\draw[<-] (mat-6-1) -- node[above] {$\feature_{2}$} +(-2cm,0);
	\draw[->] (mat-4-1) --  node[above]{$\weight_{1}$}  (mat-3-2);
	\draw[->] (mat-6-1) --  node[above]{$\weight_{2}$}  (mat-3-2);
	\draw[->] (mat-4-1) --  node[]{\hspace*{10mm}$\weight_{3}$}  (mat-6-2);
	\draw[->] (mat-6-1) --  node[above]{$\weight_{4}$}  (mat-6-2);
	\draw[->] (mat-4-1) --  node[]{\hspace*{10mm}$\weight_{5}$}  (mat-9-2);
	\draw[->] (mat-6-1) --  node[above]{$\weight_{6}$}  (mat-9-2);
	\draw[->] (mat-3-2) --  node[]{\hspace*{10mm}$\weight_{7}$}  (mat-6-3);
	\draw[->] (mat-6-2) --  node[above]{$\weight_{8}$}  (mat-6-3);
	\draw[->] (mat-9-2) --  node[above]{$\weight_{9}$}  (mat-6-3);     
	\draw[->] (mat-6-3) -- node[above] {$h^{(\weights)}(\featurevec)$} +(2.8cm,0);
	\vspace*{-4mm}
\end{tikzpicture}
\vspace*{-4mm}
\caption{\Gls{ann} representation of a predictor $h^{(\weights)}(\featurevec)$ which maps 
	the input (feature) vector $\featurevec=(\feature_{1},\feature_{2})^{T}$ to a predicted label (output) $h^{(\weights)}(\featurevec)$.
  This \gls{ann} defines a \gls{hypospace} consisting of all maps $h^{(\weights)}(\feature)$ obtained 
from all possible choices for the weights $\weights = (\weight_{1},\ldots,\weight_{9})^{T}$}
\label{fig_ANN}
\end{figure}
The first layer of the \gls{ann} in Figure \ref{fig_ANN} is referred to as the input layer. The input layer reads in 
the feature vector $\featurevec \in \mathbb{R}^{\featuredim}$ of a \gls{datapoint}. The features $\feature_{\featureidx}$ 
are then multiplied with the weights $\weight_{\featureidx,\featureidx'}$ associated with the link between the $\featureidx$th input node (``neuron'') 
with the $\featureidx'$th node in the middle (hidden) layer. The output of the $\featureidx'$-th node in the hidden layer is given by 
$s_{\featureidx'}\!=\!\actfun( \sum_{\featureidx=1}^{\featuredim} \weight_{\featureidx,\featureidx'} \feature_{\featureidx} )$ with some 
(typically non-linear) \gls{actfun} $\actfun: \mathbb{R} \rightarrow \mathbb{R}$. The argument to the \gls{actfun} is the 
weighted combination $ \sum_{\featureidx=1}^{\featuredim} \weight_{\featureidx,\featureidx'} s_{\featureidx'}$ of 
the outputs $s_{\featureidx}$ of the nodes in the previous layer. For the \gls{ann} depicted in Figure \ref{fig_ANN}, 
the output of neuron $s_{1}$ is $\actfun(z)$ with $z=\weight_{1,1}\feature_{1} + \weight_{1,2}\feature_{2}$.

The hypothesis map represented by an \gls{ann} is parametrized by the weights of the connections between neurons. 
Moreover, the resulting hypothesis map depends also on the choice for the \gls{actfun}s of the individual neurons. These 
\gls{actfun} are a design choice that can be adapted to the statistical properties of the data. However, a few particular 
choices for the \gls{actfun} have proven useful in many important application domains. Two popular choices for the 
\gls{actfun} used within \gls{ann}s are the \index{sigmoid function}sigmoid function $\actfun(z) = \frac{1}{1+\exp(-z)}$ 
or the \index{rectified linear unit}\gls{relu} $\actfun(z) = \max\{0,z\}$. \Gls{ann}s using many, say $10$, hidden layers, is often 
referred to as a ``deep net''. ML methods using hypothesis spaces obtained from \gls{deepnet}s are known as \index{deep learning}deep 
learning methods \cite{Goodfellow-et-al-2016}. 

It can be shown that an \gls{ann} with only one single (but arbitrarily large) hidden layer can approximate 
any given map $h: \featurespace \rightarrow \labelspace=\mathbb{R}$ to any desired accuracy \cite{Cybenko1989}. 
Deep learning methods often use a \gls{ann} with a relatively large number (more than hundreds) of hidden layers. 
We refer to a \gls{ann} with a relatively large number of hidden layers as a \index{deep net}\gls{deepnet}. 

There is empirical and theoretical evidence that using many hidden layers, instead of few but wide layers, is 
computationally and statistically favourable \cite{DBLP:journals/corr/EldanS15,Poggio:2017uq} and \cite[Ch. 6.4.1.]{Goodfellow-et-al-2016}.  
The hypothesis map $h^{(\weights)}$ represented by an \gls{ann} can be evaluated (to obtain the predicted label 
at the output) efficiently using message passing over the \gls{ann}. This message passing can be implemented 
using parallel and distributed computers. Moreover, the graphical representation of a parametrized hypothesis in 
the form of a \gls{ann} allows us to efficiently compute the gradient of the loss function via a (highly scalable) 
message passing procedure known as back-propagation \cite{Goodfellow-et-al-2016}. Being able to quickly compute 
gradients is instrumental for the efficiency of gradient based methods for learning a good choice for the \gls{ann} 
weights (see Chapter \ref{ch_GD}). 

\section{Maximum Likelihood}
\label{sec_max_iikelihood}

For many applications it is useful to model the observed \gls{datapoint}s $\datapoint^{(\sampleidx)}$, with $\sampleidx=1,\ldots,\samplesize$,
as \gls{iid} realizations of a \gls{rv} $\datapoint$ with probability distribution $\prob{\datapoint; \weights}$. This probability 
distribution is parametrized by a parameter vector $\weights \in \mathbb{R}^{\featuredim}$. A principled approach to 
estimating the parameter vector $\weights$ based on a set of \gls{iid} realizations $\datapoint^{(1)},\ldots,\datapoint^{(\samplesize)}\sim \prob{\vz; \weights}$ is \index{maximum likelihood estimation} \gls{ml} estimation \cite{LC}. 

\Gls{ml} estimation can be interpreted as an ML problem with a \gls{hypospace} 
parametrized by the parameter vector $\weights$. Each element $h^{(\weights)}$ of the hypothesis 
space $\hypospace$ corresponds to one particular choice for the 
parameter vector $\weights$. \Gls{ml} methods use the \gls{lossfunc}
\begin{equation} 
\label{equ_loss_ML}
\loss{\datapoint}{h^{(\weights)}} \defeq - \log \prob{\datapoint; \weights}. 
\end{equation} 

A widely used choice for the probability distribution $p \big( \datapoint; \weights \big)$ is a multivariate normal (Gaussian)
distribution with mean ${\bm \mu}$ and covariance matrix ${\bf \Sigma}$, both of which constitute the 
parameter vector $\weights = ({\bm \mu}, {\bf \Sigma})$ (we have to reshape the matrix ${\bf \Sigma}$ suitably 
into a vector form). Given the \gls{iid} realizations $\datapoint^{(1)},\ldots,\datapoint^{(\samplesize)} \sim p \big( \datapoint; \weights \big)$,  
the \gls{ml} estimates $\hat{\bm \mu}$, $\widehat{\bf \Sigma}$ of the mean vector and 
the covariance matrix are obtained via 
\begin{equation}
\label{equ_max_likeli}
\hat{\bm \mu}, \widehat{\bf \Sigma} = \argmin_{{\bm \mu} \in \mathbb{R}^{\featurelen},{\bf \Sigma} \in \mathbb{S}^{\featurelen}_{+}} (1/\samplesize) \sum_{\sampleidx=1}^{\samplesize} - \log p\big(\datapoint^{(\sampleidx)}; ({\bm \mu},{\bf \Sigma})\big). 
\end{equation} 

The optimization in \eqref{equ_max_likeli} is over all possible choices for the mean vector ${\bm \mu} \in \mathbb{R}^{\featurelen}$ 
and the covariance matrix ${\bf \Sigma} \in \mathbb{S}^{\featurelen}_{+}$. Here, $\mathbb{S}^{\featurelen}_{+}$ 
denotes the set of all \gls{psd} Hermitian $\featurelen \times \featurelen$ matrices. 

The \gls{ml} problem \eqref{equ_max_likeli} can be 
interpreted as an instance of \gls{erm} \eqref{equ_def_ERM_funs} using the 
particular loss function \eqref{equ_loss_ML}. The resulting estimates are 
given explicitly as 
\begin{equation}
\label{equ_ML_mean_cov_Gauss} 
\hat{\bm \mu} = (1/\samplesize) \sum_{\sampleidx=1}^{\samplesize} \datapoint^{(\sampleidx)} \mbox{, and } \widehat{\bf \Sigma} = (1/\samplesize) \sum_{\sampleidx=1}^{\samplesize} (\datapoint^{(\sampleidx)} - \hat{\bm \mu})(\datapoint^{(\sampleidx)} - \hat{\bm \mu})^{T}.
\end{equation} 
Note that the expressions \eqref{equ_ML_mean_cov_Gauss} are only valid when 
the probability distribution of the \gls{datapoint}s is modelled as a multivariate normal distribution. 

\section{Nearest Neighbour Methods} 
\label{sec_nearest_neighbour_methods}

\index{nearest neighbour}\Gls{nn} methods are a family of ML methods that are characterized 
by a specific construction of the \gls{hypospace}. \Gls{nn} methods can be applied to regression 
problems involving numeric labels (e.g., using \gls{labelspace} $\labelspace=\mathbb{R}$ ) as well 
as for classification problems involving categorical labels (e.g., with \gls{labelspace} $\labelspace = \{-1,1\}$). 

While \gls{nn} methods can be combined with arbitrary \gls{labelspace}s, they require the \gls{featurespace} to have 
a specific structure. \Gls{nn} methods require the feature space be a \index{metric space}metric space \cite{RudinBookPrinciplesMatheAnalysis} 
that provides a measure for the distance between different feature vectors. We need a metric or distance measure to determine 
the nearest neighbour of a \gls{datapoint}. A prominent example for a metric feature space is the \gls{euclidspace} $\mathbb{R}^{\featuredim}$. 
The metric of $\mathbb{R}^{\featuredim}$ given by the Euclidean distance $\| \featurevec\!-\!\featurevec' \|$ between two vectors $\featurevec, \featurevec' \in \mathbb{R}^{\featuredim}$. 

Consider a \gls{trainset} $\dataset = \{ (\featurevec^{(\sampleidx)},\truelabel^{(\sampleidx)}) \}_{\sampleidx=1}^{\samplesize}$ that consists 
of labeled \gls{datapoint}s. Thus, for each \gls{datapoint} we know the features and the label value. Given such a \gls{trainset}, \gls{nn} 
methods construct a \gls{hypospace} that consist of piece-wise constant maps $h: \featurespace \rightarrow \labelspace$. 
For any hypothesis $h$ in that space, the function value $h(\featurevec)$ obtained for a \gls{datapoint} with features $\featurevec$ 
depends only on the (labels of the) $k$ nearest \gls{datapoint}s (smallest distance to $\featurevec$) in the \gls{trainset} $\dataset$. 
The number $k$ of \gls{nn}s used to determine the function value $h(\featurevec)$ is a design (hyper-) parameter of a \gls{nn} method. 
\Gls{nn} methods are also referred to as $k$-\gls{nn} methods to make their dependence on the parameter $k$ explicit. 

Let us illustrate \gls{nn} methods by considering a binary classification problem using an uneven number for $k$ (e.g., $k=3$ or $k=5$). 
The goal is to learn a hypothesis that predicts the binary label $\truelabel \in \{-1,1\}$ of a \gls{datapoint} based on its feature 
vector $\featurevec \in \mathbb{R}^{\featuredim}$. This learning task can make use of a \gls{trainset} $\dataset$ containing 
$\samplesize > k$ \gls{datapoint}s with known labels. Given a \gls{datapoint} with features $\featurevec$, denote by $\neighbourhood{k}$ 
a set of $k$ \gls{datapoint}s in $\dataset$ whose feature vectors have smallest distance to $\featurevec$. The number 
of \gls{datapoint}s in $\neighbourhood{k}$ whose label is $1$ is denoted $\samplesize^{(k)}_{1}$ and those with label value $-1$ is 
denoted $\samplesize^{(k)}_{-1}$. The $k$-\gls{nn} method ``learns'' a hypothesis $\hat{h}$ given by 
\begin{equation}
\label{equ_def_k_nn}
\hat{h}(\featurevec) = \begin{cases} 1 & \mbox{ if }  \samplesize^{(k)}_{1} > \samplesize^{(k)}_{-1} \\  - 1 & \mbox{ otherwise.} \end{cases}
\end{equation}  

It is important to note that, in contrast to the ML methods in Section \ref{sec_lin_reg} - Section \ref{sec_deep_learning}, 
the \gls{hypospace} of $k$-\gls{nn} depends on a labeled dataset (\gls{trainset}) $\dataset$. As a consequence, $k$-\gls{nn} 
methods need to access (and store) the \gls{trainset} whenever the compute a prediction (evaluate $h(\featurevec))$. 
To compute the prediction $h(\featurevec)$ for a \gls{datapoint} with features $\featurevec$, $k$-\gls{nn} needs to 
determine its \gls{nn}s in the \gls{trainset}. When using a large \gls{trainset} this implies a large storage 
requirement for $k$-\gls{nn} methods. Moreover, $k$-\gls{nn} methods might be prone to revealing sensitive 
information with its predictions (see Exercise \ref{ex_privacy_k_nn}). 

For a fixed $k$, \gls{nn} methods do not require any parameter tuning. Such parameter tuning (or learning) 
is required \gls{linreg}, \gls{logreg} and deep learning methods. In contrast, the hypothesis ``learnt'' by \gls{nn} 
methods is characterized point-wise, for each possible value of features $\featurevec$, by the \gls{nn} in the \gls{trainset}. 
Compared to the ML methods in Section \ref{sec_lin_reg} - Section \ref{sec_deep_learning}, \gls{nn} methods do 
not require to solve (challenging) optimization problems for model parameters. Beside their 
low computational requirements (put aside the memory requirements), \gls{nn} methods are 
also conceptually appealing as natural approximations of \gls{bayesestimator}s (see \cite{Cover67} 
and Exercise \ref{ex_k_nn_approximates_bayes}). 

\begin{figure}[htbp]
\centering
\begin{tikzpicture}
	\pgfmathsetseed{1908} 
	\def\pts{}
	\xintFor* #1 in {\xintSeq {1}{10}} \do{
		\pgfmathsetmacro{\ptx}{0.8*\maxxy*sin(#1 * 36 )} 
		\pgfmathsetmacro{\pty}{0.8*\maxxy*cos(#1 * 36 )} 
		
		\edef\pts{\pts, (\ptx,\pty)} 
	}
	\xintForpair #1#2 in \pts \do{
		\edef\pta{#1,#2}
		\begin{scope}
			\xintForpair \#3#4 in \pts \do{
				\edef\ptb{#3,#4}
				\ifx\pta\ptb\relax 
				\tikzstyle{myclip}=[];
				\else
				\tikzstyle{myclip}=[clip];
				\fi;
				\path[myclip] (#3,#4) to[half plane] (#1,#2);
			}
			\clip (-\maxxy,-\maxxy) rectangle (\maxxy,\maxxy); 
			\pgfmathsetmacro{\randhue}{rnd}
			\definecolor{randcolor}{hsb}{\randhue,.5,1}
			\fill[randcolor] (#1,#2) circle (4*\biglen); 
			\fill[draw=red,very thick] (#1,#2) circle (1.4pt); 
		\end{scope}
	}
	\node [below] at (0,0.8*\maxxy) {$\featurevec^{(\sampleidx)}$};
	\pgfresetboundingbox
	\draw (-\maxxy,-\maxxy) rectangle (\maxxy,\maxxy);
\end{tikzpicture}
\label{fig_voronoi}
\caption{A hypothesis map $h$ for $k$-\gls{nn} with $k=1$ and feature space $\featurespace = \mathbb{R}^{2}$. 
	The hypothesis map is constant over regions (indicated by the coloured areas) located around feature 
	vectors $\featurevec^{(\sampleidx)}$ (indicated by a dot) of some \gls{data} $\dataset=\{(\featurevec^{(\sampleidx)},\truelabel^{(\sampleidx)})\}$.}
\end{figure}

%
%

\section{Deep Reinforcement Learning}
\label{sec_reinflearning_methods}
Deep reinforcement learning (DRL) refers to a subset of ML problems and methods that revolve around 
the control of dynamic systems such as autonomous driving cars or cleaning robots \cite{Levine2016,SuttonEd2,Ng_Phd}. 
A DRL problem involves \gls{datapoint}s that represent the states of a dynamic system at different 
time instants $\timeidx=0,1,\ldots$. The \gls{datapoint}s representing the state at some time instant $\timeidx$ is characterized 
by the feature vector $\featurevec^{(\timeidx)}$. The entries of this feature vector are the individual features of 
the state at time $\timeidx$. These features might be obtained via sensors, onboard-cameras or other ML 
methods (that predict the location of the dynamic system). The label $\truelabel^{(\timeidx)}$ of a \gls{datapoint} 
might represent the optimal steering angle at time $\timeidx$. 

DRL methods learn a hypothesis $h$ that delivers optimal predictions $\hat{\truelabel}^{(\timeidx)} \defeq h \big( \featurevec^{(\timeidx)} \big)$ 
for the optimal steering angle $\truelabel^{(\timeidx)}$. As their name indicates, DRL methods use \gls{hypospace}s obtained 
from a \gls{deepnet} (see Section \ref{sec_deep_learning}). The quality of the prediction $\hat{\truelabel}^{(\timeidx)}$ obtained from a hypothesis 
is measured by the loss $\loss{\big(\featurevec^{(\timeidx)},\truelabel^{(\timeidx)} \big)}{h} \defeq - \reward^{(\timeidx)}$ with a reward signal $\reward^{(\timeidx)}$. 
This reward signal might be obtained from a distance (collision avoidance) sensor or low-level characteristics of an on-board camera 
snapshot. 

The (negative) reward signal $- \reward^{(\timeidx)}$ typically depends on the feature vector $\featurevec^{(\timeidx)}$ and 
the discrepancy between optimal steering direction $\truelabel^{(\timeidx)}$ (which is unknown) and its 
prediction $\hat{\truelabel}^{(\timeidx)} \defeq h \big( \featurevec^{(\timeidx)} \big)$. 
However, what sets DRL methods apart from other ML methods such as \gls{linreg} (see Section \ref{sec_lin_reg}) or 
\gls{logreg} (see Section \ref{sec_LogReg}) is that they can evaluate the \gls{lossfunc} only point-wise $\loss{\big(\featurevec^{(\timeidx)},\truelabel^{(\timeidx)} \big)}{h}$ for the specific hypothesis $h$ that has been used to compute the prediction $\hat{\truelabel}^{(\timeidx)} \defeq h \big( \featurevec^{(\timeidx)} \big)$ 
at time instant $\timeidx$. This is fundamentally different from linear regression that uses the squared error loss \eqref{equ_squared_loss} 
which can be evaluated for every possible hypothesis $h \in \hypospace$. 

\section{LinUCB}
\label{sec_lin_ucb}

ML methods are instrumental for various recommender systems \cite{LinUCB2010}. A basic form of a recommender 
system amount to chose at some time instant $\timeidx$ the most suitable item (product, song, movie) among a finite 
set of alternatives $\actionidx = 1,\ldots,\nractions$. Each alternative is characterized by a feature vector $\featurevec^{(\timeidx,\actionidx)}$ that 
varies between different time instants. 

The \gls{datapoint}s arising in recommender systems might represent time instants $\timeidx$ 
at which recommendations are computed. The \gls{datapoint} at time $\timeidx$ is characterized by 
a feature vector 
\begin{equation} 
\label{equ_stacked_feature_LinUCB}
\featurevec^{(\timeidx)} = \big( \big(\featurevec^{(\timeidx,1)}\big)^{T},\ldots,\big( \featurevec^{(\timeidx,\nractions)} \big)^{T} \big)^{T}.
\end{equation}  
The feature vector $\featurevec^{(\timeidx)}$ is obtained by stacking the feature vectors of alternatives at time $\timeidx$ 
into a single long feature vector. The label of the \gls{datapoint} $\timeidx$ is a vector of rewards $\labelvec^{(\timeidx)} \defeq \big ( \reward_{1}^{(\timeidx)}, \ldots,\reward_{\nractions}^{(\timeidx)}\big)^{T} \in \mathbb{R}^{\nractions}$. The entry $\reward_{\actionidx}^{(\timeidx)}$ represents 
the reward obtained by choosing (recommending) alternative $\actionidx$ (with features $\featurevec^{(\timeidx,\actionidx)}$) at time $\timeidx$. 
We might interpret the reward $\reward^{(\timeidx,\actionidx)}$ as an indicator if the costumer actually buys the product corresponding 
to the recommended alternative $\actionidx$. 

The ML method LinUCB (the name seems to be inspired by the terms ``linear'' and ``upper confidence bound'' (UCB)) 
aims at learning a hypothesis $h$ that allows to predict the rewards $\labelvec^{(\sampleidx)}$ based on the feature vector 
$\featurevec^{(\timeidx)}$ \eqref{equ_stacked_feature_LinUCB}. As its \gls{hypospace} $\hypospace$, LinUCB uses the space 
of linear maps from the stacked feature vectors $\mathbb{R}^{\featurelen \nractions}$ to the space of reward vectors 
$\mathbb{R}^{\nractions}$. This \gls{hypospace} can be parametrized by matrices $\mW \in \mathbb{R}^{\nractions \times \featurelen \nractions}$. 
Thus, LinUCB learns a hypothesis that computes predicted rewards via 
\begin{equation} 
\widehat{\labelvec}^{(\timeidx)} \defeq \mW \featurevec^{(\timeidx)}. 
\end{equation} 
The entries of $\widehat{\labelvec}^{(\timeidx)} = \big( \hat{\reward}_{1}^{(\timeidx)},\ldots,\hat{\reward}_{\nractions}^{(\timeidx)}\big)$ 
are predictions of the individual rewards $\reward^{(\timeidx,\actionidx)}$. It seems natural to recommend at time $\timeidx$ 
the alternative $\actionidx$ whose predicted reward is maximum. However, it turns out that this approach is sub-optimal 
as it prevents the recommender system from learning the optimal predictor map $\mW$. 

Loosely speaking, LinUCB tries out (explores) each alternative $\actionidx \in \{1,\ldots,\nractions\}$ sufficiently often to 
obtain a sufficient amount of training data for learning a good weight matrix $\mW$. At time $\timeidx$,  
LinUCB chooses the alternative $\actionidx^{(\timeidx)}$ that maximizes the quantity
\begin{equation} 
\label{equ_linucb_bound}
\hat{\reward}_{\actionidx}^{(\timeidx)} + R(\timeidx,\actionidx) \mbox{ , } \actionidx = 1,\ldots,\nractions.
\end{equation}  
We can think of the component $R(\timeidx,\actionidx)$ as a form of confidence interval. It is constructed such 
that \eqref{equ_linucb_bound} upper bounds the actual reward $\reward_{\actionidx}^{(\timeidx)}$ with a prescribed 
level of confidence (or probability). The confidence term $R(\timeidx,\actionidx)$ depends on the feature 
vectors $\featurevec^{(\timeidx',\actionidx)}$ of the alternative $\actionidx$ at previous time instants $\timeidx' < \timeidx$. 
Thus, at each time instant $\timeidx$, LinUCB chooses the alternative $\actionidx$ that results in the largest 
upper confidence bound (UCB) \eqref{equ_linucb_bound} on the reward (hence the ``UCB'' in LinUCB). We refer 
to the relevant literature on sequential learning (and decision making) for more details on the LinUCB \cite{LinUCB2010}.  

\section{Exercises}

\begin{exercise}[\Gls{logloss} and Accuracy.]
\label{equ_ex_3}
Section \ref{sec_LogReg} discussed \gls{logreg} as a ML method that learns a linear hypothesis map by minimizing 
the \gls{logloss} \eqref{equ_def_emp_risk_logreg}. The \gls{logloss} has computationally pleasant properties as it 
is \index{smooth}\gls{smooth} and \index{convex}\gls{convex}. However, in some applications we might be ultimately 
interested in the accuracy or (equivalently) the average $0/1$ loss \eqref{equ_def_0_1}. Can we upper bound 
the average $0/1$ loss using the average \gls{logloss} incurred by a given hypothesis on a given \gls{trainset}?
\end{exercise} 

\begin{exercise}[How Many Neurons?]
\label{exercise_how_many_neurons} 
Consider a predictor map $h(\feature)$ which is piece-wise linear and consisting of $1000$ pieces. 
Assume we want to represent this map by an \gls{ann} using neurons with one hidden layer of neurons 
having a \gls{relu} \gls{actfun}. The output layer consists of a single neuron with linear \gls{actfun}. 
How many neurons must the \gls{ann} contain at least ?
\end{exercise} 

\begin{exercise}[Effective Dimension of \gls{ann}.]
	\label{eff-dim_lin_act_fun} 
   Consider a \gls{ann} with $\featuredim=10$ input neurons following by three hidden layers consisting 
   of $4$, $9$ and $3$ nodes. The three hidden layers are followed by the output layer consisting of 
   a single neuron. Assume that all neurons use a linear \gls{actfun} and no bias term. What is the \gls{effdim} $\effdim{\hypospace}$ 
   of the hypothesis space $\hypospace$ that consists of all hypothesis maps that can be obtained from this \gls{ann}. 
\end{exercise} 

\begin{exercise}[Linear Classifiers.]
\label{exercise_linear_classifier} 
Consider \gls{datapoint}s characterized by feature vectors $\featurevec \in \mathbb{R}^{\featuredim}$ and 
binary labels $\truelabel \in\{-1,1\}$. We are interested in finding a good linear classifier which is such that 
the feature vectors resulting in $h(\featurevec) = 1$ is a half-space. Which of the methods 
discussed in this chapter aim at learning a linear classifier?
\end{exercise}

\begin{exercise}[Data Dependent \Gls{hypospace}.]
\label{exercise_data_dep_hypospace} 
Consider a ML application involving \gls{datapoint}s with features $\featurevec \in \mathbb{R}^{6}$ 
and a numeric label $\truelabel \in \mathbb{R}$. We learn a hypothesis by minimizing the average loss incurred on a 
training set $\dataset = \big\{\big(\featurevec^{(1)},\truelabel^{(1)}\big),\ldots,\big(\featurevec^{(\samplesize)},\truelabel^{(\samplesize)}\big)\big\}$. 
Which of the following ML methods uses a \gls{hypospace} that depends on the dataset $\dataset$?
\begin{itemize} 
	\item  \gls{logreg}
	\item  \gls{linreg} 
	\item k-\gls{nn} 
\end{itemize}
\end{exercise} 

\begin{exercise}[Triangle.]
Consider the \gls{ann} in Figure \ref{fig_ANN} using the \gls{relu} \gls{actfun} (see Figure \ref{fig_activate_neuron}). 
Show that there is a particular choice for the weights $\weights =(\weight_{1},\ldots,\weight_{9})^{T}$ 
such that the resulting hypothesis map $h^{(\weights)}(\feature)$ is a triangle as depicted in Figure \ref{fig_triangle}. 
Can you also find a choice for the weights $\weights =(\weight_{1},\ldots,\weight_{9})^{T}$ that produce the same triangle shape if 
we replace the \gls{relu} \gls{actfun} with the linear function $\actfun(z) =10 \cdot z$?
\begin{figure}[htbp]
	\begin{center}
		\begin{tikzpicture}
				\draw[blue,  thick, domain=-3:-1] plot (\x,  {0*\x});
				\draw[blue,  thick, domain=-1:0] plot (\x,  {1+\x}); 
				\draw[blue,  thick, domain=0:1] plot (\x,  {1-\x});    
				\draw[blue,  thick, domain=1:3] plot (\x,  {0});           
				\draw[->] (-3.5,0) -- (3.5,0) node[right] {$\feature$};
				\draw[->] (0,-0.25) -- (0,1.5) node[above] {$h(\feature)$};
				\foreach \y/\ytext in {0/0, 1/1}
				\draw[shift={(0,\y)}] (2pt,0pt) -- (-2pt,0pt) node[left] {$\ytext$};  
				\foreach \x/\xtext in {-3/-3,-2/-2,-1/-1,0/0,1/1, 2/2,3/3}
				\draw[shift={(\x,0)}] (0pt,2pt) -- (0pt,-2pt) node[below] {$\xtext$};  
		\end{tikzpicture}
		\vspace*{-3mm}
	\end{center}
	\caption{A hypothesis map $h: \mathbb{R} \rightarrow \mathbb{R}$ with the shape of a triangle.}
	\label{fig_triangle}
\end{figure}
\end{exercise}

\begin{exercise}[Approximate Triangle with Gaussians]
Try to approximate the hypothesis map depicted in Figure \ref{fig_triangle} by an element of 
$\hypospace_{\rm Gauss}$ (see \eqref{equ_def_Gauss_hypospace}) using $\sigma=1/10$, 
$\featuredim=10$ and $\mu_{\featureidx} = -1 + (2\featureidx/10)$. 
\end{exercise}

\begin{exercise}[Privacy Leakage in $k$-\gls{nn}]
\label{ex_privacy_k_nn}
Consider a $k$-\gls{nn} method for a binary classification problem. We use $k=1$ and a given \gls{trainset} 
whose \gls{datapoint}s characterize humans. Each human is characterized by a feature vector and label 
that indicates sensitive information (e.g., some sickness). Assume that you have access to the feature 
vectors of the \gls{datapoint}s in the \gls{trainset} but not to their labels. Can you infer the label value of a 
\gls{datapoint} in the \gls{trainset} based on the prediction that you obtained based on your feature vector?
\end{exercise} 

\begin{exercise}[$k$-\gls{nn} Approximates \gls{bayesestimator}]
\label{ex_k_nn_approximates_bayes}
Consider a binary classification problem involving \gls{datapoint}s that are characterized by feature vectors $\featurevec \in \mathbb{R}^{\featuredim}$ 
and binary labels $\truelabel \in \{-1,1\}$. We have access to a labeled \gls{trainset} $\dataset$ of size $\samplesize$. 
Show that the $k$-\gls{nn} hypothesis \eqref{equ_def_k_nn} is obtained from the 
\gls{bayesestimator} \eqref{equ_def_Bayes_est_binary_class} by approximating or estimating the conditional 
probability distribution $\prob{\featurevec|\truelabel}$ via the density estimator \cite[Sec. 2.5.2.]{BishopBook}
\begin{equation} 
\hat{p} (\featurevec | \truelabel ) \defeq (k/\samplesize)  \frac{1}{{\rm vol}(R_{k})}. 
\end{equation} 
Here, ${\rm vol}(R)$ denotes the volume of a ball with radius $R$ and $R_{k}$ 
is the distance between $\featurevec$ and the $k$th nearest feature vector of a \gls{datapoint} in $\dataset$.  
\end{exercise} 

\chapter{Empirical Risk Minimization}
\label{ch_Optimization}

\begin{figure}[htbp]
	\begin{center}
		\begin{tikzpicture}
			\begin{axis}
				[ylabel= {},
				xlabel={predictor $h\in \hypospace$},
				axis x line=center,
				axis y line=center,
				yscale=1.2,
				xscale=1.2,  
				xtick=\empty,
				ytick=\empty,
				xlabel style={below right},
				ylabel style={above},
				xmin=-2.5,
				xmax=2.5,
				ymin=-0.1,
				ymax=1.2
				]
				\addplot [color=black, ultra thick,dashed] table [x=a, y=b, col sep=comma] {averagepredloss.csv};      
				\addplot [color=black, ultra thick] table [x=a, y=b, col sep=comma] {emprisk.csv};      
			\end{axis}
			\node [right] at (6.5,3.7) {expected loss (or risk)};
			\node [above] at (0,5.8) {\gls{emprisk} (or \gls{trainerr})};
		\end{tikzpicture}
		\vspace*{-4mm}
	\end{center}
	\caption{ML methods learn a hypothesis $h \in \hypospace$ that incur small loss 
		when predicting the label $\truelabel$ of a \gls{datapoint} based on its features $\featurevec$. \Gls{erm} 
		approximates the expected loss or risk by the \gls{emprisk} (solid curve) incurred on a 
		finite set of labeled \gls{datapoint}s (the \gls{trainset}). Note that we can compute the \gls{emprisk} 
		based on the observed \gls{datapoint}s. However, to compute the risk we would need to know 
		the underlying \gls{probdist} which is rarely the case.}
	\label{fig_ERM_idea}
\end{figure}

Chapter \ref{ch_Elements_ML} discussed three components (see Figure \ref{fig_ml_problem}): 
\begin{itemize} 
	\item  \gls{datapoint}s characterized by features $\featurevec \in \featurespace$ and labels $ \truelabel \in \labelspace$, 
	\item a \gls{hypospace} $\hypospace$ of computationally feasible maps $h:\featurespace \rightarrow \labelspace$, 
	\item and a \gls{lossfunc} $\loss{(\featurevec,\truelabel)}{h}$ that measures the discrepancy between the predicted label $h(\featurevec)$ 
	and the true label $\truelabel$. 
\end{itemize} 
Ideally we would like to learn a hypothesis $h \in \hypospace$ such that $\loss{(\featurevec,\truelabel)}{h}$ is small 
for any \gls{datapoint} $(\featurevec,\truelabel)$. However, to implement this informal goal we need to define what 
is meant precisely by ``any \gls{datapoint}''. Maybe the most widely used approach to the define the concept of ``any \gls{datapoint}'' 
is the \gls{iidasspt}. 

The \gls{iidasspt} interprets \gls{datapoint}s as realizations of \gls{iid} \gls{rv}s 
with a common \gls{probdist} $p(\featurevec,\truelabel)$. The \gls{probdist} $p(\featurevec,\truelabel)$ 
allows us to define the risk of a hypothesis $h$ as the expectation of the \gls{loss} incurred by $h$ 
on (the realizations of) a random \gls{datapoint}. We can interpret the risk of a hypothesis as a 
measure for its quality in predicting the label of``any \gls{datapoint}''. 

If we know the \gls{probdist} $p(\featurevec,\truelabel)$ from which \gls{datapoint}s are drawn (\gls{iid}), 
we can precisely determine the hypothesis with minimum risk. This optimal hypothesis, which is 
referred to as a \gls{bayesestimator}, can be read off almost directly from the posterior \gls{probdist} 
$p(\truelabel|\featurevec)$ of the label $\truelabel$ given the features $\featurevec$ of a \gls{datapoint}. 
The precise form of the \gls{bayesestimator} depends on the choice for the \gls{lossfunc}. 
When using the squared error loss, the optimal hypothesis (or \gls{bayesestimator}) is given by 
the posterior mean $h(\featurevec) = \expect \big\{ \truelabel | \featurevec \}$ \cite{LC}. 

In most ML application, we do not know the true underlying \gls{probdist} $p(\featurevec,\truelabel)$ from 
which \gls{datapoint}s are generated. Therefore, we cannot compute the \gls{bayesestimator} exactly. However, 
we can approximately compute this estimator by replacing the exact \gls{probdist} with an estimate 
or approximation. Section \ref{sec_ERM_Bayes} discusses a specific ML method that implements this approach. 

The risk of the \gls{bayesestimator} (which is the \gls{bayesrisk}) provides a useful \gls{baseline} against which 
we can compare the average loss incurred by a ML method on a set of \gls{datapoint}s. Section \ref{sec_diagnosis_ML} 
shows how to diagnose ML methods by comparing its average loss of a hypothesis on a \gls{trainset} and its 
average loss on a \gls{valset} with a \index{baseline}\gls{baseline}. 

Section \eqref{equ_sec_emp_risk_approximates_expected_loss} motivates \gls{erm} by approximating 
the risk using the \gls{emprisk} (or average loss) computed for a set of labeled (training) \gls{datapoint}s 
(see Figure \ref{fig_ERM_idea}). This approximation is justified by the \gls{lln} which characterizes the 
deviation between averages of \gls{rv}s and their expectation. Section \ref{sec_comp_stat_ERM} discusses the statistical 
and computational aspects of \gls{erm}. We then specialize the \gls{erm} for three particular ML methods 
arising from different combinations of \gls{hypospace} and \gls{lossfunc}. Section \ref{sec_ERM_lin_reg} 
discusses \gls{erm} for linear regression (see Section \ref{sec_lin_reg}). Here, \gls{erm} amounts to minimizing 
a differentiable convex function, which can be done efficiently using \gls{gdmethods} (see Chapter \ref{ch_GD}). 

We then discuss in Section \ref{sec_ERM_decision_tree} the \gls{erm} obtained for \gls{decisiontree} models. The resulting 
\gls{erm} problems becomes a discrete optimization problem which are typically much harder than convex optimization problems. 
We cannot apply \gls{gdmethods} to solve the \gls{erm} for \gls{decisiontree}s. To solve \gls{erm} for a \gls{decisiontree}, 
we essentially must try out all possible choices for the tree structure \cite{Hyafil1976}. 

Section \ref{sec_ERM_Bayes} considers the \gls{erm} obtained when learning a linear hypothesis 
using the $0/1$ loss for classification problems. The resulting \gls{erm} amounts to minimizing 
a non-differentiable and non-convex function. Instead of applying optimization methods to solve 
this \gls{erm} instance, we will instead directly construct approximations of the \gls{bayesestimator}.

Section \ref{sec_training_inference} decomposes the operation of ML methods into \index{training period}training periods 
and \index{inference period}inference periods. The training period amounts to learning a hypothesis by solving 
the \gls{erm} on a given \gls{trainset}. The resulting hypothesis is then applied to new \gls{datapoint}s, which are 
not contained in the \gls{trainset}. This application of a learnt hypothesis to \gls{datapoint}s outside the \gls{trainset} 
is referred to as the inference period. Section \ref{sec_online_learning} demonstrates how an online learning method 
can be obtained by solving the \gls{erm} sequentially as new \gls{datapoint}s come in. Online learning methods 
alternate between training and inference periods whenever new data is collected.  

\section{Approximating Risk by Empirical Risk} 
\label{equ_sec_emp_risk_approximates_expected_loss}

The \gls{datapoint}s arising in many important application domains can be modelled (or approximated) as realizations 
of \gls{iid} \gls{rv}s with a common (joint) \gls{probdist} $p(\featurevec,\truelabel)$ for the features $\featurevec$ and 
label $\truelabel$. The \gls{probdist} $p(\featurevec,\truelabel)$ used in this \gls{iidasspt} allows us to define the 
expected loss or \gls{risk} of a hypothesis $h \in \hypospace$ as 
\begin{equation}
	\label{equ_def_risk} 
	\expect \big \{ \loss{(\featurevec,\truelabel)}{h} \}. 
\end{equation}
It seems reasonable to learn a hypothesis $h$ such that its risk \eqref{equ_def_risk} is minimal,
\begin{equation}
	\label{equ_def_risk_min} 
	\bayeshypothesis \defeq \argmin_{h \in \hypospace} \expect \big \{ \loss{(\featurevec,\truelabel)}{h} \}.
\end{equation}
We refer to any hypothesis $\bayeshypothesis$ that achieves the minimum risk \eqref{equ_def_risk_min} as 
a \gls{bayesestimator} \cite{LC}. Note that the \gls{bayesestimator} $\bayeshypothesis$ depends on both, 
the \gls{probdist} $p(\featurevec,\truelabel)$ and the \gls{lossfunc}. When using the squared 
error loss \eqref{equ_squared_loss} in \eqref{equ_def_risk_min}, the \gls{bayesestimator} $\bayeshypothesis$ 
is given by the posterior mean of $\truelabel$ given the features $\featurevec$ (see \cite[Ch. 7]{papoulis}). 

Risk minimization \eqref{equ_def_risk_min} cannot be used for the design of ML methods whenever we do not know 
the \gls{probdist} $p(\featurevec,\truelabel)$. If we do not know the \gls{probdist} $p(\featurevec,\truelabel)$, 
which is the rule for many ML applications, we cannot evaluate the expectation in \eqref{equ_def_risk}. 
One exception to this rule is if the \gls{datapoint}s are synthetically generated by drawing realizations 
from a given \gls{probdist} $p(\featurevec,\truelabel)$. 

The idea of \gls{erm} is to approximate the expectation in \eqref{equ_def_risk_min} with an average loss (the \gls{emprisk}) 
incurred on a given set of \gls{datapoint}s (the ``\gls{trainset}''), 
\begin{equation} 
\nonumber
\dataset = \big\{ \big(\featurevec^{(1)}, \truelabel^{(1)}  \big), \ldots,  \big(\featurevec^{(\samplesize)}, \truelabel^{(\samplesize)}  \big) \}. 
\end{equation}
As discussed in Section \ref{sec_empirical_risk}, this approximation is justified by 
the \gls{lln}. We obtain \gls{erm} by replacing the risk in the minimization problem \eqref{equ_def_risk_min} with 
the \gls{emprisk} \eqref{eq_def_emp_error_101}, 
\begin{align}
	\label{equ_def_ERM_funs}
	\hat{h} & \in \argmin_{h \in \hypospace} \emperror(h|\dataset) \nonumber \\ 
	& \stackrel{\eqref{eq_def_emp_error_101}}{=}  \argmin_{h \in \hypospace} (1/\samplesize) \sum_{\sampleidx=1}^{\samplesize} \loss{(\featurevec^{(\sampleidx)},\truelabel^{(\sampleidx)})}{h}.
\end{align}
As the notation in \eqref{equ_def_ERM_funs} indicates there might be several different 
hypotheses that minimize $\emperror(h|\dataset)$. We denote by $\hat{h}$ any of them. 
Mathematically, \gls{erm} \eqref{equ_def_ERM_funs} is just an instance of an optimization 
problem \cite{BoydConvexBook}. The optimization domain in \eqref{equ_def_ERM_funs} is 
the \gls{hypospace} $\hypospace$ of a ML method, the \index{objective function}objective 
(or cost) function is the \gls{emprisk} \eqref{eq_def_emp_error_101}. ML methods that 
learn a hypothesis via \gls{erm} \eqref{equ_def_ERM_funs} are instances of optimization 
algorithms \cite{OptMLBook}.  

\Gls{erm} \eqref{equ_def_ERM_funs} is a form of \index{learning by trial and error}``learning by trial and error''. 
A (hypothetical) instructor (or supervisor) provides us the labels $\truelabel^{(\sampleidx)}$ for the \gls{datapoint}s in $\dataset$ 
which are characterized by features $\featurevec^{(\sampleidx)}$. This dataset serves as a \gls{trainset} in the following sense. 
We use a current guess for a good hypothesis $h$ to predict the labels $\truelabel^{(\sampleidx)}$ of the \gls{datapoint}s in $\dataset$ 
only from their features $\featurevec^{(\sampleidx)}$ . We then determine average loss $\emperror(h|\dataset)$ that is incurred by the 
predictions $\hat{\truelabel}^{(\sampleidx)} = h\big( \featurevec^{(\sampleidx)} \big)$. If the \gls{trainerr} $\emperror(h|\dataset)$ 
is too large, we try out another hypothesis map $h'$ different from $h$ with the hope of achieving a smaller \gls{trainerr} $\emperror(h'|\dataset)$. 

We highlight that \gls{erm} \eqref{equ_def_ERM_funs} is motivated by the \gls{lln}. The \gls{lln}, in turn, is only 
useful if the \gls{datapoint}s generated within an ML application can be well modelled as realizations of \gls{iid} \gls{rv}s. 
This \gls{iidasspt} is one of the most widely used working assumptions for the design and analysis of ML methods. 
However, there are many important application domains involving \gls{datapoint}s that clearly violate this \gls{iidasspt}. 

One example for \gls{noniid} \gls{data} is time series data that consists of temporally ordered (consecutive) \gls{datapoint}s \cite{Brockwell91,Luetkepol2005}. Each \gls{datapoint} in a time series might represent a specific time interval or single time instants. 
Another example for \gls{noniid} data arises in \index{active learning}active learning where ML methods 
actively choose (or query) new \gls{datapoint}s \cite{Cohn1996}. For a third example of \gls{noniid} data, 
we refer to \gls{fl} applications that involve collections (networks) of data generators with different 
statistical properties \cite{pmlr-v54-mcmahan17a,JungNetExp2020,LocalizedLinReg2019,Tran2020,SattlerClusteredFL2020}. 
A detailed discussion of ML methods for \gls{noniid} data is beyond the scope of this book. 

\section{Computational and Statistical Aspects of ERM}
\label{sec_comp_stat_ERM}

Solving the optimization problem \eqref{equ_def_ERM_funs} provides two things. First, the 
minimizer $\hat{h}$ is a predictor which performs optimal on the \gls{trainset} $\dataset$. 
Second, the corresponding objective value $\emperror(\hat{h}|\dataset)$ (the ``\gls{trainerr}'') 
can be used to estimate for the risk or expected loss of $\hat{h}$. However, as we will discuss 
in Chapter \ref{ch_overfitting_regularization}, for some datasets $\dataset$, the \gls{trainerr} 
$\emperror(\hat{h}|\dataset)$ obtained for $\dataset$ can be very different from the 
expected loss (risk) of $\hat{h}$ when applied to new \gls{datapoint}s which are not contained in $\dataset$. 

The \gls{iidasspt} implies that the \gls{trainerr} $\emperror(h|\dataset)$ is only a noisy approximation 
of the risk $\risk{h}$. The \gls{erm} solution $\hat{h}$ is a minimizer of this noisy approximation and 
therefore in general different from the \gls{bayesestimator} which minimizes the risk itself. 
Even if the hypothesis $\hat{h}$ delivered by \gls{erm} \eqref{equ_def_ERM_funs} has 
small \gls{trainerr} $\emperror(\hat{h}|\dataset)$, it might have unacceptably large risk $\risk{\hat{h}}$. 
We refer to such a situation as overfitting and will discuss techniques for detecting and avoiding it 
in Chapter \ref{ch_validation_selection}.  

Many important ML methods use hypotheses that are parametrized by a parameter vector $\weights$. For 
each possible parameter vector, we obtain a hypothesis $h^{(\weights)}(\featurevec)$. Such a parametrization 
is used in \gls{linreg} methods which learn a linear hypothesis $h^{(\weights)}(\featurevec) = \weights^{T} \featurevec$
 with some parameter vector $\weights$. Another example for such a parametrization is obtained from \gls{ann}s 
 with the weights assigned to inputs of individual (artificial) neurons (see Figure \ref{fig_ANN}). 

For ML methods that use a parametrized hypothesis $h^{(\weights)}(\featurevec)$, 
we can reformulate the optimization problem \eqref{equ_def_ERM_funs} 
as an optimization of the parameter vector, 
\begin{equation}
	\label{eq_def_ERM_weight}
	\widehat{\weights} = \argmin_{\weights \in \mathbb{R}^{\featuredim}} f(\weights) \mbox{ with } f(\weights) \defeq
	 \underbrace{(1/\samplesize) \sum_{\sampleidx=1}^{\samplesize} \loss{(\featurevec^{(\sampleidx)},\truelabel^{(\sampleidx)})}{h^{(\weights)}}}_{\emperror\big(h^{(\weights)} |\dataset\big)}. 
\end{equation}
The objective function $f(\weights)$ in \eqref{eq_def_ERM_weight} is the \gls{emprisk} $\emperror\big(h^{(\weights)} |\dataset\big)$ 
incurred  by the hypothesis $h^{(\weights)}$ when applied to the \gls{datapoint}s in the dataset $\dataset$. 
The optimization problems \eqref{eq_def_ERM_weight} and \eqref{equ_def_ERM_funs} 
are fully equivalent. Given the optimal parameter vector $\widehat{\weights}$ solving \eqref{eq_def_ERM_weight}, 
the hypothesis $h^{(\widehat{\weights})}$ solves \eqref{equ_def_ERM_funs}. 

We highlight that the precise shape of the objective function $f(\weights)$ in \eqref{eq_def_ERM_weight} 
depends on the parametrization of the hypothesis space $\hypospace$. The parametrization is the precise 
rule that assigns a hypothesis map $h^{(\weights)}$ to a given parameter vector $\weights$. The shape of $f(\weights)$ 
depends also on the choice for the loss function $\loss{(\featurevec^{(\sampleidx)},\truelabel^{(\sampleidx)})}{h^{(\weights)}}$. 
As depicted in Figure \ref{fig_diff_types_bojec}, the different combinations of parametrized \gls{hypospace} 
and loss functions can result in objective functions with fundamentally different properties such that their 
optimization is more or less difficult. 

The objective function $f(\weights)$ for the \gls{erm} obtained for \gls{linreg} (see Section \ref{sec_lin_reg}) is differentiable 
and convex and can therefore be minimized using simple \gls{gdmethods} (see Chapter \ref{ch_GD}). In contrast, 
the objective function $f(\weights)$ of \gls{erm} obtained for \gls{ladregression} or the \gls{svm} (see Section \ref{sec_lad} and \ref{sec_SVM}) 
is non-differentiable but still convex. The minimization of such functions is more challenging but still tractable as there exist 
efficient convex optimization methods which do not require differentiability of the objective function \cite{ProximalMethods}. 

The objective function $f(\weights)$ obtained for \gls{ann} are typically highly non-convex with many local minima (see Figure \ref{fig_diff_types_bojec}). 
The optimization of \index{non-convex}non-convex objective function is in general more difficult than 
optimizing convex objective functions. However, it turns out that despite the non-convexity, iterative \gls{gdmethods} 
can still be successfully applied to solve the resulting \gls{erm} \cite{Goodfellow-et-al-2016}. 
The implementation of \gls{erm} might be even more challenging for ML methods that use \gls{decisiontree}s  
or the $0/1$ \gls{loss} \eqref{equ_def_0_1}.  
Indeed, the \gls{erm} obtained for ML methods using \gls{decisiontree}s or the $0/1$ \gls{loss} \eqref{equ_def_0_1}
involve non-differentiable objective functions which are harder to minimize compared with \gls{smooth}  
functions (see Section \ref{sec_ERM_decision_tree}). 


\begin{figure}[htbp]
	\begin{center}
		\scalebox{0.8}{
			\begin{tikzpicture}[scale=0.5]
					\draw[thick, domain=-3:3]  node[left=2cm,below] {\gls{smooth} and \gls{convex}} plot ({\x-5},  {0.2*\x*\x}) node [above] {$f(\vw)$} ;
					\draw[thick, domain=-3:3] plot ({\x+5},  {0.1*\x*\x*\x +1}) node[below=2cm] {smooth and non-convex} ;                   
					\draw[thick, domain=-3:0] plot ({\x-5},  {-\x-5}) ;
					\draw[thick, domain=0:3] plot ({\x-5},  {-5}) node[left=3cm,below] {\gls{nonsmooth} and \gls{convex}};
					\draw[thick, domain=-3:0] plot ({\x+5},  { -4})  ;         
					\draw[thick, domain=0:3] plot ({\x+5},  { -5}) node[below] {\gls{nonsmooth} and non-convex} ;         
		\end{tikzpicture}}
	\end{center}
	\caption{Different types of objective functions that arise in \gls{erm} for different combinations of \gls{hypospace} and \gls{lossfunc}.}
	\label{fig_diff_types_bojec}
\end{figure}

\section{ERM for Linear Regression}
\label{sec_ERM_lin_reg}

As discussed in Section \ref{sec_lin_reg}, linear regression methods learn a linear 
hypothesis $h^{(\weights)}(\featurevec) = \weights^{T} \featurevec$ with minimum 
squared error loss \eqref{equ_squared_loss}. For \gls{linreg}, the \gls{erm} problem \eqref{eq_def_ERM_weight} becomes 
\begin{align}
	\label{equ_def_cost_MSE}
	\widehat{\weights} & = \argmin_{\weights \in \mathbb{R}^{\featuredim}} (1/\samplesize) \sum_{\samplesize=1}^{\samplesize} \big(\truelabel^{(\sampleidx)} \!-\! \weights^{T}  \featurevec^{(\sampleidx)}  \big)^2.
\end{align} 
Here, $\samplesize=|\dataset|$ denotes the \gls{samplesize} of the \gls{trainset} $\dataset$. 
The objective function $f(\weights)$ in \eqref{equ_def_cost_MSE} is \index{convex}convex and \index{smooth}smooth. 
Such a function can be minimized using the \gls{gdmethods} discussed in Chapter \ref{ch_GD}. 

We can rewrite the \gls{erm} problem \eqref{equ_def_cost_MSE} more concisely 
by stacking the labels $\truelabel^{(\sampleidx)}$ and feature vectors $\featurevec^{(\sampleidx)}$, for 
$\sampleidx=1,\ldots,\samplesize$, into a ``label vector'' $\labelvec$ and ``feature matrix'' $\featuremtx$, 
\begin{align}
	\label{equ_def_vec_matrix}
	\labelvec & = ( \truelabel^{(1)},\ldots, \truelabel^{(\samplesize)})^{T} \in \mathbb{R}^{\samplesize} \mbox{, and } \nonumber \\ 
	\featuremtx & = \begin{pmatrix} \featurevec^{(1)},\ldots,\featurevec^{(\samplesize)} \end{pmatrix}^{T}\in \mathbb{R}^{\samplesize \times \featuredim}.
\end{align}
This allows us to rewrite the objective function in \eqref{equ_def_cost_MSE} as 
\begin{equation}
	\label{equ_min_lin_pred_vec_mat}
(1/\samplesize) \sum_{\samplesize=1}^{\samplesize} \big(\truelabel^{(\sampleidx)} \!-\! \weights^{T}  \featurevec^{(\sampleidx)}  \big)^2 = (1/\samplesize) \| \labelvec - \featuremtx\weights \|^{2}_{2}.
\end{equation} 

Inserting \eqref{equ_min_lin_pred_vec_mat} into \eqref{equ_def_cost_MSE}, 
allows to rewrite the \gls{erm} problem for \gls{linreg} as 
\begin{align}
	\label{equ_def_cost_MSE_least_squard_errors}
	\widehat{\weights} & = \argmin_{\weights \in \mathbb{R}^{\featuredim}} (1/\samplesize) \| \labelvec - \featuremtx\weights \|^{2}_{2}. 
\end{align} 
The formulation \eqref{equ_def_cost_MSE_least_squard_errors} allows for an interesting geometric 
interpretation of linear regression. Solving \eqref{equ_def_cost_MSE_least_squard_errors} amounts to 
finding a vector $\featuremtx\weights$ (with feature matrix $\featuremtx$ \eqref{equ_def_vec_matrix}), 
that is closest (in the Euclidean norm) to the label vector $\labelvec \in \mathbb{R}^{\samplesize}$ \eqref{equ_def_vec_matrix}. 
The solution to this approximation problem is precisely the orthogonal projection of the vector $\labelvec$ 
onto the subspace of $\mathbb{R}^{\samplesize}$ that is spanned by the columns of the 
feature matrix $\featuremtx$ (see Figure \ref{fig_orthogo_ERM_linreg_norma}). 

\begin{figure}
	\begin{center}
		\begin{tikzpicture}[dot/.style={circle,inner sep=1pt,fill,label={#1},name=#1},
			extended line/.style={shorten >=-#1,shorten <=-#1},
			extended line/.default=1cm]
			\coordinate (A) at (-5,0);
			\coordinate (optw) at (3,0) ; 
			\coordinate (error) at (3,2); 
			\coordinate (labelvec) at (3,4); 
			\coordinate (B) at (5,0);
			\node[right] at (error)   (a) {$\| \labelvec - \featuremtx \widehat{\weights}\|$};
			\draw [dashed,line width=0.5pt]  ([xshift=-1cm]A)  --  ([xshift=-1cm]A) node [left]{$\{ \featuremtx\weights: \weights \in \mathbb{R}^{\featuredim}\}$} ;
			\draw [dashed,line width=0.5pt] ([xshift=-1cm]A) -- ([xshift=1cm]B) ; 
			\draw [->,,line width=2pt] (A) -- (labelvec) node[above right]{$\labelvec$} ; 
			\draw [dotted,,line width=2pt] (labelvec) -- (optw); 
			\draw [->,,line width=2pt] (A) -- (optw) node[above right]{$\featuremtx\widehat{\weights}$} ; 
		\end{tikzpicture}
	\end{center}
	\caption{The \gls{erm} \eqref{equ_def_cost_MSE_least_squard_errors} for \gls{linreg} amounts to an orthogonal 
		projection of the label vector $\labelvec=\big(\truelabel^{(1)},\ldots,\truelabel^{(\samplesize)}\big)^{T}$ on the 
		subspace spanned by the columns of the feature matrix $\featuremtx = \big(\featurevec^{(1)},\ldots,\featurevec^{(\samplesize)} \big)^{T}$.}
	\label{fig_orthogo_ERM_linreg_norma}
\end{figure}

To solve the optimization problem \eqref{equ_def_cost_MSE_least_squard_errors}, it is convenient to rewrite it 
as the quadratic problem 
\begin{align}
	\label{equ_quadr_form_linreg}
	& \min_{\weights \in \mathbb{R}^{\featuredim}} \underbrace{(1/2) \weights^{T} \mathbf{Q} \weights - \vq^{T}  \weights}_{= f(\weights)} \nonumber \\
	& \mbox{ with } \mathbf{Q}= (1/\samplesize) \featuremtx^{T} \featuremtx, \mathbf{q} =(1/\samplesize) \featuremtx^{T} \labelvec. 
\end{align} 
Since $f(\weights)$ is a differentiable and convex function, a necessary and sufficient condition for 
$\widehat{\weights}$ to be a minimizer $f(\widehat{\weights})\!=\!\min_{\weights \in \mathbb{R}^{\featuredim}} f(\weights)$ is the 
{\bf zero-gradient condition} \cite[Sec. 4.2.3]{BoydConvexBook}
\begin{equation}
	\label{equ_zero_gradient}
	\nabla f(\widehat{\weights}) = \mathbf{0}.
\end{equation} 

Combining \eqref{equ_quadr_form_linreg} with \eqref{equ_zero_gradient}, yields the following necessary 
and sufficient condition for a parameter vector $\widehat{\weights}$ to solve the \gls{erm} \eqref{equ_def_cost_MSE},
\begin{equation}
	\label{equ_zero_gradient_lin_reg}
	(1/\samplesize) \featuremtx^{T} \featuremtx\widehat{\weights} = (1/\samplesize) \featuremtx^{T} \labelvec.  
\end{equation} 
This condition can be rewritten as 
\begin{equation}
	\label{equ_zero_gradient_lin_reg_normal_condition}
	(1/\samplesize) \featuremtx^{T} \big( \labelvec - \featuremtx \widehat{\weights} \big) = \mathbf{0}.  
\end{equation} 
As indicated in Figure \ref{fig_orthogo_ERM_linreg_norma}, the optimality condition 
\eqref{equ_zero_gradient_lin_reg_normal_condition} requires the vector $$\big( \labelvec - \featuremtx\widehat{\vw}\big) = \big(\big(\truelabel^{(1)}-\hat{\truelabel}^{(1)}\big),\ldots,\big(\truelabel^{(\samplesize)}-\hat{\truelabel}^{(\samplesize)}\big)  \big)^{T},$$
whose entries are the prediction errors for the \gls{datapoint}s in the \gls{trainset}, to be orthogonal (or normal) 
to the subspace spanned by the columns of the feature matrix $\featuremtx$. In view of this geometric interpretation, 
we refer to \eqref{equ_zero_gradient_lin_reg_normal_condition} as a \index{normal equation}``normal equation''. 

It can be shown that, for any given feature matrix $\featuremtx$ and label vector $\labelvec$, there always 
exists at least one optimal parameter vector $\widehat{\weights}$ which solves \eqref{equ_zero_gradient_lin_reg}. 
The optimal parameter vector might not be unique, i.e., there might be several different parameter vectors 
achieving the minimum in \eqref{equ_def_cost_MSE}. However, 
every vector $\widehat{\weights}$ which solves \eqref{equ_zero_gradient_lin_reg} 
achieves the same minimum \gls{emprisk}
\begin{equation}
	\label{equ_emp_risk_lin_proje}
	\emperror(h^{(\widehat{\weights})} \mid \dataset) = \min_{\weights \in \mathbb{R}^{\featuredim}} \emperror(h^{(\weights)} \mid \dataset) = \|  (\mathbf{I}- \mathbf{P}) \labelvec \|^{2}.
\end{equation} 
Here, we used the orthogonal projection matrix $\mathbf{P} \in \mathbb{R}^{\samplesize \times \samplesize}$ 
on the linear span of the feature matrix $\featuremtx = (\featurevec^{(1)},\ldots,\featurevec^{(\samplesize)})^{T} \in \mathbb{R}^{\samplesize \times \featuredim}$ (see \eqref{equ_def_vec_matrix}). The linear span of a matrix $\mA=(\va^{(1)},\ldots,\va^{(\samplesize)}) \in \mathbb{R}^{\featuredim \times \samplesize}$, denoted as ${\rm span } \big\{\mA\}$, is the subspace of $\mathbb{R}^{\featuredim}$ consisting of all 
linear combinations of the columns $\va^{(r)} \in \mathbb{R}^{\featuredim}$ of $\mA$. 

If the columns of the feature matrix $\featuremtx$ (see \eqref{equ_def_vec_matrix}) are linearly independent, 
which implies that the matrix $\featuremtx^{T} \featuremtx$ is invertible, the projection matrix $\mathbf{P}$ is given explicitly as 
\begin{equation} 
	\nonumber
	\mathbf{P} = \featuremtx \big( \featuremtx^{T} \featuremtx \big)^{-1} \featuremtx^{T}. 
\end{equation} 
Moreover, the solution of \eqref{equ_zero_gradient_lin_reg} is then unique and given by 
\begin{equation}
	\label{equ_close_form_lin_reg}
	\widehat{\weights} = \big(  \featuremtx^{T} \featuremtx \big)^{-1} \featuremtx^{T} \labelvec. 
\end{equation}
The closed-form solution \eqref{equ_close_form_lin_reg} requires the inversion of the $\featuredim \times \featuredim$ matrix $\featuremtx^{T} \featuremtx$. 

Note that formula \eqref{equ_close_form_lin_reg} is only valid if the matrix $\featuremtx^{T} \featuremtx$ is invertible. 
The feature matrix $\featuremtx$ is determined by the \gls{datapoint}s obtained in a ML application. Its properties 
are therefore not under the control of a ML method and it might well happen that the matrix $\featuremtx^{T} \featuremtx$ 
is not invertible. As a point in case, the matrix $\featuremtx^{T} \featuremtx$ cannot be invertible for any 
dataset containing fewer \gls{datapoint}s than the number of features used to characterize \gls{datapoint}s (this is 
referred to as \index{high-dimensional data}high-dimensional data). Moreover, the matrix $\featuremtx^{T} \featuremtx$ 
is not invertible if there two co-linear features $\feature_{\featureidx},\feature_{\featureidx'}$ such that 
$\feature_{\featureidx} = \beta \feature_{\featureidx'}$ holds for any \gls{datapoint} with some constant $\lrate \in \mathbb{R}$. 

Let us now consider a dataset such that the feature matrix $\featuremtx$ is not full column-rank and, in turn, the matrix 
$\featuremtx^{T} \featuremtx$ is not invertible. In this case we cannot use \eqref{equ_close_form_lin_reg} to compute 
the optimal parameter vector since the inverse of $\featuremtx^{T} \featuremtx$ does not exist. Moreover, in this case, there 
are infinitely many different parameter vectors that solve \eqref{equ_zero_gradient_lin_reg}, i.e., the corresponding linear hypothesis map incurs minimum average squared error loss on the \gls{trainset}. Section \ref{sec_data_augmentation} explains the benefits of using weights with 
small Euclidean norm. The parameter vector $\widehat{\weights}$ solving the \gls{linreg} optimality condition \eqref{equ_zero_gradient_lin_reg} 
and having minimum Euclidean norm among all such vectors is given by 
\begin{equation}
	\widehat{\weights} = \big(  \featuremtx^{T} \featuremtx \big)^{\dagger} \featuremtx^{T} \labelvec. 
\end{equation} 
Here, $\big(  \featuremtx^{T} \featuremtx \big)^{\dagger}$ denotes the \index{pseudoinverse}pseudoinverse (or the 
Moore–Penrose inverse) of $\featuremtx^{T} \featuremtx$ (see \cite{golub96,Golub1980}).  

Computing the (pseudo-)inverse of $\featuremtx^{T} \featuremtx$ can be computationally challenging for large 
number $\featuredim$ of features. Figure \ref{fig_snapshot_pixels} depicts a simple ML problem where the 
number of features is already in the millions. The computational complexity of inverting the matrix $\featuremtx^{T} \featuremtx$ 
depends crucially on its \gls{condnr}. We refer to a matrix as ill-conditioned if its \gls{condnr} is much larger than 1. 
In general, ML methods do not have any control on the \gls{condnr} of the matrix $\featuremtx^{T} \featuremtx$. Indeed, 
this matrix is determined solely by the (features of the) \gls{datapoint}s fed into the ML method. 

Section \ref{sec_GD_linear_regression} will discuss a method for computing the optimal parameter vector $\widehat{\weights}$ 
that does not require any matrix inversion. This method, referred to as \gls{gd} constructs a sequence 
$\weights^{(0)}, \weights^{(1)},\ldots$ of increasingly accurate approximations of $\widehat{\weights}$. 
This iterative method has two major benefits compared to evaluating the formula \eqref{equ_close_form_lin_reg} 
using direct \index{matrix inversion}matrix inversion, such as Gauss-Jordan elimination \cite{golub96}. 

First, \gls{gd} typically requires significantly fewer arithmetic operations compared to direct matrix inversion. 
This is crucial in modern ML applications involving large feature matrices. Second, \gls{gd} does 
not break when the matrix $\featuremtx$ is not full rank and the formula \eqref{equ_close_form_lin_reg} 
cannot be used any more. 

\section{ERM for Decision Trees}
\label{sec_ERM_decision_tree}

Consider \gls{erm} \eqref{equ_def_ERM_funs} for a regression problem with \gls{labelspace} $\labelspace=\mathbb{R}$ and 
feature space $\featurespace= \mathbb{R}^{\featuredim}$ and the \gls{hypospace} defined by \gls{decisiontree}s(see Section \ref{sec_decision_trees}). 
In stark contrast to \gls{erm} for \gls{linreg} or \gls{logreg}, \gls{erm} for \gls{decisiontree}s amounts to a discrete optimization problem. 
Consider the particular \gls{hypospace} $\hypospace$ depicted in Figure \ref{fig_hypospace_DT_depth_2}. 
This \gls{hypospace} contains a finite number of different hypothesis maps. Each individual hypothesis map 
corresponds to a particular \gls{decisiontree}. 

For the small \gls{hypospace} $\hypospace$ in Figure \ref{fig_hypospace_DT_depth_2}, \gls{erm} is easy. 
Indeed, we just have to evaluate the \gls{emprisk} (``training error'') $\emperror(h)$ for each hypothesis in $\hypospace$ 
and pick the one yielding the smallest \gls{emprisk}. However, when allowing for a very large (deep) \gls{decisiontree}, the 
computational complexity of exactly solving the \gls{erm} becomes intractable \cite{HYAFIL197615}. A popular 
approach to learn a \gls{decisiontree} is to use greedy algorithms which try to expand (grow) a given 
\gls{decisiontree} by adding new branches to leaf nodes in order to reduce the average loss on the 
\gls{trainset} (see \cite[Chapter 8]{IntroSLR} for more details). 

\begin{center}
	\framebox[0.96\textwidth]{
		\parbox{0.9\textwidth}{
			The idea behind many \gls{decisiontree} learning methods is quite simple: try out expanding a \gls{decisiontree} 
			by replacing a leaf node with a decision node (implementing another ``test'' on the feature vector) in order to 
			reduce the overall \gls{emprisk} much as possible. 
	}}
\end{center}

Consider the labeled dataset $\dataset$ depicted in Figure \ref{fig_growingatree} and a given \gls{decisiontree} for 
predicting the label $\truelabel$ based on the features $\featurevec$. We might first try a hypothesis obtained 
from the simple tree shown in the top of Figure \ref{fig_growingatree}. This hypothesis does not allow to achieve 
a small average loss on the \gls{trainset} $\dataset$. Therefore, we might grow the tree by replacing a leaf node 
with a decision node. According to Figure \ref{fig_growingatree}, to so obtained larger \gls{decisiontree} provides 
a hypothesis that is able to perfectly predict the labels of the \gls{trainset} (it achieves zero \gls{emprisk}).

\begin{figure}[htbp]
	\begin{center}
		\begin{minipage}{0.4\textwidth}
			\scalebox{0.9}{
				\begin{tikzpicture}[auto,scale=0.8]
					\draw [thick] (5,5) circle (0.1cm)node[anchor=west] {\hspace*{0mm}$\featurevec^{(3)}$};
					\draw [dashed] (0,0) rectangle (3,6) ;
					\draw [dashed] (3,0) rectangle (6,6) ;
					\draw [thick] (1,1) circle (0.1cm)node[anchor=west] {\hspace*{0mm}$\featurevec^{(4)}$};
					\draw [thick] (5,1) circle (0.1cm)node[anchor=west,above] {\hspace*{0mm}$\featurevec^{(2)}$};
					\draw [thick] (1,5) rectangle ++(0.1cm,0.1cm) node[anchor=west,above] {\hspace*{0mm}$\featurevec^{(1)}$};
					\draw[->] (-0.5,0) -- (6.5,0) node[right] {$\feature_{1}$};
					\draw[->] (0,-0.5) -- (0,6.5) node[above] {$\feature_{2}$};
					\foreach \y/\ytext in {0/0, 1/1,2/2,3/3,4/4,5/5,6/6} \draw[shift={(0,\y)}] (2pt,0pt) -- (-2pt,0pt) node[left] {$\ytext$};  
					\foreach \x/\xtext in{0/0, 1/1,2/2,3/3,4/4,5/5,6/6}\draw[shift={(\x,0)}] (0pt,2pt) -- (0pt,-2pt) node[below] {$\xtext$};  
			\end{tikzpicture}}
		\end{minipage}
		\begin{minipage}{0.5\textwidth}
			\scalebox{0.9}{
				\begin{tikzpicture} [->,>=stealth',level/.style={sibling distance = 2cm/#1,
						level distance = 1.5cm},scale=0.9]
					\vspace*{-30mm} 
					\tikzstyle{level 1}=[sibling distance=30mm]
					\tikzstyle{level 2}=[sibling distance=30mm]
					\node [env] {$\feature_{1}\!\leq\!3$?}
					child  {node [env] {$h(\featurevec)\!=\!\circ$} 
						edge from parent node [left,align=center] {no} }
					child { node [env]  {$h(\featurevec)\!=\!\Box$}
						edge from parent node [right,align=center] {yes}};
			\end{tikzpicture}}
		\end{minipage}
	\end{center}
	\vspace*{5mm}
	\begin{minipage}{0.2\textwidth}
		\begin{tikzpicture}[auto,scale=0.5]
			\draw [thick] (5,5) circle (0.2cm)node[anchor=west] {\hspace*{0mm}$\featurevec^{(3)}$};
			\draw [dashed] (3,0) rectangle (3,3) ;
			\draw [dashed] (0,0) rectangle (3,6) ;
			\draw [dashed] (3,0) rectangle (6,3) ;
			\draw [dashed] (3,3) rectangle (6,6) ;
			\draw [thick] (1,1) circle (0.2cm)node[anchor=west] {\hspace*{0mm}$\featurevec^{(4)}$};
			\draw [thick] (5,1) circle (0.2cm)node[anchor=west,above] {\hspace*{0mm}$\featurevec^{(2)}$};
			\draw [thick] (1,5) rectangle ++(0.2cm,0.2cm) node[anchor=west,above] {\hspace*{0mm}$\featurevec^{(1)}$};
			\draw[->] (-0.5,0) -- (6.5,0) node[right] {$\feature_{1}$};
			\draw[->] (0,-0.5) -- (0,6.5) node[above] {$\feature_{2}$};
		\end{tikzpicture}
	\end{minipage}
	\begin{minipage}{0.2\textwidth}
		\scalebox{0.5}{
			\begin{tikzpicture} [->,>=stealth',level/.style={sibling distance = 2cm/#1,
					level distance = 1.5cm},scale=1.5]
				\vspace*{-30mm} 
				\tikzstyle{level 1}=[sibling distance=20mm]
				\tikzstyle{level 2}=[sibling distance=25mm]
				\node [env] {$\feature_{1}\!\leq\!3$?}
				child  {node [env] {$\feature_{2}\!\leq\!3?$} 
					child {node [env] {$h(\featurevec)\!=\!\circ$}  edge from parent node [left,align=center] {no}}
					child {node [env] {$h(\featurevec)\!=\!\circ$}   edge from parent node [left,align=center] {yes}}
					edge from parent node [left,align=center] {no} }
				child { node [env]  {$h(\featurevec)\!=\!\Box$}
					edge from parent node [right,align=center] {yes}};
		\end{tikzpicture}}
	\end{minipage}
	\hspace*{10mm}
	\begin{minipage}{0.2\textwidth}
		\scalebox{1}{
			\begin{tikzpicture}[auto,scale=0.5]
				\draw [thick] (5,5) circle (0.2cm)node[anchor=west] {\hspace*{0mm}$\featurevec^{(3)}$};
				\draw [dashed] (3,0) rectangle (3,3) ;
				\draw [dashed] (0,0) rectangle (3,6) ;
				\draw [dashed] (0,0) rectangle (3,3) ;
				\draw [dashed] (0,3) rectangle (3,6) ;
				\draw [dashed] (3,0) rectangle (6,6) ;
				\draw [thick] (1,1) circle (0.2cm)node[anchor=west] {\hspace*{0mm}$\featurevec^{(4)}$};
				\draw [thick] (5,1) circle (0.2cm)node[anchor=west,above] {\hspace*{0mm}$\featurevec^{(2)}$};
				\draw [thick] (1,5) rectangle ++(0.2cm,0.2cm) node[anchor=west,above] {\hspace*{0mm}$\featurevec^{(1)}$};
				\draw[->] (-0.5,0) -- (6.5,0) node[right] {$\feature_{1}$};
				\draw[->] (0,-0.5) -- (0,6.5) node[above] {$\feature_{2}$};
		\end{tikzpicture}}
	\end{minipage}
	\hspace*{2mm}
	\begin{minipage}{0.25\textwidth}
		\scalebox{0.5}{
			\begin{tikzpicture} [->,>=stealth',level/.style={sibling distance = 2cm/#1,
					level distance = 1.5cm},scale=1.5]
				\vspace*{-30mm} 
				\tikzstyle{level 1}=[sibling distance=20mm]
				\tikzstyle{level 2}=[sibling distance=25mm]
				\node [env] {$\feature_{1}\!\leq\!3$?}
				child  {node [env] {$h(\featurevec)\!=\!\circ$} 
					edge from parent node [left,align=center] {no} }
				child { node [env]  {$\feature_{2}\!\leq\!3?$ }
					child {node [env] {$h(\featurevec)\!=\!\Box$}   edge from parent node [left,align=center] {no} }
					child {node [env] {$h(\featurevec)\!=\!\circ$}   edge from parent node [left,align=center] {yes} }
					edge from parent node [right,align=center] {yes}};
		\end{tikzpicture}}
	\end{minipage}
	\caption{Consider a given labeled dataset and the \gls{decisiontree} in the top row. We then grow the 
		\gls{decisiontree} by expanding one of its two leaf nodes. The bottom row shows the resulting 
		\gls{decisiontree}s, along with their decision boundaries. Each \gls{decisiontree} in the bottom row is 
		obtained by expanding a different leaf node of the \gls{decisiontree} in the top row.}
	\label{fig_growingatree}
\end{figure}

One important aspect of methods that learn a \gls{decisiontree} by sequentially growing the tree is the question of 
when to stop growing. A natural stopping criterion might be obtained from the limitations in computational resources, 
i.e., we can only afford to use \gls{decisiontree}s up to certain maximum depth. Besides the computational 
limitations, we also face statistical limitations for the maximum size of \gls{decisiontree}s. ML methods that allow 
for very deep  \gls{decisiontree}s, which represent highly complicated maps, tend to overfit the \gls{trainset} (see Figure \ref{fig_decisiontree_overfits} and Chapter \ref{ch_overfitting_regularization}). In particular, Even if a deep \gls{decisiontree} incurs small average 
loss on the \gls{trainset}, it might incur large loss when predicting the labels of \gls{datapoint}s outside 
the \gls{trainset}. 

\section{ERM for Bayes Classifiers} 
\label{sec_ERM_Bayes} 

The family of ML methods referred to as \gls{bayesestimator} uses the $0/1$ loss \eqref{equ_def_0_1} to 
measuring the quality of a classifier $h$. The resulting \gls{erm} is 
\begin{align}
	\label{equ_approx_bayes_class}
	\hat{h} & = \argmin_{h \in \hypospace} (1/\samplesize) \sum_{\sampleidx=1}^{\samplesize} \loss{(\featurevec^{(\sampleidx)},y^{(\sampleidx)})}{h}   \nonumber \\ 
	& \stackrel{\eqref{equ_def_0_1}}{=}  \argmin_{h \in \hypospace} (1/\samplesize) \sum_{\sampleidx=1}^{\samplesize} \mathcal{I} ( h(\featurevec^{(\sampleidx)}) \neq y^{(\sampleidx)}). 
\end{align}
The objective function in this optimization problem is non-differentiable and non-convex (see Figure \ref{fig_diff_types_bojec}). 
This prevents us from using \gls{gdmethods} (see Chapter \ref{ch_GD}) to solve \eqref{equ_approx_bayes_class}.

We will now approach the \gls{erm} \eqref{equ_approx_bayes_class} via a different 
route by interpreting the \gls{datapoint}s $(\featurevec^{(\sampleidx)},\truelabel^{(\sampleidx)})$ as 
realizations of \gls{iid} \gls{rv}s with the common \gls{probdist} $p(\featurevec,\truelabel)$. 

As discussed in Section \ref{sec_lossfct}, the \gls{emprisk} obtained using $0/1$ loss approximates 
the error probability $\prob { \hat{\truelabel} \neq \truelabel }$ with the predicted 
label $\hat{\truelabel} = 1$ for $h(\featurevec) > 0$ and $\hat{\truelabel} = -1$ otherwise (see \eqref{equ_0_1_approx_prob}). 
Thus, we can approximate the \gls{erm} \eqref{equ_approx_bayes_class} as 
\begin{align}
	\label{equ_approx_bayes_class_approx}
	\hat{h}    & \stackrel{\eqref{equ_0_1_approx_prob}}{\approx}\argmin_{h \in \hypospace} \prob{ \hat{\truelabel} \neq \truelabel} . 
\end{align}
Note that the hypothesis $h$, which is the optimization variable 
in \eqref{equ_approx_bayes_class_approx}, enters into the objective 
function of \eqref{equ_approx_bayes_class_approx} via the definition 
of the predicted label $\hat{\truelabel}$, which is $\hat{\truelabel} = 1 $ if $h(\featurevec) > 0$ 
and $\hat{\truelabel} =-1$ otherwise. 

It turns out that if we would know the \gls{probdist} $p(\featurevec,\truelabel)$, which 
is required to compute $\prob{ \hat{\truelabel} \neq \truelabel}$, the solution of \eqref{equ_approx_bayes_class_approx} 
can be found via elementary Bayesian decision theory \cite{PoorDetEst}. In particular, 
the optimal classifier $h(\featurevec)$ is such that $\hat{\truelabel}$ achieves the maximum ``a-posteriori'' 
probability $p(\hat{\truelabel}|\featurevec)$ of the label being $\hat{\truelabel}$, given (or conditioned on) the features 
$\featurevec$. 

Since we typically do not know the \gls{probdist} $p(\featurevec,\truelabel)$, we have to estimate (or approximate) it 
from the observed data points $(\featurevec^{(\sampleidx)},\truelabel^{(\sampleidx)})$. This estimation is feasible 
if the \gls{datapoint}s can be considered (approximated) as realizations of \gls{iid} \gls{rv}s with a common \gls{probdist} 
$p(\featurevec,\truelabel)$. We can then estimate (the parameters) of the \gls{probdist} $p(\featurevec,\truelabel)$ 
using \gls{ml} methods (see Section \ref{sec_max_iikelihood}). For numeric features and labels, a widely-used 
parametric \gls{probdist} $p(\featurevec,\truelabel)$ is the multivariate normal (Gaussian) distribution. In particular, 
conditioned on the label $\truelabel$, the feature vector $\featurevec$ is a Gaussian random vector with 
mean ${\bm \mu}_{\truelabel}$ and covariance ${\bf \Sigma}$, 
\begin{equation}
	\label{equ_prob_model_Bayes}
	p(\featurevec|\truelabel) = \mathcal{N}(\featurevec;{\bm \mu}_{\truelabel},{\bf \Sigma}).\footnote{We use the shorthand $\mathcal{N}(\featurevec;{\bm \mu},{\bf \Sigma})$ to denote the 
		\gls{pdf} $$p(\featurevec) = \frac{1}{\sqrt{{\rm det} (2 \pi {\bf \Sigma})}} \exp\big(- (1/2) (\featurevec\!-\!{\bm \mu})^{T}{\bf \Sigma}^{-1}(\featurevec\!-\!{\bm \mu}) \big)$$ of 
		a Gaussian random vector $\featurevec$ with mean ${\bm \mu} = \expect \{ \featurevec \}$ and covariance matrix ${\bf \Sigma} = \expect \big\{(\featurevec\!-\!{\bm \mu})  (\featurevec\!-\!{\bm \mu})^{T} \big\}$.}
\end{equation} 

The conditional expectation of the features $\featurevec$, given (conditioned on) the label $\truelabel$ of a data point, 
is ${\bm \mu}_{1}$ if $\truelabel=1$, while for $\truelabel=-1$ the conditional mean of $\featurevec$ is ${\bm \mu}_{-1}$. 
In contrast, the conditional covariance matrix ${\bf \Sigma} = \expect\{ (\featurevec-{\bm \mu}_{\truelabel})(\featurevec-{\bm \mu}_{\truelabel})^{T}|\truelabel \}$ 
of $\featurevec$ is the same for both values of the label $\truelabel \in \{-1,1\}$. The conditional \gls{probdist} 
$p(\featurevec|\truelabel)$ of the feature vector, given the label $\truelabel$, is multivariate normal. In contrast, the marginal 
distribution of the features $\featurevec$ is a \gls{gmm}. We will revisit \gls{gmm}s later in Section \ref{sec_soft_clustering} where 
we will see that they are a great tool for \index{soft clustering} \gls{softclustering}. 

For this probabilistic model of features and labels, the optimal classifier minimizing the 
error probability $\prob{\hat{\truelabel} \neq \truelabel}$ is $\hat{\truelabel}\!=\!1$ for 
$h(\featurevec)\!>\!0$ and $\hat{\truelabel}\!=\!-1$ for $h(\featurevec)\!\leq\!0$ using the classifier map 
\begin{equation}
	\label{equ_classif_Bayes}
	h(\featurevec) = \weights^{T} \featurevec \mbox{ with } \weights =  {\bf \Sigma}^{-1} ({\bm \mu}_{1} - {\bm \mu}_{-1}). 
\end{equation}
Carefully note that this expression is only valid if the matrix ${\bf \Sigma}$ is 
invertible.

We cannot implement the classifier \eqref{equ_classif_Bayes} directly, since we do not know the 
true values of the class-specific mean vectors ${\bm \mu}_{1}$, ${\bm \mu}_{-1}$ and covariance 
matrix ${\bf \Sigma}$. Therefore, we have to replace those unknown parameters with some 
estimates $\hat{\bm \mu}_{1}$, $\hat{\bm \mu}_{-1}$ and $\widehat{\bf \Sigma}$. A principled 
approach is to use the maximum likelihood estimates (see \eqref{equ_ML_mean_cov_Gauss}) 
\begin{align}
	\hat{\bm \mu}_{1}  & = (1/\samplesize_{1}) \sum_{\sampleidx=1}^{\samplesize} \mathcal{I}(y^{(\sampleidx)}=1) \featurevec^{(\sampleidx)}, \nonumber \\
	\hat{\bm \mu}_{-1}  & = (1/\samplesize_{-1}) \sum_{\sampleidx=1}^{\samplesize} \mathcal{I}(y^{(\sampleidx)}=-1) \featurevec^{(\sampleidx)}, \nonumber \\
	\hat{\bm \mu} & =  (1/\samplesize) \sum_{\sampleidx=1}^{\samplesize} \featurevec^{(\sampleidx)}, \nonumber \\
	\mbox{and } \widehat{\bf \Sigma} & = (1/\samplesize) \sum_{\sampleidx=1}^{\samplesize} (\vz^{(\sampleidx)} - \hat{\bm \mu})(\vz^{(\sampleidx)} - \hat{\bm \mu})^{T}, \label{ML_est_naive_Bayes}
\end{align}
with $\samplesize_{1} = \sum_{\sampleidx=1}^{\samplesize}  \mathcal{I}(\truelabel^{(\sampleidx)}=1)$ 
denoting the number of datapoints with label $\truelabel=1$ ($\samplesize_{-1}$ is defined similarly). 
Inserting the estimates \eqref{ML_est_naive_Bayes} into \eqref{equ_classif_Bayes} yields the 
implementable classifier 
\begin{equation}
	\label{equ_classif_Bayes_impl}
	h(\featurevec) = \weights^{T} \featurevec \mbox{ with } \weights =  \widehat{\bf \Sigma}^{-1} (\hat{\bm \mu}_{1} - \hat{\bm \mu}_{-1}). 
\end{equation} 
We highlight that the classifier \eqref{equ_classif_Bayes_impl} is only well-defined if the estimated 
covariance matrix $\widehat{\bf \Sigma}$ \eqref{ML_est_naive_Bayes} is invertible. This requires 
to use a sufficiently large number of training datapoints such that $\samplesize \geq \featuredim$. 

We derived the classifier \eqref{equ_classif_Bayes_impl} as an approximate solution 
to the \gls{erm} \eqref{equ_approx_bayes_class}. The classifier \eqref{equ_classif_Bayes_impl} 
partitions the feature space $\mathbb{R}^{\featuredim}$ into two half-spaces. One 
half-space consists of feature vectors $\featurevec$ for which the hypothesis \eqref{equ_classif_Bayes_impl} 
is non-negative and, in turn, $\hat{y}=1$. The other half-space is constituted by feature 
vectors $\featurevec$ for which the hypothesis \eqref{equ_classif_Bayes_impl} is negative and, in turn, 
$\hat{\truelabel}=-1$. Figure \ref{fig_lin_dec_boundary} illustrates these two half-spaces and the 
decision boundary between them. 

The \gls{bayesestimator} \eqref{equ_classif_Bayes_impl} is another instance of a linear 
classifier like \gls{logreg} and the \gls{svm}. Each of these methods learns a 
linear hypothesis $h(\featurevec)=\weights^{T} \featurevec$, whose decision boundary (vectors $\featurevec$ with $h(\featurevec)=0$) 
is a hyperplane (see Figure \ref{fig_lin_dec_boundary}). However, these methods 
use different loss functions for assessing the quality of a particular linear hypothesis 
$h(\featurevec)=\weights \featurevec$ (which defined the decision boundary via $h(\featurevec)=0$). 
Therefore, these three methods typically learn classifiers with different decision 
boundaries. 

For the estimator $\widehat{\bf \Sigma}$ \eqref{equ_ML_mean_cov_Gauss} 
to be accurate (close to the unknown covariance matrix) we need a number 
of datapoints (sample size) which is at least of the order $\featuredim^{2}$. 
This sample size requirement might be infeasible for applications with only few 
datapoints available. 

The maximum likelihood estimate $\widehat{\bf \Sigma}$ \eqref{ML_est_naive_Bayes} 
is not invertible whenever $\samplesize < \featuredim$. In this case, the expression 
\eqref{equ_classif_Bayes_impl} becomes useless. To cope with small sample size 
$\samplesize < \featuredim$ we can simplify the model \eqref{equ_prob_model_Bayes} 
by requiring the covariance to be diagonal ${\bf \Sigma} = {\rm diag} (\sigma_{1}^{2}, \ldots, \sigma_{\featuredim}^{2})$. 
This is equivalent to modelling the individual features $x_{1},\ldots,x_{\featuredim}$ 
of a \gls{datapoint} as conditionally independent, given its label $\truelabel$. The resulting special case of a \gls{bayesestimator} 
is often referred to as a \index{naive Bayes} ``naive Bayes'' classifier. 

We finally highlight that the classifier \eqref{equ_classif_Bayes_impl} is obtained using the 
generative model \eqref{equ_prob_model_Bayes} for the data. Therefore, \gls{bayesestimator} 
belong to the family of \index{generative mehtods}generative ML methods which involve modelling the data generation. 
In contrast, \gls{logreg} and the \gls{svm} do not require a generative model for the 
\gls{datapoint}s but aim directly at finding the relation between features $\featurevec$ 
and label $\truelabel$ of a \gls{datapoint}. These methods belong therefore to the 
family of \index{discriminative methods}discriminative ML methods.

Generative methods such as those learning a \gls{bayesestimator} are preferable for applications 
with only very limited amounts of labeled data. Indeed, having a generative model such as \eqref{equ_prob_model_Bayes} 
allows us to synthetically generate more labeled data by generating random features and labels 
according to the \gls{probdist} \eqref{equ_prob_model_Bayes}. We refer to \cite{NIPS2001_2020} 
for a more detailed comparison between generative and discriminative methods.

\section{Training and Inference Periods} 
\label{sec_training_inference} 

Some ML methods repeat the cycle in Figure \ref{fig_AlexMLBP} in a 
highly irregular fashion. Consider a large image collection which we 
use to learn a hypothesis about how cat images look like. It might 
be reasonable to adjust the hypothesis by fitting a model to the image 
collection. This fitting or training amounts to repeating the 
cycle in Figure \ref{fig_AlexMLBP} during some specific time period 
(the ``training time'') for a large number. 

After the training period, we only apply the hypothesis to predict the labels of new images. This second phase is also known 
as inference period and might be much longer compared to the training period. Ideally, we would like to only have a very short 
training period to learn a good hypothesis and then only use the hypothesis for inference. 

\section{Online Learning} 
\label{sec_online_learning}

In it most basic form, \gls{erm} requires a given set of labeled \gls{datapoint}s, which we refer 
to as the \gls{trainset}. However, some ML methods can access data only in a sequential 
fashion. As a point in case, consider time series data such as daily minimum and maximum 
temperatures recorded by a \gls{fmi} weather station. Such a time series consists of a 
sequence of \gls{datapoint}s that are generated at successive time instants. 

\index{online learning}Online learning studies ML methods that learn (or optimize) a hypothesis 
incrementally as new data arrives. This mode of operation is quite different from ML methods 
that learn a hypothesis at once by solving an \gls{erm} problem. These different operation modes 
corresponds to different frequencies of iterating the basic ML cycle depicted in Figure \ref{fig_AlexMLBP}. 
Online learning methods start a new cycle in Figure \ref{fig_AlexMLBP} whenever a new \gls{datapoint} arrives (e.g., 
we have recorded the minimum and maximum temperate of a day that just ended). 


We now present an online learning variant of \gls{linreg} (see Section \ref{sec_lin_reg}) which is suitable for 
time series data with \gls{datapoint}s $\big(\featurevec^{(\timeidx)},\truelabel^{(\timeidx)}\big)$ gathered sequentially (over time). 
In particular, the \gls{datapoint}s $\big(\featurevec^{(\timeidx)},\truelabel^{(\timeidx)}\big)$ become available (are gathered) at a 
discrete time instants $\timeidx=1,2,3\ldots$. 

Let us stack the feature vectors and labels of all \gls{datapoint}s available at time $\timeidx$ 
into feature matrix $\featuremtx^{(\timeidx)}$ and label vector $\labelvec^{(\timeidx)}$, respectively. 
The feature matrix and label vector for the first three time instants are 
\begin{align}
	\label{equ_def_feature_label_three_instants}
	\timeidx&=1: \quad &\featuremtx^{(1)} & \defeq \big( \featurevec^{(1)} \big)^{T}  \mbox{ , } & \labelvec^{(1)} & = \big( \truelabel^{(1)} \big)^{T} \mbox{, } \\ 
	\timeidx&=2: \quad &\featuremtx^{(2)} & \defeq \big( \featurevec^{(1)},\featurevec^{(2)} \big)^{T}  \mbox{ , } & \labelvec^{(2)} & = \big( \truelabel^{(1)},\truelabel^{(2)} \big)^{T} \mbox{, } \\ 
	\timeidx&=3: \quad &\featuremtx^{(3)} & \defeq \big( \featurevec^{(1)},\featurevec^{(2)},\featurevec^{(3)} \big)^{T}  \mbox{ , } & \labelvec^{(3} & = \big( \truelabel^{(1)},\truelabel^{(2)},\truelabel^{(3)} \big)^{T}.
\end{align} 

As detailed in Section \ref{sec_lin_reg}, \gls{linreg} aims at learning the weights $\weights$ of a linear 
map $h(\featurevec)  \defeq \weights^{T} \featurevec$ such that the squared error loss $\big( \truelabel - h(\featurevec) \big)$ 
is as small as possible. This informal goal of \gls{linreg} is made precise by the \gls{erm} problem \eqref{equ_def_cost_MSE} 
which defines the optimal weights via incurring minimum average squared error loss (\gls{emprisk}) on a 
given \gls{trainset} $\dataset$. These optimal \gls{weights} are given by the solutions of \eqref{equ_zero_gradient_lin_reg_normal_condition}. 
When the feature vectors of datapoints in $\dataset$ are linearly independent, we obtain the closed-form 
expression \eqref{equ_close_form_lin_reg} for the optimal \gls{weights}. 

Inserting the feature matrix $\featuremtx^{(\timeidx)}$ and label vector $\labelvec^{(\timeidx)}$ \eqref{equ_def_feature_label_three_instants} into \eqref{equ_close_form_lin_reg}, yields 
\begin{equation} 
	\label{equ_opt_weight_time_t}
	\widehat{\weights}^{(\timeidx)} = \big(  \big(\featuremtx^{(\timeidx)}\big)^{T} \featuremtx^{(\timeidx)} \big)^{-1}  \big(\featuremtx^{(\timeidx)}\big)^{T}  \labelvec^{(\timeidx)}. 
\end{equation}
For each time instant we can evaluate the RHS of \eqref{equ_opt_weight_time_t} to 
obtain the parameter vector $\widehat{\weights}^{(\timeidx)}$ that minimizes the average squared error loss 
over all \gls{datapoint}s gathered up to time $\timeidx$. 
However, computing $\widehat{\weights}^{(\timeidx)}$ via direct evaluation of the RHS in \eqref{equ_opt_weight_time_t} 
for each new time instant $\timeidx$ misses an opportunity for recycling computations done already at earlier time 
instants. 

Let us now show how to (partially) reuse the computations used to evaluate \eqref{equ_opt_weight_time_t} for 
time $\timeidx$ in the evaluation of \eqref{equ_opt_weight_time_t} for the next time instant $\timeidx+1$. 
To this end, we first rewrite the matrix $ \mQ^{(\timeidx)} \defeq \big(\featuremtx^{(\timeidx)}\big)^{T} \featuremtx^{(\timeidx)}$ as 
\begin{equation} 
	\mQ^{(\timeidx)}  = \sum_{\itercntr=1}^{\timeidx} \featurevec^{(\itercntr)} \big(\featurevec^{(\itercntr)} \big)^{T}. 
\end{equation}
Since $\mQ^{(\timeidx\!+\!1)} =  \mQ^{(\timeidx)}  +  \featurevec^{(\timeidx\!+\!1)} \big(\featurevec^{(\timeidx\!+\!1)} \big)^{T}$, 
we can use a well-known identity for matrix inverses (see \cite{Bartlett51,MeyerSIAM73}) to obtain 
\begin{equation} 
	\label{equ_matrix_invers}
	\big( \mQ^{(\timeidx+1)} \big)^{-1} =\big( \mQ^{(\timeidx)} \big)^{-1} + \frac{\big( \mQ^{(\timeidx)} \big)^{-1}  \featurevec^{(\timeidx+1)} \big(\featurevec^{(\timeidx+1)} \big)^{T}\big( \mQ^{(\timeidx)} \big)^{-1}}{1- \big(\featurevec^{(\timeidx+1)} \big)^{T} \big( \mQ^{(\timeidx)} \big)^{-1} \featurevec^{(\timeidx+1)} }. 
\end{equation}

Inserting \eqref{equ_matrix_invers} into \eqref{equ_opt_weight_time_t} yields the following relation between optimal parameter vectors 
at consecutive time instants $\timeidx$ and $\timeidx+1$,  
\begin{equation}
	\label{equ_def_update_recuriveLS}
	\widehat{\weights}^{(\timeidx+1)} =\widehat{\weights}^{(\timeidx)} - \big( \mQ^{(\timeidx+1)} \big)^{-1}\featurevec^{(\timeidx+1)}  \big(\big(\featurevec^{(\timeidx+1)} \big)^{T}\widehat{\weights}^{(\timeidx)}  - \truelabel^{(\timeidx+1)} \big) . 
\end{equation} 
Note that neither evaluating the RHS of \eqref{equ_def_update_recuriveLS} nor evaluating the RHS 
of \eqref{equ_matrix_invers} requires to actually invert a matrix of with more than one entry (we can 
think of a scalar number as $1 \times 1$ matrix). In contrast, evaluating the RHS \eqref{equ_opt_weight_time_t} 
requires to invert the matrix $\mQ^{(\timeidx)} \in \mathbb{R}^{\featurelen \times \featurelen}$. We 
obtain an online algorithm for \gls{linreg} via computing the updates \eqref{equ_def_update_recuriveLS} 
and \eqref{equ_matrix_invers} for each new time instant $\timeidx$. Another online method for \gls{linreg} 
will be discussed at the end of Section \ref{sec_sgd}. 

\section{Weighted ERM}

Consider a ML method that uses some \gls{hypospace} $\hypospace$ and \gls{lossfunc} $\lossfun$ to measure the 
quality predictions obtained from a specific hypothesis when applied to a \gls{datapoint}. A principled approach to 
learn a useful hypothesis is via \gls{erm} \eqref{equ_def_ERM_funs} using a \gls{trainset} 
$$\dataset= \big\{ \big( \featurevec^{(1)}, \truelabel^{(1)} \big),\ldots,\big( \featurevec^{(\samplesize)}, \truelabel^{(\samplesize)} \big) \big\}.$$.

For some applications it might be useful to modify the \gls{erm} principle \eqref{equ_def_ERM_funs} by putting different 
weights on the \gls{datapoint}s. In particular, for each \gls{datapoint} $\big( \featurevec^{(\sampleidx)}, \truelabel^{(\sampleidx)} \big)$ 
we specify a non-negative weight $\sampleweight{\sampleidx} \in \mathbb{R}_{+}$. 
Weighted \gls{erm} is obtained from \gls{erm} \eqref{equ_def_ERM_funs} by replacing the average \gls{loss} 
over the \gls{trainset} with a \index{weighted average loss}weighted average loss, 
\begin{align}
	\label{equ_def_wERM_funs}
	\hat{h} & \in \argmin_{h \in \hypospace} \sum_{\sampleidx=1}^{\samplesize} \sampleweight{\sampleidx}\loss{(\featurevec^{(\sampleidx)},\truelabel^{(\sampleidx)})}{h}.
\end{align}
Note that we obtain \gls{erm} \eqref{equ_def_ERM_funs} as the special case of weighted \gls{erm} \eqref{equ_def_wERM_funs} 
for the weights $\sampleweight{\sampleidx} = 1/\samplesize$. 

We might interpret the weight $\sampleweight{\sampleidx}$ as a measure for the importance or relevance of 
the \gls{datapoint} $\big( \featurevec^{(\sampleidx)}, \truelabel^{(\sampleidx)} \big)$ for the hypothesis $\hat{h}$ 
learnt via \eqref{equ_def_wERM_funs}. The extreme case $\sampleweight{\sampleidx}=0$ means that the \gls{datapoint} 
$\big( \featurevec^{(\sampleidx)}, \truelabel^{(\sampleidx)} \big)$ becomes irrelevant for learning a hypothesis via \eqref{equ_def_wERM_funs}. 
This could be useful if the \gls{datapoint} $\big( \featurevec^{(\sampleidx)}, \truelabel^{(\sampleidx)} \big)$ represents an outlier 
that violates the \gls{iidasspt} which is satisfied by most of the other \gls{datapoint}s. Thus, using suitable weights in \eqref{equ_def_wERM_funs} 
could make the resulting ML method robust against outliers in the \gls{trainset}. Note that we have discussed another 
strategy (via the choice for the \gls{lossfunc}) to achieve robustness against outliers in Section \ref{sec_lad}. 

Another use-case of weighted \gls{erm} \eqref{equ_def_wERM_funs} is for applications where the risk of a hypothesis is defined 
using a \gls{probdist} that is different form the \gls{probdist} of the \gls{datapoint}s in the \gls{trainset}. 
Thus, the \gls{datapoint}s conform to an \gls{iidasspt} with underlying \gls{probdist} $p(\featurevec,\truelabel)$. 
However, we would like to measure the quality of a hypothesis via the expected loss or risk using a different \gls{probdist} $p'(\featurevec,\truelabel)$, 
\begin{equation}
\label{equ_target_p_prime_risk}
\expect_{p'} \big \{ \loss{(\featurevec,\truelabel)}{h} \} = \int   \loss{(\featurevec,\truelabel)}{h}  d p'(\featurevec,\truelabel)  
\end{equation} 
Having a different \gls{probdist} $p'(\featurevec,\truelabel) (\neq p (\featurevec,\truelabel)))$ to define 
the overall quality (risk) of a hypothesis might be beneficial for binary classification problems with imbalanced data. 
Indeed, using the average loss (which approximates the risk under $p(\featurevec,\truelabel)$) might not be a useful 
quality measure if one class is over-represented in the \gls{trainset} (see Section \ref{sec_confustion_matrix}). It can 
be shown that, under mild conditions, the weighted average loss in \eqref{equ_def_wERM_funs} approximates \eqref{equ_target_p_prime_risk} when 
using the weights $\sampleweight{\sampleidx} =  p'\big(\featurevec^{(\sampleidx)},\truelabel^{(\sampleidx)}\big)/ p\big(\featurevec^{(\sampleidx)},\truelabel^{(\sampleidx)}\big)$ \cite[Sec. 11.1.4]{BishopBook}. 

\newpage
\section{Exercise} 
\label{sec_exercise_chap_4} 
\vspace*{-10mm}
\
\begin{exercise}[Uniqueness in Linear Regression]
\label{ex_chap_4_lin_reg} 
What conditions on a \gls{trainset} ensure that there is a unique optimal linear hypothesis map for \gls{linreg}? 
\end{exercise}
\begin{exercise}[Uniqueness in Linear Regression II]
	\label{ex_chap_4_lin_reg_II} 
	\Gls{linreg} uses the squared error loss \eqref{equ_squared_loss} to measure the quality of a linear hypothesis map. 
	 We learn the weights $\weights$ of a linear map via \gls{erm} using a \gls{trainset} $\dataset$ that 
	 consists of $\samplesize=100$ \gls{datapoint}s. Each \gls{datapoint} is characterized by $\featurelen=5$ \gls{features} 
	 and a numeric \gls{label}. Is there a unique choice for the weights $\weights$ that results in a linear 
	 predictor with minimum average squared error loss on the \gls{trainset} $\dataset$)?
\end{exercise}
\begin{exercise}[A Simple Linear Regression Problem.]
	\label{ex_chap_4_bias}
	Consider a \gls{trainset} of $\samplesize$ datapoints, each characterized by a single numeric feature $\feature$ 
	and numeric label $\truelabel$. We learn hypothesis map of the form $h(\feature) = \feature + b$ 
	with some bias $b \in \mathbb{R}$. Can you write down a formula for the optimal $b$, that minimizes the
	average squared error on training data $\big(\feature^{(1)},\truelabel^{(1)} \big),\ldots,\big(\feature^{(\samplesize)},\truelabel^{(\samplesize)}\big)$.   
\end{exercise}
\begin{exercise}[Simple Least Absolute Deviation Problem.]
	\label{ex_chap_4_bias_absoluate_error}
	Consider \gls{datapoint}s characterized by single numeric feature $\feature$ 
	and label $\truelabel$. We learn a hypothesis map of the form $h(\feature) = \feature + b$ 
	with some bias $b \in \mathbb{R}$. Can you write down a formula 
	for the optimal $b$, that minimizes the average absolute error on 
	training data $\big(\feature^{(1)},\truelabel^{(1)} \big),\ldots,\big(\feature^{(\samplesize)},\truelabel^{(\samplesize)}\big)$.   
\end{exercise}
\begin{exercise}[Polynomial Regression.]
	\label{ex_chap_4_poly_reg} 
	Consider polynomial regression for \gls{datapoint}s with a single numeric feature $\feature \in \mathbb{R}$ and numeric 
	label $\truelabel$. Here, polynomial regression is equivalent to \gls{linreg} using the transformed feature 
	vectors $\featurevec = \big(\feature^{0},\feature^{1},\ldots,\feature^{\featuredim-1}\big)^{T}$. Given a dataset 
	$ \dataset= \big(\feature^{(1)},\truelabel^{(1)}\big),\ldots,\big(\feature^{(\samplesize)},\truelabel^{(\samplesize)}\big)$, 
	we construct the feature matrix $\featuremtx =\big(\featurevec^{(1)},\ldots,\featurevec^{(\samplesize)}\big) \in \mathbb{R}^{\samplesize \times \samplesize}$ with its $\sampleidx$th column given by the feature vector $\featurevec^{(\sampleidx)}$. 
	Verify that this feature matrix is a Vandermonde matrix \cite{Gautschi1988}? How is the determinant of the 
	feature matrix related to the features and labels of \gls{datapoint}s in the dataset $\dataset$?
\end{exercise}
\begin{exercise}[Training Error is not Expected Loss.]
	\label{ex_chap_4_empriskapp} 
	Consider a \gls{trainset} that consists of \gls{datapoint}s $\big(\feature^{(\sampleidx)},\truelabel^{(\sampleidx)} \big)$, 
	for $\sampleidx = 1,\ldots,\samplesize=100$, that are obtained as realizations of \gls{iid} \gls{rv}s. The common 
	\gls{probdist} of these \gls{rv}s is defined by a random \gls{datapoint} $(\feature,\truelabel)$. The feature $\feature$ of this 
	random \gls{datapoint} is a standard Gaussian \gls{rv} with zero mean and unit variance. The label of a \gls{datapoint} 
	is modelled as $\truelabel = \feature + e$ with Gaussian  noise $e \sim \mathcal{N}(0,1)$. The feature $\feature$ 
	and noise $e$ are statistically independent. We evaluate the specific hypothesis $h(\feature)=0$ (which outputs $0$ 
	no matter what the feature value $\feature$ is) by
	the \gls{trainerr} $\trainerror = (1/\samplesize) \sum_{\sampleidx=1}^{\samplesize} \big( \truelabel^{(\sampleidx)} - h \big( \feature^{(\sampleidx)} \big) \big)^2$. Note that $\trainerror$ is the average squared error loss \eqref{equ_squared_loss} incurred by hypothesis $h$ on the datapoints $\big(\feature^{(\sampleidx)},\truelabel^{(\sampleidx)} \big)$, for $\sampleidx = 1,\ldots,\samplesize=100$. 
	What is the probability that the \gls{trainerr} $\trainerror$ is at least $20$ \% larger 
	than the expected (squared error) loss $\expect \big\{ \big( \truelabel - h(\feature) \big)^{2} \big \}$? 
	What is the mean (expected value) and variance of the \gls{trainerr} ?
\end{exercise}
\begin{exercise}[Optimization Methods as Filters.]
	\label{ex_chap4_opt_methods_shift_invariatn_filter} 
Let us consider a fictional (idel) optimization method that can be represented as 
a filter $\mathcal{F}$. This filter $\mathcal{F}$ reads in a real-valued objective function $f(\cdot)$, defined for 
all parameter vectors  $\vw\in\mathbb{R}^{\featuredim}$. The output of the filter $\mathcal{F}$ is another 
real-valued function $\hat{f}(\vw)$ that is defined point-wise as 
\begin{equation} 
\hat{f}(\vw) = \begin{cases} 1 & \mbox{ , if } \vw \mbox{ is a local minimum of } f(\cdot) \\ 
	0 & \mbox{, otherwise.} \end{cases}
\end{equation} 
Verify that the filter $\mathcal{F}$ is shift or translation invariant, i.e., $\mathcal{F}$ commutes with a translation 
$f'(\weights) \defeq f(\weights + \weights^{(o)})$ with an arbitrary but fixed (reference) vector $\weights^{(o)} \in \mathbb{R}^{\featuredim}$. 
\end{exercise}
\begin{exercise}[Linear Regression with Sample Weighting.]
Consider a \gls{linreg} method that uses \gls{erm} to learn weights $\widehat{\weights}$ of a linear hypothesis map 
$h(\featurevec) =\weights^{T} \featurevec$. The weights are learnt by minimizing the average squared error 
loss incurred by $h$ on a \gls{trainset} that is constituted by the \gls{datapoint}s $\big( \featurevec^{(\sampleidx)}, \truelabel^{(\sampleidx)} \big)$ for $\sampleidx=1,\ldots, 100$. Someimtes it is useful to assign sample-weights $\sampleweight{\sampleidx}$ to the \gls{datapoint}s and  
learn $\widehat{\weights}$. These sample-weights reflect varying levels of importance or relevance of different \gls{datapoint}s. 
For simplicity we use the sample weights $\sampleweight{\sampleidx} = 2 \alpha \in [0,1]$ for $\sampleidx=1,\ldots,50$ and 
$\sampleweight{\sampleidx} = 2(1 - \alpha)$ for  $\sampleidx=51,\ldots,100$. Can you find a closed-form expression (similar 
to \eqref{equ_close_form_lin_reg}) for the weights $\widehat{\weights}^{(\alpha)}$ that minimize the weighted average 
squared error $f(\weights) \defeq (1/50)\sum_{\sampleidx=1}^{50} \alpha \big( \truelabel^{(\sampleidx)} -  \weights^{T} \featurevec^{(\sampleidx)} \big)^{2}
+  (1/50)\sum_{\sampleidx=51}^{100} (1-\alpha) \big( \truelabel^{(\sampleidx)} -  \weights^{T} \featurevec^{(\sampleidx)} \big)^{2}$ for different $\alpha$?
\end{exercise}

\newpage
\chapter{Gradient-Based Learning}
\label{ch_GD}

This chapter discusses an important family of optimization methods for solving \gls{erm} \eqref{eq_def_ERM_weight} 
with a parametrized \gls{hypospace} (see Chapter \ref{sec_comp_stat_ERM}). The common theme of these methods 
is to construct local approximations of the objective function in \eqref{eq_def_ERM_weight}. These local approximations 
are obtained from the \index{gradient}\gls{gradient}s of the objective function. \Gls{gdmethods} have gained popularity 
recently as an efficient technique for tuning the parameters of \gls{deepnet}s within deep learning methods \cite{Goodfellow-et-al-2016}. 

Section \ref{sec_basic_GD_iteration} discusses \gls{gd} as the most basic form of \gls{gdmethods}. The idea 
of \gls{gd} is to update the weights by locally optimizing a linear approximation of the objective function. This 
update is referred to as a \gls{gd} step and provides the main algorithmic primitive of \gls{gdmethods}. 
One key challenge for a good use of \gls{gdmethods} is the appropriate extend of the local 
approximations. This extent is controlled by a \gls{stepsize} parameter that is used in the basic \gls{gd} step. 
Section \ref{equ_sec_gd_step_size} discusses some approaches for choosing this \gls{stepsize}. 
Section \ref{sec_when_to_stop} discusses a second main challenge in using \gls{gdmethods} 
which is to decide when to stop repeating the \gls{gd} steps. 

Section \ref{sec_GD_linear_regression} and Section \ref{sec_GD_logistic_regression} spell out \gls{gd} for two instances of 
\gls{erm} arising from \gls{linreg} and \gls{logreg}, respectively. 
The beneficial effect of data normalization on the convergence speed of \gls{gdmethods} is briefly 
discussed in Section \ref{sec_data_normalization}. As explained in Section \ref{sec_sgd}, the use of stochastic 
approximations enables \gls{gdmethods} for applications involving massive amounts of data (``big data''). 
Section \ref{sec_adv_gd_methods} develops some intuition for advanced \gls{gdmethods} that 
exploit the information gathered during previous iterations.

\section{The Basic Gradient Step}
\label{sec_basic_GD_iteration}

Let us rewrite \gls{erm} \eqref{eq_def_ERM_weight} as the optimization problem  
\begin{equation} 
	\label{equ_obj_emp_risk_GD}
 \min_{\weights \in \mathbb{R}^{\featuredim}}	f(\weights) \defeq(1/\samplesize) \sum_{\sampleidx=1}^{\samplesize} \loss{(\featurevec^{(\sampleidx)},\truelabel^{(\sampleidx)})}{h^{(\weights)}}. 
\end{equation}
From now on we tacitly assume that each individual \gls{loss} 
\begin{equation} 
	\label{equ_def_componentn_loss_gd}
f_{\sampleidx}(\weights) \defeq \loss{(\featurevec^{(\sampleidx)},\truelabel^{(\sampleidx)})}{h^{(\weights)}}
\end{equation} 
arising in \eqref{equ_obj_emp_risk_GD} represents a \index{differentiable}\gls{differentiable} function of the 
parameter vector $\weights$. Trivially, differentiability of the components \eqref{equ_def_componentn_loss_gd} 
implies differentiability of the overall objective function $f(\weights)$ \eqref{equ_obj_emp_risk_GD}. 

Two important examples of \gls{erm} involving such differentiable \gls{lossfunc}s are \gls{linreg} and \gls{logreg}. 
In contrast, the hinge loss \eqref{equ_hinge_loss} used by the \gls{svm} results in a non-differentiable objective 
function $f(\weights)$ \eqref{equ_obj_emp_risk_GD}. However, it is possible to (significantly) extend the scope 
of \gls{gdmethods} to non-differentiable functions by replacing the concept of a \gls{gradient} with that 
of a \index{subgradient}\gls{subgradient}.


Gradient based methods are iterative. They construct a sequence of parameter vectors $\weights^{(0)} \rightarrow \weights^{(1)} \dots$ 
that hopefully converge to a minimizer $\overline{\weights}$ of $f(\weights)$, 
\begin{equation}
	\label{equ_def_opt_weight}
f(\overline{\weights}) = \bar{f} \defeq \min_{\weights \in \mathbb{R}^{\featuredim}} f(\weights).  
\end{equation} 
Note that there might be several different optimal parameter vectors $\overline{\weights}$ that 
satisfy the optimality condition \eqref{equ_def_opt_weight}.  We want the sequence generated 
by a gradient based method to converge towards any of them. The vectors $\weights^{(\itercntr)}$ are 
(hopefully) increasingly, with increasing iteration $\itercntr$, more accurate approximation for a 
minimizer $\overline{\weights}$ of \eqref{equ_def_opt_weight}.


Since the objective function $f(\weights)$ is differentiable, we can approximate it locally around the  
vector $\weights^{(\itercntr)}$ using a tangent hyperplane that passes through the point $\big(\weights^{(\itercntr)},f\big(\weights^{(\itercntr)}\big) \big) \in \mathbb{R}^{\featuredim+1}$. The normal vector of this hyperplane is given by $\mathbf{n} = (\nabla f\big(\weights^{(\itercntr)}\big) ,-1)$ 
(see Figure \ref{fig_smooth_function}). The first component of the normal vector is the gradient $\nabla f(\weights)$ of the objective 
function $f(\weights)$ evaluated at the point $\weights^{(\itercntr)}$. Our main use of the gradient $\nabla f\big(\weights^{(\itercntr)}\big)$ will be 
to construct a linear approximation \cite{RudinBookPrinciplesMatheAnalysis}
\begin{equation} 
	\label{equ_linear_approx_diff}
	f(\weights) \approx f\big(\weights^{(\itercntr)}\big) + \big(\weights-\weights^{(\itercntr)} \big)^{T} \nabla f\big(\weights^{(\itercntr)}\big)  \mbox{ for }\weights \mbox{ sufficiently close to } \weights^{(\itercntr)}.
\end{equation}  

Requiring the objective function $f(\weights)$ in \eqref{equ_linear_approx_diff} to be differentiable is the same 
as requiring the validity of the local linear approximation \eqref{equ_linear_approx_diff} at every possible vector $\weights^{(\itercntr)}$. 
It turns out that differentiability alone is not very helpful for the design and analysis of gradient based methods. 

Gradient based methods are most useful for finding the minimum of 
differentiable functions $f(\weights)$ that are also \gls{smooth}. 
Informally, a differentiable function $f(\weights)$ is smooth if the gradient $\nabla f(\weights)$ does 
not change too rapidly as a function of the argument $\weights$. A quantitative version  of the smoothness 
concept refers to a function as $\beta$-smooth if its gradient is Lipschitz continuous with Lipschitz 
constant $\beta> 0 $ \cite[Sec. 3.2]{CvxBubeck2015}, 
\begin{equation}
	\label{equ_def_beta_smooth}
 \| \nabla f(\weights) - \nabla f(\weights') \| \leq \beta \| \weights - \weights' \|.
\end{equation} 
Note that if a function $f(\weights)$ is $\beta$ smooth, it is also $\beta'$ smooth for any $\beta' > \beta$. 
The smallest $\beta$ such that \eqref{equ_def_beta_smooth} is satisfied depends on the features 
and labels of \gls{datapoint}s used in \eqref{equ_obj_emp_risk_GD} as well as on the choice for the \gls{lossfunc}. 

\begin{figure}[htbp]
	\begin{center}
		\begin{tikzpicture}[scale=0.9]
			\node [right] at (-4.1,1.7) {$f(\weights)$} ;
			\draw[ultra thick, domain=-4.1:4.1] plot (\x,  {(1/4)*\x*\x});
			\draw[dashed, thick, domain=1:3.6] plot (\x,  {\x - 1}) node[right] {$ f\big(\weights^{(\itercntr)}\big)\!+\!\big(\weights\!-\!\weights^{(\itercntr)}\big)^{T} \nabla f\big(\weights^{(\itercntr)}\big)$};
			\draw [fill] (2,1) circle [radius=0.1] node[right] {$f\big(\weights^{(\itercntr)}\big)$};
			\draw[thick,->] (3,2) -- (3.5,1.5) node[right] {$\mathbf{n}$};
		\end{tikzpicture}
	\end{center}
	\caption{A differentiable function $f(\weights)$ can be approximated locally around a point $\weights^{(\itercntr)}$ using a 
		hyperplane whose normal vector $\mathbf{n} = (\nabla f\big(\weights^{(\itercntr)}\big),-1)$ is determined by the 
		gradient $\nabla f\big(\weights^{(\itercntr)}\big)$ \cite{RudinBookPrinciplesMatheAnalysis}.}
	\label{fig_smooth_function}
\end{figure}

Consider a current guess or approximation $\weights^{(\itercntr)}$ for the optimal parameter vector $\overline{\weights}$ \eqref{equ_def_opt_weight}. 
We would like to find a new (better) parameter vector $\weights^{(\itercntr+1)}$ that has smaller objective 
value $f(\weights^{(\itercntr+1)}) < f\big(\weights^{(\itercntr)}\big)$ than the current guess $\weights^{(\itercntr)}$. 
The approximation \eqref{equ_linear_approx_diff} suggests to choose the next guess $\weights = \weights^{(\itercntr+1)}$ such that $\big(\weights^{(\itercntr+1)}-\weights^{(\itercntr)} \big)^{T} \nabla f\big(\weights^{(\itercntr)}\big)$ is negative. We can 
achieve this by the \gls{gd} step
\begin{equation} 
	\label{equ_def_GD_step}
	\weights^{(\itercntr\!+\!1)} = \weights^{(\itercntr)} - \lrate \nabla f(\weights^{(\itercntr)})
\end{equation} 
with a sufficiently small \gls{stepsize} $\lrate>0$. Figure \ref{fig_basic_GD_step} illustrates the \gls{gd} step \eqref{equ_def_GD_step} 
which is the elementary computation of gradient based methods. 

The \gls{stepsize} $\lrate$ in \eqref{equ_def_GD_step} must be sufficiently small to ensure the 
validity of the linear approximation \eqref{equ_linear_approx_diff}. In the context of ML, 
the \gls{gd} \gls{stepsize} parameter $\lrate$ is also referred to as \gls{learnrate}. Indeed, the 
\gls{stepsize} $\lrate$ determines the amount of progress during a \gls{gd} step towards learning 
the optimal parameter vector $\overline{\weights}$. 

We need to emphasize that the interpretation of the \gls{stepsize} $\lrate$ as a \gls{learnrate} is only 
useful when the \gls{stepsize} is sufficiently small. Indeed, when increasing the \gls{stepsize} $\lrate$ in \eqref{equ_def_GD_step} 
beyond a critical value (that depends on the properties of the objective function $f(\weights)$), the iterates \eqref{equ_def_GD_step} 
move away from the optimal parameter vector $\overline{\weights}$. Nevertheless, from now on we will consequently 
use the term \gls{learnrate} for $\lrate$. 

The idea of \gls{gdmethods} is to repeat the \gls{gd} step \eqref{equ_def_GD_step} for a 
sufficient number of iterations (repetitions) to obtain a sufficiently accurate approximation of the 
optimal parameter vector $\overline{\weights}$ \eqref{equ_def_opt_weight}. It turns out that this 
is feasible for a sufficiently small \gls{learnrate} and if the objective function is smooth and convex. 
Section \ref{equ_sec_gd_step_size} discusses precise conditions on the \gls{learnrate} such that 
the iterates produced by the \gls{gd} step converge to the optimum parameter vector, i.e., $\lim_{\itercntr \rightarrow \infty} f(\weights^{(\itercntr)}) = f\big(\overline{\weights}\big)$. 

\begin{figure}
	\begin{center}
		\begin{tikzpicture}[scale=0.9]
			\draw[loosely dotted] (-4,0) grid (4,4);
			\draw[blue, ultra thick, domain=-4.1:4.1] plot (\x,  {(1/4)*\x*\x});
			\draw[red, thick, domain=3:4.1] plot (\x,  {2*\x - 4});
			\draw[->] (4,4) -- node[right] {$\nabla f(\weights^{(\itercntr)})$} (4,2);
			\draw[->] (4,4) -- node[above] {$-\lrate \nabla f(\weights^{(\itercntr)})$} (2,4);
			\draw[->] (4,2) -- node[below] {$1$} (3,2) ;
			\draw[->] (-4.25,0) -- (4.25,0) node[right] {$\weights$};
			\draw[->] (0,-0.25) -- (0,4.25) node[above] {$f(\weights)$};
			\draw[shift={(4,0)}] (0pt,2pt) -- (0pt,-2pt) node[below] {$\weights^{(\itercntr)}$};
			\draw[shift={(2,0)}] (0pt,2pt) -- (0pt,-2pt) node[below] {$\weights^{(\itercntr\!+\!1)}$};
			\foreach \y/\ytext in {1/1, 2/2, 3/3, 4/4}
			\draw[shift={(0,\y)}] (2pt,0pt) -- (-2pt,0pt) node[left] {$\ytext$};  
		\end{tikzpicture}
	\end{center}
	\caption{A \gls{gd} step \eqref{equ_def_GD_step} updates a current guess or approximation $\weights^{(\itercntr)}$ 
		for the optimum parameter vector $\overline{\weights}$ \eqref{equ_def_opt_weight} by adding the correction 
		term $-\lrate \nabla f(\weights^{(\itercntr)})$. The updated parameter vector $\weights^{(\itercntr+1)}$ is (typically) an improved 
		approximation of the minimizer $\overline{\weights}$.}
	\label{fig_basic_GD_step}
\end{figure}

To implement the \gls{gd} step \eqref{equ_def_GD_step} we need to choose a useful \gls{learnrate} $\lrate$. Moreover, 
executing the \gls{gd} step \eqref{equ_def_GD_step} requires to compute the gradient $\nabla f(\weights^{(\itercntr)})$. 
Both tasks can be computationally challenging as discussed in Section \ref{equ_sec_gd_step_size} and \ref{sec_sgd}. 
For the objective function \eqref{equ_obj_emp_risk_GD} obtained in \gls{linreg} and \gls{logreg}, we can obtain closed-form 
expressions for the gradient $\nabla f(\weights)$ (see Section \ref{sec_GD_linear_regression} and \ref{sec_GD_logistic_regression}). 

In general, we do not have closed-form expressions for the gradient of the objective function 
\eqref{equ_obj_emp_risk_GD} arising from a non-linear \gls{hypospace}. One example for such 
a \gls{hypospace} is  obtained from a \gls{ann}, which is used by deep learning methods (see Section \ref{sec_deep_learning}). 
The empirical success of deep learning methods might be partially attributed to the availability of 
an efficient algorithm for computing the gradient $\nabla f(\weights^{(\itercntr)})$. This algorithm 
is known as \index{back-propagation}back-propagation \cite{Goodfellow-et-al-2016}. 

\section{Choosing the Learning Rate} 
\label{equ_sec_gd_step_size}

\begin{figure}[hbtp]
	\begin{center}
		\begin{minipage}{0.45\columnwidth}
			\begin{tikzpicture}[xscale=0.4,yscale=0.6]
				\draw[blue, ultra thick, domain=-4.1:4.1] plot (\x,  {(1/4)*\x*\x});
				\draw[] (4,4) circle [radius=0.1];
				\node [right] at (4,4) {$f(\weights^{(\itercntr)})$};
				\draw[] (3.8,3.61) circle [radius=0.1];
				\node [left] at (3.8,3.61) {$f(\weights^{(\itercntr\!+\!1)})$};
				\draw[] (3.65,3.33) circle [radius=0.1];
				\node [right] at (3.65,3.33) {$f(\weights^{(\itercntr\!+\!2)})$};
				\node [below] at (0,-0.2) {(a)};
			\end{tikzpicture}
		\end{minipage}
		\begin{minipage}{0.45\columnwidth}
			\begin{tikzpicture}[xscale=0.4,yscale=0.6]
				\draw[blue, ultra thick, domain=-4.1:4.1] plot (\x,  {(1/4)*\x*\x});
				\draw[] (1,0.25) circle [radius=0.1] node [right] (A) {$f(\weights^{(\itercntr)})$} ;
				\draw[] (-2,1) circle [radius=0.1] node [left] (B) {$f(\weights^{(\itercntr\!+\!1)})$} ;
				\draw[] (3,2.25) circle [radius=0.1]  node  [right] (C) {$f(\weights^{(\itercntr\!+\!2)})$} ;
				\draw[->,dashed] (-2,1) -- (3,2.25) node [midway,above] {\eqref{equ_def_GD_step}};
				\draw[->,dashed]  (1,0.25) -- (-2,1) node [midway,above] {\eqref{equ_def_GD_step}};
				\node [below] at (0,-0.2) {(b)};
			\end{tikzpicture}
		\end{minipage}
	\end{center}
	\caption{Effect of choosing bad values for the \gls{learnrate} $\lrate$ in the \gls{gd} step\eqref{equ_def_GD_step}. 
		(a) If the \gls{learnrate} $\lrate$ in the \gls{gd} step \eqref{equ_def_GD_step} is chosen too small, the 
		iterations make very little progress towards the optimum or even fail to reach the optimum at all. 
		(b) If the \gls{learnrate} $\lrate$ is chosen too large, the iterates $\weights^{(\itercntr)}$ 
		might not converge at all (it might happen that $f(\weights^{(\itercntr\!+\!1)}) > f(\weights^{(\itercntr)})$!). }
	\label{fig_small_large_lrate}
\end{figure}

The choice of the \gls{learnrate} $\lrate$ in the \gls{gd} step \eqref{equ_def_GD_step} has a significant impact on 
the performance of Algorithm \ref{alg:gd_linreg}. If we choose the \gls{learnrate} $\lrate$ too large, the \gls{gd} 
steps \eqref{equ_def_GD_step} diverge (see Figure \ref{fig_small_large_lrate}-(b)) and, in turn, Algorithm \ref{alg:gd_linreg} 
fails to deliver a satisfactory approximation of the optimal weights $\overline{\weights}$. 

If we choose the \gls{learnrate} $\lrate$ too small (see Figure \ref{fig_small_large_lrate}-(a)), the 
updates \eqref{equ_def_GD_step} make only very little progress towards approximating the optimal 
parameter vector $\overline{\weights}$. In applications that require real-time processing of data streams, it 
might be possible to repeat the \gls{gd} steps only for a moderate number. Thus If the \gls{learnrate} is chosen 
too small, Algorithm \ref{alg:gd_linreg} will fail to deliver a good approximation within an acceptable 
number of iterations (runtime of Algorithm \ref{alg:gd_linreg}). 

Finding a (nearly) optimal choice for the \gls{learnrate} $\lrate$ of \gls{gd} can be a challenging task. 
Many sophisticated approaches for tuning the \gls{learnrate} of \gls{gdmethods} have been 
proposed \cite[Chapter 8]{Goodfellow-et-al-2016}. A detailed discussion of these approaches is 
beyond the scope of this book. We will instead discuss two sufficient conditions on the \gls{learnrate} 
which guarantee the convergence of the \gls{gd} iterations to the optimum of a smooth and convex objective function 
\eqref{equ_obj_emp_risk_GD}. 

The first condition applies to an objective function that is $\beta$-smooth (see \eqref{equ_def_beta_smooth}) with 
known constant $\beta$ (not necessarily the smallest constant such that \eqref{equ_def_beta_smooth} holds). Then, 
the iterates $\weights^{(\itercntr)}$ generated by the \gls{gd} step \eqref{equ_def_GD_step} with a \gls{learnrate} 
\begin{equation}
	\label{equ_suff_cond_lrate_beta}
	 \lrate < 2/\beta,
\end{equation}
satisfy \cite[Thm. 2.1.13]{nesterov04}
\begin{equation} 
\label{equ_convergence_rate_inverse_k-GD}
f \big( \weights^{(\itercntr)}  \big) - \bar{f} \leq \frac{2(f \big( \weights^{(0)}  \big) - \bar{f}) \sqeuclnorm{\weights^{(0)} -\overline{\weights}}}{2\sqeuclnorm{ \weights^{(0)} - \overline{\weights}}+\itercntr(f \big( \weights^{(0)}  \big) -\bar{f}) \lrate(2-\beta\lrate)}.
	\end{equation} 
The bound \eqref{equ_convergence_rate_inverse_k-GD} not only tells us that \gls{gd} iterates converge to an 
optimal parameter vector but also characterize the convergence speed or rate. The sub-optimality 
$f \big( \weights^{(\itercntr)}  \big) - \min_{\weights} f(\weights)$ in terms of objective function value 
decreases inversely (like ``$1/\itercntr$'') with the number $\itercntr$ of \gls{gd} steps \eqref{equ_def_GD_step}. 
Convergence bounds like \eqref{equ_convergence_rate_inverse_k-GD} can be used to specify a 
stopping criterion, i.e., to determine the number of \gls{gd} steps to be computed (see Section \ref{sec_when_to_stop}). 

The condition \eqref{equ_suff_cond_lrate_beta} and the bound \eqref{equ_convergence_rate_inverse_k-GD} is only useful 
if we can verify $\beta$ smoothness \eqref{equ_def_beta_smooth} assumption for a reasonable constant $\beta$. 
Verifying \eqref{equ_convergence_rate_inverse_k-GD} only for a very large $\beta$ results in the 
bound \eqref{equ_convergence_rate_inverse_k-GD} being too loose (pessimistic). When we use a 
loose bound \eqref{equ_convergence_rate_inverse_k-GD} to determine the number of \gls{gd} steps, 
we might compute an unnecessary large number of  \gls{gd} steps \eqref{equ_def_GD_step}. 

One elegant approach to verify if a differentiable function $f(\weights)$ is $\beta$ smooth \eqref{equ_def_beta_smooth} 
is via the \index{Hessian}Hessian matrix $ \nabla^{2} f(\weights) \in \mathbb{R}^{\featuredim \times \featuredim}$ if it exists. 
The entries of this Hessian matrix are the second-order partial derivatives $\frac{\partial f(\weights)}{\partial \weight_{\featureidx} \partial \weight_{\featureidx'}}$ 
of the function $f(\weights)$. 

Consider an objective function $f(\weights)$ \eqref{equ_obj_emp_risk_GD} that is convex and twice-differentiable 
with \gls{psd} Hessian $\nabla^{2} f(\weights)$. If the maximum \gls{eigenvalue} $\eigval{\rm max} \big( \nabla^{2} f(\weights) \big)$
of the Hessian is upper bounded uniformly (for all $\weights$) by the constant $\beta>0$, then $f(\weights)$ is $\beta$ \gls{smooth} \eqref{equ_def_beta_smooth} \cite{CvxBubeck2015}. This implies, in turn via \eqref{equ_suff_cond_lrate_beta}, the sufficient condition  
\begin{equation} 
	\label{equ_GD_conv_guarantee}
	\lrate \leq \frac{2}{\eigval{\rm max} \big( \nabla^{2} f(\weights) \big) }\mbox{ for all } \vw \in \mathbb{R}^{\featuredim}
\end{equation} 
for the \gls{gd} \gls{learnrate} such that the \gls{gd} steps converge to the minimum of the objective function $f(\weights)$. 

It is important to note that the condition \eqref{equ_GD_conv_guarantee} guarantees convergence of the \gls{gd} steps 
for any possible initialization $\weights^{(0)}$. Note that the usefulness of the condition \eqref{equ_GD_conv_guarantee} 
depends on the difficulty of computing the Hessian matrix $\nabla^{2} f(\weights)$. Section \ref{sec_GD_linear_regression} 
and Section \ref{sec_GD_logistic_regression} will present closed-form expressions for the Hessian of the objective function \eqref{equ_obj_emp_risk_GD} obtained for \gls{linreg} and \gls{logreg}. These closed-form expressions involve the feature vectors and labels of the \gls{datapoint}s in 
the \gls{trainset} $\dataset = \big\{ \big(\featurevec^{(1)},\truelabel^{(1)} \big),\ldots,\big(\featurevec^{(\samplesize)},\truelabel^{(\samplesize)} \big) \big\}$ used in \eqref{equ_obj_emp_risk_GD}.

While it might be computationally challenging to determine the maximum (in absolute value) \gls{eigenvalue} 
$\eigval{\rm max} \big( \nabla^{2} f(\weights) \big)$ for arbitrary $\weights$, it might still be feasible to find an 
upper bound $U$ for it. If we know such an upper bound $U \geq \eigval{\rm max} \big( \nabla^{2} f(\weights) \big)$ 
(valid for all $\weights \in \mathbb{R}^{\featuredim}$), the \gls{learnrate} $\lrate =1/U$ still ensures convergence 
of the \gls{gd} steps \eqref{equ_def_GD_step}.

Up to know we have assumed a fixed (constant) \gls{learnrate} $\lrate$ that is used for each repetition of the \gls{gd} 
steps \eqref{equ_def_GD_step}. However, it might be useful to vary or adjust the \gls{learnrate} as the \gls{gd} 
steps \eqref{equ_def_GD_step} proceed. Thus, we might use a different \gls{learnrate} $\lrate_{\itercntr}$ for 
each iteration $\itercntr$ of \eqref{equ_def_GD_step}. Such a varying \gls{learnrate} is useful for a variant of \gls{gd} 
that uses stochastic approximation (see Section \ref{sec_sgd}). However, we might use a varying \gls{learnrate} also to 
avoid the burden of verifying $\beta$ smoothness  \eqref{equ_def_beta_smooth} with a tight (small) $\beta$. 
The \gls{gd} steps \eqref{equ_def_GD_step} with the \gls{learnrate}  $\lrate_{\itercntr} \defeq 1/\itercntr$ converge 
to the optimal parameter vector $\overline{\weights}$ as long as we can ensure a bounded gradient 
$\| \nabla f(\weights) \| \leq U$ for a sufficiently large neighbourhood of $\overline{\weights}$ \cite{nesterov04}. 

\section{When To Stop?} 
\label{sec_when_to_stop}

One main challenge in the successful application of \gls{gd} is to decide when to stop iterating 
(or repeating) the \gls{gd} step \eqref{equ_def_GD_step}. Maybe the most simple approach is to monitor 
the decrease in the objective function $f(\weights^{(\itercntr)})$ and to stop if the decrease 
$f(\weights^{(\itercntr-1)})-f(\weights^{(\itercntr)})$ falls below a threshold. However, the ultimate 
goal of a ML method is not to minimize the objective function $f(\weights)$ in \eqref{equ_obj_emp_risk_GD}. 
Indeed, the objective function represents the average loss of a hypothesis $h^{(\weights)}$ 
incurred on a \gls{trainset}. However, the ultimate goal of a ML method is to learn a parameter vector 
$\weights$ such that the resulting hypothesis accurately predicts any \gls{datapoint}, including those 
outside the \gls{trainset}. 

We will see in Chapter \ref{ch_validation_selection} how to use validation 
techniques to probe a hypothesis outside the \gls{trainset}. These validation techniques provide a 
validation error $\tilde{f}(\weights)$ that estimates the average loss of a hypothesis with parameter vector $\weights$. 
Early stopping techniques monitor the validation error $\tilde{f}(\weights^{(\itercntr)})$ as the \gls{gd} iterations $\itercntr$ 
proceed to decide when to stop iterating.
 
Another possible stopping criterion is to use a fixed number of iterations or \gls{gd} steps. This fixed number of iterations 
can be chosen based on convergence bounds such as \eqref{equ_convergence_rate_inverse_k-GD} in order to 
guarantee a prescribed sub-optimality of the final iterate $\weights^{(\itercntr)}$. A slightly more convenient 
convergence bound can be obtained from \eqref{equ_convergence_rate_inverse_k-GD} when using the 
the \gls{learnrate} $\lrate = 1/\beta$ in the \gls{gd} step \eqref{equ_def_GD_step} \cite{CvxBubeck2015}, 
\begin{equation} 
f \big( \weights^{(\itercntr)}  \big) -\bar{f} \leq \frac{2\beta \sqeuclnorm{\weights^{(0)} -\overline{\weights}}}{\itercntr} \mbox{ for } \itercntr=1,2,\ldots.
\end{equation} 



\section{GD for Linear Regression} 
\label{sec_GD_linear_regression}

We now present a gradient based method for learning the parameter vector for a linear hypothesis (see \eqref{equ_lin_hypospace}) 
\begin{equation} 
	\label{equ_def_lin_pred_GD}
	h^{(\weights)}(\featurevec) = \weights^{T} \featurevec.
\end{equation}
The \gls{erm} principle tells us to choose the parameter vector $\weights$ in \eqref{equ_def_lin_pred_GD} 
by minimizing the average squared error loss \eqref{equ_squared_loss}
\begin{equation} 
	\label{equ_def_cost_linreg}
	\emperror(h^{(\weights)}| \dataset) \stackrel{\eqref{eq_def_ERM_weight}}{=} (1/\samplesize) \sum_{\sampleidx=1}^{\samplesize} (\truelabel^{(\sampleidx)} - \weights^{T} \featurevec^{(\sampleidx)})^{2}. 
\end{equation}
The average squared error loss \eqref{equ_def_cost_linreg} is computed by applying the predictor $h^{(\weights)}(\featurevec)$ 
to labeled \gls{datapoint}s in a \gls{trainset} $\dataset=\{ (\featurevec^{(\sampleidx)}, \truelabel^{(\sampleidx)}) \}_{\sampleidx=1}^{\samplesize}$. 
An optimal parameter vector $\overline{\weights}$ for \eqref{equ_def_lin_pred_GD} is obtained as 
\begin{equation} 
	\label{equ_smooth_problem_linreg}
	\overline{\weights} = \argmin_{\weights \in \mathbb{R}^{\featuredim}} f(\weights) \mbox{ with } f(\weights) = (1/\samplesize) \sum_{\sampleidx=1}^{\samplesize} \big(\truelabel^{(\sampleidx)} - \weights^{T} \featurevec^{(\sampleidx)}\big)^{2}. 
\end{equation} 

The objective function $f(\weights)$ in \eqref{equ_smooth_problem_linreg} is convex and smooth.  
We can therefore use \gls{gd} \eqref{equ_def_GD_step} to  solve \eqref{equ_smooth_problem_linreg} 
iteratively, i.e., by constructing a sequence of parameter vectors that converge to an optimal parameter vector $\overline{\weights}$. 
To implement \gls{gd}, we need to compute the gradient $\nabla f(\weights)$. 

The gradient of the objective function in \eqref{equ_smooth_problem_linreg} is given by 
\begin{equation}
	\label{equ_gradient_linear_regression}
	\nabla f(\weights) = -(2/\samplesize) \sum_{\sampleidx=1}^{\samplesize} \big(\truelabel^{(\sampleidx)} - \weights^{T} \featurevec^{(\sampleidx)} \big) \featurevec^{(\sampleidx)}.
\end{equation} 
By inserting \eqref{equ_gradient_linear_regression} into the basic \gls{gd} iteration \eqref{equ_def_GD_step}, we obtain Algorithm \ref{alg:gd_linreg}. 
\begin{algorithm}[htbp]
	\caption{\Gls{linreg} via \gls{gd}}\label{alg:gd_linreg}
	\begin{algorithmic}[1]
		\renewcommand{\algorithmicrequire}{\textbf{Input:}}
		\renewcommand{\algorithmicensure}{\textbf{Output:}}
		\Require dataset $\dataset=\{ (\featurevec^{(\sampleidx)}, \truelabel^{(\sampleidx)}) \}_{\sampleidx=1}^{\samplesize}$ ; \gls{learnrate} $\lrate >0$. 
		\Statex\hspace{-6mm}{\bf Initialize:} set $\weights^{(0)}\!\defeq\!\mathbf{0}$; set iteration counter $\itercntr\!\defeq\!0$   
		\Repeat 
		\State $\itercntr \defeq \itercntr +1$    (increase iteration counter) 
		\State  $\weights^{(\itercntr)} \defeq \weights^{(\itercntr\!-\!1)} + \lrate (2/\samplesize) \sum_{\sampleidx=1}^{\samplesize} \big(\truelabel^{(\sampleidx)} - \big(\weights^{(\itercntr\!-\!1)}\big)^{T} \featurevec^{(\sampleidx)}\big) \featurevec^{(\sampleidx)}$  (do a GD step \eqref{equ_def_GD_step})
		\Until stopping criterion met 
		\Ensure $\weights^{(\itercntr)}$ (which approximates $\overline{\weights}$ in \eqref{equ_smooth_problem_linreg})
	\end{algorithmic}
\end{algorithm}

Let us have a closer look on the update in step $3$ of Algorithm \ref{alg:gd_linreg}, which is 
\begin{equation}
	\label{equ_update_gd_linreg}
	\weights^{(\itercntr)} \defeq \weights^{(\itercntr\!-\!1)} + \lrate (2/\samplesize) \sum_{\sampleidx=1}^{\samplesize} \big(\truelabel^{(\sampleidx)} - \big(\weights^{(\itercntr\!-\!1)})^{T} \featurevec^{(\sampleidx)} \big) \featurevec^{(\sampleidx)}. 
\end{equation}
The update \eqref{equ_update_gd_linreg} has an appealing form as it amounts to correcting the previous 
guess (or approximation) $\weights^{(\itercntr\!-\!1)}$ for the optimal parameter vector $\overline{\weights}$ 
by the correction term 
\begin{equation}
	\label{equ_corr_term_linreg}
	(2\lrate/\samplesize) \sum_{\sampleidx=1}^{\samplesize} \underbrace{(\truelabel^{(\sampleidx)} - \big(\weights^{(\itercntr\!-\!1)})^{T} \featurevec^{(\sampleidx)})}_{e^{(\sampleidx)}} \featurevec^{(\sampleidx)}. 
\end{equation}
The correction term \eqref{equ_corr_term_linreg} is a weighted average of the feature vectors $\featurevec^{(\sampleidx)}$ using weights $(2\lrate/\samplesize) \cdot e^{(\sampleidx)}$. 
These weights consist of the global factor $(2\lrate/\samplesize)$ (that applies equally to all 
feature vectors $\featurevec^{(\sampleidx)}$) and a sample-specific factor $e^{(\sampleidx)} = (\truelabel^{(\sampleidx)} - \big(\weights^{(\itercntr\!-\!1)})^{T} \featurevec^{(\sampleidx)})$, which is the prediction (approximation) error obtained by the linear predictor $h^{(\weights^{(\itercntr\!-\!1)})}(\featurevec^{(\sampleidx)}) =   \big(\weights^{(\itercntr\!-\!1)})^{T} \featurevec^{(\sampleidx)}$ when predicting the label $\truelabel^{(\sampleidx)}$ from the features $\featurevec^{(\sampleidx)}$. 

We can interpret the \gls{gd} step \eqref{equ_update_gd_linreg} as an instance of ``learning by trial 
and error''. Indeed, the \gls{gd} step amounts to first ``trying out'' (trial) the predictor $h(\featurevec^{(\sampleidx)}) = \big(\weights^{(\itercntr\!-\!1)})^{T}\featurevec^{(\sampleidx)}$.  The predicted values are then used to correct the weight 
vector $\weights^{(\itercntr\!-\!1)}$ according to the error $e^{(\sampleidx)} = \truelabel^{(\sampleidx)} - \big(\weights^{(\itercntr\!-\!1)})^{T} \featurevec^{(\sampleidx)}$. 

The choice of the \gls{learnrate} $\lrate$ used for Algorithm \ref{alg:gd_linreg} can be based on the condition \eqref{equ_GD_conv_guarantee} 
with the Hessian $\nabla^{2} f(\weights)$ of the objective function $f(\weights)$ underlying \gls{linreg} (see \eqref{equ_smooth_problem_linreg}). 
This Hessian is given explicitly as  
\begin{equation}
	\label{equ_hessian_linreg}
	\nabla^{2} f(\weights) = (1/\samplesize) \featuremtx^{T} \featuremtx, 
\end{equation}
with the feature matrix $\featuremtx=\big(\featurevec^{(1)},\ldots,\featurevec^{(\samplesize)}\big)^{T} \in \mathbb{R}^{\samplesize \times \featuredim}$ (see \eqref{equ_def_vec_matrix}). 
Note that the Hessian \eqref{equ_hessian_linreg} does not depend on the parameter vector $\weights$. 

Comparing \eqref{equ_hessian_linreg} with \eqref{equ_GD_conv_guarantee}, one particular strategy for choosing 
the \gls{learnrate} in Algorithm \ref{alg:gd_linreg} is to (i) compute the matrix product $ \featuremtx^{T} \featuremtx$, (ii) compute the 
maximum \gls{eigenvalue} $\eigval{\rm max}\big(  (1/\samplesize)  \featuremtx^{T} \featuremtx \big)$ of this product and (iii) set 
the \gls{learnrate} to $\lrate =1/\eigval{\rm max} \big(  (1/\samplesize)  \featuremtx^{T} \featuremtx \big)$. 

While it might be challenging to compute the maximum \gls{eigenvalue} $\eigval{\rm max} \big(  (1/\samplesize)  \featuremtx^{T} \featuremtx \big)$, 
it might be easier to find an upper bound $U$ for it.\footnote{The problem of computing a full \gls{eigenvalue} decomposition of 
$\featuremtx^{T} \featuremtx$ has essentially the same complexity as \gls{erm} via directly solving \eqref{equ_zero_gradient_lin_reg}, 
which we want to avoid by using the ``cheaper'' \gls{gd} Algorithm \ref{alg:gd_linreg}.} Given such 
an upper bound $U \geq \eigval{\rm max} \big( (1/\samplesize)  \featuremtx^{T} \featuremtx \big)$, 
the \gls{learnrate} $\lrate =1/U$ still ensures convergence of the \gls{gd} steps. Consider a dataset $\{(\featurevec^{(\sampleidx)},\truelabel^{(\sampleidx)})\}_{\sampleidx=1}^{\samplesize}$ with 
normalized features, i.e., $\| \featurevec^{(\sampleidx)}\| = 1$ for all $\sampleidx =1,\ldots,\samplesize$. 
This implies, in turn, the upper bound $U= 1$, i.e., $ 1 \geq  \lambda_{\rm max} \big(  (1/\samplesize)  \featuremtx^{T} \featuremtx \big)$. 
We can then ensure convergence of the iterates $\weights^{(\itercntr)}$ (see \eqref{equ_update_gd_linreg}) by 
choosing the \gls{learnrate} $\lrate =1$.

{\bf Time-Data Tradeoffs.} 
The number of \gls{gd} steps required by Algorithm \ref{alg:gd_linreg} to ensure a prescribed 
sub-optimality depends crucially on the \gls{condnr} of $\featuremtx^{T} \featuremtx$. What 
can we say about the \gls{condnr}? In general, we have not control over this quantity as the matrix $\featuremtx$ consists 
of the feature vectors of arbitrary \gls{datapoint}s. However, it is often useful to model the feature vectors 
as realizations of \gls{iid} random vectors. It is then possible to bound the probability of the feature matrix 
having a sufficiently small \gls{condnr}. These bounds can then be used to choose the step-size such that 
convergence is guaranteed with sufficiently large probability. The usefulness of these bounds typically 
depends on the ratio $\featurelen/\samplesize$. For increasing sample-size, these bounds allow 
to use larger step-sizes and, in turn, result in faster convergence of GD algorithm. Thus, we obtain 
a trade-off between the runtime of Algorithm \ref{alg:gd_linreg} and the number of \gls{datapoint}s that 
we feed into it \cite{Oymak2018}.


\section{GD for Logistic Regression}
\label{sec_GD_logistic_regression}

\Gls{logreg} learns a linear hypothesis $h^{(\weights)}$ that is used to classify \gls{datapoint}s by predicting their 
binary label. The quality of such a \gls{linclass} is measured by the \gls{logloss} \eqref{equ_log_loss}. 
The \gls{erm} principle suggest to learn the parameter vector $\weights$ by minimizing the average 
\gls{logloss} \eqref{equ_def_emp_risk_logreg} obtained for a \gls{trainset} $\dataset= \{ (\featurevec^{(\sampleidx)}, \truelabel^{(\sampleidx)}) \}_{\sampleidx=1}^{\samplesize}$. 
The \gls{trainset} consists of \gls{datapoint}s with features $\featurevec^{(\sampleidx)} \in \mathbb{R}^{\featuredim}$ 
and binary labels $\truelabel^{(\sampleidx)} \in \{-1,1\}$. 

We can rewrite \gls{erm} for \gls{logreg} as the optimization problem 
\begin{align} 
	\label{equ_smooth_problem_logeg}
	\overline{\weights}& = \argmin_{\weights \in \mathbb{R}^{\featuredim}} f(\weights) \nonumber \\ 
	\mbox{ with } f(\weights) & = (1/\samplesize) \sum_{\sampleidx=1}^{\samplesize} \log\big( 1\!+\!\exp \big( - \truelabel^{(\sampleidx)} \weights^{T} \featurevec^{(\sampleidx)} \big)\big). 
\end{align} 
The objective function $f(\weights)$ is differentiable and therefore we can use \gls{gd} \eqref{equ_def_GD_step} 
to solve \eqref{equ_smooth_problem_logeg}. We can write down the \gls{gradient} of the objective function in \eqref{equ_smooth_problem_logeg} in 
closed-form as 
\begin{equation}
	\label{equ_gradient_logistic_regression}
	\nabla f(\weights) = (1/\samplesize) \sum_{\sampleidx=1}^{\samplesize} \frac{-\truelabel^{(\sampleidx)}}{1 + \exp ( \truelabel^{(\sampleidx)} \weights^{T} \featurevec^{(\sampleidx)})} \featurevec^{(\sampleidx)}.
\end{equation} 
Inserting \eqref{equ_gradient_logistic_regression} into the \gls{gd} step \eqref{equ_def_GD_step} yields Algorithm \ref{alg:gd_logreg}. 
\begin{algorithm}[htbp]
	\caption{\Gls{logreg} via \gls{gd}}\label{alg:gd_logreg}
	\begin{algorithmic}[1]
		\renewcommand{\algorithmicrequire}{\textbf{Input:}}
		\renewcommand{\algorithmicensure}{\textbf{Output:}}
		\Require   labeled dataset $\dataset=\{ (\featurevec^{(\sampleidx)}, \truelabel^{(\sampleidx)}) \}_{\sampleidx=1}^{\samplesize}$ containing feature vectors 
		$\featurevec^{(\sampleidx)} \in \mathbb{R}^{\featuredim}$ and labels $\truelabel^{(\sampleidx)} \in \mathbb{R}$; \gls{gd} learning rate $\lrate >0$. 
		\Statex\hspace{-6mm}{\bf Initialize:}set $\weights^{(0)}\!\defeq\!\mathbf{0}$; set iteration counter $\itercntr\!\defeq\!0$   
		\Repeat 
		\State $\itercntr\!\defeq\! \itercntr\!+\!1$    (increase iteration counter) 
		\State \label{equ_step_updat_logreg_GD}  $\weights^{(\itercntr)} \defeq \weights^{(\itercntr\!-\!1)}\!+\!\lrate (1/\samplesize) \sum_{\sampleidx=1}^{\samplesize} \frac{\truelabel^{(\sampleidx)}}{1\!+\!\exp \big( \truelabel^{(\sampleidx)} \big(\weights^{(\itercntr\!-\!1)}\big)^{T} \featurevec^{(\sampleidx)}\big)} \featurevec^{(\sampleidx)}$  (do a \gls{gd} step \eqref{equ_def_GD_step})
		\Until stopping criterion met 
		\Ensure $\weights^{(\itercntr)}$, which approximates a solution $\overline{\weights}$ of \eqref{equ_smooth_problem_logeg})
	\end{algorithmic}
\end{algorithm}

Let us have a closer look on the update in step \eqref{equ_step_updat_logreg_GD} 
of Algorithm \ref{alg:gd_logreg}. This step amounts to computing
\begin{equation}
	\label{equ_update_logreg_GD}
	\weights^{(\itercntr)} \defeq \weights^{(\itercntr\!-\!1)} + \lrate (1/\samplesize) \sum_{\sampleidx=1}^{\samplesize} \frac{\truelabel^{(\sampleidx)}}{1 + \exp \big( \truelabel^{(\sampleidx)} \big( \vw^{(\itercntr\!-\!1)}  \big)^{T} \featurevec^{(\sampleidx)}\big)} \featurevec^{(\sampleidx)}. 
\end{equation}
Similar to the \gls{gd} step \eqref{equ_update_gd_linreg} for \gls{linreg}, also the \gls{gd} step 
\eqref{equ_update_logreg_GD} for \gls{logreg} can be interpreted as an implementation of the 
trial-and-error principle. Indeed, \eqref{equ_update_logreg_GD} corrects the previous guess (or approximation) $\weights^{(\itercntr\!-\!1)}$ for 
the optimal parameter vector $\overline{\weights}$ by the correction term 
\begin{equation}
	\label{equ_correction_logreg_GD}
	(\lrate/\samplesize) \sum_{\sampleidx=1}^{\samplesize} \underbrace{ \frac{\truelabel^{(\sampleidx)}}{1 + \exp ( \truelabel^{(\sampleidx)} \weights^{T} \featurevec^{(\sampleidx)})}}_{e^{(\sampleidx)}} \featurevec^{(\sampleidx)}. 
\end{equation}
The correction term \eqref{equ_correction_logreg_GD} is a weighted average of the feature vectors $\featurevec^{(\sampleidx)}$.  
The feature vector $\featurevec^{(\sampleidx)}$ is weighted by the factor $(\lrate/\samplesize) \cdot e^{(\sampleidx)}$. 
These weighting factors are a product of the global factor $(\lrate/\samplesize)$ that applies equally to all 
feature vectors $\featurevec^{(\sampleidx)}$. The global factor is multiplied by a \gls{datapoint}-specific factor 
$e^{(\sampleidx)} =  \frac{\truelabel^{(\sampleidx)}}{1 + \exp ( \truelabel^{(\sampleidx)} \weights^{T} \featurevec^{(\sampleidx)})}$, which 
quantifies the error of the classifier  $h^{(\weights^{(\itercntr\!-\!1)})}(\featurevec^{(\sampleidx)}) =   \big(\weights^{(\itercntr\!-\!1)})^{T} \featurevec^{(\sampleidx)}$ for a single \gls{datapoint} with true label $\truelabel^{(\sampleidx)} \in \{-1,1\}$ and features $\featurevec^{(\sampleidx)} \in \mathbb{R}^{\featuredim}$. 

We can use the sufficient condition \eqref{equ_GD_conv_guarantee} for the convergence of \gls{gd} steps 
to guide the choice of the \gls{learnrate} $\lrate$ in Algorithm \ref{alg:gd_logreg}. To apply condition 
\eqref{equ_GD_conv_guarantee}, we need to determine the Hessian $\nabla^{2} f(\weights)$ matrix of the 
objective function $f(\weights)$ underlying \gls{logreg} (see \eqref{equ_smooth_problem_logeg}). 
Some basic calculus reveals (see \cite[Ch. 4.4.]{hastie01statisticallearning})
\begin{equation}
	\label{equ_hessian_logreg}
	\nabla^{2} f(\weights) =  (1/\samplesize) \mX^{T} \mD \featuremtx. 
\end{equation}
Here, we used the feature matrix $\featuremtx=\big(\featurevec^{(1)},\ldots,\featurevec^{(\samplesize)}\big)^{T} \in \mathbb{R}^{\samplesize \times \featuredim}$ (see \eqref{equ_def_vec_matrix}) 
and the diagonal matrix $\mD = {\rm diag} \{d_{1},\ldots,d_{\samplesize}\} \in \mathbb{R}^{\samplesize \times \samplesize}$ with diagonal elements 
\begin{equation}
	d_{\sampleidx} = \frac{1}{1+\exp(-\weights^{T} \featurevec^{(\sampleidx)})} \bigg(1- \frac{1}{1+\exp(-\weights^{T} \featurevec^{(\sampleidx)})}  \bigg). \label{equ_diag_entries_log_reg} 
\end{equation} 
We highlight that, in contrast to the Hessian \eqref{equ_hessian_linreg} of the objective function 
arising in linear regression, the Hessian \eqref{equ_hessian_logreg} of \gls{logreg} varies 
with the parameter vector $\weights$. This makes the analysis of Algorithm \ref{alg:gd_logreg} and the 
optimal choice for the \gls{learnrate} $\lrate$ more difficult compared to Algorithm \ref{alg:gd_linreg}. 
At least, we can ensure convergence of \eqref{equ_update_logreg_GD} (towards a solution 
of \eqref{equ_smooth_problem_logeg}) for the \gls{learnrate} $\lrate=1$ if we normalize feature vectors 
such that $\| \featurevec^{(\sampleidx)} \|=1$. This follows from the fact the diagonal entries 
\eqref{equ_diag_entries_log_reg} take values in the interval $[0,1]$. 

\section{Data Normalization} 
\label{sec_data_normalization} 

The number of \gls{gd} steps \eqref{equ_def_GD_step} required to reach the minimum (within a prescribed accuracy) 
of the objective function \eqref{equ_def_cost_MSE} depends crucially on the \gls{condnr} \cite{CvxBubeck2015,JungFixedPoint}
\begin{equation}
	\label{equ_def_condition_number} 
	\kappa( \featuremtx^{T} \featuremtx) \defeq \eigval{\rm max}/\eigval{\rm min}. 
\end{equation} 
Here, we use the largest and smallest \gls{eigenvalue} of the matrix $\featuremtx^{T} \featuremtx$, 
denoted as $ \eigval{\rm max}$ and $\eigval{\rm min}$, respectively. 
The \gls{condnr} \eqref{equ_def_condition_number} is only well-defined if the columns of the 
feature matrix $\featuremtx$ \eqref{equ_def_vec_matrix} (which are the feature vectors $\featurevec^{(\sampleidx)}$), 
are linearly independent. In this case the \gls{condnr} is lower bounded 
as $1 \leq \kappa( \featuremtx^{T} \featuremtx)$. 

It can be shown that the \gls{gd} steps \eqref{equ_def_GD_step} converge faster for 
smaller \gls{condnr} $\kappa( \featuremtx^{T} \featuremtx)$ \cite{JungFixedPoint}. Thus, \gls{gd} 
will be faster for datasets with a feature matrix $\featuremtx$ such that  $\kappa( \featuremtx^{T} \featuremtx) \approx 1$. 
It is therefore often beneficial to pre-process the feature vectors using a normalization 
(or standardization) procedure as detailed in Algorithm \ref{alg_reshaping}. 

\begin{algorithm}[htbp]
	\caption{``Data Normalization''}\label{alg_reshaping}
	\begin{algorithmic}[1]
		\renewcommand{\algorithmicrequire}{\textbf{Input:}}
		\renewcommand{\algorithmicensure}{\textbf{Output:}}
		\Require   labeled dataset $\dataset = \{(\featurevec^{(\sampleidx)},y^{(\sampleidx)})\}_{\sampleidx=1}^{\samplesize}$
		\vspace*{2mm}
		\State remove sample means $\widehat{\featurevec}=(1/\samplesize) \sum_{\sampleidx=1}^{\samplesize}  \featurevec^{(\sampleidx)}$ from features, i.e., 
		\begin{equation}
			\nonumber
			\featurevec^{(\sampleidx)} \defeq \featurevec^{(\sampleidx)} - \widehat{\featurevec} \mbox{ for  }  \sampleidx=1,\ldots,\samplesize
		\end{equation} 
		\State normalise features to have unit variance,  
		\begin{equation} 
			\nonumber
			\hat{\feature}^{(\sampleidx)}_{\featureidx} \defeq \feature^{(\sampleidx)}_{\featureidx}/ \hat{\sigma}   \mbox{ for  } \featureidx=1,\ldots,\featuredim \mbox{ and } \sampleidx=1,\ldots,\samplesize
		\end{equation}
		with the empirical (sample) variance $\hat{\sigma}_{\featureidx}^{2}  =(1/\samplesize) \sum_{\sampleidx=1}^{\samplesize} \big( \feature^{(\sampleidx)}_{\featureidx} \big)^{2}$ 
		\Ensure normalized feature vectors $\{\hat{\featurevec}^{(\sampleidx)}\}_{\sampleidx=1}^{\samplesize}$
	\end{algorithmic}
\end{algorithm}
Algorithm \ref{alg_reshaping} transforms the original feature vectors $\featurevec^{(\sampleidx)}$ into new feature vectors $\widehat{\featurevec}^{(\sampleidx)}$ 
such that the new feature matrix $\widehat{\featuremtx} = (\widehat{\featurevec}^{(1)},\ldots,\widehat{\featurevec}^{(\samplesize)})^{T}$ 
is better conditioned than the original feature matrix, i.e., $\kappa( \widehat{\featuremtx}^{T} \widehat{\featuremtx}) <  \kappa( \featuremtx^{T} \featuremtx)$.

\section{Stochastic GD}
\label{sec_sgd}
Consider the \gls{gd} steps \eqref{equ_def_GD_step} for minimizing the \gls{emprisk} \eqref{equ_obj_emp_risk_GD}. 
The gradient $\nabla f(\weights)$ of the objective function \eqref{equ_obj_emp_risk_GD} has a particular structure. 
Indeed, this gradient is a sum 
\vspace*{-2mm}
\begin{equation}
	\label{eq_gradient_sum}
	\nabla f(\weights) = (1/\samplesize) \sum_{\sampleidx=1}^{\samplesize} \nabla f_{\sampleidx}(\vw)   \mbox{ with } f_{\sampleidx}(\weights) \defeq \loss{(\featurevec^{(\sampleidx)},\truelabel^{(\sampleidx)})}{h^{(\weights)}}.
\end{equation} 
Each component of the sum \eqref{eq_gradient_sum} corresponds to one particular \gls{datapoint}s 
$(\featurevec^{(\sampleidx)},\truelabel^{(\sampleidx)})$, for $\sampleidx=1,\ldots,\samplesize$. 
We need to compute a sum of the form \eqref{eq_gradient_sum} for each new \gls{gd} step \eqref{equ_def_GD_step}. 

Computing the sum in \eqref{eq_gradient_sum} can be computationally challenging for at least 
two reasons. First, computing the sum is challenging for very large datasets with $\samplesize$ 
in the order of billions. Second, for datasets which are stored in different data 
centres located all over the world, the summation would require a huge amount of 
network resources. Moreover, the finite transmission rate of communication networks 
limits the rate by which the \gls{gd} steps \eqref{equ_def_GD_step} can be executed. 

The idea of \index{stochastic gradient descent} \gls{stochGD} is to replace the exact 
gradient $\nabla f(\weights)$ \eqref{eq_gradient_sum} by an approximation that is easier 
to compute than a direct evaluation of \eqref{eq_gradient_sum}. The word ``stochastic'' 
in the name \gls{stochGD} hints already at the use of a stochastic approximation $\noisygrad(\weights) 
\approx \nabla f(\weights)$. It turns out that using a \gls{gradient} approximation $\noisygrad(\weights)$ 
can result in significant savings in computational complexity while incurring a graceful degradation 
in the overall optimization accuracy. The optimization accuracy (distance to minimum of $f(\weights)$) 
depends crucially on the ``\gls{gradient} noise''
\begin{equation} 
	\label{equ_def_gradient_noise_generic}
	\varepsilon \defeq \nabla f(\weights)- \noisygrad(\weights). 
\end{equation} 

The elementary step of most \gls{stochGD} methods is obtained from the \gls{gd} step \eqref{equ_def_GD_step} 
by replacing the exact \gls{gradient} $\nabla f(\weights)$ with some stochastic approximation $\noisygrad(\weights)$, 
\begin{equation}
	\label{equ_SGD_update}
	\weights^{(\itercntr\!+\!1)} = \weights^{(\itercntr)} - \lrate_{\itercntr} \noisygrad \big( \weights^{(\itercntr)} \big), 
\end{equation} 
As the notation in \eqref{equ_SGD_update} indicates, \gls{stochGD} methods use a \gls{learnrate} $\lrate_{\itercntr}$ 
that varies between different iterations. 

To avoid accumulation of the \gls{gradient} noise \eqref{equ_def_gradient_noise_generic} during the \gls{stochGD} 
updates \eqref{equ_SGD_update}, \gls{stochGD} methods typically decrease the \gls{learnrate} $\lrate_{\itercntr}$ as 
the iterations proceed. The precise dependence of the \gls{learnrate} $\lrate_{\itercntr}$ on the iteration index $\itercntr$ 
is referred to as a \index{learning rate schedule}\gls{learnrate} schedule \cite[Chapter 8]{Goodfellow-et-al-2016}. 
One possible choice for the \gls{learnrate} schedule is $\lrate_{\itercntr}\!\defeq\!1/\itercntr$ \cite{Murata98astatistical}. 
Exercise \ref{ex_SGD_learning_rate} discusses conditions on the \gls{learnrate} schedule that guarantee convergence 
of the updates \gls{stochGD} to the minimum of $f(\weights)$. 

The approximate (``noisy'') gradient $\noisygrad(\weights)$ can be obtained by different randomization strategies. 
The most basic form of \gls{stochGD} constructs the gradient approximation $\noisygrad(\weights)$ by replacing the 
sum \eqref{eq_gradient_sum} with a randomly selected component, 
\begin{equation} 
\noisygrad(\weights) \defeq \nabla f_{\hat{\sampleidx}}(\weights).
\end{equation} 
The index $\hat{\sampleidx}$ is chosen randomly from the set $\{1,\ldots,\samplesize\}$. The resulting \Gls{stochGD} 
method then repeats the update 
\begin{equation}
	\label{equ_SGD_update_basic_form}
	\weights^{(\itercntr\!+\!1)} = \weights^{(\itercntr)} - \lrate \nabla f_{\hat{\sampleidx}_{\itercntr}}(\weights^{(\itercntr)}), 
\end{equation} 
sufficiently often. Every update \eqref{equ_SGD_update_basic_form} uses a ``fresh'' randomly chosen (drawn)
index $\hat{\sampleidx}_{\itercntr}$. Formally, the indices $\hat{\sampleidx}_{\itercntr}$ are realizations 
of \gls{iid} \gls{rv}s whose common \gls{probdist} is the uniform distribution over the index 
set $\{1,\ldots,\samplesize\}$. 

Note that \eqref{equ_SGD_update_basic_form} replaces the summation over the \gls{trainset} during 
the \gls{gd} step \eqref{equ_def_GD_step} by randomly selecting a single component of this sum. 
The resulting savings in computational complexity can be significant when the \gls{trainset} consists 
of a large number of \gls{datapoint}s that might be stored in a distributed fashion (in the ``cloud''). 
The saving in computational complexity of \gls{stochGD} comes at the cost of introducing a non-zero 
gradient noise 
\begin{align}
	\label{equ_gradient_noise_simple_form}
	\varepsilon_{\itercntr} & \stackrel{\eqref{equ_def_gradient_noise_generic}}{=} \nabla f( \weights^{(\itercntr)} )  - \noisygrad  \big( \weights^{(\itercntr)} \big) \nonumber \\ 
	& = \nabla f( \weights^{(\itercntr)} ) -  \nabla f_{\hat{\sampleidx}_{\itercntr}}(\weights). 
\end{align} 

{\bf Mini-Batch \Gls{stochGD}.} Let us now discuss a variant of \gls{stochGD} that tries to reduce the approximation 
error (gradient noise) \eqref{equ_gradient_noise_simple_form} arising in the \gls{stochGD} step \eqref{equ_SGD_update_basic_form}. 
The idea behind this variant, referred to as mini-batch \gls{stochGD}, is quite simple. Instead of using only a single randomly selected 
component $\nabla f_{\sampleidx}(\weights)$ (see \eqref{eq_gradient_sum}) for constructing a gradient 
approximation, mini-batch \gls{stochGD} uses several randomly chosen components. 

We summarize mini-batch \gls{stochGD} in Algorithm \ref{alg:minibatch_gd} which requires an integer batch size $\batchsize$ 
as input parameter. Algorithm \ref{alg:minibatch_gd} repeats the \gls{stochGD} step  \eqref{equ_SGD_update} using a 
gradient approximation that is constructed from a randomly selected subset $\batch = \{ \sampleidx_{1},\ldots,\sampleidx_{\batchsize}\}$ (a ``batch''), 
\begin{equation}
	\label{equ_def_gradient_approx_mini_batch}
	\noisygrad  \big( \weights  \big) = (1/\batchsize) \sum_{\sampleidx' \in \batch}  \nabla f_{\sampleidx'}(\weights). 
\end{equation} 
For each new iteration of Algorithm \ref{alg:minibatch_gd}, a new batch $\batch$ is generated by a random generator.  
\begin{algorithm}[htbp]
	\caption{Mini-Batch \gls{stochGD}}\label{alg:minibatch_gd}
	\begin{algorithmic}[1]
		\renewcommand{\algorithmicrequire}{\textbf{Input:}}
		\renewcommand{\algorithmicensure}{\textbf{Output:}}
		\Require components $f_{\sampleidx}(\weights)$, for $\sampleidx=1,\ldots,\samplesize$ of objective function $f(\weights)=\sum_{\sampleidx=1}^{\samplesize} f_{\sampleidx}(\weights)$ ; batch size $\batchsize$; \gls{learnrate} schedule $\lrate_{\itercntr} >0$. 
		\Statex\hspace{-6mm}{\bf Initialize:} set $\vw^{(0)}\!\defeq\!\mathbf{0}$; set iteration counter $\itercntr\!\defeq\!0$   
		\Repeat 
		\State randomly select a batch $\batch = \{\sampleidx_{1},\ldots,\sampleidx_{\batchsize}\} \subseteq \{1,\ldots,\samplesize\}$ of indices 
		that select a subset of components $f_{\sampleidx}$
		\State compute an approximate gradient $\noisygrad  \big( \weights^{(\itercntr)} \big)$ using \eqref{equ_def_gradient_approx_mini_batch}
		\State $\itercntr \defeq \itercntr +1$    (increase iteration counter) 
		\State  $\weights^{(\itercntr)} \defeq \weights^{(\itercntr\!-\!1)} - \lrate_{\itercntr} \noisygrad  \big( \weights^{(\itercntr-1)} \big)$ \label{equ_sGD_step_minibtatch}
		\Until stopping criterion met 
		\Ensure $\weights^{(\itercntr)}$ (which approximates $\argmin_{\weights \in \mathbb{R}^{\featuredim}} f(\weights)$ ))
	\end{algorithmic}
\end{algorithm}
Note that Algorithm \ref{alg:minibatch_gd} includes the basic \gls{stochGD} variant \eqref{equ_SGD_update_basic_form} 
as a special case for the batch size $\batchsize=1$. Another special case is $\batchsize= \samplesize$, where 
the \gls{stochGD} step \ref{equ_sGD_step_minibtatch} in Algorithm \ref{alg:minibatch_gd} becomes an 
ordinary \gls{gd} step \eqref{equ_def_GD_step}.

{\bf Online Learning.} A main motivation for the \gls{stochGD} step \eqref{equ_SGD_update_basic_form} is 
that a \gls{trainset} is already collected but so large that the sum in \eqref{eq_gradient_sum} is 
computationally intractable. Another variant of \gls{stochGD} is obtained for sequential (time-series) data. In particular, consider 
\gls{datapoint}s that are gathered sequentially, one new \gls{datapoint} $\big( \featurevec^{(\timeidx)},\truelabel^{(\timeidx)} \big)$ 
at each time instant $\timeidx=1,2,\ldots.$. With each new \gls{datapoint} $\big( \featurevec^{(\timeidx)},\truelabel^{(\timeidx)} \big)$  
we can access a new component $f_{\timeidx}(\weights) = \loss{(\featurevec^{(\timeidx)},\truelabel^{(\timeidx)})}{h^{(\weights)}}$ (see \eqref{equ_obj_emp_risk_GD}). For sequential data, we can use a slight modification of the \gls{stochGD} step \eqref{equ_SGD_update} 
to obtain an online learning method (see Section \ref{sec_online_learning}). 
This online variant of \gls{stochGD} computes, at each time instant $\timeidx$,  
\begin{equation}
	\label{equ_SGD_update_timeiedx}
	\weights^{(\timeidx)} \defeq \weights^{(\timeidx\!-\!1)} - \lrate_{\timeidx} \nabla f_{\timeidx}(\weights^{(\timeidx\!-\!1)}).
\end{equation} 

\section{Advanced Gradient-Based Methods}
\label{sec_adv_gd_methods}

The main idea underlying \gls{gd} and \gls{stochGD} is to approximate the objective function \eqref{equ_obj_emp_risk_GD} locally 
around a current guess or approximation $\weights^{(\itercntr)}$ for the optimal weights. This 
local approximation is a tangent hyperplane whose normal vector is determined by the gradient 
$\nabla f (\weights^{(\itercntr)}$. We then obtain an updated (improved) approximation by minimizing 
this local approximation, i.e., by doing a \gls{gd} step \eqref{equ_def_GD_step}. 

The idea of advanced gradient methods \cite[Ch. 8]{Goodfellow-et-al-2016} is to exploit the information provided by the 
gradients $\nabla f\big(\weights^{(\itercntr')}\big)$ at previous iterations $\itercntr'=1,\ldots,\itercntr$, 
to build an improved local approximation of $f(\weights)$ around a current iterate $\weights^{(\itercntr)}$. 
Figure \ref{fig_improved_local_approx} indicates such an improved local approximation of $f(\weights)$ 
which is non-linear (e.g., quadratic). These improved local approximations can be used to adapt 
the \gls{learnrate} during the \gls{gd} steps \eqref{equ_def_GD_step}. 

Advanced \gls{gdmethods} use improved local approximations to modify the \gls{gradient} $\nabla f\big(\weights^{(\itercntr')}\big)$ 
to obtain an improved update direction. Figure \ref{fig_improved_local_approx_better_directions} depicts the contours of an objective 
function $f(\weights)$ for which the gradient $\nabla f\big(\weights^{(\itercntr)}\big)$  points only weakly 
towards the optimal parameter vector $\overline{\weights}$ (minimizer of $f(\weights)$). The \gls{gradient} history $\nabla f\big(\weights^{(\itercntr')}\big)$, for$\itercntr'=1,\ldots,\itercntr$, allows to detect such an unfavourable geometry 
of the objective function. Moreover, the \gls{gradient} history can be used to ``correct'' the 
update direction $\nabla f\big(\weights^{(\itercntr)}\big)$ to obtain an improved update 
direction towards the optimum parameter vector $\overline{\weights}$.


\tikzset{global scale/.style={
		scale=#1,
		every node/.append style={scale=#1}
	}
}
\begin{figure}[htbp] 
\begin{center}
\begin{tikzpicture}[global scale = 0.7]             
	
	\draw[blue,line width=1.5pt] (0,0) .. controls (-5,0.2) and  (-6,2) .. (-7,5)
	
	node[sloped, inner sep=0cm, above, pos=0.15,    
	anchor=south west,
	minimum height=(10.5)*0.3cm, minimum width=(10.5)*.3cm](N){}        
	
	node[sloped, inner sep=0cm, above, pos=0.85,
	anchor=south west,
	minimum height=(10.5)*0.3cm, minimum width=(10.5)*.3cm](M){};       
	
	\draw[blue,line width=1.5pt, dashed] (N.south west) .. controls (-5,1.2) and  (-6,2.8) .. (M.south west);
	
	\path [draw, blue!50, very thick]($(N.south west)!0.8!(N.south east)$) -- ($(N.south west)!0.8!180:(N.south east)$);           
	\path [draw, blue!50, very thick]($(M.south west)!0.5!(M.south east)$) -- ($(M.south west)!0.5!180:(M.south east)$);
	
	\draw[color=blue!50, fill=blue!50](N.south west) circle (0.12);
	\draw[color=blue!50, fill=blue!50](M.south west) circle (0.12);
	
	\node[font=\fontsize{20}{0}\selectfont] at (-7.5,3.7) {$\weights^{(\itercntr\!-\!1)}$};
	\node[font=\fontsize{20}{0}\selectfont] at (-5.4,1) {$f(\weights)$};
	\node[font=\fontsize{20}{0}\selectfont] at (-2,-0.4) {$\weights^{(\itercntr)}$};
	
\end{tikzpicture}
\caption{Advanced \gls{gdmethods} use the gradients at previous iterations to construct an 
improved (non-linear) local approximation (dashed) of the objective function $f(\weights)$ (solid). }
\label{fig_improved_local_approx}
\end{center}
\end{figure}

\begin{figure}[htbp]
\tikzset{global scale/.style={
		scale=#1,
		every node/.append style={scale=#1}
	}
}
\begin{center}
\begin{tikzpicture}[global scale = 1]             
	\draw[color=blue, rotate=20] (0,0) ellipse (2.4 and 0.3);
	
	\draw[color=blue, rotate=20] (0,0) ellipse (3.3 and 0.6);
	\draw[color=red, fill=red] (1.9,1.15) circle (0.12);
	
	\draw [stealth-, very thick, color=red](1.1,-0.7) -- (1.93,1.2);
	\draw [stealth-, dashed, very thick, color=red](2.6,-1.2) -- (1.9,1.2);
	\node[font=\fontsize{12}{0}\selectfont] at (0,0) {\textcolor{red}{$\star$}};
	\node[font=\fontsize{10}{0}\selectfont] at (1.1,-1.1) {adapted};
	\node[font=\fontsize{10}{0}\selectfont] at (1.1,-1.55) {direction};
	\node[font=\fontsize{10}{0}\selectfont] at (3.5,0) {\gls{gd}};
	\node[font=\fontsize{10}{0}\selectfont] at (3.5,-0.45) {direction};
	\node[font=\fontsize{10}{0}\selectfont] at (3.55,-1) {$-\bigtriangledown f(\weights^{(\itercntr)})$};
	\node[font=\fontsize{10}{0}\selectfont] at (2.1,1.6) {$\weights^{(\itercntr)}$};
\end{tikzpicture}
\end{center} 
\caption{Advanced \gls{gdmethods} use improved (non-linear) local approximations of the objective 
	function $f(\weights)$ to ``nudge'' the update direction towards the optimal parameter 
	vector $\overline{\weights}$. The update direction of plain vanilla \gls{gd} \eqref{equ_def_GD_step} 
	is the negative \gls{gradient} $-\nabla f\big( \weights^{(\itercntr)}\big)$. For some 
	objective functions the negative \gls{gradient} might be only weakly correlated with 
	the straight direction from $\weights^{(\itercntr)}$ towards the optimal parameter vector ($\star$).  }
\label{fig_improved_local_approx_better_directions}
\end{figure} 

\newpage
\section{Exercises}

\begin{exercise}[Use Knowledge About Problem Class.]
Consider the space $\mathcal{P}$ of sequences $f = (f[0],f[1],\ldots)$ that have the following properties 
\begin{itemize} 
\item for each sequence there is an index $\sampleidx^{(f)}$ such that $f$ is monotone increasing for indices $\sampleidx' \geq \sampleidx^{(f)}$ 
and monotone decreasing for indices $\sampleidx' \geq \sampleidx^{(f)}$
\item any change point $\itercntr$ of $f$, where $f[\itercntr] \neq f[\itercntr\!+\!1]$ can only occrur at integer multiples of 
$100$, e.g., $\itercntr\!=\!100$ or $\itercntr\!=\!300$. 
\end{itemize} 
Given a function $f \in \mathcal{P}$ and starting point $\itercntr_{0}$ our goal is to find the 
minimum value of $\min_{\itercntr} f[\itercntr] = f[\itercntr^{(f)}]$ as quickly as possible. 
Can you constuct an iterative algorithm that can access a given function $f$ only by querying the values $f[\itercntr], f[\itercntr\!-\!1]$ and $f[\itercntr\!+\!1]$ for any given index $\itercntr$. 
\end{exercise} 

\begin{exercise}[\Gls{learnrate} Schedule for \gls{stochGD}]
	\label{ex_SGD_learning_rate}
	Let us learn a linear hypothesis $h(\featurevec) = \weights^{T} \featurevec$ using \gls{datapoint}s 
	that arrive sequentially at discrete time instants $\timeidx=0,1,\ldots$. At time $\timeidx$, we gather  
	a new \gls{datapoint} $\big(\featurevec^{(\itercntr)},\truelabel^{(\itercntr)} \big)$. The \gls{datapoint}s 
	can be modelled as realizations of \gls{iid} copies of a \gls{rv} $\big(\featurevec,\truelabel\big)$. 
	The \gls{probdist} of the features $\featurevec$ is a standard 
	multivariate normal distribution $\mathcal{N}(\mathbf{0},\mathbf{I})$. 
	The label of a random\footnote{More precisely, a \gls{datapoint} that is obtained as the realization of \gls{rv}.} \gls{datapoint} is related 
	to its features via $\truelabel = \overline{\weights}^{T} \featurevec+ \varepsilon$ with some fixed but unknown 
	true parameter vector $\overline{\weights}$. The additive noise $\varepsilon \sim \mathcal{N}(0,1)$ follows 
	a standard normal distribution. We use \gls{stochGD} to learn the parameter vector $\weights$ of a linear hypothesis, 
	\begin{equation}
		\label{equ_SGD_update_online_exercise}
		\weights^{(\timeidx\!+\!1)} = \weights^{(\timeidx)} - \lrate_{\timeidx}  \big ( \big(\weights^{(\timeidx)}\big)^{T} \featurevec^{(\timeidx)} - \truelabel^{(\timeidx)} \big) \featurevec^{(\timeidx)}.
	\end{equation} 
	with \gls{learnrate} schedule $\lrate_{\timeidx} = \beta/\timeidx^{\gamma}$. Note that we compute a new 
	\gls{stochGD} iteration \eqref{equ_SGD_update_online_exercise} for each new time instant $\timeidx$. 
	What conditions on the hyper-parameters $\beta, \gamma$ ensure that $\lim\limits_{\timeidx \rightarrow \infty} \weights^{(\timeidx)} = \overline{\weights}$ in distribution?
\end{exercise}

\begin{exercise}[ImageNet.]
The ``ImageNet'' database contains more than $10^{6}$ images \cite{ImageNet}. These 
images are labeled according to their content (e.g., does the image show a dog?) and stored 
as a file of size at least $4$ kilobytes. We want to learn 
a classifier that allows to predict if an image shows a dog or not. To learn this classifier 
we run \gls{gd} for \gls{logreg} on a small computer that has $32$ kilobytes memory 
and is connected to the internet with bandwidth of $1$ Mbit/s. Therefore, for each 
single \gls{gd} update \eqref{equ_def_GD_step} it must essentially download all images 
in ImageNet. How long would such a single \gls{gd} update take ? 
\end{exercise}

\begin{exercise}[Apple or No Apple?]
\label{ex_GD_distributed_data}
Consider \gls{datapoint}s being images of size of $1024 \times 1024$ pixels. 
Each image is characterized by the RGB pixel color intensities (value range $0,\ldots,255$ resuling in $24$
 bits for each pixel), which we stack into a feature vector $\featurevec \in \mathbb{R}^{\featurelen}$.
We assign each image the label $\truelabel=1$ if it shows an apple and $\truelabel=-1$ if it does not show an apple. 
We use \gls{logreg} to learn a linear hypothesis $h(\featurevec) = \weights^{T} \featurevec$ to classify 
an image according to $\hat{\truelabel} =1$ if $h(\featurevec) \geq 0$. The \gls{trainset} consists 
of $\samplesize=10^{10}$ labeled images which are stored in the cloud. We implement 
the ML method on our own laptop which is connected to the internet with a bandwidth  
of at most $100$ Mbps. Unfortunately we can only store at most five images on our 
computer. How long does it take at least to complete one single \gls{gd} step ?
\end{exercise}

\begin{exercise}[Feature Normalization To Speed Up \gls{gd}]
	\label{ex_GD_data normalization}
	Consider the dataset with feature vectors $\featurevec^{(1)}=(100,0)^{T} \in \mathbb{R}^{2}$ 
	and $\featurevec^{(2)} = (0,1/10)^{T} \in \mathbb{R}^{2}$ which we stack into the matrix $\featuremtx = (\featurevec^{(1)},\featurevec^{(2)})^{T}$. 
	What is the \gls{condnr} of $\featuremtx^{T} \featuremtx$? What is the \gls{condnr} of $\big(\widehat{\featuremtx}\big)^{T} \widehat{\featuremtx}$ with the matrix $\widehat{\featuremtx}=(\widehat{\featurevec}^{(1)},\widehat{\featurevec}^{(2)})^{T}$ constructed from the normalized feature vectors $\widehat{\featurevec}^{(\sampleidx)}$ delivered by Algorithm \ref{alg_reshaping}. 
\end{exercise}

\begin{exercise}[Convergence of \Gls{gd} Steps]
\label{ex_GD_convergence_arbitrary}
Consider a differentiable objective function $f(\weights)$ whose argument is a parameter vector $\weights \in \mathbb{R}^{\featuredim}$. 
We make no assumption about smoothness or convexity. Thus, the function $f(\weights)$ might be non-convex and might also be 
not $\beta$ smooth. However, the gradient $\nabla f(\weights)$ is uniformly upper bounded $\| \nabla f(\weights) \| \leq 100$ for every 
$\weights$. Starting from some initial vector $\weights^{(0)}$ we construct a sequence of parameter vectors 
using \gls{gd} steps, 
\begin{equation}
	\label{equ_def_gd_exericse}
\weights^{(\itercntr+1)} =  \weights^{(\itercntr)}  - \lrate_{\itercntr } \nabla f \big(  \weights^{(\itercntr)}  \big). 
\end{equation}
The \gls{learnrate} $\lrate_{\itercntr }$ in \eqref{equ_def_gd_exericse} is allowed to vary between different iterations. 
Can you provide sufficient conditions on the evolution of the \gls{learnrate} $\lrate_{\itercntr }$, as iterations proceed, 
that ensure convergence of the sequence $\weights^{(0)}, \weights^{(1)}, \dots$.   
\end{exercise}

\newpage
\chapter{Model Validation and Selection} 
\label{ch_validation_selection}

\begin{figure}[htbp]
	\begin{center}
		\begin{tikzpicture}[ycomb]
			\draw[color=green,line width=26pt]
			plot coordinates{(0,3)};
			\node [below] at (0,0) {\gls{trainerr}} ; 
			\draw[color=red,line width=26pt]
			plot coordinates{(5,5)};
			\node [below] at (5,0) {validation error} ; 
			\draw[dashed,line width=2] (-1,2) -- (7,2) node[right,text width=5cm]{baseline or \index{benchmark}benchmark \\ (e.g., \gls{bayesrisk}, existing ML methods or human performance)};
		\end{tikzpicture}
	\end{center}
	\caption{We can \index{diagnose ML methods}diagnose a ML method by comparing its \gls{trainerr} 
		with its \gls{valerr}. Ideally both are on the same level as a \index{baseline}\gls{baseline} (or 
		benchmark error level).}
	\label{fig_bars_val_sel}
\end{figure}

Chapter \ref{ch_Optimization} discussed \gls{erm} as a principled approach to learning a good hypothesis 
out of a \gls{hypospace} or \gls{model}. \gls{erm} based methods learn a hypothesis $\hat{h} \in \hypospace$ 
that incurs minimum average loss on a set of labeled \gls{datapoint}s that serve as the \gls{trainset}. 
We refer to the average loss incurred by a hypothesis on the \gls{trainset} as the \gls{trainerr}. 
The minimum average \gls{loss} achieved by a hypothesis that solves the \gls{erm} might be referred to as 
the \gls{trainerr} of the overall ML method. This overall ML method is defined by the choice of 
\gls{hypospace} (or \gls{model}) and \gls{lossfunc}  (see Chapter \ref{ch_some_examples}). 

\Gls{erm} is sensible only if the \gls{trainerr} of a hypothesis is an reliable approximation 
for its loss incurred on \gls{datapoint}s outside the \gls{trainset}. Whether the training 
error of a hypothesis is a reliable approximation for its \gls{loss} on \gls{datapoint}s outside the 
\gls{trainset} depends on both, the statistical properties of the \gls{datapoint}s generated 
by an ML application and on the \gls{hypospace} used by the ML method.  

ML methods often use \gls{hypospace}s with a large \gls{effdim} (see Section \ref{sec_hypo_space}). As an example 
consider \gls{linreg} (see Section \ref{sec_lin_reg}) with \gls{datapoint}s having a large number $\featuredim$ of 
features (this setting is referred to as the \index{high-dimensional regime}\gls{highdimregime}). The \gls{effdim} of 
the linear hypothesis space \eqref{equ_lin_hypospace}, which is used by \gls{linreg}, is equal to the number $\featuredim$ 
of features. Modern technology allows to collect a huge number of features about individual \gls{datapoint}s which 
implies, in turn, that the \gls{effdim} of \eqref{equ_lin_hypospace} is large. Another example of a high-dimensional \gls{hypospace} 
arises in deep learning methods using a \gls{hypospace} are constituted by all maps represented by an \gls{ann} 
with billions of tunable \gls{parameters}. 

A high-dimensional \gls{hypospace} is very likely to contain a hypothesis that perfectly 
fits any given \gls{trainset}. Such a hypothesis achieves a very small \gls{trainerr} but might 
incur a large loss when predicting the labels of a \gls{datapoint} that is not included in \gls{trainset}. 
Thus, the (minimum) \gls{trainerr} achieved by a hypothesis learnt by \gls{erm} can be misleading. 
We say that a ML method, such as \gls{linreg} using too many features, overfits the \gls{trainset} 
when it learns a hypothesis (e.g., via \gls{erm}) that has small \gls{trainerr} but incurs 
much larger loss outside the \gls{trainset}.  


Section \ref{sec_overfitting_sec_6} shows that \gls{linreg} will overfit a \gls{trainset} 
as soon as the number of features of a \gls{datapoint} exceeds the size of the \gls{trainset}. 
Section \ref{sec_validate_predictor} demonstrates how to validate a learnt hypothesis by 
computing its average loss on \gls{datapoint}s which are not contained in the \gls{trainset}. 
We refer to the set of \gls{datapoint}s used to validate the learnt hypothesis as a \index{validation set}\gls{valset}. 
If a ML method overfits the \gls{trainset}, it learns a hypothesis whose \gls{trainerr} is much 
smaller than its \gls{valerr}. We can detect if a ML method overfits by comparing its \gls{trainerr} with 
its \gls{valerr} (see Figure \ref{fig_bars_val_sel}). 

We can use the \gls{valerr} not only to detect if a ML method overfits. The \gls{valerr} 
can also be used as a quality measure for the \gls{hypospace} or \gls{model} used by the ML method.
This is analogous to the concept of a \gls{lossfunc} that allows us to evaluate the quality 
of a hypothesis $h\!\in\!\hypospace$. Section \ref{sec_modsel} shows how to select between 
ML methods using different \gls{model}s by comparing their \gls{valerr}s. 

Section \ref{sec_gen_linreg} uses a simple probabilistic model for the data to study the 
relation between the \gls{trainerr} of a learnt hypothesis and its expected loss (see \eqref{equ_def_risk}). 
This probabilistic analysis reveals the interplay between the data, the \gls{hypospace} 
and the resulting \gls{trainerr} and \gls{valerr} of a ML method. 

Section \ref{sec_the_bootsrap} discusses the \index{bootsrap}\gls{bootstrap}  
as a simulation based alternative to the probabilistic analysis of Section \ref{sec_gen_linreg}. 
While Section \ref{sec_gen_linreg} assumes a specific \gls{probdist} of the \gls{datapoint}s, 
the \gls{bootstrap} does not require the specification of a \gls{probdist} underlying the data.

As indicated in Figure \ref{fig_bars_val_sel}, for some ML applications, we might have a 
\index{baseline} \gls{baseline} (or \index{benchmark}benchmark) for the achievable performance 
of ML methods. Such a \gls{baseline} might be obtained from existing ML methods, human performance 
levels or from a probabilistic model (see Section \ref{sec_gen_linreg}). Section \ref{sec_diagnosis_ML} details 
how the comparison between \gls{trainerr}, \gls{valerr} and (if available) a \gls{baseline} 
informs possible improvements for a ML method. These improvements might be obtained by 
collecting more \gls{datapoint}s, using more features of \gls{datapoint}s or by changing the \gls{hypospace} (or \gls{model}). 

Having a \gls{baseline} for the expected \gls{loss}, such as the \gls{bayesrisk}, allows to tell if a ML 
method already provides satisfactory results. If the \gls{trainerr} and the \gls{valerr} of a ML method 
are close to the \gls{baseline}, there might be little point in trying to further improve the ML method. 



\section{Overfitting}
\label{sec_overfitting_sec_6}

We now have a closer look at the occurrence of overfitting in \gls{linreg} methods. As discussed in  
Section \ref{sec_lin_reg}, \gls{linreg} methods learn a linear hypothesis $h(\featurevec) = \weights^{T} \featurevec$ 
which is parametrized by the parameter vector $\weights \in \mathbb{R}^{\featurelen}$. 
The learnt hypothesis is then used to predict the numeric label $\truelabel \in \mathbb{R}$ 
of a \gls{datapoint} based on its feature vector $\featurevec \in \mathbb{R}^{\featurelen}$. 
\Gls{linreg} aims at finding a parameter vector $\widehat{\vw}$ with minimum 
average squared error loss incurred on a \gls{trainset}
$$\dataset = \big\{ \big(\featurevec^{(1)},\truelabel^{(1)}\big),\ldots,\big(\featurevec^{(\samplesize)},\truelabel^{(\samplesize)}\big)  \big\}.$$ 

The \gls{trainset} $\dataset$ consists of $\samplesize$ \gls{datapoint}s $\big(\featurevec^{(\sampleidx)},\truelabel^{(\sampleidx)}\big)$, 
for $\sampleidx=1,\ldots,\samplesize$, with known label values $\truelabel^{(\sampleidx)}$. 
We stack the feature vectors $\featurevec^{(\sampleidx)}$ and labels $\truelabel^{(\sampleidx)}$, respectively, 
of the \gls{datapoint}s in the \gls{trainset} into the feature matrix $\featuremtx=(\featurevec^{(1)},\ldots,\featurevec^{(\samplesize)})^{T}$ 
and label vector $\labelvec=(\truelabel^{(1)},\ldots,\truelabel^{(\samplesize)})^{T}$. 

The \gls{erm} \eqref{equ_emp_risk_lin_proje} of \gls{linreg} is solved by any parameter vector $\widehat{\weights}$ 
that solves \eqref{equ_zero_gradient_lin_reg}. The (minimum) \gls{trainerr} of the hypothesis $h^{(\widehat{\weights})}$ 
is obtained as 
\begin{align}
	\emperror(h^{(\widehat{\weights})} \mid \dataset) & \stackrel{\eqref{eq_def_ERM_weight}}{=} 
	\min_{\weights \in \mathbb{R}^{\featuredim}} \emperror(h^{(\weights)} | \dataset) \nonumber \\ 
	& \stackrel{\eqref{equ_emp_risk_lin_proje}}{=} \sqeuclnorm{ (\mathbf{I}- \mathbf{P}) \labelvec }.
\end{align} 
Here, we used the orthogonal projection matrix $\mathbf{P}$ on the \index{linear span}linear span 
\begin{equation} 
	\nonumber
	{\rm span}\{ \featuremtx \} = \big\{  \featuremtx \va : \va \in \mathbb{R}^{\featuredim} \big\} \subseteq \mathbb{R}^{\samplesize} , 
\end{equation}
of the feature matrix $\featuremtx = (\featurevec^{(1)},\ldots,\featurevec^{(\samplesize)})^{T} \in \mathbb{R}^{  \samplesize \times \featuredim}$. 

In many ML applications we have access to a huge number of individual features to characterize a \gls{datapoint}. 
As a point in case, consider a \gls{datapoint} which is a snapshot obtained from a modern smartphone camera. These 
cameras have a resolution of several megapixels. Here, we can use millions of pixel colour intensities as its features. 
For such applications, it is common to have more features for \gls{datapoint}s than the size of the \gls{trainset},  
\begin{equation} 
	\label{equ_condition_overfitting}
	\featuredim  \geq \samplesize. 
\end{equation} 
Whenever \eqref{equ_condition_overfitting} holds, the feature vectors $\featurevec^{(1)},\ldots,\featurevec^{(\samplesize)} \in \mathbb{R}^{\featuredim}$ 
of the \gls{datapoint}s in $\dataset$ are typically linearly independent. As a case in point, if the feature vectors 
$\featurevec^{(1)},\ldots,\featurevec^{(\samplesize)} \in \mathbb{R}^{\featuredim}$ are realizations of \gls{iid} \gls{rv}s with a 
continuous \gls{probdist}, these vectors are linearly independent with probability one \cite{Muirhead1982}. 

If the feature vectors $\featurevec^{(1)},\ldots,\featurevec^{(\samplesize)} \in \mathbb{R}^{\featuredim}$ 
are linearly independent, the span of the feature matrix $\featuremtx =  (\featurevec^{(1)},\ldots,\featurevec^{(\samplesize)})^{T}$ 
coincides with $\mathbb{R}^{\samplesize}$ which implies, in turn, $\mathbf{P} = \mathbf{I}$. 
Inserting $\mathbf{P} = \mathbf{I}$ into \eqref{equ_emp_risk_lin_proje} yields 
\begin{equation}
	\label{eq_zero_trianing_error}
	\emperror(h^{(\widehat{\weights})} \mid \dataset) = 0.
\end{equation} 
As soon as the number $\samplesize= | \dataset|$ of training \gls{datapoint}s does 
not exceed the number $\featuredim$ of features that characterize \gls{datapoint}s, there 
is (with probability one) a linear predictor $h^{(\widehat{\weights})}$ achieving zero \gls{trainerr}(!).  

\begin{figure}[htbp]
	\centering
	\includegraphics[width=\textwidth]{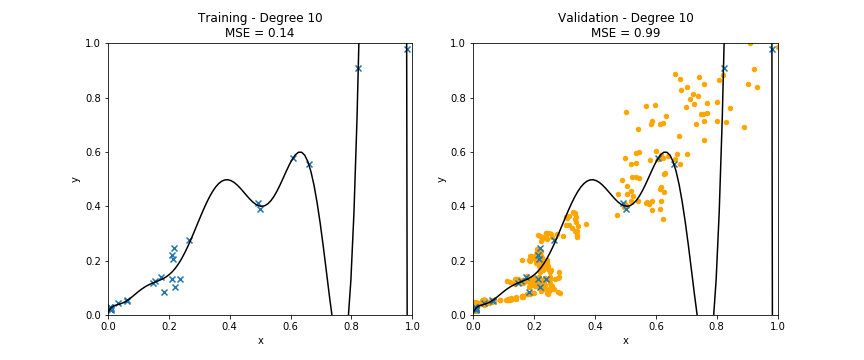}  
	\caption{Polynomial regression learns a polynomial map with degree $\featurelen-1$ 
		by minimizing its average loss on a \gls{trainset} (blue crosses). Using high-degree polynomials 
		(large $\featurelen$) results in a small \gls{trainerr}. However, the learnt high-degree polynomial 
		performs poorly on \gls{datapoint}s outside the \gls{trainset} (orange dots).}
	\label{fig_polyn_training}
\end{figure}

While the hypothesis $h^{(\widehat{\weights})}$ achieves zero \gls{trainerr}, it will typically incur a 
non-zero average prediction error $\truelabel - h^{(\widehat{\weights})}(\featurevec)$ on \gls{datapoint}s $(\featurevec,\truelabel)$ 
outside the \gls{trainset} (see Figure \ref{fig_polyn_training}). Section \ref{sec_gen_linreg} 
will make this statement more precise by using a probabilistic model for the \gls{datapoint}s 
within and outside the \gls{trainset}. 

Note that \eqref{eq_zero_trianing_error} also applies if the features $\featurevec$ and labels $y$ 
of \gls{datapoint}s are completely unrelated. Consider an ML problem with  
\gls{datapoint}s whose labels $\truelabel$ and features are realizations of a \gls{rv} that are statistically 
independent. Thus, in a very strong sense, the features $\featurevec$ contain no information 
about the label of a \gls{datapoint}. Nevertheless, as soon as the number of features exceeds 
the size of the \gls{trainset}, such that \eqref{equ_condition_overfitting} holds, \gls{linreg} methods 
will learn a hypothesis with zero \gls{trainerr}. 

We can easily extend the above discussion about the occurrence of overfitting in linear 
regression to other methods that combine \gls{linreg} with a feature map. 
Polynomial regression, using \gls{datapoint}s with a single feature $z$, combines linear 
regression with the \gls{featuremap} $\rawfeature \mapsto \featuremapvec(\rawfeature) \defeq \big(\rawfeature^{0},\ldots,\rawfeature^{\featurelen-1}\big)^{T}$ 
as discussed in Section \ref{sec_polynomial_regression}. 

It can be shown that whenever \eqref{equ_condition_overfitting} holds and the features 
$\rawfeature^{(1)},\ldots,\rawfeature^{(\samplesize)}$ of the \gls{trainset} are all different, 
the feature vectors $\featurevec^{(1)}\defeq \featuremapvec \big(\rawfeature^{(1)}\big),\ldots, \featurevec^{(\samplesize)}\defeq \featuremapvec \big(\rawfeature^{(\samplesize)}\big)$ are \index{linearly independent}linearly independent. 
This implies, in turn, that polynomial regression is guaranteed to find a hypothesis with zero \gls{trainerr} 
whenever $\samplesize \leq \featurelen$ and the \gls{datapoint}s in the \gls{trainset} have different feature values. 

\section{Validation}
\label{sec_validate_predictor}

\begin{figure}
	\begin{center}
		\begin{tikzpicture}[scale=0.9]
			\draw [thick] (-1,5) rectangle ++(0.1cm,0.1cm) node[anchor=west,above] {\hspace*{0mm}$\big(\featurevec^{(1)},\truelabel^{(1)}\big)$};
			\draw [thick] (1,3.5) rectangle ++(0.1cm,0.1cm) node[anchor=west,above] {\hspace*{0mm}$\big(\featurevec^{(2)},\truelabel^{(2)}\big)$};
			\draw [thick] (3,2.5) rectangle ++(0.1cm,0.1cm) node[anchor=west,above] {\hspace*{0mm}$\big(\featurevec^{(3)},\truelabel^{(3)}\big)$};
			\draw [dashed] {[rounded corners] (0,7) -- (6,2)  -- (2,0) -- (-3,5) -- (0,7) };
			\node[anchor=east,above] at (2,1) {$\trainset$};
			\draw [thick] (7,4) rectangle ++(0.1cm,0.1cm) node[anchor=west,above] {\hspace*{0mm}$\big(\featurevec^{(4)},\truelabel^{(4)}\big)$};
			\draw [thick] (6,6) rectangle ++(0.1cm,0.1cm) node[anchor=west,above] {\hspace*{0mm}$\big(\featurevec^{(5)},\truelabel^{(5)}\big)$};
			\node[anchor=east,above] at (5,5) {$\valset$};
			\draw [dashed] {[rounded corners] (4,5) -- (5,8)  -- (7,8) -- (9,3) -- (4,5) };
		\end{tikzpicture}
	\end{center}
	\caption{We split the dataset $\dataset$ into two subsets, a \gls{trainset} $\trainset$ and a \gls{valset}
		$\valset$. We use the \gls{trainset} to learn (find) the hypothesis $\hat{h}$ with minimum 
		\gls{emprisk} $\emperror(\hat{h}| \trainset)$ on the \gls{trainset} \eqref{equ_def_ERM_funs}. 
		We then validate $\hat{h}$ by computing its average loss $\emperror(\hat{h}| \dataset^{(\rm val)})$  
		on the \gls{valset} $\valset$. The average loss $\emperror(\hat{h}| \valset)$ obtained on the validation 
		set is the \gls{valerr}. Note that $\hat{h}$ depends on the \gls{trainset} $\trainset$ but is completely 
		independent of the \gls{valset} $\valset$.}
	\label{fig_split_train_val}
\end{figure}

Consider an ML method that uses \gls{erm} \eqref{equ_def_ERM_funs} to learn a hypothesis $\hat{h} \in \hypospace$ 
out of the \gls{hypospace} $\hypospace$. The discussion in Section \ref{sec_overfitting_sec_6} revealed that the 
\gls{trainerr} of a learnt hypothesis $\hat{h}$ can be a poor indicator for the performance of $\hat{h}$ for \gls{datapoint}s 
outside the \gls{trainset}. The hypothesis $\hat{h}$ tends to ``look better'' on the \gls{trainset} over which it has been tuned 
within \gls{erm}.
The basic idea of validating the predictor $\hat{h}$ is simple: 
\begin{itemize} 
	\item first we learn a hypothesis $\hat{h}$ using \gls{erm} on a \gls{trainset} and  
	\item then we compute the average \gls{loss} of $\hat{h}$ on \gls{datapoint}s that do not belong to the \gls{trainset}.
 \end{itemize} 
Thus, validation means to compute the average \gls{loss} of a hypothesis using \gls{datapoint}s that 
have not been used in \gls{erm} to learn that hypothesis. 

Assume we have access to a dataset of $\samplesize$ \gls{datapoint}s, 
$$\dataset = \big\{ \big(\featurevec^{(1)},\truelabel^{(1)}\big),\ldots,\big(\featurevec^{(\samplesize)},\truelabel^{(\samplesize)}\big)  \big\}.$$ 
Each \gls{datapoint} is characterized by a feature vector $\featurevec^{(\sampleidx)}$ and a label $\truelabel^{(\sampleidx)}$. 
Algorithm \ref{alg:validated_ERM} outlines how to learn and validate a hypothesis $h\in \hypospace$ 
by splitting the dataset $\dataset$ into a \gls{trainset} and a \gls{valset}. The random shuffling in step 
\ref{alg_shuffle_step} of Algorithm \ref{alg:validated_ERM} ensures  the \gls{iidasspt} for the shuffled data. 
Section \ref{sec_size_val_set} shows next how the \gls{iidasspt} ensures that the \gls{valerr} \eqref{equ_def_training_val_val} 
approximates the expected \gls{loss} of the hypothesis $\hat{h}$. The hypothesis $\hat{h}$ is learnt via \gls{erm} on the \gls{trainset} 
during step \ref{equ_step_train_val_ERM} of Algorithm \ref{alg:validated_ERM}. 

\begin{algorithm}[htbp]
\caption{Validated \gls{erm}}\label{alg:validated_ERM}
\begin{algorithmic}[1]
	\renewcommand{\algorithmicrequire}{\textbf{Input:}}
	\renewcommand{\algorithmicensure}{\textbf{Output:}}
	\Require   model $\hypospace$, loss function $\lossfun$, dataset $\dataset=\big\{ \big(\featurevec^{(1)}, \truelabel^{(1)}\big),\ldots,\big(\featurevec^{(\samplesize)}, \truelabel^{(\samplesize)}\big) \big\}$; split ratio $\splitratio$
	\State randomly shuffle the \gls{datapoint}s in $\dataset$ \label{alg_shuffle_step}
	\State create the \gls{trainset} $\trainset$ using the first $\samplesize_{t}\!=\! \lceil\splitratio \samplesize\rceil$ \gls{datapoint}s,
	$$\trainset = \big\{ \big(\featurevec^{(1)}, \truelabel^{(1)}\big),\ldots,\big(\featurevec^{(\samplesize_{t})}, \truelabel^{(\samplesize_{t})}\big) \big\}.$$
	\State create the \gls{valset} $\valset$ by the $\samplesize_v = \samplesize - \samplesize_t$ remaining \gls{datapoint}s, 
	$$\valset = \big\{ \big(\featurevec^{(\samplesize_{t}+1)}, \truelabel^{(\samplesize_{t}+1)}\big),\ldots,\big(\featurevec^{(\samplesize)}, \truelabel^{(\samplesize)}\big) \big\}.$$
	\State  \label{equ_step_train_val_ERM} learn hypothesis $\hat{h}$ via \gls{erm} on the \gls{trainset}, 
	\begin{equation} 
		\label{equ_def_hat_h_fitting}
		\hat{h} \defeq \argmin_{h\in \hypospace} \emperror\big(h| \trainset \big)
	\end{equation} 
	\State compute the \gls{trainerr}
	\begin{equation} 
		\label{equ_def_training_error_val}
		\trainerror \defeq \emperror\big(\hat{h}| \trainset \big) = (1/\samplesize_{t}) \sum_{\sampleidx=1}^{\samplesize_{t}} \loss{(\featurevec^{(\sampleidx)},\truelabel^{(\sampleidx)})}{\hat{h}}. 
	\end{equation} 
	\State compute the \gls{valerr}
	\begin{equation} 
		\label{equ_def_training_val_val}
		\valerror \defeq \emperror\big(\hat{h}| \valset \big)=   (1/\samplesize_{v}) \sum_{\sampleidx=\samplesize_{t}+1}^{\samplesize} \loss{(\featurevec^{(\sampleidx)},\truelabel^{(\sampleidx)})}{\hat{h}}. 
	\end{equation} 
	\Ensure learnt hypothesis $\hat{h}$, \gls{trainerr} $\trainerror$, \gls{valerr} $\valerror$
\end{algorithmic}
\end{algorithm}


\subsection{The Size of the Validation Set}
\label{sec_size_val_set}

The choice of the split ratio $\splitratio \approx \samplesize_{t}/ \samplesize$ in Algorithm \ref{alg:validated_ERM} 
is often based on trial and error. We try out different choices for the split ratio and pick 
the one with the smallest \gls{valerr}. It is difficult to make a precise statement on how 
to choose the split ratio which applies broadly \cite{Larsen1999}. This difficulty stems 
from the fact that the optimal choice for $\rho$ depends on the precise statistical properties 
of the \gls{datapoint}s. 

One approach to determine the required size of the \gls{valset} is to use a probabilistic model for the \gls{datapoint}s. 
The \gls{iidasspt} is maybe the most widely used probabilistic model within ML. Here, we interpret \gls{datapoint}s 
as the realizations of \gls{iid} \gls{rv}s. These \gls{iid} \gls{rv}s have a common (joint) \gls{probdist} 
$p(\featurevec,\truelabel)$ over possible features $\featurevec$ and labels $\truelabel$ of a \gls{datapoint}.  
Under the \gls{iidasspt}, the \gls{valerr} $\valerror$ \eqref{equ_def_training_val_val} also becomes a realization 
of a \gls{rv}. The expectation (or mean) $\expect \{ \valerror \}$ of this \gls{rv} is precisely the \gls{risk} $\expect\{ \loss{(\featurevec,\truelabel)} {\hat{h}} \}$ of $\hat{h}$ (see \eqref{equ_def_risk}). 

Within the above \gls{iidasspt}, the \gls{valerr} $\valerror$ becomes a realization of a \gls{rv} that 
fluctuates around its mean $\expect \{ \valerror \}$. We can quantify this fluctuation using the variance 
$$\sigma_{\valerror}^{2} \defeq \expect \big\{ \big( \valerror -\expect \{ \valerror \}\big)^{2} \big\}.$$ 
Note that the validation error is the average of the realizations $\loss{(\featurevec^{(\sampleidx)},\truelabel^{(\sampleidx)})}{\hat{h}}$ 
of \gls{iid} \gls{rv}s. The \gls{probdist} of the \gls{rv} $\loss{(\featurevec,\truelabel)}{\hat{h}}$ is 
determined by the \gls{probdist} $p(\featurevec,\truelabel)$, the choice of loss function and the 
hypothesis $\hat{h}$. In general, we do not know $p(\featurevec,\truelabel)$ and, in turn, also do not know 
the \gls{probdist} of $\loss{(\featurevec,\truelabel)}{\hat{h}}$. 

If we know an upper bound $U$ on the variance of the (random) loss $\loss{(\featurevec^{(\sampleidx)},\truelabel^{(\sampleidx)})}{\hat{h}}$, 
we can bound the variance of $\valerror$ as $$ \sigma_{\valerror}^{2}  \leq U/\samplesize_{v}.$$ We can then, in turn, 
ensure that the variance $\sigma_{\valerror}^{2}$ of the validation error $\valerror$ does not exceed a given threshold 
$\eta$, say $\eta = (1/100) \trainerror^2$, by using a \gls{valset} of size 
\begin{equation} 
\label{equ_lower_bound_variance}
\samplesize_{v} \geq U/ \eta. 
\end{equation} 

The lower bound \eqref{equ_lower_bound_variance} is only useful if we can determine an upper 
bound $U$ on the variance of the \gls{rv} $\loss{(\featurevec,\truelabel)}{\hat{h}}$ where $\big(\featurevec,\truelabel\big)$ is a \gls{rv} 
with \gls{probdist} $p(\featurevec,\truelabel)$. An upper bound on the variance of $\loss{(\featurevec,\truelabel)}{\hat{h}}$ 
can be derived using probability theory if we know an accurate probabilistic model $p(\featurevec,\truelabel)$ 
for the \gls{datapoint}s. Such a probabilistic model might be provided by application-specific scientific 
fields such as biology or psychology. Another option is to estimate the variance of $\loss{(\featurevec,\truelabel)}{\hat{h}}$ 
using the sample variance of the actual loss values $\loss{(\featurevec^{(1)},\truelabel^{(1)})}{\hat{h}},\ldots, \loss{(\featurevec^{(\samplesize)},\truelabel^{(\samplesize)})}{\hat{h}}$ 
obtained for the dataset $\dataset$. 

\subsection{$k$-Fold Cross Validation}

Algorithm \ref{alg:validated_ERM} uses the most basic form of splitting a given dataset $\dataset$ into a \gls{trainset} 
and a \gls{valset}. Many variations and extensions of this basic splitting approach have been proposed 
and studied (see \cite{Efron97} and Section \ref{sec_the_bootsrap}). One very popular extension of the single 
split into \gls{trainset} and \gls{valset} is known as \gls{kCV} \cite[Sec. 7.10]{hastie01statisticallearning}. 
We summarize \gls{kCV} in Algorithm \ref{alg:kfoldCV_ERM} below.

Figure \ref{fig_k_fold_CV} illustrates the key principle behind \gls{kCV}. First, we divide the entire dataset 
evenly into $\nrfolds$ subsets which are referred to as ``folds''. The learning (via \gls{erm}) and validation 
of a hypothesis out of a given \gls{hypospace} $\hypospace$ is then repeated $\nrfolds$ times. During each 
repetition, we use one fold as the \gls{valset} and the remaining $\nrfolds-1$ folds as a \gls{trainset}. We then 
average the values of the \gls{trainerr} and \gls{valerr} obtained for each repetition (fold).

The average (over all $\nrfolds$ folds) \gls{valerr} delivered by \gls{kCV} tends to better estimate 
the expected loss or risk \eqref{equ_def_risk} compared to the \gls{valerr} obtained from a single 
split in Algorithm \ref{alg:validated_ERM}. Consider a dataset that consists of a relatively small number 
of \gls{datapoint}s. If we use a single split of this small dataset into a \gls{trainset} and \gls{valset}, we 
might be very unlucky and choose \gls{datapoint}s for the \gls{valset} which are \gls{outlier}s 
and not representative for the statistical properties of most \gls{datapoint}s. The effect of such 
an unlucky split is typically averaged out when using \gls{kCV}. 

\begin{figure}
\begin{center}
	\begin{tikzpicture}
		\matrix (M) [matrix of nodes,
		nodes={minimum height = 7mm, minimum width = 2.5cm, outer sep=0, anchor=center, draw},
		column 1/.style={nodes={draw=none}, minimum width = 4cm},
		row sep=1mm, column sep=-\pgflinewidth, nodes in empty cells,
		e/.style={fill=yellow!10}
		]
		{
			fold $1$ & $\valset\!=\!\dataset_{1}$ & & & & \\
			fold $2$ & & $\valset\!=\!\dataset_{2}$ & & & \\
			fold $3$ & & & $\valset\!=\!\dataset_{3}$ & & \\
			fold $4$ & & & &  $\valset\!=\!\dataset_{4}$ & \\
			fold $5$ & & & & &  $\valset\!=\!\dataset_{5}$ \\
		};
		\draw (M-1-2.north west) ++(0,2mm) coordinate (LT) edge[|<->|, >= latex] node[above]{dataset $\dataset=\big\{ \big(\featurevec^{(1)},\truelabel^{(1)}\big),\ldots, \big(\featurevec^{(\samplesize)},\truelabel^{(\samplesize)}\big) \big\}$} (LT-|M-1-6.north east);
	\end{tikzpicture}
\end{center}
\caption{Illustration of $\nrfolds$-fold CV for $\nrfolds=5$. We evenly partition the entire 
	dataset $\dataset$ into $\nrfolds=5$ subsets (or folds) $\dataset_{1},\ldots,\dataset_{5}$. We then 
	repeat the validated \gls{erm} Algorithm \ref{alg:validated_ERM} for $\nrfolds=5$ times. The $\foldidx$th 
	repetition uses the $\foldidx$th fold $\dataset_{\foldidx}$ as the \gls{valset} and the remaining 
	$\nrfolds\!-\!1(=4)$ folds as the \gls{trainset} for \gls{erm} \eqref{equ_def_ERM_funs}.}
\label{fig_k_fold_CV}
\end{figure}

\begin{algorithm}[htbp]
\caption{\gls{kCV} \gls{erm}}\label{alg:kfoldCV_ERM}
\begin{algorithmic}[1]
	\renewcommand{\algorithmicrequire}{\textbf{Input:}}
	\renewcommand{\algorithmicensure}{\textbf{Output:}}
	\Require   model $\hypospace$, loss function $\lossfun$, dataset $\dataset=\big\{ \big(\featurevec^{(1)}, \truelabel^{(1)}\big),\ldots,\big(\featurevec^{(\samplesize)}, \truelabel^{(\samplesize)}\big) \big\}$; number $\nrfolds$ of folds
	\State randomly shuffle the \gls{datapoint}s in $\dataset$ \label{alg_shuffle_step_kCV}
	\State divide the shuffled dataset $\dataset$ into $\nrfolds$ folds $\dataset_{1},\ldots,\dataset_{\nrfolds}$ 
	            of size $\foldsize=\lceil\samplesize/\nrfolds\rceil$, 
	\begin{equation}
		\hspace*{-12mm}\dataset_{1}\!=\!\big\{ \big(\featurevec^{(1)}, \truelabel^{(1)}\big),\ldots, \big(\featurevec^{(\foldsize)}, \truelabel^{(\foldsize)}\big)\} ,\ldots,\dataset_{k}\!=\!\big\{ \big(\featurevec^{((\nrfolds\!-\!1)\foldsize+1)}, \truelabel^{((\nrfolds\!-\!1)\foldsize+1)}\big),\ldots, \big(\featurevec^{(\samplesize)}, \truelabel^{(\samplesize)}\big)\}
	\end{equation} 
	\For{ fold index $\foldidx=1,\ldots,\nrfolds$ }
	\State use $\foldidx$th fold as the validation set $\valset=\dataset_{\foldidx}$
	\State use rest as the \gls{trainset} $\trainset=\dataset \setminus \dataset_{\foldidx}$
	\State  learn hypothesis $\hat{h}$ via \gls{erm} on the \gls{trainset}, 
	\begin{equation} 
		\label{equ_def_hat_h_fitting_cv}
		\hat{h}^{(\foldidx)} \defeq \argmin_{h\in \hypospace} \emperror\big(h| \trainset \big)
	\end{equation} 
	\State compute the \gls{trainerr}
	\begin{equation} 
		\label{equ_def_training_error_val_cv}
		\trainerror^{(\foldidx)} \defeq \emperror\big(\hat{h}| \trainset \big) = (1/\big|\trainset\big|) \sum_{\sampleidx \in \trainset} \loss{(\featurevec^{(\sampleidx)},\truelabel^{(\sampleidx)})}{h}. 
	\end{equation} 
	\State compute {\bf validation error}
	\begin{equation} 
		\label{equ_def_training_val_val_cv}
		\valerror^{(\foldidx)} \defeq \emperror\big(\hat{h}| \valset \big)=   (1/\big|\valset\big|) \sum_{\sampleidx \in \valset} \loss{(\featurevec^{(\sampleidx)},\truelabel^{(\sampleidx)})}{\hat{h}}. 
	\end{equation} 
	\EndFor
	
	\State compute average training and validation errors 
	$$\trainerror \defeq (1/\nrfolds) \sum_{\foldidx=1}^{\nrfolds} 	\trainerror^{(\foldidx)}\mbox{, and }\valerror \defeq (1/\nrfolds) \sum_{\foldidx=1}^{\nrfolds} 	\valerror^{(\foldidx)}$$
	\State pick a learnt hypothesis $\hat{h} \defeq \hat{h}^{(\foldidx)}$ for some $\foldidx \in \{1,\ldots,\nrfolds\}$
	\Ensure learnt hypothesis $\hat{h}$; average \gls{trainerr} $\trainerror$; average validation error $\valerror$
\end{algorithmic}
\end{algorithm}

\subsection{Imbalanced Data} 

The simple validation approach discussed above requires the validation set to be 
a good representative for the overall statistical properties of the data. This might 
not be the case in applications with discrete valued labels and some of the label 
values being very rare. We might then be interested in having a good estimate of 
the conditional risks $\expect \{ \loss{(\featurevec,\truelabel)}{h} | \truelabel=\truelabel'\}$ where $\truelabel'$ is one of the rare 
label values. This is more than requiring a good estimate for the risk $\expect \{ \loss{(\featurevec,\truelabel)}{h} \}$. 

Consider \gls{datapoint}s characterized by a feature vector $\featurevec$ and binary label $\truelabel \in \{-1,1\}$. 
Assume we aim at learning a hypothesis $h(\featurevec) = \weights^{T} \featurevec$ to classify \gls{datapoint}s 
as $\hat{\truelabel}=1$ if $h(\featurevec) \geq 0$ while $\hat{\truelabel}=-1$ otherwise. The learning is based 
on a dataset $\dataset$ which contains only one single (!) \gls{datapoint} with $\truelabel=-1$. If 
we then split the dataset into training and validation set, it is with high probability that 
the validation set does not include any \gls{datapoint} with label value $\truelabel=-1$. This cannot happen 
when using $\nrfolds$-fold CV since the single \gls{datapoint} must be in one of the validation folds. 
However, even the applicability of \gls{kCV} for such an imbalanced dataset is limited since 
we evaluate the performance of a hypothesis $h(\featurevec)$ using only one single 
\gls{datapoint} with $\truelabel=-1$. The resulting \gls{valerr} will be dominated by the \gls{loss}  
of $h(\featurevec)$ incurred on \gls{datapoint}s from the majority class (those with true label value $\truelabel=1$). 

To learn and validate a hypothesis with imbalanced data, it might be useful to to generate synthetic 
\gls{datapoint}s to enlarge the minority class. This can be done using \gls{dataug} techniques which 
we discuss in Section \ref{sec_data_augmentation}. Another option is to choose a \gls{lossfunc} that 
takes the different frequencies of label values into account. Let us illustrate this approach in what 
follows by an illustrative example. 

Consider an imbalanced dataset of size $\samplesize=100$, which contains $90$ \gls{datapoint}s with 
label $\truelabel=1$ but only $10$ \gls{datapoint}s with label $\truelabel=-1$. We might want to 
put more weight on wrong predictions obtained for \gls{datapoint}s from the minority class (with true label 
value $\truelabel=-1$). This can be done by using a much larger value for the \gls{loss} $\loss{(\featurevec,\truelabel=-1)}{h(\featurevec)=1}$ 
than for the \gls{loss} $\loss{(\featurevec,\truelabel=1)}{h(\featurevec)=-1}$ incurred by incorrectly 
predicting the label of a \gls{datapoint} from the majority class (with true label value $\truelabel=1$).

\section{Model Selection}
\label{sec_modsel}

Chapter \ref{ch_some_examples} illustrated how many well-known ML methods are obtained 
by different combinations of a \gls{hypospace} or \gls{model}, \gls{lossfunc} and data representation. 
While for many ML applications there is often a natural choice for the \gls{lossfunc} and data 
representation, the right choice for the \gls{model} is typically less obvious. We now discuss how 
to use the validation methods of Section \ref{sec_validate_predictor} to choose between different 
candidate models. 

Consider \gls{datapoint}s characterized by a single numeric feature $\feature\in \mathbb{R}$ and numeric label $\truelabel\in \mathbb{R}$. 
If we suspect that the relation between feature $\feature$ and label $\truelabel$ is non-linear, we might use 
polynomial regression which is discussed in Section \ref{sec_polynomial_regression}. Polynomial 
regression uses the \gls{hypospace} $\hypospace_{\rm poly}^{(\featuredim)}$ with some maximum 
degree $\featuredim$. Different choices for the maximum degree $\featuredim$ yield a different 
\gls{hypospace}: $\hypospace^{(1)} = \mathcal{H}_{\rm poly}^{(0)},\hypospace^{(2)} = \mathcal{H}_{\rm poly}^{(1)},\ldots,\hypospace^{(\nrmodels)} = \hypospace_{\rm poly}^{(\nrmodels)}$. 

Another ML method that learns non-linear hypothesis map is Gaussian basis regression (see Section \ref{sec_linbasreg}). 
Here, different choices for the variance $\sigma$ and shifts $\mu$ of the Gaussian basis function \eqref{equ_basis_Gaussian} 
result in different \gls{hypospace}s. For example, $\hypospace^{(1)} = \mathcal{H}^{(2)}_{\rm Gauss}$ with $\sigma=1$ and $\mu_{1}=1$ and $\mu_{2}=2$, $\hypospace^{(2)} = \hypospace^{(2)}_{\rm Gauss}$ with $\sigma = 1/10$, $\mu_{1}=10$, $\mu_{2}= 20$.

Algorithm \ref{alg:model_selection} summarizes a simple method to choose between different candidate 
models $\hypospace^{(1)},\hypospace^{(2)},\ldots,\hypospace^{(\nrmodels)}$. The idea is to first learn 
and validate a hypothesis $\hat{h}^{(\modelidx)}$ separately for each model $\hypospace^{(\modelidx)}$ 
using Algorithm \ref{alg:kfoldCV_ERM}. For each model $\hypospace^{(\modelidx)}$, we learn the hypothesis 
$\hat{h}^{(\modelidx)}$ via \gls{erm} \eqref{equ_def_hat_h_fitting} and then compute its \gls{valerr}
$\valerror^{(\modelidx)}$ \eqref{equ_def_training_val_val}. We then choose the hypothesis $\hat{h}^{(\hat{\modelidx})}$ 
from those model $\hypospace^{(\hat{\modelidx})}$ which resulted in the smallest \gls{valerr} $\valerror^{(\hat{\modelidx})} = \min_{\modelidx=1,\ldots,\nrmodels} \valerror^{(\modelidx)}$. 

The workflow of Algorithm \ref{alg:model_selection} is similar to the workflow of \gls{erm}. 
Remember that the idea of \gls{erm} is to learn a hypothesis out of a set of different candidates (the \gls{hypospace}). 
The quality of a particular hypothesis $h$ is measured using the (average) loss incurred on some \gls{trainset}. 
We use the same principle for model selection but on a higher level. Instead of learning a hypothesis 
within a \gls{hypospace}, we choose (or learn) a \gls{hypospace} within a set of candidate 
\gls{hypospace}s. The quality of a given \gls{hypospace} is measured by the validation error \eqref{equ_def_training_val_val}. 
To determine the validation error of a \gls{hypospace}, we first learn the hypothesis $\hat{h} \in \hypospace$ via  
\gls{erm} \eqref{equ_def_hat_h_fitting} on the \gls{trainset}. Then, we obtain the \gls{valerr} as the average loss of 
$\hat{h}$ on the \gls{valset}. 

The final hypothesis $\hat{h}$ delivered by the model selection Algorithm \ref{alg:model_selection} not 
only depends on the \gls{trainset} used in \gls{erm} (see \eqref{equ_def_hat_h_fitting_cv}). This hypothesis $\hat{h}$ has 
also been chosen based on its validation error which is the average loss on the \gls{valset} in \eqref{equ_def_training_val_val_cv}. 
Indeed, we compared this validation error with the \gls{valerr}s of other models to pick the 
model $\hypospace^{(\hat{\modelidx})}$ (see step \ref{step_pick_optimal_model}) which contains $\hat{h}$. 
Since we used the \gls{valerr} \eqref{equ_def_training_val_val_cv} of $\hat{h}$ to learn it, we 
cannot use this \gls{valerr} as a good indicator for the general performance of $\hat{h}$. 

To estimate the general performance of the final hypothesis $\hat{h}$ delivered by Algorithm \ref{alg:model_selection} we 
must try it out on a test set. The test set, which is constructed in step \ref{equ_construct_test_set_algmodsel} 
of Algorithm \ref{alg:model_selection}, consists of \gls{datapoint}s that are neither contained  
in the \gls{trainset} \eqref{equ_def_hat_h_fitting_cv} nor the \gls{valset} \eqref{equ_def_training_val_val_cv} 
used for training and validating the candidate models $\hypospace^{(1)},\ldots,\hypospace^{(\nrmodels)}$. 
The average loss of the final hypothesis on the test set is referred to as the test error. The test error is computed 
in the step \ref{step_compute_test_error_mod_selection} of Algorithm \ref{alg:model_selection}.

\begin{algorithm}[htbp]
\caption{Model Selection}\label{alg:model_selection}
\begin{algorithmic}[1]
	\renewcommand{\algorithmicrequire}{\textbf{Input:}}
	\renewcommand{\algorithmicensure}{\textbf{Output:}}
	\Require   list of candidate models $\hypospace^{(1)},\ldots,\hypospace^{(\nrmodels)}$, 
	loss function $\lossfun$, dataset $\dataset=\big\{ \big(\featurevec^{(1)}, \truelabel^{(1)}\big),\ldots,\big(\featurevec^{(\samplesize)}, \truelabel^{(\samplesize)}\big) \big\}$; 
	number $\nrfolds$ of folds, \gls{testset} fraction $\rho$
	\State randomly shuffle the \gls{datapoint}s in $\dataset$ \label{alg_shuffle_step_model-sel}
	\State determine size $\samplesize' \defeq \lceil \rho \samplesize \rceil$ of test set
	\State construct a \gls{testset} \label{step_construct_test_set_mo_sel}
	$$\testset= \big\{\big(\featurevec^{(1)}, \truelabel^{(1)}\big),\ldots,\big(\featurevec^{(\samplesize')}, \truelabel^{(\samplesize')}\big) \big\}$$ \label{equ_construct_test_set_algmodsel}
	\State construct a \gls{trainset} and a \gls{valset}, 
	$$\dataset^{(\rm trainval)} = \big\{\big(\featurevec^{(\samplesize'+1)}, \truelabel^{(\samplesize'+1)}\big),\ldots,\big(\featurevec^{(\samplesize)}, \truelabel^{(\samplesize)}\big) \big\}$$
	\For{ model index $\modelidx=1,\ldots,\nrmodels$ }
	\State run Algorithm \ref{alg:kfoldCV_ERM} using $\hypospace=\hypospace^{(\modelidx)}$, dataset $\dataset=\dataset^{(\rm trainval)}$, 
	\gls{lossfunc} $\lossfun$ 
	and $\nrfolds$ folds
	\State Algorithm \ref{alg:kfoldCV_ERM} delivers hypothesis $\hat{h}$ and \gls{valerr} $\valerror$
	\State store learnt hypothesis $\hat{h}^{(\modelidx)}\defeq\hat{h}$ and \gls{valerr} $\valerror^{(\modelidx)}\defeq\valerror$
	
	\EndFor
	
	\State pick model $\hypospace^{(\hat{\modelidx})}$ with minimum validation error 
	$\valerror^{(\hat{\modelidx})}\!=\!\min_{\modelidx=1,\ldots,\nrmodels}\valerror^{(\modelidx)}$ \label{step_pick_optimal_model}
	
	\State define optimal hypothesis $\hat{h}= \hat{h}^{(\hat{\modelidx})}$ 
	
	\State compute {\bf test error}  \label{step_compute_test_error_mod_selection}
	\begin{equation} 
		\label{equ_def_training_error_val_test}
		\testerror \defeq \emperror\big(\hat{h}| \testset \big) = (1/\big|\testset\big|) \sum_{\sampleidx \in \testset} \loss{(\featurevec^{(\sampleidx)},\truelabel^{(\sampleidx)})}{\hat{h}}. 
	\end{equation}

	\Ensure  hypothesis $\hat{h}$; \gls{trainerr} $\trainerror^{(\hat{\modelidx})}$; validation error $\valerror^{(\hat{\modelidx})}$, 
	test error $\testerror$. 
\end{algorithmic}
\end{algorithm}

Sometimes it is beneficial to use different loss functions for the training and the validation of a hypothesis. 
As an example, consider \gls{logreg} and the \gls{svm} which have been discussed in Sections \ref{sec_LogReg} and 
\ref{sec_SVM}, respectively. Both methods use the same model which is the space of linear hypothesis 
maps $h(\featurevec) = \weights^{T} \featurevec$. The main difference between these two methods is 
in their choice for the \gls{lossfunc}. 
\Gls{logreg} minimizes the (average) \gls{logloss} \eqref{equ_log_loss} on the \gls{trainset} to learn the 
hypothesis $h^{(1)}(\featurevec)= \big( \weights^{(1)} \big)^{T} \featurevec $ with a parameter vector $\weights^{(1)}$. 
The \gls{svm} instead minimizes the (average) \gls{hingeloss} \eqref{equ_hinge_loss} 
on the \gls{trainset} to learn the hypothesis $h^{(2)}(\featurevec) = \big( \weights^{(2)} \big)^{T} \featurevec$ with a parameter 
vector $\weights^{(2)}$. It is inconvenient to compare the usefulness of the two hypotheses $h^{(1)}(\featurevec)$ and 
$h^{(2)}(\featurevec)$ using different \gls{lossfunc}s to compute their \gls{valerr}s. This comparison is more convenient if 
we instead compute the \gls{valerr}s for $h^{(1)}(\featurevec)$ and $h^{(2)}(\featurevec)$ using the average $0/1$ 
loss \eqref{equ_def_0_1}. 

Algorithm \ref{alg:model_selection} requires as one of its inputs a given list of candidate models. The 
longer this list, the more computation is required from Algorithm \ref{alg:model_selection}. Sometimes 
it is possible to prune the list of candidate models by removing models that are very unlikely to have 
minimum validation error. 

Consider polynomial regression which uses as the model the space $\hypospace_{\rm poly}^{(r)}$ 
of polynomials with maximum degree $\polydegree$ (see \eqref{equ_def_poly_hyposapce}). For $\polydegree=1$, 
$\hypospace_{\rm poly}^{(\polydegree)}$ is the space of polynomials with maximum degree one (which are linear maps), $h(\feature)=\weight_{2}\feature+\weight_{1}$. For $\polydegree=2$, $\hypospace_{\rm poly}^{(\polydegree)}$ is the 
space of polynomials with maximum degree two, $h(\feature)=\weight_{3}\feature^2 +\weight_{2}\feature+\weight_{1}.$

The polynomial degree $\polydegree$ parametrizes a nested set of models, 
$$\hypospace_{\rm poly}^{(\polydegree)} \subset \hypospace_{\rm poly}^{(\polydegree)} \subset \ldots.$$
For each degree $\polydegree$, we learn a hypothesis $h^{(r)} \in \hypospace_{\rm poly}^{(\polydegree)}$ with minimum average 
loss (\gls{trainerr}) $\trainerror^{(\polydegree)}$ on a \gls{trainset} (see \eqref{equ_def_training_error_val}). To validate 
the learnt hypothesis $h^{(\polydegree)}$, we compute its average loss (\gls{valerr}) $\valerror^{(\polydegree)}$ 
on a \gls{valset} (see \eqref{equ_def_training_val_val}). 

\begin{figure}[htbp] 
\centering
\includegraphics[width=0.6\textwidth]{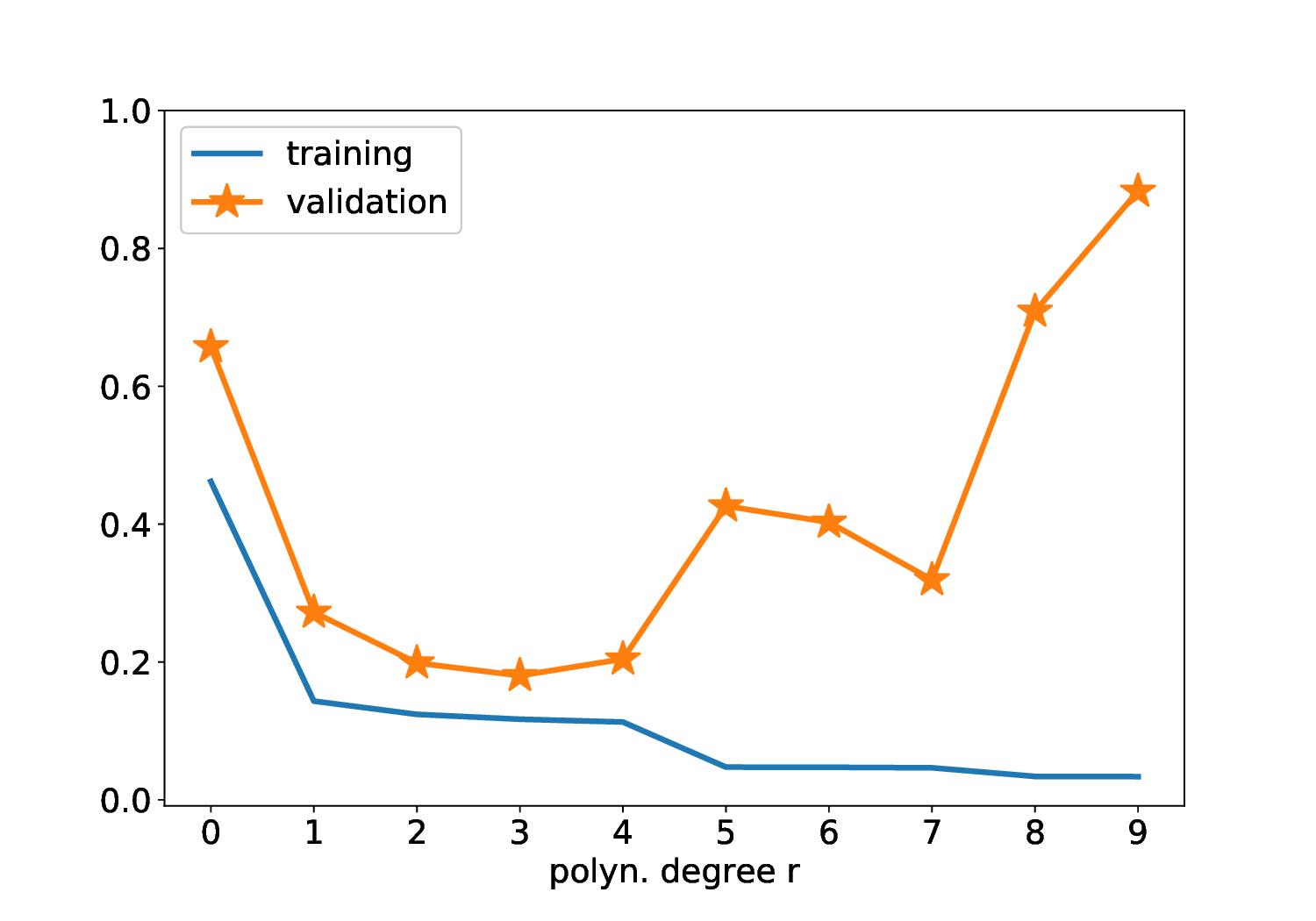}
\caption{The \gls{trainerr} and validation error obtained from polynomial regression using 
	different values $\polydegree$ for the maximum polynomial degree.} 
\label{fig_trainvalvsdegree}
\end{figure}

Figure \ref{fig_trainvalvsdegree} depicts the typical dependency of the training and validation errors on the 
polynomial degree $\polydegree$. The \gls{trainerr} $\trainerror^{(\polydegree)}$ decreases monotonically 
with increasing polynomial degree $\polydegree$. To illustrate this monotonic decrease, we consider the two specific 
choices $\polydegree=3$ and $\polydegree=5$ with corresponding models $\hypospace^{(\polydegree)}_{\rm poly}$ 
and $\hypospace^{(\polydegree)}_{\rm poly}$. Note that $\hypospace^{(3)}_{\rm poly} \subset \hypospace^{(5)}_{\rm poly}$ 
since any polynomial with degree not exceeding $3$ is also a polynomial with degree not exceeding $5$. 
Therefore, the \gls{trainerr} \eqref{equ_def_training_error_val} obtained when minimizing over the larger 
model $ \hypospace^{(5)}_{\rm poly}$ can only decrease but never increase compared to \eqref{equ_def_training_error_val} 
using the smaller model $ \hypospace^{(3)}_{\rm poly}$

Figure \ref{fig_trainvalvsdegree} indicates that the validation error $\valerror^{(\polydegree)}$ (see \eqref{equ_def_training_val_val}) 
behaves very different compared to the \gls{trainerr} $\trainerror^{(\polydegree)}$. Starting with degree $\polydegree=0$, the validation error 
first decreases with increasing degree $\polydegree$. As soon as the degree $\polydegree$ is increased 
beyond a critical value, the validation error starts to increase with increasing $\polydegree$. For very large 
values of $\polydegree$, the \gls{trainerr}  becomes almost negligible while the validation error becomes 
very large. In this regime, polynomial regression overfits the \gls{trainset}. 

\begin{figure}[htbp] 
\centering
\includegraphics[width=0.6\textwidth]{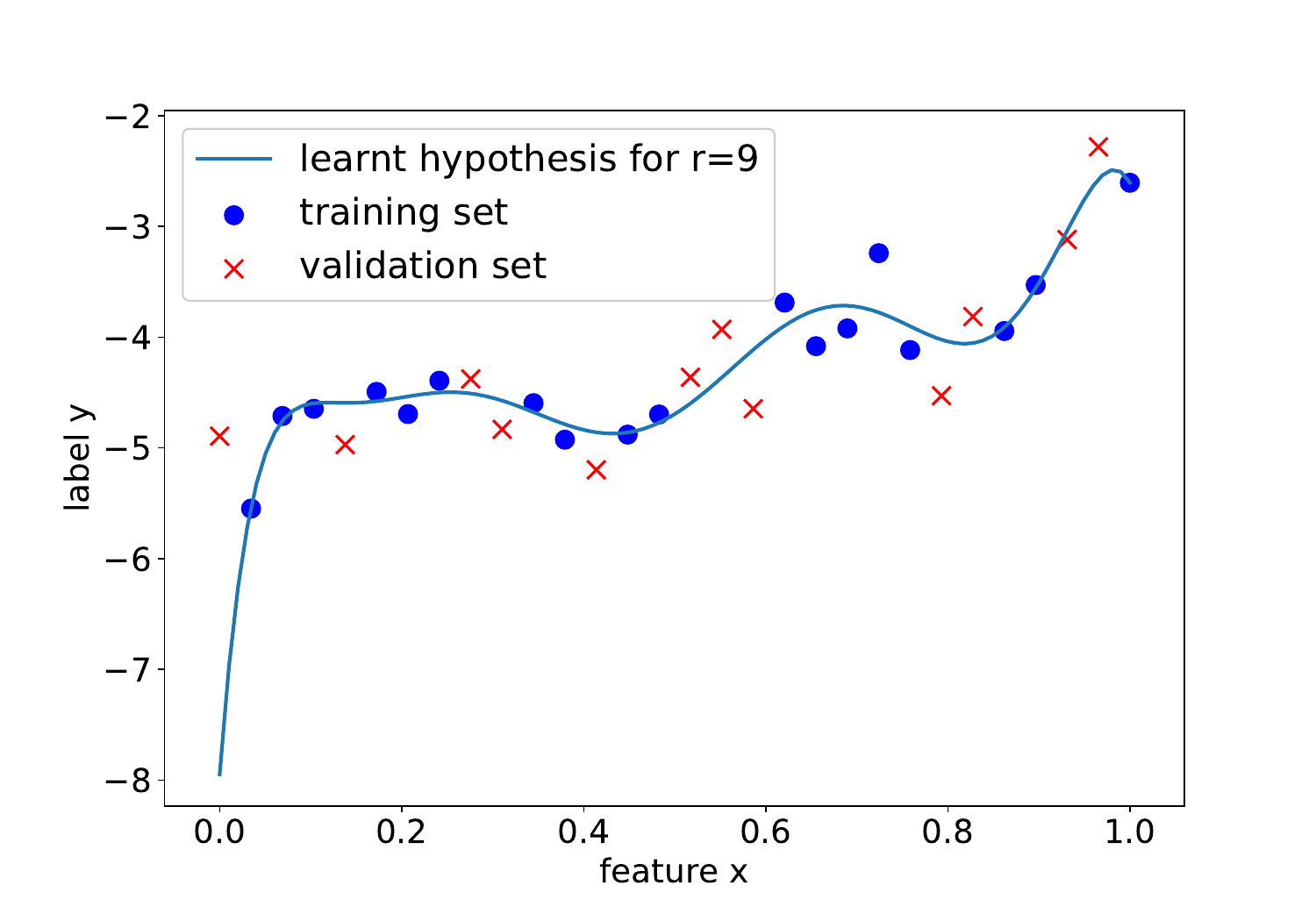}
\caption{A hypothesis $\hat{h}$ which is a polynomial with degree not larger than $\polydegree=9$. The 
	hypothesis has been learnt by minimizing the average loss on the \gls{trainset}. Note the rapid change of $\hat{h}$ for feature 
	values $\feature \approx 0$. } 
\label{fig_polyregdegree9}
\end{figure}

Figure \ref{fig_polyregdegree9} illustrates the overfitting of polynomial regression when using a maximum 
degree that is too large. In particular, Figure \ref{fig_polyregdegree9} depicts a learnt hypothesis which is a degree $9$ 
polynomial that fits very well the \gls{trainset}, resulting in a very small \gls{trainerr}. To achieve this low \gls{trainerr} 
the resulting polynomial has an unreasonable high rate of change for feature values $\feature \approx 0$. This results in 
large prediction errors for \gls{datapoint}s with feature values $x \approx 0$. 

\newpage
\section{A Probabilistic Analysis of Generalization} 
\label{sec_gen_linreg}
\emph{More Data Beats Clever Algorithms ?; More Data Beats Clever Feature Selection?}

A key challenge in ML is to ensure that a hypothesis that predicts well the labels on a \gls{trainset} (which has 
been used to learn that hypothesis) will also predict well the labels of \gls{datapoint}s outside the \gls{trainset}. 
We say that a ML method generalizes well if it learns a hypothesis $\hat{h}$ that performs not significantly 
worse on \gls{datapoint}s outside the \gls{trainset}. In other words, the loss incurred by $\hat{h}$ for 
\gls{datapoint}s outside the \gls{trainset} is not much larger than the average \gls{loss} of $\hat{h}$ incurred 
on the \gls{trainset}. 

We now study the generalization of \gls{linreg} methods (see Section \ref{sec_lin_reg}) using an \gls{iidasspt}. 
In particular, we interpret \gls{datapoint}s as \gls{iid} realizations of \gls{rv}s that have the same distribution as 
a random \gls{datapoint} $\datapoint=(\featurevec,\truelabel)$. The feature vector $\featurevec$ is then a 
realization of a standard Gaussian \gls{rv} with zero mean and covariance 
being the identity matrix, i.e., $\featurevec \sim \mathcal{N}(\mathbf{0}, \mathbf{I})$. 

The label $\truelabel$ of a random \gls{datapoint} is related to its features $\featurevec$ via a linear Gaussian model
\begin{equation} 
\label{equ_linear_obs_model}
\truelabel = \overline{\weights}^{T}  \featurevec + \varepsilon \mbox{, with noise } \varepsilon \sim \mathcal{N}(0,\sigma^{2}).
\end{equation} 
We assume the noise variance $\sigma^{2}$ fixed and known. This is a simplifying assumption and in 
practice we would need to estimate the noise variance from data \cite{Cohen2002}. Note that, within 
our probabilistic model, the error component $\varepsilon$ in \eqref{equ_linear_obs_model} is intrinsic 
to the data and cannot be overcome by any ML method. We highlight that the probabilistic model for 
the observed \gls{datapoint}s is just a modelling assumption. This assumption allows us to study some 
fundamental behaviour of ML methods. There are principled methods (``statistical tests'') that allow to 
determine if a given dataset can be accurately modelled using \eqref{equ_linear_obs_model} \cite{HuberApproxModels}. 

We predict the label $\truelabel$ from the features $\featurevec$ using a linear hypothesis $h(\featurevec)$ 
that depends only on the first $\modelidx$ features $\feature_{1},\ldots,\feature_{\modelidx}$. Thus, we 
use the \gls{hypospace} 
\vspace*{-2mm}
\begin{equation}
\label{equ_generalization_hypospace_r}
\hypospace^{(\modelidx)} = \{ h^{(\weights)}(\featurevec)= (\weights^{T},\mathbf{0}^{T}) \featurevec \mbox{ with } \weights \in \mathbb{R}^{\modelidx} \}.   
\vspace*{-1mm}
\end{equation}
Note that each element $h^{(\weights)} \in \hypospace^{(\modelidx)}$ corresponds to a particular 
choice of the parameter vector $\weights \in \mathbb{R}^{\modelidx}$. 

The model parameter $\modelidx \in \{0,\ldots,\featuredim\}$ coincides with the \gls{effdim} of 
the \gls{hypospace} $\hypospace^{(\modelidx)}$. For $\modelidx< \featuredim$, the \gls{hypospace} $\hypospace^{(\modelidx)}$ 
is a proper (strict) subset of the space of linear hypothesis maps \eqref{equ_def_hypo_linear_pred} 
used within \gls{linreg} (see Section \ref{sec_lin_reg}). Moreover, the parameter $\modelidx$ indexes 
a nested sequence of models, 
\begin{equation}
\hypospace^{(0)} \subseteq \hypospace^{(1)} \subseteq \ldots \subseteq \hypospace^{(\featuredim)}.  \nonumber
\end{equation} 

The quality of a particular predictor $h^{(\weights)} \in \hypospace^{(\modelidx)}$ is measured via the 
average squared error $\emperror (h^{(\weights)} \mid \trainset)$ incurred on the labeled \gls{trainset}
\begin{equation} 
\label{equ_def_train_set_prob_analysis_generatlization}
\trainset= \{ \big(\featurevec^{(1)}, \truelabel^{(1)}\big), \ldots, \big(\featurevec^{(\samplesize_{t})}, \truelabel^{(\samplesize_{t})}\big)  \}. 
\end{equation} 
We interpret \gls{datapoint}s in the \gls{trainset} $\trainset$ as well as any other \gls{datapoint} 
outside the \gls{trainset} as realizations of \gls{iid} \gls{rv}s with a common \gls{probdist}. This 
common \gls{probdist} is a multivariate normal (Gaussian) distribution, 
\begin{equation} 
\label{equ_toy_model_iid}
\featurevec, \featurevec^{(\sampleidx)} \mbox{ \gls{iid} with } \featurevec, \featurevec^{(\sampleidx)} \sim \mathcal{N}(\mathbf{0}, \mathbf{I}). 
\end{equation} 
The labels $\truelabel^{(\sampleidx)},\truelabel$ are related to the features of \gls{datapoint}s via (see \eqref{equ_linear_obs_model})
\begin{equation} 
\label{equ_labels_training_data}
\truelabel^{(\sampleidx)} = \overline{\weights}^{T}  \featurevec^{(\sampleidx)} + \varepsilon^{(\sampleidx)}\mbox{, and } \truelabel = \overline{\weights}^{T} \featurevec + \varepsilon. 
\end{equation}  
Here, the noise terms $\varepsilon, \varepsilon^{(\sampleidx)} \sim \mathcal{N}(0,\sigma^{2})$ are realizations 
of \gls{iid} Gaussian \gls{rv}s with zero mean and variance $\sigma^{2}$. 

Chapter \ref{ch_Optimization} showed that the \gls{trainerr} $\emperror (h^{(\weights)} \mid \trainset)$ is 
minimized by the predictor $h^{(\widehat{\weights})}(\featurevec) =  \widehat{\weights}^{T} \mathbf{I}_{\modelidx \times \featuredim} \featurevec$, 
that uses the parameter vector 
\begin{equation}
\label{equ_optimal_weight_closed_form}
\widehat{\weights} =  \big(\big(\featuremtx^{(\modelidx)}\big)^{T} \featuremtx^{(\modelidx)} \big)^{-1} \big(\featuremtx^{(\modelidx)}\big)^{T} \labelvec. 
\end{equation} 
Here we used the (restricted) feature matrix $\mX^{(\modelidx)}$ and the label vector $\labelvec$  defined as, respectively, 
\begin{align}
\label{equ_def_feature_matrix_r}
\mX^{(\modelidx)}& \!=\!(\featurevec^{(1)},\ldots,\featurevec^{(\samplesize_{\rm t})})^{T} \mathbf{I}_{\featuredim \times \modelidx}\!\in\!\mathbb{R}^{\samplesize_{\rm t} \times \modelidx} \mbox{, and }  \nonumber \\
\vy& \!=\!\big(\truelabel^{(1)},\ldots,\truelabel^{(\samplesize_{\rm t})}\big)^{T}\!\in\!\mathbb{R}^{\samplesize_{\rm t}}.
\end{align} 

It will be convenient to tolerate a slight abuse of notation and denote both, 
the length-$\modelidx$ vector \eqref{equ_optimal_weight_closed_form} as well as the 
zero-padded parameter vector $\big(\widehat{\weights}^{T},\mathbf{0}^{T}\big)^{T} \in \mathbb{R}^{\featuredim}$, 
by $\widehat{\weights}$. This allows us to write 
\begin{equation} 
h^{(\widehat{\weights})}(\featurevec) = \widehat{\weights}^{T} \featurevec. 
\end{equation}

We highlight that the formula \eqref{equ_optimal_weight_closed_form} for the optimal weight 
vector $\widehat{\weights}$ is only valid if the matrix $\big(\featuremtx^{(\modelidx)}\big)^{T} \featuremtx^{(\modelidx)}$ 
is invertible. Within our toy model (see \eqref{equ_toy_model_iid}), this is true with probability one 
whenever $\samplesize_{\rm t} \geq \modelidx$. Indeed, for $\samplesize_{\rm t} \geq \modelidx$ the 
truncated feature vectors $\mathbf{I}_{\modelidx \times \featuredim} \featurevec^{(1)}, \ldots, \mathbf{I}_{\modelidx \times \featuredim} \featurevec^{(\samplesize_{t})}$, 
which are \gls{iid} realizations of a Gaussian \gls{rv}, are linearly independent with probability one \cite{BertsekasProb,Gallager13}. 

In what follows, we consider the case $\samplesize_{\rm t} > \modelidx$ such that the 
formula \eqref{equ_optimal_weight_closed_form} is valid (with probability one). The more 
challenging \gls{highdimregime} $\samplesize_{\rm t} \leq \modelidx$ will be 
studied in Chapter \ref{ch_overfitting_regularization}.

The optimal parameter vector $\widehat{\weights}$ (see \eqref{equ_optimal_weight_closed_form}) 
depends on the \gls{trainset} $\trainset$ via the feature matrix $\featuremtx^{(\modelidx)}$ 
and label vector $\labelvec$ (see \eqref{equ_def_feature_matrix_r}). Therefore, 
since we model the \gls{datapoint}s in the \gls{trainset} as realizations of \gls{rv}s, the parameter vector $\widehat{\weights}$ 
\eqref{equ_optimal_weight_closed_form} is the realization of a \gls{rv}. For each specific  
realization of the \gls{trainset} $\trainset$, we obtain a specific realization of the 
optimal parameter vector $\widehat{\weights}$. 

The probabilistic model \eqref{equ_linear_obs_model} relates the features 
$\featurevec$ of a \gls{datapoint} to its label $\truelabel$ via some (unknown) true parameter vector $\overline{\weights}$. 
Intuitively, the best linear hypothesis would be $h(\featurevec) =\widehat{\weights}^{T} \featurevec$ with 
parameter vector $\widehat{\weights} = \overline{\weights}$. However, in general this will not be 
achievable since we have to compute $\widehat{\weights}$ based on the features 
$\featurevec^{(\sampleidx)}$ and noisy labels $\truelabel^{(\sampleidx)}$ of the \gls{datapoint}s in the \gls{trainset} $\dataset$. 

The parameter vector $\widehat{\vw}$ delivered by \gls{erm} \eqref{equ_def_cost_MSE} typically 
results in a non-zero \gls{esterr}
\begin{equation}
\label{equ_def_est_error}
\Delta \weights \defeq \widehat{\weights} - \overline{\weights}. 
\end{equation} 
The \gls{esterr} \eqref{equ_def_est_error} is the realization of a \gls{rv} since the learnt 
parameter vector $\widehat{\weights}$ (see \eqref{equ_optimal_weight_closed_form}) is itself a 
realization of a \gls{rv}.

{\bf The Bias and Variance Decomposition.} 
The prediction accuracy of $h^{(\widehat{\weights})}$, using the learnt parameter vector \eqref{equ_optimal_weight_closed_form}, 
depends crucially on the \index{mean squared estimation error} \gls{msee}
\begin{equation}
\label{equ_def_est_err}
\mseesterr \defeq \expect \{  \sqeuclnorm{\Delta \weights } \} \stackrel{\eqref{equ_def_est_error}}{=}  
\expect \big\{ \sqeuclnorm{\widehat{\weights} - \overline{\weights}} \big \}. 
\end{equation}
We will next decompose the \gls{msee} $\mseesterr$ into two components, which are referred to as 
a \gls{variance} term and a \gls{bias} term. The variance term quantifies the random fluctuations or the 
parameter vector obtained from \gls{erm} on the \gls{trainset} \eqref{equ_def_train_set_prob_analysis_generatlization}. 
The \gls{bias} term characterizes the systematic (or expected) deviation between the true parameter  
vector $\overline{\weights}$ (see \eqref{equ_linear_obs_model}) and the (expectation of the) 
learnt parameter vector $\widehat{\weights}$.
 
Let us start with rewriting \eqref{equ_def_est_err} using elementary manipulations as 
\begin{align}
\mseesterr & \stackrel{\eqref{equ_def_est_err}}{=} \expect \big\{  \sqeuclnorm{\widehat{\weights} - \overline{\weights}} \big \}\big \}  \nonumber \\[2mm]
&=  \expect \bigg\{  \sqeuclnorm{\big( \widehat{\weights} -  \expect \big\{ \widehat{\weights} \big\}\big) - \big( \overline{\weights} - \expect \big\{  \widehat{\weights}  \big\} \big)} \bigg \}.  \nonumber
\end{align}
We can develop the last expression further by expanding the squared Euclidean norm, 
\begin{align}
\mseesterr &=  \expect \big\{ \sqeuclnorm{\widehat{\weights} - \expect \{ \widehat{\weights} \} }  \big\}  - 2 \expect \big \{  \big( \widehat{\weights} -  \expect \big\{ \widehat{\weights} \big\} \big)^{T} \big( \overline{\weights} - \expect \big\{  \widehat{\weights} \big\} \big)  \big\} + \expect \big\{ \sqeuclnorm{\overline{\weights} - \expect \big\{ \widehat{\weights} \big\} } \big\}\nonumber \\[4mm]
&=  \expect \big\{ \sqeuclnorm{\widehat{\weights} - \expect \{ \widehat{\weights} \} }  \big\}  - 2  \big( \underbrace{\expect \big \{  \widehat{\weights}  \big\} -  \expect \big\{ \widehat{\weights} \big\}}_{=\mathbf{0}} \big)^{T} \big( \overline{\weights} - \expect \big\{  \widehat{\weights} \big\} \big)  \big\} + \expect \big\{ \sqeuclnorm{\overline{\weights} - \expect \big\{ \widehat{\weights} \big\} } \big\}\nonumber \\[4mm]
 &=   \underbrace{ \expect \big\{  \sqeuclnorm{ \widehat{\weights} - \expect \{ \widehat{\weights} \} } \big\} }_{\mbox{\small{\gls{variance} } }\varianceterm}  
 + \underbrace{ \expect \big\{  \sqeuclnorm{ \overline{\weights} - \expect \{ \widehat{\weights}   \}  } \big\} }_{\mbox{\small{\gls{bias} } } \biasterm^2}. \label{equ_bias_var_decomp}
\end{align} 
The first component in \eqref{equ_bias_var_decomp} represents the (expected) \gls{variance} of the 
learnt parameter vector $\widehat{\weights}$ \eqref{equ_optimal_weight_closed_form}.
Note that, within our probabilistic model, the \gls{trainset} \eqref{equ_def_train_set_prob_analysis_generatlization} is 
the realization of a \gls{rv} since it is constituted by \gls{datapoint}s that are \gls{iid} realizations of \gls{rv}s 
(see \eqref{equ_toy_model_iid} and \eqref{equ_linear_obs_model}). 

The second component in \eqref{equ_bias_var_decomp} is referred to as a \index{bias} \gls{bias} term. 
The parameter vector $\widehat{\weights}$ is computed from a randomly fluctuating \gls{trainset} via \eqref{equ_optimal_weight_closed_form} 
and is therefore itself fluctuating around its expectation $\expect \big \{\widehat{\weights}\}$. The \gls{bias} term 
is the Euclidean distance between this expectation $\expect \big \{\widehat{\weights}\}$ and the true parameter 
vector $\overline{\weights}$ relating features and label of a \gls{datapoint} via \eqref{equ_linear_obs_model}. 

The \gls{bias} term $\biasterm^{2}$ and the \gls{variance} $\varianceterm$ in \eqref{equ_bias_var_decomp} 
both depend on the model complexity parameter $\modelidx$ but in a fundamentally different manner. 
The bias term $\biasterm^{2}$ typically decreases with increasing $\modelidx$ while the variance $\varianceterm$ 
increases with increasing $\modelidx$. 
In particular, the \gls{bias} term is given as
\vspace*{-3mm}
\begin{equation}
\label{equ_def_bias_term}
\biasterm^{2} = \sqeuclnorm{\overline{\weights} - \expect \{ \widehat{\weights} \} } = \sum_{\featureidx=\modelidx+1}^{\featuredim} \overline{\weight}_{\featureidx}^2, 
\end{equation} 
The \gls{bias} term \eqref{equ_def_bias_term} is zero if and only if 
\begin{equation}
	\label{equ_def_zero_bias_weght_elements}
\overline{\weight}_{\featureidx}=0 \mbox{ for any index } \featureidx=\modelidx+1,\ldots,\featuredim. 
\end{equation} 

The necessary and sufficient condition \eqref{equ_def_zero_bias_weght_elements} 
for zero bias is equivalent to $h^{(\overline{\weights})} \in \hypospace^{(\modelidx)}$. 
Note that the condition \eqref{equ_def_zero_bias_weght_elements} depends on both, 
the model parameter $\modelidx$ and the true parameter vector $\overline{\weights}$. 
While the model parameter $\modelidx$ is under control, the true parameter vector $\overline{\weights}$ 
is not under our control but determined by the underlying data generation process. 
The only way to ensure \eqref{equ_def_zero_bias_weght_elements} for every possible 
parameter vector $\overline{\weights}$ in \eqref{equ_linear_obs_model} is to use $\modelidx=\featuredim$, i.e., 
to use all available features $\feature_{1},\ldots,\feature_{\featuredim}$ of a \gls{datapoint}. 

When using the model $\hypospace^{(\modelidx)}$ with $\modelidx < \featuredim$, we cannot guarantee 
a zero \gls{bias} term since we have no control over the true underlying parameter vector $\overline{\weights}$ 
in \eqref{equ_linear_obs_model}. In general, the \gls{bias} term decreases with an 
increasing model size $\modelidx$ (see Figure \ref{fig_bias_variance}). We highlight that 
the \gls{bias} term does not depend on the variance $\sigma^{2}$ of the noise $\varepsilon$ 
in our toy model \eqref{equ_linear_obs_model}. 

Let us now consider the \gls{variance} term in \eqref{equ_bias_var_decomp}. Using the statistical 
independence of the features and labels of \gls{datapoint}s (see \eqref{equ_linear_obs_model}, 
\eqref{equ_toy_model_iid} and \eqref{equ_labels_training_data}), one can show that\footnote{This derivation is 
	not very difficult but rather lengthy. For more details about the derivation of \eqref{equ_variance_term_toy_model} 
	we refer to the literature \cite{BertsekasProb,Luetkepol2005}.}
\begin{equation}
\label{equ_variance_term_toy_model}
\varianceterm = \expect \big\{ \sqeuclnorm{\widehat{\weights} - \expect\{ \widehat{\weights} \} } \big\} = \big( \biasterm^2+ \sigma^{2}\big) \mbox{tr} \left\{ \expect \left\{\big( \big(\featuremtx^{(\modelidx)}\big)^{T} \featuremtx^{(\modelidx)} \big)^{-1} \right\} \right\}.
\end{equation} 
By \eqref{equ_toy_model_iid}, the matrix $\left(\big(\featuremtx^{(\modelidx)}\big)^{T} \mX^{(\modelidx)} \right)^{-1}$ is 
a realization of a (matrix-valued) \gls{rv} with an \index{inverse Wishart distribution} inverse Wishart distribution \cite{Mardia1979}. 
For $\samplesize_{\rm t} > \modelidx+1$, its expectation is given as 
\begin{equation} 
\label{equ_expr_expect_inv-wishart}
\expect\{ \big(\big(\featuremtx^{(\modelidx)} \big)^{T} \featuremtx^{(\modelidx)} \big)^{-1} \} = 1/(\samplesize_{\rm t}-\modelidx-1) \mathbf{I}.
\end{equation} 

By inserting \eqref{equ_expr_expect_inv-wishart} and $\mbox{tr} \{ \mathbf{I} \} = \modelidx$ into \eqref{equ_variance_term_toy_model}, 
\begin{equation} 
\label{equ_formulae_variance_toy_model}
 \varianceterm = \expect \left\{ \sqeuclnorm{\widehat{\weights} - \expect\left\{ \widehat{\weights} \right\}} \right\} = \big(  \biasterm^{2} +\sigma^{2}\big) \modelidx/(\samplesize_{\rm t}-\modelidx-1). 
\end{equation} 
The \gls{variance} \eqref{equ_formulae_variance_toy_model} typically increases with increasing 
model complexity $\modelidx$ (see Figure \ref{fig_bias_variance}). In contrast, the \gls{bias} term 
\eqref{equ_def_bias_term} decreases with increasing $\modelidx$. 

The opposite dependence of \gls{variance} and \gls{bias} on the model complexity results in a 
\index{bias-variance tradeoff} \gls{bias}-\gls{variance} trade-off. Choosing a model (\gls{hypospace}) 
with small \gls{bias} will typically result in large \gls{variance} and vice versa. In general, the 
choice of model must balance between a small \gls{variance} and a small \gls{bias}. 

\begin{figure}
\begin{center}
	\begin{tikzpicture}
		\draw [thick, <->] (1,3.5) -- (1,0) -- (10.2,0);<
		\draw[red, ultra thick, domain=1:10] plot (\x,  {3*\x^-1});
		\draw[blue, ultra thick, domain=1:10] plot (\x,  {3*\x/10});
		\draw[ultra thick, domain=1:10,dashed] plot (\x,  {3*\x^(-1)+3*\x/10});
		\node [right, color=red] at (10.2,0.3) {\gls{bias}};
		\node [right,color=blue] at (5.2,1.4) {\gls{variance}};
		\node [below] at (5,0) {model complexity $\modelidx$};
		\node [above] at (5,2.2) {$\mseesterr$};
	\end{tikzpicture}
\end{center}
\caption{The \gls{msee} $\mseesterr$ incurred by \gls{linreg} can be decomposed into a 
	\gls{bias} term $\biasterm^{2}$ and a \gls{variance} term $\varianceterm$ (see \eqref{equ_bias_var_decomp}). 
	These two components depend on the model complexity $\modelidx$ in an opposite 
	manner which results in a \gls{bias}-\gls{variance} trade-off.}
\label{fig_bias_variance}
\end{figure}

{\bf Generalization.}
Consider a \gls{linreg} method that learns the linear hypothesis $h(\featurevec) = \widehat{\weights}^{T} \featurevec$ using 
the parameter vector \eqref{equ_optimal_weight_closed_form}. The parameter vector $\widehat{\weights}^{T}$ \eqref{equ_optimal_weight_closed_form} 
results in a linear hypothesis with minimum \gls{trainerr}, i.e., minimum average loss on the \gls{trainset}. 
However, the ultimate goal of ML is to find a hypothesis that predicts well the label of any \gls{datapoint}. 
In particular, we want the hypothesis $h(\featurevec) = \widehat{\weights}^{T} \featurevec$ to generalize 
well to \gls{datapoint}s outside the \gls{trainset}. 

We quantify the generalization capability of $h(\featurevec) = \widehat{\weights}^{T} \featurevec$  by 
its expected prediction loss 
\begin{equation} 
\label{equ_def_expected_pred_loss} 
\error_{\rm pred} = \expect \big\{ \big( \truelabel - \underbrace{\widehat{\weights}^{T} \featurevec}_{ = \hat{\truelabel}} \big)^2 \big\}. 
\end{equation} 
Note that $\error_{\rm pred}$ is a measure for the performance of a ML method and not of a specific hypothesis. 
Indeed, the learnt parameter vector $\widehat{\weights}$ is not fixed but depends on the \gls{datapoint}s in the \gls{trainset}. These 
\gls{datapoint}s are modelled as realizations of \gls{iid} \gls{rv}s and, in turn, the learnt parameter vector $\widehat{\weights}$ 
becomes a realization of a \gls{rv}. Thus, in some sense, the expected prediction loss \eqref{equ_def_expected_pred_loss} characterizes 
the overall ML method that reads in a \gls{trainset} and delivers (learn) a linear hypothesis with parameter vector $\widehat{\weights}$ \eqref{equ_optimal_weight_closed_form}. In contrast, the \gls{risk} \eqref{equ_def_risk} introduced in 
Chapter \ref{ch_Optimization} characterizes the performance of a specific (fixed) hypothesis $h$ 
without taking into account a learning process that delivered $h$ based on \gls{data}. 

Let us now relate the expected prediction loss \eqref{equ_def_expected_pred_loss} of the linear 
hypothesis $h(\featurevec) = \widehat{\weights}^{T} \featurevec$ to the \gls{bias} and \gls{variance} 
of \eqref{equ_optimal_weight_closed_form}, 
\begin{align} 
\label{equ_decomp_E_pred_toy_model}
\error_{\rm pred} 
& \stackrel{\eqref{equ_linear_obs_model}}{=} \expect \{ \Delta \weights^{T} \featurevec \featurevec^{T} \Delta \weights \} + \sigma^{2}   \nonumber \\
& \stackrel{(a)}{=} \expect \{ \expect \{ \Delta \weights^{T} \featurevec \featurevec^{T} \Delta \weights \mid \trainset \} \} + \sigma^{2}  \nonumber \\
& \stackrel{(b)}{=} \expect \{ \Delta \weights^{T} \Delta \weights \}  + \sigma^{2}  \nonumber \\
& \stackrel{\eqref{equ_def_est_error},\eqref{equ_def_est_err}}{=} \mseesterr + \sigma^{2} \nonumber \\
& \stackrel{\eqref{equ_bias_var_decomp}}{=} \biasterm^{2} + \varianceterm + \sigma^{2}. 
\end{align} 
Here, step (a) uses the law of iterated expectation (see, e.g., \cite{BertsekasProb}). Step (b) uses that 
the feature vector $\featurevec$ of a ``new'' \gls{datapoint} is a realization of a \gls{rv} which is 
statistically independent of the \gls{datapoint}s in the \gls{trainset} $\trainset$. We also used 
our assumption that $\featurevec$ is the realization of a \gls{rv} with zero mean and covariance 
matrix $\expect \{ \featurevec \featurevec^{T}\}=\mathbf{I}$ (see \eqref{equ_toy_model_iid}). 

According to \eqref{equ_decomp_E_pred_toy_model}, the average (expected) prediction error 
$\error_{\rm pred}$ is the sum of three components: (i) the bias $\biasterm^{2}$, (ii) the 
variance $\varianceterm$ and (iii) the noise variance $\sigma^{2}$. Figure \ref{fig_bias_variance} illustrates 
the typical dependency of the bias and variance on the model \eqref{equ_generalization_hypospace_r}, 
which is parametrized by the model complexity $\modelidx$. Note that the model complexity parameter $\modelidx$ 
in \eqref{equ_generalization_hypospace_r} coincides with the effective model dimension $\effdim{\hypospace^{(\modelidx)}}$ (see Section \ref{sec_hypo_space}). 

The bias and variance, whose sum is the estimation error $\error_{\rm est}$, can be influenced by 
varying the model complexity $\modelidx$ which is a design parameter. The noise variance $\sigma^{2}$ is the 
intrinsic accuracy limit of our toy model \eqref{equ_linear_obs_model} and is not under the control of 
the ML engineer. It is impossible for any ML method (no matter how computationally expensive) 
to achieve, on average, a prediction error smaller than the noise variance $\sigma^{2}$. Carefully 
note that this statement only applies if the \gls{datapoint}s arising in a ML application can be (reasonably well) 
modelled as realizations of \gls{iid} \gls{rv}s.   

We highlight that our statistical analysis, resulting the formulas for bias \eqref{equ_def_bias_term}, \gls{variance} \eqref{equ_formulae_variance_toy_model} 
and the average prediction error \eqref{equ_decomp_E_pred_toy_model}, applies only if the observed \gls{datapoint}s can be 
well modelled using the probabilistic model specified by \eqref{equ_linear_obs_model}, \eqref{equ_toy_model_iid} 
and \eqref{equ_labels_training_data}. The validity of this probabilistic model can to be verified by principled 
statistical model validation techniques \cite{Young93,Vasicek76}. Section \ref{sec_the_bootsrap} discusses a fundamentally different approach 
to analyzing the statistical properties of a ML method. Instead of a probabilistic model, this approach uses random sampling 
techniques to synthesize \gls{iid} copies of given (small) \gls{datapoint}s. We can approximate the expectation 
of some relevant quantity, such as the loss $\loss{\big(\featurevec,\truelabel \big) }{h}$, using an average over 
synthetic data \cite{hastie01statisticallearning}.

The qualitative behaviour of estimation error in Figure \ref{fig_bias_variance} depends on the definition for the model complexity. 
Our concept of \gls{effdim} (see Section \ref{sec_hypo_space}) coincides with most other notions of model complexity 
for the linear \gls{hypospace} \eqref{equ_generalization_hypospace_r}. However, for more complicated models 
such as deep nets it is often not obvious how \gls{effdim} is related to more tangible quantities such 
as total number of tunable \gls{weights} or the number of artificial neurons. Indeed, the \gls{effdim} might also depend 
on the specific learning algorithm such as \gls{stochGD}. Therefore, for deep nets, if we would plot estimation error 
against number of tunable \gls{weights} we might observe a behaviour of estimation error fundamentally different from 
the shape in Figure \ref{fig_bias_variance}. One example for such un-intuitive behaviour is known as ``double descent phenomena'' \cite{Belkin15849}.

\section{The Bootstrap} 
\label{sec_the_bootsrap}
\emph{basic idea of bootstrap: use \gls{histogram} of dataset  
as the underlying \gls{probdist}; generate new \gls{datapoint}s by 
random sampling (with replacement) from that distribution. }

Consider learning a hypothesis $\hat{h} \in \hypospace$ by minimizing the 
average loss incurred on a \gls{dataset} $\dataset=\{ \big(\featurevec^{(1)},\truelabel^{(1)}\big),\ldots,\big(\featurevec^{(\samplesize)},\truelabel^{(\samplesize)}\big)\}$. 
The \gls{datapoint}s $\big(\featurevec^{(\sampleidx))},\truelabel^{(\sampleidx)}\big)$ are modelled 
as realizations of \gls{iid} \gls{rv}s. Let use denote the (common) probability 
distribution of these \gls{rv}s by $p(\featurevec,\truelabel)$. 

If we interpret the \gls{datapoint}s $\big(\featurevec^{(\sampleidx))},\truelabel^{(\sampleidx)}\big)$ as realizations of 
\gls{rv}s, also the learnt hypothesis $\hat{h}$ is a realization  
of a \gls{rv}. Indeed, the hypothesis $\hat{h}$ is obtained by 
solving an optimization problem \eqref{equ_def_ERM_funs} that involves 
realizations of \gls{rv}s. The bootstrap is a method for estimating 
(parameters of) the \gls{probdist} $p(\hat{h})$ \cite{hastie01statisticallearning}.  

Section \ref{sec_gen_linreg} used a probabilistic model for \gls{datapoint}s to derive (the parameters of) 
the \gls{probdist} $p(\hat{h})$. Note that the analysis in Section \ref{sec_gen_linreg} only applies to 
the specific probabilistic model \eqref{equ_toy_model_iid}, \eqref{equ_labels_training_data}. In contrast, 
the \index{bootstrap}\gls{bootstrap} can be used for \gls{datapoint}s drawn from an arbitrary \gls{probdist}. 

The core idea behind the \gls{bootstrap} is to use the \gls{histogram} $\hat{p}(\datapoint)$ 
of the \gls{datapoint}s in $\dataset$ to generate $\nrbootstraps$ new datasets $\dataset^{(1)},\ldots,\dataset^{(\nrbootstraps)}$. 
Each dataset is constructed such that is has the same size as the original dataset $\dataset$. 
For each dataset $\dataset^{(\bootstrapidx)}$, we solve a separate \gls{erm} \eqref{equ_def_ERM_funs} 
to obtain the hypothesis $\hat{h}^{(\bootstrapidx)}$. The hypothesis $\hat{h}^{(\bootstrapidx)}$ is a realization 
of a \gls{rv} whose distribution is determined by the \gls{histogram} $\hat{p}(\datapoint)$ 
as well as the \gls{hypospace} and the \gls{lossfunc} used in the \gls{erm} \eqref{equ_def_ERM_funs}.

\section{Diagnosing ML} 
\label{sec_diagnosis_ML}

\emph{diagnose ML methods by comparing \gls{trainerr} with \gls{valerr} and (if available) some \gls{baseline}; 
	\gls{baseline} can be obtained via the \gls{bayesrisk} when using a probabilistic model (such as the \gls{iidasspt}) 
	or human performance or the performance of existing ML methods ("experts" in regret framework)  }

In what follows, we tacitly assume that \gls{datapoint}s can (to a good approximation) be 
interpreted as realizations of \gls{iid} \gls{rv}s (see Section \ref{equ_prob_models_data}). 
This ``\gls{iidasspt}'' underlies \gls{erm} \eqref{equ_def_ERM_funs} as the guiding principle 
for learning a hypothesis with small risk \eqref{equ_def_risk}. This assumption also motivates 
to use the average loss \eqref{equ_def_training_val_val} on a \gls{valset} as an estimate for 
the \gls{risk}. More fundamentally, we need the \gls{iidasspt} to define the concept of \gls{risk} 
as a measure for how well a hypothesis predicts the labels of arbitrary \gls{datapoint}s. 

Consider a ML method which uses Algorithm \ref{alg:validated_ERM} (or Algorithm \ref{alg:kfoldCV_ERM}) 
to learn and validate the hypothesis $\hat{h} \in \hypospace$. Besides the learnt hypothesis 
$\hat{h}$, these algorithms also deliver the \gls{trainerr} $\trainerror$ and the validation 
error $\valerror$. As we will see shortly, we can diagnose ML methods to some extent just 
by comparing training with validation errors. This diagnosis is further enabled if we know a 
\gls{baseline} $\benchmarkerror$ . 

One important source of a \gls{baseline} $\benchmarkerror$ are probabilistic models 
for the \gls{datapoint}s (see Section \ref{sec_gen_linreg}). Given a probabilistic model, which 
specifies the \gls{probdist} $p(\featurevec,\truelabel)$ of the features and label of \gls{datapoint}s, 
we can compute the minimum achievable \gls{risk} \eqref{equ_def_risk}. Indeed, the minimum 
achievable risk is precisely the expected loss of the \gls{bayesestimator} $\hat{h}(\featurevec)$ 
of the label $\truelabel$, given the features $\featurevec$ of a \gls{datapoint}. The \gls{bayesestimator} 
$\hat{h}(\featurevec)$ is fully determined by the \gls{probdist} $p(\featurevec,\truelabel)$ of 
the features and label of a (random) \gls{datapoint} \cite[Chapter 4]{LC}. 

A further potential source for a \gls{baseline} $\benchmarkerror$ is an existing, but for some reason 
unsuitable, ML method. This existing ML method might be computationally too expensive to be used 
for the ML application at end. However, we might still use its statistical properties as a benchmark. 
The 

We might also use the performance of human experts as a \gls{baseline}. 
If we want to develop a ML method that detects certain type of skin cancers from 
images of the skin, a benchmark might be the current classification accuracy 
achieved by experienced dermatologists \cite{Esteva2017}. 

We can diagnose a ML method by comparing the \gls{trainerr} $\trainerror$ 
with the validation error $\valerror$ and (if available) the benchmark $\benchmarkerror$.
\begin{itemize} 
\item $\trainerror \approx \valerror \approx \benchmarkerror$: The \gls{trainerr} is on the same 
level as the validation error and the benchmark error. There is not much to improve here since 
the validation error is already on the desired error level. Moreover, the \gls{trainerr} is not much 
smaller than the validation error which indicates that there is no overfitting. It seems we have 
obtained a ML method that achieves the benchmark error level. 

\item $\valerror \gg \trainerror$: The \gls{valerr} is significantly larger than the \gls{trainerr}. 
It seems that the \gls{erm} \eqref{equ_def_ERM_funs} results in a hypothesis $\hat{h}$ that overfits 
the \gls{trainset}. The loss incurred by $\hat{h}$ on \gls{datapoint}s outside the \gls{trainset}, such as 
those in the \gls{valset}, is significantly worse. This is an indicator for overfitting which can be 
addressed either by reducing the \gls{effdim} of the \gls{hypospace} or by increasing the 
size of the \gls{trainset}. To reduce the \gls{effdim} of the \gls{hypospace} we have different options 
depending on the used \gls{model}. We might use a small number of features in a linear model \eqref{equ_lin_hypospace}, 
a smaller maximum depth of \gls{decisiontree}s (Section \ref{sec_decision_trees}) or a 
fewer layers in an \gls{ann} (Section \ref{sec_deep_learning}). 
One very elegant means for reducing the \gls{effdim} of a \gls{hypospace} is by limiting the 
number of \gls{gd} steps used in \gls{gdmethods}. This optimization based shrinking of a \gls{hypospace} 
is referred to as \index{early stopping}early stopping. More generally, we can reduce the \gls{effdim} 
of a \gls{hypospace} via \gls{regularization} techniques (see Chapter \ref{ch_overfitting_regularization}).

\item $\trainerror \approx \valerror\gg \benchmarkerror$: The \gls{trainerr} is on the same level as the \gls{valerr}
and both are significantly larger than the \gls{baseline}. Since the \gls{trainerr} is not much smaller than the \gls{valerr}, 
the learnt hypothesis seems to not overfit the \gls{trainset}. However, the \gls{trainerr} achieved by the learnt hypothesis 
is significantly larger than the benchmark error level. There can be several reasons for this to happen. First, it might be that 
the \gls{hypospace} used by the ML method is too small, i.e., it does not include a hypothesis that provides a good approximation 
for the relation between features and label of a \gls{datapoint}. The remedy for this situation is to use a larger \gls{hypospace}, e.g., 
by including more features in a linear model, using higher polynomial degrees in polynomial regression, using deeper 
\gls{decisiontree}s or having larger \gls{ann}s (\gls{deepnet}s). Another reason for the \gls{trainerr} being too large is 
that the optimization algorithm used to solve \gls{erm} \eqref{equ_def_ERM_funs} is not working properly. 

When using \gls{gdmethods} (see Section \ref{sec_GD_linear_regression}) to solve \gls{erm}, one 
reason for $\trainerror \gg \benchmarkerror$ could be that the \gls{learnrate} $\lrate$ in the \gls{gd} step \eqref{equ_def_GD_step} 
is chosen too small or too large (see Figure \ref{fig_small_large_lrate}-(b)). This can be solved by adjusting the \gls{learnrate} 
by trying out several different values and using the one resulting in the smallest \gls{trainerr}. Another option is derive 
optimal values for the \gls{learnrate} based on a probabilistic model for how the \gls{datapoint}s are generated. 
One example for such a probabilistic model is the \gls{iidasspt} that has been used in Section \ref{sec_gen_linreg} 
to analyze \gls{linreg} methods. 

\item $\trainerror \gg \valerror $: The \gls{trainerr} is significantly larger than the \gls{valerr} (see Exercise \ref{ex_val_err_smaller_train_err}).
The idea of \gls{erm} \eqref{equ_def_ERM_funs} is to approximate the risk \eqref{equ_def_risk} of a 
hypothesis by its average loss on a \gls{trainset} $\dataset = \{ (\featurevec^{(\sampleidx)},\truelabel^{(\sampleidx)}) \}_{\sampleidx=1}^{\samplesize}$.
The mathematical underpinning for this approximation is the \gls{lln} which characterizes the 
average of (realizations of) \gls{iid} \gls{rv}s. The quality and usefulness of this approximation 
depends on the validity of two conditions. First, the \gls{datapoint}s used for computing the average 
loss should be such that they would be typically obtained as realizations of \gls{iid} \gls{rv}s with a 
common \gls{probdist}. Second, the number of \gls{datapoint}s used for computing the average \gls{loss} 
must be sufficiently large. 

Whenever the \gls{datapoint}s behave different than the the realizations of \gls{iid} \gls{rv}s or 
if the size of the \gls{trainset} or \gls{valset} is too small, the interpretation (comparison) of \gls{trainerr} 
and \gls{valerr}s becomes more difficult. As an extreme case, it might then be that the \gls{valerr} consists of 
\gls{datapoint}s for which every hypothesis incurs small average loss. Here, we might try to increase 
the size of the \gls{valset} by collecting more labeled \gls{datapoint}s or by using \gls{dataug} (see Section \ref{sec_data_augmentation}). 
If the size of \gls{trainset} and \gls{valset} are large but we still obtain $\trainerror \gg \valerror $, one should 
verify if \gls{datapoint}s in these sets conform to the \gls{iidasspt}. There are principled statistical 
methods that allow to test if an \gls{iidasspt} is satisfied (see \cite{Luetkepol2005} and references therein). 
\end{itemize}


\section{Exercises} 

\begin{exercise}[Validation Set Size.] 
\label{ex_val_set_size}
Consider a \gls{linreg} problem with \gls{datapoint}s $(\feature,\truelabel)$ characterized by a scalar 
feature $\feature$ and a numeric label $\truelabel$. Assume \gls{datapoint}s are realizations of \gls{iid} 
\gls{rv}s whose common \gls{probdist} is multivariate normal with zero-mean and covariance 
matrix $\mathbf{C} = \begin{pmatrix} \sigma^2_{\feature} & \sigma_{\feature,\truelabel} \\  \sigma_{\feature,\truelabel} & \sigma^{2}_{\truelabel} \end{pmatrix}$. The entries of this covariance matrix are the variance $\sigma^2_{\feature}$ of the (zero-mean) feature, 
the variance $\sigma^2_{\feature}$ of the (zero-mean) label and the covariance between feature and label of a random 
\gls{datapoint}. How many \gls{datapoint}s do 
we need to include in a \gls{valset} such that with probability of at least $0.8$ the \index{validation error}\gls{valerr} 
of a given hypothesis $h$ does not deviate by more than $20$ percent from its expected loss?
\end{exercise}

\begin{exercise}[Validation Error Smaller Than Training Error?]
\label{ex_val_err_smaller_train_err}
\Gls{linreg} learns a linear hypothesis map $\hat{h}$ having minimal average squared error on 
a \gls{trainset}. The learnt hypothesis $\hat{h}$ is then validated on a \gls{valset} which is 
different from the \gls{trainset}. Can you construct a \gls{trainset} and \gls{valset} such that 
the \gls{valerr} of $\hat{h}$ is strictly smaller than the \gls{trainerr} of $\hat{h}$?
\end{exercise}

\begin{exercise}[When is Validation Set Too Small?]
The usefulness of the \gls{valerr} as an indicator for the performance of a hypothesis 
depends on the size of the \gls{valset}. Experiment with different ML methods and 
datasets to find out the minimum required size for the \gls{valset}.
\end{exercise}  

\begin{exercise}[Too many Features?]
Consider \gls{datapoint}s that are characterized by $\featuredim=1000$ numeric 
features $\feature_{1},\ldots,\feature_{\featuredim} \in \mathbb{R}$ and a numeric 
label $\truelabel \in \mathbb{R}$. We want to learn a linear hypothesis map $h(\featurevec)  = \weights^{T} \featurevec$ 
for predicting the label of a \gls{datapoint} based on its features. Could it be beneficial 
to constrain the learnt hypothesis by requiring it to only depend on the first $5$ features of a \gls{datapoint}? 
\end{exercise}  

\begin{exercise}[Benchmark via Probability Theory.]
Consider \gls{datapoint}s that are characterized with single numeric feature $\feature$ and label $\truelabel$. 
We model the feature and label of a \gls{datapoint} as \gls{iid} realizations of a Gaussian random vector $\datapoint \sim \mathcal{N}(0,\mathbf{C})$ 
with zero mean and covariance matrix $\mathbf{C}$. The optimal hypothesis $\hat{h}(\feature)$ to predict the label $\truelabel$ 
given the feature $\feature$ is the conditional expectation of the (unobserved) label $\truelabel$ given the (observed) feature $\feature$. 
How is the expected squared error loss of this optimal hypothesis (which is the \gls{bayesestimator}) related to the covariance 
matrix $\mathbf{C}$ of the Gaussian random vector $\datapoint$. 
\end{exercise}

\chapter{Regularization}
\label{ch_overfitting_regularization}
\emph{Keywords: Data Augmentation. Robustness. Semi-Supervised Learning. Transfer Learning. Multitask Learning.}
\vspace*{-3mm}
\begin{figure}[htbp]
	\begin{center}
		\begin{tikzpicture}[auto,scale=1.3]
			\draw[line width=0.4mm,->] (0,-0.5) -- (0,3.5) node[above] {label $\truelabel$};
			\draw[line width=0.4mm,->] (-0.5,0) -- (5,0) node[right] {feature $\feature$};
			\draw [thick] (1,2) rectangle ++(0.05cm,0.05cm)  node[anchor=west,above]  {\hspace*{0mm}$(\feature^{(1)},\truelabel^{(1)})$}; 
			\draw [thick] (2,3) rectangle ++(0.05cm,0.05cm) node[anchor=west,above] {\hspace*{0mm}$(\feature^{(2)},\truelabel^{(2)})$};
			\draw [thick] (3,2) rectangle ++(0.05cm,0.05cm) ; 
			\draw [thick] (4,2.5) rectangle ++(0.05cm,0.05cm) ;
			\draw [red,line width=0.4mm,dashed] (0,0) -- (0.95,0) -- (0.95,2) -- (1.05,2) -- (1.05,0) --  (1.95,0) -- 
			(1.95,3) -- (2.05,3) -- (2.05,0) --  (2.95,0) -- (2.95,2) -- (3.05,2) -- (3.05,0)-- (3.95,0) -- (3.95,2.5) -- (4.05,2.5) -- (4.05,0) --  (4.5,0) ; 
			\node [anchor=west] at (4.1,1) {$\hat{h}(\feature)$}; 
			
		\end{tikzpicture}
	\end{center}
	\caption{
		The non-linear hypothesis map $\hat{h}$ perfectly predicts the labels of four \gls{datapoint}s in a \gls{trainset} and therefore 
		has vanishing \gls{trainerr}. Despite perfectly fitting the \gls{trainset}, the hypothesis $\hat{h}$ delivers the trivial (and useless) 
		prediction $\hat{\truelabel}=\hat{h}(\feature)=0$ for \gls{datapoint}s outside the \gls{trainset}. Regularization techniques 
		help to prevent ML methods from learning such a map $\hat{h}$.  
	} 
	\label{fig_regular}
\end{figure}

Many ML methods use the principle of \gls{erm} (see Chapter \ref{ch_Optimization}) 
to learn a hypothesis out of a \gls{hypospace} by minimizing the average loss 
(\gls{trainerr}) on a set of labeled \gls{datapoint}s (which constitute a \gls{trainset}). 
Using \gls{erm} as a guiding principle for ML methods makes sense only if the \gls{trainerr} 
is a good indicator for its loss incurred outside the \gls{trainset}. 

Figure \ref{fig_regular} illustrates a typical scenario for a modern ML method which uses 
a large \gls{hypospace}. This large \gls{hypospace} includes highly non-linear maps 
which can perfectly resemble any dataset of modest size. However, there might be 
non-linear maps for which a small \gls{trainerr} does not guarantee accurate predictions 
for the labels of \gls{datapoint}s outside the \gls{trainset}. 

Chapter \ref{ch_validation_selection} discussed validation techniques to verify if a 
hypothesis with small \gls{trainerr} will predict also well the labels of \gls{datapoint}s 
outside the \gls{trainset}. These validation techniques, including Algorithm \ref{alg:validated_ERM} 
and Algorithm \ref{alg:kfoldCV_ERM}, probe the hypothesis $\hat{h} \in \hypospace$ delivered 
by \gls{erm} on a validation set. The \gls{valset} consists of \gls{datapoint}s which have not 
been used in the \gls{trainset} of \gls{erm} \eqref{equ_def_ERM_funs}. The \gls{valerr}, which 
is the average loss of the hypothesis on the \gls{datapoint}s in the \gls{valset}, serves as an 
estimate for the average error or risk \eqref{equ_def_risk} of the hypothesis $\hat{h}$. 

This chapter discusses \index{regularization}\gls{regularization} as an alternative to validation techniques. 
In contrast to validation, regularization techniques do not require having a separate 
validation set which is not used for the \gls{erm} \eqref{equ_def_ERM_funs}. This makes 
regularization attractive for applications where obtaining a separate \gls{valset} is 
difficult or costly (where labelled data is scarce). 

Instead of probing a hypothesis $\hat{h}$ on a \gls{valset}, regularization techniques estimate (or approximate) 
the loss increase when applying $\hat{h}$ to \gls{datapoint}s outside the \gls{trainset}. The loss increase is estimated 
by adding a regularization term to the \gls{trainerr} in \gls{erm} \eqref{equ_def_ERM_funs}. 

Section \ref{sec_reg_ERM} discusses the resulting regularized \gls{erm}, which we will refer to as 
\gls{srm}. It turns out that the \gls{srm} is equivalent to \gls{erm} using a smaller (pruned) \gls{hypospace}. 
The amount of pruning depends on the weight of the regularization term relative to the \gls{trainerr}. 
For an increasing weight of the regularization term, we obtain a stronger pruning resulting in a 
smaller effective \gls{hypospace}. 

Section \ref{sec_robustness} constructs \gls{regularization} terms by requiring the resulting ML method 
to be robust against (small) random perturbations of the \gls{datapoint}s in a \gls{trainset}. Here, 
we replace each \gls{datapoint} of a \gls{trainset} by the realization of a \gls{rv} that fluctuates 
around this \gls{datapoint}. This construction allows to interpret \gls{regularization} as a (implicit) form of \gls{dataug}. 

Section \ref{sec_data_augmentation} discusses \gls{dataug} methods as 
a simulation-based implementation of regularization. \Gls{dataug} adds a certain 
number of perturbed copies to each \gls{datapoint} in the \gls{trainset}. One way to 
construct perturbed copies of a \gls{datapoint} is to add the realization of a \gls{rv} 
to its features. 

Section \ref{sec_prob_mod_regularization} analyzes the effect of regularization 
for linear regression using a simple probabilistic model for \gls{datapoint}s. This 
analysis parallels our previous study of the validation error of linear regression in Section \ref{sec_gen_linreg}. 
Similar to Section \ref{sec_gen_linreg}, we reveal a trade-off between the bias and variance 
of the hypothesis learnt by regularized \gls{linreg}. This trade- off was traced out by a discrete model parameter 
(the \gls{effdim}) in Section \ref{sec_gen_linreg}. In contrast, regularization offers a continuous 
trade-off between bias and variance via a continuous regularization parameter. 

\index{semi-supervised learning}\Gls{ssl} uses (large amounts of) unlabeled \gls{datapoint}s to 
support the learning of a hypothesis from (a small number of) labeled \gls{datapoint}s \cite{SemiSupervisedBook}. 
Section \ref{sec_ssl_regularization} discusses \gls{ssl} methods that use the statistical properties 
of unlabeled \gls{datapoint}s to construct useful regularization terms. These regularization terms are 
then used in \gls{srm} with a (typically small) set of labeled \gls{datapoint}s.

Multitask learning exploits similarities between different but related learning tasks \cite{Caruana:1997wk}. 
We can formally define a \index{learning task}learning task by a particular choice for the \gls{lossfunc} (see Section \ref{sec_lossfct}) . 
The primary role of a \gls{lossfunc} is to score the quality of a hypothesis map. However, the \gls{lossfunc} 
also encapsulates the choice for the label of a \gls{datapoint}. For learning tasks defined for a single 
underlying data generation process it is reasonable to assume that the same subset of features is 
relevant for those learning tasks. One example for a ML application involving several related learning 
tasks is \gls{multilabelclass} (see Section \ref{sec_labels}). Indeed, each individual label of a \gls{datapoint} 
represents an separate learning task. Section \ref{sec_mtl_regularization} shows how multitask learning 
can be implemented using \gls{regularization} methods. The loss incurred in different learning 
tasks serves mutual regularization terms in a joint \gls{srm} for all learning tasks.

Section \ref{sec_transfer_learning} shows how regularization can be used for {\bf transfer learning}. Like multitask 
learning also transfer learning exploits relations between different learning tasks. In contrast to multitask 
learning, which jointly solves the individual learning tasks, transfer learning solves the learning tasks 
sequentially. The most basic form of transfer learning is to fine tune a pre-trained model. A pre-trained model 
can be obtained via \gls{erm} \eqref{equ_def_ERM_funs} in a (``source'') learning task for which we have a 
large amount of labeled training data. The fine-tuning is then obtained via \gls{erm} \eqref{equ_def_ERM_funs} 
in the (``target'') learning task of interest for which we might have only a small amount of labeled training data. 

\section{Structural Risk Minimization} 
\label{sec_reg_ERM}

Section \ref{sec_hypo_space} defined the \gls{effdim} $\effdim{\hypospace}$ of a 
\gls{hypospace} $\hypospace$ as the maximum number of \gls{datapoint}s that can be perfectly 
fit by some hypothesis $h \in \hypospace$. As soon as the \gls{effdim} of the hypothesis 
space in \eqref{equ_def_ERM_funs} exceeds the number $\samplesize$ of training \gls{datapoint}s, we can 
find a hypothesis that perfectly fits the training data. However, a hypothesis that perfectly fits the 
training data might deliver poor predictions for \gls{datapoint}s outside the \gls{trainset} (see Figure \ref{fig_regular}). 

Modern ML methods typically use a \gls{hypospace} with large \gls{effdim} \cite{Wain2019,BuhlGeerBook}. 
Two well-known examples for such methods is \gls{linreg} (see Section \ref{sec_lin_reg}) using a large number of 
features and deep learning with \gls{ann}s using a large number (billions) of artificial neurons (see Section \ref{sec_deep_learning}). 
The \gls{effdim} of these methods can be easily on the order of billions ($10^{9}$) if not larger \cite{ShalevMLBook}. 
To avoid overfitting during the naive use of \gls{erm} \eqref{equ_def_ERM_funs} we would require a \gls{trainset} 
containing at least as many \gls{datapoint}s as the \gls{effdim} of the \gls{hypospace}. However, in practice we 
often do not have access to a \gls{trainset} consisting of billions of labeled \gls{datapoint}s. The challenge is typically 
in the labelling process which often requires human labour. 

It seems natural to combat overfitting of a ML method by pruning its \gls{hypospace} $\hypospace$. 
We prune $\hypospace$ by removing some of the hypothesis in $\hypospace$ to obtain the smaller 
\gls{hypospace} $\hypospace' \subset \hypospace$. 
We then replace \gls{erm} \eqref{equ_def_ERM_funs} with the restricted (or pruned) \gls{erm}
\begin{equation}
	\label{equ_ERM_fun_pruned}
	\hat{h} = \argmin_{h \in \hypospace'} \emperror(h|\dataset) \mbox{ with pruned \gls{hypospace} } \hypospace' \!\subset\!\hypospace. 
\end{equation}
The \gls{effdim} of the pruned \gls{hypospace} $\hypospace'$ is typically much smaller than 
the \gls{effdim} of the original (large) \gls{hypospace} $\hypospace$,  $\effdim{\hypospace'} \ll \effdim{\hypospace}$. 
For a given size $\samplesize$ of the \gls{trainset}, the risk of overfitting in \eqref{equ_ERM_fun_pruned} is 
much smaller than the risk of overfitting in \eqref{equ_def_ERM_funs}. 

Let us illustrate the idea of pruning for \gls{linreg} using the \gls{hypospace} \eqref{equ_lin_hypospace} constituted by 
linear maps $h(\featurevec) = \weights^{T} \featurevec$. The \gls{effdim} of \eqref{equ_lin_hypospace}  is equal to the number 
of features, $\effdim{\hypospace} = \featuredim$. The \gls{hypospace} $\hypospace$ might be too large if we use a 
large number $\featurelen$ of features, leading to overfitting. We prune \eqref{equ_lin_hypospace} by retaining only 
linear hypotheses $h(\featurevec) = \big(\weights'\big)^T \featurevec$ with parameter vectors $\weights'$ 
satisfying $\weight'_{3} = \weights_{4}'= \ldots = \weights_{\featurelen}'=0$. Thus, the \gls{hypospace} $\hypospace'$ 
is constituted by all linear maps that only depend on the first two features $\feature_{1},\feature_{2}$ of a \gls{datapoint}. 
The \gls{effdim} of $\hypospace'$ is  dimension is $\effdim{\hypospace'}=2$ instead of $\effdim{\hypospace}=\featurelen$. 

Pruning the \gls{hypospace} is a special case of a more general strategy which we  refer to as \gls{srm} \cite{VapnikBook}. 
The idea behind \gls{srm} is to modify the \gls{trainerr} in \gls{erm} \eqref{equ_def_ERM_funs} to favour hypotheses 
which are more smooth or regular in a specific sense. By enforcing a smooth hypothesis, a ML 
methods becomes less sensitive, or more robust, to small perturbations of \gls{datapoint}s in the \gls{trainset}. 
Section \ref{sec_robustness} discusses the intimate relation between the robustness (against perturbations 
of the \gls{datapoint}s in the \gls{trainset}) of a ML method and its ability to generalize to \gls{datapoint}s outside the \gls{trainset}. 

We measure the smoothness of a hypothesis using a \index{regularizer}regularizer $\regularizer(h) \in \mathbb{R}_{+}$. 
Roughly speaking, the value $\regularizer(h)$ measures the irregularity or variation of a predictor map $h$. 
The (design) choice for the regularizer depends on the precise definition of what is meant by regularity 
or variation of a hypothesis. Section \ref{sec_data_augmentation} discusses how a particular choice for the 
regularizer $\regularizer(h)$ arises naturally from a probabilistic model for \gls{datapoint}s. 

We obtain \gls{srm} by adding the scaled regularizer $\regparam \regularizer(h)$ to the \gls{erm} \eqref{equ_def_ERM_funs} , 
\begin{align}
	\label{equ_ERM_fun_regularized}
	\hat{h} & = \argmin_{h \in \hypospace} \big[ \emperror(h|\dataset)  + \regparam \regularizer(h) \big] \nonumber \\
	&   \stackrel{\eqref{eq_def_emp_error_101}}{=}  \argmin_{h \in \hypospace} \big[(1/\samplesize) \sum_{\sampleidx=1}^{\samplesize} \loss{(\featurevec^{(\sampleidx)},\truelabel^{(\sampleidx)})}{h}+ \regparam \regularizer(h)\big]. 
\end{align} 
We can interpret the penalty term $\regparam \regularizer(h)$ in \eqref{equ_ERM_fun_regularized} as an estimate 
(or approximation) for the increase, relative to the \gls{trainerr} on $\dataset$, of the average loss of a hypothesis $\hat{h}$ 
when it is applied to \gls{datapoint}s outside $\dataset$. Another interpretation of the term $\regparam \regularizer(h)$ will be 
discussed in Section \ref{sec_data_augmentation}.

The \index{regularization parameter} regularization parameter $\regparam$ allows us to trade between a 
small \gls{trainerr} $\emperror(h^{(\vw)}|\dataset)$ and small regularization term $\regularizer(h)$, which 
enforces smoothness or regularity of $h$. If we choose a large value for $\regparam$, irregular or hypotheses $h$, 
with large $\regularizer(h)$, are heavily ``punished'' in \eqref{equ_ERM_fun_regularized}. Thus, 
increasing the value of $\regparam$ results in the solution (minimizer) of \eqref{equ_ERM_fun_regularized} having 
smaller $\regularizer(h)$. On the other hand, choosing a small value for $\regparam$ in \eqref{equ_ERM_fun_regularized} puts more 
emphasis on obtaining a hypothesis $h$ incurring a small \gls{trainerr}. 
For the extreme case $\regparam =0$, the \gls{srm} \eqref{equ_ERM_fun_regularized} reduces to ERM \eqref{equ_def_ERM_funs}. 

The pruning approach \eqref{equ_ERM_fun_pruned} is intimately related to the \gls{srm} \eqref{equ_ERM_fun_regularized}. 
They are, in a certain sense, {\bf dual} to each other. First, note that \eqref{equ_ERM_fun_regularized} reduces to the 
pruning approach \eqref{equ_ERM_fun_pruned} when using the regularizer $\regularizer(h) = 0$ for all $h \in \hypospace'$ 
, and $\regularizer(h) = \infty$ otherwise, in \eqref{equ_ERM_fun_regularized}. In the other direction, for many important 
choices for the regularizer $\regularizer(h)$, there is a restriction $\hypospace^{(\regparam)} \subset \hypospace$ such that 
the solutions of \eqref{equ_ERM_fun_pruned} and \eqref{equ_ERM_fun_regularized} coincide (see Figure \ref{fig_soft_pruning_regularization}). 
The relation between the optimization problems \eqref{equ_ERM_fun_pruned} and \eqref{equ_ERM_fun_regularized} can 
be made precise using the theory of convex duality (see \cite[Ch. 5]{BoydConvexBook} and \cite{BertsekasNonLinProgr}).

\begin{figure} 
	\begin{center}
		\begin{tikzpicture}
			\node [above] at (0,2) {$\regparam=0$};
			\draw [line width=2,fill=black!10] (2,0) -- (0,2) -- (-2,0) -- (0,-2) -- cycle;
			\node [] at (0,0) {$\hypospace^{(\regparam\!=\!0)}$};
			\node [above] at (5.5,2) {$\regparam=1$};
			\filldraw [line width=2,fill=black!10] (7,0) -- (5.5,1.5) -- (4,0) -- (5.5,-1.5) -- cycle;
			\node [] at (5.5,0) {$\hypospace^{(\regparam\!=\!1)}$};
			\node [above] at (10,2) {$\regparam=10$};
			\draw [line width=2,fill=black!10] (11.,0) -- (10,1) -- (9,0) -- (10,-1) -- cycle;
			\node [] at (10,0) {$\hypospace^{(\regparam\!=\!10)}$};
		\end{tikzpicture}
	\end{center}
	\caption{Adding the scaled regularizer $\regparam \regularizer(h)$ to the \gls{trainerr} in the objective 
		function of \gls{srm} \eqref{equ_ERM_fun_regularized} is equivalent to solving \gls{erm} \eqref{equ_ERM_fun_pruned} 
		with a pruned \gls{hypospace} $\hypospace^{(\regparam)}$.}
	\label{fig_soft_pruning_regularization}
\end{figure}

%

For a \gls{hypospace} $\hypospace$ whose elements $h \in \hypospace$ are parametrized by 
a parameter vector $\weights \in \mathbb{R}^{\featuredim}$, we can rewrite \gls{srm} \eqref{equ_ERM_fun_regularized} as 
\begin{align} 
	\label{equ_rerm_weight}
	\widehat{\weights}^{(\regparam)}  & = \argmin_{\weights \in  \mathbb{R}^{\featurelen}} \big[ \emperror(h^{(\weights)}|\dataset)+ \regparam \regularizer(\weights)\big] \nonumber \\ 
	&  = \argmin_{\weights \in  \mathbb{R}^{\featurelen}} \big[(1/\samplesize) \sum_{\sampleidx=1}^{\samplesize} \loss{(\featurevec^{(\sampleidx)},\truelabel^{(\sampleidx)})}{h^{(\weights)}} + \regparam \regularizer(\weights) \big]. 
\end{align}
For the particular choice of squared error loss \eqref{equ_squared_loss}, linear \gls{hypospace} \eqref{equ_lin_hypospace} 
and regularizer $\regularizer(\weights)=\| \weights \|_{2}^{2}$, \gls{srm} \eqref{equ_rerm_weight} specializes to 
\begin{align} 
	\label{equ_rerm_ridge_regression}
	\widehat{\weights}^{(\regparam)}  & = \argmin_{\weights \in  \mathbb{R}^{\featurelen}} \big[(1/\samplesize) \sum_{\sampleidx=1}^{\samplesize} \big( \truelabel^{(\sampleidx)} - \weights^{T} \featurevec^{(\sampleidx)}\big)^{2} + \regparam \| \weights \|_{2}^{2}\big]. 
\end{align}
The special case \eqref{equ_rerm_ridge_regression} of \gls{srm} \eqref{equ_rerm_weight} is known 
as \gls{ridgeregression} \cite{hastie01statisticallearning}. 

\index{ridge regression}\Gls{ridgeregression} \eqref{equ_rerm_ridge_regression} is equivalent to (see \cite[Ch. 5]{BertsekasNonLinProgr}) 
\begin{equation} 
	\label{equ_restr_ERM}
	\widehat{\weights}^{(\regparam)} = \argmin_{h^{(\weights)} \in \hypospace^{(\regparam)}}  (1/\samplesize) \sum_{\sampleidx=1}^{\samplesize}  \big(\truelabel^{(\sampleidx)} - h^{(\weights)}(\featurevec^{(\sampleidx)})  \big)^2
\end{equation}  
with the restricted \gls{hypospace}
\begin{align} 
	\label{equ_hyposapce_lambda}
	\hypospace^{(\regparam)} & \defeq \{ h^{(\weights)}: \mathbb{R}^{\featuredim} \rightarrow \mathbb{R}: h^{(\vw)}(\featurevec) = \weights^{T} \featurevec  \mbox{, with some } \weights \in \mathbb{R}^{\featuredim}, \| \weights \|_{2}^{2} \leq C(\regparam) \} \subset \hypospace^{(\featuredim)}. 
\end{align} 
For any given value $\regparam$ of the regularization parameter in \eqref{equ_rerm_ridge_regression}, there is a number $C(\regparam)$ such that 
solutions of \eqref{equ_rerm_ridge_regression} coincide with the solutions of \eqref{equ_restr_ERM}. Thus, ridge 
regression \eqref{equ_rerm_ridge_regression} is equivalent to \gls{linreg} with a pruned version $\hypospace^{(\regparam)}$ 
of the linear hypothesis space \eqref{equ_lin_hypospace}. The size of the pruned \gls{hypospace} $\hypospace^{(\regparam)}$ \eqref{equ_hyposapce_lambda} varies continuously with $\regparam$.

Another popular special case of \gls{erm} \eqref{equ_rerm_weight} is obtained for the regularizer $\regularizer(\weights)=\| \weights \|_{1}$ 
and known as the Lasso \cite{HastieWainwrightBook}
\begin{align} 
	\label{equ_rerm_Lasso}
	\widehat{\weights}^{(\regparam)}  & = \argmin_{\weights \in  \mathbb{R}^{\featurelen}} \big[(1/\samplesize) \sum_{\sampleidx=1}^{\samplesize} \big( \truelabel^{(\sampleidx)} - \weights^{T} \featurevec^{(\sampleidx)}\big)^{2} + \regparam \| \weights \|_{1}\big]. 
\end{align}
Ridge regression \eqref{equ_rerm_ridge_regression} and the Lasso \eqref{equ_rerm_Lasso} have fundamentally different 
computational and statistical properties. Ridge regression \eqref{equ_rerm_ridge_regression} uses a smooth and convex 
objective function that can be minimized using efficient \gls{gd} methods. The objective function of Lasso \eqref{equ_rerm_Lasso} 
is also convex but non-smooth and therefore requires more advanced optimization methods. The increased 
computational complexity of \index{Lasso}Lasso \eqref{equ_rerm_Lasso} comes at the benefit of typically delivering 
a hypothesis with a smaller expected loss than those obtained from \index{ridge regression}\gls{ridgeregression} \cite{BuhlGeerBook,HastieWainwrightBook}.


%



\section{Robustness} 
\label{sec_robustness} 

Section \ref{sec_reg_ERM} motivates regularization as a soft variant of model selection. Indeed, the regularization term 
in \gls{srm} \eqref{equ_ERM_fun_regularized} is equivalent to \gls{erm} \eqref{equ_ERM_fun_pruned} using a pruned (reducing) 
\gls{hypospace}. We now discuss an alternative view on regularization as a means to make ML methods robust. 

The ML methods discussed in Chapter \ref{ch_Optimization} rest on the idealizing assumption that 
we have access to the true label values and feature values of labeled \gls{datapoint}s (that form a \gls{trainset}). 
These methods learn a hypothesis $h \in \hypospace$ with minimum average loss (\gls{trainerr}) 
incurred for \gls{datapoint}s in the \gls{trainset}. In practice, the acquisition of label and feature values 
might be prone to errors. These errors might stem from the measurement device itself (hardware 
failures or thermal noise in electronic devices) or might be due to human mistakes such as labelling errors. 

Let us assume for the sake of exposition that the label values $\truelabel^{(\sampleidx)}$ 
in the \gls{trainset} are accurate but that the features $\featurevec^{(\sampleidx)}$ are a perturbed version 
of the true features of the $\sampleidx$th \gls{datapoint}. Thus, instead of having observed the 
\gls{datapoint} $\big( \featurevec^{(\sampleidx)}, \truelabel^{(\sampleidx)} \big)$ we could have equally well observed 
the \gls{datapoint} $\big( \featurevec^{(\sampleidx)}+\bm{\varepsilon}, \truelabel^{(\sampleidx)} \big)$ in the \gls{trainset}. 
Here, we have modelled the perturbations in the features using a \gls{rv} $\bm{\varepsilon}$. The probability 
distribution of the perturbation $\bm{\varepsilon}$ is a design parameter that controls robustness 
properties of the overall ML method. We will study a particular choice for this distribution in Section \ref{sec_data_augmentation}. 

A robust ML method should learn a hypothesis that incurs a small loss not only for a 
specific \gls{datapoint} $\big( \featurevec^{(\sampleidx)}, y^{(\sampleidx)} \big)$ but also for perturbed 
\gls{datapoint}s $\big( \featurevec^{(\sampleidx)}+\bm{\varepsilon}, y^{(\sampleidx)} \big)$. 
Therefore, it seems natural to replace the loss $\loss{\big( \featurevec^{(\sampleidx)}, y^{(\sampleidx)} \big)}{h}$, 
incurred on the $\sampleidx$th \gls{datapoint} in the \gls{trainset}, with the expectation 
\begin{equation} 
	\label{equ_def_expe_perturb_robust}
	\expect \big\{ \loss{ \big( \featurevec^{(\sampleidx)}+\bm{\varepsilon}, y^{(\sampleidx)} \big)}{h}\big\}.
\end{equation} 
The expectation \eqref{equ_def_expe_perturb_robust} is computed using the probability distribution 
of the perturbation $\bm{\varepsilon}$. We will show in Section \ref{sec_data_augmentation} that minimizing 
the average of the expectation \eqref{equ_def_expe_perturb_robust}, for $\sampleidx=1,\ldots,\samplesize$, is equivalent 
to the \gls{srm} \eqref{equ_ERM_fun_regularized}. 

Using the expected loss \eqref{equ_def_expe_perturb_robust} is not the only possible approach to make 
a ML method robust. Another approach to make a ML method robust is known as \index{bagging}\gls{bagging}. 
The idea of \gls{bagging} is to use the bootstrap method (see Section \ref{sec_the_bootsrap} and \cite[Ch. 8]{hastie01statisticallearning}) 
to construct a finite number of perturbed copies $\dataset^{(1)},\ldots,\dataset^{(\augparam)}$ 
of the original \gls{trainset} $\dataset$. 

We then learn (e.g, using \gls{erm}) a separate hypothesis $h^{(\augidx)}$  for each perturbed copy 
$\dataset^{(\augidx)}$, $\augidx = 1,\ldots, \augparam$. This results in a whole ensemble 
of different hypotheses $h^{(\augidx)}$ which might even belong to different hypothesis spaces. For example, one 
the hypothesis $h^{(1)}$ could be a linear map (see Section \ref{sec_lin_reg}) and the hypothesis  $h^{(2)}$ could 
be obtained from an \gls{ann} (see Section \ref{sec_deep_learning}).  

The final hypothesis delivered by \gls{bagging} is obtained by combining or aggregating (e.g., using the average) the predictions 
$h^{(\augidx)}\big(\featurevec\big)$ delivered by each hypothesis $h^{(\augidx)}$, for $\augidx=1,\ldots,\augparam$ in the ensemble. The ML method referred to as \index{random forest}\gls{randomforest} uses \gls{bagging}  
to learn an ensemble of \gls{decisiontree}s (see Chapter \ref{sec_decision_trees}). The individual 
predictions obtained from the different \gls{decisiontree}s forming a \gls{randomforest} are then 
combined (e.g., using an average for numeric labels or a majority vote for finite-valued labels), 
to obtain a final prediction \cite{hastie01statisticallearning}.

\section{Data Augmentation} 
\label{sec_data_augmentation} 
\index{data augmentation}

ML methods using \gls{erm} \eqref{equ_def_ERM_funs} are prone to overfitting as soon 
as the \gls{effdim} of the \gls{hypospace} $\hypospace$ exceeds the number $\samplesize$ of \gls{datapoint}s 
in the \gls{trainset}. Section \ref{sec_modsel} and Section \ref{sec_reg_ERM} 
approached this by modifying either the model or the loss function by adding a regularization term. 
Both approaches prune the \gls{hypospace} $\hypospace$ underlying a ML method to reduce 
the \gls{effdim} $\effdim{\hypospace}$. Model selection does this reduction in a discrete 
fashion while regularization implements a soft ``shrinking'' of the \gls{hypospace}. 

Instead of trying to reduce the \gls{effdim} we could also try to increase the number 
$\samplesize$ of \gls{datapoint}s in the \gls{trainset} used for \gls{erm} \eqref{equ_def_ERM_funs}. 
We now discuss how to synthetically generate new labeled \gls{datapoint}s by exploiting 
statistical symmetries of data. 

The data arising in many ML applications exhibit intrinsic symmetries and invariances at least in 
some approximation. The rotated image of a cat still shows a cat. The temperature measurement 
taken at a given location will be similar to another measurement taken $10$ milliseconds later. 
\Gls{dataug} exploits such symmetries and invariances to augment the raw data with 
additional synthetic data. 

Let us illustrate \gls{dataug} using an application that involves \gls{datapoint}s characterized 
by features $\featurevec \in \mathbb{R}^{\featuredim}$ and number labels $y \in \mathbb{R}$. We assume 
that the data generating process is such that \gls{datapoint}s with close feature values have the same 
label. Equivalently, this assumption is requiring the resulting ML method to be robust against small 
perturbations of the feature values (see Section \ref{sec_robustness}). 
This suggests to augment a \gls{datapoint} $\big(\featurevec,\truelabel\big)$ by several synthetic \gls{datapoint}s 
\begin{equation} 
	\label{equ_def_copies_aug}
	\big(\featurevec+{\bm \varepsilon}^{(1)},\truelabel\big),\ldots,\big(\featurevec+{\bm \varepsilon}^{(\augparam)},\truelabel\big), 
\end{equation}
with ${\bm \varepsilon}^{(1)},\ldots,{\bm \varepsilon}^{(\augparam)}$ being realizations of \gls{iid} 
random vectors with the same probability distribution $p({\bm \varepsilon})$. 

Given a (raw) dataset $\dataset = \big\{ \big(\featurevec^{(1)},\truelabel^{(1)}\big),\ldots, \big(\featurevec^{(\samplesize)},\truelabel^{(\samplesize)}\big) \} $ 
we denote the associated augmented dataset by 
\begin{align} 
	\label{equ_def_augmented_dataset}
	\dataset' = \big\{ &\big(\featurevec^{(1,1)},y^{(1)}\big), \ldots, \big(\featurevec^{(1,\augparam)},\truelabel^{(1)}\big), \nonumber \\ 
	&\big(\featurevec^{(2,1)},\truelabel^{(2)}\big), \ldots, \big(\featurevec^{(2,\augparam)},\truelabel^{(2)}\big), \nonumber \\ 
	& \ldots \nonumber \\ 
	&\big(\featurevec^{(\samplesize,1)},y^{(\samplesize)}\big), \ldots, \big(\featurevec^{(\samplesize,\augparam)},\truelabel^{(\samplesize)}\big) \}. 
\end{align} 
The size of the augmented dataset $\dataset'$ is $ \samplesize' = \augparam \times \samplesize$. 
For a sufficiently large augmentation parameter $\augparam$, the augmented sample size $\samplesize'$ is 
larger than the \gls{effdim} $\featurelen$ of the \gls{hypospace} $\hypospace$. We then learn a 
hypothesis via \gls{erm} on the augmented dataset, 
\begin{align}
	\label{equ_def_ERM_funs_aug}
	\hat{h} & = \argmin_{h \in \hypospace} \emperror(h|\dataset') \nonumber \\ 
	& \stackrel{\eqref{equ_def_augmented_dataset}}{=}  \argmin_{h \in \hypospace} (1/\samplesize') \sum_{\sampleidx=1}^{\samplesize} \sum_{\augidx=1}^{\augparam} \loss{(\featurevec^{(\sampleidx,\augidx)},y^{(\sampleidx,\augidx)})}{h} \nonumber \\ 
	& \stackrel{\eqref{equ_def_copies_aug}}{=}  \argmin_{h \in \hypospace} (1/\samplesize) \sum_{\sampleidx=1}^{\samplesize} (1/\augparam)\sum_{\augidx=1}^{\augparam} \loss{(\featurevec^{(\sampleidx)}+{\bm \varepsilon}^{(b)},y^{(\sampleidx)})}{h}. 
\end{align}
We can interpret data-augmented \gls{erm} \eqref{equ_def_ERM_funs_aug} as a data-driven 
form of regularization (see Section \ref{sec_reg_ERM}). The regularization is implemented 
by replacing, for each \gls{datapoint} $\big(\featurevec^{(\sampleidx)},y^{(\sampleidx)}\big) \in \dataset$, 
the loss $\loss{(\featurevec^{(\sampleidx)},y^{(\sampleidx)})}{h}$ with the average 
loss $(1/\augparam)\sum_{\augidx=1}^{\augparam} \loss{(\featurevec^{(\sampleidx)}+{\bm \varepsilon}^{(b)},y^{(\sampleidx)})}{h}$ 
over the augmented \gls{datapoint}s that accompany $\big(\featurevec^{(\sampleidx)},y^{(\sampleidx)}\big) \in \dataset$. 

Note that in order to implement \eqref{equ_def_ERM_funs_aug} we need to first generate 
$\augparam$ realizations  ${\bm \varepsilon}^{(b)} \in \mathbb{R}^{\featurelen}$ of \gls{iid}
random vectors with common probability distribution $p({\bm \varepsilon})$. 
This might be computationally costly for a large $\augparam, \featurelen$. 
However, when using a large augmentation parameter $\augparam$, we might 
use the approximation  
\begin{align}
	\label{equ_approx_augm_loss_expect}
	(1/\augparam)\sum_{\augidx=1}^{\augparam} \loss{(\featurevec^{(\sampleidx)}+{\bm \varepsilon}^{(\augidx)},y^{(\sampleidx)})}{h} \approx  
	\expect \big\{ \loss{(\featurevec^{(\sampleidx)}+{\bm \varepsilon},y^{(\sampleidx)})}{h} \big\}. 
\end{align}
This approximation is made precise by a key result of probability theory, known as the \gls{lln}. 
We obtain an instance of \gls{erm} by inserting \eqref{equ_approx_augm_loss_expect} into \eqref{equ_def_ERM_funs_aug}, 
\begin{align}
	\label{equ_def_ERM_funs_aug_approx}
	\hat{h}  = \argmin_{h \in \hypospace} (1/\samplesize) \sum_{\sampleidx=1}^{\samplesize} \expect\big\{ \loss{(\featurevec^{(\sampleidx)}+{\bm \varepsilon},\truelabel^{(\sampleidx)})}{h} \big\}. 
\end{align}

The usefulness of \eqref{equ_def_ERM_funs_aug_approx} as an approximation to the augmented 
\gls{erm} \eqref{equ_def_ERM_funs_aug} depends on the difficulty of computing the expectation $\expect\big\{ \loss{(\featurevec^{(\sampleidx)}+{\bm \varepsilon},\truelabel^{(\sampleidx)})}{h} \big\}$. The complexity of computing this expectation depends 
on the choice of loss function and the choice for the probability distribution $p({\bm \varepsilon})$.  

Let us study \eqref{equ_def_ERM_funs_aug_approx} for the special case linear regression with squared error loss 
\eqref{equ_squared_loss} and linear \gls{hypospace} \eqref{equ_lin_hypospace}, 
\begin{align}
	\label{equ_def_ERM_funs_aug_approx_linreg}
	\hat{h}  = \argmin_{h^{(\vw)} \in \hypospace^{(\featuredim)}} (1/\samplesize) \sum_{\sampleidx=1}^{\samplesize} \expect\big\{ \big( y^{(\sampleidx)} - \vw^{T} \big(\featurevec^{(\sampleidx)}+{\bm \varepsilon} \big) \big)^{2} \big\}. 
\end{align}
We use perturbations ${\bm \varepsilon}$ drawn a multivariate normal distribution with zero mean and covariance matrix $\sigma^{2} \mathbf{I}$, 
\begin{equation} 
	\label{equ_augm_mvn_standard}
	{\bm \varepsilon} \sim \mathcal{N}(\mathbf{0},\sigma^{2} \mathbf{I}). 
\end{equation}
We develop \eqref{equ_def_ERM_funs_aug_approx_linreg} further by using 
\begin{equation} 
	\label{equ_uncorr_augmentation_implicit}
	\expect\{\big( \truelabel^{(\sampleidx)} - \weights^{T} \featurevec^{(\sampleidx)} \big) {\bm \varepsilon}  \} = \mathbf{0}. 
\end{equation}
The identity \eqref{equ_uncorr_augmentation_implicit} uses that the \gls{datapoint}s $\big(\featurevec^{(\sampleidx)},\truelabel^{(\sampleidx)}\big)$ 
are fixed and known (deterministic) while ${\bm \varepsilon}$ is a zero-mean random vector. 
Combining \eqref{equ_uncorr_augmentation_implicit} with \eqref{equ_def_ERM_funs_aug_approx_linreg}, 
\begin{align} 
	\label{equ_implicit_aug_regu_11}
	\expect\big\{ \big( y^{(\sampleidx)} - \weights^{T} \big(\featurevec^{(\sampleidx)}+{\bm \varepsilon} \big) \big)^{2} \big\}
	& = \big( \truelabel^{(\sampleidx)} - \weights^{T}\featurevec^{(\sampleidx)} \big) ^{2} \!+\!\sqeuclnorm{ \weights } \, \expect \big\{ \sqeuclnorm{ {\bm \varepsilon} } \big\} \nonumber \\[5mm]
	&=  \big( \truelabel^{(\sampleidx)} - \weights^{T}\featurevec^{(\sampleidx)} \big) ^{2}  + \featurelen \sqeuclnorm{ \weights }  \sigma^{2}. 
\end{align}
where the last step used $\expect \big\{  \sqeuclnorm { {\bm \varepsilon} } \big\} \stackrel{\eqref{equ_augm_mvn_standard}}{=} \featurelen \sigma^{2}$. 
%
Inserting \eqref{equ_implicit_aug_regu_11} into \eqref{equ_def_ERM_funs_aug_approx_linreg}, 
\begin{align}
	\label{equ_def_ERM_funs_aug_approx_ridge}
	\hat{h}  = \argmin_{h^{(\weights)} \in \hypospace^{(\featuredim)}} (1/\samplesize) \sum_{\sampleidx=1}^{\samplesize} \big( \truelabel^{(\sampleidx)} - \vw^{T}\featurevec^{(\sampleidx)} \big) ^{2}  + \featurelen \sqeuclnorm{ \weights }  \sigma^{2}.
\end{align}
We have obtained \eqref{equ_def_ERM_funs_aug_approx_ridge} as an approximation of 
the augmented \gls{erm} \eqref{equ_def_ERM_funs_aug} for the special case of squared error 
loss \eqref{equ_squared_loss} and the linear \gls{hypospace} \eqref{equ_lin_hypospace}. 
This approximation uses the \gls{lln} \eqref{equ_approx_augm_loss_expect} and becomes more 
accurate for increasing augmentation parameter $\augparam$. 

Note that \eqref{equ_def_ERM_funs_aug_approx_ridge} is nothing but \gls{ridgeregression}
\eqref{equ_rerm_ridge_regression} using the regularization parameter $\regparam  =\featurelen \sigma^{2}$. 
Thus, we can interpret \gls{ridgeregression} as implicit \gls{dataug} \eqref{equ_def_augmented_dataset} 
by applying random perturbations \eqref{equ_def_copies_aug} to the feature vectors in the 
original \gls{trainset} $\dataset$.

The regularizer $\regularizer(\weights) = \sqeuclnorm{ \weights }$ in \eqref{equ_def_ERM_funs_aug_approx_ridge} 
arose naturally from the specific choice for the probability distribution \eqref{equ_augm_mvn_standard} 
of the random perturbation ${\bm \varepsilon}^{(\sampleidx)}$ in \eqref{equ_def_copies_aug} and using the 
squared error loss. Other choices for this probability distribution or the loss function result in different regularizers. 

Augmenting \gls{datapoint}s with random perturbations distributed according \eqref{equ_augm_mvn_standard} 
treat the features of a \gls{datapoint} independently. For application domains that generate \gls{datapoint}s with 
highly correlated features it might be useful to augment \gls{datapoint}s using random perturbations ${\bm \varepsilon}$ 
(see \eqref{equ_def_copies_aug}) distributed as
\begin{equation} 
	\label{equ_augm_mvn_cov_matris}
	{\bm \varepsilon} \sim \mathcal{N}(\mathbf{0},\mC). 
\end{equation}
The covariance matrix $\mC$ of the perturbation ${\bm \varepsilon}$ can be chosen 
using domain expertise or estimated (see Section \ref{sec_ssl_regularization}). 
Inserting the distribution \eqref{equ_augm_mvn_cov_matris} into \eqref{equ_def_ERM_funs_aug_approx}, 
\begin{align}
	\label{equ_def_ERM_funs_aug_approx_ridge_cov}
	\hat{h}  = \argmin_{h^{(\weights)} \in \hypospace^{(\featuredim)}} \bigg[  (1/\samplesize) \sum_{\sampleidx=1}^{\samplesize} \big( \truelabel^{(\sampleidx)} - \weights^{T}\featurevec^{(\sampleidx)} \big) ^{2}  +  \weights^{T}  \mC  \weights \bigg] .
\end{align}
Note that \eqref{equ_def_ERM_funs_aug_approx_ridge_cov} reduces to ordinary \gls{ridgeregression}
\eqref{equ_def_ERM_funs_aug_approx_ridge} for the choice $\mC = \sigma^{2} \mathbf{I}$.

\section{Statistical and Computational Aspects of Regularization}
\label{sec_prob_mod_regularization}

The goal of this section is to develop a better understanding for the effect of the regularization 
term in \gls{srm} \eqref{equ_rerm_weight}. We will analyze the solutions of \gls{ridgeregression} \eqref{equ_rerm_ridge_regression} 
which is the special case of \gls{srm} using the linear \gls{hypospace} \eqref{equ_lin_hypospace} 
and squared error loss \eqref{equ_squared_loss}. 
Using the feature matrix $\mX\!=\!\big(\featurevec^{(1)},\ldots,\featurevec^{(\samplesize)}\big)^{T}$ and label vector $\labelvec\!=\!(\truelabel^{(1)},\ldots,\truelabel^{(\samplesize)})^{T}$, 
we can rewrite \eqref{equ_rerm_ridge_regression} more compactly as 
\begin{equation} 
	\label{equ_def_rlr_w_opt}
	\widehat{\weights}^{(\regparam)} = \argmin_{\weights \in \mathbb{R}^{\featuredim}} \big[ (1/\samplesize) \sqeuclnorm{\labelvec - \featuremtx \vw} + \regparam \sqeuclnorm{ \vw }\big].
\end{equation} 
The solution of \eqref{equ_def_rlr_w_opt} is given by 
\begin{equation}
	\label{equ_close_form_reglinreg}
	\widehat{\weights}^{(\regparam)} = (1/\samplesize) \big((1/\samplesize) \featuremtx^{T} \featuremtx + \regparam \mathbf{I} \big)^{-1} \featuremtx^{T} \labelvec. 
\end{equation}
For $\regparam\!=\!0$, \eqref{equ_close_form_reglinreg} reduces to the formula \eqref{equ_optimal_weight_closed_form} for 
the optimal \gls{weights} in linear regression (see \eqref{equ_rerm_ridge_regression} and \eqref{equ_def_cost_MSE}). 
Note that for $\regparam>0$, the formula \eqref{equ_close_form_reglinreg} is always valid, even when $\featuremtx^{T} \featuremtx$ is 
singular (not invertible). For $\regparam> 0$ the optimization problem \eqref{equ_def_rlr_w_opt} (and \eqref{equ_rerm_ridge_regression}) 
has the unique solution \eqref{equ_close_form_reglinreg}. 

To study the statistical properties of the predictor $h^{(\widehat{\weights}^{(\regparam)})}(\featurevec) = \big(\widehat{\weights}^{(\regparam)}\big)^{T} \featurevec$
(see \eqref{equ_close_form_reglinreg}) we use the probabilistic toy model \eqref{equ_linear_obs_model}, \eqref{equ_toy_model_iid} and \eqref{equ_labels_training_data} that we used already in Section \ref{sec_gen_linreg}. We interpret the training data $\dataset^{(\rm train)} = \{ (\featurevec^{(\sampleidx)},\truelabel^{(\sampleidx)}) \}_{\sampleidx=1}^{\samplesize}$ as realizations of \gls{iid} \gls{rv}s whose 
distribution is defined by \eqref{equ_linear_obs_model}, \eqref{equ_toy_model_iid} and \eqref{equ_labels_training_data}. 

We can then define the average prediction error of \gls{ridgeregression} as 
\begin{equation} 
	\error_{\rm pred}^{(\regparam)} \defeq \expect \bigg\{ \bigg( \truelabel - h^{(\widehat{\weights}^{(\regparam)})}(\featurevec) \bigg)^{2} \bigg\}. 
\end{equation} 
As shown in Section \ref{sec_gen_linreg}, the error $\error_{\rm pred}^{(\regparam)}$ is the sum of three 
components: the bias, the variance and the noise variance $\sigma^{2}$ (see \eqref{equ_decomp_E_pred_toy_model}). 
The bias of $\widehat{\weights}^{(\regparam)}$ is  
\begin{equation} 
	\label{equ_bias_reg_lin_reg}
	\biasterm^{2} = \expect \bigg\{ \sqeuclnorm{ (\mathbf{I} - ( \featuremtx^{T} \featuremtx +  \samplesize \regparam \mathbf{I})^{-1} \featuremtx^{T} \featuremtx  )  \overline{\weights} } \bigg\}. 
\end{equation} 
For sufficiently large size $\samplesize$ of the \gls{trainset}, we can use the approximation 
\begin{equation} 
	\label{equ_approx_Gram_large_N}
	\mX^{T} \mX  \approx \samplesize \mathbf{I} 
\end{equation} 
such that \eqref{equ_bias_reg_lin_reg} can be approximated as 
\begin{align} 
	\label{equ_bias_reg_lin_reg_approx}
	\biasterm^{2} & \approx \sqeuclnorm{(\mathbf{I}\!-\!(\mathbf{I}\!+\!\regparam \mathbf{I})^{-1} ) \overline{\weights} }\nonumber \\
	& =  \sum_{\featureidx=1}^{\featuredim} \frac{\regparam}{1+\regparam} \overline{\weight}_{\featureidx}^2 .
\end{align} 
Let us compare the (approximate) \gls{bias} term \eqref{equ_bias_reg_lin_reg_approx} of \gls{ridgeregression} 
with the \gls{bias} term \eqref{equ_def_bias_term} of ordinary \gls{linreg} (which is the extreme 
case of \gls{ridgeregression} with $\regparam=0$). The \gls{bias} term \eqref{equ_bias_reg_lin_reg_approx} increases 
with increasing \gls{regularization} parameter $\regparam$ in \gls{ridgeregression} \eqref{equ_rerm_ridge_regression}. 
Sometimes the increase in \gls{bias} is outweighed by the reduction in \gls{variance}. The \gls{variance} 
typically decreases with increasing $\regparam$ as shown next.

The \gls{variance} of \gls{ridgeregression} \eqref{equ_rerm_ridge_regression} satisfies 
\begin{align}
	\label{equ_variance_reglinreg}
	\varianceterm & =( \sigma^{2}/\samplesize^{2}) \times \nonumber \\ 
	& \hspace*{-7mm} {\rm tr} \big\{  \expect \{  ( (1/\samplesize) \featuremtx^{T} \featuremtx\!+\!\regparam \mathbf{I} )^{-1} \featuremtx^{T} \featuremtx ( (1/\samplesize) \featuremtx^{T} \featuremtx\!+\!\regparam \mathbf{I} )^{-1}   \} \big\}.
\end{align}
Inserting the approximation \eqref{equ_approx_Gram_large_N} into \eqref{equ_variance_reglinreg},  
\begin{equation}
	\label{equ_variance_reglinreg_approx}
	\varianceterm \approx   ( \sigma^{2}/\samplesize^{2})  {\rm tr} \big\{  \expect \{  ( \mathbf{I}\!+\!\regparam \mathbf{I} )^{-1} \featuremtx^{T} \featuremtx ( \mathbf{I} \!+\!\regparam \mathbf{I} )^{-1}   \} \big\}
	= \sigma^{2} (1/\samplesize) (\featuredim/(1\!+\!\regparam)^2).
\end{equation} 
According to \eqref{equ_variance_reglinreg_approx}, the \gls{variance} of $\widehat{\weights}^{(\regparam)}$ 
decreases with increasing regularization parameter $\regparam$ of \gls{ridgeregression} \eqref{equ_rerm_ridge_regression}. 
This is the opposite behaviour as observed for the \gls{bias} \eqref{equ_bias_reg_lin_reg_approx}, which increases 
with increasing $\regparam$. By comparing the \gls{variance} approximation \eqref{equ_variance_reglinreg_approx} 
with the \gls{variance} \eqref{equ_formulae_variance_toy_model} of \gls{linreg} suggests to interpret the ratio $\featuredim/(1\!+\!\regparam)^2$ as an effective number of features used by \gls{ridgeregression}. Increasing 
the regularization parameter $\regparam$ decreases the effective number of features. 

Figure \ref{fig_bias_variance_lambda} illustrates the trade-off between the \gls{bias} $\biasterm^{2}$ \eqref{equ_bias_reg_lin_reg_approx} of \gls{ridgeregression}, which increases for increasing $\regparam$, 
and the \gls{variance} $\varianceterm$ \eqref{equ_variance_reglinreg_approx} 
which decreases with increasing $\regparam$. Note that we have seen another example for a bias-variance 
trade-off in Section \ref{sec_gen_linreg}. This trade-off was traced out by a discrete (model complexity) 
parameter $\modelidx \in \{1,2,\ldots\}$ (see \eqref{equ_generalization_hypospace_r}). In stark contrast to discrete 
model selection, the \gls{bias}-\gls{variance} trade-off for \gls{ridgeregression} is traced out by the continuous regularization 
parameter $\regparam \in \mathbb{R}_{+}$.

\begin{figure}
	\begin{center}
		\begin{tikzpicture}
			\draw [thick, <->] (0,3.5) -- (0,0) -- (6.2,0);
			\draw[red, ultra thick, domain=0:5] plot (\x,  {2*\x/(1+\x)}) node [above] {bias of $\widehat{\weights}^{(\regparam)}$};
			\draw[blue, ultra thick, domain=0:5] plot (\x,  {2/(1+\x)}) node [above]  {variance of $\widehat{\weights}^{(\regparam)}$};
			\node [below] at (5,0) {regularization parameter $\regparam$};
		\end{tikzpicture}
	\end{center}
	\caption{The \gls{bias} and \gls{variance} of ridge regression \eqref{equ_rerm_ridge_regression} 
		depend on the regularization parameter $\regparam$ in an opposite manner resulting in a bias-variance trade-off.}
	\label{fig_bias_variance_lambda}
\end{figure}


The main statistical effect of the regularization term in \gls{ridgeregression} is to 
balance the bias with the variance to minimize the average prediction error of 
the learnt hypothesis. There is also a computational effect or adding a regularization term. 
Roughly speaking, the regularization term serves as a pre-conditioning of the optimization 
problem and, in turn, reduces the computational complexity of solving \gls{ridgeregression} \eqref{equ_def_rlr_w_opt}. 

The objective function in \eqref{equ_def_rlr_w_opt} is a smooth (infinitely often differentiable) convex function. 
We can therefore use \gls{gd} to solve \eqref{equ_def_rlr_w_opt} efficiently (see Chapter \ref{ch_GD}). Algorithm \ref{alg:gd_reglinreg} summarizes the application of \gls{gd} to \eqref{equ_def_rlr_w_opt}. The computational 
complexity of Algorithm \ref{alg:gd_reglinreg} depends crucially on the number of \gls{gd} iterations required 
to reach a sufficiently small neighbourhood of the solutions to \eqref{equ_def_rlr_w_opt}. 
Adding the regularization term $\regparam \| \weights \|^{2}_{2}$ to the objective function of \gls{linreg} 
speeds up \gls{gd}. To verify this claim, we first rewrite \eqref{equ_def_rlr_w_opt} as the quadratic problem 
\begin{align} 
	\label{equ_quadr_form_reglinreg}
	\min_{\weights \in \mathbb{R}^{\featuredim}} & \underbrace{(1/2) \weights^{T} \mQ \weights - \vq^{T}  \weights}_{= f(\weights)} \nonumber \\ 
	& \mbox{ with } \mathbf{Q}= (1/\samplesize) \mX^{T} \mX + \regparam \mathbf{I}, \vq =(1/\samplesize) \mX^{T} \labelvec. 
\end{align} 
This is similar to the quadratic optimization problem \eqref{equ_quadr_form_linreg} underlying linear regression but 
with a different matrix $\mQ$. The computational complexity (number of iterations) required by \gls{gd} (see \eqref{equ_def_GD_step}) to solve \eqref{equ_quadr_form_reglinreg} up to a prescribed accuracy depends 
crucially on the \gls{condnr} $\kappa(\mQ) \geq 1$ of the \gls{psd} matrix $\mQ$ \cite{JungFixedPoint}. 
The smaller the \gls{condnr} $\kappa(\mQ)$, the fewer iterations are required 
by \gls{gd}. We refer to a matrix with a small \gls{condnr} as being ``well-conditioned''. 

The \gls{condnr} of the matrix $\mQ$ in \eqref{equ_quadr_form_reglinreg} is given by 
\begin{equation} 
	\label{equ_def_cond_number_ridgereg}
	\kappa(\mQ) = \frac{\eigval{\rm max}((1/\samplesize)\mX^{T} \mX) + \regparam} {\eigval{\rm min}((1/\samplesize)\mX^{T} \mX)+ \regparam}. 
\end{equation} 
According to \eqref{equ_def_cond_number_ridgereg}, the \gls{condnr} $\kappa(\mQ)$ tends to one 
for increasing regularization parameter $\regparam$,
\begin{equation}
	\label{equ_cond_numer_goes_1_rlr}
	\lim_{\regparam \rightarrow \infty} \frac{\eigval{\rm max}((1/\samplesize)\mX^{T} \mX) + \regparam} {\eigval{\rm min}((1/\samplesize)\mX^{T} \mX)+ \regparam} =1. 
\end{equation} 
Thus, the number of required \gls{gd} iterations in Algorithm \ref{alg:gd_reglinreg} decreases with increasing 
regularization parameter $\regparam$.

\begin{algorithm}[htbp]
	\caption{Regularized \Gls{linreg} via \gls{gd}}\label{alg:gd_reglinreg}
	\begin{algorithmic}[1]
		\renewcommand{\algorithmicrequire}{\textbf{Input:}}
		\renewcommand{\algorithmicensure}{\textbf{Output:}}
		\Require  dataset $\dataset=\{ (\featurevec^{(\sampleidx)}, \truelabel^{(\sampleidx)}) \}_{\sampleidx=1}^{\samplesize}$; \gls{gd} \gls{learnrate} $\lrate >0$. 
		\Statex\hspace{-6mm}{\bf Initialize:}set $\weights^{(0)}\!\defeq\!\mathbf{0}$; set iteration counter $\itercntr\!\defeq\!0$   
		\Repeat 
		\State $\itercntr \defeq \itercntr +1$    (increase iteration counter) 
		\State  $\weights^{(\itercntr)} \defeq (1- \lrate \regparam) \weights^{(\itercntr\!-\!1)} + \lrate (2/\samplesize) \sum_{\sampleidx=1}^{\samplesize} (\truelabel^{(\sampleidx)} - \big(\weights^{(\itercntr\!-\!1)})^{T} \featurevec^{(\sampleidx)}) \featurevec^{(\sampleidx)}$  (do a GD step \eqref{equ_def_GD_step})
		\Until stopping criterion met 
		\Ensure $\weights^{(\itercntr)}$ (which approximates $\widehat{\weights}^{(\regparam)}$ in \eqref{equ_def_rlr_w_opt})
	\end{algorithmic}
\end{algorithm}

\section{Semi-Supervised Learning} 
\label{sec_ssl_regularization}
Consider the task of predicting the numeric label $y$ of a \gls{datapoint} $\vz=\big(\featurevec,y\big)$ based on its feature vector $\featurevec\!=\!\big(x_{1},\ldots,x_{\featurelen}\big)^{T} \in \mathbb{R}^{\featurelen}$. At our disposal are two datasets $\dataset^{(u)}$ 
and $\dataset^{(l)}$. For each datapoint in $\dataset^{(u)}$ we only know the feature vector. We therefore refer 
to $\dataset^{(u)}$ as ``unlabelled data''. For each datapoint in $\dataset^{(l)}$ we know both, the feature vector $\featurevec$ 
and the label $y$. We therefore refer to $\dataset^{(l)}$ as ``labeled data''. 

\Gls{ssl} methods exploit the information provided by unlabelled data $\dataset^{(u)}$ to 
support the learning of a hypothesis based on minimizing its \gls{emprisk} on the labelled (training) data 
$\dataset^{(l)}$. The success of \gls{ssl} methods depends on the statistical properties of the data generated 
within a given application domain. Loosely speaking, the information provided by the probability distribution 
of the features must be relevant for the ultimate task of predicting the label $y$ from the the features $\featurevec$ \cite{SemiSupervisedBook}. 

Let us design a \gls{ssl} method, summarized in Algorithm \ref{alg:ssl_linreg} below, using the \gls{dataug} perspective 
from Section \ref{sec_data_augmentation}. The idea is the augment the (small) labeled dataset $\dataset^{(l)}$ by 
adding random perturbations fo the features vectors of \gls{datapoint} in $\dataset^{(l)}$. This is reasonable for 
applications where feature vectors are subject to inherent measurement or modelling errors. Given a \gls{datapoint} 
with vector $\featurevec$ we could have equally well observed a feature vector $\featurevec + {\bm \varepsilon}$ with some small 
random perturbation ${\bm \varepsilon} \sim \mathcal{N}(\mathbf{0}, \mathbf{C})$. To estimate the covariance 
matrix $\mC$, we use the sample covariance matrix of the feature vectors in the (large) unlabelled dataset $\dataset^{(u)}$. 
We then learn a hypothesis using the augmented (regularized) ERM \eqref{equ_def_ERM_funs_aug_approx_ridge_cov}.

\begin{algorithm}[htbp]
	\caption{A Semi-Supervised Learning Algorithm}\label{alg:ssl_linreg}
	\begin{algorithmic}[1]
		\renewcommand{\algorithmicrequire}{\textbf{Input:}}
		\renewcommand{\algorithmicensure}{\textbf{Output:}}
		\Require  labeled dataset $\dataset^{(l)}=\{ (\featurevec^{(\sampleidx)}, y^{(\sampleidx)}) \}_{\sampleidx=1}^{\samplesize}$;  unlabeled dataset $\dataset^{(u)}=\{ \widetilde{\featurevec}^{(\sampleidx)} \}_{\sampleidx=1}^{\samplesize'}$
		\State compute $\mC$ via sample covariance on $\dataset^{(u)}$, 
		\begin{equation} 
			\mC \defeq (1/\samplesize') \sum_{\sampleidx=1}^{\samplesize'} \big(\widetilde{\featurevec}^{(\sampleidx)}\!-\!\widehat{\featurevec} \big) \big(\widetilde{\featurevec}^{(\sampleidx)}\!-\!\widehat{\featurevec} \big)^{T}  \mbox{ with }  \widehat{\featurevec} \defeq (1/\samplesize') \sum_{\sampleidx=1}^{\samplesize'} \widetilde{\featurevec}^{(\sampleidx)}. 
		\end{equation}
		\State  compute (e.g. using \gls{gd}) 
		\begin{align}
			\label{equ_def_ERM_funs_aug_approx_ridge_cov_weights}
			\widehat{\vw}  \defeq \argmin_{\vw \in \mathbb{R}^{\featuredim}} \bigg[  (1/\samplesize) \sum_{\sampleidx=1}^{\samplesize} \big( y^{(\sampleidx)} - \weights^{T}\featurevec^{(\sampleidx)} \big) ^{2}  +  \vw^{T}  \mC  \vw \bigg] .
		\end{align}
		\Ensure hypothesis $\hat{h}(\featurevec) = \big( 	\widehat{\vw} )^{T} \featurevec$ 
	\end{algorithmic}
\end{algorithm}

%
%
%
%

\section{Multitask Learning} 
\label{sec_mtl_regularization}

Consider a specific learning task of finding a hypothesis $h$ with minimum (expected) loss $\loss{(\featurevec,\truelabel)}{h}$. 
Note that the loss incurred by $h$ for a specific \gls{datapoint} depends on the definition for the label of a \gls{datapoint}. 
We can obtain different learning tasks for the same \gls{datapoint}s by using different choices or 
definitions for the label of a \gls{datapoint}. Multitask learning exploits the similarities between different 
learning tasks to jointly solve them. Let us next discuss a simple example of a multitask learning problem. 

Consider a \gls{datapoint} $\vz$ representing a hand-drawing that is collected via 
the online game \url{https://quickdraw.withgoogle.com/}. The features of a \gls{datapoint} are 
the pixel intensities of the bitmap which is used to store the hand-drawing. As label we could 
use the fact if a hand-drawing shows an apple or not. This results in the learning task $\task^{(1)}$. 
Another choice for the label of a hand-drawing could be the fact if a hand-drawing shows a fruit 
at all or not. This results in another learning task $\task^{(2)}$ which is similar but 
different from the task $\task^{(1)}$. 

The idea of multitask learning is that a reasonable hypothesis $h$ for a learning task 
should also do well for a related learning tasks. Thus, we can use the loss incurred 
on similar learning tasks as a regularization term for learning a hypothesis for the 
learning task at hand. Algorithm \ref{alg:mtl} is a straightforward implementation of 
this idea for a given dataset that gives rise to $\nrtasks$ related learning tasks $\task^{(1)},\ldots,\task^{(\nrtasks)}$. 
For each individual learning task $\task^{(\taskidx')}$ it uses the loss on the remaining learning 
tasks $\task^{(\taskidx)}$, with $\taskidx \neq \taskidx'$, as regularization term in \eqref{equ_def_ERM_mt}. 

\begin{algorithm}[htbp]
	\caption{A Multitask Learning Algorithm}\label{alg:mtl}
	\begin{algorithmic}[1]
		\renewcommand{\algorithmicrequire}{\textbf{Input:}}
		\renewcommand{\algorithmicensure}{\textbf{Output:}}
		\Require dataset $\dataset = \{\datapoint^{(1)},\ldots,\datapoint^{(\samplesize)} \}$; $\nrtasks$ learning tasks with \gls{lossfunc}s $\lossfun^{(1)},\ldots,\lossfun^{(\nrtasks)}$, \gls{hypospace} $\hypospace$
		\State learn a hypothesis $\hat{h}$ via
		\begin{align}
			\label{equ_def_ERM_mt}
			\hat{h} \defeq \argmin_{h \in \hypospace} \sum_{\taskidx=1}^{\nrtasks} \sum_{\sampleidx=1}^{\samplesize} \lossfun^{(\taskidx)}\big(\datapoint^{(\sampleidx)},h\big).
		\end{align}
		\Ensure hypothesis $\hat{h}$ 
	\end{algorithmic}
\end{algorithm}

The applicability of Algorithm \ref{alg:mtl} is somewhat limited as it aims at finding a single hypothesis that does 
well for all $\nrtasks$ learning tasks simultaneously. For certain application domains it might be more reasonable 
to not learn a single hypothesis for all learning tasks but to learn a separate hypothesis $h^{(\taskidx)}$ for each 
learning task $\taskidx=1,\ldots,\nrtasks$. However, these separate hypotheses typically might still share some 
structural similarities.\footnote{One important example for such a structural similarity in the case of linear predictors 
	$h^{(\taskidx)}(\featurevec) =\big(\weights^{(\taskidx)} \big)^{T} \featurevec$is that the parameter vectors $\vw^{(\nrtasks)}$ have a 
	small joint support. Requiring the parameter vectors to have a small joint support is equivalent to requiring the stacked vector $\widetilde{\weights}=\big(\weights^{(1)},\ldots,\weights^{(\nrtasks)}  \big) $ to be block (group) sparse \cite{BlockSparsityEldarTSP}.}  
We can enforce different notion of similarities between the hypotheses $h^{(\taskidx)}$ by adding a regularization term to
the loss functions of the tasks. 

Algorithm \ref{alg:mtl_reg} generalizes Algorithms \ref{alg:mtl} by learning a separate 
hypothesis for each task $\taskidx$ while requiring these hypotheses to be structurally similar. 
The structural (dis-)similarity between the hypotheses is measured by a regularization term $\regularizer$ 
in \eqref{equ_def_ERM_mt_reg}.

\begin{algorithm}[htbp]
	\caption{A Multitask Learning Algorithm}\label{alg:mtl_reg}
	\begin{algorithmic}[1]
		\renewcommand{\algorithmicrequire}{\textbf{Input:}}
		\renewcommand{\algorithmicensure}{\textbf{Output:}}
		\Require dataset $\dataset = \{\datapoint^{(1)},\ldots,\datapoint^{(\samplesize)} \}$ with $\nrtasks$ associated learning tasks with loss functions $\lossfun^{(1)},\ldots,\lossfun^{(\nrtasks)}$, \gls{hypospace} $\hypospace$
		\State learn a hypothesis $\hat{h}$ via
		\begin{align}
			\label{equ_def_ERM_mt_reg}
			\hat{h}^{(1)},\ldots,\hat{h}^{(\nrtasks)} \defeq \argmin_{h^{(1)},\ldots,h^{(\nrtasks)} \in \hypospace} \sum_{\taskidx=1}^{\nrtasks} \sum_{\sampleidx=1}^{\samplesize} \lossfun^{(\taskidx)}\big(\vz^{(\sampleidx)},h^{(\taskidx)}\big) + \regparam \regularizer \big(h^{(1)},\ldots,h^{(\nrtasks)}\big).
		\end{align}
		\Ensure hypotheses $\hat{h}^{(1)},\ldots,\hat{h}^{(\nrtasks)}$
	\end{algorithmic}
\end{algorithm}

\section{Transfer Learning} 
\label{sec_transfer_learning}
Regularization is also instrumental for transfer learning to capitalize on synergies between different 
related learning tasks \cite{Pan2010,howard-ruder-2018-universal}. Transfer learning is enabled by 
constructing regularization terms for a learning task by using the result of a previous leaning task. 
While multitask learning methods solve many related learning tasks simultaneously, transfer learning 
methods operate in a sequential fashion. 

Let us illustrate the idea of transfer learning using two learning tasks which differ signifcantly in their 
intrinsic difficulty. Informally, we consider a learning task to be easy if we can easily gather large 
amounts of labeled (training) data for that task. Consider the learning task $\task^{(1)}$ of predicting 
whether an image shows a cat or not. For this learning task we can easily gather a large \gls{trainset} 
$\dataset^{(1)}$ using via image collections of animals. Another (related) learning task $\task^{(2)}$ is 
to predict whether an image shows a cat of a particular breed, with a particular body height and with a 
specific age. The learning task $\task^{(2)}$ is more dificult than $\task^{(1)}$ since we have only 
a very limited amount of cat images for which we know the particular breed, body height and precise 
age of the depicted cat. 

\section{Exercises} 
\begin{exercise}[Ridge Regression is a Quadratic Problem.] 
	\label{ex_2_0}
	Consider the linear \gls{hypospace} consisting of linear maps parameterized by 
	\gls{weights} $\weights$. We try to find the best linear map by minimizing the regularized average 
	squared error loss (\gls{emprisk}) incurred on a \gls{trainset} 
	$$\dataset \defeq \big \{ (\featurevec^{(1)},\truelabel^{(1)}),(\featurevec^{(2)},\truelabel^{(2)}),\ldots,(\featurevec^{(\samplesize)},\truelabel^{(\samplesize)}) \big \}.$$ 
	Ridge reression augments the average squared error loss on $\dataset$ by the regularizer $\| \weights \|^{2}$, yielding the
	 following learning problem 
	$$ \min_{\weights \in \mathbb{R}^{\featurelen}} f(\weights) =  (1/\samplesize)\sum_{\sampleidx=1}^{\samplesize}\big( \truelabel^{(\sampleidx)} - \weights^{T} \featurevec^{(\sampleidx)} \big)  + \regparam \| \weights \|^{2}_{2}.$$
	Is it possible to rewrite the objective function $f(\weights)$ as a convex quadratic function 
	$f(\weights) = \weights^{T} \mathbf{C} \weights + \vb \weights + c$? If this is possible, how are the 
	matrix $\mathbf{C}$, vector $\vb$ and constant $c$ related to the feature vectors 
	and labels of the training data ? 
\end{exercise} 

\begin{exercise}[Regularization or Model Selection.]
	Consider \gls{datapoint}s, each characterized by $\featurelen=10$ features $\featurevec \in \mathbb{R}^{\featurelen}$ 
	and a single numeric label $\truelabel$. We want to learn a linear hypothesis $h(\featurevec) = \weights^{T} \featurevec$ by 
	minimizing the average squared error on the \gls{trainset} $\dataset$ of size $\samplesize=4$. 
	We could learn such a hypothesis by two approaches. The first approach is to split the dataset 
	into a \gls{trainset} and a \gls{valset}. Then we consider all models that consists of 
	linear hypotheses with weight vectors having at most two non-zero \gls{weights}. 
	Each of these models corresponds to a different subset of two \gls{weights} that might be 
	non-zero. Find the model resulting in the smallest validation errors (see Algorithm \ref{alg:validated_ERM}). 
	Compute the average loss of the resulting optimal linear hypothesis on some \gls{datapoint}s that have 
	neither been used in the \gls{trainset} nor the \gls{valset}. Compare this average loss (``test error'') with the 
	average loss obtained on the same \gls{datapoint}s by the hypothesis learnt by \gls{ridgeregression} \eqref{equ_rerm_ridge_regression}. 
\end{exercise} 

\chapter{Clustering} 
\label{ch_Clustering}

\begin{figure}[htbp]
	\begin{center}
		\begin{tikzpicture}[auto,scale=0.8]
			\draw [thick] (5,2.5) circle (0.1cm)node[anchor=west] {\hspace*{0mm}$\featurevec^{(3)}$};
			\draw [thick] (4,2) circle (0.1cm)node[anchor=west] {\hspace*{0mm}$\featurevec^{(4)}$};
			\draw [thick] (5,1) circle (0.1cm)node[anchor=west,above] {\hspace*{0mm}$\featurevec^{(2)}$};
			\draw [thick] (1,5)circle (0.1cm) node[anchor=west,above] {\hspace*{0mm}$\featurevec^{(1)}$};
			\draw [thick] (1,3.5)circle (0.1cm)node[anchor=west,above] {\hspace*{0mm}$\featurevec^{(5)}$};
			\draw [thick] (1,2.5) circle (0.1cm)node[anchor=west,above] {\hspace*{0mm}$\featurevec^{(6)}$};
			\draw [thick] (2,4) circle (0.1cm)node[anchor=west,above] {\hspace*{0mm}$\featurevec^{(7)}$};
			\draw[->] (-0.5,0) -- (6.5,0) node[right] {$\feature_{\rm g}$};
			\draw[->] (0,-0.5) -- (0,6.5) node[above] {$\feature_{\rm r}$};
		\end{tikzpicture}
	\end{center}
	\caption{Each circle represents an image which is characterized by its average redness $\feature_{\rm r}$ and average 
		greenness $\feature_{\rm g}$. The $\sampleidx$-th image is depicted by a circle located at the 
		point $\featurevec^{(\sampleidx)} = \big( \feature_{\rm r}^{(\sampleidx)}, \feature_{\rm g}^{(\sampleidx)}\big)^{T} \in \mathbb{R}^{2}$. 
		It seems that the images can be grouped into two clusters.
	} 
	\label{fig_scatterplot_clustering}
\end{figure}

So far we focused on ML methods that use the \gls{erm} principle and lean a hypothesis by minimizing the 
discrepancy between its predictions and the true labels on a training set. These methods are referred 
to as supervised methods as they require labeled \gls{datapoint}s for which the true label values have been 
determined by some human (who serves as a ``supervisor''). This and the following chapter discus ML methods 
which do not require to know the label of any \gls{datapoint}. These methods are often referred to as 
	``unsupervised'' since they do not require a ``supervisor'' to provide the label values for any \gls{datapoint}. 

One important family of unsupervised ML methods aim at clustering a given set of \gls{datapoint}s such as those 
depicted in Figure \ref{fig_scatterplot_clustering}.  The basic idea of clustering is to decompose a set of \gls{datapoint}s 
into few subsets or \gls{cluster}s that consist of similar \gls{datapoint}s. For the \gls{dataset} in Figure \ref{fig_scatterplot_clustering} 
it seems reasonable to define two clusters, one \gls{cluster} $\big\{ \featurevec^{(1)}, \featurevec^{(5)}, \featurevec^{(6)}, \featurevec^{(7)} \big\}$ 
and a second \gls{cluster} $\big\{ \featurevec^{(2)}, \featurevec^{(3)}, \featurevec^{(4)}, \featurevec^{(8)} \big\}$ . 

Formally, clustering methods learn a hypothesis that assign each \gls{datapoint} either to precisely 
one cluster (see Section \ref{sec_hard_clustering}) or several clusters with different degrees of belonging 
(see Section \ref{sec_soft_clustering}). Different clustering methods use different measures for 
the similarity between \gls{datapoint}s. For \gls{datapoint}s characterized by (numeric) Euclidean 
feature vectors, the similarity between \gls{datapoint}s can be naturally defined in terms of the 
Euclidean distance between feature vectors. Section \ref{sec_connect_clustering} discusses 
\gls{clustering} methods that use notions of similarity that are not based on a \gls{euclidspace}.

There is a strong conceptual link between \gls{clustering} methods and the \gls{classification} methods discussed 
in Chapter \ref{ch_some_examples}. Both type of methods learn a hypothesis that reads in the features 
of a \gls{datapoint} and delivers a prediction for some quantity of interest. In classification methods, this quantity 
of interest is some generic label of a \gls{datapoint}. For \gls{clustering} methods, this quantity of interest for a \gls{datapoint} 
is the \gls{cluster} assignment (for \gls{hardclustering}) of the \gls{dob} (for \gls{softclustering}). 
A main difference between \gls{clustering} and \gls{classification} is that clustering methods do 
not require the true label (cluster assignment or \gls{dob}) of a single \gls{datapoint}. 

Classification methods learn a good hypothesis via minimizing their average loss incurred on a training set 
of labeled \gls{datapoint}s. In contrast, \gls{clustering} methods do not have access to a single labeled \gls{datapoint}. 
To find the correct labels (cluster assignments) clustering methods rely solely on the intrinsic 
geometry of the \gls{datapoint}s. We will see that \gls{clustering} methods use this intrinsic geometry 
to define an \gls{emprisk} incurred by a candidate hypothesis. Like \gls{classification} methods, also clustering methods 
use an instance of the \gls{erm} principle (see Chapter \ref{ch_Optimization}) to find a good hypothesis (clustering). 

%
%

This chapter discusses two main flavours of \gls{clustering} methods: 
\begin{itemize} 
	\item \index{hard clustering}\gls{hardclustering} (see Section \ref{sec_hard_clustering}) 
	\item and \index{soft clustering}\gls{softclustering} methods (see Section \ref{sec_soft_clustering}).
\end{itemize}
Hard \gls{clustering} methods learn a hypothesis $h$ that reads in the feature vector $\featurevec$ of a 
\gls{datapoint} and delivers a predicted \gls{cluster} assignment $\hat{\truelabel}=h(\featurevec) \in \{1,\ldots,\nrcluster\}$. Thus, 
\index{hardclustering} assigns each \gls{datapoint} to one single \gls{cluster}. Section \ref{sec_hard_clustering} will 
discuss one of the most widely-used \gls{hardclustering} algorithms which is known as \gls{kmeans}. 

In contrast to \gls{hardclustering} methods, \gls{softclustering} methods assign each \gls{datapoint} 
to several \gls{cluster}s with varying \gls{dob}. These methods learn a hypothesis that delivers a vector 
$\hat{\labelvec}=\big(\hat{\truelabel}_{1},\ldots,\hat{\truelabel}_{\nrcluster}\big)^{T}$ with entry $\hat{\truelabel}_{\clusteridx} \in [0,1]$ being the 
predicted degree by which the \gls{datapoint} belongs to the $\clusteridx$-th cluster. 
Hard clustering is an extreme case of \gls{softclustering} where we enforce each \gls{dob} to 
take only values in $\{0,1\}$. Moreover, \gls{hardclustering} requires that for each \gls{datapoint} only of the 
corresponding \gls{dob} (one for each cluster) is non-zero. 

The main focus of this chapter is on methods that require \gls{datapoint}s being represented by numeric 
feature vectors (see Sections \ref{sec_hard_clustering} and \ref{sec_soft_clustering}). These methods define 
the similarity between \gls{datapoint}s using the Euclidean distance between their feature vectors. Some applications 
generate \gls{datapoint}s for which it is not obvious how to obtain numeric feature vectors such that their 
Euclidean distances reflect the similarity between \gls{datapoint}s. It is then desirable to use a more flexible 
notion of similarity which does not require to determine (useful) numeric feature vectors of \gls{datapoint}s. 

Maybe the most fundamental concept to represent similarities between \gls{datapoint}s is a similarity graph. 
The nodes of the similarity graph are the individual \gls{datapoint}s of a dataset. Similar \gls{datapoint}s are 
connected by edges (links) that might be assigned some weight that quantities the amount of similarity. 
Section \ref{sec_connect_clustering} discusses clustering methods that use a graph to represent similarities 
between \gls{datapoint}s. 


\newpage
\section{Hard Clustering with \gls{kmeans}}
\label{sec_hard_clustering}

Consider a dataset $\dataset$ which consists of $\samplesize$ \gls{datapoint}s that are indexed by $\sampleidx=1,\ldots,\samplesize$. 
The \gls{datapoint}s are characterized via their numeric feature vectors $\featurevec^{(\sampleidx)} \in \mathbb{R}^{\featuredim}$, for $\sampleidx=1,\ldots,\samplesize$. It will be convenient for the following discussion if we identify a \gls{datapoint} 
with its feature vector. In particular, we refer by $\featurevec^{(\sampleidx)}$ to the $\sampleidx$-th \gls{datapoint}. 
\Gls{hardclustering} methods decompose (or cluster) the dataset into a given number $\nrcluster$ of different clusters 
$\cluster^{(1)},\ldots,\cluster^{(\nrcluster)}$. These methods assign each \gls{datapoint} $\featurevec^{(\sampleidx)}$ 
to one and only one cluster $\cluster^{(\clusteridx)}$ with the cluster index $\clusteridx \in \{1,\ldots,\nrcluster\}$. 

Let us define for each \gls{datapoint} its label $\truelabel^{(\sampleidx)} \in \{1,\ldots,\nrcluster\}$ as the index of 
the cluster to which the $\sampleidx$th \gls{datapoint} actually belongs to. The $\clusteridx$-th cluster consists of 
all \gls{datapoint}s with $\truelabel^{(\sampleidx)}=\clusteridx$, 
\begin{equation} 
	\label{equ_def_individual_cluster}
	\cluster^{(\clusteridx)} \defeq \big\{ \sampleidx \in \{1,\ldots,\samplesize\} : \truelabel^{(\sampleidx)} = \clusteridx  \big\}.
\end{equation} 
We can interpret \gls{hardclustering} methods as ML methods that compute predictions $\hat{\truelabel}^{(\sampleidx)}$ 
for the (``correct'') cluster assignments $\truelabel^{(\sampleidx)}$. The predicted cluster assignments result in 
the predicted clusters   
\begin{equation} 
	\label{equ_def_individual_cluster}
	\widehat{\cluster}^{(\clusteridx)} \defeq \big\{ \sampleidx \in \{1,\ldots,\samplesize\} : \hat{\truelabel}^{(\sampleidx)} = \clusteridx  \big\}\mbox{, for } \clusteridx =1,\ldots,\nrcluster.
\end{equation} 
We now discuss a \gls{hardclustering} method which is known as \index{$k$-means}\gls{kmeans}. 
This method does not require the knowledge of the label or (true) cluster assignment $\truelabel^{(\sampleidx)}$ 
for any \gls{datapoint} in $\dataset$. This method computes predicted cluster assignments $\hat{\truelabel}^{(\sampleidx)}$ 
based solely from the intrinsic geometry of the feature vectors $\featurevec^{(\sampleidx)} \in \mathbb{R}^{\featuredim}$ 
for all $\sampleidx=1,\ldots,\samplesize$. Since it does not require any labeled \gls{datapoint}s, \gls{kmeans} is often 
referred to as being an unsupervised method. However, note that \gls{kmeans} requires the number $\nrcluster$ of 
clusters to be given as an input (or hyper-) parameter.

The \gls{kmeans} method represents the $\clusteridx$-th cluster $\widehat{\cluster}^{(\clusteridx)}$ 
by a representative feature vector $\clustermean^{(\clusteridx)} \in \mathbb{R}^{\featuredim}$. It seems 
reasonable to assign \gls{datapoint}s in $\dataset$ to clusters $\widehat{\cluster}^{(\clusteridx)}$ such 
that they are well concentrated around the cluster representatives $\clustermean^{(\clusteridx)}$. 
We make this informal requirement precise by defining the clustering error
\begin{equation}
	\label{equ_def_emp_risk_kmeans}
	\vspace*{-2mm}
	\emperror \big( \{\clustermean^{(\clusteridx)}\}_{\clusteridx=1}^{\nrcluster},\{\hat{y}^{(\sampleidx)}\}_{\sampleidx=1}^{\samplesize} \mid \dataset \big)
	=(1/\samplesize) \sum_{\sampleidx=1}^{\samplesize} {\left\|\vx^{(\sampleidx)}-\clustermean^{(\hat{y}^{(\sampleidx)})}\right\|^2}.
\end{equation} 
Note that the clustering error $\emperror$ \eqref{equ_def_emp_risk_kmeans} depends on both, the cluster 
assignments $\hat{\truelabel}^{(\sampleidx)}$, which define the cluster \eqref{equ_def_individual_cluster}, and the cluster 
representatives $\clustermean^{(\clusteridx)}$, for $\clusteridx=1,\ldots,\nrcluster$. 

Finding the optimal cluster means $\{\clustermean^{(\clusteridx)}\}_{\clusteridx=1}^{\nrcluster}$ and cluster assignments 
$\{\hat{\truelabel}^{(\sampleidx)}\}_{\sampleidx=1}^{\samplesize}$ that minimize the clustering error  
\eqref{equ_def_emp_risk_kmeans} is computationally challenging. The difficulty stems from the fact that the clustering error 
is a non-convex function of the cluster means and assignments. While jointly optimizing the cluster means and assignments 
is hard, separately optimizing either the cluster means for given assignments or vice-versa is easy. In what follows, we present 
simple closed-form solutions for these sub-problems. The $k$-means method simply combines these solutions in an alternating 
fashion. 

It can be shown that for given predictions (cluster assignments) $\hat{\truelabel}^{(\sampleidx)}$, the clustering 
error \eqref{equ_def_emp_risk_kmeans} is minimized by setting the cluster representatives equal to the {\bf cluster means} \cite{BishopBook}
\begin{equation}
	\label{eq_def_mean_optimal} 
	\clustermean^{(\clusteridx)}\defeq \big(1/|\widehat{\cluster}^{(\clusteridx)}| \big) \sum_{\hat{\truelabel}^{(\sampleidx)} = \clusteridx} \featurevec^{(\sampleidx)}.
\end{equation} 
To evaluate \eqref{eq_def_mean_optimal} we need to know the predicted cluster assignments $\hat{\truelabel}^{(\sampleidx)}$. 
The crux is that the optimal predictions $\hat{\truelabel}^{(\sampleidx)}$, in the sense of minimizing clustering error \eqref{equ_def_emp_risk_kmeans}, 
depend themselves on the choice for the cluster representatives $\clustermean^{(\clusteridx)}$. In particular, for 
given cluster representative $\clustermean^{(\clusteridx)}$ with $\clusteridx=1,\ldots,\nrcluster$, the clustering error 
is minimized by the cluster assignments
\begin{equation}
	\label{equ_def_clustera_assgt-nearst_mean} 
	\hat{\truelabel}^{(\sampleidx)} \in \argmin_{\clusteridx \in \{1,\ldots,\nrcluster\}} \big\| \featurevec^{(\sampleidx)} - \clustermean^{(\clusteridx)} \big\|. 
\end{equation} 
Here, we denote by $\argmin\limits_{\clusteridx' \in \{1,\ldots,\nrcluster\}} \| \featurevec^{(\sampleidx)} - \clustermean^{(\clusteridx')} \|$ the 
set of all cluster indices $\clusteridx \in \{1,\ldots,\nrcluster\}$ such that $ \| \featurevec^{(\sampleidx)} - \clustermean^{(\clusteridx)} \| =  \min_{\clusteridx' \in \{1,\ldots,\nrcluster\}} \| \featurevec^{(\sampleidx)} - \clustermean^{(\clusteridx')} \|$.

Note that \eqref{equ_def_clustera_assgt-nearst_mean} assigns the $\sampleidx$th datapoint to those  
cluster $\cluster^{(\clusteridx)}$ whose cluster mean $\clustermean^{(\clusteridx)}$ is nearest (in Euclidean distance) 
to $\featurevec^{(\sampleidx)}$. Thus, if we knew the optimal cluster representatives, we could predict the cluster 
assignments using \eqref{equ_def_clustera_assgt-nearst_mean}. However, we do not know the optimal cluster 
representatives unless we have found good predictions for the cluster assignments $\hat{\truelabel}^{(\sampleidx)}$ (see \eqref{eq_def_mean_optimal}). 

To recap: We have characterized the optimal choice \eqref{eq_def_mean_optimal} for the cluster representatives 
for given cluster assignments and the optimal choice \eqref{equ_def_clustera_assgt-nearst_mean} for the cluster 
assignments for given cluster representatives. It seems natural, starting from some initial guess for the cluster representatives, 
to alternate between the cluster assignment update \eqref{equ_def_clustera_assgt-nearst_mean} and the update \eqref{eq_def_mean_optimal} 
for the cluster means. This alternating optimization strategy is illustrated in Figure \ref{fig_flow_kmeans} and summarized in 
Algorithm \ref{alg:kmeans}. Note that Algorithm \ref{alg:kmeans}, which is maybe the most basic variant of $k$-means, 
simply alternates between the two updates \eqref{eq_def_mean_optimal} and \eqref{equ_def_clustera_assgt-nearst_mean} 
until some stopping criterion is satisfied.

\begin{figure} 
	\begin{center}
\tikzset{global scale/.style={
		scale=#1,
		every node/.append style={scale=#1}
	}
}
\begin{tikzpicture}[global scale = 1,           
	squarednode/.style={rectangle, draw=blue, line width=2pt, minimum size=15mm, font=\fontsize{15}{15}\selectfont, align=center},      
	myarrow/.style={line width = 1mm, draw = blue!80, -triangle 45, postaction={draw, line width = 2mm, shorten >=3mm, -}}]         
	
	
	\node[squarednode] (initial) {initial choice for \\ cluster means};      
	\node[squarednode, rounded corners] (ucassignment) [below of=initial, xshift=3cm, yshift=-1.5cm]  {update cluster \\ assignment \eqref{equ_cluster_assign_update}};       
	\node[squarednode, rounded corners] (ucmeans) [right=of ucassignment, xshift=0.5cm] {update cluster \\ means \eqref{equ_cluster_mean_update}};        
	\node[font=\fontsize{18}{0}\selectfont, above of=initial, yshift=0.5cm] (kmeans) {``k-Means''};        
	
	\draw[myarrow, rounded corners] (initial.east) -| ([yshift=0.2cm]ucassignment.north) [xshift=2cm, yshift=2cm];          
	\draw[myarrow] ([xshift=0.5cm, yshift=-0.5cm]ucassignment.south) to [bend right=45] ([xshift=-0.5cm, yshift=-0.5cm]ucmeans.south);        
	\draw[myarrow] ([xshift=-0.5cm, yshift=0.5cm]ucmeans.north) to [bend right=45] ([xshift=0.5cm, yshift=0.5cm]ucassignment.north);          
	
\end{tikzpicture}
	\end{center}
\caption{\label{fig_flow_kmeans} The workflow of $k$-means. Starting from an initial guess or estimate 
for the cluster means, the cluster assignments and cluster means are updated (improved) in an alternating fashion.}
\end{figure}

\begin{algorithm}[htbp]
	\caption{``\gls{kmeans}''}\label{alg:kmeans}
	\begin{algorithmic}[1]
		\renewcommand{\algorithmicrequire}{\textbf{Input:}}
		\renewcommand{\algorithmicensure}{\textbf{Output:}}
		\Require   dataset $\dataset=\{ \featurevec^{(\sampleidx)}\}_{\sampleidx=1}^{\samplesize}$; number $\nrcluster$ of clusters; 
		initial cluster means $\clustermean^{(\clusteridx)}$ for $\clusteridx=1,\ldots,\nrcluster$.
		\Repeat
		\State\label{equ_step_cluster_asst_update} for each datapoint $\featurevec^{(\sampleidx)}$, $\sampleidx\!=\!1,\ldots,\samplesize$, do  
		\begin{equation} 
			\label{equ_cluster_assign_update}
			\hat{\truelabel}^{(\sampleidx)} \defeq \argmin\limits_{\clusteridx' \in \{1,\ldots,\nrcluster\}} \| \featurevec^{(\sampleidx)} - \clustermean^{(\clusteridx')} \| \quad \mbox{  (update cluster assignments)} 
		\end{equation}
		\State\label{equ_step_cluster_update} for each cluster $\clusteridx\!=\!1,\ldots,\nrcluster$ do 
		\begin{equation}
			\label{equ_cluster_mean_update} 
			\clustermean^{(\clusteridx)} \defeq \frac{1}{|\{\sampleidx: \hat{\truelabel}^{(\sampleidx)}= \clusteridx\}|}  \sum_{\sampleidx: \hat{\truelabel}^{(\sampleidx)} = \clusteridx} \featurevec^{(\sampleidx)} \quad  \mbox{  (update cluster means) } 
		\end{equation} 
		\Until\label{equ_def_stop_criterion_kmeans} stopping criterion is met 
		\State compute final clustering error $\error^{(\nrcluster)} \defeq (1/\samplesize) \sum_{\sampleidx=1}^{\samplesize} {\left\|\featurevec^{(\sampleidx)}-\clustermean^{(\hat{\truelabel}^{(\sampleidx)})}\right\|^2}$
		\Ensure cluster means $\clustermean^{(\clusteridx)}$, for $\clusteridx=1,\ldots,\nrcluster$, cluster assignments $\hat{\truelabel}^{(\sampleidx)} \in \{1,\ldots,\nrcluster\}$, for $\sampleidx=1,\ldots,\samplesize$, final clustering error $\error^{(\nrcluster)}$ 
	\end{algorithmic}
\end{algorithm}

Algorithm \ref{alg:kmeans} requires the specification of the number $\nrcluster$ of clusters and initial choices for the 
cluster means $\clustermean^{(\clusteridx)}$, for $\clusteridx=1,\ldots,\nrcluster$. Those quantities are hyper-parameters 
that must be tuned to the specific geometry of the given dataset $\dataset$. This tuning can be based on probabilistic models 
for the dataset and its cluster structure (see Section \ref{equ_prob_models_data} and \cite{KulisJordan2012,WadeBayesianCluster2018}). 
Alternatively, if Algorithm \ref{alg:kmeans} is used as pre-processing within an overall supervised ML method (see Chapter \ref{ch_some_examples}), 
the validation error (see Section \ref{sec_modsel}) of the overall method might guide the choice of the number $\nrcluster$ of clusters. 

{\bf Choosing Number of Clusters.} The choice for the number $\nrcluster$ of clusters typically depends on the role of the 
clustering method within an overall ML application. If the clustering method serves as a pre-processing for a supervised ML problem,
we could try out different values of the number $\nrcluster$ and determine, for each choice $\nrcluster$, the corresponding validation 
error. We then pick the value of $\nrcluster$ which results in the smallest validation error. If the clustering 
method is mainly used as a tool for data visualization, we might prefer a small number of clusters.  
The choice for the number $\nrcluster$ of clusters can also be guided by the so-called ``elbow-method''. 
Here, we run the $k$-means Algorithm \ref{alg:kmeans} for several different choices of $\nrcluster$. For each 
value of $\nrcluster$, Algorithm \ref{alg:kmeans} delivers a clustering with clustering error 
$$\error^{(\nrcluster)} = \emperror \big(\{\clustermean^{(\clusteridx)}\}_{\clusteridx=1}^{\nrcluster},\{\hat{\truelabel}^{(\sampleidx)}\}_{\sampleidx=1}^{\samplesize} \mid \dataset \big).$$ 
We then plot the minimum empirical error $\error^{(\nrcluster)}$ as a function of the number $\nrcluster$ of clusters. 
Figure \ref{fig_ellbow} depicts an example for such a plot which typically starts with a steep decrease 
for increasing $\nrcluster$ and then flattening out for larger values of $\nrcluster$. Note that for $\nrcluster \geq \samplesize$ 
we can achieve zero clustering error since each datapoint $\featurevec^{(\sampleidx)}$ can be assigned to a separate cluster $\cluster^{(\clusteridx)}$ 
whose mean coincides with that datapoint, $\featurevec^{(\sampleidx)} = \clustermean^{(\clusteridx)}$. 

\begin{figure}[htbp]
	\begin{center}
		\begin{tikzpicture}
			\begin{axis}
				[ylabel=$\error^{(\nrcluster)}$,
				xlabel=$\nrcluster$]
				\addplot[domain=1:2, ultra thick] {10-x^3};
				\addplot[domain=2:10, ultra thick] {2+1/5-x/10};
			\end{axis}
		\end{tikzpicture}
	\end{center}
	\caption{The clustering error $\error^{(\nrcluster)}$ achieved by \gls{kmeans} for increasing number $\nrcluster$ of clusters.}
	\label{fig_ellbow}
\end{figure}

{\bf Cluster-Means Initialization.} We briefly mention some popular strategies for choosing the initial cluster means in Algorithm \ref{alg:kmeans}. 
One option is to initialize the cluster means with realizations of \gls{iid} random vectors whose probability distribution is 
matched to the dataset $\dataset = \{ \vx^{(\sampleidx)} \}_{\sampleidx=1}^{\samplesize}$ (see Section \ref{sec_max_iikelihood}). 
For example, we could use a multivariate normal distribution $\mathcal{N}(\vx;\widehat{\clustermean}, \widehat{\clustercov})$ 
with the sample mean $\widehat{\clustermean} = (1/\samplesize) \sum_{\sampleidx=1}^{\samplesize} \featurevec^{(\sampleidx)}$ and 
the sample covariance $\widehat{\clustercov} = (1/\samplesize) \sum_{\sampleidx=1}^{\samplesize} (\featurevec^{(\sampleidx)}\!-\!\widehat{\clustermean}) (\featurevec^{(\sampleidx)}\!-\!\widehat{\clustermean})^{T}$. Alternatively, we could choose the initial cluster means $\clustermean^{(\clusteridx)}$ 
by selecting $\nrcluster$ different \gls{datapoint}s $\featurevec^{(\sampleidx)}$ from $\dataset$. This selection process might combine 
random choices with an optimization of the distances between cluster means \cite{Arthur2007}. Finally, the cluster means 
might also be chosen by evenly partitioning the principal component of the dataset (see Chapter \ref{ch_FeatureLearning}).  

{\bf Interpretation as \gls{erm}.} For a practical implementation of Algorithm \ref{alg:kmeans} we need to decide 
when to stop updating the cluster means and assignments (see \eqref{equ_cluster_assign_update} and \eqref{equ_cluster_mean_update}). 
To this end it is useful to interpret Algorithm \ref{alg:kmeans} as a method for iteratively minimizing 
the clustering error \eqref{equ_def_emp_risk_kmeans}. As can be verified easily, the updates \eqref{equ_cluster_assign_update} and \eqref{equ_cluster_mean_update} always modify (update) the cluster means or assignments in such a way 
that the clustering error \eqref{equ_def_emp_risk_kmeans} is never increased. Thus, each new iteration 
of Algorithm \ref{alg:kmeans} results in cluster means and assignments with a smaller (or the same) clustering 
error compared to the cluster means and assignments obtained after the previous iteration. Algorithm \ref{alg:kmeans} 
implements a form of \gls{erm} (see Chapter \ref{ch_Optimization}) using the clustering error \eqref{equ_def_emp_risk_kmeans} 
as the \gls{emprisk} incurred by the predicted cluster assignments $\hat{y}^{(\sampleidx)}$. Note that after completing 
a full iteration of Algorithm \ref{alg:kmeans}, the cluster means $\big\{ \clustermean^{(\clusteridx)} \big\}_{\clusteridx=1}^{\nrcluster}$ 
are fully determined by the cluster assignments $\big\{ \hat{y}^{(\sampleidx)} \big\}_{\sampleidx=1}^{\samplesize}$ via \eqref{equ_cluster_mean_update}. 
It seems natural to terminate Algorithm \ref{alg:kmeans} if the decrease in the clustering error achieved by 
the most recent iteration is below a prescribed (small) threshold. 


{\bf Clustering and Classification.} There is a strong conceptual link between Algorithm \ref{alg:kmeans} and 
classification methods (see e.g. Section \ref{sec_nearest_neighbour_methods}). Both methods essentially 
learn a hypothesis $h(\featurevec)$ that maps the feature vector $\featurevec$ to a predicted label $\hat{\truelabel}=h(\featurevec)$ from 
a finite set. The practical meaning of the label values is different for Algorithm \ref{alg:kmeans} 
and classification methods. For classification methods, the meaning of the label values is essentially 
defined by the training set (of labeled \gls{datapoint}s) used for \gls{erm} \eqref{equ_def_ERM_funs}. On the other hand, clustering methods 
use the predicted label $\hat{\truelabel}=h(\featurevec)$ as a cluster index. 

Another main difference between Algorithm \ref{alg:kmeans} and most classification methods is the choice for the 
\gls{emprisk} used to evaluate the quality or usefulness of a given hypothesis $h(\cdot)$. Classification methods 
typically use an average loss over labeled \gls{datapoint}s in a training set as \gls{emprisk}. In contrast, Algorithm \ref{alg:kmeans} 
uses the clustering error \eqref{equ_def_emp_risk_kmeans} as a form of \gls{emprisk}. Consider a hypothesis that resembles 
the cluster assignments $\hat{\truelabel}^{(\sampleidx)}$ obtained after completing an iteration in Algorithm \ref{alg:kmeans}, 
$\hat{\truelabel}^{(\sampleidx)} = h\big( \featurevec^{(\sampleidx)} \big)$. Then we can rewrite the resulting clustering error 
achieved after this iteration as  
\begin{equation}
	\label{equ_emp_risk_clustering}
	\emperror\big( h | \dataset \big) = (1/\samplesize) \sum_{\sampleidx=1}^{\samplesize} \left\|\featurevec^{(\sampleidx)}-  \frac{\sum_{\sampleidx' \in \dataset^{(\sampleidx)}} \featurevec^{(\sampleidx')}}{\big| \dataset^{(\sampleidx)} \big|} \right\|^2. \mbox{ with } \dataset^{(\sampleidx)} \defeq \big\{ \sampleidx' : h\big(\featurevec^{(\sampleidx)} \big) =  h\big(\featurevec^{(\sampleidx')} \big) \big\}.
\end{equation}
Note that the $\sampleidx$-th summand in \eqref{equ_emp_risk_clustering} depends on 
the entire \gls{dataset} $\dataset$ and not only on (the features of) the $\sampleidx$-th \gls{datapoint} $\featurevec^{(\sampleidx)}$. 

{\bf Some Practicalities.} For a practical implementation of Algorithm \ref{alg:kmeans} we need to fix three issues. 
\begin{itemize} 
	\item Issue 1 (``tie-breaking''): We need to specify what to do if several different cluster indices $\clusteridx\!\in\!\{1,\ldots,\nrcluster\}$ 
	achieve the minimum value in the cluster assignment update \eqref{equ_cluster_assign_update} during step \ref{equ_step_cluster_asst_update}. 
	\item Issue 2 (``empty cluster''): The cluster assignment update  \eqref{equ_cluster_assign_update} in step \ref{equ_step_cluster_update} of Algorithm \ref{alg:kmeans} might result in a cluster $\clusteridx$ with no datapoints associated with it, $|\{ \sampleidx: \hat{y}^{(\sampleidx)} = \clusteridx \}|=0$. 
	For such a cluster $\clusteridx$, the update \eqref{equ_cluster_mean_update} is not well-defined. 
	\item Issue 3 (``stopping criterion''): We need to specify a criterion used in step \ref{equ_def_stop_criterion_kmeans} of Algorithm \ref{alg:kmeans} 
	to decide when to stop iterating. 
\end{itemize}
Algorithm \ref{alg:kmeansimpl} is obtained from Algorithm \ref{alg:kmeans} by fixing those three issues \cite{Gray1980}. Step \ref{equ_step_cluster_assg_kmeans2} of Algorithm \ref{alg:kmeansimpl} solves the first issue mentioned above (``tie breaking''), 
arising when there are several cluster clusters whose means have minimum distance to a \gls{datapoint} $\featurevec^{(\sampleidx)}$, 
by assigning $\vx^{(\sampleidx)}$ to the cluster with smallest cluster index (see \eqref{equ_cluster_assign_update2}).  
\begin{algorithm}[htbp]
	\caption{``$k$-Means II'' (slight variation of ``Fixed Point Algorithm'' in \cite{Gray1980})}\label{alg:kmeansimpl}
	\begin{algorithmic}[1]
		\renewcommand{\algorithmicrequire}{\textbf{Input:}}
		\renewcommand{\algorithmicensure}{\textbf{Output:}}
		\Require dataset  $\dataset=\{ \featurevec^{(\sampleidx)}\}_{\sampleidx=1}^{\samplesize}$; 
		number $\nrcluster$ of \gls{cluster}s; tolerance $\varepsilon \geq 0$; initial \gls{cluster} 
		means $\big\{ \clustermean^{(\clusteridx)} \big\}_{\clusteridx=1}^{\nrcluster}$ 
		\State {\bf Initialize.} set iteration counter $\itercntr \defeq 0$;  $\error_{0}\defeq 0$
		\Repeat 
		\State\label{equ_step_cluster_assg_kmeans2} for all datapoints $\sampleidx\!=\!1,\ldots,\samplesize$, 
		\begin{equation} 
			\label{equ_cluster_assign_update2}
			\hat{\truelabel}^{(\sampleidx)} \defeq \min \{  \argmin\limits_{\clusteridx' \in \{1,\ldots,\nrcluster\}} \| \featurevec^{(\sampleidx)} - \clustermean^{(\clusteridx')} \| \} \quad \mbox{  (update cluster assignments)} 
		\end{equation}
		\State\label{equ_def_cluster_indicators} for all clusters $\clusteridx\!=\!1,\ldots,\nrcluster$, update 
		the activity indicator 
		$$b^{(\clusteridx)} \defeq \begin{cases} 1 & \mbox{ if } |\{\sampleidx: \hat{y}^{(\sampleidx)}= \clusteridx\}| > 0 \\ 0 & \mbox{ else.} \end{cases}$$
		\State for all $\clusteridx\!=\!1,\ldots,\nrcluster$ with $b^{(\clusteridx)}=1$,  
		\begin{equation}
			\label{equ_cluster_mean_update2} 
			\clustermean^{(\clusteridx)} \defeq \frac{1}{|\{ \sampleidx: \hat{\truelabel}^{(\sampleidx)}= \clusteridx\}|}  \sum_{\{ \sampleidx: \hat{\truelabel}^{(\sampleidx)} = \clusteridx\}} \featurevec^{(\sampleidx)}  \quad \mbox{  (update cluster means) } 
		\end{equation} 
		\State $\itercntr \defeq \itercntr+1$  (increment iteration counter) 
		\vspace*{4mm}
		\State\label{equ_def_clustering_error_kmeans2} $\error_{\itercntr} \defeq \emperror \big( \{\clustermean^{(\clusteridx)}\}_{\clusteridx=1}^{\nrcluster},\{\hat{\truelabel}^{(\sampleidx)}\}_{\sampleidx=1}^{\samplesize} \mid \dataset \big)$  (evaluate clustering error \eqref{equ_def_emp_risk_kmeans})
		\vspace*{4mm}
		\Until\label{step_stop_criterion_kmeans2} $\itercntr > 1$ \mbox{and} $\error_{\itercntr\!-\!1} - \error_{\itercntr} \leq \varepsilon$  (check for sufficient decrease in clustering error)
		\vspace*{4mm}
		\State\label{equ_def_step_final_clustering_kmeans2}$\error^{(\nrcluster)} \defeq (1/\samplesize) \sum_{\sampleidx=1}^{\samplesize} {\left\|\featurevec^{(\sampleidx)}-\clustermean^{(\hat{\truelabel}^{(\sampleidx)})}\right\|^2}$ (compute final clustering error)
		\Ensure cluster assignments $\hat{\truelabel}^{(\sampleidx)}\!\in\!\{1,\ldots,\nrcluster\}$, cluster means $\clustermean^{(\clusteridx)}$, 
		clustering error $\error^{(\nrcluster)}$. 
	\end{algorithmic}
\end{algorithm}
Step \ref{equ_def_cluster_indicators} of Algorithm \ref{alg:kmeansimpl} resolves the ``empty cluster'' issue 
by computing the variables $b^{(\clusteridx)} \in \{0,1\}$ for $\clusteridx=1,\ldots,\nrcluster$. The variable $b^{(\clusteridx)}$ 
indicates if the cluster with index $\clusteridx$ is active ($b^{(\clusteridx)}= 1$) or the cluster $\clusteridx$ is inactive 
($b^{(\clusteridx)}=0$). The cluster $\clusteridx$ is defined to be inactive if there are no \gls{datapoint}s assigned to it 
during the preceding cluster assignment step \eqref{equ_cluster_assign_update2}. The cluster activity indicators 
$b^{(\clusteridx)}$ allows to restrict the cluster mean updates \eqref{equ_cluster_mean_update2} only to the clusters 
$\clusteridx$ with at least one \gls{datapoint} $\featurevec^{(\sampleidx)}$. To obtain a stopping criterion, step \ref{equ_def_clustering_error_kmeans2} 
Algorithm \ref{alg:kmeansimpl} monitors the clustering error $\error_{\itercntr}$ incurred by the 
cluster means and assignments obtained after $\itercntr$ iterations. Algorithm \ref{alg:kmeansimpl} continues updating cluster assignments \eqref{equ_cluster_assign_update2} and cluster means \eqref{equ_cluster_mean_update2} as long as the decrease 
is above a given threshold $\varepsilon \geq0$. 

For Algorithm \ref{alg:kmeansimpl} to be useful we must ensure that the stopping criterion is met within a finite number of iterations. 
In other words, we must ensure that the clustering error decrease can be made arbitrarily small within a sufficiently 
large (but finite) number of iterations. To this end, it is useful to represent Algorithm \ref{alg:kmeansimpl} as a 
fixed-point iteration 
\begin{equation}
	\label{equ_fixed_point_clustering}
	\{ \hat{\truelabel}^{(\sampleidx)} \}_{\sampleidx=1}^{\samplesize} \mapsto \mathcal{P}  \{ \hat{\truelabel}^{(\sampleidx)} \}_{\samplesize=1}^{\samplesize}. 
\end{equation}
The operator $\mathcal{P}$, which depends on the dataset $\dataset$, reads in a list of cluster assignments and delivers 
an improved list of cluster assignments aiming at reducing the associated clustering error \eqref{equ_def_emp_risk_kmeans}. 
Each iteration of Algorithm \ref{alg:kmeansimpl} updates the cluster assignments $\hat{\truelabel}^{(\sampleidx)}$ by applying the operator $\mathcal{P}$. Representing Algorithm \ref{alg:kmeansimpl} as a fixed-point iteration \eqref{equ_fixed_point_clustering} allows for 
an elegant proof of the convergence of Algorithm \ref{alg:kmeansimpl} within a finite number of iterations (even for $\varepsilon = 0$) \cite[Thm. 2]{Gray1980}. 

\begin{figure}[htbp]
	\begin{center}
		\includegraphics[width=14cm]{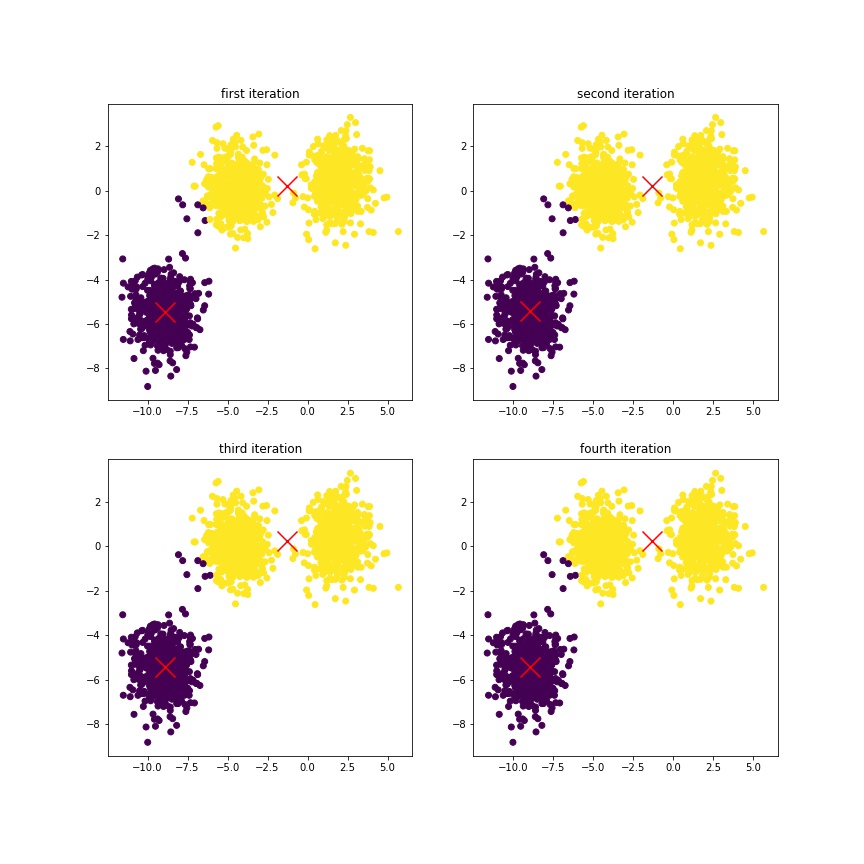}  
		\vspace*{-6mm}
	\end{center}
	\caption{The evolution of cluster means \eqref{equ_cluster_mean_update} and cluster assignments \eqref{equ_cluster_assign_update} 
		(depicted as large dot and large cross) during the first four iterations of \gls{kmeans} Algorithm \ref{alg:kmeansimpl}.}
	\label{fig:first_iter_kmeans}
\end{figure}

Figure \ref{fig:first_iter_kmeans} depicts the evolution of the cluster assignments and cluster means 
during the iterations Algorithm \ref{alg:kmeansimpl}. Each subplot corresponds to one iteration of 
Algorithm \ref{alg:kmeansimpl} and depicts the cluster means before that iteration and the 
clustering assignments (via the marker symbols) after the corresponding iteration. In particular, the 
upper left subplot depicts the cluster means before the first iteration (which are the initial cluster means) 
and the cluster assignments obtained after the first iteration of Algorithm \ref{alg:kmeansimpl}. 

Consider running Algorithm \ref{alg:kmeansimpl} with tolerance $\varepsilon=0$ (see step \ref{step_stop_criterion_kmeans2}) such that the 
iterations are continued until there is no decrease in the clustering error $\error^{(\itercntr)}$ (see step \ref{equ_def_clustering_error_kmeans2} 
of Algorithm \ref{alg:kmeansimpl}). As discussed above, Algorithm \ref{alg:kmeansimpl} will terminate after a finite number of iterations. 
Moreover, for $\varepsilon=0$, the delivered cluster assignments $\big\{ \hat{\truelabel}^{(\sampleidx)} \big\}_{\sampleidx=1}^{\samplesize}$ 
are fully determined by the delivered clustered means $\big\{ \clustermean^{(\clusteridx)} \big\}_{\clusteridx=1}^{\nrcluster}$, 
\begin{equation} 
	\label{equ_condition_clusterassmean_kmeans2}
	\hat{\truelabel}^{(\sampleidx)} = \min \{  \argmin\limits_{\clusteridx' \in \{1,\ldots,\nrcluster\}} \| \featurevec^{(\sampleidx)} - \clustermean^{(\clusteridx')} \| \}.
\end{equation} 
Indeed, if \eqref{equ_condition_clusterassmean_kmeans2} does not hold one can show the final iteration $r$ would still 
decrease the clustering error and the stopping criterion in step \ref{step_stop_criterion_kmeans2} would not be met. 

If cluster assignments and cluster means satisfy the condition \eqref{equ_condition_clusterassmean_kmeans2}, 
we can rewrite the clustering error \eqref{equ_def_emp_risk_kmeans} as a function of the cluster means solely, 
\begin{equation}
	\label{equ_def_clustering_error_means}
	\emperror\big( \big\{ \clustermean^{(\clusteridx)} \big\}_{\clusteridx=1}^{\nrcluster} | \dataset \big) \defeq (1/\samplesize) \sum_{\sampleidx=1}^{\samplesize} 
	\min\limits_{\clusteridx' \in \{1,\ldots,\nrcluster\}} \| \featurevec^{(\sampleidx)} - \clustermean^{(\clusteridx')} \|^{2} .
\end{equation} 
Even for cluster assignments and cluster means that do not satisfy \eqref{equ_condition_clusterassmean_kmeans2}, 
we can still use \eqref{equ_def_clustering_error_means} to lower bound the clustering error \eqref{equ_def_emp_risk_kmeans}, 
$$\emperror\big( \big\{ \clustermean^{(\clusteridx)} \big\}_{\clusteridx=1}^{\nrcluster} | \dataset \big)\leq \emperror \big( \{\clustermean^{(\clusteridx)}\}_{\clusteridx=1}^{\nrcluster},\{\hat{\truelabel}^{(\sampleidx)}\}_{\sampleidx=1}^{\samplesize} \mid \dataset \big)$$. 

Algorithm \ref{alg:kmeansimpl} iteratively improves the cluster means in order to minimize \eqref{equ_def_clustering_error_means}. 
Ideally, we would like Algorithm \ref{alg:kmeansimpl} to deliver cluster means that achieve the global minimum of \eqref{equ_def_clustering_error_means} (see Figure \ref{fig_emp_risk_k_means}). However, for some combination of dataset $\dataset$ and initial cluster means, 
Algorithm \ref{alg:kmeansimpl} delivers cluster means that form only a local optimum of $\emperror\big( \big\{ \clustermean^{(\clusteridx)} \big\}_{\clusteridx=1}^{\nrcluster} | \dataset \big)$ which is strictly worse (larger) than its global optimum (see Figure \ref{fig_emp_risk_k_means}). 

The tendency of Algorithm \ref{alg:kmeansimpl} to get trapped around a local minimum of \eqref{equ_def_clustering_error_means} 
depends on the initial choice for cluster means. It is therefore useful to repeat Algorithm \ref{alg:kmeansimpl} several times, 
with each repetition using a different initial choice for the cluster means. We then pick the cluster assignments 
$\{ \hat{\truelabel}^{(\sampleidx)} \}_{\sampleidx=1}^{\samplesize}$ obtained for the repetition that resulted in the smallest 
clustering error $\error^{(\nrcluster)}$ (see step \ref{equ_def_step_final_clustering_kmeans2}). 

\begin{figure}[htbp]
	\begin{center}
		\begin{tikzpicture}
			\draw[  thick, domain=-1:2.5] plot (2*\x,  {\x^2-3*\x^3+\x+\x^4}) node [right] {$\emperror \big( \{\clustermean^{(\clusteridx)}\}_{\clusteridx=1}^{\nrcluster} \mid \dataset \big)$}; 
			\draw[  thick,fill=blue]  (-1.1,-0.45) circle (0.05cm) node[below] {local minimum}; 
			\draw[  thick,fill=blue]  (3.8,-2) circle (0.05cm) node[below] {global minimum}; 
		\end{tikzpicture}
	\end{center}
	\caption{The clustering error \eqref{equ_def_clustering_error_means} is a non-convex function of the \gls{cluster} means  $\{\clustermean^{(\clusteridx)}\}_{\clusteridx=1}^{\nrcluster}$. Algorithm \ref{alg:kmeansimpl} iteratively updates cluster 
		means to minimize the clustering error but might get trapped around one of its local minimum.} \label{fig_emp_risk_k_means}
\end{figure}

\newpage
\section{Soft Clustering with Gaussian Mixture Models}
\label{sec_soft_clustering}

Consider a dataset $\dataset=\{\featurevec^{(1)},\ldots,\featurevec^{(\samplesize)}\}$ 
that we wish to group into a given number of $\nrcluster$ different clusters. The \gls{hardclustering} 
methods discussed in Section \ref{sec_hard_clustering} deliver cluster assignments $\hat{\truelabel}^{(\sampleidx)}$, for $\sampleidx=1,\ldots,\samplesize$. 
The cluster assignment $\hat{\truelabel}^{(\sampleidx)}$ is the index of the cluster to which the $\sampleidx$th 
\gls{datapoint} $\featurevec^{(\sampleidx)}$ is assigned to. These cluster assignments $\hat{\truelabel}$ 
provide rather coarse-grained information. Two \gls{datapoint}s $\featurevec^{(\sampleidx)}, \featurevec^{(\sampleidx')}$ 
might be assigned to the same cluster $\clusteridx$ although their distances to the cluster mean $\clustermean^{(\clusteridx)}$ 
might differ significantly. Intuitively, these two \gls{datapoint}s have a different \index{degree of belonging}\gls{dob} to the cluster $\clusteridx$. 

For some clustering applications it is desirable to quantify the degree by which a \gls{datapoint} belongs to a cluster. 
Soft clustering methods use a continues range, such as the closed interval $[0,1]$, of possible values for 
the \gls{dob}. In contrast, hard clustering methods use only two possible values for the \gls{dob} to a specific cluster, 
either ``full belonging'' or no ``belonging at all''. While hard clustering methods assign a given \gls{datapoint} 
to precisely one cluster, soft clustering methods typically assign a \gls{datapoint} to several different 
clusters with non-zero \gls{dob}.  

This chapter discusses soft clustering methods that compute, for each \gls{datapoint} $\featurevec^{(\sampleidx)}$ in the dataset $\dataset$, 
a vector $\widehat{\labelvec}^{(\sampleidx)}=\big(\hat{\truelabel}_{1}^{(\sampleidx)},\ldots,\hat{\truelabel}_{\nrcluster}^{(\sampleidx)}\big)^{T}$. 
We can interpret the entry $\hat{\truelabel}_{\clusteridx}^{(\sampleidx)} \in [0,1]$ as the degree by which the \gls{datapoint} $\featurevec^{(\sampleidx)}$ 
belongs to the cluster $\cluster^{(\clusteridx)}$. For $\hat{\truelabel}_{\clusteridx}^{(\sampleidx)} \approx 1$, we are quite confident 
in the \gls{datapoint} $\featurevec^{(\sampleidx)}$ belonging to cluster $\cluster^{(\clusteridx)}$. In contrast, 
for $\hat{\truelabel}_{\clusteridx}^{(\sampleidx)} \approx 0$, we are quite confident that the \gls{datapoint} $\featurevec^{(\sampleidx)}$ 
is outside the cluster $\cluster^{(\clusteridx)}$. 

A widely used soft clustering method uses a probabilistic model for the \gls{datapoint}s 
$\dataset = \{ \featurevec^{(\sampleidx)} \}_{\sampleidx=1}^{\samplesize}$. Within this model,  
each cluster $\cluster^{(\clusteridx)}$, for $\clusteridx=1,\ldots,\nrcluster$, is represented 
by a multivariate normal distribution \cite{BertsekasProb}
\begin{equation}
	\label{equ_def_mvn_distribution}
	\mathcal{N} \big(\featurevec ; \clustermean^{(\clusteridx)}, \clustercov^{(\clusteridx)} \big) \!=\! \frac{1}{\sqrt{{\rm det } \{ 2 \pi  \clustercov \} }} \exp\big( - (1/2) \big(\featurevec\!-\!\clustermean^{(\clusteridx)} \big)^{T}  \big( \clustercov^{(\clusteridx)} \big)^{-1} \big(\featurevec\!-\!\clustermean^{(\clusteridx)}\big)  \big). 
\end{equation} 
The probability distribution \eqref{equ_def_mvn_distribution} is parametrized by a cluster-specific 
mean vector $\clustermean^{(\clusteridx)}$ and an (invertible) cluster-specific covariance matrix 
$\clustercov^{(\clusteridx)}$.\footnote{Note that the expression \eqref{equ_def_mvn_distribution} is 
	only valid for an invertible (non-singular) covariance matrix $\clustercov$.} 
Let us interpret a specific \gls{datapoint} $\featurevec^{(\sampleidx)}$ as a realization 
drawn from the probability distribution \eqref{equ_def_mvn_distribution} of a 
specific cluster $\clusteridx^{(\sampleidx)}$, 
\begin{equation} 
	\label{equ_cond_dist_GMM}
	\featurevec^{(\sampleidx)} \sim  \mathcal{N} \big(\featurevec ; \clustermean^{(\clusteridx)}, \clustercov^{(\clusteridx)} \big) \mbox{ with cluster index } \clusteridx=\clusteridx^{(\sampleidx)}.
\end{equation}
We can think of $\clusteridx^{(\sampleidx)}$ as the true index of the cluster to which the \gls{datapoint} $\featurevec^{(\sampleidx)}$ 
belongs to. The variable $\clusteridx^{(\sampleidx)}$ selects the cluster distributions \eqref{equ_def_mvn_distribution} from which 
the feature vector $\featurevec^{(\sampleidx)}$ has been generated (drawn). We will therefore refer to the variable $\clusteridx^{(\sampleidx)}$ 
as the (true) cluster assignment for the $\sampleidx$th \gls{datapoint}. Similar to the feature vectors $\featurevec^{(\sampleidx)}$ 
we also interpret the cluster assignments $\clusteridx^{(\sampleidx)}$, for $\sampleidx=1,\ldots,\samplesize$ 
as realizations of \gls{iid} \gls{rv}s. 

In contrast to the feature vectors $\featurevec^{(\sampleidx)}$, we do not observe (know) the true cluster 
indices $\clusteridx^{(\sampleidx)}$. After all, the goal of soft clustering is to estimate the cluster 
indices $\clusteridx^{(\sampleidx)}$. We obtain a \gls{softclustering} method by estimating the cluster indices 
$\clusteridx^{(\sampleidx)}$ based solely on the \gls{datapoint}s in $\dataset$. To compute 
these estimates we assume that the (true) cluster indices $\clusteridx^{(\sampleidx)}$ 
are realizations of \gls{iid} \gls{rv}s with the common probability distribution (or probability mass function)
\begin{equation} 
	\label{equ_def_cluster_prob}
	p_{\clusteridx} \defeq \prob{\clusteridx^{(\sampleidx)}=\clusteridx} \mbox{ for } \clusteridx=1,\ldots,\nrcluster.
\end{equation}  
The (prior) probabilities $p_{\clusteridx}$, for $\clusteridx=1,\ldots,\nrcluster$, are either assumed known or 
estimated from data \cite{LC,BertsekasProb}. The choice for the probabilities $p_{\clusteridx}$ could reflect 
some prior knowledge about different sizes of the clusters. For example, if cluster $\cluster^{(1)}$ is known 
to be larger than cluster $\cluster^{(2)}$, we might choose the prior probabilities such that $p_{1} > p_{2}$. 

The probabilistic model given by \eqref{equ_cond_dist_GMM}, \eqref{equ_def_cluster_prob} is referred to as a  
a \index{Gaussian mixture model}\gls{gmm}. This name is quite natural as the common 
marginal distribution for the feature vectors $\featurevec^{(\sampleidx)}$, for $\sampleidx=1,\ldots,\samplesize$, is 
a (additive) mixture of multivariate normal (Gaussian) distributions, 
\begin{equation} 
	\label{equ_def_GMM}
	\prob{\featurevec^{(\sampleidx)}} = \sum_{\clusteridx=1}^{\nrcluster} \underbrace{\prob{\clusteridx^{(\sampleidx)}=\clusteridx}}_{p_{\clusteridx}}  \underbrace{\prob{\featurevec^{(\sampleidx)} | \clusteridx^{(\sampleidx)}=\clusteridx}}_{\mathcal{N}(\featurevec^{(\sampleidx)};\clustermean^{(\clusteridx)}, \clustercov^{(\clusteridx)})}. 
\end{equation} 
As already mentioned, the cluster assignments $\clusteridx^{(\sampleidx)}$ are hidden (unobserved) 
\gls{rv}s. We thus have to infer or estimate these variables from the observed 
\gls{datapoint}s $\featurevec^{(\sampleidx)}$ which realizations or \gls{iid} \gls{rv}s
with the common distribution \eqref{equ_def_GMM}. 

The \gls{gmm} (see \eqref{equ_cond_dist_GMM} and \eqref{equ_def_cluster_prob}) lends naturally to a rigorous 
definition for the degree $\truelabel^{(\sampleidx)}_{\clusteridx}$ by which \gls{datapoint} $\featurevec^{(\sampleidx)}$ belongs 
to cluster $\clusteridx$.\footnote{Remember that the \gls{dob} $\truelabel^{(\sampleidx)}_{\clusteridx}$ 
	is considered as the (unknown) label value of a \gls{datapoint}. The choice or definition for the labels 
	of \gls{datapoint}s is a design choice. In particular, we can define the labels of \gls{datapoint}s using a 
	hypothetical probabilistic model such as the \gls{gmm}.}
Let us define the label value  $\truelabel^{(\sampleidx)}_{\clusteridx}$ as the ``a-posteriori'' probability of the cluster 
assignment $\clusteridx^{(\sampleidx)}$ being equal to $\clusteridx \in \{1,\ldots,\nrcluster\}$: 
\begin{equation}
	\label{equ_def_deg_belonging_prob}
	\truelabel^{(\sampleidx)}_{\clusteridx} \defeq \prob{ \clusteridx^{(\sampleidx)} = \clusteridx | \dataset}.
\end{equation} 
By their very definition \eqref{equ_def_deg_belonging_prob}, the degrees of belonging $y^{(\sampleidx)}_{\clusteridx}$ 
always sum to one, 
\begin{equation} 
	\label{equ_dob_sum_to_one}
	\sum_{\clusteridx=1}^{\nrcluster} \truelabel^{(\sampleidx)}_{\clusteridx}=1 \mbox{ for each } \sampleidx=1,\ldots,\samplesize.
\end{equation}  
We emphasize that we use the conditional cluster probability \eqref{equ_def_deg_belonging_prob}, conditioned 
on the dataset $\dataset$, for defining the \gls{dob} $\truelabel^{(\sampleidx)}_{\clusteridx}$. 
This is reasonable since the \gls{dob} $\truelabel^{(\sampleidx)}_{\clusteridx}$ 
depends on the overall (cluster) geometry of the \gls{dataset} $\dataset$.

The definition \eqref{equ_def_deg_belonging_prob} for the label values (\gls{dob}s) 
$\truelabel^{(\sampleidx)}_{\clusteridx}$ involves the \gls{gmm} parameters $\{\clustermean^{(\clusteridx)},\clustercov^{(\clusteridx)},p_{\clusteridx}\}_{\clusteridx=1}^{\nrcluster}$ (see \eqref{equ_def_GMM}). 
Since we do not know these parameters beforehand we cannot evaluate the conditional probability in \eqref{equ_def_deg_belonging_prob}. 
A principled approach to solve this problem is to evaluate \eqref{equ_def_deg_belonging_prob} with 
the true \gls{gmm} parameters replaced by some estimates $\{\widehat{\clustermean}^{(\clusteridx)},\widehat{\clustercov}^{(\clusteridx)},\hat{p}_{\clusteridx}\}_{\clusteridx=1}^{\nrcluster}$. 
Plugging in the \gls{gmm} parameter estimates into \eqref{equ_def_deg_belonging_prob} provides 
us with predictions $\hat{\truelabel}^{(\sampleidx)}_{\clusteridx}$ for the degrees of belonging. 
However, to compute the \gls{gmm} parameter estimates we would have already needed 
the degrees of belonging $\truelabel^{(\sampleidx)}_{\clusteridx}$. This situation is similar to 
hard clustering where ultime goals is to jointly optimize cluster means and assignments (see Section \ref{sec_hard_clustering}). 

Similar to the spirit of Algorithm \ref{alg:kmeans} for hard clustering, we solve the 
above dilemma of \gls{softclustering} by an alternating optimization scheme. This scheme, which is illustrated in Figure \ref{fig_flow_softclustering}, 
alternates between updating (optimizing) the predicted degrees of belonging (or soft cluster assignments) $\hat{\truelabel}^{(\sampleidx)}_{\clusteridx}$, 
for $\sampleidx=1,\ldots,\samplesize$ and $\clusteridx=1,\ldots,\nrcluster$, given the current \gls{gmm} parameter estimates $\{\widehat{\clustermean}^{(\clusteridx)},\widehat{\clustercov}^{(\clusteridx)},\hat{p}_{\clusteridx}\}_{\clusteridx=1}^{\nrcluster}$ and 
then updating (optimizing) these \gls{gmm} parameter estimates based on the updated predictions $\hat{\truelabel}^{(\sampleidx)}_{\clusteridx}$. 
We summarize the resulting soft clustering method in Algorithm \ref{alg:softclustering}. Each iteration of 
Algorithm \ref{alg:softclustering} consists of an update \eqref{equ_update_soft_cluster_assignm} for the degrees 
of belonging followed by an update (step \ref{equ_GMM_update_step}) for the \gls{gmm} parameters. 

\begin{figure} 
	\begin{center}

\tikzset{global scale/.style={
		scale=#1,
		every node/.append style={scale=#1}
	}
}
\begin{tikzpicture}[global scale = 1,           
	squarednode/.style={rectangle, draw=blue!80, line width=2pt, minimum size=15mm, font=\fontsize{15}{15}\selectfont, align=center},      
	myarrow/.style={line width = 1mm, draw = blue!80, -triangle 45, postaction={draw, line width = 2mm, shorten >=3mm, -}}]            
	
	
	\node[squarednode] (initial) {initial choice for \\ cluster means, cov. \\ and effective size};      
	\node[squarednode, rounded corners] (ucassignment) [below of=initial, xshift=3cm, yshift=-1.5cm]  {update soft cluster \\ assignment};    
	\node[squarednode, rounded corners] (ucmeans) [right=of ucassignment, xshift=0.5cm] {update cluster \\ means, cov. and eff. size};        
	\node[font=\fontsize{18}{0}\selectfont, below of=ucassignment, yshift=-0.5cm, xshift=-0.5cm] (yc) {$\hat{\truelabel}^{(\sampleidx)}_{\clusteridx}$};          
	\node[font=\fontsize{18}{0}\selectfont, above of=ucmeans, yshift=0.5cm, xshift=2cm] (uc) {$\{\widehat{\clustermean}^{(\clusteridx)},\widehat{\clustercov}^{(\clusteridx)},\hat{p}_{\clusteridx}\}_{\clusteridx=1}^{\nrcluster}$};         
	
	\draw[myarrow, rounded corners] (initial.east) -| ([yshift=0.2cm]ucassignment.north) [xshift=2cm, yshift=2cm];          
	\draw[myarrow] ([xshift=0.5cm, yshift=-0.5cm]ucassignment.south) to [bend right=45] ([xshift=-0.5cm, yshift=-0.5cm]ucmeans.south);      
	\draw[myarrow] ([xshift=-0.5cm, yshift=0.5cm]ucmeans.north) to [bend right=45] ([xshift=0.5cm, yshift=0.5cm]ucassignment.north);         
	
\end{tikzpicture}
	\end{center}
	\caption{\label{fig_flow_softclustering} The workflow of the soft clustering Algorithm \ref{alg:softclustering}. 
		Starting from an initial guess or estimate for the cluster parameters, the soft cluster assignments 
		and cluster parameters are updated (improved) in an alternating fashion.}
\end{figure}



\begin{figure}[hbtp]
	\begin{center}
		\begin{tikzpicture}[scale=0.4]
			\draw [thick] \boundellipse{0,0}{10}{5} node[right]  {$\clustermean^{(1)}$};
			\fill (0,0) circle (2pt) ; 
			\node [right] at (0,5.5) {$ \clustercov^{(1)}$} ; 
			\draw [thick] \boundellipse{11,1}{-2}{4} node[right]  {$\clustermean^{(2)}$};
			\fill (11,1) circle (2pt) ; 
			\node [right] at (11,5.5) {$ \clustercov^{(2)}$} ; 
			\draw [thick] \boundellipse{-9,4}{2}{3} node[left,xshift=3mm,yshift=3mm]  {$\,\,\clustermean^{(3)}$}; 
			\fill (-9,4) circle (2pt) ; 
			\node [right] at (-9,8) {$ \clustercov^{(3)}$} ; 
		\end{tikzpicture}
	\end{center}
	\label{fig_GMM_elippses}
	\caption{The \gls{gmm} \eqref{equ_cond_dist_GMM}, \eqref{equ_def_cluster_prob} results in a 
		probability distribution \eqref{equ_def_GMM} for (feature vectors of) \gls{datapoint}s which is a 
		weighted sum of multivariate normal distributions $\mathcal{N}(\clustermean^{(\clusteridx)}, \clustercov^{(\clusteridx)})$. 
		The weight of the $\clusteridx$-th component is the cluster probability 
		$ \prob{\clusteridx^{(\sampleidx)}=\clusteridx}$.}
\end{figure}

To analyze Algorithm \ref{alg:softclustering} it is helpful to interpret (the features of) \gls{datapoint}s $\featurevec^{(\sampleidx)}$ 
as realizations of \gls{iid} \gls{rv}s distributed according to a \gls{gmm} \eqref{equ_cond_dist_GMM}-\eqref{equ_def_cluster_prob}. 
We can then understand Algorithm \ref{alg:softclustering} as a method for estimating the \gls{gmm} parameters based on observing realizations 
drawn from the \gls{gmm} \eqref{equ_cond_dist_GMM}-\eqref{equ_def_cluster_prob}. We can estimate the 
parameters of a probability distribution using the \gls{ml} method (see Section \ref{sec_max_iikelihood} and \cite{kay,LC}). 
As its name suggests, \gls{ml} methods estimate the \gls{gmm} parameters by maximizing the probability (density) 
\begin{equation} 
\label{equ_def_prob_den_GMM}
p \big(\dataset; \{ \clustermean^{(\clusteridx)}, \clustercov^{(\clusteridx)}, p_{\clusteridx} \}_{\clusteridx=1}^{\nrcluster}\big)
\end{equation} 
of actually observing the \gls{datapoint}s in the dataset $\dataset$. 

It can be shown that Algorithm \ref{alg:softclustering} is an instance of a generic approximate maximum likelihood technique 
referred to as \index{expectation maximization}expectation maximization \gls{em} (see \cite[Chap. 8.5]{hastie01statisticallearning} for more details). 
In particular, each iteration of Algorithm \ref{alg:softclustering} updates the \gls{gmm} parameter estimates such that 
the corresponding probability density \eqref{equ_def_prob_den_GMM} does not decrease \cite{XuJordan1996}. If 
we denote the \gls{gmm} parameter estimate obtained after $\itercntr$ iterations of Algorithm \ref{alg:softclustering} by 
${\bm \theta}^{(\itercntr)}$ \cite[Sec. 8.5.2]{hastie01statisticallearning}, 
\begin{equation}
\label{equ_proability_increasing_GMM}
p \big(\dataset;{\bm \theta}^{(\itercntr\!+\!1)} \big) \geq p \big(\dataset;{\bm \theta}^{(\itercntr)} \big)
\end{equation} 

\begin{algorithm}[htbp]
\caption{``A Soft-Clustering Algorithm'' \cite{BishopBook}}\label{alg:softclustering}
\begin{algorithmic}[1]
	\renewcommand{\algorithmicrequire}{\textbf{Input:}}
	\renewcommand{\algorithmicensure}{\textbf{Output:}}
	\Require   dataset $\dataset=\{ \featurevec^{(\sampleidx)}\}_{\sampleidx=1}^{\samplesize}$; number $\nrcluster$ of clusters, initial \gls{gmm} parameter estimates $\{\widehat{\clustermean}^{(\clusteridx)},\widehat{\clustercov}^{(\clusteridx)},\hat{p}_{\clusteridx}\}_{\clusteridx=1}^{\nrcluster}$  
	\Repeat
	\vspace*{2mm}
	\State \label{equ_update_degree_of_belonging} for each $\sampleidx=1,\ldots,\samplesize$ and $\clusteridx=1,\ldots,\nrcluster$, update degrees of belonging
	\vspace*{-1mm}
	\begin{align}
		\label{equ_update_soft_cluster_assignm}
		\hat{\truelabel}_{\clusteridx}^{(\sampleidx)} & \defeq  \frac{\hat{p}_{\clusteridx} \mathcal{N}(\featurevec^{(\sampleidx)};\widehat{\clustermean}^{(\clusteridx)},\widehat{\clustercov}^{(\clusteridx)})}{\sum_{\clusteridx'=1}^{\nrcluster} \hat{p}_{\clusteridx'}\mathcal{N}(\featurevec^{(\sampleidx)};\widehat{\clustermean}^{(\clusteridx')},\widehat{\clustercov}^{(\clusteridx')})} 
	\end{align}
	\State \label{equ_GMM_update_step} for each $\clusteridx \in \{1,\ldots,\nrcluster\}$, update \gls{gmm} parameter estimates:
	\begin{itemize} 
		\item $\hat{p}_{\clusteridx}\!\defeq\!\samplesize_{\clusteridx} /\samplesize$ with 
		effective cluster size $\samplesize_{\clusteridx} \defeq \sum\limits_{\sampleidx=1}^{\samplesize} \hat{\truelabel}_{\clusteridx}^{(\sampleidx)}$  (cluster probability) 
		\item $\widehat{\clustermean}^{(\clusteridx)} \defeq (1/\samplesize_{\clusteridx}) \sum\limits_{\sampleidx=1}^{\samplesize} \hat{\truelabel}_{\clusteridx}^{(\sampleidx)} \featurevec^{(\sampleidx)}$  (cluster mean)
		\item $\widehat{\clustercov}^{(\clusteridx)}  \defeq (1/\samplesize_{\clusteridx}) {\sum\limits_{\sampleidx=1}^{\samplesize} \hat{\truelabel}_{\clusteridx}^{(\sampleidx)} \big(\featurevec^{(\sampleidx)}\!-\!\widehat{\clustermean}^{(\clusteridx)}\big)   \big(\featurevec^{(\sampleidx)}\!-\!\widehat{\clustermean}^{(\clusteridx)}\big)^{T} }$ (cluster covariance matrix)
	\end{itemize}
	\vspace*{1mm}
	\Until stopping criterion met \label{equ_conv_soft_clustering_algo}
	\Ensure predicted degrees of belonging $\widehat{\labelvec}^{(\sampleidx)}=(\hat{\truelabel}_{1}^{(\sampleidx)},\ldots,\hat{y}_{\nrcluster}^{(\sampleidx)})^{T}$ for $\sampleidx=1,\ldots,\samplesize$. 
\end{algorithmic}
\end{algorithm}

As for Algorithm \ref{alg:kmeans}, we can also interpret Algorithm \ref{alg:softclustering} 
as an instance of the \gls{erm} principle discussed in Chapter \ref{ch_Optimization}. Indeed, maximizing the 
probability density \eqref{equ_def_prob_den_GMM} is equivalent to 
minimizing the \gls{emprisk} 
\begin{equation} 
\label{equ_def_emp_risk_soft_clustering}
\emperror \big({\bm \theta}  \mid \dataset \big) \defeq
- \log  p \big( \dataset; {\bm \theta}  \big) \mbox{ with \gls{gmm} parameters }  {\bm \theta} \defeq \{ \clustermean^{(\clusteridx)}, \clustercov^{(\clusteridx)}, p_{\clusteridx} \}_{\clusteridx=1}^{\nrcluster} 
\end{equation} 
The \gls{emprisk} \eqref{equ_def_emp_risk_soft_clustering} is the negative logarithm 
of the probability (density) \eqref{equ_def_prob_den_GMM} of observing the dataset $\dataset$ as \gls{iid} realizations 
of the \gls{gmm} \eqref{equ_def_GMM}. The monotone increase in the probability density \eqref{equ_proability_increasing_GMM} 
achieved by the iterations of Algorithm \ref{alg:softclustering} translate into a monotone decrease of the \gls{emprisk}, 
\begin{equation} 
\label{equ_emprisk_decreasing_GMM}
\emperror \big({\bm \theta}^{(\itercntr)}  \mid \dataset \big) \leq \emperror \big({\bm \theta}^{(\itercntr-1)}  \mid \dataset \big) \mbox{ with iteration counter } \itercntr. 
\end{equation} 

The monotone decrease \eqref{equ_emprisk_decreasing_GMM} in the \gls{emprisk} \eqref{equ_def_emp_risk_soft_clustering} 
achieved by the iterations of Algorithm \ref{alg:softclustering} naturally lends to a stopping criterion. Let $\error_{\itercntr}$ denote 
the \gls{emprisk} \eqref{equ_def_emp_risk_soft_clustering} achieved by the \gls{gmm} parameter 
estimates ${\bm \theta}^{(\itercntr)}$ obtained after $\itercntr$ iterations in Algorithm \ref{alg:softclustering}. 
Algorithm \ref{alg:softclustering} stops iterating as soon as the decrease $\error_{\itercntr\!-\!1} - \error_{\itercntr}$ 
achieved by the $\itercntr$-th iteration of Algorithm \ref{alg:softclustering} falls below a given (positive) threshold $\varepsilon>0$. 

Similar to Algorithm \ref{alg:kmeans}, also Algorithm \ref{alg:softclustering} might get trapped in local minima 
of the underlying \gls{emprisk}. The \gls{gmm} parameters delivered by Algorithm \ref{alg:softclustering} might only be a local 
minimum of \eqref{equ_def_emp_risk_soft_clustering} but not the global minimum (see Figure \ref{fig_emp_risk_k_means} 
for the analogous situation in \gls{hardclustering}). As for \gls{hardclustering} Algorithm \ref{alg:kmeans}, we typically repeat Algorithm \ref{alg:softclustering} 
several times. During each repetition of Algorithm \ref{alg:softclustering}, we use a different (randomly chosen) initialization for 
the \gls{gmm} parameter estimates ${\bm \theta} = \{ \widehat{\clustermean}^{(\clusteridx)}, \widehat{\clustercov}^{(\clusteridx)}, \hat{p}_{\clusteridx} \}_{\clusteridx=1}^{\nrcluster} $. Each repetition of Algorithm \ref{alg:softclustering} results in a potentially different 
set of \gls{gmm} parameter estimates and degrees of belongings $\hat{y}^{(\sampleidx)}_{\clusteridx}$. We then use the 
results for that repetition that achieves the smallest \gls{emprisk} \eqref{equ_def_emp_risk_soft_clustering}.

%

Let us point out an interesting link between soft clustering methods based on \gls{gmm} (see Algorithm \ref{alg:softclustering}) 
and \gls{hardclustering} with \gls{kmeans} (see Algorithm \ref{alg:kmeans}). Consider the \index{Gaussian mixture model (GMM)} \gls{gmm} \eqref{equ_cond_dist_GMM} with prescribed cluster covariance matrices 
\begin{equation}
\label{equ_def_special_case}
\clustercov^{(\clusteridx)}= \sigma^{2} \mathbf{I} \mbox{ for all } \clusteridx \in \{1,\ldots,\nrcluster\}, 
\end{equation} 
with some given variance $\sigma^{2}>0$. 
We assume the cluster covariance matrices in the \gls{gmm} to be given by \eqref{equ_def_special_case} and 
therefore can replace the covariance matrix updates in Algorithm \ref{alg:softclustering} 
with the assignment $\widehat{\clustercov}^{(\clusteridx)} \defeq  \sigma^{2} \mathbf{I}$. 
It can be verified easily that for sufficiently small variance $\sigma^{2}$ in \eqref{equ_def_special_case}, the update \eqref{equ_update_soft_cluster_assignm} tends to enforce $\hat{y}_{\clusteridx}^{(\sampleidx)} \in \{0,1\}$. In other words, 
each \gls{datapoint} $\featurevec^{(\sampleidx)}$ becomes then effectively associated with exactly one single cluster $\clusteridx$ 
whose cluster mean $\widehat{\clustermean}^{(\clusteridx)}$ is nearest to $\featurevec^{(\sampleidx)}$. For $\sigma^{2} \rightarrow 0$, 
the \gls{softclustering} update \eqref{equ_update_soft_cluster_assignm} in Algorithm \ref{alg:softclustering} reduces to the (hard) 
cluster assignment update \eqref{equ_cluster_assign_update} in \gls{kmeans} Algorithm \ref{alg:kmeans}. We can interpret 
Algorithm \ref{alg:kmeans} as an extreme case of Algorithm \ref{alg:softclustering} that is obtained by fixing the covariance 
matrices in the \gls{gmm}  to $\sigma^{2} \mathbf{I}$ with a sufficiently small $\sigma^{2}$.

{\bf Combining \gls{gmm} with \gls{linreg}.} Let us sketch how Algorithm \ref{alg:softclustering} could be combined 
with \gls{linreg} methods (see Section \ref{sec_lin_reg}). The idea is to first compute the \gls{dob}s to the clusters 
for each \gls{datapoint}. We then learn separate linear predictors for each cluster using the \gls{dob}s as \gls{weights} 
for the individual loss terms in the \gls{trainerr}. To predict the label of a new \gls{datapoint}, we first 
compute the predictions obtained for each cluster-specific linear hypothesis. These cluster-specific predictions are then averaged 
using the \gls{dob}s for the new \gls{datapoint} as weights. 

\section{Connectivity-based Clustering}
\label{sec_connect_clustering}

The clustering methods discussed in Sections \ref{sec_hard_clustering} and \ref{sec_soft_clustering} can 
only be applied to \gls{datapoint}s which are characterized by numeric feature vectors. These methods define 
the similarity between \gls{datapoint}s using the Euclidean distance between the feature vectors of these 
\gls{datapoint}s. As illustrated in Figure \ref{fig_GMM_kmeans_geometry}, these methods can only produce 
``Euclidean shaped'' clusters that are contained either within hyper-spheres (Algorithm \ref{alg:kmeans}) 
or hyper-ellipsoids (Algorithm \ref{alg:softclustering}). 

\begin{figure}[htbp]
\begin{center}
	\begin{minipage}{0.3\linewidth}
		\begin{tikzpicture}[scale=0.2]
			\draw [thick] \boundellipse{0,0}{6}{6} node[right]  {$\clustermean^{(1)}$};
			\fill (0,0) circle (2pt) ; 
			\node [] at (4,-8) {(a)} ; 
			\node [right] at (0,5.5) {$ \clustercov^{(1)}$} ; 
			\draw [thick] \boundellipse{11,1}{3}{3} node[right]  {\hspace*{-2mm}$\clustermean^{(2)}$};
			\fill (11,1) circle (2pt) ; 
			\node [right] at (11,5.5) {$ \clustercov^{(2)}$} ; 
			\draw [thick] \boundellipse{-9,4}{4}{4} node[left,xshift=3mm,yshift=3mm]  {$\clustermean^{(3)}$}; 
			\fill (-9,4) circle (2pt) ; 
			\node [right] at (-9,8) {$ \clustercov^{(3)}$} ; 
		\end{tikzpicture}
	\end{minipage}
	\hspace*{25mm}
	\begin{minipage}{0.3\linewidth}
		\begin{tikzpicture}[scale=0.2]
			\draw [thick] \boundellipse{0,0}{10}{5} node[right]  {$\clustermean^{(1)}$};
			\fill (0,0) circle (6pt) ; 
			\node [right] at (0,5.5) {$ \clustercov^{(1)}$} ; 
			\draw [thick] \boundellipse{11,1}{-2}{4} node[right]  {$\clustermean^{(2)}$};
			\fill (11,1) circle (6pt) ; 
			\node [right] at (11,5.5) {$\clustercov^{(2)}$} ; 
			\draw [thick] \boundellipse{-9,4}{2}{3} node[left,xshift=3mm,yshift=3mm]  {$\,\,\clustermean^{(3)}$}; 
			\fill (-9,4) circle (6pt) ; 
			\node [] at (4,-8) {(b)} ; 
			\node [right] at (-9,8) {$ \clustercov^{(3)}$} ; 
		\end{tikzpicture}
	\end{minipage}
\end{center}
\caption{(a): Cartoon of typical cluster shapes delivered by \gls{kmeans} Algorithm \ref{alg:kmeansimpl}. 
	(b): Cartoon of typical cluster shapes delivered by soft clustering Algorithm \ref{alg:softclustering}.	\label{fig_GMM_kmeans_geometry}}
\end{figure}

Some applications generate \gls{datapoint}s for which the construction of useful numeric features is difficult. 
Even if we can easily obtain numeric features for \gls{datapoint}s, the Euclidean distances between the 
resulting feature vectors might not reflect the actual similarities between \gls{datapoint}s. As a case in point, 
consider \gls{datapoint}s representing text documents. We could use the histogram of a respecified list of 
words as numeric features for a text document. In general, a small Euclidean distance between histograms 
of text documents does not imply that the text documents have similar meanings. Moreover, clusters 
of similar text documents might have highly complicated shapes in the space of feature vectors 
that cannot be grouped within hyper-ellipsoids. For datasets with such ``non-Euclidean'' cluster 
shapes, \gls{kmeans} or \gls{gmm} are not suitable as clustering methods. We should then replace 
the Euclidean distance between feature vectors with another concept to determine or measure the 
similarity between \gls{datapoint}s. 

Connectivity-based clustering methods do not require any numeric features of \gls{datapoint}s. These methods cluster 
\gls{datapoint}s based on explicitly specifying for any two different \gls{datapoint}s if they are similar and to what extend. 
A convenient mathematical tool to represent similarities between the \gls{datapoint}s of a dataset $\dataset$ 
is a weighted undirected graph $\graph=\big(\nodes,\edges\big)$. We refer to this graph as the similarity graph 
of the dataset $\dataset$ (see Figure \ref{fig_connect_clustering}). The nodes $\nodes$ in this similarity graph $\graph$ 
represent \gls{datapoint}s in $\dataset$ and the undirected edges connect nodes that represent similar \gls{datapoint}s. 
The extend of the similarity is represented by the \gls{weights} $W_{\nodeidx,\nodeidx'}$ for each edge $\{\nodeidx,\nodeidx'\} \in \edges$. 

Given a similarity graph $\graph$ of a dataset, connectivity-based clustering methods determine clusters as 
subsets of nodes that are well connected within the cluster but weakly connected between different clusters. 
Different concepts for quantifying the connectivity between nodes in a graph yield different clustering methods. 
Spectral clustering methods use eigenvectors of a graph Laplacian matrix to measure the connectivity between nodes \cite{Luxburg2007,Ng2001}.  
Flow-based clustering methods measure the connectivity between two nodes via the amount of flow that can be 
routed between them \cite{JungLocalGraphClustering}. Note that we might use these connectivity measures to 
construct meaningful numerical feature vectors for the nodes in the empirical graph. These feature vectors can 
then be fed into the hard-clustering Algorithm \ref{alg:kmeansimpl} or the soft clustering Algorithm \ref{alg:softclustering} 
(see Figure \ref{fig_connect_clustering}). 

The algorithm \gls{dbscan} considers two \gls{datapoint}s $\sampleidx,\sampleidx'$ as connected if one of 
them (say $\sampleidx$) is a \index{core node} core node and the other node ($\sampleidx'$) 
can be reached via a sequence (path) of connected core nodes $$\sampleidx^{(1)},\ldots,\sampleidx^{(\itercntr)} \mbox{ , with } \{\sampleidx,\sampleidx^{(1)}\}, \{\sampleidx^{(1)},\sampleidx^{(2)}\},\ldots,\{\sampleidx^{(\itercntr)},\sampleidx'\} \in \edges.$$
 \Gls{dbscan} considers a node to be a core node if it has a sufficiently large number of neighbours \cite{DBSCAN}. 
The minimum number of neighbours required for a node to be considered a core node is a hyper-parameter of 
 \gls{dbscan}. When \gls{dbscan} is applied to \gls{datapoint}s with numeric feature vectors, it defines 
 two \gls{datapoint}s as connected if the Euclidean distance between their feature vectors does not exceed 
 a given threshold $\varepsilon$ (see Figure \ref{fig_DBSCAN}). 

In contrast to \gls{kmeans} and \gls{gmm}, \gls{dbscan} does not require the number of 
clusters to be specified. The number of clusters is determined automatically by  \gls{dbscan} and 
depends on its hyper-parameters. \Gls{dbscan} also performs an implicit outlier detection. The outliers 
delivered by \gls{dbscan} are those \gls{datapoint}s which do not belong to the same cluster as any other 
\gls{datapoint}. 

\begin{figure}[htbp]
	\subfloat[][]{
			\begin{tikzpicture}[auto,scale=1.2]
				\tikzstyle{vertex}=[circle,fill=black!25,minimum size=17pt,inner sep=0pt]
				\foreach \name/\x/\y in {1/0/0, 2/0/2, 3/0/-2, 4/2/0, 5/3/0, 6/5/0, 7/5/2, 8/5/-2}
				\node[vertex] (G-\name) at (\x,\y) {$\name$};
				\foreach \from/\to in {1/2, 1/4, 4/2, 4/5, 1/3/, 3/4}
				\draw[line width=1mm] (G-\from) -- (G-\to); 
				\foreach \from/\to in {5/6, 6/7, 6/8, 5/7, 5/8}
				\draw[line width=1mm] (G-\from) -- (G-\to); 
				\node[above right= 0.0cm and -0.1cm of G-1] () {$\featurevec^{(1)}$};
			\end{tikzpicture}
			\label{fig-a_tmp_graph_similarity}
		}
		\hspace*{5mm}
		\subfloat[][]{
				\begin{tikzpicture}[auto,scale=0.6]
					\draw [thick] (5.8,0.5) circle (0.1cm)node[anchor=west] {\hspace*{0mm}$\featurevec^{(1)}$};
					\draw [thick] (4.4,-0.3) circle (0.1cm)node[anchor=east] {\hspace*{0mm}$\featurevec^{(2)}$};
					\draw [thick] (5.2,-0.2) circle (0.1cm)node[anchor=west,below] {\hspace*{0mm}$\featurevec^{(3)}$};
					\draw [thick] (5.4,0.5) circle (0.1cm)node[anchor=west,above] {\hspace*{0mm}$\featurevec^{(4)}$};
					\draw [thick] (-0.4,4.8)circle (0.1cm) node[anchor=west,above] {\hspace*{0mm}$\featurevec^{(5)}$};
					\draw [thick] (0.7,5)circle (0.1cm)node[anchor=west,above] {\hspace*{0mm}$\featurevec^{(6)}$};
					\draw [thick] (0.3,5.6) circle (0.1cm)node[anchor=west,above] {\hspace*{0mm}$\featurevec^{(7)}$};
					\draw [thick] (0.5,4) circle (0.1cm)node[anchor=west,above] {\hspace*{0mm}$\featurevec^{(8)}$};
					\draw[->] (-0.5,0) -- (7.5,0) node[right] {$\feature_{1}$};
					\draw[->] (0,-0.5) -- (0,7.5) node[above] {$\feature_{2}$};
				\end{tikzpicture}
				\label{fig-a_tmp_graph_similarity_features}}
		\caption{\label{fig_connect_clustering} Connectivity-based clustering can be obtained by constructing features $\featurevec^{(\sampleidx)}$ that are (approximately) identical for well-connected \gls{datapoint}s. \protect\subref{fig-a_tmp_graph_similarity}: 
			A similarity graph for a dataset $\dataset$ consists of nodes representing individual \gls{datapoint}s and 
			edges that connect similar \gls{datapoint}s. \protect\subref{fig-a_tmp_graph_similarity_features} Feature 
			vectors of well-connected \gls{datapoint}s have small Euclidean distance. }
	\end{figure} 
	
	\begin{figure} 
		\begin{center}
			\begin{tikzpicture}[scale=11/5]
				\tikzstyle{every node}=[font=\small]
				\node[] (C1) at (0,0) {};
				\node[right=2 cm of C1] (C2)  {};
				\node[below=1cm of C2] (C2_1)  {};
				\node[above =1cm of C2] (C2_2)  {};
				\node[right=2 cm of C2] (C3)  {};
				\node[right=2 cm of C3] (C4)  {};
				\draw (C3) circle (2pt);
				\draw (C2_1) circle (2pt);
				\draw (C2_2) circle (2pt);
				\draw (C1) circle (2pt);
				\draw (C2) circle (2pt);
				\draw [dashed] (C2) circle (0.8cm);
				\draw (C4) circle (2pt);
				\node[above right = -0.1cm and 0.0cm of C1,font=\fontsize{8}{0}\selectfont,anchor=south east]  {$\featurevec^{(1)}$}; 
				\node[above right = -0.2cm and 0.0cm of C2,font=\fontsize{8}{0}\selectfont,anchor=south east]  {$\featurevec^{(2)}$}; 
				\draw [line width=0.3mm,-] (C1)--(C2);
				\draw [line width=0.3mm,-] (C2)--(C3);
				\draw [line width=0.3mm,-] (C3)--(C4);
				\path [line width=0.3mm,-]  (C2) edge node [ right] {$<\varepsilon$}(C2_1);
				\draw [line width=0.3mm,-] (C2_2)--(C2);
			\end{tikzpicture}
			\caption{\label{fig_DBSCAN} \index{DBSCAN}\Gls{dbscan} assigns two \gls{datapoint}s to the same 
				cluster if they are reachable. Two \gls{datapoint}s $\featurevec^{(\sampleidx)},\featurevec^{(\sampleidx')}$ 
				are reachable if there is a path of \gls{datapoint}s from $\featurevec^{(\sampleidx')}$ to $\featurevec^{(\sampleidx)}$. 
				This path consists of a sequence of \gls{datapoint}s that are within a distance of $\varepsilon$. Moreover, 
				each \gls{datapoint} on this path must be a core point which has at least a given number of neighbouring 
				\gls{datapoint}s within the distance $\varepsilon$.}
		\end{center}
	\end{figure} 
	
	\section{Clustering as Preprocessing}
	\label{sec_clust_preproc}
	
	In applications it might be benefiial to combine clustering methods with supervised methods such as \gls{linreg}.  
	As a point in case consider a dataset that consists of \gls{datapoint}s obtained from two different data generation 
	processes. Let us denote the \gls{datapoint}s generated by one process by $\dataset^{(1)}$ and the other one by $\dataset^{(2)}$. 
	Each datapoint is characterzed by features and a label. While there would be an accurate lienar hypothesis 
	for predicting the label of datapoints in $\dataset^{(1)}$ and another linear hypothesis for $\dataset^{(2)}$ these 
	two are very different. 
	
	We could try to use clustering methods to assign any given \gls{datapoint} to the corresponding 
	data generation process. If we are lucky, the resulting clusters resemble (approximately) the two 
	data generation processes $\dataset^{(1)}$ and $\dataset^{(2)}$. Once we have successfully 
	clustered the \gls{datapoint}s, we can learn a separate (tailored) hypothesis for ach cluster. 
	More generally, we can use the predicted cluster assignments obtained from the methods of 
	Section \ref{sec_hard_clustering} - \ref{sec_connect_clustering} as additional features for each \gls{datapoint}. 
	
	Let us illustrate the above ideas by combining Algorithm \ref{alg:kmeans} with \gls{linreg}. 
	We first group \gls{datapoint}s into a given number $\nrcluster$ of clusters and then learn 
	separate linear predictors $h^{(\clusteridx)}(\featurevec)=\big( \weights^{(\clusteridx)} \big)^{T} \featurevec$ 
	for each cluster $\clusteridx=1,\ldots,\nrcluster$. To predict the label of a new \gls{datapoint} 
	with features $\featurevec$, we first assign to the cluster $\clusteridx'$ with the nearest
	cluster mean. We then use the linear predictor $h^{(\clusteridx')}$ assigned to cluster 
	$\clusteridx'$ to compute the predicted label $\hat{\truelabel} = h^{(\clusteridx')}(\featurevec)$.  
	
\section{Exercises} 
	
\begin{exercise}[Monotonicity of \gls{kmeans} Updates.] 
Show that the cluster means and assignments updates \eqref{equ_cluster_mean_update} 
and \eqref{equ_cluster_assign_update} never increase the clustering error \eqref{equ_def_emp_risk_kmeans}. 
\end{exercise}
	
\begin{exercise}[How to choose $\nrcluster$ in \gls{kmeans}?] 
Discuss and experiment with different strategies for choosing the number $\nrcluster$ 
of clusters in \gls{kmeans} Algorithm \ref{alg:kmeansimpl}. 
\end{exercise}
	
\begin{exercise}[Local Minima.] 
Apply the \gls{hardclustering} Algorithm \ref{alg:kmeansimpl} 
to the dataset $(-10,1),(10,1),(-10,-1),(10,-1)$ with initial cluster means $(0,1),(0,-1)$ 
and tolerance $\varepsilon=0$. For this initialization, will Algorithm \ref{alg:kmeansimpl} 
get trapped in a local minimum of the clustering error \eqref{equ_def_clustering_error_means}?
\end{exercise}
	
\begin{exercise}[Image Compression with \gls{kmeans}.]
Apply \gls{kmeans} to image compression. Consider image pixels as \gls{datapoint}s whose 
features are RGB intensities. We obtain a simple image compression format by, instead of 
storing RGB pixel values, storing the cluster means (which are RGB triplets) and the cluster 
index for each pixel. Try out different values for the numberf $\nrcluster$ of clusters 
and discuss the resulting trade off between achievable reconstruction quality and storage 
size.   
\end{exercise}
	
\begin{exercise}[Compression with \gls{kmeans}.]
Consider $\samplesize=10000$ datapoints $\featurevec^{(1)},\ldots,\featurevec^{(\samplesize)}$ 
which are represented by numeric feature vectors of length two. We apply \gls{kmeans} 
to cluster the data set into $\nrcluster = 5$ clusters. How many bits do we need to store the 
resulting cluster assignments?
\end{exercise}

\chapter{Feature Learning} 
\label{ch_FeatureLearning}

\begin{quote}
	``Solving Problems By Changing the Viewpoint.''
\end{quote}

\begin{figure}[htbp]
	\begin{center}
		\includegraphics[width=0.7\textwidth]{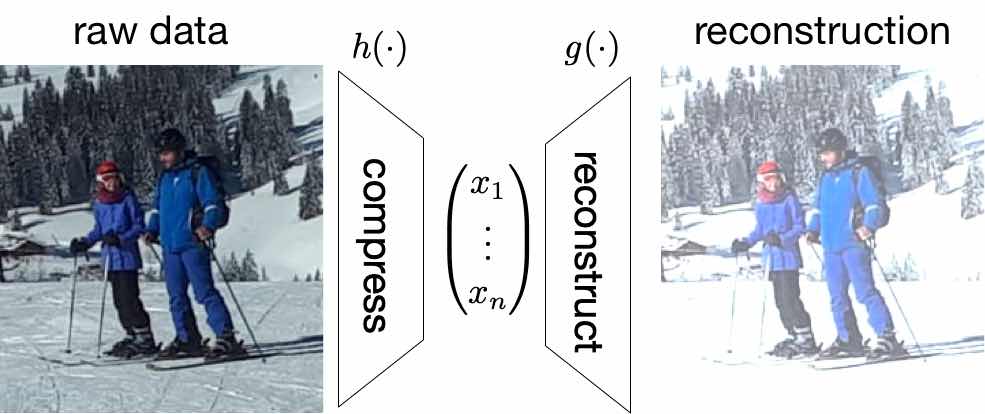}  
	\end{center}
	\caption{Dimensionality reduction methods aim at finding a map $h$ which maximally 
		compresses the raw data while still allowing to accurately reconstruct the original 
		datapoint from a small number of features $x_{1},\ldots,x_{\featuredim}$.}
	\label{fig_dimred}
\end{figure}



Chapter \ref{ch_Elements_ML} defined features as those properties of a \gls{datapoint} that can 
be measured or computed easily. Sometimes the choice of features follows naturally from the 
available hard- and software. For example, we might use the numeric measurement $\rawfeature \in \mathbb{R}$ 
delivered by a sensing device as a feature. However, we could augment this single feature with 
new features such as the powers $\rawfeature^{2}$ and $\rawfeature^{3}$ or adding a constant $\rawfeature+5$. 
Each of these computations produces a new feature. Which of these additional features are most useful?

Feature learning methods automate the choice of finding good features. These methods learn a hypothesis 
map that reads in some representation of a \gls{datapoint} and transforms it to a set of features. Feature learning 
methods differ in the precise format of the original data representation as well as the format of the delivered features. 
This chapter mainly discusses feature learning methods that require \gls{datapoint}s being represented by $\featurelenraw$ 
numeric raw features and deliver a set of $\featurelen$ new numeric features. We will denote the raw features and the learnt 
new features by $\rawfeaturevec=\big(\rawfeature_{1},\ldots,\rawfeature_{\featurelenraw} \big)^{T} \in \mathbb{R}^{\featurelenraw}$ and $\featurevec=\big(\feature_{1},\ldots,\feature_{\featurelen}\big)^{T} \in \mathbb{R}^{\featurelen}$, respectively. 

Many ML application domains generate \gls{datapoint}s for which can access a huge number of raw features. 
Consider \gls{datapoint}s being snapshots generated by a smartphone. It seems natural to use the pixel colour intensities 
as the raw features of the snapshot. Since modern smartphone have ``Megapixel cameras'', the pixel intensities would 
provide us with millions of raw features. It might seem a good idea to use as many raw features of a \gls{datapoint} as possible 
since more features should offer more information about a \gls{datapoint} and its label $\truelabel$. There are, however, 
two pitfalls in using an unnecessarily large number of features. The first one is a computational pitfall and the second one 
is a statistical pitfall. 

Computationally, using very long feature vectors $\featurevec \in \mathbb{R}^{\featuredim}$ (with $\featuredim$ being billions), 
might result in prohibitive computational resource requirements (bandwidth, storage, time) of the resulting ML method. 
Statistically, using a large number of features makes the resulting ML methods more prone to 
overfitting. For example, \gls{linreg} will typically overfit when using feature vectors $\featurevec\!\in\!\mathbb{R}^{\featuredim}$ 
whose length $\featuredim$ exceeds the number $\samplesize$ of labeled \gls{datapoint}s used 
for training (see Chapter \ref{ch_overfitting_regularization}). 

Both from a computational and a statistical perspective, it is beneficial to use only the maximum necessary 
amount of features. The challenge is to select those features which carry most of the relevant 
information required for the prediction of the label $\truelabel$. Finding the most relevant features 
out of a huge number of raw features is the goal of dimensionality reduction methods. 
Dimensionality reduction methods form an important sub-class of feature learning methods.  
These methods learn a hypothesis $h(\rawfeaturevec)$ that maps a long raw feature vector 
$\rawfeaturevec \in \mathbb{R}^{\featurelenraw}$ to a new (short) feature vector $\featurevec \in \mathbb{R}^{\featurelen}$ 
with $\featurelen \ll \featurelenraw$. 

Beside avoiding overfitting and coping with limited computational resources, dimensionality reduction 
can also be useful for data visualization. Indeed, if the resulting feature vector has length $\featurelen=2$, 
we depict \gls{datapoint}s in the two-dimensional plane in form of a \gls{scatterplot}.  

We will discuss the basic idea underlying dimensionality reduction methods in Section \ref{sec_dim_red}. 
Section \ref{sec_pca} presents one particular example of a dimensionality reduction method that computes 
relevant features by a linear transformation of the raw feature vector. Section \ref{sec_lda} discusses a method 
for dimensionality reduction that exploits the availability of labelled \gls{datapoint}s. Section \ref{equ_random_projections} 
shows how randomness can be used to obtain computationally cheap dimensionality reduction .

Most of this chapter discusses dimensionality reduction methods that determine a small number 
of relevant features from a large set of raw features. However, sometimes it might be useful to go 
the opposite direction. There are applications where it might be beneficial to construct a large (even infinite) 
number of new features from a small set of raw features. Section \ref{sec_dim_increas} will showcase 
how computing additional features can help to improve the prediction accuracy of ML methods. 

\section{Basic Principle of Dimensionality Reduction} 
\label{sec_dim_red}

The efficiency of ML methods depends crucially on the choice of features that are used to characterize \gls{datapoint}s. 
Ideally we would like to have a small number of highly relevant features to characterize \gls{datapoint}s. If we use too 
many features we risk to waste computations on exploring irrelevant features. If we use too few features 
we might not have enough information to predict the label of a \gls{datapoint}. For a given number $\featurelen$ 
of features, dimensionality reduction methods aim at learning an (in a certain sense) optimal map from the \gls{datapoint} 
to a feature vector of length $\featurelen$. 

Figure \ref{fig_dimred} illustrates the basic idea of dimensionality reduction methods. Their goal  
is to learn (or find) a ``compression'' map $h(\cdot): \mathbb{R}^{\featurelenraw} \rightarrow \mathbb{R}^{\featuredim}$ 
that transforms a (long) raw feature vector $\rawfeaturevec \in \mathbb{R}^{\featurelenraw}$ to a (short) feature vector 
$\featurevec=(\feature_{1},\ldots,\feature_{\featuredim})^{T} \defeq h(\rawfeaturevec)$ (typically $\featuredim \ll \featurelenraw$). 

The new feature vector $\featurevec = h(\rawfeaturevec)$ serves as a compressed representation (or code) 
for the original feature vector $\rawfeaturevec$. We can reconstruct the raw feature vector using a 
reconstruction map $\reconstrmap(\cdot): \mathbb{R}^{\featurelen} \rightarrow \mathbb{R}^{\featurelenraw}$. 
The reconstructed raw features $\widehat{\rawfeaturevec} \defeq \reconstrmap(\featurevec) = \reconstrmap(h(\rawfeaturevec))$ 
will typically differ from the original raw feature vector $\rawfeaturevec$. In general, we obtain a non-zero reconstruction error 
\begin{equation} 
	\label{equ_def_reconst_error}
	\underbrace{\widehat{\rawfeaturevec}}_{=\reconstrmap(h(\rawfeaturevec)))}  - \rawfeaturevec.
\end{equation} 

Dimensionality reduction methods learn a compression map $h(\cdot)$ such that the reconstruction error \eqref{equ_def_reconst_error}
is minimized. In particular, for a dataset $\dataset = \big\{ \rawfeaturevec^{(1)},\ldots,\rawfeaturevec^{(\samplesize)} \big\}$, 
we measure the quality of a pair of compression map $h$ and reconstruction map $\reconstrmap$ by the average reconstruction error 
\begin{equation} 
	\label{equ_def_emp_error_dimred}
	\emperror\big(h,\reconstrmap| \dataset \big) \defeq (1/\samplesize) \sum_{\sampleidx=1}^{\samplesize} \loss{\rawfeaturevec^{(\sampleidx)}}{\reconstrmap\big(h\big(\rawfeaturevec^{(\sampleidx)}\big)\big)}. 
\end{equation} 
Here, $\loss{\rawfeaturevec}{\reconstrmap\big(h\big(\rawfeaturevec^{(\sampleidx)}\big)}$ denotes a loss function that is used 
to measure the reconstruction error $\underbrace{\reconstrmap\big(h\big(\rawfeaturevec^{(\sampleidx)}\big)\big)}_{\widehat{\rawfeaturevec}}  - \rawfeaturevec$. 
Different choices for the loss function in \eqref{equ_def_emp_error_dimred} result in different dimensionality 
reduction methods. One widely-used choice for the loss is the squared Euclidean norm 
\begin{equation} 
	\label{equ_def_squared_eucl_norm}
	\loss{\rawfeaturevec}{g\big(h\big(\rawfeaturevec\big)\big)} \defeq \big\| \rawfeaturevec -g\big(h\big(\rawfeaturevec\big)\big)  \big\|^{2}_{2}. 
\end{equation} 

Practical dimensionality reduction methods have only finite computational resources. Any practical 
method must therefore restrict the set of possible compression and reconstruction maps to small 
subsets $\hypospace$ and $\hypospace^{*}$, respectively. These subsets are the \gls{hypospace}s 
for the compression map $h \in \hypospace$ and the reconstruction map $\reconstrmap \in \hypospace^{*}$. 
Feature learning methods differ in their choice for these \gls{hypospace}s. 

Dimensionality reduction methods learn a compression map by solving  
\begin{align} 
	\label{equ_optimaization_dimred}
	\hat{h} & = \argmin_{h \in \hypospace} \min_{\reconstrmap \in \hypospace^{*}} \emperror\big(h,\reconstrmap| \dataset \big)  \nonumber \\ 
	& \stackrel{\eqref{equ_def_emp_error_dimred}}{=} \argmin_{h \in \hypospace} \min_{\reconstrmap\in \hypospace^{*}} (1/\samplesize) \sum_{\sampleidx=1}^{\samplesize} \loss{\rawfeaturevec^{(\sampleidx)}}{\reconstrmap\big(h\big(\rawfeaturevec^{(\sampleidx)}\big)\big)}. 
\end{align} 
We can interpret \eqref{equ_optimaization_dimred} as a (typically non-linear) approximation problem. 
The optimal compression map $\hat{h}$ is such that the reconstruction$\reconstrmap(\hat{h}(\rawfeaturevec))$, 
with a suitably chosen reconstruction map $\reconstrmap$, approximates the original raw feature vector $\rawfeaturevec$ 
as good as possible. Note that we use a single compression map $h(\cdot)$ and a single 
reconstruction map $\reconstrmap(\cdot)$ for all \gls{datapoint}s in the dataset $\dataset$. 

We obtain variety of dimensionality methods by using different choices for the \gls{hypospace}s $\hypospace, \hypospace^{*}$ 
and loss function in \eqref{equ_optimaization_dimred}. Section \ref{sec_pca} discusses a method that solves \eqref{equ_optimaization_dimred} 
for $\hypospace, \hypospace^{*}$ constituted by linear maps and the loss \eqref{equ_def_squared_eucl_norm}. Deep \index{autoencoder}autoencoders 
are another family of dimensionality reduction methods that solve \eqref{equ_optimaization_dimred} with 
$\hypospace, \hypospace^{*}$ constituted by non-linear maps that are represented by deep neural 
networks \cite[Ch. 14]{Goodfellow-et-al-2016}. 

\section{Principal Component Analysis} 
\label{sec_pca}

We now consider the special case of dimensionality reduction where the compression and reconstruction map 
are required to be linear maps. Consider a \gls{datapoint} which is characterized by a (typically very long) 
raw feature vector $\rawfeaturevec\!=\!\big( \rawfeature_{1},\ldots,\rawfeature_{\featurelenraw} \big)^{T} \!\in\! \mathbb{R}^{\featurelenraw}$ 
of length $\featurelenraw$. The length $\featurelenraw$ of the raw feature vector might be easily of 
the order of millions. To obtain a small set of relevant features $\featurevec\!=\!\big(\feature_{1},\ldots,\feature_{\featuredim} \big)^{T}\!\in\! \mathbb{R}^{\featuredim}$, we apply a linear transformation to the raw feature vector,  
\begin{equation} 
	\label{equ_feat_learning_matrix}
	\featurevec = \mathbf{W} \rawfeaturevec.
\end{equation}
Here, the ``compression'' matrix $\mathbf{W} \in \mathbb{R}^{\featuredim \times \featurelenraw}$ maps 
(in a linear fashion) the (long) raw feature vector $\rawfeaturevec \in \mathbb{R}^{\featurelenraw}$ to the (shorter) 
feature vector $\featurevec \in \mathbb{R}^{\featuredim}$.

It is reasonable to choose the compression matrix $\mathbf{W} \in \mathbb{R}^{\featuredim \times \featurelenraw}$ 
in \eqref{equ_feat_learning_matrix} such that the resulting features $\featurevec \in \mathbb{R}^{\featuredim}$ 
allow to approximate the original \gls{datapoint} $\rawfeaturevec \in \mathbb{R}^{\featurelenraw}$ as 
accurate as possible. We can approximate (or recover) the \gls{datapoint} $\rawfeaturevec \in \mathbb{R}^{\featurelenraw}$ 
back from the features $\featurevec$ by applying a reconstruction operator $\mathbf{R} \in \mathbb{R}^{\featurelenraw \times \featuredim}$, which is 
chosen such that 
\begin{equation} 
	\label{equ_approx_linear_PCA}
	\rawfeaturevec \approx \mathbf{R} \featurevec \stackrel{\eqref{equ_feat_learning_matrix}}{=} \mathbf{R} \mathbf{W} \rawfeaturevec. 
\end{equation} 

The approximation error $\emperror \big( \mathbf{W},\mathbf{R} \mid \dataset\big)$ resulting 
when \eqref{equ_approx_linear_PCA} is applied to each \gls{datapoint} in a 
dataset $\dataset=  \{ \rawfeaturevec^{(\sampleidx)}\}_{\sampleidx=1}^{\samplesize}$ is then
\begin{equation} 
	\label{equ_app_error_PCA}
	\emperror \big( \mathbf{W},\mathbf{R} \mid \dataset\big) = (1/\samplesize) \sum_{\sampleidx=1}^{\samplesize} \| \rawfeaturevec^{(\sampleidx)} - \mathbf{R} \mathbf{W} \rawfeaturevec^{(\sampleidx)} \|^{2}. 
\end{equation} 

One can verify that the approximation error $\emperror \big( \mathbf{W},\mathbf{R} \mid \dataset\big)$ 
can only by minimal if the compression matrix $\mathbf{W}$ is of the form
\begin{equation}
	\label{equ_def_PCA_W}
	\mathbf{W} = \mathbf{W}_{\rm PCA} \defeq \big( \eigvecCov^{(1)},\ldots, \eigvecCov^{(\featuredim)} \big)^{T} \in \mathbb{R}^{\featuredim \times \featurelenraw}, 
\end{equation} 
with $\featuredim$ orthonormal vectors $\eigvecCov^{(\featureidx)}$, for $\featureidx=1,\ldots,\featuredim$. 
The vectors $\eigvecCov^{(\featureidx)}$ are the \gls{eigenvector}s corresponding to the $\featuredim$ largest 
\gls{eigenvalue}s of the \index{sample covariance matrix} sample covariance matrix
\begin{equation} 
	\label{equ_def_Q_PCA}
	\mQ  \defeq (1/\samplesize) \mathbf{Z}^{T} \mathbf{Z} \in \mathbb{R}^{\featurelenraw \times \featurelenraw}. 
\end{equation}
Here we used the data matrix $\mathbf{Z}\!=\!\big( \rawfeaturevec^{(1)}, \ldots, \rawfeaturevec^{(\samplesize)} \big)^{T}\!\in\!\mathbb{R}^{\samplesize \times \featurelenraw}$.\footnote{Some authors define the data matrix as $\mZ\!=\!\big( \widetilde{\rawfeaturevec}^{(1)}, \ldots, \widetilde{\rawfeaturevec}^{(\samplesize)} \big)^{T}\!\in\!\mathbb{R}^{\samplesize \times D}$  using ``centered'' \gls{datapoint}s $\rawfeaturevec^{(\sampleidx)} - \widehat{\vm}$ 
	obtained by subtracting the average $\widehat{\vm} = (1/\samplesize) \sum_{\sampleidx=1}^{\samplesize} \rawfeaturevec^{(\sampleidx)}$.}
It can be verified easily, using the definition \eqref{equ_def_Q_PCA}, that the matrix $\mathbf{Q}$ is \gls{psd}. 
As a \gls{psd} matrix, $\mQ$ has an \gls{evd} of the form \cite{Strang2007}
\vspace*{-2mm}
\begin{equation}
	\label{equ_EVD_Q_PCA}
	\mQ=\big(\eigvecCov^{(1)},\ldots,\eigvecCov^{(\featurelenraw)}\big)  \begin{pmatrix}\eigval{1} &  \ldots & 0 \\ 0 & \ddots & 0  \\ 0 & \ldots & \eigval{\featurelenraw} 
	\end{pmatrix} \big(\eigvecCov^{(1)},\ldots,\eigvecCov^{(\featurelenraw)}\big)^{T} 
\end{equation} 
with real-valued \gls{eigenvalue}s $\eigval{1} \geq \eigval{2}\geq \ldots \geq \eigval{\featurelenraw} \geq 0$ 
and orthonormal \gls{eigenvector}s $\{ \eigvecCov^{(\featureidx)} \}_{\featureidx=1}^{\featurelenraw}$. 

The feature vectors $\featurevec^{(\sampleidx)}$ are obtained by applying the compression matrix $\mathbf{W}_{\rm PCA}$ 
\eqref{equ_def_PCA_W} to the raw feature vectors $\rawfeaturevec^{(\sampleidx)}$. We refer to the entries of the 
learnt feature vector $\featurevec^{(\sampleidx)}= \mW_{\rm PCA} \rawfeaturevec^{(\sampleidx)}$ (see \eqref{equ_def_PCA_W}) 
as the principal components (PC) of the raw feature vectors $\rawfeaturevec^{(\sampleidx)}$. Algorithm \ref{alg_PCA} 
summarizes the overall procedure of determining the compression matrix \eqref{equ_def_PCA_W} and computing 
the learnt feature vectors $\featurevec^{(\sampleidx)}$. This procedure is known as \gls{pca}. 
\begin{algorithm}[htbp]
	\caption{\gls{pca}}\label{alg_PCA}
	\begin{algorithmic}[1]
		\renewcommand{\algorithmicrequire}{\textbf{Input:}}
		\renewcommand{\algorithmicensure}{\textbf{Output:}}
		\Require  dataset  $\dataset=\{ \rawfeaturevec^{(\sampleidx)} \in \mathbb{R}^{\featurelenraw} \}_{\sampleidx=1}^{\samplesize}$; number $\featuredim$ of PCs. 
		\State compute the \gls{evd} \eqref{equ_EVD_Q_PCA} to obtain orthonormal \gls{eigenvector}s $\big(\eigvecCov^{(1)},\ldots,\eigvecCov^{(\featurelenraw)}\big)$ 
		corresponding to (decreasingly ordered) eigenvalues $\eigval{1} \geq \eigval{2} \geq \ldots \geq \eigval{\featurelenraw} \geq 0$
		\vspace*{2mm}
		\State construct compression matrix $\mW_{\rm PCA}  \defeq \big( \eigvecCov^{(1)},\ldots, \eigvecCov^{(\featuredim)} \big)^{T} \in \mathbb{R}^{\featuredim \times \featurelenraw}$ 
		\vspace*{2mm}
		\State compute feature vector $\featurevec^{(\sampleidx)} = \mW_{\rm PCA} \rawfeaturevec^{(\sampleidx)}$ whose entries are PC of $\rawfeaturevec^{(\sampleidx)}$ 
		\vspace*{2mm}
		\State compute approximation error $\emperror^{(\rm PCA)} = \sum_{\featureidx = \featuredim+1}^{\featurelenraw} \eigval{\featureidx}$ (see \eqref{equ_approx_error_PCA}). 
		\vspace*{2mm}
		\Ensure $\featurevec^{(\sampleidx)}$, for $\sampleidx=1,\ldots,\samplesize$, and the approximation error $\emperror^{(\rm PCA)}$. 
	\end{algorithmic}
\end{algorithm}
The length $\featuredim \in \{0,\ldots,\featurelenraw)$ of the delivered feature vectors $\featurevec^{(\sampleidx)}$, 
for $\sampleidx=1,\ldots,\samplesize$, is an input (or hyper-) parameter of Algorithm \ref{alg_PCA}. Two extreme cases 
are $\featuredim=0$ (maximum compression) and $\featuredim=\featurelenraw$ (no compression). 
We finally note that the choice for the orthonormal \gls{eigenvector}s in \eqref{equ_def_PCA_W} 
might not be unique. Depending on the sample covariance matrix $\mQ$, there might different sets 
of orthonormal vectors that correspond to the same eigenvalue of $\mQ$. Thus, for a given length $\featuredim$ 
of the new feature vectors, there might be several different matrices $\mW$ that achieve the same 
(optimal) reconstruction error $\emperror^{(\rm PCA)}$.  

Computationally, Algorithm \ref{alg_PCA} essentially amounts to an \gls{evd} of the sample covariance matrix $\mQ$ 
\eqref{equ_def_Q_PCA}. The \gls{evd} of $\mQ$ provides not only the optimal compression matrix 
$\mW_{\rm PCA}$ but also the measure $\emperror^{(\rm PCA)}$ for the information loss incurred by 
replacing the original \gls{datapoint}s $\rawfeaturevec^{(\sampleidx)} \in \mathbb{R}^{\featurelenraw}$ 
with the shorter feature vector $\featurevec^{(\sampleidx)} \in \mathbb{R}^{\featuredim}$. 
We quantify this information loss by the approximation error obtained when using the compression 
matrix $\mathbf{W}_{\rm PCA}$ (and corresponding reconstruction matrix 
$\mathbf{R}_{\rm opt} = \mathbf{W}_{\rm PCA}^{T}$), 
\begin{equation} 
	\label{equ_approx_error_PCA}
	\emperror^{(\rm PCA)} \defeq \emperror \big( \mathbf{W}_{\rm PCA},\underbrace{\mathbf{R}_{\rm opt}}_{=\mathbf{W}_{\rm PCA}^{T}}\mid \dataset\big) = \sum_{\featureidx = \featuredim+1}^{\featurelenraw} \eigval{\featureidx}. 
\end{equation} 
As depicted in Figure \ref{fig_ellbow_PCA}, the approximation error  $\emperror^{(\rm PCA)}$ decreases with increasing 
number $\featuredim$ of PCs used for the new features \eqref{equ_feat_learning_matrix}. For the extreme case $\featuredim\!=\!0$, 
where we completely ignore the raw feature vectors $\rawfeaturevec^{(\sampleidx)}$, the optimal reconstruction error is 
$\emperror^{(\rm PCA)} = (1/\samplesize) \sum_{\sampleidx=1}^{\samplesize} \big\| \rawfeaturevec^{(\sampleidx)} \big\|^{2}$. 
The other extreme case $\featuredim\!=\!\featurelenraw$ allows to use the raw features directly as the new features $\featurevec^{(\sampleidx)}\!=\!\rawfeaturevec^{(\sampleidx)}$. This extreme case means no compression at all, and 
trivially results in a zero reconstruction error $\emperror^{(\rm PCA)}\!=\!0$. 

\begin{figure}[htbp]
	\begin{center}
		\begin{tikzpicture}
			\begin{axis}
				[ylabel=$\emperror^{(\rm PCA)}$,
				xlabel=$\featuredim$,xticklabels={0,,,,,,$\featurelenraw$}]
				\addplot[domain=0:2, ultra thick] {10-x^3-1-1/5};
				\addplot[domain=2:10, ultra thick] {1-x/10};
			\end{axis}
		\end{tikzpicture}
	\end{center}
	\caption{Reconstruction error $\emperror^{(\rm PCA)}$ (see \eqref{equ_approx_error_PCA}) 
		of \gls{pca} for varying number $\featuredim$ of PCs.}
	\label{fig_ellbow_PCA}
\end{figure}

\subsection{Combining \gls{pca} with \gls{linreg}} 
\label{sec_comb_PcA_Linreg}

One important use case of \gls{pca} is as a pre-processing step within an overall ML problem 
such as \gls{linreg} (see Section \ref{sec_lin_reg}). As discussed in Chapter 
\ref{ch_overfitting_regularization}, \gls{linreg} methods are prone to overfitting 
whenever the \gls{datapoint}s are characterized by raw feature vectors $\rawfeaturevec$ 
whose length $\featurelenraw$ exceeds the number $\samplesize$ of labeled \gls{datapoint}s 
used in \gls{erm}. 

One simple but powerful strategy to avoid overfitting is to preprocess the raw 
features $\rawfeaturevec^{(\sampleidx)}\in \mathbb{R}^{\featurelenraw}$, for $\sampleidx=1,\ldots,\samplesize$ 
by applying \gls{pca}. Indeed,  \gls{pca} Algorithm \ref{alg_PCA} delivers feature vectors 
$\featurevec^{(\sampleidx)} \in \mathbb{R}^{\featuredim}$ of prescribed length $\featuredim$. Thus, 
choosing the parameter $\featuredim$ such that $\featuredim < \samplesize$ will typically prevent 
 the follow-up \gls{linreg} method from overfitting. 

\subsection{How To Choose Number of PC?} 
There are several aspects which can guide the choice for the number $\featuredim$ 
of PCs to be used as features. 
\begin{itemize}
	\item To generate data visualizations we might use either $\featuredim=2$ or $\featuredim=3$. 
	\item We should choose $\featuredim$ sufficiently small such that the overall ML method fits the available computational resources.  
	\item Consider using \gls{pca} as a pre-processing for \gls{linreg} (see Section \ref{sec_lin_reg}). 
	In particular, we can use the learnt feature vectors $\featurevec^{(\sampleidx)}$ delivered by \gls{pca} as the 
	feature vectors of \gls{datapoint}s in plain \gls{linreg} methods. 
	To avoid overfitting, we should choose $\featuredim < \samplesize$ (see Chapter \ref{ch_overfitting_regularization}). 
	\item Choose $\featuredim$ large enough such that the resulting approximation error 
	$\emperror^{(\rm PCA)}$ is reasonably small (see Figure \ref{fig_ellbow_PCA}). 
\end{itemize}

\subsection{Data Visualisation}

If we use \gls{pca} with $\featuredim=2$, we obtain feature vectors $\featurevec^{(\sampleidx)} = \mW_{\rm PCA} \rawfeaturevec^{(\sampleidx)}$ (see \eqref{equ_feat_learning_matrix}) which can be depicted as points in a \gls{scatterplot} (see Section \ref{equ_subsection_scatterplot}). 
As an example, consider \gls{datapoint}s $\datapoint^{(\sampleidx)}$ obtained from historic recordings of Bitcoin statistics. 
Each \gls{datapoint} $\datapoint^{(\sampleidx)} \in \mathbb{R}^{\featurelenraw}$ is a vector of length $\featurelenraw=6$. 
It is difficult to visualise points in an \gls{euclidspace} $\mathbb{R}^{\featurelenraw}$ of dimension $\featurelenraw > 2$. 
Therefore, we apply \gls{pca} with $\featuredim=2$ which results in feature vectors $\featurevec^{(\sampleidx)} \in \mathbb{R}^{2}$. 
These new feature vectors (of length $2$) can be depicted conveniently as the \gls{scatterplot} in Figure \ref{fig_scatterplot_visualization}. 

\begin{figure}[htbp]
	\begin{center}
		\begin{tikzpicture}
			\begin{axis}[
				axis x line=middle,
				axis y line=middle,
				xmax=5000,
				enlarge y limits=true,
				enlarge x limits=true,
				width=10cm, height=8cm,    
				grid = major,
				grid style={dashed, gray!30},
				ylabel=first PC $x_{1}$,
				xlabel=second PC $x_{2}$,
				]        
				\addplot[only marks,mark=+] table [x=pc1, y=pc2, col sep = comma] {twopc.csv};
			\end{axis}
		\end{tikzpicture}
	\end{center}
	\caption{A \gls{scatterplot} of \gls{datapoint}s with feature vectors $\featurevec^{(\sampleidx)} = \big(\feature_{1}^{(\sampleidx)},\feature_{2}^{(\sampleidx)}\big)^{T}$ 
		whose entries are the first two PCs of the Bitcoin statistics $\rawfeaturevec^{(\sampleidx)}$ of the $\sampleidx$-th day.} 
	\label{fig_scatterplot_visualization}
\end{figure}

\subsection{Extensions of \gls{pca}}
We now briefly discuss variants and extensions of the basic \gls{pca} method.  
\begin{itemize}
	\item {\bf Kernel \gls{pca} \cite[Ch.14.5.4]{hastie01statisticallearning}:} The \gls{pca} method is 
	most effective if the raw feature vectors of \gls{datapoint}s are concentrated around a $\featurelen$-dimensional 
	linear subspace of $\mathbb{R}^{\featurelenraw}$. Kernel \gls{pca} extends \gls{pca} to \gls{datapoint}s that are 
	located near a low-dimensional manifold which might be highly non-linear. This is achieved by applying \gls{pca}  
	to transformed feature vectors instead of the original raw feature vectors. Kernel \gls{pca} first applies a (typically non-linear) feature 
	map $\featuremap$ to the raw feature vectors $\rawfeaturevec^{(\sampleidx)}$ (see Section \ref{sec_kernel_methods}) 
	and applies \gls{pca} to the transformed feature vectors $\featuremap \big( \rawfeaturevec^{(\sampleidx)} \big)$, for $\sampleidx=1,\ldots,\samplesize$. 
	
	\vspace*{2mm}
	\item {\bf Robust \gls{pca} \cite{RobustPCA}:} The basic \gls{pca} Algorithm \ref{alg_PCA} is sensitive 
	to \gls{outlier}s, i.e., a small number of \gls{datapoint}s with significantly different statistical properties 
	than the bulk of \gls{datapoint}s. This sensitivity might be attributed to the properties of the squared 
	Euclidean norm \eqref{equ_def_squared_eucl_norm} which is used in \gls{pca} 
	to measure the reconstruction error \eqref{equ_def_reconst_error}. We have seen in Chapter \ref{ch_some_examples} 
	that \gls{linreg} (see Section \ref{sec_lin_reg} and \ref{sec_lad}) can be made robust 
	against outliers by replacing the squared error loss with another \gls{lossfunc}. In a similar spirit, robust \gls{pca} 
	replaces the squared Euclidean norm with another norm that is less sensitive to having very large reconstruction 
	errors \eqref{equ_def_reconst_error} for a small number of \gls{datapoint}s (which are outliers). 
	\vspace*{2mm}
	\item {\bf Sparse \gls{pca} \cite[Ch.14.5.5]{hastie01statisticallearning}:} The basic \gls{pca} method transforms the raw feature vector 
	$\rawfeaturevec^{(\sampleidx)}$ of a \gls{datapoint} to a new (shorter) feature vector $\featurevec^{(\sampleidx)}$. In general each 
	entry $\feature^{(\sampleidx)}_{\featureidx}$ of the new feature vector will depend on each entry of the raw feature vector $\rawfeaturevec^{(\sampleidx)}$. 
	More precisely, the new feature $\feature^{(\sampleidx)}_{\featureidx}$ depends on all raw features $\rawfeature^{(\sampleidx)}_{\featureidx'}$ for 
	which the corresponding entry $W_{\featureidx,\featureidx'}$ of the matrix $\mW= \mW_{\rm PCA}$ \eqref{equ_def_PCA_W} is non-zero. 
	For most \gls{dataset}s, all entries of the matrix $\mW_{\rm PCA}$ will typically be non-zero. 
	
	In some applications of linear dimensionality reduction we would like to construct new features that depend only on a 
	small subset of raw features. Equivalently we would like to learn a linear compression map $\mW$ \eqref{equ_feat_learning_matrix} 
	such that each row of $\mW$ contains only few non-zero entries. To this end, sparse \gls{pca} enforces the rows of the 
	compression matrix $\mW$ to contain only a small number of non-zero entries. This enforcement can be implement 
	either using additional constraints on $\mW$ or by adding a penalty term to the reconstruction error \eqref{equ_app_error_PCA}.  
	\vspace*{2mm}
	\item {\bf Probabilistic \gls{pca} \cite{Roweis98emalgorithms,probabilistic-principal-component-analysis}:} 
	We have motivated \gls{pca} as a method for learning an optimal linear compression map (matrix) \eqref{equ_feat_learning_matrix} 
	such that the compressed feature vectors allows to linearly reconstruct the original raw feature vector 
	with minimum reconstruction error \eqref{equ_app_error_PCA}. Another interpretation of \gls{pca} is that of 
	a method that learns a subspace of $\mathbb{R}^{\featurelenraw}$ that best fits the distribution of 
	the raw feature vectors $\rawfeaturevec^{(\sampleidx)}$, for $\sampleidx=1,\ldots,\samplesize$. This optimal subspace is 
	precisely the subspace spanned by the rows of $\mathbf{W}_{\rm PCA}$ \eqref{equ_def_PCA_W}. 
	
	\index{probabilistic principal component analysis}\Gls{ppca} interprets the raw feature vectors $\rawfeaturevec^{(\sampleidx)}$ 
	as realizations of \gls{iid} \gls{rv}s. These realizations are modelled as 
	\begin{equation}
		\label{equ_def_PPCA_model}
		\rawfeaturevec^{(\sampleidx)} = \mW^{T} \featurevec^{(\sampleidx)} + \bm{\varepsilon}^{(\sampleidx)} \mbox{, for } \sampleidx=1,\ldots,\samplesize. 
	\end{equation} 
	Here, $\mW \in \mathbb{R}^{\featuredim \times \featurelenraw}$ is some unknown matrix with orthonormal rows. The rows of $\mW$ span 
	the subspace around which the raw features are concentrated. The vectors $\featurevec^{(\sampleidx)}$ in \eqref{equ_def_PPCA_model} 
	are realizations of \gls{iid} \gls{rv}s whose common probability distribution is $\mathcal{N}(\mathbf{0}, \mathbf{I})$. The 
	vectors $\bm{\varepsilon}^{(\sampleidx)}$ are realizations of \gls{iid} \gls{rv}s whose common probability distribution is $\mathcal{N}(\mathbf{0}, \sigma^{2} \mathbf{I})$ with some fixed but unknown variance $\sigma^{2}$. Note that $\mW$ and $\sigma^{2}$ parametrize 
	the joint probability distribution of the feature vectors $ \rawfeaturevec^{(\sampleidx)}$ via \eqref{equ_def_PPCA_model}. 
	\Gls{ppca} amounts to maximum likelihood estimation (see Section \ref{sec_max_iikelihood}) of the parameters $\mW$ and $\sigma^{2}$. 
	This maximum likelihood estimation problem can be solved using computationally efficient estimation techniques such as \gls{em} \cite[Appendix B]{probabilistic-principal-component-analysis}. The implementation of \gls{ppca} via \gls{em} also offers a principled approach 
	to handle missing data. Roughly speaking, the \gls{em} method allows to use the probabilistic model \eqref{equ_def_PPCA_model} 
	to estimate missing raw features \cite[Sec.\ 4.1]{probabilistic-principal-component-analysis}. 
\end{itemize}

\section{Feature Learning for Non-Numeric Data}
\label{sec_discrete_embeddings} 
We have motivated dimensionality reduction methods as transformations of (very long) raw feature vectors 
to a new (shorter) feature vector $\featurevec$ such that it allows to reconstruct the raw features $\rawfeaturevec$ with minimum reconstruction 
error \eqref{equ_def_reconst_error}. To make this requirement precise we need to define a measure for the size 
of the reconstruction error and specify the class of possible reconstruction maps. \Gls{pca} uses the squared Euclidean 
norm \eqref{equ_app_error_PCA} to measure the reconstruction error and only allows for linear reconstruction 
maps \eqref{equ_approx_linear_PCA}. 

Alternatively, we can view dimensionality reduction as the generation of new feature vectors $\featurevec^{(\sampleidx)}$ 
that maintain the intrinsic geometry of the \gls{datapoint}s with their raw feature vectors $\rawfeaturevec^{(\sampleidx)}$. 
Different dimensionality reduction methods use different concepts for characterizing the ``intrinsic geometry'' of \gls{datapoint}s. 
\Gls{pca} defines the intrinsic geometry of \gls{datapoint}s using the squared Euclidean distances between feature vectors. 
Indeed, \Gls{pca} produces feature vectors $\featurevec^{(\sampleidx)}$ such that for \gls{datapoint}s whose raw feature vectors 
have small squared Euclidean distance, also the new feature vectors  $\featurevec^{(\sampleidx)}$ will have small squared 
Euclidean distance. 

Some application domains generate \gls{datapoint}s for which the Euclidean distances between raw feature vectors 
does not reflect the intrinsic geometry of \gls{datapoint}s. As a point in case, consider \gls{datapoint}s representing scientific 
articles which can be characterized by the relative frequencies of words from some given set of relevant words (dictionary). 
A small Euclidean distance between the resulting raw feature vectors typically does not imply that the corresponding 
text documents are similar. Instead, the similarity between two articles might depend on the number of authors that 
are contained in author lists of both papers. We can represent the similarities between all articles using a similarity 
graph whose nodes represent \gls{datapoint}s which are connected by an edge (link) if they are similar (see Figure \ref{fig_connect_clustering}). 

Consider a dataset $\dataset = \big(\datapoint^{(1)},\ldots,\datapoint^{(\samplesize)} \big)$ whose intrinsic geometry is 
characterized by an unweighted \gls{simgraph} $\graph = \big(\nodes \defeq \{1,\ldots,\samplesize\},\edges\big)$. 
The node $\sampleidx \in \nodes$ represents the $\sampleidx$-th \gls{datapoint} $\datapoint^{(\sampleidx)}$. Two 
nodes are connected by an undirected edge if the corresponding \gls{datapoint}s are similar. 

We would like to find short feature vectors $\featurevec^{(\sampleidx)}$, for $\sampleidx=1,\ldots,\samplesize$, such 
that two \gls{datapoint}s $\sampleidx,\sampleidx'$, whose feature vectors $\featurevec^{(\sampleidx)}, \featurevec^{(\sampleidx')}$ 
have small Euclidean distance, are well-connected to each other. This informal requirement must be made 
precise by a measure for how well two nodes of an undirected graph are connected. We refer the 
reader to literature on network theory for an overview and details of various connectivity 
measures \cite{NewmannBook}. 

Let us discuss a simple but powerful technique to map the nodes $\sampleidx \in \nodes$ of an 
undirected graph $\graph$ to (short) feature vectors $\featurevec^{(\sampleidx)} \in \mathbb{R}^{\featuredim}$. 
This map is such that the Euclidean distances between the feature vectors of two nodes reflect their connectivity 
within $\graph$. This technique uses the \index{Laplacian matrix}\gls{LapMat} $\mL \in \mathbb{R}^{(\sampleidx)}$ 
which is defined for an undirected graph $\graph$ (with node set $\nodes = \{1,\ldots,\samplesize\}$) element-wise 
\begin{equation} 
	\label{equ_def_laplacian_sim_graph}
	L_{\sampleidx,\nodeidx'} \defeq \begin{cases} - 1 & \mbox{ , if } \{\sampleidx,\sampleidx'\} \in \edges \\ 
		\nodedegree{\sampleidx}  & \mbox{ , if } \sampleidx=\sampleidx' \\
		0 & \mbox{ otherwise.} \end{cases}. 
\end{equation}
Here, $\nodedegree{\sampleidx} \defeq \big| \{ \sampleidx': \{\sampleidx,\sampleidx'\} \in \edges \} \big|$ 
denotes the number of neighbours (the degree) of node $\sampleidx \in \nodes$. It can be shown 
that the \gls{LapMat} $\mL$ is \gls{psd} \cite[Proposition 1]{Luxburg2007}. 
Therefore we can find a set of orthonormal \gls{eigenvector}s 
\begin{equation} 
	\label{equ_def_eigvec_alap_featLearn}
  \eigvecCov^{(1)},\ldots,\eigvecCov^{(\samplesize)} \in \mathbb{R}^{\samplesize}
\end{equation}
with corresponding (ordered in a non-decreasing fashion) \gls{eigenvalue}s $\eigval{1} \leq \ldots \leq \eigval{\samplesize}$ of $\mL$. 

For a given number $\featuredim$, we construct the feature vector 
$$\featurevec^{(\sampleidx)} \defeq \big( \eigvecCoventry_{\sampleidx}^{(1)},\ldots,\eigvecCoventry_{\sampleidx}^{(\featurelen)} \big)^{T}$$
for the $\sampleidx$th \gls{datapoint}. Here, we used the entries of of the first $\featuredim$ eigenvectors \eqref{equ_def_eigvec_alap_featLearn}. 
It can be shown that the Euclidean distances between the feature vectors $\featurevec^{(\sampleidx)}$, for $\sampleidx=1,\ldots,\samplesize$, 
reflect the connectivities between \gls{datapoint}s $\sampleidx=1,\ldots,\samplesize$ in the \gls{simgraph} $\graph$. 
For a more precise statement of this informal claim we refer to the excellent tutorial \cite{Luxburg2007}. 

To summarize, we can construct numeric feature vectors for (non-numeric ) \gls{datapoint}s via the \gls{eigenvector}s 
of the \gls{LapMat} of a \gls{simgraph} for the \gls{datapoint}s. Algorithm \ref{alg_dimred_discrete} summarizes this 
feature learning method which requires as its input a \gls{simgraph} for the \gls{datapoint}s and the desired number $\featurelen$ 
of numeric features. Note that Algorithm \ref{alg_dimred_discrete} does not make any use of the Euclidean distances 
between raw feature vectors and uses solely the \gls{simgraph} $\graph$ to determine the intrinsic geometry of $\dataset$. 

\begin{algorithm}[htbp]
	\caption{Feature Learning for Non-Numeric Data}\label{alg_dimred_discrete}
	\begin{algorithmic}[1]
		\renewcommand{\algorithmicrequire}{\textbf{Input:}}
		\renewcommand{\algorithmicensure}{\textbf{Output:}}
		\Require  \gls{dataset} $\dataset=\{ \rawfeaturevec^{(\sampleidx)} \in \mathbb{R}^{\featurelenraw} \}_{\sampleidx=1}^{\samplesize}$; 
		\gls{simgraph} $\graph$; 
		number $\featuredim$ of features to be constructed for each \gls{datapoint}. 
		\State construct the \gls{LapMat} $\mL$ of the \gls{simgraph} (see \eqref{equ_def_laplacian_sim_graph})
		\vspace*{2mm}
		\State compute \gls{evd} of $\mL$ to obtain $\featurelen$ orthonormal \gls{eigenvector}s \eqref{equ_def_eigvec_alap_featLearn}
		corresponding to the smallest \gls{eigenvalue}s of $\mL$ 
		\vspace*{2mm}
		\State for each \gls{datapoint} $\sampleidx$, construct feature vector 
		\begin{equation} 
			\featurevec^{(\sampleidx)} \defeq \big( \eigvecCoventry^{(1)}_{\sampleidx}, \ldots, \eigvecCoventry^{(\featurelen)}_{\sampleidx} \big)^{T} \in \mathbb{R}^{\featurelen} 
		\end{equation}
		\vspace*{2mm}
		\Ensure $\featurevec^{(\sampleidx)}$, for $\sampleidx=1,\ldots,\samplesize$
	\end{algorithmic}
\end{algorithm}

\section{Feature Learning for Labeled Data}
\label{sec_lda}

We have discussed \gls{pca} as a linear dimensionality reduction method. \Gls{pca} learns a compression matrix 
that maps raw features $\rawfeaturevec^{(\sampleidx)}$ of \gls{datapoint}s to new (much shorter) feature vectors 
$\featurevec^{(\sampleidx)}$. The feature vectors $\featurevec^{(\sampleidx)}$ determined by \gls{pca} depend solely 
on the raw feature vectors $\rawfeaturevec^{(\sampleidx)}$ of a given dataset $\dataset$. In particular, \gls{pca} determines 
the compression matrix such that the new features allow for a linear reconstruction \eqref{equ_approx_linear_PCA} 
with minimum reconstruction error \eqref{equ_app_error_PCA}.

For some application domains we might not only have access to raw feature vectors but also to the label 
values $\truelabel^{(\sampleidx)}$ of the \gls{datapoint}s in $\dataset$. Indeed, dimensionality reduction 
methods might be used as pre-processing step within a regression or classification problem that involves a labeled \gls{trainset}. 
However, in its basic form, \gls{pca} (see Algorithm \ref{alg_PCA}) does not allow to exploit the information 
provided by available labels $\truelabel^{(\sampleidx)}$ of \gls{datapoint}s $\rawfeaturevec^{(\sampleidx)}$. 
For some datasets, \gls{pca} might deliver feature vectors that are not very relevant for the overall task of 
predicting the label of a \gls{datapoint}.  

Let us now discuss a modification of \gls{pca} that exploits the information provided by available labels of the 
\gls{datapoint}s. The idea is to learn a linear construction map (matrix) $\mW$ such that the new feature vectors 
$\featurevec^{(\sampleidx)} = \mW \rawfeaturevec^{(\sampleidx)}$ allow to predict the label $\truelabel^{(\sampleidx)}$ 
as good as possible. We restrict the prediction to be linear, 
\begin{equation} 
	\label{equ_lin_predictor_dimred}
	\hat{\truelabel}^{(\sampleidx)} \defeq \vr^{T} \featurevec^{(\sampleidx)}  = \vr^{T} \mW  \rawfeaturevec^{(\sampleidx)},
\end{equation} 
with some weight vector $\vr \in \mathbb{R}^{\featuredim}$. 

While \gls{pca} is motivated by minimizing the reconstruction error \eqref{sec_dim_red}, we now aim 
at minimizing the prediction error $\hat{\truelabel}^{(\sampleidx)} - \truelabel^{(\sampleidx)}$. In particular, we assess 
the usefulness of a given pair of construction map $\mW$ and predictor $\vr$ (see \eqref{equ_lin_predictor_dimred}), 
using the \gls{emprisk} 
\begin{align}
	\label{equ_lda_prediction_error} 
	\emperror \big( \mathbf{W},\mathbf{r} \mid \dataset\big) & \defeq  (1/\samplesize) \sum_{\sampleidx=1}^{\samplesize} \big( \truelabel^{(\sampleidx)} -  \hat{\truelabel}^{(\sampleidx)} \big)^{2} \nonumber \\ 
	& \stackrel{\eqref{equ_lin_predictor_dimred}}{=}  (1/\samplesize) \sum_{\sampleidx=1}^{\samplesize} \big( \truelabel^{(\sampleidx)} - \mathbf{r}^{T} \mathbf{W} \rawfeaturevec^{(\sampleidx)} \big)^{2}. 
\end{align} 
to guide the learning of a compressing matrix $\mW$ and corresponding linear predictor \gls{weights} $\vr$ (\eqref{equ_lin_predictor_dimred}). 

The optimal matrix $\mW$ that minimizes the \gls{emprisk} \eqref{equ_lda_prediction_error} can be obtained via 
the \gls{evd} \eqref{equ_EVD_Q_PCA} of the sample covariance matrix $\mQ$ \eqref{equ_def_Q_PCA}. Note that we have 
used the \gls{evd} of $\mQ$ already for \gls{pca} in Section \ref{sec_pca} (see \eqref{equ_def_PCA_W}). Remember that \gls{pca} 
uses the $\featuredim$ eigenvectors $\eigvecCov^{(1)},\ldots,\eigvecCov^{(\featurelen)}$ corresponding to the $\featurelen$ 
largest eigenvalues of $\mQ$. In contrast, to minimize \eqref{equ_lda_prediction_error}, we need to use a different 
set of eigenvectors in the rows of $\mW$ in general. To find the right set of $\featuredim$ eigenvectors, we need 
the sample cross-correlation vector 
\begin{equation}
	\label{equ_cross_correlation_PCA}
	\vq \defeq (1/\samplesize) \sum_{\sampleidx=1}^{\samplesize} \truelabel^{(\sampleidx)} \rawfeaturevec^{(\sampleidx)}. 
\end{equation} 
The entry $q_{\featureidx}$ of the vector $\vq$ estimates the correlation between the raw 
feature $\rawfeature^{(\sampleidx)}_{\featureidx}$ and the label $\truelabel^{(\sampleidx)}$. 
We then define the index set 
\begin{equation}
	\mathcal{S} \defeq \{ \featureidx_{1},\ldots,\featureidx_{\featurelen}\} \mbox{ such that } \big(q_{\featureidx}\big)^2/\eigval{\featureidx} \geq \big(q_{\featureidx'}\big)^2/\eigval{\featureidx'} \mbox{ for any } \featureidx \in \mathcal{S}, \featureidx' \in \{1,\ldots,\featurelenraw\} \notin \mathcal{S}. 
\end{equation}
It can then be shown that the rows of the optimal compression matrix $\mW$ are the \gls{eigenvector}s 
$\eigvecCov^{(\featureidx)}$ with $\featureidx \in \mathcal{S}$. We summarize the overall feature learning method 
in Algorithm \ref{alg_PCA_labeled}. 

\begin{algorithm}[htbp]
	\caption{Linear Feature Learning for Labeled Data}\label{alg_PCA_labeled}
	\begin{algorithmic}[1]
		\renewcommand{\algorithmicrequire}{\textbf{Input:}}
		\renewcommand{\algorithmicensure}{\textbf{Output:}}
		\Require  dataset $ \big(\rawfeaturevec^{(1)},\truelabel^{(1)} \big),\ldots,\big(\rawfeaturevec^{(\samplesize)},\truelabel^{(\samplesize)} \big)$ with raw features $\rawfeaturevec^{(\sampleidx)} \in \mathbb{R}^{\featurelenraw}$ and numeric labels $\truelabel^{(\sampleidx)} \in \mathbb{R}$ ; length $\featuredim$ of new feature vectors. 
		\State compute \gls{evd} \eqref{equ_EVD_Q_PCA} of the sample covariance matrix \eqref{equ_def_Q_PCA} to obtain orthonormal eigenvectors $\big(\eigvecCov^{(1)},\ldots,\eigvecCov^{(\featurelenraw)}\big)$ 
		corresponding to (decreasingly ordered) eigenvalues $\eigval{1}\geq \eigval{2}\geq \ldots \geq \eigval{\featurelenraw} \geq 0$
		\vspace*{2mm}
		\State compute the sample cross-correlation vector \eqref{equ_cross_correlation_PCA} and, in turn, the sequence 
		\begin{equation}
			\label{equ_sequence_cross_cov_matrix}
			\big(q_{1}\big)^2/\eigval{1},  \ldots,  \big(q_{\featurelenraw}\big)^2/\eigval{\featurelenraw}
		\end{equation} 
		\vspace*{2mm}
		\State determine indices $\featureidx_{1},\ldots,\featureidx_{\featurelen}$ of $\featurelen$ largest elements in \eqref{equ_sequence_cross_cov_matrix}
		\vspace*{2mm}
		\State construct compression matrix $\mW  \defeq \big( \eigvecCov^{(\featureidx_{1})},\ldots, \eigvecCov^{(\featureidx_{\featuredim})} \big)^{T} \in \mathbb{R}^{\featuredim \times \featurelenraw}$ 
		\vspace*{2mm}
		\State compute feature vector $\featurevec^{(\sampleidx)} = \mW \rawfeaturevec^{(\sampleidx)}$  
		\vspace*{2mm}
		\Ensure $\featurevec^{(\sampleidx)}$, for $\sampleidx=1,\ldots,\samplesize$, and compression matrix $\mW$. 
	\end{algorithmic}
\end{algorithm}

The main focus of this section is on regression problems that involve \gls{datapoint}s with numeric labels 
(e.g., from the \gls{labelspace} $\labelspace = \mathbb{R}$). Given the raw features and labels of the \gls{datapoint} 
in the dataset $\dataset$, Algorithm \ref{alg_PCA_labeled} determines new feature vectors $\featurevec^{(\sampleidx)}$ 
that allow to linearly predict a numeric label with minimum squared error. A similar approach can be used for 
classification problems involving \gls{datapoint}s with a finite \gls{labelspace} $\labelspace$. Linear (or Fisher) 
discriminant analysis aims at constructing a compression matrix $\mW$ such that the learnt features 
$\featurevec = \mW \rawfeaturevec$ of a \gls{datapoint} allow to predict its label $\truelabel$ as accurately 
as possible \cite{hastie01statisticallearning}. 

\section{Privacy-Preserving Feature Learning} 
\label{equ_pp_feature_learning} 
Many important application domains of ML involve sensitive data that is subject to data protection law \cite{Wachter:2019wn}. 
Consider a health-care provider (such as a hospital) holding a large database of patient records. From a ML perspective this 
databases is nothing but a (typically large) set of \gls{datapoint}s representing individual patients. The \gls{datapoint}s 
are characterized by many features including personal identifiers (name, social security number), bio-physical parameters 
as well as examination results . We could apply ML to learn a predictor for the risk of particular disease given the features of a \gls{datapoint}. 

Given large patient databases, the ML methods might not be implemented locally at the hospital but using cloud computing. 
However, data protection requirements might prohibit the transfer of raw patient records that allow to match individuals with 
bio-physical properties. In this case we might apply feature learning methods to construct new features for each patient such 
that they allow to learn an accurate hypothesis for predicting a disease but do not allow to identify sensitive properties of the 
patient such as its name or a social security number. 

Let us formalize the above application by characterizing each \gls{datapoint} (patient in the hospital database) using raw feature 
vector $\rawfeaturevec^{(\sampleidx)} \in \mathbb{R}^{\featurelenraw}$ and a sensitive numeric property $\pi^{(\sampleidx)}$. 
We would like to find a compression map $\mW$ such that the resulting features $\featurevec^{(\sampleidx)} = \mW \rawfeaturevec^{(\sampleidx)}$ 
do not allow to accurately predict the sensitive property $\pi^{(\sampleidx)}$. The prediction of the sensitive property 
is restricted to be a linear $\hat{\pi}^{(\sampleidx)} \defeq \vr^{T} \featurevec^{(\sampleidx)}$ with some weight vector $\vr$. 

Similar to Section \ref{sec_lda} we want to find a compression matrix $\mW$ that transforms, in a linear fashion, the 
raw feature vector $\rawfeaturevec \in \mathbb{R}^{\featurelenraw}$ to a new feature vector $\featurevec \in \mathbb{R}^{\featurelen}$. 
However the design criterion for the optimal compression matrix $\mW$ was different in Section \ref{sec_lda} where 
the new feature vectors should allow for an accurate linear prediction of the label. In contrast, here we want to construct 
feature vectors such that there is no accurate linear predictor of the sensitive property $\pi^{(\sampleidx)}$. 

As in Section \ref{sec_lda}, the optimal compression matrix $\mW$ is given row-wise by the \gls{eigenvector}s of the 
sample covariance matrix \eqref{equ_def_Q_PCA}. However, the choice of which eigenvectors to use is different 
and based on the entries of the sample cross-correlation vector
\begin{equation}
	\label{equ_cross_correlation_privacypreseringPCA}
	\vc \defeq (1/\samplesize) \sum_{\sampleidx=1}^{\samplesize} \pi^{(\sampleidx)} \rawfeaturevec^{(\sampleidx)}. 
\end{equation} 
We summarize the construction of the optimal privacy-preserving compression matrix and corresponding new 
feature vectors in Algorithm \ref{alg_PCA_privacypreserving}. 
\begin{algorithm}[htbp]
	\caption{Privacy Preserving Feature Learning}\label{alg_PCA_privacypreserving}
	\begin{algorithmic}[1]
		\renewcommand{\algorithmicrequire}{\textbf{Input:}}
		\renewcommand{\algorithmicensure}{\textbf{Output:}}
		\Require  dataset $ \big(\rawfeaturevec^{(1)},\truelabel^{(1)} \big),\ldots,\big(\rawfeaturevec^{(\samplesize)},\truelabel^{(\samplesize)} \big)$; each 
		\gls{datapoint} characterized by raw features $\rawfeaturevec^{(\sampleidx)} \in \mathbb{R}^{\featurelenraw}$ and (numeric) sensitive property $\pi^{(\sampleidx)} \in \mathbb{R}$; number $\featuredim$ of new features. 
		\State compute the \gls{evd} \eqref{equ_EVD_Q_PCA} of the sample covariance matrix \eqref{equ_def_Q_PCA} to obtain orthonormal eigenvectors $\big(\eigvecCov^{(1)},\ldots,\eigvecCov^{(\featurelenraw)}\big)$ 
		corresponding to (decreasingly ordered) eigenvalues $\eigval{1} \geq \eigval{2} \geq \ldots \geq \eigval{\featurelenraw} \geq 0$
		\vspace*{2mm}
		\State compute the sample cross-correlation vector \eqref{equ_cross_correlation_privacypreseringPCA} and, in turn, the sequence 
		\begin{equation}
			\label{equ_sequence_cross_cov_matrix_privacypreserving}
			\big(c_{1}\big)^2/\eigval{1},  \ldots,  \big(c_{\featurelenraw}\big)^2/\eigval{\featurelenraw}
		\end{equation} 
		\vspace*{2mm}
		\State determine indices $\featureidx_{1},\ldots,\featureidx_{\featurelen}$ of $\featurelen$ smallest elements in \eqref{equ_sequence_cross_cov_matrix_privacypreserving}
		\vspace*{2mm}
		\State construct compression matrix $\mW  \defeq \big( \eigvecCov^{(\featureidx_{1})},\ldots, \eigvecCov^{(\featureidx_{\featuredim})} \big)^{T} \in \mathbb{R}^{\featuredim \times \featurelenraw}$ 
		\vspace*{2mm}
		\State compute feature vector $\featurevec^{(\sampleidx)} = \mW \rawfeaturevec^{(\sampleidx)}$  
		\vspace*{2mm}
		\Ensure feature vectors $\featurevec^{(\sampleidx)}$, for $\sampleidx=1,\ldots,\samplesize$, and compression matrix $\mW$. 
	\end{algorithmic}
\end{algorithm}

Algorithm \ref{alg_PCA_privacypreserving} learns a map $\mW$ to extract privacy-preserving features out of the raw feature vector 
of a \gls{datapoint}. These new features are privacy-preserving as they do not allow to accurately predict (in a linear fashion) a 
sensitive property $\pi$ of the \gls{datapoint}. Another formalization for the preservation of privacy can be obtained using  
information-theoretic concepts. This information-theoretic approach interprets \gls{datapoint}s, their feature vector and sensitive property, 
as realizations of \gls{rv}s. It is then possible to use the mutual information between new features $\featurevec$ and the sensitive (private) property $\pi$ 
as an optimization criterion for learning a compression map $h$ (Section \ref{sec_dim_red}). The resulting feature learning method (referred 
to as privacy-funnel) differs from Algorithm \ref{alg_PCA_privacypreserving} not only in the optimization criterion for the compression map 
but also in that it allows it to be non-linear \cite{MakhdoumiFunnel2014,Shkel2021}.

\section{Random Projections} 
\label{equ_random_projections}

Note that \gls{pca} uses an \gls{evd} of the sample covariance matrix $\mathbf{Q}$ \eqref{equ_def_Q_PCA}.
The computational complexity (e.g., measured by number of multiplications and additions) for computing this \gls{evd} 
is lower bounded by $\featuredim \min\{ \featurelenraw^{2}, \samplesize^{2} \}$ \cite{golub96,TippingProbPCA}. 
This computational complexity can be prohibitive for ML applications with $\featurelenraw$ 
and $\samplesize$ being of the order of millions or even billions. 

There is a computationally cheap alternative to \gls{pca} (Algorithm \ref{alg_PCA}) for finding a useful 
compression matrix $\mathbf{W}$ in \eqref{equ_feat_learning_matrix}. This alternative is  
to construct the compression matrix $\mathbf{W}$ entry-wise 
\begin{equation} 
	\label{equ_def_iid_random_matrix}
	W_{\featureidx,\featureidx'} \defeq a_{\featureidx,\featureidx'} \mbox{ with  } a_{\featureidx,\featureidx'} \sim \prob{a}. 
\end{equation} 
The matrix entries \eqref{equ_def_iid_random_matrix} are realizations $a_{i,j}$ of \gls{iid} \gls{rv}s 
with some common probability distribution $\prob{a}$. Different choices for the probability distribution $\prob{a}$ 
have been studied in the literature \cite{RauhutFoucartCS}. The Bernoulli distribution is used to obtain a compression 
matrix with binary entries. Another popular choice for $\prob{a}$ is the multivariate normal (Gaussian) distribution. 

Consider \gls{datapoint}s whose raw feature vectors $\rawfeaturevec$ are located near a $\sparsity$-dimensional subspace 
of $\mathbb{R}^{\featurelenraw}$. The feature vectors $\featurevec$ obtained via \eqref{equ_feat_learning_matrix} using 
a random matrix \eqref{equ_def_iid_random_matrix} allows to reconstruct the raw feature vectors $\rawfeaturevec$ 
with high probability whenever 
\begin{equation} 
	\featuredim \geq C \sparsity \log \featurelenraw. 
\end{equation}
The constant $C$ depends on the maximum tolerated reconstruction error $\eta$ (such that 
$\| \widehat{\rawfeaturevec} - \rawfeaturevec \|_{2}^{2} \leq \eta$ for any \gls{datapoint}) and the probability 
that the features $\featurevec$ (see \eqref{equ_def_iid_random_matrix}) allow for a maximum reconstruction 
error $\eta$ \cite[Theorem 9.27.]{RauhutFoucartCS}. 



\section{Dimensionality Increase} 
\label{sec_dim_increas} 

The focus of this chapter is on dimensionality reduction methods that learn a \gls{featuremap} delivering new feature 
vectors which are (significantly) shorter than the raw feature vectors. However, it might sometimes be beneficial to 
learn a \gls{featuremap} that delivers new feature vectors which are longer than the raw feature vectors. We have already 
discussed two examples for such feature learning methods in Sections \ref{sec_polynomial_regression} and \ref{sec_kernel_methods}. 
Polynomial regression maps a single raw feature $\rawfeature$ to a feature vector containing the powers of the raw feature $z$. 
This allows to use apply linear predictor maps to the new feature vectors to obtain predictions that depend non-linearly 
on the raw feature $\rawfeature$. Kernel methods might even use a \gls{featuremap} that delivers feature vectors belonging to 
an infinite-dimensional \gls{hilbertspace} \cite{LearningKernelsBook}. 

Mapping raw feature vectors into higher-dimensional (or even infinite-dimensional) spaces might be useful if the 
intrinsic geometry of the \gls{datapoint}s is simpler when looked at in the higher-dimensional space. Consider a 
binary classification problem where \gls{datapoint}s are highly inter-winded in the original feature space (see Figure \ref{fig_kernelmethods}). 
Loosely speaking, mapping into higher-dimensional feature space might "flatten-out" a non-linear decision boundary 
between \gls{datapoint}s. We can then apply linear classifiers to the higher-dimensional features to achieve accurate 
predictions. 


\section{Exercises} 

\begin{exercise}[Computational Burden of Many Features]
	Discuss the computational complexity of \gls{linreg}. How much computation do we need 
	to compute the linear predictor that minimizes the average squared error on a training set?
\end{exercise}

\begin{exercise}[Power Iteration] 
	The key computational step of \gls{pca} amounts to an \gls{evd} of the \gls{psd} matrix \eqref{equ_def_Q_PCA}.  
	Consider an arbitrary initial vector $\eigvecCov^{(\itercntr)}$ and the sequence obtained by iterating
	\begin{equation}
		\eigvecCov^{(\itercntr+1)} \defeq \mQ \eigvecCov^{(\itercntr)} / \big\|  \mQ \eigvecCov^{(\itercntr)} \big\|. 
	\end{equation}
	What (if any) conditions on the initialization $\eigvecCov^{(\itercntr)}$ ensure that 
	the sequence $\eigvecCov^{(\itercntr)}$ converges to the \gls{eigenvector} $\eigvecCov^{(1)}$ of 
	$\mQ$ that corresponds to its largest \gls{eigenvalue} $\eigval{1}$?
\end{exercise}

\begin{exercise}{Linear Classifiers with High-Dimensional Features} 
	Consider a training set $\dataset$ consisting of $\samplesize=10^{10}$ 
	labeled \gls{datapoint}s $\big( \rawfeaturevec^{(1)}, \truelabel^{(1)} \big), \ldots, \big( \rawfeaturevec^{(\samplesize)}, \truelabel^{(\samplesize)} \big)$ with raw feature vectors $\rawfeaturevec^{(\sampleidx)} \in \mathbb{R}^{4000}$ and 
	binary labels $\truelabel^{(\sampleidx)} \in \{-1,1\}$. Assume we have used a feature learning method to obtain the new features $\featurevec^{(\sampleidx)} \in \{0,1\}^{\featurelen}$ with $\featurelen=\samplesize$ and such that the only non-zero entry of $\featurevec^{(\sampleidx)}$ is $\feature^{(\sampleidx)}_{\sampleidx} = 1$, for $\sampleidx=1,\ldots,\samplesize$. 
	Can you find a linear classifier that perfectly classifies the training set?
\end{exercise}

\chapter{Transparent and Explainable ML}
\label{chap_explainable_ML}

The successful deployment of ML methods depends on their transparency or \gls{explainability}. 
We formalize the notion of an explanation and its effect using a simple probabilistic model 
in Section \ref{sec_model_agn_xml}. Roughly speaking, an explanation is any artefact. 
such as a list of relevant features or a reference \gls{datapoint} from a \gls{trainset}, that 
coneys information about a ML method and its predictions. Put differently, explaining a ML 
method should reduce the uncertainty (of a human end-user) about its predictions. 

\Gls{xml} is umbrella term for techniques that make ML method transparent or explainable. 
Providing explanations for the predictions of a ML method is particularly important when these  
predictions inform decision making \cite{Cheng2019}. It is increasingly becoming a legal requirement 
to provide explanations for automated decision making systems \cite{Hacker:2020aa}. 

Even for applications where predictions are not directly used to inform far-reaching decisions, 
providing explanations is important. The human end users have an intrinsic desire for explanations 
that resolve the uncertainty about the prediction. This is known as the ``need for closure'' in psychology \cite{DeBacker2006,Kagan1972}. 
Beside legal and psychological requirements, providing explanations for predictions might also 
be useful for validating and verifying ML methods. Indeed, the explanations of ML methods (and its predictions) 
can point the user (which might be a ``domain expert'') to incorrect modelling assumptions used 
by the ML method \cite{WinklerMelanoma}. 

\Gls{xml} is challenging since explanations must be tailored (personalized) to human end-users 
with varying backgrounds and in different contexts \cite{LiaoXAI2021}. The user background includes 
the formal education as well as the individual digital literacy. Some users might have received 
university-level education in ML, while other users might have no relevant formal training (such 
as an undergraduate course in linear algebra). \Gls{linreg} with few features might be perfectly 
interpretable for the first group but be considered a ``black box'' for the latter. To enable tailored 
explanations we need to model the user background as relevant for understanding the ML predictions. 

This chapter discusses \gls{xml} methods that have access to some user signal or feedback for 
some \gls{datapoint}s. Such a user signal might be obtained in various ways, including answers 
to surveys or bio-physical measurements collected via wearables or medical diagnostics. 
The user signal is used to determine (to some extent) the end-user background and, in turn, to 
tailor the delivered explanations for this end-user. 

Existing \gls{xml} methods can be roughly divided into two categories. The first category is referred 
to as \index{model agnostic}``model-agnostic'' \cite{Cheng2019}). Model-agnostic methods do not require 
knowledge of the detailed work principles of a ML method. These methods do not require 
knowledge of the \gls{hypospace} used by a ML method but learn how to explain its predictions 
by observing them on a \gls{trainset} \cite{Chen2018}. 

A second category of \gls{xml} methods, sometimes referred to as ``white-box'' methods \cite{Cheng2019}, 
uses ML methods that are considered as intrinsically explainable. The intrinsic \gls{explainability} of a ML 
method depends crucially on its choice for the \gls{hypospace} (see Section \ref{sec_hypo_space}). 
This chapter discusses one recent method from each of the two \gls{xml} categories \cite{JunXML2020,JuEERM2020}. 
The common theme of both methods is the use of \index{information theory}information-theoretic 
concepts to measure the usefulness of explanations \cite{coverthomas}. 

Section \ref{sec_model_agn_xml} discusses a recently proposed model-agnostic approach 
to \gls{xml} that constructs tailored explanations for the predictions of a given ML method \cite{JunXML2020}. 
This approach does not require any details about the internal mechanism of a ML method whose 
predictions are to be explained. Rather, this approach only requires a (sufficiently large) \gls{trainset} 
of \gls{datapoint}s for which the predictions of the ML method are known. 

To tailor the explanations to a particular user, we use the values of a user (feedback) signal provided 
for the \gls{datapoint}s in the \gls{trainset}. Roughly speaking, the explanations are chosen such that 
they maximally reduce the ``surprise'' or uncertainty that the user has about the predictions of the ML method.  

Section \ref{sec_eerm} discusses an example for a ML method that uses a \gls{hypospace} that is 
intrinsically explainable \cite{JuEERM2020}. We construct an explainable \gls{hypospace} by appropriate 
pruning of a given \gls{hypospace} such as linear maps (see Section \ref{sec_lin_reg}) 
or non-linear maps represented by either an \gls{ann} (see Section \ref{sec_deep_learning}) or \gls{decisiontree}s
(see Section \ref{sec_decision_trees}). This pruning is implemented via adding a regularization term 
to \gls{erm} \eqref{equ_def_ERM_funs}, resulting in an instance of \gls{srm} \eqref{equ_ERM_fun_regularized} 
which we refer to as \gls{eerm}. The regularization term favours hypotheses that are explainable to a 
user. Similar to the method in Section \ref{sec_model_agn_xml}, the \gls{explainability} of a map is quantified by 
information theoretic quantities. For example, if the original \gls{hypospace} is the set of linear maps 
using a large number of features, the regularization term might favour maps that depend only on few 
features that are interpretable. Hence, we can interpret \gls{eerm} as a feature learning method 
that aims at learning relevant and interpretable features (see Chapter \ref{ch_FeatureLearning}). 

\section{Personalized Explanations for ML Methods}
\label{sec_model_agn_xml}

Consider a ML application involving \gls{datapoint}s with features $\featurevec = \big(\feature_{1},\ldots,\feature_{\featurelen}\big)^{T} \in \mathbb{R}^{\featurelen}$ 
and label $\truelabel \in \mathbb{R}$. We use a ML method that reads in some labelled \gls{datapoint}s 
\begin{equation}
	\big(\featurevec^{(1)},\truelabel^{(1)}\big),\big(\featurevec^{(2)},\truelabel^{(2)}\big),\ldots,\big(\featurevec^{(\samplesize)},\truelabel^{(\samplesize)}\big), 
\end{equation}
and learns a hypothesis 
\begin{equation} 
\label{equ_pred_map}
h(\cdot): \mathbb{R}^{\featurelen} \rightarrow \mathbb{R}: \featurevec \mapsto \hat{\truelabel}=h(\featurevec). 
\end{equation}  
The precise working principle of this ML method for how to learn this hypothesis $h$ is not relevant in what follows. 
\begin{figure}[htbp]
	\hspace*{5mm}
	\begin{center}
		\begin{tikzpicture}[node distance=1cm]
			\coordinate (OR) at (0.00, 1.50);
			\node[punkt] (data) {user $\user$ consumig prediction $\hat{\truelabel}$};
			\node[above=of data] (dummy) {};
			\node[punkt,above=1cm of dummy] (hypothesis) {ML method};
			\node[right=1.4cm of dummy] (t) {prediction $\hat{\truelabel}$} ; 
			\node[left=1.4cm of dummy] (g) {explanation $\explanation$} ; 
			\draw [->,line width=0.5mm] (hypothesis.east) to [out=0,in=90] (t.north);
			\draw [->,line width=0.5mm] (t.south) to [out=270,in=0] (data.east);
			\draw [<-,line width=0.5mm] (data.west) to [out=180,in=270] (g.south);
			\draw [<-,line width=0.5mm] (g.north) to [out=90,in=180] (hypothesis.west);
		\end{tikzpicture}
		\caption{An explanation $\explanation$ provides additional information $I (\hat{\truelabel},\explanation|\user)$ to a user $\user$ 
			about the prediction $\hat{\truelabel}$.}
		\label{fig_explainable_ML}
	\end{center}
\end{figure}

The learnt predictor $h(\featurevec)$ is applied to the features of a \gls{datapoint} to obtain the 
predicted label  $\hat{\truelabel}\!\defeq\!h(\featurevec)$. The prediction $\hat{\truelabel}$ is then 
delivered to a human end-user (see Figure \ref{fig_explainable_ML}). Depending on the ML application, 
this end-user might be a streaming service subscriber \cite{GomezUribe2016}, a \index{dermatologist}dermatologist \cite{Esteva2017} or 
a \index{city planner}city planner \cite{Yang2019}. 

Human users of ML methods often have some conception or model for the relation between 
features $\featurevec$ and label $\truelabel$ of a \gls{datapoint}. This intrinsic model might vary significantly 
between users with different (social or educational) background. We will model the user 
understanding of a \gls{datapoint} by a ``user summary'' $\user \in \mathbb{R}$. The summary 
is obtained by a (possibly stochastic) map from the features $\featurevec$ of a \gls{datapoint}. 
For ease of exposition, we focus on summaries obtained by a deterministic map
\begin{equation}
	\label{eq_def_user_summary}
	\user(\cdot): \mathbb{R}^{\featurelen} \rightarrow \mathbb{R}: \featurevec \mapsto \user \defeq \user(\featurevec).
\end{equation} 
However, the resulting \gls{xml} method can be extended to user feedback $\user$ modelled as a stochastic maps.  
In this case, the user feedback $\user$ is characterized by a probability distribution $p(\user| \featurevec)$. 

The user feedback $\user$ is determined by the features $\featurevec$ of a \gls{datapoint}.  
We might think of the value $\user$ for a specific \gls{datapoint} as a signal that reflects how the 
human end-user interprets (or perceives) the \gls{datapoint}, given her knowledge (including formal education) 
and the context of the ML application. We do not assume any knowledge about the details for how 
the signal value $\user$ is formed for a specific \gls{datapoint}. In particular, we do not know any 
properties of the map $\user(\cdot): \featurevec \mapsto \user$. 

The above approach is quite flexible as it allows for very different forms of user summaries. 
The user summary could be the prediction obtained from a simplified model, such as \gls{linreg} 
using few features that the user anticipates as being relevant. Another example for a user summary 
$\user$ could be a higher-level feature, such as eye spacing in facial pictures, that the user 
considers relevant \cite{Jeong2015}. 

Note that, since we allow for an arbitrary map in \eqref{eq_def_user_summary}, 
the user summary $\user(\featurevec)$ obtained for a random \gls{datapoint} with features $\featurevec$ 
might be correlated with the prediction $\hat{\truelabel}=h(\featurevec)$. As an extreme case, consider 
a very knowledgable user that is able to predict the label of any \gls{datapoint} from its  
features as well as the ML method itself. In this case, the maps \eqref{equ_pred_map} and 
\eqref{eq_def_user_summary} might be nearly identical.  However, in general the predictions 
delivered by the learnt hypothesis \eqref{equ_pred_map} will be different from the user 
summary $\user(\featurevec)$. 

We formalize the act of explaining a prediction $\hat{\truelabel} = h(\featurevec)$ as presenting 
some additional quantity $\explanation$ to the user (see Figure \ref{fig_explainable_ML}). This 
explanation $\explanation$ can be any artefact that helps the user to understand the 
prediction $\hat{\truelabel}$, given her understanding $\user$ of the \gls{datapoint}. Loosely speaking, 
the aim of providing explanation $\explanation$ is to reduce the uncertainty of the user $\user$ 
about the prediction $\hat{\truelabel}$ \cite{Kagan1972}. 

For the sake of exposition, we construct explanations $\explanation$ that are obtained via a deterministic map 
\begin{equation}
	\label{equ_def_explanation}
	\explanation(\cdot): \mathbb{R}^{\featurelen} \rightarrow \mathbb{R}: \featurevec \mapsto \explanation \defeq \explanation(\featurevec), 
\end{equation} 
from the features $\featurevec$ of a \gls{datapoint}. However, the \gls{xml} methods in this chapter 
can be generalized without difficulty to handle explanations obtained from a stochastic map. 
In the end, we only require the specification of the conditional probability distribution $p(\explanation|\featurevec)$. 

The explanation $\explanation$ \eqref{equ_def_explanation} depends only on the features $\featurevec$  
but not explicitly on the prediction $\hat{\truelabel}$. However, our method for constructing the map \eqref{equ_def_explanation} 
takes into account the properties of the predictor map $h(\featurevec)$ \eqref{equ_pred_map}. 
In particular, Algorithm \ref{alg:xml} below requires as input the predicted labels $\hat{\truelabel}^{(\sampleidx)}$ 
for a set of \gls{datapoint}s (that serve as a \gls{trainset} for our method). 

To obtain comprehensible explanations that can be computed efficiently, 
we must typically restrict the space of possible explanations to a small subset $\mathcal{F}$ of maps 
\eqref{equ_def_explanation}. This is conceptually similar to the restriction of the space of possible 
predictor functions in a ML method to a small subset of maps which is known as the \gls{hypospace}.

\subsection{Probabilistic Data Model for XML}
\label{sec_prob_model_XML}

In what follows, we model \gls{datapoint}s as realizations of \gls{iid} \gls{rv}s with common (joint) probability 
distribution $p(\featurevec,\truelabel)$ of features and label (see Section \ref{equ_prob_models_data}). Modelling the 
\gls{datapoint}s as realizations of \gls{rv}s implies that the user summary $\user$, prediction $\hat{\truelabel}$ and 
explanation $\explanation$ are also realizations of \gls{rv}s. The joint distribution $p(\user,\hat{\truelabel},\explanation,\featurevec,\truelabel)$ conforms with the 
Bayesian network \cite{Pearl1988} depicted in Figure \ref{fig_simple_prob_ML}. Indeed,  
\begin{equation} 
	\label{equ_joint_prob_factor}
	p(\user,\hat{\truelabel},\explanation,\featurevec,\truelabel) = p(\user|\featurevec) \cdot p(\explanation|\featurevec)  \cdot p(\hat{\truelabel}|\featurevec) \cdot p(\featurevec,\truelabel). 
\end{equation}

We measure the amount of additional information provided by an explanation $\explanation$ for 
a prediction $\predictedlabel$ to some user $\user$ via the conditional mutual information (MI) \cite[Ch. 2 and 8]{coverthomas}
\begin{equation} 
	\label{eq_def_surprise}
	I(\explanation;\hat{\truelabel}|\user) \defeq  \expect \bigg\{ \log \frac{p( \hat{\truelabel},\explanation|\user)}{p(\hat{\truelabel}|\user)p(\explanation|\user)} \bigg\}.  
\end{equation} 
The conditional MI $I(\explanation;\predictedlabel|\user)$ can also be interpreted as a measure 
for the amount by which the explanation $\explanation$ reduces the uncertainty about the prediction $\hat{\truelabel}$ 
which is delivered to some user $\user$. Providing the explanation $\explanation$ serves the apparent 
human need to understand observed phenomena, such as the predictions from a ML method \cite{Kagan1972}. 

\begin{figure}[htbp]
	\hspace*{0mm}
	\begin{center}
		\begin{tikzpicture}[
			node distance=1cm and 0cm,
			mynode/.style={draw,ellipse,text width=1.8cm,align=center}, 
			mynode1/.style={draw,ellipse,text width=2.5cm,align=center}
			]
			\node[mynode1] (dp) {\gls{datapoint} $(\featurevec,\truelabel)$};
			\draw [color=gray,thick,dashed](-5,-4) rectangle (5,-1.0);
			\node at (-3.9,-3.5) [above=1mm, right=0mm] {some user};
			\node[mynode,below right=1cm and 1cm of dp] (exp) {explanation $\explanation$};
			\node[mynode,below left=1cm and 1cm of dp] (user) {user signal $\user$};
			\node[mynode,below=of dp] (pred) {prediction $\hat{\truelabel}=h(\featurevec)$};
			\path (dp) edge[-latex] (exp)
			(dp) edge[-latex] (user) ; 
			
			\draw (dp) edge[-latex] (pred);
		\end{tikzpicture}
	\end{center}
	\vspace*{0mm}
	\caption{A simple probabilistic graphical model (a \index{Bayesian network}Bayesian network \cite{LauritzenGM,koller2009probabilistic}) 
		for \gls{xml}. We interpret \gls{datapoint}s (with features $\featurevec$ 
		and label $\truelabel$) along with the user summary $\user$, $\explanation$ and predicted label $\hat{\truelabel}$ as
		 realizations of \gls{rv}s. These \gls{rv}s satisfy conditional independence relations encoded by the directed links 
		 of the graph \cite{koller2009probabilistic}. Given the \gls{datapoint}, the predicted label $\hat{\truelabel}$, 
		 the explanation $\explanation$ and the user summary $\user$ are conditionally independent. This conditional 
		 independence is trivial if all these quantities are obtained from deterministic maps applied to the features $\featurevec$ of the \gls{datapoint}.}
	\label{fig_simple_prob_ML}
\end{figure}
\vspace*{-3mm}

\subsection{Computing Optimal Explanations}
\label{sec_optimal_explanation} 
\vspace*{-1mm}

Capturing the effect of an explanation using the probabilistic model \eqref{eq_def_surprise} 
offers a principled approach to computing an optimal explanation $\explanation$. We require the optimal 
explanation $\explanation^{*}$ to maximize the conditional MI \eqref{eq_def_surprise} 
between the explanation $\explanation$ and the prediction $\hat{\truelabel}$ conditioned on the user 
summary $\user$ of the \gls{datapoint}. 

Formally, an optimal explanation $\explanation^{*}$ solves
\begin{equation}
	\label{equ_opt_explanation}
	I(\explanation^{*};\predictedlabel|\user) =  \sup_{\explanation \in \mathcal{F}} I(\explanation;\predictedlabel|\user). 
\end{equation} 
The choice for the subset $\mathcal{F}$ of valid explanations offers a trade-off 
between comprehensibility, informativeness and computational cost incurred 
by an explanation $\explanation^{*}$ (solving \eqref{equ_opt_explanation}).

The maximization problem \eqref{equ_opt_explanation} for obtaining optimal explanations 
is similar to the approach in \cite{Chen2018}. However, while \cite{Chen2018} uses the unconditional 
MI between explanation and prediction, \eqref{equ_opt_explanation} uses the conditional 
MI given the user summary $\user$. Therefore, \eqref{equ_opt_explanation} delivers  
personalized explanations that are tailored to the user who is characterized by the summary $\user$.

It is important to note that the construction \eqref{equ_opt_explanation} allows for many different 
forms of explanations. An explanation could be a subset of features of a \gls{datapoint} (see \cite{Ribeiro2016} 
and Section \ref{sec_optimal_explanation}). More generally, explanations could be 
obtained from simple local statistics (averages) of features that are considered related, 
such as nearby pixels in an image or consecutive amplitude values of an audio signal. Instead 
of individual features, carefully chosen \gls{datapoint}s from a \gls{trainset} can 
also serve as an explanation \cite{Mcinerney18,Ribeiro2018}. 

Let us illustrate the concept of optimal explanations \eqref{equ_opt_explanation} 
using \gls{linreg}. We model the features $\featurevec$ as a realization of 
a multivariate normal random vector with zero mean and covariance matrix $\mC_{\feature}$, 
\begin{equation} 
	\label{equ_feature_vector_Gaussian}
	\featurevec \sim \mathcal{N}(\mathbf{0},\mC_{\feature}).
\end{equation} 
The predictor and the user summary are linear functions   
\begin{equation} 
	\label{equ_pred_summary}
	\predictedlabel \defeq \weights^{T} \featurevec \mbox{, and } \user \defeq \mathbf{v}^{T} \featurevec. 
\end{equation}

We construct explanations via subsets of individual features $\feature_{\featureidx}$ 
that are considered most relevant for a user to understand the prediction $\predictedlabel$ (see \cite[Definition 2]{Montavon2018} and \cite{Molnar2019}). 
Thus, we consider explanations of the form 
\begin{equation} 
	\label{equ_def_explanation_subset}
	e \defeq \{ \feature_{\featureidx} \}_{\featureidx \in \explainset} \mbox{ with some subset } \explainset \subseteq \{1,\ldots,\featurelen\}. 
\end{equation}

The complexity of an explanation $e$ is measured by the number $|\explainset|$ 
of features that contribute to it. We limit the complexity of explanations 
by a fixed (small) sparsity level, 
\begin{equation} 
	| \explainset | \leq \sparsity (\ll \featurelen). 
\end{equation}

Modelling the feature vector $\featurevec$ as Gaussian \eqref{equ_feature_vector_Gaussian} implies 
that the prediction $\hat{\truelabel}$ and user summary $u$ obtained from \eqref{equ_pred_summary} is jointly 
Gaussian for a given $\explainset$ \eqref{equ_def_explanation}. 
Basic properties of multivariate normal distributions \cite[Ch. 8]{coverthomas}, 
allow to develop \eqref{equ_opt_explanation} as 
\begin{align}
	\label{equ_sup_mi_Gauss}
	\max_{\substack{\explainset  \subseteq \{1,\ldots,\featurelen\} \\ |\explainset|  \leq s }} & I(\explanation;\hat{\truelabel}|\user)  \nonumber \\ 
	&=  h(\hat{\truelabel}|\user) - h(\hat{\truelabel}|\user,\explainset) \nonumber \\[3mm]
	&= (1/2) \log \mC_{\hat{\truelabel}|\user} - (1/2) \log \det \mC_{\hat{\truelabel}|\user,\trainset} \nonumber \\[3mm]
	& =  (1/2) \log \sigma^2_{\hat{\truelabel}|\user} - (1/2) \log  \sigma^{2}_{\hat{\truelabel}|\user,\trainset}. 
\end{align}
Here, $\sigma^2_{\hat{\truelabel}|\user}$ denotes the conditional variance of the prediction $\hat{\truelabel}$, 
conditioned on the user summary $\user$. Similarly, $\sigma^{2}_{\hat{\truelabel}|\user,\explainset}$ denotes the conditional 
variance of $\hat{\truelabel}$, conditioned on the user summary $\user$ and the subset $\{\feature_{\featureidx}\}_{\featureidx\in \explainset}$ of features. 
The last step in \eqref{equ_sup_mi_Gauss} follows from the fact that $\hat{\truelabel}$ is a scalar 
random variable. 

The first component of the final expression of \eqref{equ_sup_mi_Gauss} does not depend on 
the index set $\explainset$ used to construct the explanation $\explanation$ (see \eqref{equ_def_explanation_subset}). 
Therefore, the optimal choice for $\explainset$ solves
\begin{equation}
	\label{equ_sup_m_sigma}
	\sup_{|\explainset| \leq \sparsity}  - (1/2) \log  \sigma^{2}_{\hat{\truelabel}|\user,\explainset}. 
\end{equation} 
The maximization \eqref{equ_sup_m_sigma} is equivalent to 
\begin{equation} 
	\label{equ_min_variance}
	\inf_{|\explainset| \leq \sparsity}  \sigma^{2}_{\hat{\truelabel}|\user,\explainset}. 
\end{equation}

In order to solve \eqref{equ_min_variance}, we relate the conditional variance $\sigma^{2}_{\hat{\truelabel}|\user,\explainset}$ 
to a particular decomposition
\begin{equation} 
	\label{equ_def_linear_model}
	\hat{\truelabel} = \eta \user + \sum_{\featureidx \in \explainset} \beta_{\featureidx} \feature_{\featureidx} + \varepsilon. 
\end{equation}
For an optimal choice of the coefficients $\eta$ and $\beta_{\featureidx}$, 
the variance of the error term in \eqref{equ_def_linear_model} is given by $\sigma^{2}_{\hat{\truelabel}|\user,\explainset}$. 
Indeed, 
\begin{equation} 
	\label{equ_def_optimal_coef_linmodel}
	\min_{\eta,\beta_{\featureidx} \in \mathbb{R}} \expect\big\{ \big(\hat{\truelabel} - \eta \user - \sum_{\featureidx \in \explainset} \beta_{\featureidx} \feature_{\featureidx} \big)^{2}\big\} =  \sigma^{2}_{\hat{\truelabel}|\user,\explanation}. 
\end{equation} 

Inserting \eqref{equ_def_optimal_coef_linmodel} into \eqref{equ_min_variance}, an optimal 
choice $\explainset$ (of feature) for the explanation of prediction $\hat{\truelabel}$ to user $\user$ 
is obtained from  
\begin{align} 
	& \inf_{|\explainset| \leq \sparsity}  \min_{\eta,\beta_{\featureidx} \in \mathbb{R}} \expect\big\{ \big(\hat{\truelabel} - \eta \user - \sum_{\featureidx \in \explainset} \beta_{\featureidx} x_{\featureidx} \big)^{2}\big\}  \label{equ_final_opt_E} \\
	&= \min_{ \| {\bm \beta} \|_{0} \leq \sparsity }  \expect\big\{ \big(\hat{\truelabel} - \eta \user - {\bm \beta}^{T} \featurevec \big)^{2}\big\}   \label{equ_final_opt_beta}.
\end{align} 
An optimal subset $\explainset_{\rm opt}$ of features defining the explanation $\explanation$ \eqref{equ_def_explanation_subset}
is obtained from any solution ${\bm \beta}_{\rm opt}$ of \eqref{equ_final_opt_beta} via
\begin{equation}
	\label{equ_opt_expl_support}
	\explainset_{\rm opt} = \supp {\bm \beta}_{\rm opt}. 
\end{equation}

Section \ref{sec_optimal_explanation} uses the probabilistic model \eqref{equ_feature_vector_Gaussian} to 
construct optimal explanations via the (support of the) solutions ${\bm \beta}_{\rm opt}$ of the sparse \gls{linreg} 
problem \eqref{equ_final_opt_beta}. To obtain a practical algorithm for computing (approximately) optimal explanations \eqref{equ_opt_expl_support}, 
we approximate the expectation in \eqref{equ_final_opt_beta} using an average over the \gls{trainset} 
$\big(\featurevec^{(\sampleidx)},\predictedlabel^{(\sampleidx)},\user^{(\sampleidx)}\big)$, for $\sampleidx=1,\ldots,\samplesize$. 
This resulting method for computing personalized explanations is summarized in Algorithm \ref{alg:xml}.

\begin{algorithm}[htbp]
	\caption{XML Algorithm}\label{alg:xml}
	\begin{algorithmic}[1]
		\renewcommand{\algorithmicrequire}{\textbf{Input:}}
		\renewcommand{\algorithmicensure}{\textbf{Output:}}
		\Require explanation complexity $\sparsity$, \gls{trainset} $\big(\featurevec^{(\sampleidx)},\predictedlabel^{(\sampleidx)},\user^{(\sampleidx)}\big)$ for $\sampleidx=1,\ldots,\samplesize$
		\State compute $\widehat{\bm \beta}$ by solving 
		\begin{equation}
			\label{equ_P0}
			\widehat{\bm \beta} \in  	 \argmin_{\eta\!\in\!\mathbb{R},\| {\bm \beta} \|_{0} \leq \sparsity } (1/\samplesize) \sum_{\sampleidx=1}^{\samplesize}  \big(\hat{\truelabel}^{(\sampleidx)} \!-\! \eta \user^{(\sampleidx)}\!-\!{\bm \beta}^{T} \featurevec^{(\sampleidx)} \big)^{2}
		\end{equation}
		\Ensure feature set $\widehat{\explainset} \defeq {\rm supp} \widehat{\bm \beta}$
	\end{algorithmic}
\end{algorithm}
Algorithm \ref{alg:xml} is interactive in the sense that the user has to provide a feedback signal $\user^{(\sampleidx)}$ 
for the \gls{datapoint}s with features $\featurevec^{(\sampleidx)}$. Based on the user feedback 
$\user^{(\sampleidx)}$, for $\sampleidx=1,\ldots,\samplesize$, Algorithm \ref{alg:xml} learns an optimal 
subset $\explainset$ of features \eqref{equ_def_explanation_subset} that  are used for the explanation 
of predictions. 

The sparse regression problem \eqref{equ_P0} becomes intractable for large feature length $\featurelen$. 
However, if the features are weakly correlated with each other and the user summary $\user$, the solutions of 
\eqref{equ_P0} can be found by efficient convex optimization methods. One popular method to (approximately) 
solve sparse regression \eqref{equ_P0} is the \gls{lasso} (see Section \ref{sec_lasso}), 
\begin{equation} 
	\label{equ_Lasso}
	\widehat{\bm \beta} \!\in\!\argmin_{ \eta\!\in\!\mathbb{R},{\bm \beta} \in \mathbb{R}^{\featurelen}}  (1/\samplesize) \sum_{\sampleidx=1}^{\samplesize}  \big(\predictedlabel^{(\sampleidx)} \!-\! \eta \user^{(\sampleidx)}\!-\!{\bm \beta}^{T} \featurevec^{(\sampleidx)} \big)^{2}\!+\!\regparam \| {\bm \beta} \|_{1}.  
\end{equation}
There is large body of work that studies the choice of \gls{lasso} parameter $\regparam$ in \eqref{equ_Lasso} 
such that solutions \eqref{equ_Lasso} coincide with the solutions of \eqref{equ_P0} (see \cite{HastieWainwrightBook,GeerBuhlConditions} and 
references therein). The proper choice for $\regparam$ typically requires knowledge of statistical properties of data. If such a 
probabilistic model is not available, the choice of $\regparam$ can be guided by simple validation techniques 
(see Section \ref{sec_validate_predictor}).  

\section{Explainable Empirical Risk Minimization} 
\label{sec_eerm}

Section \ref{sec_reg_ERM} discussed \gls{srm} \eqref{equ_ERM_fun_pruned} as a method for pruning the 
\gls{hypospace} $\hypospace$ used in \gls{erm} \eqref{equ_def_ERM_funs}. This pruning is implemented either 
via a (hard) constraint as in \eqref{equ_ERM_fun_pruned} or by adding a regularization term to the \gls{trainerr} 
as in \eqref{equ_ERM_fun_regularized}. The idea of \gls{srm} is to avoid (prune away) 
hypothesis maps that perform good on the \gls{trainset} but poorly outside, i.e., they do not generalize well. 
Here, we will use another criterion for steering the pruning and construction of \gls{regularization} terms. 
In particular, we use the (intrinsic) \gls{explainability} of a hypotheses map as a \gls{regularization} term. 

To make the notion of \gls{explainability} precise we use again the probabilistic model of Section \ref{sec_prob_model_XML}. 
We interpret \gls{datapoint}s as realizations of \gls{iid} \gls{rv}s with common (joint) probability distribution $p(\featurevec,\truelabel)$ 
of features $\featurevec$ and label $\truelabel$. A quantitative measure the intrinsic \gls{explainability} of a 
hypothesis $h \in \hypospace$ is the conditional (differential) entropy \cite[Ch. 2 and 8]{coverthomas}
\begin{equation} 
	\label{eq_def_explainability}
	\condent(\predictedlabel|\user) \defeq -  \expect\bigg\{ \log p(\predictedlabel|\user) \bigg\}.  
\end{equation} 
The conditional entropy \eqref{eq_def_explainability} indicates the uncertainty about the 
prediction $\predictedlabel$, given the user summary $\hat{\user}=\user(\featurevec)$. Smaller 
values $\condent(\predictedlabel;u)$ correspond to smaller levels of uncertainty in the predictions 
$\predictedlabel$ that is experienced by user $\user$. 

We obtain \gls{eerm} by requiring a sufficiently small conditional entropy \eqref{eq_def_explainability} of 
a hypothesis, 
\begin{equation} 
	\label{equ_def_EERM}
	\hat{\predictor} \in \argmin_{\predictor \in \hypospace} \emperror \big( \predictor \big) \quad \mbox{s.t. } \quad \condent(\predictedlabel|\hat{u}) \leq \eta. 
\end{equation}
The random variable $\hat{\truelabel}=\predictor(\featurevec)$ in the constraint of \eqref{equ_def_EERM} is obtained 
by applying the predictor map $\predictor \in \hypospace$ to the features.  The constraint $\condent(\predictedlabel|\hat{u}) \leq \eta$ 
in \eqref{equ_def_EERM} enforces the learnt hypothesis $\hat{\predictor}$ to be sufficiently explainable in 
the sense that the conditional entropy $\condent(\hat{\predictor}|\hat{u}) \leq \eta$ does not exceed a 
prescribed level $\eta$. 

Let us now consider the special case of \gls{eerm} \eqref{equ_def_EERM} for the linear \gls{hypospace}
\begin{equation} 
	\label{equ_linear_predictor_map}
	\predictor^{(\weights)}(\featurevec) \defeq \weights^{T} \mathbf{x} \mbox{ with some parameter vector } \weights \in \mathbb{R}^{\featurelen}. 
\end{equation} 
Moreover, we assume that the features $\featurevec$ of a \gls{datapoint} and 
its user summary $u$ are jointly Gaussian with mean zero and covariance matrix $\mathbf{C}$, 
\begin{equation} 
	\label{equ_Gaussian_feature_summary}
	\big(\featurevec^{T},\hat{\user}\big)^{T} \sim \mathcal{N}(\mathbf{0},\mathbf{C}). 
\end{equation} 
Under the assumptions \eqref{equ_linear_predictor_map} and \eqref{equ_Gaussian_feature_summary} (see \cite[Ch. 8]{coverthomas}), 
\begin{align}
	\label{equ_codent_EERM}
	\condent(\hat{\user}|\predictedlabel) &= (1/2) \log  \sigma^{2}_{\predictedlabel|\hat{\user}}. 
\end{align}
Here, we used the conditional variance $\sigma^{2}_{\predictedlabel|\hat{u}}$ of $\predictedlabel$ given the 
random user summary $\user=\user(\featurevec)$. 
Inserting \eqref{equ_codent_EERM} into the generic form of \gls{eerm} \eqref{equ_def_EERM},
\begin{equation} 
	\label{equ_def_EERM_Gaussian}
	\hat{\predictor} \in \argmin_{\predictor \in \hypospace} \emperror(h) \quad \mbox{s.t. } \quad \log  \sigma^{2}_{\hat{\truelabel}|\hat{\user}} \leq \eta. 
\end{equation}

By the monotonicity of the logarithm, \eqref{equ_def_EERM_Gaussian} is equivalent to 
\begin{equation} 
	\label{equ_def_EERM_Gaussian_1}
	\hat{\predictor} \in \argmin_{\predictor \in \mathcal{H}} \emperror(\predictor) \quad \mbox{s.t. } \quad   \sigma^{2}_{\hat{y}|\hat{u}} \leq \explanation^{(\eta)}. 
\end{equation}
To further develop \eqref{equ_def_optimal_coef_linmodel}, we use the identity
\begin{equation} 
	\label{equ_def_optimal_coef_linmodel}
	\min_{\eta \in \mathbb{R}} \expect\big\{ \big(\hat{\truelabel} - \eta \user \big)^{2}\big\} =  \sigma^{2}_{\hat{\truelabel}|\hat{\user}}. 
\end{equation} 
The identity \eqref{equ_def_optimal_coef_linmodel} relates the conditional variance $\sigma^{2}_{\hat{\truelabel}|\hat{\user}}$ 
to the minimum mean squared error that can be achieved by estimating $\hat{\truelabel}$ using a 
linear estimator $\eta \hat{\user}$ with some $\eta \in \mathbb{R}$. 
Inserting \eqref{equ_def_optimal_coef_linmodel} and \eqref{equ_linear_predictor_map} into \eqref{equ_def_EERM_Gaussian_1}, 
\begin{equation} 
	\label{equ_def_EERM_Gaussian_2}
	\hat{\predictor} \in \argmin_{\weights \in \mathbb{R}^{\featurelen},\eta \in \mathbb{R}} \emperror(\predictor^{(\weights)}) \quad \mbox{s.t. } \quad     \expect\big\{ \underbrace{\big(\weights^{T}\featurevec}_{\stackrel{\eqref{equ_linear_predictor_map}}{=} \hat{\truelabel}} - \eta \hat{\user} \big)^{2}\big\} \leq \explanation^{(\eta)}. 
\end{equation}

The inequality constraint in \eqref{equ_def_EERM_Gaussian_2} is convex \cite[Ch. 4.2.]{BoydConvexBook}. 
For squared error loss, the objective function $\emperror(\predictor^{(\weights)})$ is also convex. 
Thus, for linear least squares regression, we can reformulate \eqref{equ_def_EERM_Gaussian_2} 
as an equivalent (dual) unconstrained problem \cite[Ch. 5]{BoydConvexBook}
\begin{equation} 
	\label{equ_def_ELLRM_Gaussian_3}
	\hat{\predictor} \in \argmin_{\weights \in \mathbb{R}^{\featurelen},\eta \in \mathbb{R}} \explainset(\predictor^{(\weights)})  + \regparam    \expect\big\{ \big(\weights^{T}\featurevec- \eta \hat{\user} \big)^{2}\big\}. 
\end{equation}
By convex duality, for a given threshold $\explanation^{(\eta)}$ in \eqref{equ_def_EERM_Gaussian_2}, we can find 
a value for $\regparam$ in \eqref{equ_def_ELLRM_Gaussian_3} such that \eqref{equ_def_EERM_Gaussian_2} 
and \eqref{equ_def_ELLRM_Gaussian_3} have the same solutions \cite[Ch. 5]{BoydConvexBook}. 
Algorithm \ref{alg:eerm} below is obtained from \eqref{equ_def_ELLRM_Gaussian_3} by approximating 
the expectation $ \expect\big\{ \big(\weights^{T}\featurevec- \eta \hat{\user} \big)^{2}\big\}$ with 
an average over the \gls{datapoint}s $\big(\featurevec^{(\sampleidx)},\hat{\truelabel}^{(\sampleidx)},\hat{\user}^{(\sampleidx)}\big)$ for $\sampleidx=1,\ldots,\samplesize$. 

\begin{algorithm}[htbp]
	\caption{Explainable Linear Least Squares Regression}\label{alg:eerm}
	\begin{algorithmic}[1]
		\renewcommand{\algorithmicrequire}{\textbf{Input:}}
		\renewcommand{\algorithmicensure}{\textbf{Output:}}
		\Require \gls{explainability} parameter $\regparam$, \gls{trainset} $\big(\featurevec^{(\sampleidx)},\hat{\truelabel}^{(\sampleidx)},\hat{\user}^{\sampleidx)}\big)$ for $\sampleidx=1,\ldots,\samplesize$
		\State solve 
		\begin{equation}
			\label{equ_P0_ellr}
			\hspace*{-7mm}\widehat{\weights} \!\in\! \argmin_{\eta\!\in\!\mathbb{R},\weights \in \mathbb{R}^{\featurelen} } (1/\samplesize) \sum_{\sampleidx=1}^{\samplesize}  
			\underbrace{\big(\hat{\truelabel}^{(\sampleidx)} \!-\! \weights^{T} \featurevec^{(\sampleidx)} \big)^{2}}_{\mbox{\gls{emprisk}}} + \regparam \underbrace{( \weights^{T} \featurevec^{(\sampleidx)} - \eta \hat{\user}^{(i)})^{2}}_{\mbox{\gls{explainability}}} 
		\end{equation}
		\Ensure \gls{weights} $\widehat{\weights}$ of explainable linear hypothesis 
	\end{algorithmic}
\end{algorithm}

\section{Exercises} 
\begin{exercise}[Convexity of Explainable \Gls{linreg}] 
	Rewrite the optimization problem \eqref{equ_P0_ellr} as an equivalent quadratic 
	optimization problem $\min_{\vv \in \mathbb{R}^{\featuredim}} \vv^{T} \mathbf{Q} \vv + \vv^{T} \vq$. Identify the 
	matrix $\mQ \in \mathbb{R}^{\featuredim \times \featuredim}$ and the vector $\vq \in \mathbb{R}^{\featuredim}$. 
\end{exercise}

\printglossaries
\printindex

\bibliographystyle{plain}
\bibliography{Literature}



\end{document}